%% file: main.tex
\documentclass[runningheads]{llncs}

\usepackage{eccv}

\usepackage{eccvabbrv}
\usepackage{tabularx}
\usepackage[table]{xcolor}
\usepackage{makecell}
\usepackage{booktabs}
\usepackage{microtype}
\usepackage{multirow}
\usepackage{makecell}
\usepackage[export]{adjustbox}
\usepackage{wrapfig}
\usepackage[percentage]{overpic}
\usepackage[export]{adjustbox}

\usepackage{tikz}
\usetikzlibrary{positioning, fit, backgrounds, shapes.arrows, calc}

\definecolor{safetyColor}{HTML}{0d6a82}
\definecolor{biodiversitycolor}{HTML}{fb4d3d}
\definecolor{3dmodelColor}{HTML}{345995}
\definecolor{applicationColor}{HTML}{e7901a}
\definecolor{backgroundColor}{HTML}{2a9d8f}
\definecolor{challengeColor}{HTML}{606c38}
\definecolor{applicationBackground}{HTML}{ffb703}

\usepackage{graphicx}
\usepackage{booktabs}

\usepackage[accsupp]{axessibility}  %

\definecolor{myorange}{HTML}{f57600}

\usepackage[backend=biber, style=lncs,maxcitenames=1,minnames=1,maxalphanames=4,minalphanames=3, url = false]{biblatex}
\DefineBibliographyStrings{english}{%
  andothers = {\normalfont et\addabbrvspace al\adddot}
}
\AtEveryBibitem{
  \clearfield{urlyear}
  \clearfield{urlmonth}
}

\usepackage[breaklinks,colorlinks,citecolor=eccvblue]{hyperref}
\usepackage{hyperref}

\usepackage{orcidlink}

\usepackage{xspace} %

\begin{document}

\input{defs}

\title{SEAR: \underline{S}imple and \underline{E}fficient \underline{A}daptation of Visual Geometric Transformers for Unpaired \underline{R}GB-Thermal 3D Reconstruction}

\titlerunning{SEAR: Simple and Efficient Adaptation for RGB-T 3D Reconstruction}

\author{
  Vsevolod Skorokhodov\inst{1}\orcidlink{0009-0002-9840-761X} \and
  Chenghao Xu\inst{2}\orcidlink{0009-0007-2521-4493} \and
  Shuo Sun\inst{3}\orcidlink{0000-0003-0790-6510} \and
  Olga Fink\inst{2}\orcidlink{0000-0002-9546-1488} \and
  Malcolm Mielle\inst{1}\orcidlink{0000-0002-3079-0512}
}

\authorrunning{V.~Skorokhodov et al.}

\institute{
  Schindler EPFL Lab, Lausanne, Switzerland
  \and EPFL, Lausanne, Switzerland
  \and Örebro University, Örebro, Sweden
}

\maketitle

\input{sec/1_abstract}

\input{sec/2_introduction}
\input{sec/3_related_works}
\input{sec/4_method}
\input{sec/5_experiments}
\input{sec/6_conclusion}

\section*{Acknowledgements}
\input{sec/acknowledgements}

\printbibliography

\clearpage
\appendix
\section*{Appendix}
\addcontentsline{toc}{section}{Appendix}
\input{supplementary/2_impl_details}

\input{supplementary/3_dataset}
\input{supplementary/4_metrics_colmap}

\input{supplementary/5_more_visualization}
\input{supplementary/6_per_scene_table}

\input{supplementary/7_two_view_metrics}

\input{supplementary/8_more_results_align_rgb_thermal}
\input{supplementary/11_thermal_only}
\input{supplementary/14_quality_thermal_and_rgb_independently}
\input{supplementary/15_results_full_average}

\input{supplementary/16_data_split}

\end{document}

%% file: defs.tex
\definecolor{best}{RGB}{200, 200, 255}
\definecolor{secondbest}{RGB}{240, 240, 255}

\newcommand{\ours}[0]{SEAR\xspace}
\newcommand{\datasetname}[0]{SEAR dataset\xspace}
\newcommand{\modelname}[0]{\ours}
\newcommand{\rgbt}[0]{RGB-T\xspace} %
\newcommand{\rgbthermal}[0]{RGB+thermal\xspace}
\newcommand{\orebrodataset}[0]{RF\xspace}

\newif\ifdraft
\drafttrue

\ifdraft
\newcommand{\MM}[1]{{\color{red}{\bf PF: #1}}}
\newcommand{\mm}[1]{{\color{red} #1}}
\newcommand{\SevS}[1]{{\color{orange}{\bf SS: #1}}}
\newcommand{\sevs}[1]{{\color{orange} #1}}
\newcommand{\ivan}[1]{{\color{purple}[TODO (@ivan): #1]}}
\else
\newcommand{\MM}[1]{}
\newcommand{\mm}[1]{#1}
\newcommand{\SevS}[1]{}
\newcommand{\sevs}[1]{#1}
\fi

\newcommand{\apref}[1]{Appx.~\ref*{#1}}

%% file: sec/1_abstract.tex
\begin{abstract}
  Foundational feed-forward visual geometry models enable accurate and efficient camera pose estimation and scene reconstruction by learning strong scene priors from massive RGB datasets.
  However, their effectiveness drops when applied to mixed sensing modalities, such as RGB-thermal (RGB-T) images.
  We observe that while a visual geometry grounded transformer pretrained on RGB data generalizes well to thermal-only reconstruction, it struggles to align RGB and thermal modalities when processed jointly.
  To address this, we propose SEAR, a simple yet efficient fine-tuning strategy that adapts a pretrained geometry transformer to multimodal RGB-T inputs.
  Despite being trained on a relatively small RGB-T dataset, our approach significantly outperforms state-of-the-art methods for 3D reconstruction and camera pose estimation, achieving significant improvements over all metrics and delivering higher detail and consistency between modalities with negligible overhead in inference time compared to the original pretrained model.
  Notably, SEAR enables reliable multimodal pose estimation and reconstruction even under challenging conditions, such as low lighting and dense smoke.
  We validate our architecture through extensive ablation studies and demonstrate how the model aligns both modalities.
  Additionally, we introduce a new dataset featuring RGB and thermal sequences captured at different times, viewpoints, and illumination conditions, providing a robust benchmark for future work in multimodal 3D scene reconstruction.
  Code and models are publicly available at \url{https://doi.org/10.5281/ZENODO.21077295}.

  \keywords{Multimodal Pose Estimation \and Thermal imagery \and Feed-Forward Neural Network}
\end{abstract}

%% file: sec/2_introduction.tex
\section{Introduction}
\label{sec:intro}

Recent work introduced large-scale 3D geometric foundation models as adaptable solutions for in-the-wild 3D vision tasks.
Unlike traditional photogrammetry pipelines---which depend on feature matching and iterative optimization---these models use transformers to predict camera poses and scene structure from sparse RGB inputs in a single feed-forward pass~\cite{wang2025vggt}.
Pretrained on diverse datasets, they achieve strong cross-scene generalization, addressing classical limitations like computational inefficiency and sensitivity to image quality and environmental conditions.
Despite these advances, current geometric foundation models rely solely on RGB inputs, limiting their robustness in real-world scenarios where complementary modalities are needed.
E.g., thermal cameras capture long-wave infrared radiation, enabling applications such as low-light reconstruction~\cite{xuLeveragingThermalModality2025}, thermal simulation~\cite{chassaingThermoxelsVoxelBasedMethod2025}, and infrastructure inspection~\cite{hassanThermoNeRFMultimodalNeural2025}.

When applying a pretrained feed-forward model (e.g., VGGT~\cite{wang2025vggt}) to mixed RGB-thermal inputs, we observe independent reconstructions that fail to align into a coherent 3D representation (see \cref{fig:mm_point_clouds}).
This indicates that pretrained models encode strong geometric priors, but lack explicit cross-modal consistency for pose estimation and reconstruction.
Most prior works address this issue via either synchronous data collection~\cite{hassanThermoNeRFMultimodalNeural2025,chassaingThermoxelsVoxelBasedMethod2025,luThermalGaussianThermal3D2024a, thermalnerf}, or post-hoc alignment~\cite{tuzcuouglu2024xoftr}.
In simultaneous capture, using paired images from both modalities, a ``strong'' modality (e.g., RGB) infers poses for a ``weaker'' one (e.g., thermal) using classical pipelines like COLMAP~\cite{schonbergerStructurefrommotionRevisited2016a}.
In practice, synchronous captures necessitate complex or expensive sensor setups, and real-world data is often asynchronous (e.g., thermal imagery at sunrise for steady-state facades~\cite{waseemPhysicsInformedNeuralNetworks2025} vs. RGB at daytime for visual clarity~\cite{hassanThermoNeRFMultimodalNeural2025}).
Post-hoc alignment methods, on the other hand, use cross-modal feature matching to align poses/images using specialized descriptors; those methods are thus sensitive to image, feature, and match quality, leading to lower robustness.
Thus, estimating consistent multimodal camera poses and 3D scene reconstruction in realistic scenarios is still a challenge.

In this work, we address camera pose estimation and scene reconstruction from unpaired RGB and thermal (RGB-T) images.
Our key insight is that, since VGGT performs well on each modality independently when processed jointly but lacks geometric consistency, large-scale multimodal retraining should be unnecessary.
Instead, a pretrained model can be adapted with minimal parameter updates to bridge the modality gap.
Our contributions are threefold:
\begin{itemize}
  \item
    We propose \ours{}, a lightweight fine-tuning strategy for cross-modal pose estimation and 3D reconstruction, based on LoRA-based adapters and a specific batching strategy.
    Our method requires fewer than 5\% of the original model's parameters and preserves the inference speed of the base model.
  \item We show that \ours{} only requires a modest multimodal dataset for tuning ($\sim$15,000 pairs of RGB and thermal images), making it practical for real-world applications where collecting large multimodal datasets is challenging.
  \item
    We present a new dataset comprising 9 scenes (1,890 images) with distinct RGB/thermal trajectories captured under varying illumination and viewpoints.
    This dataset provides a new benchmark for evaluating RGB-T cross-spatial multimodal reconstruction in challenging conditions.
\end{itemize}

We demonstrate large improvements against eight state-of-the-art baselines in pose estimation and point cloud reconstruction metrics.
Complementary ablation studies further support our key design choices.

%% file: sec/3_related_works.tex
\section{Related Works}
\label{sec:related_works}

\label{subsec:vggt_pipeline}

RGB-T imaging enables low-light scene mapping and non-invasive infrastructure inspection (e.g., detecting thermal defects via finite element analysis~\cite{chassaingThermoxelsVoxelBasedMethod2025}).
Though NeRF~\cite{hassanThermoNeRFMultimodalNeural2025,liSPECNERFMultiSpectralNeural2024} and Gaussian Splatting~\cite{luThermalGaussianThermal3D2024a,liuThermalGSDynamic3D2025} have been extended to RGB-T data, they depend on known camera poses---typically estimated via SfM~\cite{schonbergerStructurefrommotionRevisited2016a} or single-modality feed-forward networks, which fail for cross-modal datasets (see \cref{fig:mm_point_clouds}).
Thus, most prior works either assume paired multimodal datasets, using RGB to obtain pose estimates for other modalities~\cite{hassanThermoNeRFMultimodalNeural2025,chassaingThermoxelsVoxelBasedMethod2025,luThermalGaussianThermal3D2024a} or are limited by the complexity of performing feature matching between the two modalities~\cite{cordonnier2026unpairedrgbthermalgaussiansplattingusing}.

To estimate camera poses and scene reconstruction, traditional 3D reconstruction relies on iterative optimization (e.g., SfM~\cite{schonbergerStructurefrommotionRevisited2016a} or SLAM~\cite{sunHighFidelitySLAMUsing2024}), on the detection of salient features (e.g., ORB~\cite{mur-artalORBSLAMVersatileAccurate2015}), and on computationally expensive matching algorithms (e.g., PnP~\cite{wuPnPProblemRevisited2006} or RANSAC~\cite{fischlerRandomSampleConsensus1981}).
While these methods can struggle with robustness and generalization, recent feed-forward models use transformers to predict camera poses and 3D geometry in a single pass~\cite{wangDust3rGeometric3d2024,leroyGroundingImageMatching2025,wang2025vggt}, achieving strong zero-shot generalization.
However, these methods are trained on unimodal inputs (e.g., RGB), and their use for applications where multimodal data (e.g., thermal, depth) is necessary for robustness remains limited.

Retraining large models for multimodal tasks is computationally prohibitive for most machine learning practitioners.
Furthermore, RGB and thermal datasets for scene reconstruction are scarce (e.g., ThermoNeRF's 20 scenes~\cite{hassanThermoNeRFMultimodalNeural2025}), with larger RGB+thermal datasets focusing on segmentation/navigation.
Parameter-efficient fine-tuning (PEFT) methods address this by reducing the number of trainable parameters while preserving performance.

Key approaches include Adapter Layers~\cite{houlsbyParameterefficientTransferLearning2019} (lightweight modules inserted between transformer layers), Prompt Tuning~\cite{lesterPowerScaleParameterEfficient2021} (optimizes continuous prompts in embedding space), and Low-Rank Adaptation (LoRA)~\cite{hu2022lora} (decomposes weight updates into low-rank matrices).
While PEFT has been applied in NLP~\cite{hu2022lora} and visual segmentation~\cite{miEmpowerVisionApplications2025}, existing work focuses on task specialization---not modality expansion.
No prior work explores PEFT to adapt pretrained 3D geometric models to multimodal reconstruction.
By bridging this gap, our work lays the foundation for resource-efficient, generalizable multimodal reconstruction.

%% file: sec/4_method.tex
\section{Preliminaries: VGGT}
\label{sec:preliminary}

VGGT~\cite{wang2025vggt} is a transformer-based model able to estimate, from a sequence of $N$ RGB images observing a 3D scene, the intrinsic and extrinsic camera parameters, a depth map, a point map, and a grid of $C$-dimensional features for point tracking.
DINOv2~\cite{oquab2023dinov2} performs feature extraction under the hood, producing a sequence of tokens for each input frame.
After that, a learnable \textit{camera token} is concatenated to each set of tokens of each frame.
The resulting token sequence is processed by 24 self-attention blocks, with alternating (frame-wise and global) attention (AA) blocks.
The tokens from the last layer are passed to the camera parameters prediction head, while intermediate tokens from layers 4, 11, 17, and 23 are concatenated and passed to depth, point map, and tracking prediction heads.
For more details, refer to the original paper~\cite{wang2025vggt}.

\section{Methodology}

\begin{figure}[t]
  \centering
  \resizebox{0.9\textwidth}{!}{%
    \input{images/MethodCameraToken/flowchart.tex}
 }

  \caption{\ours architecture.
 RGB and thermal images are first tokenized using DINOv2.
 For each modality, camera-specific tokens are concatenated with the corresponding DINO tokens.
 The combined tokens are then processed by an Alternating-Attention (AA) module with LoRA adapters.
 Finally, the refined tokens are passed to separate prediction heads for camera parameter estimation and depth estimation.
 Trainable parameters are highlighted with a flame symbol.
 }
  \label{fig:thermo_vggt_architecture}
\end{figure}
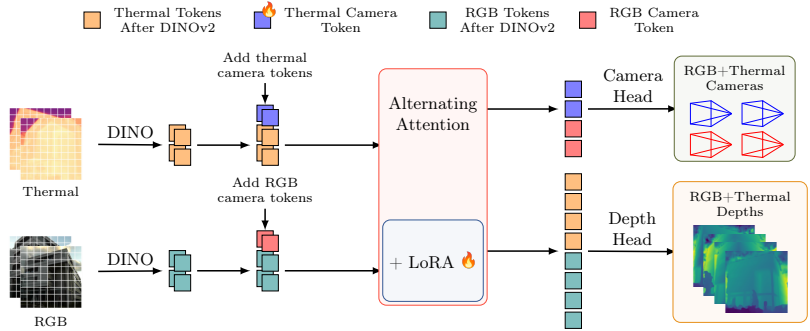

While VGGT can be used to estimate RGB or thermal 3D scenes, estimated RGB-T scenes often result in two disjoint reconstructions (one per modality), misaligned in pose and scale (see \cref{fig:mm_point_clouds}).
We hypothesize that this knowledge gap can be closed by fine-tuning the pretrained model, drastically reducing computational cost compared to retraining.

We introduce a framework (see \cref{fig:thermo_vggt_architecture}) to adapt VGGT~\cite{wang2025vggt} for joint estimation of RGB and thermal intrinsic/extrinsic camera parameters and depth maps.
We introduce two innovations:
1) Lightweight LoRA adapters in the AA module to bridge the domain gap between RGB and thermal inputs (\cref{sec:method:lora})
2) A batching strategy designed to prevent the model from relying on RGB-T image correspondences (\cref{sec:training_procedure}).
This simple, albeit effective, methodology, combined with a small amount of training data (\cref{sec:dataset}), enables RGB-T scene reconstruction and camera pose estimation from unpaired multimodal RGB and thermal images, with negligible memory and inference overhead.

\subsection{LoRA Integration}\label{sec:method:lora}

We fine-tune the RGB-pretrained VGGT model using LoRA to adapt it for mixed RGB+thermal inputs.
LoRA modules are integrated into all linear and multi-head attention layers of the alternating-attention (AA) module.
This approach preserves the model's pretrained RGB knowledge, since the original weights are frozen; the original RGB and thermal data processing is left unchanged, pushing the model to focus on alignment between the modalities to reduce the loss during training.
The features produced by the AA module are then processed by the prediction heads to estimate camera parameters and depth maps with their uncertainties.
The adapter follows the default LoRA initialization scheme~\cite{hu2022lora}.

\label{subsec:feature_extraction}

The original VGGT model uses DINOv2~\cite{oquab2023dinov2} as a tokenizer and visual feature extractor for RGB images.
For each frame, a camera token is appended to the image tokens, and the sequence is processed by the AA module.
Prior work~\cite{fanGeneralizableThermalbasedDepth2024} has shown that while DINOv2 is able to extract meaningful thermal features without retraining, the feature distributions of the two modalities differ.
Hence, we use DINOv2 as a tokenizer for both modalities and introduce two learnable thermal camera tokens---a counterpart to the two original VGGT camera tokens (now termed the RGB camera token)---to align both modalities' feature spaces before the AA module.
These tokens encode the intrinsic/extrinsic features for thermal frames---the first token is for the first frame of the sequence, while the other is for all subsequent frames.
RGB and thermal images are independently tokenized using DINOv2 before the RGB and learnable thermal camera tokens are respectively appended to the RGB and thermal tokens.
Then, the augmented sequences are fed into the AA module.
By assigning distinct camera tokens to each modality, the model learns modality-specific representations.
Since these tokens are used in both frame-wise and global interactions in the AA module, they enable differentiated processing of RGB and thermal inputs.
To preserve the pretrained model's behavior during early tuning, we initialize the thermal camera token with the RGB camera token's weights.

We optimize our model using the loss functions from VGGT, formulated as a multitask loss:
\begin{equation}\label{eq:loss}
\mathcal{L} = \lambda_{camera}\mathcal{L}_{camera} + \mathcal{L}_{depth}.
\end{equation}
\noindent
Where $\mathcal{L}_{camera}$ is the Huber loss for camera pose estimation, $\mathcal{L}_{depth}$ is an uncertainty-aware loss for depth prediction, and $\lambda_{camera}$ is a weighting coefficient set to $5.0$, as in the original VGGT model.

\subsection{Data Pre-processing and Batching}
\label{sec:training_procedure}

We apply asymmetric augmentation pipelines for RGB and thermal inputs.
For both RGB and thermal images, we apply random cropping, random aspect ratio adjustments (uniformly sampled from $[0.33, 1.0]$), Gaussian noise, Gaussian blur, and random sharpness adjustments.
For RGB images, we further apply color jittering and grayscale conversion.
For thermal images, we apply random linear transformations of pixel intensities followed by random exponential scaling of pixel values with degrees sampled uniformly in [$1/1.5, 1.5$].
Additionally, we apply random $90^\circ$ rotations to both RGB and thermal images.

To ensure generalization to inputs of varying ratios and sizes, we construct batches of independent RGB and thermal images---i.e., without shared camera poses---sampled from a single RGB+thermal scene.
This constraint prevents the model from relying on trivial RGB-thermal correspondences during training, instead forcing it to learn inter-modal relationships across viewpoints.
The ratio of thermal to RGB images $\tau$ within each batch is randomly sampled from a uniform distribution $U(0,1)$, enabling the model to process arbitrary combinations of RGB and thermal inputs.

%% file: images/MethodCameraToken/flowchart.tex
\begin{tikzpicture}[
    x=0.95cm,y=0.95cm,
    token/.style={draw,minimum width=0.30cm,minimum height=0.30cm},
    therm/.style={token,fill=orange!50},
    thermcam/.style={token,fill=blue!50},
    rgb/.style={token,fill=teal!50},
    rgbcam/.style={token,fill=red!50},
    smalllabel/.style={font=\scriptsize,align=center},
    medlabel/.style={font=\small,align=center},
    tinylabel/.style={font=\tiny,align=center},
    node distance=12mm and 10mm,
    arr/.style={thick, -latex},
    box/.style={minimum width=#1, minimum height=10mm, align=center},
    box/.default=20mm,
    circlebox/.style={draw, circle, minimum size=8mm, align=center}
  ]

  \node[therm] (leg1) at (1.0,6.3) {};
  \node[smalllabel,anchor=west] at (1.3,6.3) {Thermal Tokens\\After DINOv2};
  \node[thermcam] (leg2) at (4.3,6.3) {};
  \node[smalllabel,anchor=west] at (4.6,6.3) {Thermal Camera\\Token};
  \node[box=5mm, right=-3mm of leg2, yshift=2mm] (fire1) {\includegraphics[width=3mm]{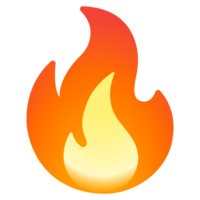}};

  \node[rgb] (leg3) at (7.7,6.3) {};
  \node[smalllabel,anchor=west] at (8.0,6.3) {RGB Tokens\\After DINOv2};
  \node[rgbcam] (leg4) at (10.6,6.3) {};
  \node[smalllabel,anchor=west] at (10.9,6.3) {RGB Camera\\Token};

  \node[box=15mm] at (0.0
  , 4.0)(thermal1) {\includegraphics[width=15mm]{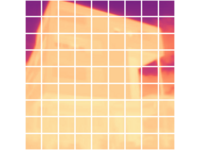}};
  \node[box=15mm] at (0.2
  , 3.8)(thermal2) {\includegraphics[width=15mm]{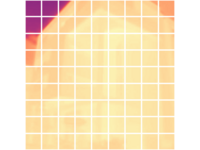}};
  \node[smalllabel,below=-1mm of thermal2] (thermaltext) {Thermal};
  \node[fit=(thermal1)(thermal2), inner sep=0pt] (timgBB) {};

  \node[box=15mm, below=10mm of thermal1](rgb1) {\includegraphics[width=15mm]{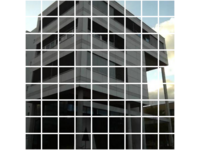}};
  \node[box=15mm, below=10mm of thermal2] (rgb2) {\includegraphics[width=15mm]{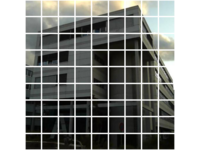}};
  \node[smalllabel, below=20mm of thermaltext] {RGB};
  \node[fit=(rgb1)(rgb2), inner sep=0pt] (rimgBB) {};

  \node[therm, right=13mm of timgBB, yshift=2mm] (dt11) {};
  \node[therm, below=0.5mm of dt11] (dt12) {};
  \node[therm, right=-2.2mm of dt11, yshift=-0.8mm] (dt21) {};
  \node[therm, below=0.5mm of dt21] (dt22) {};
  \node[fit=(dt11)(dt12)(dt21)(dt22), inner sep=3pt] (dtBB) {};

  \node[rgb, below=20mm of dt11] (rgb11) {};
  \node[rgb, below=20mm of dt12] (rgb12) {};
  \node[rgb, below=20mm of dt21] (rgb21) {};
  \node[rgb, below=20mm of dt22] (rgb22) {};
  \node[fit=(rgb11)(rgb12)(rgb21)(rgb22), inner sep=3pt] (drBB) {};

  \node[therm, right=13mm of dt11] (wcam_dt11) {};
  \node[therm, below=0.5mm of wcam_dt11] (wcam_dt12) {};
  \node[therm, right=13mm of dt21] (wcam_dt21) {};
  \node[therm, below=0.5mm of wcam_dt21] (wcam_dt22) {};
  \node[thermcam, above=0.5mm of wcam_dt11] (camt1) {};
  \node[thermcam, above=0.5mm of wcam_dt21] (camt2) {};
  \node[fit=(wcam_dt11)(wcam_dt12)(wcam_dt21)(wcam_dt22), inner sep=3pt] (thermal_wcamBB) {};

  \node[rgb, below=20mm of wcam_dt11] (wcam_rgb11) {};
  \node[rgb, below=20mm of wcam_dt12] (wcam_rgb12) {};
  \node[rgb, below=20mm of wcam_dt21] (wcam_rgb21) {};
  \node[rgb, below=20mm of wcam_dt22] (wcam_rgb22) {};
  \node[rgbcam, above=0.5mm of wcam_rgb11] (rcam1) {};
  \node[rgbcam, above=0.5mm of wcam_rgb21] (rcam2) {};
  \node[fit=(wcam_rgb11)(wcam_rgb12)(wcam_rgb21)(wcam_rgb22), inner sep=3pt] (rgb_wcamBB) {};

  \node[smalllabel, above=4mm of camt1] (addt) {Add thermal\\camera tokens};
  \draw[arr] (addt) -- (camt1);

  \node[smalllabel, above=4mm of rcam1] (addr) {Add RGB\\camera tokens};
  \draw[arr] (addr) -- (rcam1);

  \begin{scope}[local bounding box=attention_block]
    \node[box=17mm, right=20mm of camt1, yshift=-0mm, minimum height=15mm, rounded corners] (attention) {Alternating\\Attention};
    \node[box=18.5mm, draw=3dmodelColor, below=of attention, minimum height=15mm, rounded corners, fill=3dmodelColor!5] (lora) {+ LoRA \includegraphics[width=3mm]{images/MethodCameraToken/fire.png}};
  \end{scope}

  \node[thermcam, right=15mm of attention_block, yshift=18mm] (fincamt1) {};
  \node[thermcam, below=0.5mm of fincamt1] (fincamt2) {};
  \node[rgbcam, below=0.5mm of fincamt2] (fincamr1) {};
  \node[rgbcam, below=0.5mm of fincamr1] (fincamr2) {};
  \node[fit=(fincamt1)(fincamt2)(fincamr1)(fincamr1), inner sep=3pt] (fincamBB) {};

  \node[therm, below=3.0mm of fincamr2] (fint11) {};
  \node[therm, below=0.5mm of fint11] (fint12) {};
  \node[therm, below=0.5mm of fint12] (fint21) {};
  \node[therm, below=0.5mm of fint21] (fint22) {};
  \node[rgb, below=0.5mm of fint22] (finr11) {};
  \node[rgb, below=0.5mm of finr11] (finr12) {};
  \node[rgb, below=0.5mm of finr12] (finr21) {};
  \node[rgb, below=0.5mm of finr21] (finr22) {};
  \node[fit=(fint11)(fint12)(fint21)(fint22)(finr11)(finr12)(finr21)(finr22), inner sep=3pt] (findepBB) {};

  \begin{scope}[local bounding box=out1_block]
    \coordinate[right=35mm of fincamBB, yshift=-3.7mm] (out1_center);
    \foreach \dx/\dy/\camcol in
    {-0.5/-0.3/red,  0.5/-0.3/red,
    -0.5/ 0.3/blue, 0.5/ 0.3/blue} {

      \begin{scope}[shift={($(out1_center)+(\dx,\dy)$)}]
        \coordinate (A) at (-1.2,-0.2);
        \coordinate (B) at (-0.8,-0.25);
        \coordinate (C) at (-0.8,0.25);
        \coordinate (D) at (-1.2,0.2);
        \coordinate (O) at (-0.4,0);

        \draw[draw=\camcol] (A) -- (B) -- (C) -- (D) -- cycle;
        \draw[draw=\camcol] (O) -- (A);
        \draw[draw=\camcol] (O) -- (B);
        \draw[draw=\camcol] (O) -- (C);
        \draw[draw=\camcol] (O) -- (D);
      \end{scope}
    }

    \node[smalllabel, above=6mm of out1_center, xshift=-8mm] {RGB+Thermal\\Cameras};

  \end{scope}

  \begin{scope}[local bounding box=out2_block]
    \coordinate[below=26mm of out1_block, xshift=-2mm] (out2_center);

    \node[box=15mm, above=1mm of out2_center] (depth_t1) {\includegraphics[width=15mm]{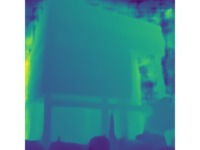}};

    \node[box=15mm, below=-12mm of depth_t1, xshift=2mm] (depth_t2) {\includegraphics[width=15mm]{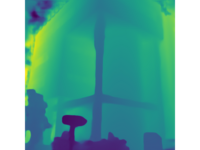}};

    \node[box=15mm, below=-12mm of depth_t2, xshift=2mm] (depth_r1) {\includegraphics[width=15mm]{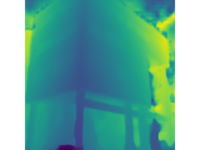}};

    \node[box=15mm, below=-12mm of depth_r1, xshift=2mm] (depth_r2) {\includegraphics[width=15mm]{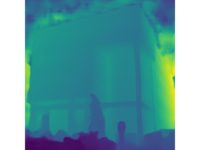}};

    \node[fit=(depth_t1)(depth_t2)(depth_r1)(depth_r2), inner sep=0pt] (depBB) {};

    \node[smalllabel, above=-1mm of depBB]{RGB+Thermal\\Depths};
  \end{scope}

  \begin{pgfonlayer}{background}

    \draw[arr](timgBB)-- node[above] {DINO}(dtBB);
    \draw[arr](rimgBB)-- node[above] {DINO}(drBB);
    \draw[arr] (dtBB) -- (thermal_wcamBB);
    \draw[arr] (drBB) -- (rgb_wcamBB);
    \draw[arr] (thermal_wcamBB) -- (thermal_wcamBB -| attention_block.west);
    \draw[arr] (rgb_wcamBB) -- (rgb_wcamBB -| attention_block.west);
    \draw[arr] (attention_block.east |- fincamBB) -- (fincamBB);
    \draw[arr] (attention_block.east |- findepBB) -- (findepBB);

    \draw[arr] (fincamBB) -- node[above, align=center] {Camera\\Head} (out1_block);
    \draw[arr] (findepBB) -- node[above, align=center] {Depth\\Head} (out2_block);

    \node[rounded corners, fill=biodiversitycolor!5, draw=biodiversitycolor,  inner sep=2pt, fit=(attention_block)] {};
    \node[rounded corners, fill=challengeColor!5, draw=challengeColor,  inner sep=2pt, fit=(out1_block)] {};
    \node[rounded corners, fill=applicationColor!5, draw=applicationColor,  inner sep=2pt, fit=(out2_block)] {};
  \end{pgfonlayer}

\end{tikzpicture}

%% file: sec/5_experiments.tex
\section{Datasets}
\label{sec:dataset}

\subsection{Novel Challenging Cross-Spatial \rgbt Dataset}\label{subsec:complex_dataset}

We introduce a novel multimodal \rgbthermal dataset of 1,890 paired images across nine diverse scenes, captured using a FLIR One Pro LT~\cite{FLIR_One_Pro_Product_Page_2026}.
The dataset includes residential and outdoor environments (e.g., buildings, structures) as well as objects (e.g., metallic items, a telescope), which are visualized in \cref{fig:oursdataset}.
To assess cross-spatial reconstruction, each scene includes paired RGB-thermal images captured along two distinct trajectories.
The dataset is publicly available~\cite{SearDataset}.

The dataset consists of two subsets, differentiated by trajectory characteristics and environmental conditions.
The first subset consists of six scenes (upper section of \cref{fig:oursdataset}) captured under consistent, well-lit conditions (e.g., daytime or artificially illuminated indoor settings).
Trajectories are non-intersecting, simulating independent data collection by separate sensors.
Ground-truth poses are estimated using VGGT on all RGB images from both trajectories.
The second subset (bottom section of \cref{fig:oursdataset}) consists of three scenes captured under varying lighting (i.e., outdoor scenes spanning day/night).
Since half of all scenes are low-illumination, RGB-based camera pose estimation is unreliable.
Given the high uncertainty of pose estimation from thermal images, this subset is used for qualitative assessment due to the lack of reliable ground truth.

\input{images/OursDataset/two_trajectories}

\subsection{Publicly Available Datasets}\label{sec:subsec:data}

Despite the scarcity of RGB-thermal datasets with ground-truth poses, we show that fine-tuning our model requires only limited real-world data.
Our training set includes 66 scenes (around 15,000 RGB-thermal image pairs), aggregated from five publicly available datasets and randomly split into 48 training and 18 evaluation scenes: ThermoScenes~\cite{hassanThermoNeRFMultimodalNeural2025} 17(train)/5(val), ThermalNeRF~\cite{thermalnerf} 7/2, ThermalGaussian~\cite{luThermalGaussianThermal3D2024a} 11/3, ThermalMix~\cite{thermalmix} 4/2, and the Radar Forest (\orebrodataset)~\cite{kubelka2026vikinghilldatasetlidarradarcamera} Dataset 9/6---scene names per split are in the Supplementary Material (Sec. K).
We exclude datasets not designed for scene reconstruction (e.g., autonomous driving datasets like~\cite{lee2022vivid++, ms2}).

For ThermoScenes, ThermalGaussian, and ThermalMix, RGB and thermal images are paired with identical extrinsics.
We derive ground-truth extrinsics, intrinsics, and depth from VGGT reconstructions on RGB images, then transfer these to corresponding thermal frames.
Since our goal is cross-modal reconstruction from unpaired data (not improving VGGT's pose estimation), using VGGT-derived RGB poses as ground truth does not bias validation.
In ThermalNeRF~\cite{thermalnerf}, RGB and thermal images are paired via a known rigid transformation.
We estimate RGB extrinsics using VGGT and derive thermal extrinsics by applying the rigid transform.
Thermal depth maps are generated by projecting RGB depth estimates into 3D world points and reprojecting them onto the thermal frame.
The \orebrodataset dataset includes high-precision camera poses captured via a motion capture system and sparse LiDAR point clouds---both serving as ground truth in evaluation, eliminating the need for ground truth estimation using VGGT.
We generate depth maps by projecting LiDAR points into the camera frame using the provided extrinsics and intrinsics.

\subsection{Data Split of Evaluation Scenes}
\label{sec:dataset:spliteval}

Validation uses a fixed thermal ratio of $\tau = 0.5$ (to avoid modality imbalance) and a batch size equal to the total number of frames per scene.
As for training, we ensure that no RGB and thermal images in a batch share the same pose; given paired images, each scene is evaluated twice by swapping RGB and thermal image selections.
This eliminates possible selection bias (since both runs are complementary to each other) and yields 36 evaluation cases (18 scenes $\times$ 2).

For evaluation on \datasetname, we use the dataset's split into two non-overlapping trajectories, pairing thermal images from one with RGB from the other.
Each of the 6 scenes is evaluated twice (once per modality-trajectory pair), yielding 12 evaluation cases.
We additionally include SmokeSeer~\cite{jain2025smokeseer}, which provides two RGB-thermal drone scenes (both with and without dense smoke) but lacks ground-truth poses.
We use SmokeSeer for qualitative evaluation only since, unlike other datasets, VGGT-based pose estimation fails here due to high uncertainty on smoke-occluded RGB frames.

\section{Experiments}\label{sec:experiments}

\subsection{Implementation Details}\label{subsec:implementation}

Our model's trained weights and implementation are available online~\cite{malcolmmielleSchindlerEPFLLabSEARECCV2026}.
We use the VGGT checkpoint from HuggingFace\footnote{https://huggingface.co/facebook/VGGT-1B}.
Training uses \texttt{bf-16} mixed precision, \texttt{AdamW} optimizer with learning rate of $5 \cdot 10^{-5}$ and weight decay of $10^{-2}$, linear warmup scheduling (first 10\% of epochs, LR from $0.0$ to $5 \cdot 10^{-5}$).
We train the model for $100$ epochs with a batch size of $24$ (around 2 days on a single A100).
To improve scalability and generalization to arbitrary input sizes, we follow VGGT's approach and partition the 24 frames of each batch into $N \in \{1, 2, 3, 4, 6, 12\}$ equal-length sequences.
To reduce variations due to known numerical instabilities~\cite{yuanUnderstandingMitigatingNumerical2025}, we fine-tune 13 models with different seeds and report their average metrics.

To demonstrate that bridging the multimodal knowledge gap requires only minimal fine-tuning, the LoRA rank used is $r = 64$ and the scaling factor $\alpha = 128$, yielding $\sim$50M trainable parameters---less than 5\% of the original model's size.
Since, as established in the original LoRA paper, tuning $\alpha$ is equivalent to adjusting the learning rate, we fix $\alpha$ and optimize the learning rate empirically across all scenes in the ThermoScenes dataset.

\subsection{Baselines and Metrics}

We benchmark against traditional SfM (COLMAP~\cite{schonbergerStructurefrommotionRevisited2016a} with SuperPoint + SuperGlue~\cite{sarlin2020superglue}) and deep-learning based models (DUSt3R~\cite{wangDust3rGeometric3d2024}, MASt3R~\cite{leroyGroundingImageMatching2025}, MapAnything~\cite{keetha2026mapanything}, and pretrained VGGT).
Since these methods lack multimodal support, we input RGB/thermal images as a single modality.
We also benchmark against hybrid approaches mixing feature matching and deep learning.
We use $\text{MA}_{\text{ELoFTR}}$ (ELoFTR~\cite{eloftr} trained with MatchAnything~\cite{he2025matchanything}) and $\text{MINIMA}_{\text{ROMA}}$ (RoMA~\cite{edstedt2024roma} trained with MINIMA~\cite{ren2024minima}) to establish RGB-thermal correspondences, followed by DIM~\cite{morelli2024_deep_image_matching} for 3D reconstruction and pose estimation.
We also use MP-SfM~\cite{pataki2025mp}, which builds on COLMAP and includes depth and normal predictions, as well as correspondences from a learned matching model---we use MINIMA$_{\text{ROMA}}$~\cite{ren2024minima} as the matching model.

We use publicly available implementations and official pretrained weights for all baselines and use identical data splits and computational resources (one A100 GPU) for all methods.
Classical approaches are evaluated using their default parameters.
Due to computational constraints and the fact that this is not the best-performing method, we evaluate MP-SfM only on the Public Datasets and not on the more challenging SEAR dataset. Qualitative renderings for this method are provided in the supplementary materials.

\label{subsec:metrics_description}

Following VGGT~\cite{wang2025vggt}, we evaluate the methods using the RRA@30 and RTA@30 (percentages of image pairs with relative rotation/translation errors $<30^\circ$) and AUC@30 (area under the accuracy-threshold curve of the maximum values between RRA and RTA with thresholds $[5.0, 15.0, 30.0]^\circ$).
We also report the processed frames per second (FPS) and registration rate (Reg.)~\cite{light3r} (percentage of successfully registered images).
We compute the quality metrics on registered cameras only, excluding images without predicted poses.
When LiDAR data is available, we report point cloud completeness (PCC) (the mean nearest-neighbor distance from reconstruction to LiDAR) and point cloud accuracy (PCA) (the mean nearest-neighbor distance from LiDAR to reconstruction).
We also report the Chamfer distance, i.e., the average of PCC and PCA.
To compute these metrics, we align estimated and ground-truth camera poses using Umeyama alignment~\cite{umeyama}, back-project estimated depth maps, and calculate the metrics between estimated and ground-truth point clouds.
To mitigate bias toward methods that perform well on only a subset of scenes, we report the metrics over the errors across all datasets/scenes (for scene-agnostic evaluation) in \cref{tab:concat}.

\subsection{Multimodal Camera Pose Estimation}
\label{subsec:mm_camera_pose_estimation}

\begin{figure}[ht]
  \vspace{4mm}
  \centering
  \begin{overpic}[page=4,width=0.9\linewidth]{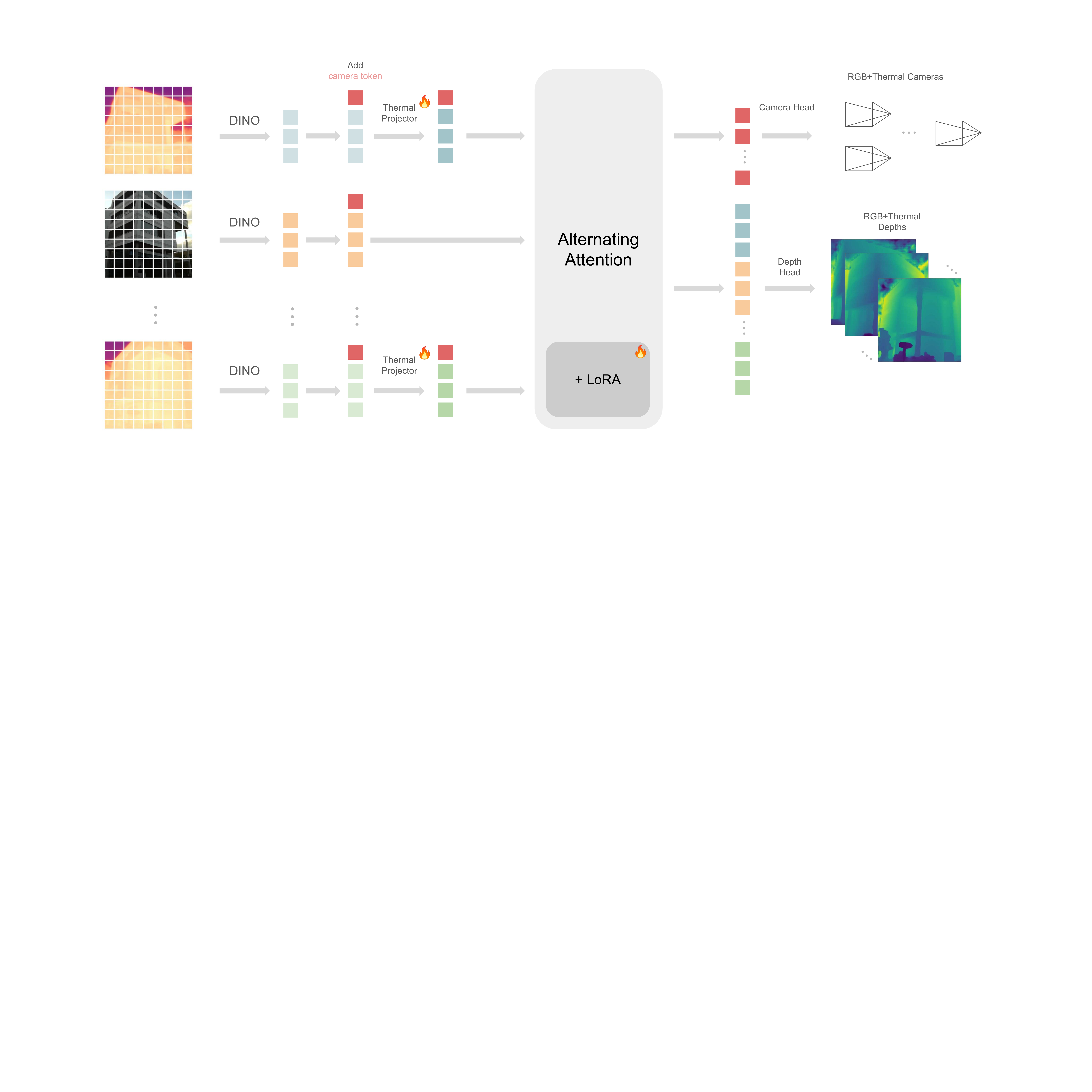}
    \put(3.5, 101){\tiny{Ground Truth}}
    \put(21, 101){\tiny{COLMAP}}
    \put(36.5, 101){\scalebox{0.5}{MINIMA$_{ROMA}$}}
    \put(52, 101){\tiny{MASt3R}}
    \put(72, 101){\tiny{VGGT}}
    \put(85, 101){\scalebox{0.6}{\ours (ours)}}
  \end{overpic}
  \caption{
    Qualitative results comparing RGB/thermal reconstructions (camera poses in red/blue); we show results for 4 methods at rows 1, 3, and 5, and zoom in on more interesting reconstruction details in rows 2, 4, and 6.
    Our method (\ours) achieves higher accuracy, consistency, and level of detail than other methods.
  }
  \label{fig:mm_point_clouds}
  \vspace{-5mm}
\end{figure}

Our method significantly outperforms all baselines (see \cref{tab:concat}) while maintaining a high registration rate. E.g., we achieve an AUC@30 of $70.0$ on the Public Datasets and of $62.8$ on our \datasetname, compared to $41.0/48.2$ for $\text{MINIMA}_{\text{ROMA}}$, the method with the second-highest performance and a strong registration rate—though COLMAP achieves better metrics, its low registration rate artificially boosts those results.
Non-multimodal approaches fail on thermal images due to the lack of discriminative features for cross-modal alignment, often producing disjoint reconstructions (see \cref{fig:mm_point_clouds})---e.g., COLMAP only registers frames of one modality.
$\text{MA}_{\text{ELoFTR}}$ produces sparse matches leading to low registration rates and poor 3D reconstructions.
Interestingly, even methods trained with grayscale images (which could be construed as a substitute for thermal images), such as VGGT and MapAnything, fail to register or estimate depth in real thermal images, leading to incorrect 3D reconstructions: MapAnything exhibits the same failure mode as VGGT, reconstructing two separate point clouds.
These failure cases underscore the necessity of both real data and specialized training strategies for successful RGB-Thermal reconstruction.
Furthermore, our method is $200\times$ faster (9.94 FPS vs. $\text{MINIMA}_{\text{ROMA}}$'s 0.05 FPS), nearly matching VGGT's speed (9.94 FPS vs. 10.46 FPS).
Unlike other methods, \ours achieves consistent RGB-T scene reconstruction.

\subsection{Multimodal Point Cloud Reconstruction}\label{subsec:mm_point_cloud_reconstruction}

As shown in \cref{tab:concat}, \ours achieves the highest point cloud metrics: PCC of $0.06$, PCA of $0.47$, and Chamfer distance of $0.27$.
The next-best method, MASt3R, only has a PCC of $0.27$, PCA of $0.66$, and Chamfer distance of $0.46$, while COLMAP+SPSG produces single-modality sparse point clouds.
For hybrid methods, the quality of 3D reconstructions is dependent on the 2D-2D correspondences' quality and density; $\text{MA}_{\text{ELoFTR}}$ only detected a small number of correspondences, leading to low metrics.
$\text{MINIMA}_{\text{ROMA}}$ finds more correspondences, but their uncertainty results in low-quality reconstructions---its Chamfer distance is only $1.03$ compared to $0.27$ for \ours.
Similarly, MP-SfM achieves a Chamfer distance of only $1.08$, due to the limited robustness of depth estimation on thermal images.

While point cloud metrics are only calculated against scenes with LiDAR ground truth, \cref{fig:mm_point_clouds} provides additional qualitative comparisons across diverse scenes.
DUSt3R, MASt3R, and VGGT frequently generate disjoint 3D representations for each modality, exhibiting inconsistencies in both scale and spatial alignment between modalities.
$\text{MINIMA}_{\text{ROMA}}$ reconstructs less detailed point clouds than our method, which reconstructs consistent multimodal point clouds.

\input{tables/metrics_concat}

\subsection{Qualitative Evaluation}

\input{images/smokeseer_with_rgb.tex}

We perform qualitative evaluation on SmokeSeer~\cite{jain2025smokeseer} and the subset of our new dataset with varying lighting conditions (see \cref{fig:mm_smokeseer}).
Under challenging conditions, \ours estimates well-aligned 3D camera poses across both RGB and thermal modalities, reconstructing the scenes with details, and demonstrating robust performance under challenging conditions (when both modalities are captured at different times or smoke is occluding the scene).
In comparison, MASt3R and $\text{MINIMA}_\text{ROMA}$ misalign RGB/thermal reconstructions (rotated, improperly scaled thermal point cloud, with camera poses not consistent between modalities).
On the other hand, VGGT sometimes achieves partial alignment (e.g., top image of \cref{fig:mm_smokeseer}), but the results are inconsistent and noisy (e.g., alignments are wrong on scenes from our new dataset), especially in the thermal domain.
Additional qualitative results are provided in the Supplementary Material.

\subsection{Varying thermal to RGB ratio}

Prior experiments assumed a balanced RGB-thermal ratio ($\tau = 0.5$).
To evaluate robustness to unequal ratios, we vary $\tau\in[0,1]$ and report camera pose estimation in \cref{fig:thermal_ratio_metrics}.
For each $\tau$, we perform three random inferences per scene with randomly selected non-corresponding RGB-thermal pairs to reduce statistical variance.
\cref{fig:thermal_ratio_metrics} shows a slow but constant performance decrease as $\tau$ increases, with a slight recovery after $\tau \approx 0.75$.

We hypothesize that the metrics decrease when $\tau \in [0.0, 0.75]$ is mostly due to thermal-specific characteristics---e.g., the ghosting effect~\cite{hassanThermoNeRFMultimodalNeural2025}, which reduces sharpness and contrast in thermal images, making the pose reconstruction more difficult.
For $\tau \in [0.75, 1.0]$, quality improves because the model increasingly operates in a near single-modality regime, reducing multimodal alignment complexity, while the larger number of thermal views provides stronger scene coverage for thermal reconstruction.
The minimum occurs at $\tau \approx 0.75$ (rather than $\tau=0.5$) since, while more thermal frames weaken pose cues, a thermal-dominant input reduces cross-modal alignment and increases thermal view coverage.

\input{images/graphs/thermal_ratio_figure.tex}

\subsection{Thermal to RGB Alignment}
\label{subsec:thermal_to_rgb_alignment}

\begin{figure}[t]
  \centering
  \resizebox{0.95\textwidth}{!}{\input{images/RGB_Thermal_Features/distances_cosine.pgf}}
  \caption{
    The blue line represents the median difference between RGB-to-RGB and RGB-to-thermal cosine similarity dependence across layers for \ours method.
    The orange line represents the same dependency for the VGGT model.
    The filled area represents the boundary from $0.25-$ to $0.75-$quantiles.
  }
  \label{fig:distance_between_thermal_rgb_features}
\end{figure}

In contrast to VGGT, we believe that our method explicitly keeps the distributions of thermal and RGB tokens in the AA module aligned.
To validate this hypothesis, we feed RGB-thermal image pairs into our model and extract intermediate outputs from the AA module's layers.
We compute the cosine similarity~\cite{robustvggt} between same-level RGB and thermal tokens (excluding camera tokens).
High similarity implies aligned distributions, while low similarity implies divergence.
We compare RGB-to-RGB cosine similarity $x_{r2r}$ (the baseline) to the cosine similarity $x_{r2t}$ between RGB and thermal images.
A large difference between $x_{r2r}$ and $x_{r2t}$ indicates a distribution mismatch, while a small difference suggests that the distributions are close.
Experiments use a batch size of $12$ and thermal ratios $\tau \in [0.25, 0.75]$.

Our results (\cref{fig:distance_between_thermal_rgb_features}) show that for the first 10 layers, both our model and VGGT show low RGB-thermal cosine similarity---interestingly, \textcite{vggt_geometry_analysis} demonstrated that VGGT reconstructs geometry post-\texttt{layer10} of the AA-module.
Beyond \texttt{layer10}, when the geometric reconstruction starts, VGGT's difference between $x_{r2r}$ and $x_{r2t}$ increases, while ours remains consistently low, suggesting our model successfully aligned RGB and thermal token distributions for reconstruction in the whole AA module.

\subsection{Ablation Studies}
\label{subsec:ablation_studies}

We perform extensive ablation studies to support our choice of architecture.
We evaluate: 1) using the non-learnable rgb camera token instead of the learnable thermal camera token, 2) LLaVA~\cite{liu2023llava}'s strategy (adding a learnable thermal projector of size 1024, initialized as the identity transformation, for thermal images), 3) adding a learnable thermal embedding, similar to positional embeddings in transformer~\cite{vaswani2017attention} (initialized as zero to preserve the original model's performance at the beginning of training), 4) only adding LoRA to global-attention layers, and 5) only applying LoRA to frame-attention layers.

\input{tables/metrics_ablation}

\input{images/statistical_significance}

We compute metrics over the aggregated scenes of Public Datasets and \datasetname, and present their mean and standard deviation, as well as statistical analysis in \cref{tab:ablation_study} and \cref{tab:pairwise-significance}.
Our statistical analysis is a one-sided Welch's t-test evaluating the null hypothesis that a given model (row) outperforms another (column) in terms of AUC@30.
Our study reveals that incorporating a learnable thermal projector or learnable thermal embeddings does not improve performance.
Specifically, our method has p-values of 0.05 (trend) and 0.02 (statistically significant) when compared to those models, supporting our decision not to include a learnable thermal projector or learnable thermal embeddings.
For the no-camera-token model, the p-value of 0.24 does not invalidate the null hypothesis.
However, when the no-camera-token model is evaluated against the thermal-embedding model, the p-value (0.09) only shows a trend, while our method achieves statistical significance ($p=0.02$).
Similarly, the no-camera-token model shows no significant difference from the thermal-projector baseline ($p=0.28$), whereas our method reaches statistical significance ($p=0.05$).
These results indicate that, although our architecture does not significantly outperform the no-camera-token model, it provides some improvements and greater stability when compared to other models.

Our method and the global-only model achieved comparable results in terms of statistical significance.
While our analysis does not indicate a clear superiority of one approach over the other, future work could explore applying LoRA layers exclusively to the global attention layer---rather than both the global and frame layers---to further reduce parameter count without compromising performance.

\subsection{LoRA Parameters}

\input{images/lora_r_study/figure}

We evaluate the impact of different values of $r$ for our LoRA layers; due to computational constraints, we run each experiment only once.
We selected $r$ to keep the number of trainable parameters low (approximately 50M, less than 5\% of the original model size).
As shown in~\cref{fig:lora_r}, while increasing the rank $r$ slightly improves the final accuracy---suggesting that higher-rank adapters provide additional representativeness---the gains are modest and only a small $r$ value is enough.
We hypothesize that this is due to VGGT's ability to process thermal images alone: the adapter only needs to estimate the \rgbthermal alignment, a comparatively smaller task.

%% file: images/OursDataset/two_trajectories.tex
\newcommand{\imageWithInset}[4]{%
  \begin{minipage}[c]{#4}
    \centering
    \scriptsize{#1} \par
    \vspace{3pt}
    \begin{tikzpicture}[inner sep=0]
      \node[anchor=south west] (main) at (0,0) {\includegraphics[width=\linewidth]{#2}};
      \node[anchor=south east, xshift=-1pt, yshift=1pt] at (main.south east) {
        \includegraphics[width=0.4\linewidth]{#3}
      };
    \end{tikzpicture}
  \end{minipage}%
}

\newcommand{\imagePlain}[3]{%
  \begin{minipage}[c]{#3}%
    \centering
    \scriptsize{#1} \par
    \vspace{3pt}
    \includegraphics[width=\linewidth]{#2}%
  \end{minipage}%
}

\newcommand{\imagePlainAndInset}[7]{%
  \begin{minipage}[c]{#3}%
    \centering
    \scriptsize{#1} \par
    \vspace{3pt}
    \includegraphics[width=\linewidth]{#2}%
  \end{minipage}%
  \hfill
  \begin{minipage}[c]{#7}%
    \centering
    \scriptsize{#4} \par
    \vspace{3pt}
    \begin{tikzpicture}[inner sep=0]
      \node[anchor=south west] (main) at (0,0) {\includegraphics[width=\linewidth]{#5}};
      \node[anchor=south east, xshift=-1pt, yshift=1pt] at (main.south east) {
        \includegraphics[width=0.4\linewidth]{#6}
      };
    \end{tikzpicture}
  \end{minipage}%
}

\begin{figure}[t]
  \centering
  \begin{subfigure}[t]{0.20\linewidth}
    \imageWithInset{}{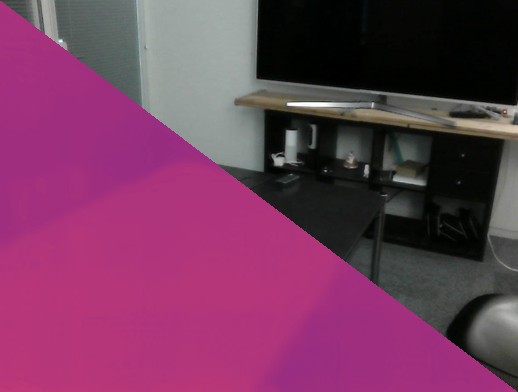}{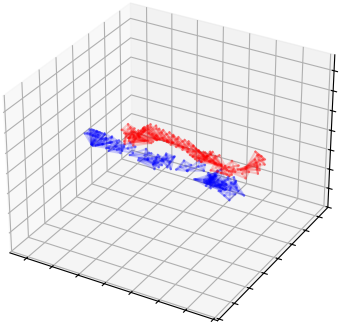}{\linewidth}
    \caption*{Conference Room}
  \end{subfigure}
  \begin{subfigure}[t]{0.20\linewidth}
    \imageWithInset{}{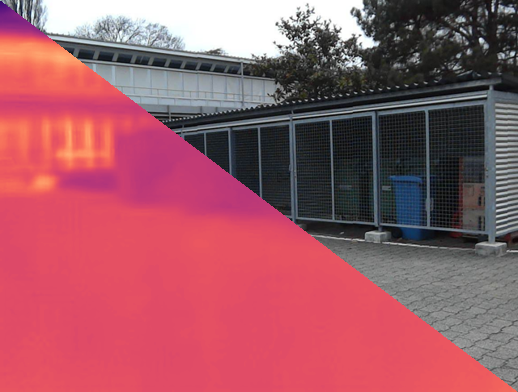}{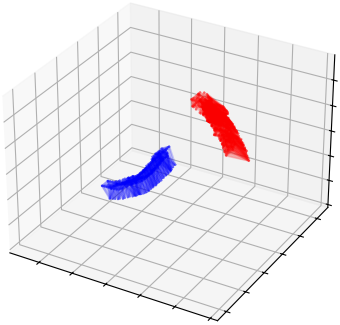}{\linewidth}
    \caption*{Metallic Container}
  \end{subfigure}
  \begin{subfigure}[t]{0.20\linewidth}
    \imageWithInset{}{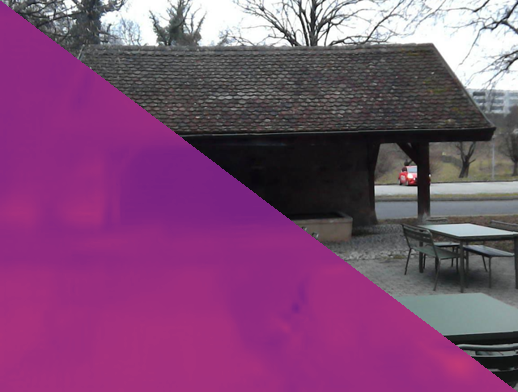}{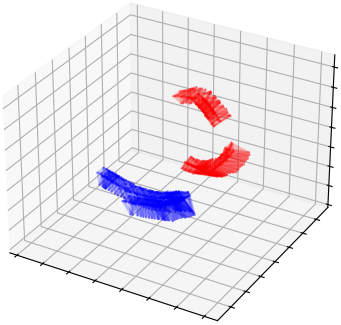}{\linewidth}
    \caption*{Old Fountain}
  \end{subfigure}
  \begin{subfigure}[t]{0.20\linewidth}
     \imageWithInset{}{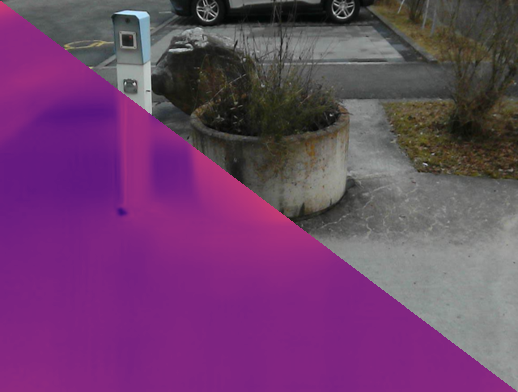}{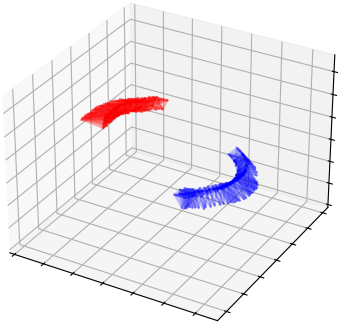}{\linewidth}
    \caption*{Parking Lot}
  \end{subfigure}
  \begin{subfigure}[t]{0.20\linewidth}
    \imageWithInset{}{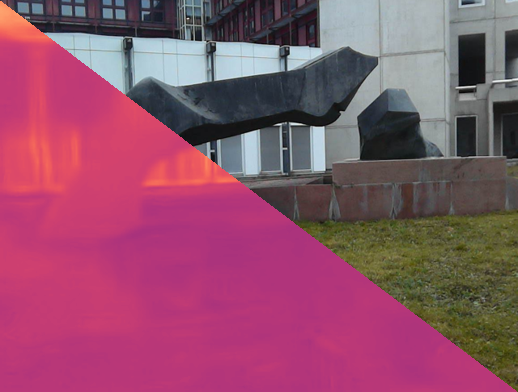}{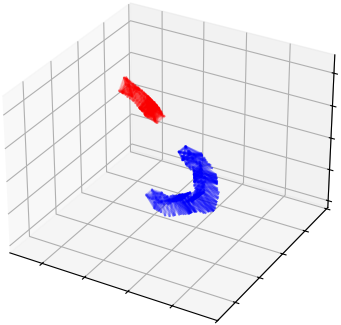}{\linewidth}
    \caption*{Statue}
  \end{subfigure}
  \begin{subfigure}[t]{0.20\linewidth}
    \imageWithInset{}{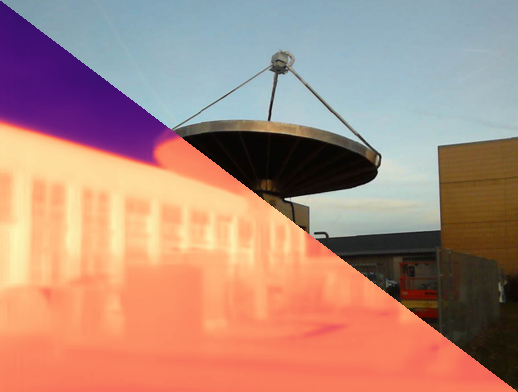}{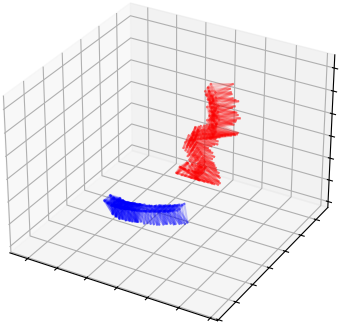}{\linewidth}
    \caption*{Telescope}
  \end{subfigure}
 \begin{subfigure}[t]{0.40\linewidth}
    \imagePlainAndInset{}{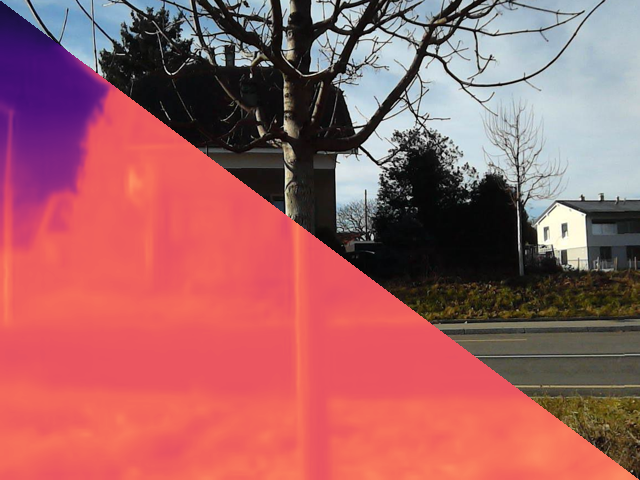}{0.5\textwidth}{}{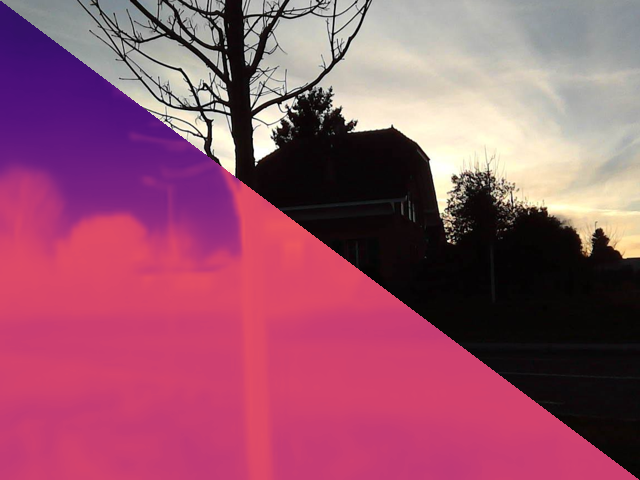}{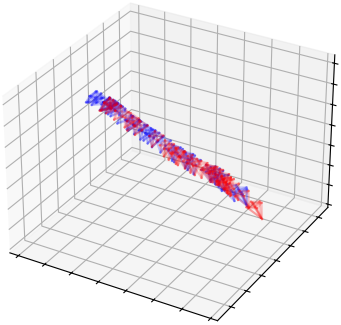}{0.5\textwidth}
    \caption*{House}
  \end{subfigure}
 \begin{subfigure}[t]{0.40\linewidth}
    \imagePlainAndInset{}{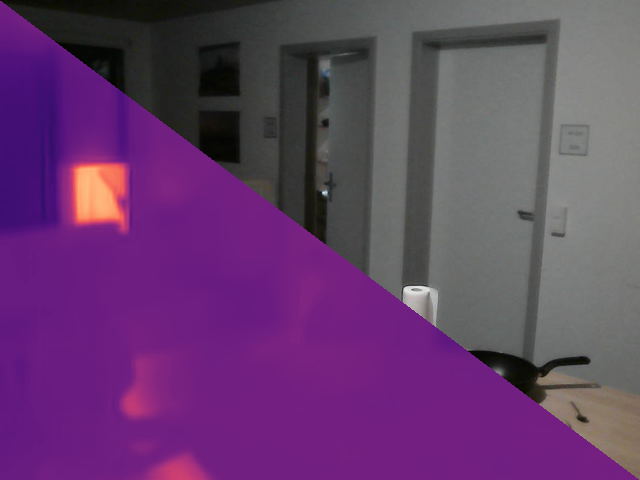}{0.5\textwidth}{}{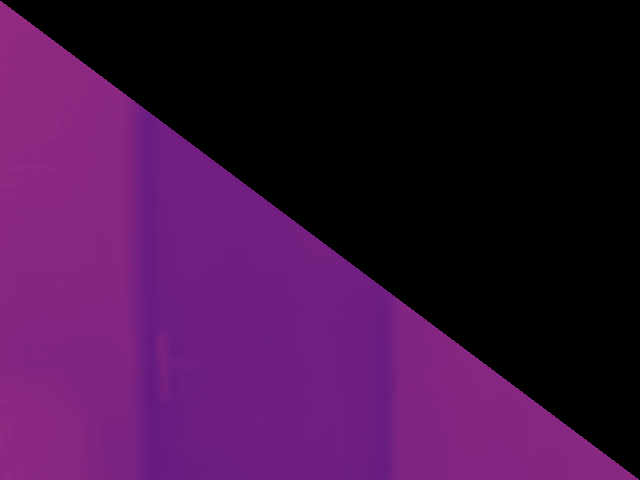}{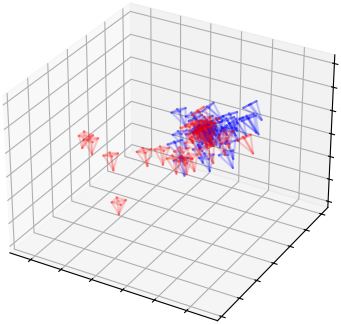}{0.5\textwidth}
\caption*{Living Room}
  \end{subfigure}
 \begin{subfigure}[t]{0.40\linewidth}
    \imagePlainAndInset{}{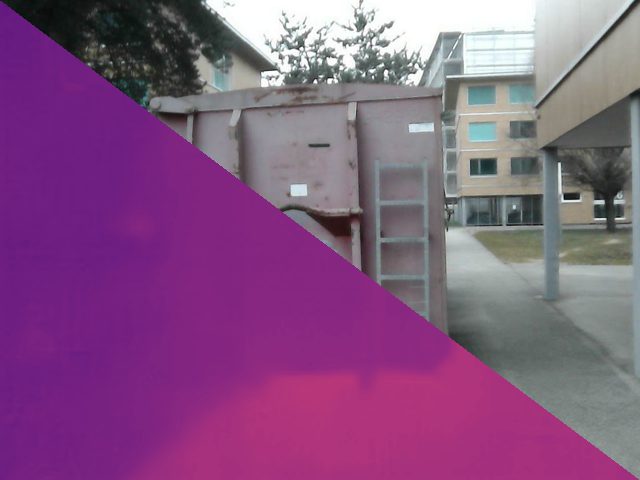}{0.5\textwidth}{}{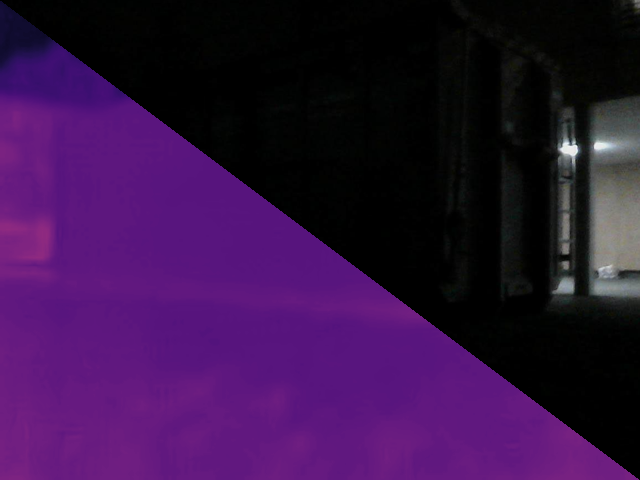}{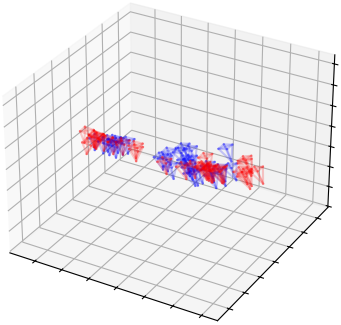}{0.5\textwidth}
\caption*{Red Container}
  \end{subfigure}

  \caption{
    The \datasetname{} includes 9 scenes, each with paired RGB-thermal images captured along two distinct trajectories.
    Ground-truth poses (red/blue for each trajectory) are estimated via VGGT on all RGB images.
    The top 6 scenes feature trajectories under similar lighting, while the bottom 3 have large lighting variations (some RGB images are near fully black).
  }
  \label{fig:oursdataset}
  \vspace{-5mm}
\end{figure}

%% file: tables/metrics_concat.tex
\begin{table}[t]
  \caption{
    Quantitative comparison of 3D reconstruction methods for all datasets. Metrics include \texttt{AUC}, \texttt{RRA}, and \texttt{RTA} @30 (higher is better, $\uparrow$), point cloud accuracy \texttt{PCA}, point cloud completeness \texttt{PCC}, \texttt{Chamfer} distance (lower is better, $\downarrow$), registration rate \texttt{Reg.} (higher is better, $\uparrow$), and frames per second \texttt{FPS} (higher is better, $\uparrow$).
    Best and second-best results are highlighted in \colorbox{best}{blue} and \colorbox{secondbest}{light blue}, respectively.
  }
  \label{tab:concat}
  \centering
  \resizebox{\linewidth}{!}{
    \begin{tabular}{l ccc ccc cc@{\hspace{8pt}}ccc cc}

      \toprule

      & \multicolumn{8}{c}{Public Datasets}
      & \multicolumn{5}{c}{\ours Dataset} \\

      \cmidrule(r){2-9} \cmidrule(l){10-14}

      Method
      & AUC $\uparrow$
      & RRA $\uparrow$
      & RTA $\uparrow$
      & PCA $\downarrow$
      & PCC $\downarrow$
      & Chamfer $\downarrow$
      & Reg (\%) $\uparrow$
      & FPS $\uparrow$
      & AUC $\uparrow$
      & RRA $\uparrow$
      & RTA $\uparrow$
      & Reg (\%) $\uparrow$
      & FPS $\uparrow$ \\

      \cmidrule(r){1-9}  \cmidrule(l){10-14}

      \makecell[l]{COLMAP\\+SPSG} &  \cellcolor{secondbest}{57.6} & \cellcolor{secondbest}{82.5} & \cellcolor{secondbest}{74.6} &  1.64 & 1.20 & 1.42 &  44.7 & 0.44 &  \cellcolor{best}{74.4} & \cellcolor{best}{99.9} & \cellcolor{best}{95.9} &  27.9 & 0.66\\

      \cmidrule(r){1-9}  \cmidrule(l){10-14}

      $\text{MA}_{\text{ELoFTR}}$ &  13.1 & 70.1 & 41.2 &  \cellcolor{secondbest}{0.49} & 5.16 & 2.82 &  21.8 & 0.21 &  15.9 & 73.7 & 53.1 &  \cellcolor{secondbest}{37.3} & 0.17\\
      $\text{MINIMA}_{\text{ROMA}}$ &  41.0 & 68.3 & 63.0 &  0.97 & 1.09 & 1.03 &  \cellcolor{secondbest}{98.7} & 0.05 &  48.2 & 65.9 & 68.9 &  \cellcolor{best}{100.0} & 0.04\\
      MP-SfM &  31.4 & 78.0 & 52.5 &  1.59 & 0.58 & 1.08 &  \cellcolor{best}{100.0} & 0.04 & - & - & - & - & - \\

      \cmidrule(r){1-9}  \cmidrule(l){10-14}

      DUSt3R &  18.9 & 45.9 & 42.1 &  0.72 & 4.72 & 2.72 &  \cellcolor{best}{100.0} & 0.66 &  18.7 & 51.0 & 39.2 &  \cellcolor{best}{100.0} & 0.67\\
      MASt3R &  30.8 & 69.7 & 55.8 &  0.66 & \cellcolor{secondbest}{0.27} & \cellcolor{secondbest}{0.46} &  \cellcolor{best}{100.0} & 0.24 &  39.1 & 54.9 & 56.3 &  \cellcolor{best}{100.0} & 0.25\\
      
      \cmidrule(r){1-9}  \cmidrule(l){10-14}

      VGGT &  22.9 & 50.7 & 48.5 &  1.22 & 2.60 & 1.91 &  \cellcolor{best}{100.0} & \cellcolor{best}{10.46} &  23.3 & 50.5 & 56.4 &  \cellcolor{best}{100.0} & \cellcolor{best}{10.51}\\
      
      MapAnything &  21.4 & 51.3 & 47.4 &  0.69 & 3.76 & 2.23 &  \cellcolor{best}{100.0} & 1.98 & 23.2 & 51.4 & 52.2 & \cellcolor{best}{100.0} & 2.12 \\
      
      \ours &  \cellcolor{best}{70.0} & \cellcolor{best}{90.6} & \cellcolor{best}{87.6} &  \cellcolor{best}{0.47} & \cellcolor{best}{0.06} & \cellcolor{best}{0.27} &  \cellcolor{best}{100.0} & \cellcolor{secondbest}{9.94} &  \cellcolor{secondbest}{62.8} & \cellcolor{secondbest}{83.7} & \cellcolor{secondbest}{84.2} &  \cellcolor{best}{100.0} & \cellcolor{secondbest}{10.22}\\ \bottomrule

    \end{tabular}
  }
\end{table}

%% file: images/smokeseer_with_rgb.tex
\begin{figure}[t]
  \centering
  \begin{subfigure}[t]{0.19\textwidth}
    \includegraphics[width=\textwidth]{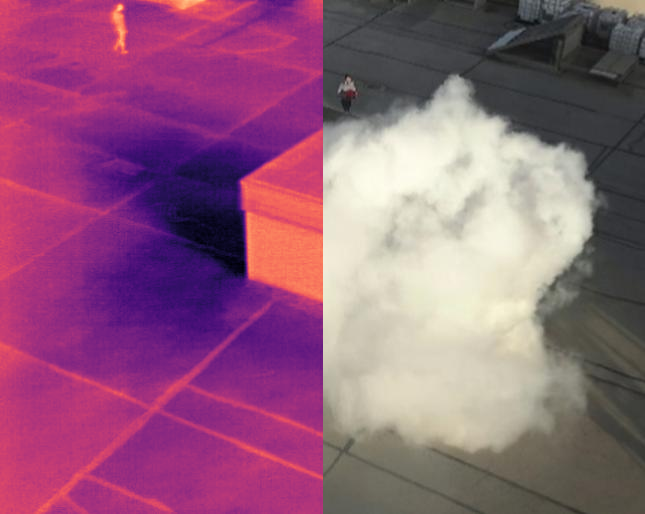}
  \end{subfigure}
  \begin{subfigure}[t]{0.19\textwidth}
    \includegraphics[width=\textwidth]{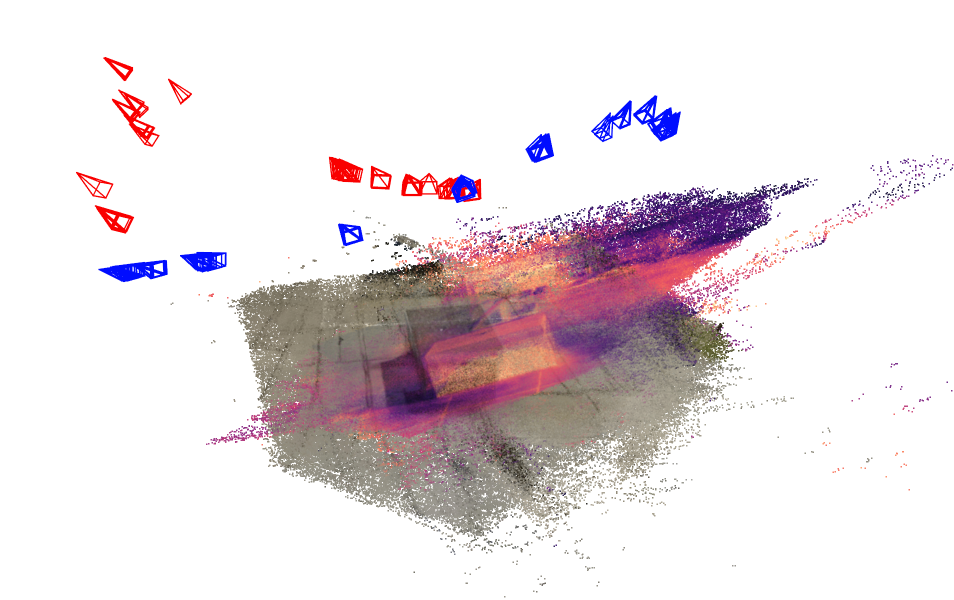}
    \label{sec:evaluation:fig:mast3r_drone_bathroom}
  \end{subfigure}
  \begin{subfigure}[t]{0.19\textwidth}
    \includegraphics[width=\textwidth]{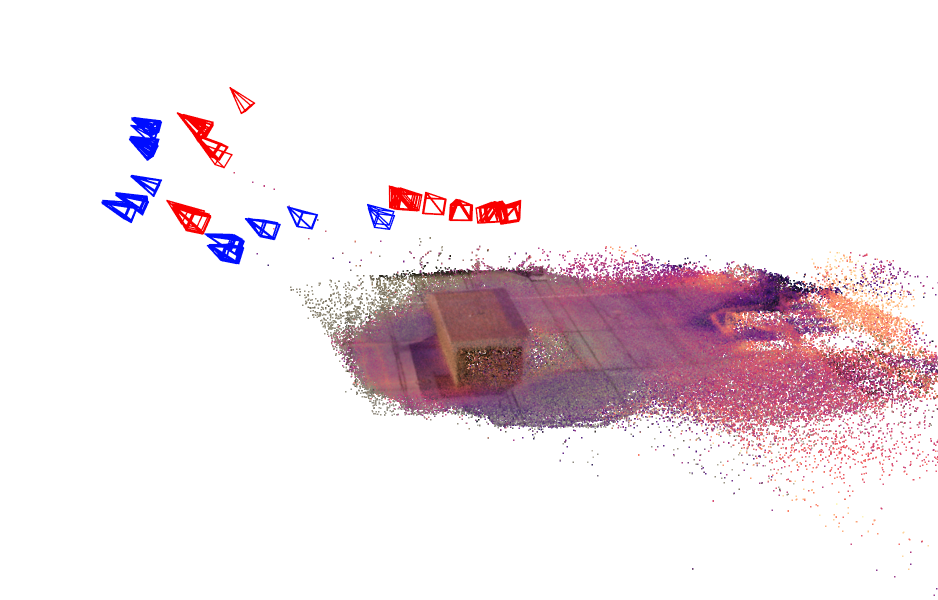}
    \label{sec:evaluation:fig:minima_drone_bathroom}
  \end{subfigure}
  \begin{subfigure}[t]{0.19\textwidth}
    \includegraphics[width=\textwidth]{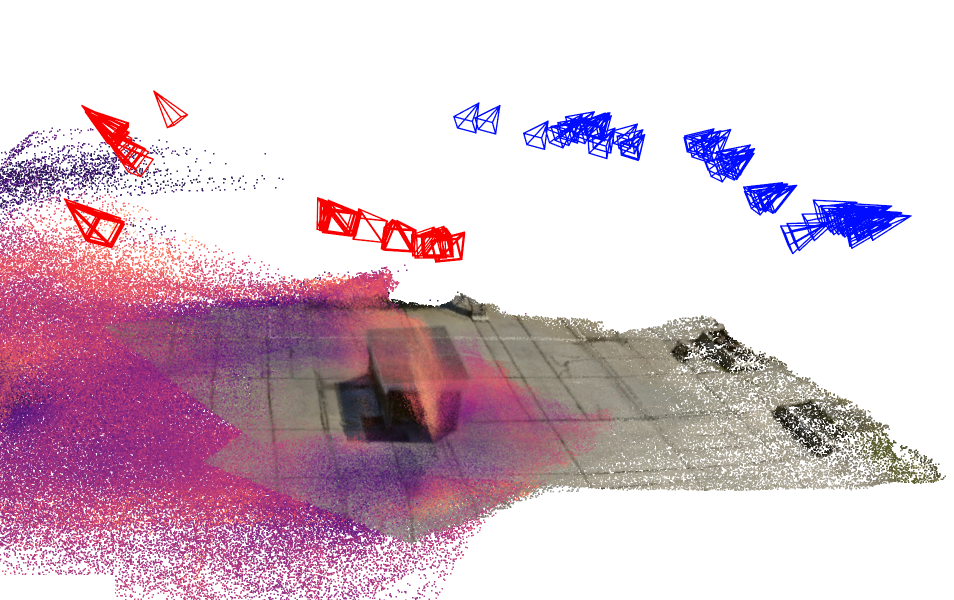}
    \label{sec:evaluation:fig:vggt_drone_bathroom}
  \end{subfigure}
  \begin{subfigure}[t]{0.19\textwidth}
    \includegraphics[width=\textwidth]{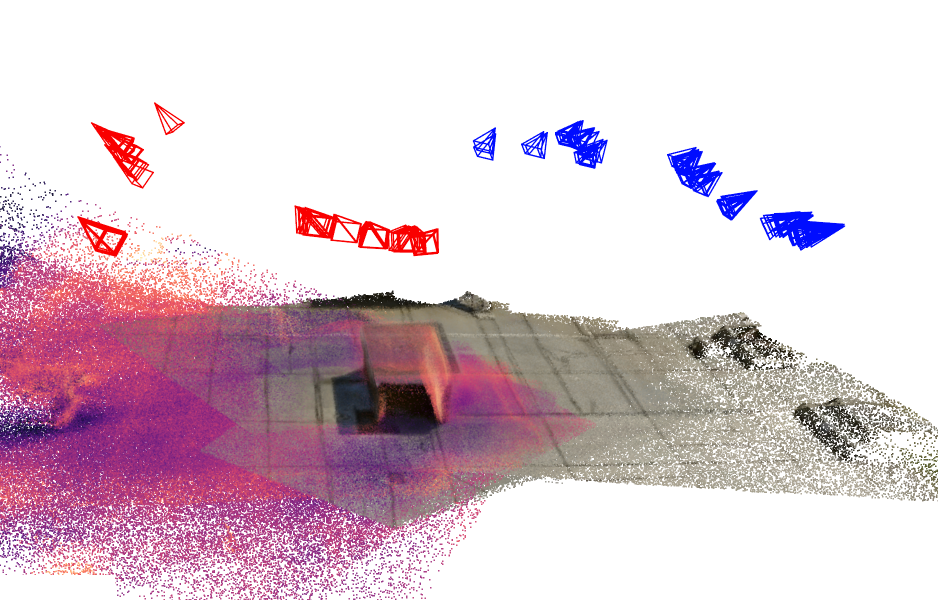}
    \label{sec:evaluation:fig:ours_drone_bathroom}
  \end{subfigure}
  \begin{subfigure}[t]{0.19\textwidth}
    \includegraphics[width=\textwidth]{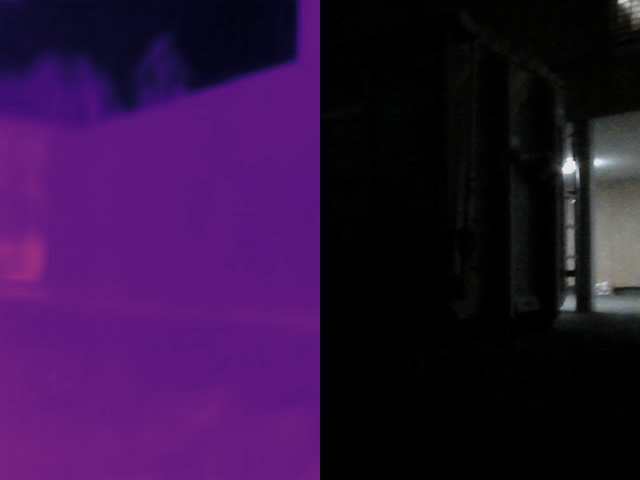}
  \end{subfigure}
  \begin{subfigure}[t]{0.19\textwidth}
    \includegraphics[width=\textwidth]{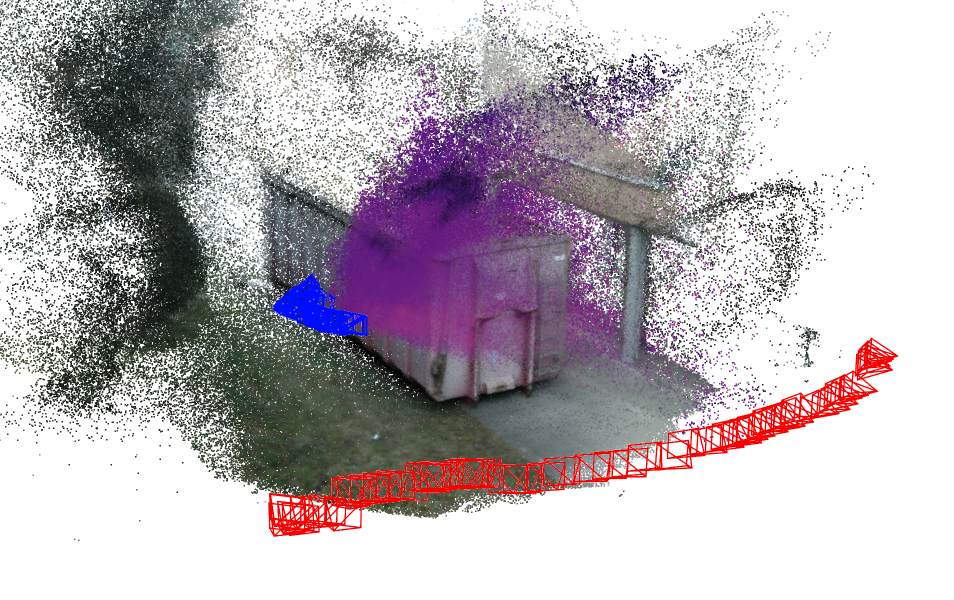}
  \end{subfigure}
  \begin{subfigure}[t]{0.19\textwidth}
    \includegraphics[width=\textwidth]{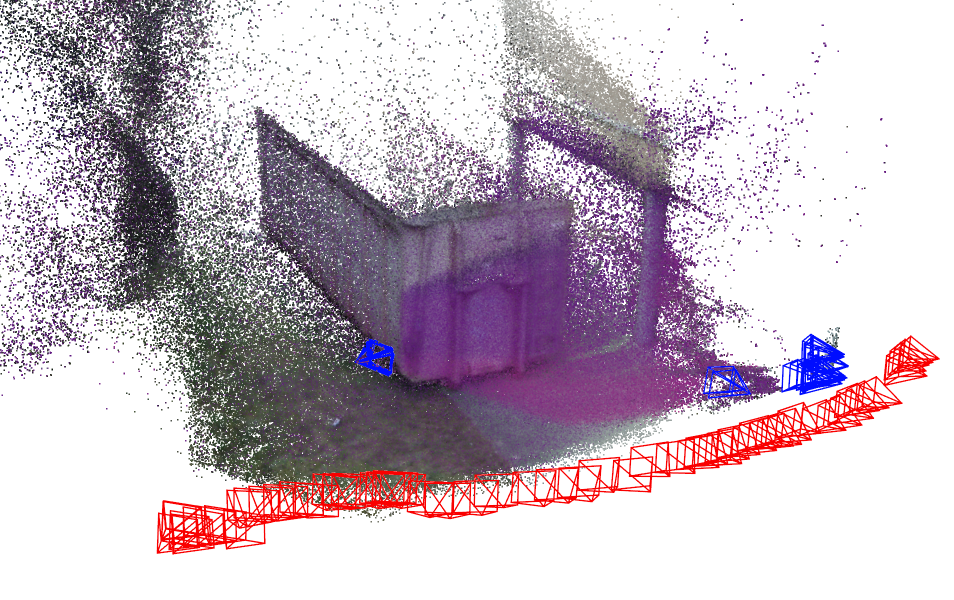}
  \end{subfigure}
  \begin{subfigure}[t]{0.19\textwidth}
    \includegraphics[width=\textwidth]{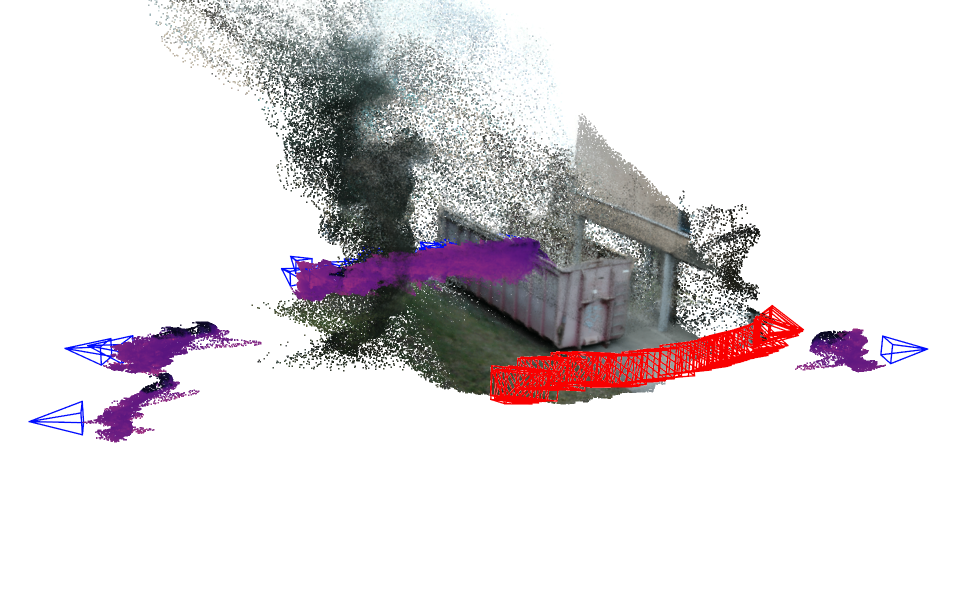}
  \end{subfigure}
  \begin{subfigure}[t]{0.19\textwidth}
    \includegraphics[width=\textwidth]{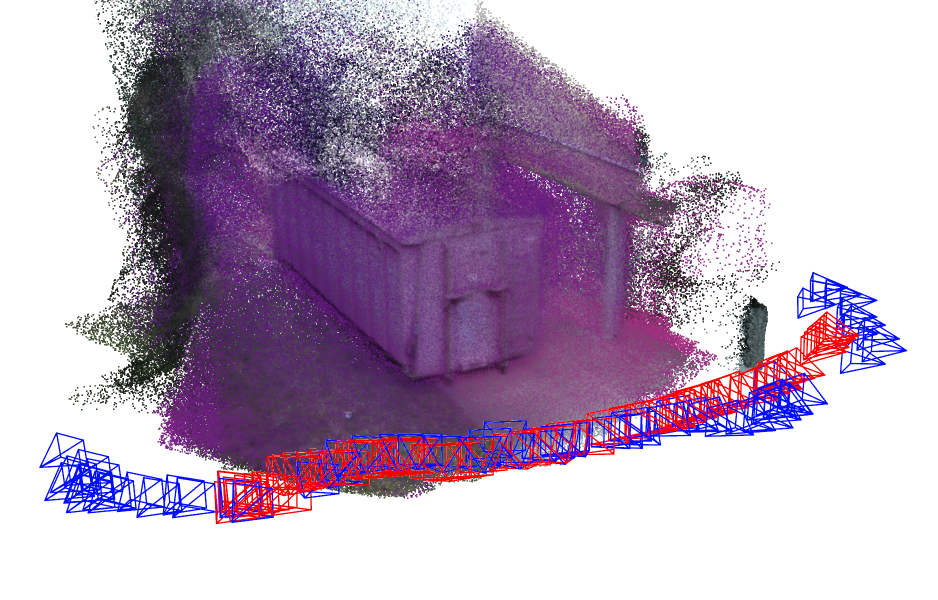}
  \end{subfigure}
  \begin{subfigure}[t]{0.19\textwidth}
    \includegraphics[width=\textwidth]{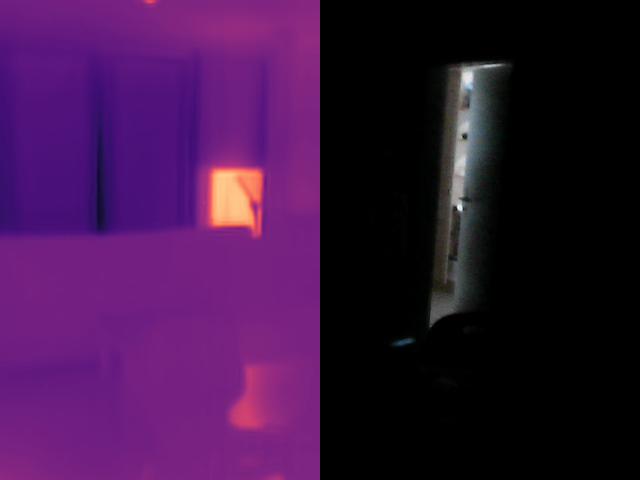}
    \caption*{thermal+RGB}
  \end{subfigure}
  \begin{subfigure}[t]{0.19\textwidth}
    \includegraphics[width=\textwidth]{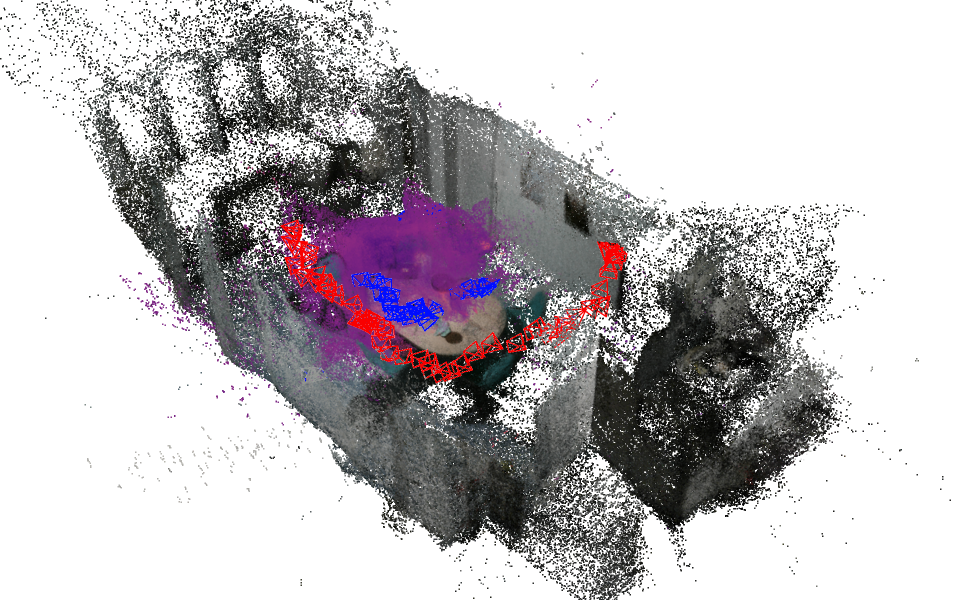}
    \caption*{MASt3R}
  \end{subfigure}
  \begin{subfigure}[t]{0.19\textwidth}
    \includegraphics[width=\textwidth]{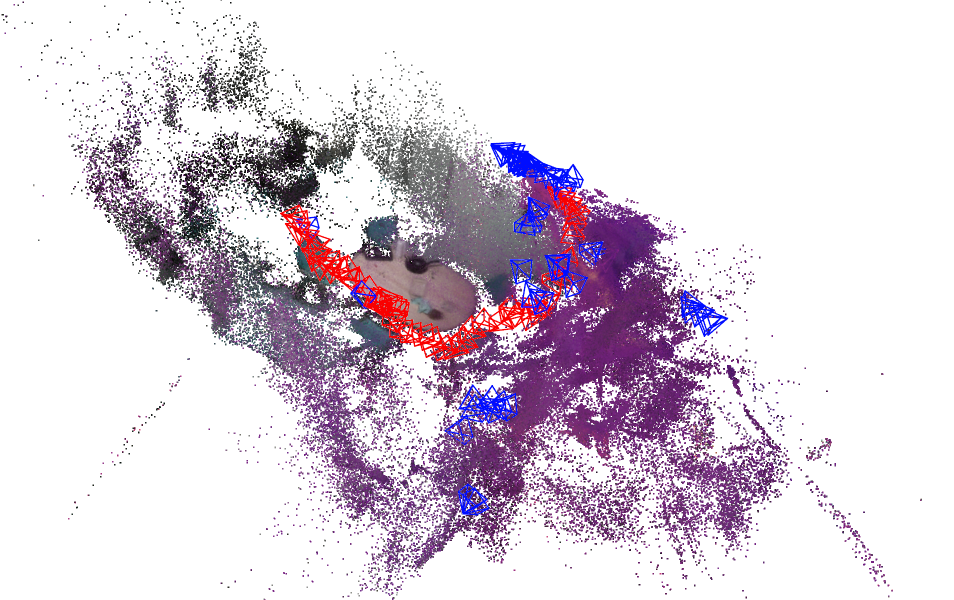}
    \caption*{MINIMA$_{ROMA}$}
  \end{subfigure}
  \begin{subfigure}[t]{0.19\textwidth}
    \includegraphics[width=\textwidth]{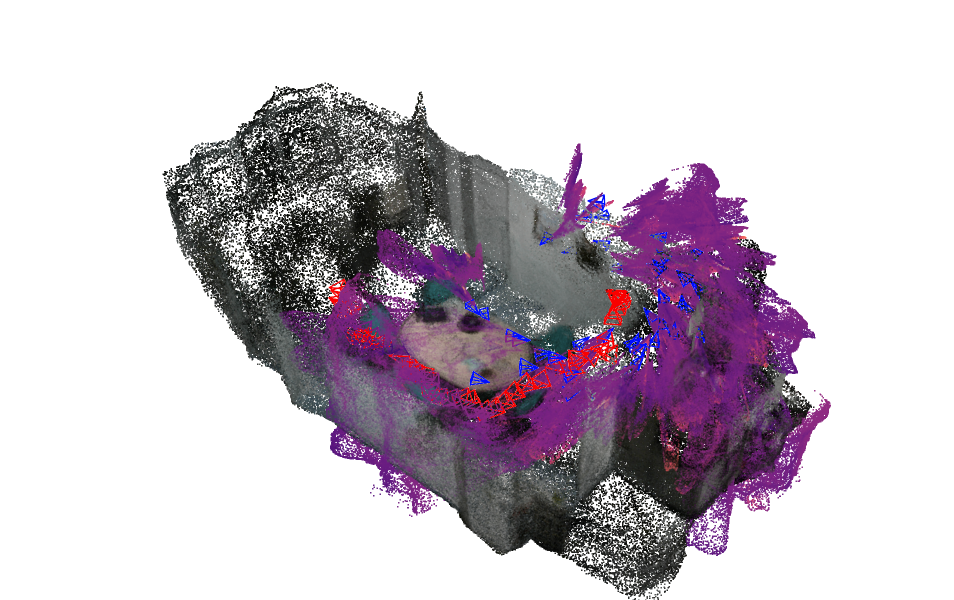}
    \caption*{VGGT}
  \end{subfigure}
  \begin{subfigure}[t]{0.19\textwidth}
    \includegraphics[width=\textwidth]{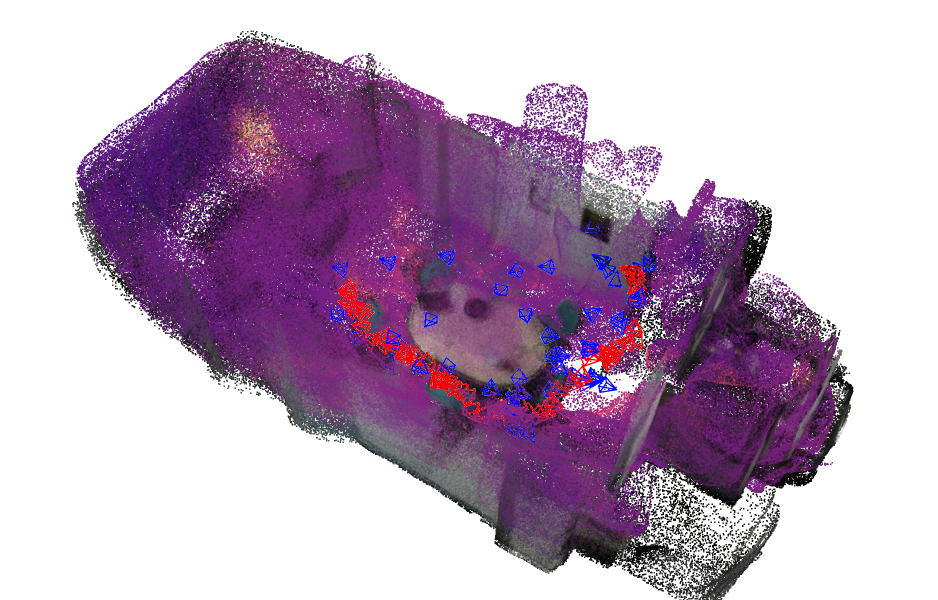}
    \caption*{\ours (ours)}
  \end{subfigure}
  \caption{
    Reconstructions from the SmokeSeer dataset (dense smoke, top) and our new dataset's scenes (lighting changes, middle and bottom rows).
    The first column shows cases where RGB images (right) are unreliable for localization, so thermal images (left) are used.
    Our method recovers the scene even with smoke (SmokeSeer) and aligns RGB and thermal camera poses in different lighting conditions (our dataset).
  }
  \vspace{-2mm}
  \label{fig:mm_smokeseer}
\end{figure}

%% file: images/graphs/thermal_ratio_figure.tex
\begin{figure*}[t]
  \centering
  \begin{subfigure}[b]{0.32\textwidth}
    \centering
    \resizebox{\linewidth}{!}{\input{images/graphs/thermal_ratio_camera_token_mAA.pgf}}
    \caption{AUC $\uparrow$}
    \label{fig:thermal_ratio_maa}
  \end{subfigure}
  \begin{subfigure}[b]{0.32\textwidth}
    \centering
    \resizebox{\linewidth}{!}{\input{images/graphs/thermal_ratio_camera_token_RRA.pgf}}
    \caption{RRA $\uparrow$}
    \label{fig:thermal_ratio_rra}
  \end{subfigure}
  \begin{subfigure}[b]{0.32\textwidth}
    \centering
    \resizebox{\linewidth}{!}{\input{images/graphs/thermal_ratio_camera_token_RTA.pgf}}
    \caption{RTA $\uparrow$}
    \label{fig:thermal_ratio_rta}
  \end{subfigure}

  \caption{
    The AUC, RRA, RTA (errors <$30^\circ$, $15^\circ$, $5^\circ$) across varying thermal-to-RGB image ratios. The filled area represents the boundary from $0.25-$ to $0.75-$quantiles estimated by bootstrapping scenes $2000$ times.
  }
  \label{fig:thermal_ratio_metrics}
  \vspace{-6mm}
\end{figure*}

%% file: images/graphs/thermal_ratio_camera_token_mAA.pgf
\begingroup%
\makeatletter%
\begin{pgfpicture}%
\pgfpathrectangle{\pgfpointorigin}{\pgfqpoint{7.450000in}{5.450000in}}%
\pgfusepath{use as bounding box, clip}%
\begin{pgfscope}%
\pgfsetbuttcap%
\pgfsetmiterjoin%
\definecolor{currentfill}{rgb}{1.000000,1.000000,1.000000}%
\pgfsetfillcolor{currentfill}%
\pgfsetlinewidth{0.000000pt}%
\definecolor{currentstroke}{rgb}{1.000000,1.000000,1.000000}%
\pgfsetstrokecolor{currentstroke}%
\pgfsetdash{}{0pt}%
\pgfpathmoveto{\pgfqpoint{0.000000in}{0.000000in}}%
\pgfpathlineto{\pgfqpoint{7.450000in}{0.000000in}}%
\pgfpathlineto{\pgfqpoint{7.450000in}{5.450000in}}%
\pgfpathlineto{\pgfqpoint{0.000000in}{5.450000in}}%
\pgfpathlineto{\pgfqpoint{0.000000in}{0.000000in}}%
\pgfpathclose%
\pgfusepath{fill}%
\end{pgfscope}%
\begin{pgfscope}%
\pgfsetbuttcap%
\pgfsetmiterjoin%
\definecolor{currentfill}{rgb}{1.000000,1.000000,1.000000}%
\pgfsetfillcolor{currentfill}%
\pgfsetlinewidth{0.000000pt}%
\definecolor{currentstroke}{rgb}{0.000000,0.000000,0.000000}%
\pgfsetstrokecolor{currentstroke}%
\pgfsetstrokeopacity{0.000000}%
\pgfsetdash{}{0pt}%
\pgfpathmoveto{\pgfqpoint{0.674954in}{0.862305in}}%
\pgfpathlineto{\pgfqpoint{7.147414in}{0.862305in}}%
\pgfpathlineto{\pgfqpoint{7.147414in}{5.231386in}}%
\pgfpathlineto{\pgfqpoint{0.674954in}{5.231386in}}%
\pgfpathlineto{\pgfqpoint{0.674954in}{0.862305in}}%
\pgfpathclose%
\pgfusepath{fill}%
\end{pgfscope}%
\begin{pgfscope}%
\pgfpathrectangle{\pgfqpoint{0.674954in}{0.862305in}}{\pgfqpoint{6.472460in}{4.369081in}}%
\pgfusepath{clip}%
\pgfsetbuttcap%
\pgfsetroundjoin%
\definecolor{currentfill}{rgb}{0.050980,0.415686,0.509804}%
\pgfsetfillcolor{currentfill}%
\pgfsetfillopacity{0.300000}%
\pgfsetlinewidth{1.003750pt}%
\definecolor{currentstroke}{rgb}{0.050980,0.415686,0.509804}%
\pgfsetstrokecolor{currentstroke}%
\pgfsetstrokeopacity{0.300000}%
\pgfsetdash{}{0pt}%
\pgfsys@defobject{currentmarker}{\pgfqpoint{0.674954in}{3.492852in}}{\pgfqpoint{7.147414in}{4.810039in}}{%
\pgfpathmoveto{\pgfqpoint{0.674954in}{4.810039in}}%
\pgfpathlineto{\pgfqpoint{0.674954in}{4.654861in}}%
\pgfpathlineto{\pgfqpoint{0.998577in}{4.534302in}}%
\pgfpathlineto{\pgfqpoint{2.293069in}{4.117275in}}%
\pgfpathlineto{\pgfqpoint{3.911184in}{3.760694in}}%
\pgfpathlineto{\pgfqpoint{5.529299in}{3.492852in}}%
\pgfpathlineto{\pgfqpoint{6.823791in}{3.624906in}}%
\pgfpathlineto{\pgfqpoint{7.147414in}{3.798553in}}%
\pgfpathlineto{\pgfqpoint{7.147414in}{4.232464in}}%
\pgfpathlineto{\pgfqpoint{7.147414in}{4.232464in}}%
\pgfpathlineto{\pgfqpoint{6.823791in}{4.081596in}}%
\pgfpathlineto{\pgfqpoint{5.529299in}{3.967789in}}%
\pgfpathlineto{\pgfqpoint{3.911184in}{4.158651in}}%
\pgfpathlineto{\pgfqpoint{2.293069in}{4.427697in}}%
\pgfpathlineto{\pgfqpoint{0.998577in}{4.711437in}}%
\pgfpathlineto{\pgfqpoint{0.674954in}{4.810039in}}%
\pgfpathlineto{\pgfqpoint{0.674954in}{4.810039in}}%
\pgfpathclose%
\pgfusepath{stroke,fill}%
}%
\begin{pgfscope}%
\pgfsys@transformshift{0.000000in}{0.000000in}%
\pgfsys@useobject{currentmarker}{}%
\end{pgfscope}%
\end{pgfscope}%
\begin{pgfscope}%
\pgfpathrectangle{\pgfqpoint{0.674954in}{0.862305in}}{\pgfqpoint{6.472460in}{4.369081in}}%
\pgfusepath{clip}%
\pgfsetbuttcap%
\pgfsetroundjoin%
\definecolor{currentfill}{rgb}{0.960784,0.462745,0.000000}%
\pgfsetfillcolor{currentfill}%
\pgfsetfillopacity{0.300000}%
\pgfsetlinewidth{1.003750pt}%
\definecolor{currentstroke}{rgb}{0.960784,0.462745,0.000000}%
\pgfsetstrokecolor{currentstroke}%
\pgfsetstrokeopacity{0.300000}%
\pgfsetdash{{3.700000pt}{1.600000pt}}{0.000000pt}%
\pgfpathmoveto{\pgfqpoint{0.674954in}{4.476823in}}%
\pgfpathlineto{\pgfqpoint{0.674954in}{4.231671in}}%
\pgfpathlineto{\pgfqpoint{0.998577in}{4.091768in}}%
\pgfpathlineto{\pgfqpoint{2.293069in}{3.608303in}}%
\pgfpathlineto{\pgfqpoint{3.911184in}{3.241529in}}%
\pgfpathlineto{\pgfqpoint{5.529299in}{3.001269in}}%
\pgfpathlineto{\pgfqpoint{6.823791in}{3.121166in}}%
\pgfpathlineto{\pgfqpoint{7.147414in}{3.241253in}}%
\pgfpathlineto{\pgfqpoint{7.147414in}{3.692521in}}%
\pgfpathlineto{\pgfqpoint{7.147414in}{3.692521in}}%
\pgfpathlineto{\pgfqpoint{6.823791in}{3.583145in}}%
\pgfpathlineto{\pgfqpoint{5.529299in}{3.475004in}}%
\pgfpathlineto{\pgfqpoint{3.911184in}{3.645706in}}%
\pgfpathlineto{\pgfqpoint{2.293069in}{3.947896in}}%
\pgfpathlineto{\pgfqpoint{0.998577in}{4.344005in}}%
\pgfpathlineto{\pgfqpoint{0.674954in}{4.476823in}}%
\pgfpathlineto{\pgfqpoint{0.674954in}{4.476823in}}%
\pgfpathclose%
\pgfusepath{stroke,fill}%
\end{pgfscope}%
\begin{pgfscope}%
\pgfpathrectangle{\pgfqpoint{0.674954in}{0.862305in}}{\pgfqpoint{6.472460in}{4.369081in}}%
\pgfusepath{clip}%
\pgfsetbuttcap%
\pgfsetroundjoin%
\definecolor{currentfill}{rgb}{0.219608,0.219608,0.219608}%
\pgfsetfillcolor{currentfill}%
\pgfsetfillopacity{0.300000}%
\pgfsetlinewidth{1.003750pt}%
\definecolor{currentstroke}{rgb}{0.219608,0.219608,0.219608}%
\pgfsetstrokecolor{currentstroke}%
\pgfsetstrokeopacity{0.300000}%
\pgfsetdash{{1.000000pt}{1.650000pt}}{0.000000pt}%
\pgfpathmoveto{\pgfqpoint{0.674954in}{3.518067in}}%
\pgfpathlineto{\pgfqpoint{0.674954in}{3.144958in}}%
\pgfpathlineto{\pgfqpoint{0.998577in}{3.000054in}}%
\pgfpathlineto{\pgfqpoint{2.293069in}{2.503163in}}%
\pgfpathlineto{\pgfqpoint{3.911184in}{2.181205in}}%
\pgfpathlineto{\pgfqpoint{5.529299in}{2.038001in}}%
\pgfpathlineto{\pgfqpoint{6.823791in}{2.122982in}}%
\pgfpathlineto{\pgfqpoint{7.147414in}{2.152298in}}%
\pgfpathlineto{\pgfqpoint{7.147414in}{2.555235in}}%
\pgfpathlineto{\pgfqpoint{7.147414in}{2.555235in}}%
\pgfpathlineto{\pgfqpoint{6.823791in}{2.502106in}}%
\pgfpathlineto{\pgfqpoint{5.529299in}{2.410682in}}%
\pgfpathlineto{\pgfqpoint{3.911184in}{2.536187in}}%
\pgfpathlineto{\pgfqpoint{2.293069in}{2.844960in}}%
\pgfpathlineto{\pgfqpoint{0.998577in}{3.348935in}}%
\pgfpathlineto{\pgfqpoint{0.674954in}{3.518067in}}%
\pgfpathlineto{\pgfqpoint{0.674954in}{3.518067in}}%
\pgfpathclose%
\pgfusepath{stroke,fill}%
\end{pgfscope}%
\begin{pgfscope}%
\pgfpathrectangle{\pgfqpoint{0.674954in}{0.862305in}}{\pgfqpoint{6.472460in}{4.369081in}}%
\pgfusepath{clip}%
\pgfsetbuttcap%
\pgfsetroundjoin%
\pgfsetlinewidth{2.007500pt}%
\definecolor{currentstroke}{rgb}{0.501961,0.501961,0.501961}%
\pgfsetstrokecolor{currentstroke}%
\pgfsetstrokeopacity{0.300000}%
\pgfsetdash{{7.400000pt}{3.200000pt}}{0.000000pt}%
\pgfpathmoveto{\pgfqpoint{0.674954in}{0.862305in}}%
\pgfpathlineto{\pgfqpoint{0.674954in}{5.231386in}}%
\pgfusepath{stroke}%
\end{pgfscope}%
\begin{pgfscope}%
\pgfsetbuttcap%
\pgfsetroundjoin%
\definecolor{currentfill}{rgb}{0.000000,0.000000,0.000000}%
\pgfsetfillcolor{currentfill}%
\pgfsetlinewidth{0.803000pt}%
\definecolor{currentstroke}{rgb}{0.000000,0.000000,0.000000}%
\pgfsetstrokecolor{currentstroke}%
\pgfsetdash{}{0pt}%
\pgfsys@defobject{currentmarker}{\pgfqpoint{0.000000in}{-0.048611in}}{\pgfqpoint{0.000000in}{0.000000in}}{%
\pgfpathmoveto{\pgfqpoint{0.000000in}{0.000000in}}%
\pgfpathlineto{\pgfqpoint{0.000000in}{-0.048611in}}%
\pgfusepath{stroke,fill}%
}%
\begin{pgfscope}%
\pgfsys@transformshift{0.674954in}{0.862305in}%
\pgfsys@useobject{currentmarker}{}%
\end{pgfscope}%
\end{pgfscope}%
\begin{pgfscope}%
\definecolor{textcolor}{rgb}{0.000000,0.000000,0.000000}%
\pgfsetstrokecolor{textcolor}%
\pgfsetfillcolor{textcolor}%
\pgftext[x=0.674954in,y=0.765082in,,top]{\color{textcolor}{\rmfamily\fontsize{25.000000}{30.000000}\selectfont\catcode`\^=\active\def^{\ifmmode\sp\else\^{}\fi}\catcode`\%=\active\def
\end{pgfscope}%
\begin{pgfscope}%
\pgfpathrectangle{\pgfqpoint{0.674954in}{0.862305in}}{\pgfqpoint{6.472460in}{4.369081in}}%
\pgfusepath{clip}%
\pgfsetbuttcap%
\pgfsetroundjoin%
\pgfsetlinewidth{2.007500pt}%
\definecolor{currentstroke}{rgb}{0.501961,0.501961,0.501961}%
\pgfsetstrokecolor{currentstroke}%
\pgfsetstrokeopacity{0.300000}%
\pgfsetdash{{7.400000pt}{3.200000pt}}{0.000000pt}%
\pgfpathmoveto{\pgfqpoint{1.969446in}{0.862305in}}%
\pgfpathlineto{\pgfqpoint{1.969446in}{5.231386in}}%
\pgfusepath{stroke}%
\end{pgfscope}%
\begin{pgfscope}%
\pgfsetbuttcap%
\pgfsetroundjoin%
\definecolor{currentfill}{rgb}{0.000000,0.000000,0.000000}%
\pgfsetfillcolor{currentfill}%
\pgfsetlinewidth{0.803000pt}%
\definecolor{currentstroke}{rgb}{0.000000,0.000000,0.000000}%
\pgfsetstrokecolor{currentstroke}%
\pgfsetdash{}{0pt}%
\pgfsys@defobject{currentmarker}{\pgfqpoint{0.000000in}{-0.048611in}}{\pgfqpoint{0.000000in}{0.000000in}}{%
\pgfpathmoveto{\pgfqpoint{0.000000in}{0.000000in}}%
\pgfpathlineto{\pgfqpoint{0.000000in}{-0.048611in}}%
\pgfusepath{stroke,fill}%
}%
\begin{pgfscope}%
\pgfsys@transformshift{1.969446in}{0.862305in}%
\pgfsys@useobject{currentmarker}{}%
\end{pgfscope}%
\end{pgfscope}%
\begin{pgfscope}%
\definecolor{textcolor}{rgb}{0.000000,0.000000,0.000000}%
\pgfsetstrokecolor{textcolor}%
\pgfsetfillcolor{textcolor}%
\pgftext[x=1.969446in,y=0.765082in,,top]{\color{textcolor}{\rmfamily\fontsize{25.000000}{30.000000}\selectfont\catcode`\^=\active\def^{\ifmmode\sp\else\^{}\fi}\catcode`\%=\active\def
\end{pgfscope}%
\begin{pgfscope}%
\pgfpathrectangle{\pgfqpoint{0.674954in}{0.862305in}}{\pgfqpoint{6.472460in}{4.369081in}}%
\pgfusepath{clip}%
\pgfsetbuttcap%
\pgfsetroundjoin%
\pgfsetlinewidth{2.007500pt}%
\definecolor{currentstroke}{rgb}{0.501961,0.501961,0.501961}%
\pgfsetstrokecolor{currentstroke}%
\pgfsetstrokeopacity{0.300000}%
\pgfsetdash{{7.400000pt}{3.200000pt}}{0.000000pt}%
\pgfpathmoveto{\pgfqpoint{3.263938in}{0.862305in}}%
\pgfpathlineto{\pgfqpoint{3.263938in}{5.231386in}}%
\pgfusepath{stroke}%
\end{pgfscope}%
\begin{pgfscope}%
\pgfsetbuttcap%
\pgfsetroundjoin%
\definecolor{currentfill}{rgb}{0.000000,0.000000,0.000000}%
\pgfsetfillcolor{currentfill}%
\pgfsetlinewidth{0.803000pt}%
\definecolor{currentstroke}{rgb}{0.000000,0.000000,0.000000}%
\pgfsetstrokecolor{currentstroke}%
\pgfsetdash{}{0pt}%
\pgfsys@defobject{currentmarker}{\pgfqpoint{0.000000in}{-0.048611in}}{\pgfqpoint{0.000000in}{0.000000in}}{%
\pgfpathmoveto{\pgfqpoint{0.000000in}{0.000000in}}%
\pgfpathlineto{\pgfqpoint{0.000000in}{-0.048611in}}%
\pgfusepath{stroke,fill}%
}%
\begin{pgfscope}%
\pgfsys@transformshift{3.263938in}{0.862305in}%
\pgfsys@useobject{currentmarker}{}%
\end{pgfscope}%
\end{pgfscope}%
\begin{pgfscope}%
\definecolor{textcolor}{rgb}{0.000000,0.000000,0.000000}%
\pgfsetstrokecolor{textcolor}%
\pgfsetfillcolor{textcolor}%
\pgftext[x=3.263938in,y=0.765082in,,top]{\color{textcolor}{\rmfamily\fontsize{25.000000}{30.000000}\selectfont\catcode`\^=\active\def^{\ifmmode\sp\else\^{}\fi}\catcode`\%=\active\def
\end{pgfscope}%
\begin{pgfscope}%
\pgfpathrectangle{\pgfqpoint{0.674954in}{0.862305in}}{\pgfqpoint{6.472460in}{4.369081in}}%
\pgfusepath{clip}%
\pgfsetbuttcap%
\pgfsetroundjoin%
\pgfsetlinewidth{2.007500pt}%
\definecolor{currentstroke}{rgb}{0.501961,0.501961,0.501961}%
\pgfsetstrokecolor{currentstroke}%
\pgfsetstrokeopacity{0.300000}%
\pgfsetdash{{7.400000pt}{3.200000pt}}{0.000000pt}%
\pgfpathmoveto{\pgfqpoint{4.558430in}{0.862305in}}%
\pgfpathlineto{\pgfqpoint{4.558430in}{5.231386in}}%
\pgfusepath{stroke}%
\end{pgfscope}%
\begin{pgfscope}%
\pgfsetbuttcap%
\pgfsetroundjoin%
\definecolor{currentfill}{rgb}{0.000000,0.000000,0.000000}%
\pgfsetfillcolor{currentfill}%
\pgfsetlinewidth{0.803000pt}%
\definecolor{currentstroke}{rgb}{0.000000,0.000000,0.000000}%
\pgfsetstrokecolor{currentstroke}%
\pgfsetdash{}{0pt}%
\pgfsys@defobject{currentmarker}{\pgfqpoint{0.000000in}{-0.048611in}}{\pgfqpoint{0.000000in}{0.000000in}}{%
\pgfpathmoveto{\pgfqpoint{0.000000in}{0.000000in}}%
\pgfpathlineto{\pgfqpoint{0.000000in}{-0.048611in}}%
\pgfusepath{stroke,fill}%
}%
\begin{pgfscope}%
\pgfsys@transformshift{4.558430in}{0.862305in}%
\pgfsys@useobject{currentmarker}{}%
\end{pgfscope}%
\end{pgfscope}%
\begin{pgfscope}%
\definecolor{textcolor}{rgb}{0.000000,0.000000,0.000000}%
\pgfsetstrokecolor{textcolor}%
\pgfsetfillcolor{textcolor}%
\pgftext[x=4.558430in,y=0.765082in,,top]{\color{textcolor}{\rmfamily\fontsize{25.000000}{30.000000}\selectfont\catcode`\^=\active\def^{\ifmmode\sp\else\^{}\fi}\catcode`\%=\active\def
\end{pgfscope}%
\begin{pgfscope}%
\pgfpathrectangle{\pgfqpoint{0.674954in}{0.862305in}}{\pgfqpoint{6.472460in}{4.369081in}}%
\pgfusepath{clip}%
\pgfsetbuttcap%
\pgfsetroundjoin%
\pgfsetlinewidth{2.007500pt}%
\definecolor{currentstroke}{rgb}{0.501961,0.501961,0.501961}%
\pgfsetstrokecolor{currentstroke}%
\pgfsetstrokeopacity{0.300000}%
\pgfsetdash{{7.400000pt}{3.200000pt}}{0.000000pt}%
\pgfpathmoveto{\pgfqpoint{5.852922in}{0.862305in}}%
\pgfpathlineto{\pgfqpoint{5.852922in}{5.231386in}}%
\pgfusepath{stroke}%
\end{pgfscope}%
\begin{pgfscope}%
\pgfsetbuttcap%
\pgfsetroundjoin%
\definecolor{currentfill}{rgb}{0.000000,0.000000,0.000000}%
\pgfsetfillcolor{currentfill}%
\pgfsetlinewidth{0.803000pt}%
\definecolor{currentstroke}{rgb}{0.000000,0.000000,0.000000}%
\pgfsetstrokecolor{currentstroke}%
\pgfsetdash{}{0pt}%
\pgfsys@defobject{currentmarker}{\pgfqpoint{0.000000in}{-0.048611in}}{\pgfqpoint{0.000000in}{0.000000in}}{%
\pgfpathmoveto{\pgfqpoint{0.000000in}{0.000000in}}%
\pgfpathlineto{\pgfqpoint{0.000000in}{-0.048611in}}%
\pgfusepath{stroke,fill}%
}%
\begin{pgfscope}%
\pgfsys@transformshift{5.852922in}{0.862305in}%
\pgfsys@useobject{currentmarker}{}%
\end{pgfscope}%
\end{pgfscope}%
\begin{pgfscope}%
\definecolor{textcolor}{rgb}{0.000000,0.000000,0.000000}%
\pgfsetstrokecolor{textcolor}%
\pgfsetfillcolor{textcolor}%
\pgftext[x=5.852922in,y=0.765082in,,top]{\color{textcolor}{\rmfamily\fontsize{25.000000}{30.000000}\selectfont\catcode`\^=\active\def^{\ifmmode\sp\else\^{}\fi}\catcode`\%=\active\def
\end{pgfscope}%
\begin{pgfscope}%
\pgfpathrectangle{\pgfqpoint{0.674954in}{0.862305in}}{\pgfqpoint{6.472460in}{4.369081in}}%
\pgfusepath{clip}%
\pgfsetbuttcap%
\pgfsetroundjoin%
\pgfsetlinewidth{2.007500pt}%
\definecolor{currentstroke}{rgb}{0.501961,0.501961,0.501961}%
\pgfsetstrokecolor{currentstroke}%
\pgfsetstrokeopacity{0.300000}%
\pgfsetdash{{7.400000pt}{3.200000pt}}{0.000000pt}%
\pgfpathmoveto{\pgfqpoint{7.147414in}{0.862305in}}%
\pgfpathlineto{\pgfqpoint{7.147414in}{5.231386in}}%
\pgfusepath{stroke}%
\end{pgfscope}%
\begin{pgfscope}%
\pgfsetbuttcap%
\pgfsetroundjoin%
\definecolor{currentfill}{rgb}{0.000000,0.000000,0.000000}%
\pgfsetfillcolor{currentfill}%
\pgfsetlinewidth{0.803000pt}%
\definecolor{currentstroke}{rgb}{0.000000,0.000000,0.000000}%
\pgfsetstrokecolor{currentstroke}%
\pgfsetdash{}{0pt}%
\pgfsys@defobject{currentmarker}{\pgfqpoint{0.000000in}{-0.048611in}}{\pgfqpoint{0.000000in}{0.000000in}}{%
\pgfpathmoveto{\pgfqpoint{0.000000in}{0.000000in}}%
\pgfpathlineto{\pgfqpoint{0.000000in}{-0.048611in}}%
\pgfusepath{stroke,fill}%
}%
\begin{pgfscope}%
\pgfsys@transformshift{7.147414in}{0.862305in}%
\pgfsys@useobject{currentmarker}{}%
\end{pgfscope}%
\end{pgfscope}%
\begin{pgfscope}%
\definecolor{textcolor}{rgb}{0.000000,0.000000,0.000000}%
\pgfsetstrokecolor{textcolor}%
\pgfsetfillcolor{textcolor}%
\pgftext[x=7.147414in,y=0.765082in,,top]{\color{textcolor}{\rmfamily\fontsize{25.000000}{30.000000}\selectfont\catcode`\^=\active\def^{\ifmmode\sp\else\^{}\fi}\catcode`\%=\active\def
\end{pgfscope}%
\begin{pgfscope}%
\definecolor{textcolor}{rgb}{0.000000,0.000000,0.000000}%
\pgfsetstrokecolor{textcolor}%
\pgfsetfillcolor{textcolor}%
\pgftext[x=3.911184in,y=0.404763in,,top]{\color{textcolor}{\rmfamily\fontsize{25.000000}{30.000000}\selectfont\catcode`\^=\active\def^{\ifmmode\sp\else\^{}\fi}\catcode`\%=\active\def
\end{pgfscope}%
\begin{pgfscope}%
\pgfpathrectangle{\pgfqpoint{0.674954in}{0.862305in}}{\pgfqpoint{6.472460in}{4.369081in}}%
\pgfusepath{clip}%
\pgfsetbuttcap%
\pgfsetroundjoin%
\pgfsetlinewidth{2.007500pt}%
\definecolor{currentstroke}{rgb}{0.501961,0.501961,0.501961}%
\pgfsetstrokecolor{currentstroke}%
\pgfsetstrokeopacity{0.300000}%
\pgfsetdash{{7.400000pt}{3.200000pt}}{0.000000pt}%
\pgfpathmoveto{\pgfqpoint{0.674954in}{1.954575in}}%
\pgfpathlineto{\pgfqpoint{7.147414in}{1.954575in}}%
\pgfusepath{stroke}%
\end{pgfscope}%
\begin{pgfscope}%
\pgfsetbuttcap%
\pgfsetroundjoin%
\definecolor{currentfill}{rgb}{0.000000,0.000000,0.000000}%
\pgfsetfillcolor{currentfill}%
\pgfsetlinewidth{0.803000pt}%
\definecolor{currentstroke}{rgb}{0.000000,0.000000,0.000000}%
\pgfsetstrokecolor{currentstroke}%
\pgfsetdash{}{0pt}%
\pgfsys@defobject{currentmarker}{\pgfqpoint{-0.048611in}{0.000000in}}{\pgfqpoint{-0.000000in}{0.000000in}}{%
\pgfpathmoveto{\pgfqpoint{-0.000000in}{0.000000in}}%
\pgfpathlineto{\pgfqpoint{-0.048611in}{0.000000in}}%
\pgfusepath{stroke,fill}%
}%
\begin{pgfscope}%
\pgfsys@transformshift{0.674954in}{1.954575in}%
\pgfsys@useobject{currentmarker}{}%
\end{pgfscope}%
\end{pgfscope}%
\begin{pgfscope}%
\definecolor{textcolor}{rgb}{0.000000,0.000000,0.000000}%
\pgfsetstrokecolor{textcolor}%
\pgfsetfillcolor{textcolor}%
\pgftext[x=0.259244in, y=1.835961in, left, base]{\color{textcolor}{\rmfamily\fontsize{25.000000}{30.000000}\selectfont\catcode`\^=\active\def^{\ifmmode\sp\else\^{}\fi}\catcode`\%=\active\def
\end{pgfscope}%
\begin{pgfscope}%
\pgfpathrectangle{\pgfqpoint{0.674954in}{0.862305in}}{\pgfqpoint{6.472460in}{4.369081in}}%
\pgfusepath{clip}%
\pgfsetbuttcap%
\pgfsetroundjoin%
\pgfsetlinewidth{2.007500pt}%
\definecolor{currentstroke}{rgb}{0.501961,0.501961,0.501961}%
\pgfsetstrokecolor{currentstroke}%
\pgfsetstrokeopacity{0.300000}%
\pgfsetdash{{7.400000pt}{3.200000pt}}{0.000000pt}%
\pgfpathmoveto{\pgfqpoint{0.674954in}{3.046845in}}%
\pgfpathlineto{\pgfqpoint{7.147414in}{3.046845in}}%
\pgfusepath{stroke}%
\end{pgfscope}%
\begin{pgfscope}%
\pgfsetbuttcap%
\pgfsetroundjoin%
\definecolor{currentfill}{rgb}{0.000000,0.000000,0.000000}%
\pgfsetfillcolor{currentfill}%
\pgfsetlinewidth{0.803000pt}%
\definecolor{currentstroke}{rgb}{0.000000,0.000000,0.000000}%
\pgfsetstrokecolor{currentstroke}%
\pgfsetdash{}{0pt}%
\pgfsys@defobject{currentmarker}{\pgfqpoint{-0.048611in}{0.000000in}}{\pgfqpoint{-0.000000in}{0.000000in}}{%
\pgfpathmoveto{\pgfqpoint{-0.000000in}{0.000000in}}%
\pgfpathlineto{\pgfqpoint{-0.048611in}{0.000000in}}%
\pgfusepath{stroke,fill}%
}%
\begin{pgfscope}%
\pgfsys@transformshift{0.674954in}{3.046845in}%
\pgfsys@useobject{currentmarker}{}%
\end{pgfscope}%
\end{pgfscope}%
\begin{pgfscope}%
\definecolor{textcolor}{rgb}{0.000000,0.000000,0.000000}%
\pgfsetstrokecolor{textcolor}%
\pgfsetfillcolor{textcolor}%
\pgftext[x=0.259244in, y=2.928231in, left, base]{\color{textcolor}{\rmfamily\fontsize{25.000000}{30.000000}\selectfont\catcode`\^=\active\def^{\ifmmode\sp\else\^{}\fi}\catcode`\%=\active\def
\end{pgfscope}%
\begin{pgfscope}%
\pgfpathrectangle{\pgfqpoint{0.674954in}{0.862305in}}{\pgfqpoint{6.472460in}{4.369081in}}%
\pgfusepath{clip}%
\pgfsetbuttcap%
\pgfsetroundjoin%
\pgfsetlinewidth{2.007500pt}%
\definecolor{currentstroke}{rgb}{0.501961,0.501961,0.501961}%
\pgfsetstrokecolor{currentstroke}%
\pgfsetstrokeopacity{0.300000}%
\pgfsetdash{{7.400000pt}{3.200000pt}}{0.000000pt}%
\pgfpathmoveto{\pgfqpoint{0.674954in}{4.139116in}}%
\pgfpathlineto{\pgfqpoint{7.147414in}{4.139116in}}%
\pgfusepath{stroke}%
\end{pgfscope}%
\begin{pgfscope}%
\pgfsetbuttcap%
\pgfsetroundjoin%
\definecolor{currentfill}{rgb}{0.000000,0.000000,0.000000}%
\pgfsetfillcolor{currentfill}%
\pgfsetlinewidth{0.803000pt}%
\definecolor{currentstroke}{rgb}{0.000000,0.000000,0.000000}%
\pgfsetstrokecolor{currentstroke}%
\pgfsetdash{}{0pt}%
\pgfsys@defobject{currentmarker}{\pgfqpoint{-0.048611in}{0.000000in}}{\pgfqpoint{-0.000000in}{0.000000in}}{%
\pgfpathmoveto{\pgfqpoint{-0.000000in}{0.000000in}}%
\pgfpathlineto{\pgfqpoint{-0.048611in}{0.000000in}}%
\pgfusepath{stroke,fill}%
}%
\begin{pgfscope}%
\pgfsys@transformshift{0.674954in}{4.139116in}%
\pgfsys@useobject{currentmarker}{}%
\end{pgfscope}%
\end{pgfscope}%
\begin{pgfscope}%
\definecolor{textcolor}{rgb}{0.000000,0.000000,0.000000}%
\pgfsetstrokecolor{textcolor}%
\pgfsetfillcolor{textcolor}%
\pgftext[x=0.259244in, y=4.020501in, left, base]{\color{textcolor}{\rmfamily\fontsize{25.000000}{30.000000}\selectfont\catcode`\^=\active\def^{\ifmmode\sp\else\^{}\fi}\catcode`\%=\active\def
\end{pgfscope}%
\begin{pgfscope}%
\pgfpathrectangle{\pgfqpoint{0.674954in}{0.862305in}}{\pgfqpoint{6.472460in}{4.369081in}}%
\pgfusepath{clip}%
\pgfsetbuttcap%
\pgfsetroundjoin%
\pgfsetlinewidth{2.007500pt}%
\definecolor{currentstroke}{rgb}{0.501961,0.501961,0.501961}%
\pgfsetstrokecolor{currentstroke}%
\pgfsetstrokeopacity{0.300000}%
\pgfsetdash{{7.400000pt}{3.200000pt}}{0.000000pt}%
\pgfpathmoveto{\pgfqpoint{0.674954in}{5.231386in}}%
\pgfpathlineto{\pgfqpoint{7.147414in}{5.231386in}}%
\pgfusepath{stroke}%
\end{pgfscope}%
\begin{pgfscope}%
\pgfsetbuttcap%
\pgfsetroundjoin%
\definecolor{currentfill}{rgb}{0.000000,0.000000,0.000000}%
\pgfsetfillcolor{currentfill}%
\pgfsetlinewidth{0.803000pt}%
\definecolor{currentstroke}{rgb}{0.000000,0.000000,0.000000}%
\pgfsetstrokecolor{currentstroke}%
\pgfsetdash{}{0pt}%
\pgfsys@defobject{currentmarker}{\pgfqpoint{-0.048611in}{0.000000in}}{\pgfqpoint{-0.000000in}{0.000000in}}{%
\pgfpathmoveto{\pgfqpoint{-0.000000in}{0.000000in}}%
\pgfpathlineto{\pgfqpoint{-0.048611in}{0.000000in}}%
\pgfusepath{stroke,fill}%
}%
\begin{pgfscope}%
\pgfsys@transformshift{0.674954in}{5.231386in}%
\pgfsys@useobject{currentmarker}{}%
\end{pgfscope}%
\end{pgfscope}%
\begin{pgfscope}%
\definecolor{textcolor}{rgb}{0.000000,0.000000,0.000000}%
\pgfsetstrokecolor{textcolor}%
\pgfsetfillcolor{textcolor}%
\pgftext[x=0.100000in, y=5.112772in, left, base]{\color{textcolor}{\rmfamily\fontsize{25.000000}{30.000000}\selectfont\catcode`\^=\active\def^{\ifmmode\sp\else\^{}\fi}\catcode`\%=\active\def
\end{pgfscope}%
\begin{pgfscope}%
\pgfpathrectangle{\pgfqpoint{0.674954in}{0.862305in}}{\pgfqpoint{6.472460in}{4.369081in}}%
\pgfusepath{clip}%
\pgfsetrectcap%
\pgfsetroundjoin%
\pgfsetlinewidth{2.509375pt}%
\definecolor{currentstroke}{rgb}{0.050980,0.415686,0.509804}%
\pgfsetstrokecolor{currentstroke}%
\pgfsetdash{}{0pt}%
\pgfpathmoveto{\pgfqpoint{0.674954in}{4.733824in}}%
\pgfpathlineto{\pgfqpoint{0.998577in}{4.626702in}}%
\pgfpathlineto{\pgfqpoint{2.293069in}{4.277117in}}%
\pgfpathlineto{\pgfqpoint{3.911184in}{3.967874in}}%
\pgfpathlineto{\pgfqpoint{5.529299in}{3.743739in}}%
\pgfpathlineto{\pgfqpoint{6.823791in}{3.864597in}}%
\pgfpathlineto{\pgfqpoint{7.147414in}{4.022726in}}%
\pgfusepath{stroke}%
\end{pgfscope}%
\begin{pgfscope}%
\pgfpathrectangle{\pgfqpoint{0.674954in}{0.862305in}}{\pgfqpoint{6.472460in}{4.369081in}}%
\pgfusepath{clip}%
\pgfsetbuttcap%
\pgfsetroundjoin%
\definecolor{currentfill}{rgb}{0.050980,0.415686,0.509804}%
\pgfsetfillcolor{currentfill}%
\pgfsetlinewidth{1.003750pt}%
\definecolor{currentstroke}{rgb}{0.050980,0.415686,0.509804}%
\pgfsetstrokecolor{currentstroke}%
\pgfsetdash{}{0pt}%
\pgfsys@defobject{currentmarker}{\pgfqpoint{-0.055556in}{-0.055556in}}{\pgfqpoint{0.055556in}{0.055556in}}{%
\pgfpathmoveto{\pgfqpoint{0.000000in}{-0.055556in}}%
\pgfpathcurveto{\pgfqpoint{0.014734in}{-0.055556in}}{\pgfqpoint{0.028866in}{-0.049702in}}{\pgfqpoint{0.039284in}{-0.039284in}}%
\pgfpathcurveto{\pgfqpoint{0.049702in}{-0.028866in}}{\pgfqpoint{0.055556in}{-0.014734in}}{\pgfqpoint{0.055556in}{0.000000in}}%
\pgfpathcurveto{\pgfqpoint{0.055556in}{0.014734in}}{\pgfqpoint{0.049702in}{0.028866in}}{\pgfqpoint{0.039284in}{0.039284in}}%
\pgfpathcurveto{\pgfqpoint{0.028866in}{0.049702in}}{\pgfqpoint{0.014734in}{0.055556in}}{\pgfqpoint{0.000000in}{0.055556in}}%
\pgfpathcurveto{\pgfqpoint{-0.014734in}{0.055556in}}{\pgfqpoint{-0.028866in}{0.049702in}}{\pgfqpoint{-0.039284in}{0.039284in}}%
\pgfpathcurveto{\pgfqpoint{-0.049702in}{0.028866in}}{\pgfqpoint{-0.055556in}{0.014734in}}{\pgfqpoint{-0.055556in}{0.000000in}}%
\pgfpathcurveto{\pgfqpoint{-0.055556in}{-0.014734in}}{\pgfqpoint{-0.049702in}{-0.028866in}}{\pgfqpoint{-0.039284in}{-0.039284in}}%
\pgfpathcurveto{\pgfqpoint{-0.028866in}{-0.049702in}}{\pgfqpoint{-0.014734in}{-0.055556in}}{\pgfqpoint{0.000000in}{-0.055556in}}%
\pgfpathlineto{\pgfqpoint{0.000000in}{-0.055556in}}%
\pgfpathclose%
\pgfusepath{stroke,fill}%
}%
\begin{pgfscope}%
\pgfsys@transformshift{0.674954in}{4.733824in}%
\pgfsys@useobject{currentmarker}{}%
\end{pgfscope}%
\begin{pgfscope}%
\pgfsys@transformshift{0.998577in}{4.626702in}%
\pgfsys@useobject{currentmarker}{}%
\end{pgfscope}%
\begin{pgfscope}%
\pgfsys@transformshift{2.293069in}{4.277117in}%
\pgfsys@useobject{currentmarker}{}%
\end{pgfscope}%
\begin{pgfscope}%
\pgfsys@transformshift{3.911184in}{3.967874in}%
\pgfsys@useobject{currentmarker}{}%
\end{pgfscope}%
\begin{pgfscope}%
\pgfsys@transformshift{5.529299in}{3.743739in}%
\pgfsys@useobject{currentmarker}{}%
\end{pgfscope}%
\begin{pgfscope}%
\pgfsys@transformshift{6.823791in}{3.864597in}%
\pgfsys@useobject{currentmarker}{}%
\end{pgfscope}%
\begin{pgfscope}%
\pgfsys@transformshift{7.147414in}{4.022726in}%
\pgfsys@useobject{currentmarker}{}%
\end{pgfscope}%
\end{pgfscope}%
\begin{pgfscope}%
\pgfpathrectangle{\pgfqpoint{0.674954in}{0.862305in}}{\pgfqpoint{6.472460in}{4.369081in}}%
\pgfusepath{clip}%
\pgfsetbuttcap%
\pgfsetroundjoin%
\pgfsetlinewidth{2.509375pt}%
\definecolor{currentstroke}{rgb}{0.960784,0.462745,0.000000}%
\pgfsetstrokecolor{currentstroke}%
\pgfsetdash{{9.250000pt}{4.000000pt}}{0.000000pt}%
\pgfpathmoveto{\pgfqpoint{0.674954in}{4.357751in}}%
\pgfpathlineto{\pgfqpoint{0.998577in}{4.218213in}}%
\pgfpathlineto{\pgfqpoint{2.293069in}{3.782202in}}%
\pgfpathlineto{\pgfqpoint{3.911184in}{3.443220in}}%
\pgfpathlineto{\pgfqpoint{5.529299in}{3.250093in}}%
\pgfpathlineto{\pgfqpoint{6.823791in}{3.364588in}}%
\pgfpathlineto{\pgfqpoint{7.147414in}{3.474857in}}%
\pgfusepath{stroke}%
\end{pgfscope}%
\begin{pgfscope}%
\pgfpathrectangle{\pgfqpoint{0.674954in}{0.862305in}}{\pgfqpoint{6.472460in}{4.369081in}}%
\pgfusepath{clip}%
\pgfsetbuttcap%
\pgfsetmiterjoin%
\definecolor{currentfill}{rgb}{0.960784,0.462745,0.000000}%
\pgfsetfillcolor{currentfill}%
\pgfsetlinewidth{1.003750pt}%
\definecolor{currentstroke}{rgb}{0.960784,0.462745,0.000000}%
\pgfsetstrokecolor{currentstroke}%
\pgfsetdash{}{0pt}%
\pgfsys@defobject{currentmarker}{\pgfqpoint{-0.055556in}{-0.055556in}}{\pgfqpoint{0.055556in}{0.055556in}}{%
\pgfpathmoveto{\pgfqpoint{-0.055556in}{-0.055556in}}%
\pgfpathlineto{\pgfqpoint{0.055556in}{-0.055556in}}%
\pgfpathlineto{\pgfqpoint{0.055556in}{0.055556in}}%
\pgfpathlineto{\pgfqpoint{-0.055556in}{0.055556in}}%
\pgfpathlineto{\pgfqpoint{-0.055556in}{-0.055556in}}%
\pgfpathclose%
\pgfusepath{stroke,fill}%
}%
\begin{pgfscope}%
\pgfsys@transformshift{0.674954in}{4.357751in}%
\pgfsys@useobject{currentmarker}{}%
\end{pgfscope}%
\begin{pgfscope}%
\pgfsys@transformshift{0.998577in}{4.218213in}%
\pgfsys@useobject{currentmarker}{}%
\end{pgfscope}%
\begin{pgfscope}%
\pgfsys@transformshift{2.293069in}{3.782202in}%
\pgfsys@useobject{currentmarker}{}%
\end{pgfscope}%
\begin{pgfscope}%
\pgfsys@transformshift{3.911184in}{3.443220in}%
\pgfsys@useobject{currentmarker}{}%
\end{pgfscope}%
\begin{pgfscope}%
\pgfsys@transformshift{5.529299in}{3.250093in}%
\pgfsys@useobject{currentmarker}{}%
\end{pgfscope}%
\begin{pgfscope}%
\pgfsys@transformshift{6.823791in}{3.364588in}%
\pgfsys@useobject{currentmarker}{}%
\end{pgfscope}%
\begin{pgfscope}%
\pgfsys@transformshift{7.147414in}{3.474857in}%
\pgfsys@useobject{currentmarker}{}%
\end{pgfscope}%
\end{pgfscope}%
\begin{pgfscope}%
\pgfpathrectangle{\pgfqpoint{0.674954in}{0.862305in}}{\pgfqpoint{6.472460in}{4.369081in}}%
\pgfusepath{clip}%
\pgfsetbuttcap%
\pgfsetroundjoin%
\pgfsetlinewidth{2.509375pt}%
\definecolor{currentstroke}{rgb}{0.219608,0.219608,0.219608}%
\pgfsetstrokecolor{currentstroke}%
\pgfsetdash{{2.500000pt}{4.125000pt}}{0.000000pt}%
\pgfpathmoveto{\pgfqpoint{0.674954in}{3.328978in}}%
\pgfpathlineto{\pgfqpoint{0.998577in}{3.175070in}}%
\pgfpathlineto{\pgfqpoint{2.293069in}{2.683783in}}%
\pgfpathlineto{\pgfqpoint{3.911184in}{2.361778in}}%
\pgfpathlineto{\pgfqpoint{5.529299in}{2.227984in}}%
\pgfpathlineto{\pgfqpoint{6.823791in}{2.316937in}}%
\pgfpathlineto{\pgfqpoint{7.147414in}{2.357226in}}%
\pgfusepath{stroke}%
\end{pgfscope}%
\begin{pgfscope}%
\pgfpathrectangle{\pgfqpoint{0.674954in}{0.862305in}}{\pgfqpoint{6.472460in}{4.369081in}}%
\pgfusepath{clip}%
\pgfsetbuttcap%
\pgfsetmiterjoin%
\definecolor{currentfill}{rgb}{0.219608,0.219608,0.219608}%
\pgfsetfillcolor{currentfill}%
\pgfsetlinewidth{1.003750pt}%
\definecolor{currentstroke}{rgb}{0.219608,0.219608,0.219608}%
\pgfsetstrokecolor{currentstroke}%
\pgfsetdash{}{0pt}%
\pgfsys@defobject{currentmarker}{\pgfqpoint{-0.055556in}{-0.055556in}}{\pgfqpoint{0.055556in}{0.055556in}}{%
\pgfpathmoveto{\pgfqpoint{0.000000in}{0.055556in}}%
\pgfpathlineto{\pgfqpoint{-0.055556in}{-0.055556in}}%
\pgfpathlineto{\pgfqpoint{0.055556in}{-0.055556in}}%
\pgfpathlineto{\pgfqpoint{0.000000in}{0.055556in}}%
\pgfpathclose%
\pgfusepath{stroke,fill}%
}%
\begin{pgfscope}%
\pgfsys@transformshift{0.674954in}{3.328978in}%
\pgfsys@useobject{currentmarker}{}%
\end{pgfscope}%
\begin{pgfscope}%
\pgfsys@transformshift{0.998577in}{3.175070in}%
\pgfsys@useobject{currentmarker}{}%
\end{pgfscope}%
\begin{pgfscope}%
\pgfsys@transformshift{2.293069in}{2.683783in}%
\pgfsys@useobject{currentmarker}{}%
\end{pgfscope}%
\begin{pgfscope}%
\pgfsys@transformshift{3.911184in}{2.361778in}%
\pgfsys@useobject{currentmarker}{}%
\end{pgfscope}%
\begin{pgfscope}%
\pgfsys@transformshift{5.529299in}{2.227984in}%
\pgfsys@useobject{currentmarker}{}%
\end{pgfscope}%
\begin{pgfscope}%
\pgfsys@transformshift{6.823791in}{2.316937in}%
\pgfsys@useobject{currentmarker}{}%
\end{pgfscope}%
\begin{pgfscope}%
\pgfsys@transformshift{7.147414in}{2.357226in}%
\pgfsys@useobject{currentmarker}{}%
\end{pgfscope}%
\end{pgfscope}%
\begin{pgfscope}%
\pgfsetrectcap%
\pgfsetmiterjoin%
\pgfsetlinewidth{2.007500pt}%
\definecolor{currentstroke}{rgb}{0.000000,0.000000,0.000000}%
\pgfsetstrokecolor{currentstroke}%
\pgfsetdash{}{0pt}%
\pgfpathmoveto{\pgfqpoint{0.674954in}{0.862305in}}%
\pgfpathlineto{\pgfqpoint{0.674954in}{5.231386in}}%
\pgfusepath{stroke}%
\end{pgfscope}%
\begin{pgfscope}%
\pgfsetrectcap%
\pgfsetmiterjoin%
\pgfsetlinewidth{2.007500pt}%
\definecolor{currentstroke}{rgb}{0.000000,0.000000,0.000000}%
\pgfsetstrokecolor{currentstroke}%
\pgfsetdash{}{0pt}%
\pgfpathmoveto{\pgfqpoint{0.674954in}{0.862305in}}%
\pgfpathlineto{\pgfqpoint{7.147414in}{0.862305in}}%
\pgfusepath{stroke}%
\end{pgfscope}%
\begin{pgfscope}%
\pgfsetbuttcap%
\pgfsetmiterjoin%
\definecolor{currentfill}{rgb}{1.000000,1.000000,1.000000}%
\pgfsetfillcolor{currentfill}%
\pgfsetlinewidth{1.003750pt}%
\definecolor{currentstroke}{rgb}{0.800000,0.800000,0.800000}%
\pgfsetstrokecolor{currentstroke}%
\pgfsetdash{}{0pt}%
\pgfpathmoveto{\pgfqpoint{0.869398in}{1.001194in}}%
\pgfpathlineto{\pgfqpoint{2.213036in}{1.001194in}}%
\pgfpathquadraticcurveto{\pgfqpoint{2.268591in}{1.001194in}}{\pgfqpoint{2.268591in}{1.056749in}}%
\pgfpathlineto{\pgfqpoint{2.268591in}{2.191056in}}%
\pgfpathquadraticcurveto{\pgfqpoint{2.268591in}{2.246612in}}{\pgfqpoint{2.213036in}{2.246612in}}%
\pgfpathlineto{\pgfqpoint{0.869398in}{2.246612in}}%
\pgfpathquadraticcurveto{\pgfqpoint{0.813843in}{2.246612in}}{\pgfqpoint{0.813843in}{2.191056in}}%
\pgfpathlineto{\pgfqpoint{0.813843in}{1.056749in}}%
\pgfpathquadraticcurveto{\pgfqpoint{0.813843in}{1.001194in}}{\pgfqpoint{0.869398in}{1.001194in}}%
\pgfpathlineto{\pgfqpoint{0.869398in}{1.001194in}}%
\pgfpathclose%
\pgfusepath{stroke,fill}%
\end{pgfscope}%
\begin{pgfscope}%
\pgfsetrectcap%
\pgfsetroundjoin%
\pgfsetlinewidth{2.509375pt}%
\definecolor{currentstroke}{rgb}{0.050980,0.415686,0.509804}%
\pgfsetstrokecolor{currentstroke}%
\pgfsetdash{}{0pt}%
\pgfpathmoveto{\pgfqpoint{0.924954in}{2.038278in}}%
\pgfpathlineto{\pgfqpoint{1.202732in}{2.038278in}}%
\pgfpathlineto{\pgfqpoint{1.480509in}{2.038278in}}%
\pgfusepath{stroke}%
\end{pgfscope}%
\begin{pgfscope}%
\pgfsetbuttcap%
\pgfsetroundjoin%
\definecolor{currentfill}{rgb}{0.050980,0.415686,0.509804}%
\pgfsetfillcolor{currentfill}%
\pgfsetlinewidth{1.003750pt}%
\definecolor{currentstroke}{rgb}{0.050980,0.415686,0.509804}%
\pgfsetstrokecolor{currentstroke}%
\pgfsetdash{}{0pt}%
\pgfsys@defobject{currentmarker}{\pgfqpoint{-0.055556in}{-0.055556in}}{\pgfqpoint{0.055556in}{0.055556in}}{%
\pgfpathmoveto{\pgfqpoint{0.000000in}{-0.055556in}}%
\pgfpathcurveto{\pgfqpoint{0.014734in}{-0.055556in}}{\pgfqpoint{0.028866in}{-0.049702in}}{\pgfqpoint{0.039284in}{-0.039284in}}%
\pgfpathcurveto{\pgfqpoint{0.049702in}{-0.028866in}}{\pgfqpoint{0.055556in}{-0.014734in}}{\pgfqpoint{0.055556in}{0.000000in}}%
\pgfpathcurveto{\pgfqpoint{0.055556in}{0.014734in}}{\pgfqpoint{0.049702in}{0.028866in}}{\pgfqpoint{0.039284in}{0.039284in}}%
\pgfpathcurveto{\pgfqpoint{0.028866in}{0.049702in}}{\pgfqpoint{0.014734in}{0.055556in}}{\pgfqpoint{0.000000in}{0.055556in}}%
\pgfpathcurveto{\pgfqpoint{-0.014734in}{0.055556in}}{\pgfqpoint{-0.028866in}{0.049702in}}{\pgfqpoint{-0.039284in}{0.039284in}}%
\pgfpathcurveto{\pgfqpoint{-0.049702in}{0.028866in}}{\pgfqpoint{-0.055556in}{0.014734in}}{\pgfqpoint{-0.055556in}{0.000000in}}%
\pgfpathcurveto{\pgfqpoint{-0.055556in}{-0.014734in}}{\pgfqpoint{-0.049702in}{-0.028866in}}{\pgfqpoint{-0.039284in}{-0.039284in}}%
\pgfpathcurveto{\pgfqpoint{-0.028866in}{-0.049702in}}{\pgfqpoint{-0.014734in}{-0.055556in}}{\pgfqpoint{0.000000in}{-0.055556in}}%
\pgfpathlineto{\pgfqpoint{0.000000in}{-0.055556in}}%
\pgfpathclose%
\pgfusepath{stroke,fill}%
}%
\begin{pgfscope}%
\pgfsys@transformshift{1.202732in}{2.038278in}%
\pgfsys@useobject{currentmarker}{}%
\end{pgfscope}%
\end{pgfscope}%
\begin{pgfscope}%
\definecolor{textcolor}{rgb}{0.000000,0.000000,0.000000}%
\pgfsetstrokecolor{textcolor}%
\pgfsetfillcolor{textcolor}%
\pgftext[x=1.702732in,y=1.941056in,left,base]{\color{textcolor}{\rmfamily\fontsize{20.000000}{24.000000}\selectfont\catcode`\^=\active\def^{\ifmmode\sp\else\^{}\fi}\catcode`\%=\active\def
\end{pgfscope}%
\begin{pgfscope}%
\pgfsetbuttcap%
\pgfsetroundjoin%
\pgfsetlinewidth{2.509375pt}%
\definecolor{currentstroke}{rgb}{0.960784,0.462745,0.000000}%
\pgfsetstrokecolor{currentstroke}%
\pgfsetdash{{9.250000pt}{4.000000pt}}{0.000000pt}%
\pgfpathmoveto{\pgfqpoint{0.924954in}{1.650917in}}%
\pgfpathlineto{\pgfqpoint{1.202732in}{1.650917in}}%
\pgfpathlineto{\pgfqpoint{1.480509in}{1.650917in}}%
\pgfusepath{stroke}%
\end{pgfscope}%
\begin{pgfscope}%
\pgfsetbuttcap%
\pgfsetmiterjoin%
\definecolor{currentfill}{rgb}{0.960784,0.462745,0.000000}%
\pgfsetfillcolor{currentfill}%
\pgfsetlinewidth{1.003750pt}%
\definecolor{currentstroke}{rgb}{0.960784,0.462745,0.000000}%
\pgfsetstrokecolor{currentstroke}%
\pgfsetdash{}{0pt}%
\pgfsys@defobject{currentmarker}{\pgfqpoint{-0.055556in}{-0.055556in}}{\pgfqpoint{0.055556in}{0.055556in}}{%
\pgfpathmoveto{\pgfqpoint{-0.055556in}{-0.055556in}}%
\pgfpathlineto{\pgfqpoint{0.055556in}{-0.055556in}}%
\pgfpathlineto{\pgfqpoint{0.055556in}{0.055556in}}%
\pgfpathlineto{\pgfqpoint{-0.055556in}{0.055556in}}%
\pgfpathlineto{\pgfqpoint{-0.055556in}{-0.055556in}}%
\pgfpathclose%
\pgfusepath{stroke,fill}%
}%
\begin{pgfscope}%
\pgfsys@transformshift{1.202732in}{1.650917in}%
\pgfsys@useobject{currentmarker}{}%
\end{pgfscope}%
\end{pgfscope}%
\begin{pgfscope}%
\definecolor{textcolor}{rgb}{0.000000,0.000000,0.000000}%
\pgfsetstrokecolor{textcolor}%
\pgfsetfillcolor{textcolor}%
\pgftext[x=1.702732in,y=1.553694in,left,base]{\color{textcolor}{\rmfamily\fontsize{20.000000}{24.000000}\selectfont\catcode`\^=\active\def^{\ifmmode\sp\else\^{}\fi}\catcode`\%=\active\def
\end{pgfscope}%
\begin{pgfscope}%
\pgfsetbuttcap%
\pgfsetroundjoin%
\pgfsetlinewidth{2.509375pt}%
\definecolor{currentstroke}{rgb}{0.219608,0.219608,0.219608}%
\pgfsetstrokecolor{currentstroke}%
\pgfsetdash{{2.500000pt}{4.125000pt}}{0.000000pt}%
\pgfpathmoveto{\pgfqpoint{0.924954in}{1.263555in}}%
\pgfpathlineto{\pgfqpoint{1.202732in}{1.263555in}}%
\pgfpathlineto{\pgfqpoint{1.480509in}{1.263555in}}%
\pgfusepath{stroke}%
\end{pgfscope}%
\begin{pgfscope}%
\pgfsetbuttcap%
\pgfsetmiterjoin%
\definecolor{currentfill}{rgb}{0.219608,0.219608,0.219608}%
\pgfsetfillcolor{currentfill}%
\pgfsetlinewidth{1.003750pt}%
\definecolor{currentstroke}{rgb}{0.219608,0.219608,0.219608}%
\pgfsetstrokecolor{currentstroke}%
\pgfsetdash{}{0pt}%
\pgfsys@defobject{currentmarker}{\pgfqpoint{-0.055556in}{-0.055556in}}{\pgfqpoint{0.055556in}{0.055556in}}{%
\pgfpathmoveto{\pgfqpoint{0.000000in}{0.055556in}}%
\pgfpathlineto{\pgfqpoint{-0.055556in}{-0.055556in}}%
\pgfpathlineto{\pgfqpoint{0.055556in}{-0.055556in}}%
\pgfpathlineto{\pgfqpoint{0.000000in}{0.055556in}}%
\pgfpathclose%
\pgfusepath{stroke,fill}%
}%
\begin{pgfscope}%
\pgfsys@transformshift{1.202732in}{1.263555in}%
\pgfsys@useobject{currentmarker}{}%
\end{pgfscope}%
\end{pgfscope}%
\begin{pgfscope}%
\definecolor{textcolor}{rgb}{0.000000,0.000000,0.000000}%
\pgfsetstrokecolor{textcolor}%
\pgfsetfillcolor{textcolor}%
\pgftext[x=1.702732in,y=1.166333in,left,base]{\color{textcolor}{\rmfamily\fontsize{20.000000}{24.000000}\selectfont\catcode`\^=\active\def^{\ifmmode\sp\else\^{}\fi}\catcode`\%=\active\def
\end{pgfscope}%
\end{pgfpicture}%
\makeatother%
\endgroup%

%% file: images/graphs/thermal_ratio_camera_token_RRA.pgf
\begingroup%
\makeatletter%
\begin{pgfpicture}%
\pgfpathrectangle{\pgfpointorigin}{\pgfqpoint{7.450000in}{5.450000in}}%
\pgfusepath{use as bounding box, clip}%
\begin{pgfscope}%
\pgfsetbuttcap%
\pgfsetmiterjoin%
\definecolor{currentfill}{rgb}{1.000000,1.000000,1.000000}%
\pgfsetfillcolor{currentfill}%
\pgfsetlinewidth{0.000000pt}%
\definecolor{currentstroke}{rgb}{1.000000,1.000000,1.000000}%
\pgfsetstrokecolor{currentstroke}%
\pgfsetdash{}{0pt}%
\pgfpathmoveto{\pgfqpoint{0.000000in}{0.000000in}}%
\pgfpathlineto{\pgfqpoint{7.450000in}{0.000000in}}%
\pgfpathlineto{\pgfqpoint{7.450000in}{5.450000in}}%
\pgfpathlineto{\pgfqpoint{0.000000in}{5.450000in}}%
\pgfpathlineto{\pgfqpoint{0.000000in}{0.000000in}}%
\pgfpathclose%
\pgfusepath{fill}%
\end{pgfscope}%
\begin{pgfscope}%
\pgfsetbuttcap%
\pgfsetmiterjoin%
\definecolor{currentfill}{rgb}{1.000000,1.000000,1.000000}%
\pgfsetfillcolor{currentfill}%
\pgfsetlinewidth{0.000000pt}%
\definecolor{currentstroke}{rgb}{0.000000,0.000000,0.000000}%
\pgfsetstrokecolor{currentstroke}%
\pgfsetstrokeopacity{0.000000}%
\pgfsetdash{}{0pt}%
\pgfpathmoveto{\pgfqpoint{0.674954in}{0.862305in}}%
\pgfpathlineto{\pgfqpoint{7.147414in}{0.862305in}}%
\pgfpathlineto{\pgfqpoint{7.147414in}{5.231386in}}%
\pgfpathlineto{\pgfqpoint{0.674954in}{5.231386in}}%
\pgfpathlineto{\pgfqpoint{0.674954in}{0.862305in}}%
\pgfpathclose%
\pgfusepath{fill}%
\end{pgfscope}%
\begin{pgfscope}%
\pgfpathrectangle{\pgfqpoint{0.674954in}{0.862305in}}{\pgfqpoint{6.472460in}{4.369081in}}%
\pgfusepath{clip}%
\pgfsetbuttcap%
\pgfsetroundjoin%
\definecolor{currentfill}{rgb}{0.050980,0.415686,0.509804}%
\pgfsetfillcolor{currentfill}%
\pgfsetfillopacity{0.300000}%
\pgfsetlinewidth{1.003750pt}%
\definecolor{currentstroke}{rgb}{0.050980,0.415686,0.509804}%
\pgfsetstrokecolor{currentstroke}%
\pgfsetstrokeopacity{0.300000}%
\pgfsetdash{}{0pt}%
\pgfsys@defobject{currentmarker}{\pgfqpoint{0.674954in}{4.434307in}}{\pgfqpoint{7.147414in}{5.231386in}}{%
\pgfpathmoveto{\pgfqpoint{0.674954in}{5.231386in}}%
\pgfpathlineto{\pgfqpoint{0.674954in}{5.231386in}}%
\pgfpathlineto{\pgfqpoint{0.998577in}{5.145166in}}%
\pgfpathlineto{\pgfqpoint{2.293069in}{4.858878in}}%
\pgfpathlineto{\pgfqpoint{3.911184in}{4.623616in}}%
\pgfpathlineto{\pgfqpoint{5.529299in}{4.434307in}}%
\pgfpathlineto{\pgfqpoint{6.823791in}{4.613212in}}%
\pgfpathlineto{\pgfqpoint{7.147414in}{4.808528in}}%
\pgfpathlineto{\pgfqpoint{7.147414in}{5.169299in}}%
\pgfpathlineto{\pgfqpoint{7.147414in}{5.169299in}}%
\pgfpathlineto{\pgfqpoint{6.823791in}{5.018021in}}%
\pgfpathlineto{\pgfqpoint{5.529299in}{4.890207in}}%
\pgfpathlineto{\pgfqpoint{3.911184in}{5.008003in}}%
\pgfpathlineto{\pgfqpoint{2.293069in}{5.101008in}}%
\pgfpathlineto{\pgfqpoint{0.998577in}{5.198634in}}%
\pgfpathlineto{\pgfqpoint{0.674954in}{5.231386in}}%
\pgfpathlineto{\pgfqpoint{0.674954in}{5.231386in}}%
\pgfpathclose%
\pgfusepath{stroke,fill}%
}%
\begin{pgfscope}%
\pgfsys@transformshift{0.000000in}{0.000000in}%
\pgfsys@useobject{currentmarker}{}%
\end{pgfscope}%
\end{pgfscope}%
\begin{pgfscope}%
\pgfpathrectangle{\pgfqpoint{0.674954in}{0.862305in}}{\pgfqpoint{6.472460in}{4.369081in}}%
\pgfusepath{clip}%
\pgfsetbuttcap%
\pgfsetroundjoin%
\definecolor{currentfill}{rgb}{0.960784,0.462745,0.000000}%
\pgfsetfillcolor{currentfill}%
\pgfsetfillopacity{0.300000}%
\pgfsetlinewidth{1.003750pt}%
\definecolor{currentstroke}{rgb}{0.960784,0.462745,0.000000}%
\pgfsetstrokecolor{currentstroke}%
\pgfsetstrokeopacity{0.300000}%
\pgfsetdash{{3.700000pt}{1.600000pt}}{0.000000pt}%
\pgfpathmoveto{\pgfqpoint{0.674954in}{5.231386in}}%
\pgfpathlineto{\pgfqpoint{0.674954in}{5.231386in}}%
\pgfpathlineto{\pgfqpoint{0.998577in}{5.140104in}}%
\pgfpathlineto{\pgfqpoint{2.293069in}{4.779552in}}%
\pgfpathlineto{\pgfqpoint{3.911184in}{4.501289in}}%
\pgfpathlineto{\pgfqpoint{5.529299in}{4.199184in}}%
\pgfpathlineto{\pgfqpoint{6.823791in}{4.296930in}}%
\pgfpathlineto{\pgfqpoint{7.147414in}{4.505489in}}%
\pgfpathlineto{\pgfqpoint{7.147414in}{4.969256in}}%
\pgfpathlineto{\pgfqpoint{7.147414in}{4.969256in}}%
\pgfpathlineto{\pgfqpoint{6.823791in}{4.784573in}}%
\pgfpathlineto{\pgfqpoint{5.529299in}{4.719574in}}%
\pgfpathlineto{\pgfqpoint{3.911184in}{4.924759in}}%
\pgfpathlineto{\pgfqpoint{2.293069in}{5.091796in}}%
\pgfpathlineto{\pgfqpoint{0.998577in}{5.194977in}}%
\pgfpathlineto{\pgfqpoint{0.674954in}{5.231386in}}%
\pgfpathlineto{\pgfqpoint{0.674954in}{5.231386in}}%
\pgfpathclose%
\pgfusepath{stroke,fill}%
\end{pgfscope}%
\begin{pgfscope}%
\pgfpathrectangle{\pgfqpoint{0.674954in}{0.862305in}}{\pgfqpoint{6.472460in}{4.369081in}}%
\pgfusepath{clip}%
\pgfsetbuttcap%
\pgfsetroundjoin%
\definecolor{currentfill}{rgb}{0.219608,0.219608,0.219608}%
\pgfsetfillcolor{currentfill}%
\pgfsetfillopacity{0.300000}%
\pgfsetlinewidth{1.003750pt}%
\definecolor{currentstroke}{rgb}{0.219608,0.219608,0.219608}%
\pgfsetstrokecolor{currentstroke}%
\pgfsetstrokeopacity{0.300000}%
\pgfsetdash{{1.000000pt}{1.650000pt}}{0.000000pt}%
\pgfpathmoveto{\pgfqpoint{0.674954in}{5.224534in}}%
\pgfpathlineto{\pgfqpoint{0.674954in}{5.218729in}}%
\pgfpathlineto{\pgfqpoint{0.998577in}{5.059038in}}%
\pgfpathlineto{\pgfqpoint{2.293069in}{4.455747in}}%
\pgfpathlineto{\pgfqpoint{3.911184in}{4.048732in}}%
\pgfpathlineto{\pgfqpoint{5.529299in}{3.770597in}}%
\pgfpathlineto{\pgfqpoint{6.823791in}{3.853102in}}%
\pgfpathlineto{\pgfqpoint{7.147414in}{3.904640in}}%
\pgfpathlineto{\pgfqpoint{7.147414in}{4.461116in}}%
\pgfpathlineto{\pgfqpoint{7.147414in}{4.461116in}}%
\pgfpathlineto{\pgfqpoint{6.823791in}{4.404740in}}%
\pgfpathlineto{\pgfqpoint{5.529299in}{4.330493in}}%
\pgfpathlineto{\pgfqpoint{3.911184in}{4.517118in}}%
\pgfpathlineto{\pgfqpoint{2.293069in}{4.781542in}}%
\pgfpathlineto{\pgfqpoint{0.998577in}{5.119875in}}%
\pgfpathlineto{\pgfqpoint{0.674954in}{5.224534in}}%
\pgfpathlineto{\pgfqpoint{0.674954in}{5.224534in}}%
\pgfpathclose%
\pgfusepath{stroke,fill}%
\end{pgfscope}%
\begin{pgfscope}%
\pgfpathrectangle{\pgfqpoint{0.674954in}{0.862305in}}{\pgfqpoint{6.472460in}{4.369081in}}%
\pgfusepath{clip}%
\pgfsetbuttcap%
\pgfsetroundjoin%
\pgfsetlinewidth{2.007500pt}%
\definecolor{currentstroke}{rgb}{0.501961,0.501961,0.501961}%
\pgfsetstrokecolor{currentstroke}%
\pgfsetstrokeopacity{0.300000}%
\pgfsetdash{{7.400000pt}{3.200000pt}}{0.000000pt}%
\pgfpathmoveto{\pgfqpoint{0.674954in}{0.862305in}}%
\pgfpathlineto{\pgfqpoint{0.674954in}{5.231386in}}%
\pgfusepath{stroke}%
\end{pgfscope}%
\begin{pgfscope}%
\pgfsetbuttcap%
\pgfsetroundjoin%
\definecolor{currentfill}{rgb}{0.000000,0.000000,0.000000}%
\pgfsetfillcolor{currentfill}%
\pgfsetlinewidth{0.803000pt}%
\definecolor{currentstroke}{rgb}{0.000000,0.000000,0.000000}%
\pgfsetstrokecolor{currentstroke}%
\pgfsetdash{}{0pt}%
\pgfsys@defobject{currentmarker}{\pgfqpoint{0.000000in}{-0.048611in}}{\pgfqpoint{0.000000in}{0.000000in}}{%
\pgfpathmoveto{\pgfqpoint{0.000000in}{0.000000in}}%
\pgfpathlineto{\pgfqpoint{0.000000in}{-0.048611in}}%
\pgfusepath{stroke,fill}%
}%
\begin{pgfscope}%
\pgfsys@transformshift{0.674954in}{0.862305in}%
\pgfsys@useobject{currentmarker}{}%
\end{pgfscope}%
\end{pgfscope}%
\begin{pgfscope}%
\definecolor{textcolor}{rgb}{0.000000,0.000000,0.000000}%
\pgfsetstrokecolor{textcolor}%
\pgfsetfillcolor{textcolor}%
\pgftext[x=0.674954in,y=0.765082in,,top]{\color{textcolor}{\rmfamily\fontsize{25.000000}{30.000000}\selectfont\catcode`\^=\active\def^{\ifmmode\sp\else\^{}\fi}\catcode`\%=\active\def
\end{pgfscope}%
\begin{pgfscope}%
\pgfpathrectangle{\pgfqpoint{0.674954in}{0.862305in}}{\pgfqpoint{6.472460in}{4.369081in}}%
\pgfusepath{clip}%
\pgfsetbuttcap%
\pgfsetroundjoin%
\pgfsetlinewidth{2.007500pt}%
\definecolor{currentstroke}{rgb}{0.501961,0.501961,0.501961}%
\pgfsetstrokecolor{currentstroke}%
\pgfsetstrokeopacity{0.300000}%
\pgfsetdash{{7.400000pt}{3.200000pt}}{0.000000pt}%
\pgfpathmoveto{\pgfqpoint{1.969446in}{0.862305in}}%
\pgfpathlineto{\pgfqpoint{1.969446in}{5.231386in}}%
\pgfusepath{stroke}%
\end{pgfscope}%
\begin{pgfscope}%
\pgfsetbuttcap%
\pgfsetroundjoin%
\definecolor{currentfill}{rgb}{0.000000,0.000000,0.000000}%
\pgfsetfillcolor{currentfill}%
\pgfsetlinewidth{0.803000pt}%
\definecolor{currentstroke}{rgb}{0.000000,0.000000,0.000000}%
\pgfsetstrokecolor{currentstroke}%
\pgfsetdash{}{0pt}%
\pgfsys@defobject{currentmarker}{\pgfqpoint{0.000000in}{-0.048611in}}{\pgfqpoint{0.000000in}{0.000000in}}{%
\pgfpathmoveto{\pgfqpoint{0.000000in}{0.000000in}}%
\pgfpathlineto{\pgfqpoint{0.000000in}{-0.048611in}}%
\pgfusepath{stroke,fill}%
}%
\begin{pgfscope}%
\pgfsys@transformshift{1.969446in}{0.862305in}%
\pgfsys@useobject{currentmarker}{}%
\end{pgfscope}%
\end{pgfscope}%
\begin{pgfscope}%
\definecolor{textcolor}{rgb}{0.000000,0.000000,0.000000}%
\pgfsetstrokecolor{textcolor}%
\pgfsetfillcolor{textcolor}%
\pgftext[x=1.969446in,y=0.765082in,,top]{\color{textcolor}{\rmfamily\fontsize{25.000000}{30.000000}\selectfont\catcode`\^=\active\def^{\ifmmode\sp\else\^{}\fi}\catcode`\%=\active\def
\end{pgfscope}%
\begin{pgfscope}%
\pgfpathrectangle{\pgfqpoint{0.674954in}{0.862305in}}{\pgfqpoint{6.472460in}{4.369081in}}%
\pgfusepath{clip}%
\pgfsetbuttcap%
\pgfsetroundjoin%
\pgfsetlinewidth{2.007500pt}%
\definecolor{currentstroke}{rgb}{0.501961,0.501961,0.501961}%
\pgfsetstrokecolor{currentstroke}%
\pgfsetstrokeopacity{0.300000}%
\pgfsetdash{{7.400000pt}{3.200000pt}}{0.000000pt}%
\pgfpathmoveto{\pgfqpoint{3.263938in}{0.862305in}}%
\pgfpathlineto{\pgfqpoint{3.263938in}{5.231386in}}%
\pgfusepath{stroke}%
\end{pgfscope}%
\begin{pgfscope}%
\pgfsetbuttcap%
\pgfsetroundjoin%
\definecolor{currentfill}{rgb}{0.000000,0.000000,0.000000}%
\pgfsetfillcolor{currentfill}%
\pgfsetlinewidth{0.803000pt}%
\definecolor{currentstroke}{rgb}{0.000000,0.000000,0.000000}%
\pgfsetstrokecolor{currentstroke}%
\pgfsetdash{}{0pt}%
\pgfsys@defobject{currentmarker}{\pgfqpoint{0.000000in}{-0.048611in}}{\pgfqpoint{0.000000in}{0.000000in}}{%
\pgfpathmoveto{\pgfqpoint{0.000000in}{0.000000in}}%
\pgfpathlineto{\pgfqpoint{0.000000in}{-0.048611in}}%
\pgfusepath{stroke,fill}%
}%
\begin{pgfscope}%
\pgfsys@transformshift{3.263938in}{0.862305in}%
\pgfsys@useobject{currentmarker}{}%
\end{pgfscope}%
\end{pgfscope}%
\begin{pgfscope}%
\definecolor{textcolor}{rgb}{0.000000,0.000000,0.000000}%
\pgfsetstrokecolor{textcolor}%
\pgfsetfillcolor{textcolor}%
\pgftext[x=3.263938in,y=0.765082in,,top]{\color{textcolor}{\rmfamily\fontsize{25.000000}{30.000000}\selectfont\catcode`\^=\active\def^{\ifmmode\sp\else\^{}\fi}\catcode`\%=\active\def
\end{pgfscope}%
\begin{pgfscope}%
\pgfpathrectangle{\pgfqpoint{0.674954in}{0.862305in}}{\pgfqpoint{6.472460in}{4.369081in}}%
\pgfusepath{clip}%
\pgfsetbuttcap%
\pgfsetroundjoin%
\pgfsetlinewidth{2.007500pt}%
\definecolor{currentstroke}{rgb}{0.501961,0.501961,0.501961}%
\pgfsetstrokecolor{currentstroke}%
\pgfsetstrokeopacity{0.300000}%
\pgfsetdash{{7.400000pt}{3.200000pt}}{0.000000pt}%
\pgfpathmoveto{\pgfqpoint{4.558430in}{0.862305in}}%
\pgfpathlineto{\pgfqpoint{4.558430in}{5.231386in}}%
\pgfusepath{stroke}%
\end{pgfscope}%
\begin{pgfscope}%
\pgfsetbuttcap%
\pgfsetroundjoin%
\definecolor{currentfill}{rgb}{0.000000,0.000000,0.000000}%
\pgfsetfillcolor{currentfill}%
\pgfsetlinewidth{0.803000pt}%
\definecolor{currentstroke}{rgb}{0.000000,0.000000,0.000000}%
\pgfsetstrokecolor{currentstroke}%
\pgfsetdash{}{0pt}%
\pgfsys@defobject{currentmarker}{\pgfqpoint{0.000000in}{-0.048611in}}{\pgfqpoint{0.000000in}{0.000000in}}{%
\pgfpathmoveto{\pgfqpoint{0.000000in}{0.000000in}}%
\pgfpathlineto{\pgfqpoint{0.000000in}{-0.048611in}}%
\pgfusepath{stroke,fill}%
}%
\begin{pgfscope}%
\pgfsys@transformshift{4.558430in}{0.862305in}%
\pgfsys@useobject{currentmarker}{}%
\end{pgfscope}%
\end{pgfscope}%
\begin{pgfscope}%
\definecolor{textcolor}{rgb}{0.000000,0.000000,0.000000}%
\pgfsetstrokecolor{textcolor}%
\pgfsetfillcolor{textcolor}%
\pgftext[x=4.558430in,y=0.765082in,,top]{\color{textcolor}{\rmfamily\fontsize{25.000000}{30.000000}\selectfont\catcode`\^=\active\def^{\ifmmode\sp\else\^{}\fi}\catcode`\%=\active\def
\end{pgfscope}%
\begin{pgfscope}%
\pgfpathrectangle{\pgfqpoint{0.674954in}{0.862305in}}{\pgfqpoint{6.472460in}{4.369081in}}%
\pgfusepath{clip}%
\pgfsetbuttcap%
\pgfsetroundjoin%
\pgfsetlinewidth{2.007500pt}%
\definecolor{currentstroke}{rgb}{0.501961,0.501961,0.501961}%
\pgfsetstrokecolor{currentstroke}%
\pgfsetstrokeopacity{0.300000}%
\pgfsetdash{{7.400000pt}{3.200000pt}}{0.000000pt}%
\pgfpathmoveto{\pgfqpoint{5.852922in}{0.862305in}}%
\pgfpathlineto{\pgfqpoint{5.852922in}{5.231386in}}%
\pgfusepath{stroke}%
\end{pgfscope}%
\begin{pgfscope}%
\pgfsetbuttcap%
\pgfsetroundjoin%
\definecolor{currentfill}{rgb}{0.000000,0.000000,0.000000}%
\pgfsetfillcolor{currentfill}%
\pgfsetlinewidth{0.803000pt}%
\definecolor{currentstroke}{rgb}{0.000000,0.000000,0.000000}%
\pgfsetstrokecolor{currentstroke}%
\pgfsetdash{}{0pt}%
\pgfsys@defobject{currentmarker}{\pgfqpoint{0.000000in}{-0.048611in}}{\pgfqpoint{0.000000in}{0.000000in}}{%
\pgfpathmoveto{\pgfqpoint{0.000000in}{0.000000in}}%
\pgfpathlineto{\pgfqpoint{0.000000in}{-0.048611in}}%
\pgfusepath{stroke,fill}%
}%
\begin{pgfscope}%
\pgfsys@transformshift{5.852922in}{0.862305in}%
\pgfsys@useobject{currentmarker}{}%
\end{pgfscope}%
\end{pgfscope}%
\begin{pgfscope}%
\definecolor{textcolor}{rgb}{0.000000,0.000000,0.000000}%
\pgfsetstrokecolor{textcolor}%
\pgfsetfillcolor{textcolor}%
\pgftext[x=5.852922in,y=0.765082in,,top]{\color{textcolor}{\rmfamily\fontsize{25.000000}{30.000000}\selectfont\catcode`\^=\active\def^{\ifmmode\sp\else\^{}\fi}\catcode`\%=\active\def
\end{pgfscope}%
\begin{pgfscope}%
\pgfpathrectangle{\pgfqpoint{0.674954in}{0.862305in}}{\pgfqpoint{6.472460in}{4.369081in}}%
\pgfusepath{clip}%
\pgfsetbuttcap%
\pgfsetroundjoin%
\pgfsetlinewidth{2.007500pt}%
\definecolor{currentstroke}{rgb}{0.501961,0.501961,0.501961}%
\pgfsetstrokecolor{currentstroke}%
\pgfsetstrokeopacity{0.300000}%
\pgfsetdash{{7.400000pt}{3.200000pt}}{0.000000pt}%
\pgfpathmoveto{\pgfqpoint{7.147414in}{0.862305in}}%
\pgfpathlineto{\pgfqpoint{7.147414in}{5.231386in}}%
\pgfusepath{stroke}%
\end{pgfscope}%
\begin{pgfscope}%
\pgfsetbuttcap%
\pgfsetroundjoin%
\definecolor{currentfill}{rgb}{0.000000,0.000000,0.000000}%
\pgfsetfillcolor{currentfill}%
\pgfsetlinewidth{0.803000pt}%
\definecolor{currentstroke}{rgb}{0.000000,0.000000,0.000000}%
\pgfsetstrokecolor{currentstroke}%
\pgfsetdash{}{0pt}%
\pgfsys@defobject{currentmarker}{\pgfqpoint{0.000000in}{-0.048611in}}{\pgfqpoint{0.000000in}{0.000000in}}{%
\pgfpathmoveto{\pgfqpoint{0.000000in}{0.000000in}}%
\pgfpathlineto{\pgfqpoint{0.000000in}{-0.048611in}}%
\pgfusepath{stroke,fill}%
}%
\begin{pgfscope}%
\pgfsys@transformshift{7.147414in}{0.862305in}%
\pgfsys@useobject{currentmarker}{}%
\end{pgfscope}%
\end{pgfscope}%
\begin{pgfscope}%
\definecolor{textcolor}{rgb}{0.000000,0.000000,0.000000}%
\pgfsetstrokecolor{textcolor}%
\pgfsetfillcolor{textcolor}%
\pgftext[x=7.147414in,y=0.765082in,,top]{\color{textcolor}{\rmfamily\fontsize{25.000000}{30.000000}\selectfont\catcode`\^=\active\def^{\ifmmode\sp\else\^{}\fi}\catcode`\%=\active\def
\end{pgfscope}%
\begin{pgfscope}%
\definecolor{textcolor}{rgb}{0.000000,0.000000,0.000000}%
\pgfsetstrokecolor{textcolor}%
\pgfsetfillcolor{textcolor}%
\pgftext[x=3.911184in,y=0.404763in,,top]{\color{textcolor}{\rmfamily\fontsize{25.000000}{30.000000}\selectfont\catcode`\^=\active\def^{\ifmmode\sp\else\^{}\fi}\catcode`\%=\active\def
\end{pgfscope}%
\begin{pgfscope}%
\pgfpathrectangle{\pgfqpoint{0.674954in}{0.862305in}}{\pgfqpoint{6.472460in}{4.369081in}}%
\pgfusepath{clip}%
\pgfsetbuttcap%
\pgfsetroundjoin%
\pgfsetlinewidth{2.007500pt}%
\definecolor{currentstroke}{rgb}{0.501961,0.501961,0.501961}%
\pgfsetstrokecolor{currentstroke}%
\pgfsetstrokeopacity{0.300000}%
\pgfsetdash{{7.400000pt}{3.200000pt}}{0.000000pt}%
\pgfpathmoveto{\pgfqpoint{0.674954in}{1.954575in}}%
\pgfpathlineto{\pgfqpoint{7.147414in}{1.954575in}}%
\pgfusepath{stroke}%
\end{pgfscope}%
\begin{pgfscope}%
\pgfsetbuttcap%
\pgfsetroundjoin%
\definecolor{currentfill}{rgb}{0.000000,0.000000,0.000000}%
\pgfsetfillcolor{currentfill}%
\pgfsetlinewidth{0.803000pt}%
\definecolor{currentstroke}{rgb}{0.000000,0.000000,0.000000}%
\pgfsetstrokecolor{currentstroke}%
\pgfsetdash{}{0pt}%
\pgfsys@defobject{currentmarker}{\pgfqpoint{-0.048611in}{0.000000in}}{\pgfqpoint{-0.000000in}{0.000000in}}{%
\pgfpathmoveto{\pgfqpoint{-0.000000in}{0.000000in}}%
\pgfpathlineto{\pgfqpoint{-0.048611in}{0.000000in}}%
\pgfusepath{stroke,fill}%
}%
\begin{pgfscope}%
\pgfsys@transformshift{0.674954in}{1.954575in}%
\pgfsys@useobject{currentmarker}{}%
\end{pgfscope}%
\end{pgfscope}%
\begin{pgfscope}%
\definecolor{textcolor}{rgb}{0.000000,0.000000,0.000000}%
\pgfsetstrokecolor{textcolor}%
\pgfsetfillcolor{textcolor}%
\pgftext[x=0.259244in, y=1.835961in, left, base]{\color{textcolor}{\rmfamily\fontsize{25.000000}{30.000000}\selectfont\catcode`\^=\active\def^{\ifmmode\sp\else\^{}\fi}\catcode`\%=\active\def
\end{pgfscope}%
\begin{pgfscope}%
\pgfpathrectangle{\pgfqpoint{0.674954in}{0.862305in}}{\pgfqpoint{6.472460in}{4.369081in}}%
\pgfusepath{clip}%
\pgfsetbuttcap%
\pgfsetroundjoin%
\pgfsetlinewidth{2.007500pt}%
\definecolor{currentstroke}{rgb}{0.501961,0.501961,0.501961}%
\pgfsetstrokecolor{currentstroke}%
\pgfsetstrokeopacity{0.300000}%
\pgfsetdash{{7.400000pt}{3.200000pt}}{0.000000pt}%
\pgfpathmoveto{\pgfqpoint{0.674954in}{3.046845in}}%
\pgfpathlineto{\pgfqpoint{7.147414in}{3.046845in}}%
\pgfusepath{stroke}%
\end{pgfscope}%
\begin{pgfscope}%
\pgfsetbuttcap%
\pgfsetroundjoin%
\definecolor{currentfill}{rgb}{0.000000,0.000000,0.000000}%
\pgfsetfillcolor{currentfill}%
\pgfsetlinewidth{0.803000pt}%
\definecolor{currentstroke}{rgb}{0.000000,0.000000,0.000000}%
\pgfsetstrokecolor{currentstroke}%
\pgfsetdash{}{0pt}%
\pgfsys@defobject{currentmarker}{\pgfqpoint{-0.048611in}{0.000000in}}{\pgfqpoint{-0.000000in}{0.000000in}}{%
\pgfpathmoveto{\pgfqpoint{-0.000000in}{0.000000in}}%
\pgfpathlineto{\pgfqpoint{-0.048611in}{0.000000in}}%
\pgfusepath{stroke,fill}%
}%
\begin{pgfscope}%
\pgfsys@transformshift{0.674954in}{3.046845in}%
\pgfsys@useobject{currentmarker}{}%
\end{pgfscope}%
\end{pgfscope}%
\begin{pgfscope}%
\definecolor{textcolor}{rgb}{0.000000,0.000000,0.000000}%
\pgfsetstrokecolor{textcolor}%
\pgfsetfillcolor{textcolor}%
\pgftext[x=0.259244in, y=2.928231in, left, base]{\color{textcolor}{\rmfamily\fontsize{25.000000}{30.000000}\selectfont\catcode`\^=\active\def^{\ifmmode\sp\else\^{}\fi}\catcode`\%=\active\def
\end{pgfscope}%
\begin{pgfscope}%
\pgfpathrectangle{\pgfqpoint{0.674954in}{0.862305in}}{\pgfqpoint{6.472460in}{4.369081in}}%
\pgfusepath{clip}%
\pgfsetbuttcap%
\pgfsetroundjoin%
\pgfsetlinewidth{2.007500pt}%
\definecolor{currentstroke}{rgb}{0.501961,0.501961,0.501961}%
\pgfsetstrokecolor{currentstroke}%
\pgfsetstrokeopacity{0.300000}%
\pgfsetdash{{7.400000pt}{3.200000pt}}{0.000000pt}%
\pgfpathmoveto{\pgfqpoint{0.674954in}{4.139116in}}%
\pgfpathlineto{\pgfqpoint{7.147414in}{4.139116in}}%
\pgfusepath{stroke}%
\end{pgfscope}%
\begin{pgfscope}%
\pgfsetbuttcap%
\pgfsetroundjoin%
\definecolor{currentfill}{rgb}{0.000000,0.000000,0.000000}%
\pgfsetfillcolor{currentfill}%
\pgfsetlinewidth{0.803000pt}%
\definecolor{currentstroke}{rgb}{0.000000,0.000000,0.000000}%
\pgfsetstrokecolor{currentstroke}%
\pgfsetdash{}{0pt}%
\pgfsys@defobject{currentmarker}{\pgfqpoint{-0.048611in}{0.000000in}}{\pgfqpoint{-0.000000in}{0.000000in}}{%
\pgfpathmoveto{\pgfqpoint{-0.000000in}{0.000000in}}%
\pgfpathlineto{\pgfqpoint{-0.048611in}{0.000000in}}%
\pgfusepath{stroke,fill}%
}%
\begin{pgfscope}%
\pgfsys@transformshift{0.674954in}{4.139116in}%
\pgfsys@useobject{currentmarker}{}%
\end{pgfscope}%
\end{pgfscope}%
\begin{pgfscope}%
\definecolor{textcolor}{rgb}{0.000000,0.000000,0.000000}%
\pgfsetstrokecolor{textcolor}%
\pgfsetfillcolor{textcolor}%
\pgftext[x=0.259244in, y=4.020501in, left, base]{\color{textcolor}{\rmfamily\fontsize{25.000000}{30.000000}\selectfont\catcode`\^=\active\def^{\ifmmode\sp\else\^{}\fi}\catcode`\%=\active\def
\end{pgfscope}%
\begin{pgfscope}%
\pgfpathrectangle{\pgfqpoint{0.674954in}{0.862305in}}{\pgfqpoint{6.472460in}{4.369081in}}%
\pgfusepath{clip}%
\pgfsetbuttcap%
\pgfsetroundjoin%
\pgfsetlinewidth{2.007500pt}%
\definecolor{currentstroke}{rgb}{0.501961,0.501961,0.501961}%
\pgfsetstrokecolor{currentstroke}%
\pgfsetstrokeopacity{0.300000}%
\pgfsetdash{{7.400000pt}{3.200000pt}}{0.000000pt}%
\pgfpathmoveto{\pgfqpoint{0.674954in}{5.231386in}}%
\pgfpathlineto{\pgfqpoint{7.147414in}{5.231386in}}%
\pgfusepath{stroke}%
\end{pgfscope}%
\begin{pgfscope}%
\pgfsetbuttcap%
\pgfsetroundjoin%
\definecolor{currentfill}{rgb}{0.000000,0.000000,0.000000}%
\pgfsetfillcolor{currentfill}%
\pgfsetlinewidth{0.803000pt}%
\definecolor{currentstroke}{rgb}{0.000000,0.000000,0.000000}%
\pgfsetstrokecolor{currentstroke}%
\pgfsetdash{}{0pt}%
\pgfsys@defobject{currentmarker}{\pgfqpoint{-0.048611in}{0.000000in}}{\pgfqpoint{-0.000000in}{0.000000in}}{%
\pgfpathmoveto{\pgfqpoint{-0.000000in}{0.000000in}}%
\pgfpathlineto{\pgfqpoint{-0.048611in}{0.000000in}}%
\pgfusepath{stroke,fill}%
}%
\begin{pgfscope}%
\pgfsys@transformshift{0.674954in}{5.231386in}%
\pgfsys@useobject{currentmarker}{}%
\end{pgfscope}%
\end{pgfscope}%
\begin{pgfscope}%
\definecolor{textcolor}{rgb}{0.000000,0.000000,0.000000}%
\pgfsetstrokecolor{textcolor}%
\pgfsetfillcolor{textcolor}%
\pgftext[x=0.100000in, y=5.112772in, left, base]{\color{textcolor}{\rmfamily\fontsize{25.000000}{30.000000}\selectfont\catcode`\^=\active\def^{\ifmmode\sp\else\^{}\fi}\catcode`\%=\active\def
\end{pgfscope}%
\begin{pgfscope}%
\pgfpathrectangle{\pgfqpoint{0.674954in}{0.862305in}}{\pgfqpoint{6.472460in}{4.369081in}}%
\pgfusepath{clip}%
\pgfsetrectcap%
\pgfsetroundjoin%
\pgfsetlinewidth{2.509375pt}%
\definecolor{currentstroke}{rgb}{0.050980,0.415686,0.509804}%
\pgfsetstrokecolor{currentstroke}%
\pgfsetdash{}{0pt}%
\pgfpathmoveto{\pgfqpoint{0.674954in}{5.231386in}}%
\pgfpathlineto{\pgfqpoint{0.998577in}{5.173271in}}%
\pgfpathlineto{\pgfqpoint{2.293069in}{4.986364in}}%
\pgfpathlineto{\pgfqpoint{3.911184in}{4.820665in}}%
\pgfpathlineto{\pgfqpoint{5.529299in}{4.676434in}}%
\pgfpathlineto{\pgfqpoint{6.823791in}{4.803713in}}%
\pgfpathlineto{\pgfqpoint{7.147414in}{4.961556in}}%
\pgfusepath{stroke}%
\end{pgfscope}%
\begin{pgfscope}%
\pgfpathrectangle{\pgfqpoint{0.674954in}{0.862305in}}{\pgfqpoint{6.472460in}{4.369081in}}%
\pgfusepath{clip}%
\pgfsetbuttcap%
\pgfsetroundjoin%
\definecolor{currentfill}{rgb}{0.050980,0.415686,0.509804}%
\pgfsetfillcolor{currentfill}%
\pgfsetlinewidth{1.003750pt}%
\definecolor{currentstroke}{rgb}{0.050980,0.415686,0.509804}%
\pgfsetstrokecolor{currentstroke}%
\pgfsetdash{}{0pt}%
\pgfsys@defobject{currentmarker}{\pgfqpoint{-0.055556in}{-0.055556in}}{\pgfqpoint{0.055556in}{0.055556in}}{%
\pgfpathmoveto{\pgfqpoint{0.000000in}{-0.055556in}}%
\pgfpathcurveto{\pgfqpoint{0.014734in}{-0.055556in}}{\pgfqpoint{0.028866in}{-0.049702in}}{\pgfqpoint{0.039284in}{-0.039284in}}%
\pgfpathcurveto{\pgfqpoint{0.049702in}{-0.028866in}}{\pgfqpoint{0.055556in}{-0.014734in}}{\pgfqpoint{0.055556in}{0.000000in}}%
\pgfpathcurveto{\pgfqpoint{0.055556in}{0.014734in}}{\pgfqpoint{0.049702in}{0.028866in}}{\pgfqpoint{0.039284in}{0.039284in}}%
\pgfpathcurveto{\pgfqpoint{0.028866in}{0.049702in}}{\pgfqpoint{0.014734in}{0.055556in}}{\pgfqpoint{0.000000in}{0.055556in}}%
\pgfpathcurveto{\pgfqpoint{-0.014734in}{0.055556in}}{\pgfqpoint{-0.028866in}{0.049702in}}{\pgfqpoint{-0.039284in}{0.039284in}}%
\pgfpathcurveto{\pgfqpoint{-0.049702in}{0.028866in}}{\pgfqpoint{-0.055556in}{0.014734in}}{\pgfqpoint{-0.055556in}{0.000000in}}%
\pgfpathcurveto{\pgfqpoint{-0.055556in}{-0.014734in}}{\pgfqpoint{-0.049702in}{-0.028866in}}{\pgfqpoint{-0.039284in}{-0.039284in}}%
\pgfpathcurveto{\pgfqpoint{-0.028866in}{-0.049702in}}{\pgfqpoint{-0.014734in}{-0.055556in}}{\pgfqpoint{0.000000in}{-0.055556in}}%
\pgfpathlineto{\pgfqpoint{0.000000in}{-0.055556in}}%
\pgfpathclose%
\pgfusepath{stroke,fill}%
}%
\begin{pgfscope}%
\pgfsys@transformshift{0.674954in}{5.231386in}%
\pgfsys@useobject{currentmarker}{}%
\end{pgfscope}%
\begin{pgfscope}%
\pgfsys@transformshift{0.998577in}{5.173271in}%
\pgfsys@useobject{currentmarker}{}%
\end{pgfscope}%
\begin{pgfscope}%
\pgfsys@transformshift{2.293069in}{4.986364in}%
\pgfsys@useobject{currentmarker}{}%
\end{pgfscope}%
\begin{pgfscope}%
\pgfsys@transformshift{3.911184in}{4.820665in}%
\pgfsys@useobject{currentmarker}{}%
\end{pgfscope}%
\begin{pgfscope}%
\pgfsys@transformshift{5.529299in}{4.676434in}%
\pgfsys@useobject{currentmarker}{}%
\end{pgfscope}%
\begin{pgfscope}%
\pgfsys@transformshift{6.823791in}{4.803713in}%
\pgfsys@useobject{currentmarker}{}%
\end{pgfscope}%
\begin{pgfscope}%
\pgfsys@transformshift{7.147414in}{4.961556in}%
\pgfsys@useobject{currentmarker}{}%
\end{pgfscope}%
\end{pgfscope}%
\begin{pgfscope}%
\pgfpathrectangle{\pgfqpoint{0.674954in}{0.862305in}}{\pgfqpoint{6.472460in}{4.369081in}}%
\pgfusepath{clip}%
\pgfsetbuttcap%
\pgfsetroundjoin%
\pgfsetlinewidth{2.509375pt}%
\definecolor{currentstroke}{rgb}{0.960784,0.462745,0.000000}%
\pgfsetstrokecolor{currentstroke}%
\pgfsetdash{{9.250000pt}{4.000000pt}}{0.000000pt}%
\pgfpathmoveto{\pgfqpoint{0.674954in}{5.231386in}}%
\pgfpathlineto{\pgfqpoint{0.998577in}{5.169106in}}%
\pgfpathlineto{\pgfqpoint{2.293069in}{4.940364in}}%
\pgfpathlineto{\pgfqpoint{3.911184in}{4.711241in}}%
\pgfpathlineto{\pgfqpoint{5.529299in}{4.470362in}}%
\pgfpathlineto{\pgfqpoint{6.823791in}{4.550383in}}%
\pgfpathlineto{\pgfqpoint{7.147414in}{4.741089in}}%
\pgfusepath{stroke}%
\end{pgfscope}%
\begin{pgfscope}%
\pgfpathrectangle{\pgfqpoint{0.674954in}{0.862305in}}{\pgfqpoint{6.472460in}{4.369081in}}%
\pgfusepath{clip}%
\pgfsetbuttcap%
\pgfsetmiterjoin%
\definecolor{currentfill}{rgb}{0.960784,0.462745,0.000000}%
\pgfsetfillcolor{currentfill}%
\pgfsetlinewidth{1.003750pt}%
\definecolor{currentstroke}{rgb}{0.960784,0.462745,0.000000}%
\pgfsetstrokecolor{currentstroke}%
\pgfsetdash{}{0pt}%
\pgfsys@defobject{currentmarker}{\pgfqpoint{-0.055556in}{-0.055556in}}{\pgfqpoint{0.055556in}{0.055556in}}{%
\pgfpathmoveto{\pgfqpoint{-0.055556in}{-0.055556in}}%
\pgfpathlineto{\pgfqpoint{0.055556in}{-0.055556in}}%
\pgfpathlineto{\pgfqpoint{0.055556in}{0.055556in}}%
\pgfpathlineto{\pgfqpoint{-0.055556in}{0.055556in}}%
\pgfpathlineto{\pgfqpoint{-0.055556in}{-0.055556in}}%
\pgfpathclose%
\pgfusepath{stroke,fill}%
}%
\begin{pgfscope}%
\pgfsys@transformshift{0.674954in}{5.231386in}%
\pgfsys@useobject{currentmarker}{}%
\end{pgfscope}%
\begin{pgfscope}%
\pgfsys@transformshift{0.998577in}{5.169106in}%
\pgfsys@useobject{currentmarker}{}%
\end{pgfscope}%
\begin{pgfscope}%
\pgfsys@transformshift{2.293069in}{4.940364in}%
\pgfsys@useobject{currentmarker}{}%
\end{pgfscope}%
\begin{pgfscope}%
\pgfsys@transformshift{3.911184in}{4.711241in}%
\pgfsys@useobject{currentmarker}{}%
\end{pgfscope}%
\begin{pgfscope}%
\pgfsys@transformshift{5.529299in}{4.470362in}%
\pgfsys@useobject{currentmarker}{}%
\end{pgfscope}%
\begin{pgfscope}%
\pgfsys@transformshift{6.823791in}{4.550383in}%
\pgfsys@useobject{currentmarker}{}%
\end{pgfscope}%
\begin{pgfscope}%
\pgfsys@transformshift{7.147414in}{4.741089in}%
\pgfsys@useobject{currentmarker}{}%
\end{pgfscope}%
\end{pgfscope}%
\begin{pgfscope}%
\pgfpathrectangle{\pgfqpoint{0.674954in}{0.862305in}}{\pgfqpoint{6.472460in}{4.369081in}}%
\pgfusepath{clip}%
\pgfsetbuttcap%
\pgfsetroundjoin%
\pgfsetlinewidth{2.509375pt}%
\definecolor{currentstroke}{rgb}{0.219608,0.219608,0.219608}%
\pgfsetstrokecolor{currentstroke}%
\pgfsetdash{{2.500000pt}{4.125000pt}}{0.000000pt}%
\pgfpathmoveto{\pgfqpoint{0.674954in}{5.221667in}}%
\pgfpathlineto{\pgfqpoint{0.998577in}{5.089347in}}%
\pgfpathlineto{\pgfqpoint{2.293069in}{4.617156in}}%
\pgfpathlineto{\pgfqpoint{3.911184in}{4.288028in}}%
\pgfpathlineto{\pgfqpoint{5.529299in}{4.052222in}}%
\pgfpathlineto{\pgfqpoint{6.823791in}{4.125488in}}%
\pgfpathlineto{\pgfqpoint{7.147414in}{4.185641in}}%
\pgfusepath{stroke}%
\end{pgfscope}%
\begin{pgfscope}%
\pgfpathrectangle{\pgfqpoint{0.674954in}{0.862305in}}{\pgfqpoint{6.472460in}{4.369081in}}%
\pgfusepath{clip}%
\pgfsetbuttcap%
\pgfsetmiterjoin%
\definecolor{currentfill}{rgb}{0.219608,0.219608,0.219608}%
\pgfsetfillcolor{currentfill}%
\pgfsetlinewidth{1.003750pt}%
\definecolor{currentstroke}{rgb}{0.219608,0.219608,0.219608}%
\pgfsetstrokecolor{currentstroke}%
\pgfsetdash{}{0pt}%
\pgfsys@defobject{currentmarker}{\pgfqpoint{-0.055556in}{-0.055556in}}{\pgfqpoint{0.055556in}{0.055556in}}{%
\pgfpathmoveto{\pgfqpoint{0.000000in}{0.055556in}}%
\pgfpathlineto{\pgfqpoint{-0.055556in}{-0.055556in}}%
\pgfpathlineto{\pgfqpoint{0.055556in}{-0.055556in}}%
\pgfpathlineto{\pgfqpoint{0.000000in}{0.055556in}}%
\pgfpathclose%
\pgfusepath{stroke,fill}%
}%
\begin{pgfscope}%
\pgfsys@transformshift{0.674954in}{5.221667in}%
\pgfsys@useobject{currentmarker}{}%
\end{pgfscope}%
\begin{pgfscope}%
\pgfsys@transformshift{0.998577in}{5.089347in}%
\pgfsys@useobject{currentmarker}{}%
\end{pgfscope}%
\begin{pgfscope}%
\pgfsys@transformshift{2.293069in}{4.617156in}%
\pgfsys@useobject{currentmarker}{}%
\end{pgfscope}%
\begin{pgfscope}%
\pgfsys@transformshift{3.911184in}{4.288028in}%
\pgfsys@useobject{currentmarker}{}%
\end{pgfscope}%
\begin{pgfscope}%
\pgfsys@transformshift{5.529299in}{4.052222in}%
\pgfsys@useobject{currentmarker}{}%
\end{pgfscope}%
\begin{pgfscope}%
\pgfsys@transformshift{6.823791in}{4.125488in}%
\pgfsys@useobject{currentmarker}{}%
\end{pgfscope}%
\begin{pgfscope}%
\pgfsys@transformshift{7.147414in}{4.185641in}%
\pgfsys@useobject{currentmarker}{}%
\end{pgfscope}%
\end{pgfscope}%
\begin{pgfscope}%
\pgfsetrectcap%
\pgfsetmiterjoin%
\pgfsetlinewidth{2.007500pt}%
\definecolor{currentstroke}{rgb}{0.000000,0.000000,0.000000}%
\pgfsetstrokecolor{currentstroke}%
\pgfsetdash{}{0pt}%
\pgfpathmoveto{\pgfqpoint{0.674954in}{0.862305in}}%
\pgfpathlineto{\pgfqpoint{0.674954in}{5.231386in}}%
\pgfusepath{stroke}%
\end{pgfscope}%
\begin{pgfscope}%
\pgfsetrectcap%
\pgfsetmiterjoin%
\pgfsetlinewidth{2.007500pt}%
\definecolor{currentstroke}{rgb}{0.000000,0.000000,0.000000}%
\pgfsetstrokecolor{currentstroke}%
\pgfsetdash{}{0pt}%
\pgfpathmoveto{\pgfqpoint{0.674954in}{0.862305in}}%
\pgfpathlineto{\pgfqpoint{7.147414in}{0.862305in}}%
\pgfusepath{stroke}%
\end{pgfscope}%
\begin{pgfscope}%
\pgfsetbuttcap%
\pgfsetmiterjoin%
\definecolor{currentfill}{rgb}{1.000000,1.000000,1.000000}%
\pgfsetfillcolor{currentfill}%
\pgfsetlinewidth{1.003750pt}%
\definecolor{currentstroke}{rgb}{0.800000,0.800000,0.800000}%
\pgfsetstrokecolor{currentstroke}%
\pgfsetdash{}{0pt}%
\pgfpathmoveto{\pgfqpoint{0.869398in}{1.001194in}}%
\pgfpathlineto{\pgfqpoint{2.213036in}{1.001194in}}%
\pgfpathquadraticcurveto{\pgfqpoint{2.268591in}{1.001194in}}{\pgfqpoint{2.268591in}{1.056749in}}%
\pgfpathlineto{\pgfqpoint{2.268591in}{2.191056in}}%
\pgfpathquadraticcurveto{\pgfqpoint{2.268591in}{2.246612in}}{\pgfqpoint{2.213036in}{2.246612in}}%
\pgfpathlineto{\pgfqpoint{0.869398in}{2.246612in}}%
\pgfpathquadraticcurveto{\pgfqpoint{0.813843in}{2.246612in}}{\pgfqpoint{0.813843in}{2.191056in}}%
\pgfpathlineto{\pgfqpoint{0.813843in}{1.056749in}}%
\pgfpathquadraticcurveto{\pgfqpoint{0.813843in}{1.001194in}}{\pgfqpoint{0.869398in}{1.001194in}}%
\pgfpathlineto{\pgfqpoint{0.869398in}{1.001194in}}%
\pgfpathclose%
\pgfusepath{stroke,fill}%
\end{pgfscope}%
\begin{pgfscope}%
\pgfsetrectcap%
\pgfsetroundjoin%
\pgfsetlinewidth{2.509375pt}%
\definecolor{currentstroke}{rgb}{0.050980,0.415686,0.509804}%
\pgfsetstrokecolor{currentstroke}%
\pgfsetdash{}{0pt}%
\pgfpathmoveto{\pgfqpoint{0.924954in}{2.038278in}}%
\pgfpathlineto{\pgfqpoint{1.202732in}{2.038278in}}%
\pgfpathlineto{\pgfqpoint{1.480509in}{2.038278in}}%
\pgfusepath{stroke}%
\end{pgfscope}%
\begin{pgfscope}%
\pgfsetbuttcap%
\pgfsetroundjoin%
\definecolor{currentfill}{rgb}{0.050980,0.415686,0.509804}%
\pgfsetfillcolor{currentfill}%
\pgfsetlinewidth{1.003750pt}%
\definecolor{currentstroke}{rgb}{0.050980,0.415686,0.509804}%
\pgfsetstrokecolor{currentstroke}%
\pgfsetdash{}{0pt}%
\pgfsys@defobject{currentmarker}{\pgfqpoint{-0.055556in}{-0.055556in}}{\pgfqpoint{0.055556in}{0.055556in}}{%
\pgfpathmoveto{\pgfqpoint{0.000000in}{-0.055556in}}%
\pgfpathcurveto{\pgfqpoint{0.014734in}{-0.055556in}}{\pgfqpoint{0.028866in}{-0.049702in}}{\pgfqpoint{0.039284in}{-0.039284in}}%
\pgfpathcurveto{\pgfqpoint{0.049702in}{-0.028866in}}{\pgfqpoint{0.055556in}{-0.014734in}}{\pgfqpoint{0.055556in}{0.000000in}}%
\pgfpathcurveto{\pgfqpoint{0.055556in}{0.014734in}}{\pgfqpoint{0.049702in}{0.028866in}}{\pgfqpoint{0.039284in}{0.039284in}}%
\pgfpathcurveto{\pgfqpoint{0.028866in}{0.049702in}}{\pgfqpoint{0.014734in}{0.055556in}}{\pgfqpoint{0.000000in}{0.055556in}}%
\pgfpathcurveto{\pgfqpoint{-0.014734in}{0.055556in}}{\pgfqpoint{-0.028866in}{0.049702in}}{\pgfqpoint{-0.039284in}{0.039284in}}%
\pgfpathcurveto{\pgfqpoint{-0.049702in}{0.028866in}}{\pgfqpoint{-0.055556in}{0.014734in}}{\pgfqpoint{-0.055556in}{0.000000in}}%
\pgfpathcurveto{\pgfqpoint{-0.055556in}{-0.014734in}}{\pgfqpoint{-0.049702in}{-0.028866in}}{\pgfqpoint{-0.039284in}{-0.039284in}}%
\pgfpathcurveto{\pgfqpoint{-0.028866in}{-0.049702in}}{\pgfqpoint{-0.014734in}{-0.055556in}}{\pgfqpoint{0.000000in}{-0.055556in}}%
\pgfpathlineto{\pgfqpoint{0.000000in}{-0.055556in}}%
\pgfpathclose%
\pgfusepath{stroke,fill}%
}%
\begin{pgfscope}%
\pgfsys@transformshift{1.202732in}{2.038278in}%
\pgfsys@useobject{currentmarker}{}%
\end{pgfscope}%
\end{pgfscope}%
\begin{pgfscope}%
\definecolor{textcolor}{rgb}{0.000000,0.000000,0.000000}%
\pgfsetstrokecolor{textcolor}%
\pgfsetfillcolor{textcolor}%
\pgftext[x=1.702732in,y=1.941056in,left,base]{\color{textcolor}{\rmfamily\fontsize{20.000000}{24.000000}\selectfont\catcode`\^=\active\def^{\ifmmode\sp\else\^{}\fi}\catcode`\%=\active\def
\end{pgfscope}%
\begin{pgfscope}%
\pgfsetbuttcap%
\pgfsetroundjoin%
\pgfsetlinewidth{2.509375pt}%
\definecolor{currentstroke}{rgb}{0.960784,0.462745,0.000000}%
\pgfsetstrokecolor{currentstroke}%
\pgfsetdash{{9.250000pt}{4.000000pt}}{0.000000pt}%
\pgfpathmoveto{\pgfqpoint{0.924954in}{1.650917in}}%
\pgfpathlineto{\pgfqpoint{1.202732in}{1.650917in}}%
\pgfpathlineto{\pgfqpoint{1.480509in}{1.650917in}}%
\pgfusepath{stroke}%
\end{pgfscope}%
\begin{pgfscope}%
\pgfsetbuttcap%
\pgfsetmiterjoin%
\definecolor{currentfill}{rgb}{0.960784,0.462745,0.000000}%
\pgfsetfillcolor{currentfill}%
\pgfsetlinewidth{1.003750pt}%
\definecolor{currentstroke}{rgb}{0.960784,0.462745,0.000000}%
\pgfsetstrokecolor{currentstroke}%
\pgfsetdash{}{0pt}%
\pgfsys@defobject{currentmarker}{\pgfqpoint{-0.055556in}{-0.055556in}}{\pgfqpoint{0.055556in}{0.055556in}}{%
\pgfpathmoveto{\pgfqpoint{-0.055556in}{-0.055556in}}%
\pgfpathlineto{\pgfqpoint{0.055556in}{-0.055556in}}%
\pgfpathlineto{\pgfqpoint{0.055556in}{0.055556in}}%
\pgfpathlineto{\pgfqpoint{-0.055556in}{0.055556in}}%
\pgfpathlineto{\pgfqpoint{-0.055556in}{-0.055556in}}%
\pgfpathclose%
\pgfusepath{stroke,fill}%
}%
\begin{pgfscope}%
\pgfsys@transformshift{1.202732in}{1.650917in}%
\pgfsys@useobject{currentmarker}{}%
\end{pgfscope}%
\end{pgfscope}%
\begin{pgfscope}%
\definecolor{textcolor}{rgb}{0.000000,0.000000,0.000000}%
\pgfsetstrokecolor{textcolor}%
\pgfsetfillcolor{textcolor}%
\pgftext[x=1.702732in,y=1.553694in,left,base]{\color{textcolor}{\rmfamily\fontsize{20.000000}{24.000000}\selectfont\catcode`\^=\active\def^{\ifmmode\sp\else\^{}\fi}\catcode`\%=\active\def
\end{pgfscope}%
\begin{pgfscope}%
\pgfsetbuttcap%
\pgfsetroundjoin%
\pgfsetlinewidth{2.509375pt}%
\definecolor{currentstroke}{rgb}{0.219608,0.219608,0.219608}%
\pgfsetstrokecolor{currentstroke}%
\pgfsetdash{{2.500000pt}{4.125000pt}}{0.000000pt}%
\pgfpathmoveto{\pgfqpoint{0.924954in}{1.263555in}}%
\pgfpathlineto{\pgfqpoint{1.202732in}{1.263555in}}%
\pgfpathlineto{\pgfqpoint{1.480509in}{1.263555in}}%
\pgfusepath{stroke}%
\end{pgfscope}%
\begin{pgfscope}%
\pgfsetbuttcap%
\pgfsetmiterjoin%
\definecolor{currentfill}{rgb}{0.219608,0.219608,0.219608}%
\pgfsetfillcolor{currentfill}%
\pgfsetlinewidth{1.003750pt}%
\definecolor{currentstroke}{rgb}{0.219608,0.219608,0.219608}%
\pgfsetstrokecolor{currentstroke}%
\pgfsetdash{}{0pt}%
\pgfsys@defobject{currentmarker}{\pgfqpoint{-0.055556in}{-0.055556in}}{\pgfqpoint{0.055556in}{0.055556in}}{%
\pgfpathmoveto{\pgfqpoint{0.000000in}{0.055556in}}%
\pgfpathlineto{\pgfqpoint{-0.055556in}{-0.055556in}}%
\pgfpathlineto{\pgfqpoint{0.055556in}{-0.055556in}}%
\pgfpathlineto{\pgfqpoint{0.000000in}{0.055556in}}%
\pgfpathclose%
\pgfusepath{stroke,fill}%
}%
\begin{pgfscope}%
\pgfsys@transformshift{1.202732in}{1.263555in}%
\pgfsys@useobject{currentmarker}{}%
\end{pgfscope}%
\end{pgfscope}%
\begin{pgfscope}%
\definecolor{textcolor}{rgb}{0.000000,0.000000,0.000000}%
\pgfsetstrokecolor{textcolor}%
\pgfsetfillcolor{textcolor}%
\pgftext[x=1.702732in,y=1.166333in,left,base]{\color{textcolor}{\rmfamily\fontsize{20.000000}{24.000000}\selectfont\catcode`\^=\active\def^{\ifmmode\sp\else\^{}\fi}\catcode`\%=\active\def
\end{pgfscope}%
\end{pgfpicture}%
\makeatother%
\endgroup%

%% file: images/graphs/thermal_ratio_camera_token_RTA.pgf
\begingroup%
\makeatletter%
\begin{pgfpicture}%
\pgfpathrectangle{\pgfpointorigin}{\pgfqpoint{7.450000in}{5.450000in}}%
\pgfusepath{use as bounding box, clip}%
\begin{pgfscope}%
\pgfsetbuttcap%
\pgfsetmiterjoin%
\definecolor{currentfill}{rgb}{1.000000,1.000000,1.000000}%
\pgfsetfillcolor{currentfill}%
\pgfsetlinewidth{0.000000pt}%
\definecolor{currentstroke}{rgb}{1.000000,1.000000,1.000000}%
\pgfsetstrokecolor{currentstroke}%
\pgfsetdash{}{0pt}%
\pgfpathmoveto{\pgfqpoint{0.000000in}{0.000000in}}%
\pgfpathlineto{\pgfqpoint{7.450000in}{0.000000in}}%
\pgfpathlineto{\pgfqpoint{7.450000in}{5.450000in}}%
\pgfpathlineto{\pgfqpoint{0.000000in}{5.450000in}}%
\pgfpathlineto{\pgfqpoint{0.000000in}{0.000000in}}%
\pgfpathclose%
\pgfusepath{fill}%
\end{pgfscope}%
\begin{pgfscope}%
\pgfsetbuttcap%
\pgfsetmiterjoin%
\definecolor{currentfill}{rgb}{1.000000,1.000000,1.000000}%
\pgfsetfillcolor{currentfill}%
\pgfsetlinewidth{0.000000pt}%
\definecolor{currentstroke}{rgb}{0.000000,0.000000,0.000000}%
\pgfsetstrokecolor{currentstroke}%
\pgfsetstrokeopacity{0.000000}%
\pgfsetdash{}{0pt}%
\pgfpathmoveto{\pgfqpoint{0.674954in}{0.862305in}}%
\pgfpathlineto{\pgfqpoint{7.147414in}{0.862305in}}%
\pgfpathlineto{\pgfqpoint{7.147414in}{5.231386in}}%
\pgfpathlineto{\pgfqpoint{0.674954in}{5.231386in}}%
\pgfpathlineto{\pgfqpoint{0.674954in}{0.862305in}}%
\pgfpathclose%
\pgfusepath{fill}%
\end{pgfscope}%
\begin{pgfscope}%
\pgfpathrectangle{\pgfqpoint{0.674954in}{0.862305in}}{\pgfqpoint{6.472460in}{4.369081in}}%
\pgfusepath{clip}%
\pgfsetbuttcap%
\pgfsetroundjoin%
\definecolor{currentfill}{rgb}{0.050980,0.415686,0.509804}%
\pgfsetfillcolor{currentfill}%
\pgfsetfillopacity{0.300000}%
\pgfsetlinewidth{1.003750pt}%
\definecolor{currentstroke}{rgb}{0.050980,0.415686,0.509804}%
\pgfsetstrokecolor{currentstroke}%
\pgfsetstrokeopacity{0.300000}%
\pgfsetdash{}{0pt}%
\pgfsys@defobject{currentmarker}{\pgfqpoint{0.674954in}{4.355929in}}{\pgfqpoint{7.147414in}{5.179283in}}{%
\pgfpathmoveto{\pgfqpoint{0.674954in}{5.179283in}}%
\pgfpathlineto{\pgfqpoint{0.674954in}{5.120460in}}%
\pgfpathlineto{\pgfqpoint{0.998577in}{5.045897in}}%
\pgfpathlineto{\pgfqpoint{2.293069in}{4.804924in}}%
\pgfpathlineto{\pgfqpoint{3.911184in}{4.570476in}}%
\pgfpathlineto{\pgfqpoint{5.529299in}{4.355929in}}%
\pgfpathlineto{\pgfqpoint{6.823791in}{4.564004in}}%
\pgfpathlineto{\pgfqpoint{7.147414in}{4.704409in}}%
\pgfpathlineto{\pgfqpoint{7.147414in}{4.991110in}}%
\pgfpathlineto{\pgfqpoint{7.147414in}{4.991110in}}%
\pgfpathlineto{\pgfqpoint{6.823791in}{4.877753in}}%
\pgfpathlineto{\pgfqpoint{5.529299in}{4.717961in}}%
\pgfpathlineto{\pgfqpoint{3.911184in}{4.884136in}}%
\pgfpathlineto{\pgfqpoint{2.293069in}{5.009134in}}%
\pgfpathlineto{\pgfqpoint{0.998577in}{5.125073in}}%
\pgfpathlineto{\pgfqpoint{0.674954in}{5.179283in}}%
\pgfpathlineto{\pgfqpoint{0.674954in}{5.179283in}}%
\pgfpathclose%
\pgfusepath{stroke,fill}%
}%
\begin{pgfscope}%
\pgfsys@transformshift{0.000000in}{0.000000in}%
\pgfsys@useobject{currentmarker}{}%
\end{pgfscope}%
\end{pgfscope}%
\begin{pgfscope}%
\pgfpathrectangle{\pgfqpoint{0.674954in}{0.862305in}}{\pgfqpoint{6.472460in}{4.369081in}}%
\pgfusepath{clip}%
\pgfsetbuttcap%
\pgfsetroundjoin%
\definecolor{currentfill}{rgb}{0.960784,0.462745,0.000000}%
\pgfsetfillcolor{currentfill}%
\pgfsetfillopacity{0.300000}%
\pgfsetlinewidth{1.003750pt}%
\definecolor{currentstroke}{rgb}{0.960784,0.462745,0.000000}%
\pgfsetstrokecolor{currentstroke}%
\pgfsetstrokeopacity{0.300000}%
\pgfsetdash{{3.700000pt}{1.600000pt}}{0.000000pt}%
\pgfpathmoveto{\pgfqpoint{0.674954in}{5.100176in}}%
\pgfpathlineto{\pgfqpoint{0.674954in}{4.990979in}}%
\pgfpathlineto{\pgfqpoint{0.998577in}{4.884697in}}%
\pgfpathlineto{\pgfqpoint{2.293069in}{4.511614in}}%
\pgfpathlineto{\pgfqpoint{3.911184in}{4.152195in}}%
\pgfpathlineto{\pgfqpoint{5.529299in}{3.872591in}}%
\pgfpathlineto{\pgfqpoint{6.823791in}{4.068190in}}%
\pgfpathlineto{\pgfqpoint{7.147414in}{4.231060in}}%
\pgfpathlineto{\pgfqpoint{7.147414in}{4.669700in}}%
\pgfpathlineto{\pgfqpoint{7.147414in}{4.669700in}}%
\pgfpathlineto{\pgfqpoint{6.823791in}{4.513437in}}%
\pgfpathlineto{\pgfqpoint{5.529299in}{4.353225in}}%
\pgfpathlineto{\pgfqpoint{3.911184in}{4.552225in}}%
\pgfpathlineto{\pgfqpoint{2.293069in}{4.812247in}}%
\pgfpathlineto{\pgfqpoint{0.998577in}{5.024024in}}%
\pgfpathlineto{\pgfqpoint{0.674954in}{5.100176in}}%
\pgfpathlineto{\pgfqpoint{0.674954in}{5.100176in}}%
\pgfpathclose%
\pgfusepath{stroke,fill}%
\end{pgfscope}%
\begin{pgfscope}%
\pgfpathrectangle{\pgfqpoint{0.674954in}{0.862305in}}{\pgfqpoint{6.472460in}{4.369081in}}%
\pgfusepath{clip}%
\pgfsetbuttcap%
\pgfsetroundjoin%
\definecolor{currentfill}{rgb}{0.219608,0.219608,0.219608}%
\pgfsetfillcolor{currentfill}%
\pgfsetfillopacity{0.300000}%
\pgfsetlinewidth{1.003750pt}%
\definecolor{currentstroke}{rgb}{0.219608,0.219608,0.219608}%
\pgfsetstrokecolor{currentstroke}%
\pgfsetstrokeopacity{0.300000}%
\pgfsetdash{{1.000000pt}{1.650000pt}}{0.000000pt}%
\pgfpathmoveto{\pgfqpoint{0.674954in}{4.672202in}}%
\pgfpathlineto{\pgfqpoint{0.674954in}{4.327450in}}%
\pgfpathlineto{\pgfqpoint{0.998577in}{4.176742in}}%
\pgfpathlineto{\pgfqpoint{2.293069in}{3.707791in}}%
\pgfpathlineto{\pgfqpoint{3.911184in}{3.348569in}}%
\pgfpathlineto{\pgfqpoint{5.529299in}{3.141343in}}%
\pgfpathlineto{\pgfqpoint{6.823791in}{3.296818in}}%
\pgfpathlineto{\pgfqpoint{7.147414in}{3.395849in}}%
\pgfpathlineto{\pgfqpoint{7.147414in}{3.912507in}}%
\pgfpathlineto{\pgfqpoint{7.147414in}{3.912507in}}%
\pgfpathlineto{\pgfqpoint{6.823791in}{3.808742in}}%
\pgfpathlineto{\pgfqpoint{5.529299in}{3.662347in}}%
\pgfpathlineto{\pgfqpoint{3.911184in}{3.806573in}}%
\pgfpathlineto{\pgfqpoint{2.293069in}{4.113045in}}%
\pgfpathlineto{\pgfqpoint{0.998577in}{4.523731in}}%
\pgfpathlineto{\pgfqpoint{0.674954in}{4.672202in}}%
\pgfpathlineto{\pgfqpoint{0.674954in}{4.672202in}}%
\pgfpathclose%
\pgfusepath{stroke,fill}%
\end{pgfscope}%
\begin{pgfscope}%
\pgfpathrectangle{\pgfqpoint{0.674954in}{0.862305in}}{\pgfqpoint{6.472460in}{4.369081in}}%
\pgfusepath{clip}%
\pgfsetbuttcap%
\pgfsetroundjoin%
\pgfsetlinewidth{2.007500pt}%
\definecolor{currentstroke}{rgb}{0.501961,0.501961,0.501961}%
\pgfsetstrokecolor{currentstroke}%
\pgfsetstrokeopacity{0.300000}%
\pgfsetdash{{7.400000pt}{3.200000pt}}{0.000000pt}%
\pgfpathmoveto{\pgfqpoint{0.674954in}{0.862305in}}%
\pgfpathlineto{\pgfqpoint{0.674954in}{5.231386in}}%
\pgfusepath{stroke}%
\end{pgfscope}%
\begin{pgfscope}%
\pgfsetbuttcap%
\pgfsetroundjoin%
\definecolor{currentfill}{rgb}{0.000000,0.000000,0.000000}%
\pgfsetfillcolor{currentfill}%
\pgfsetlinewidth{0.803000pt}%
\definecolor{currentstroke}{rgb}{0.000000,0.000000,0.000000}%
\pgfsetstrokecolor{currentstroke}%
\pgfsetdash{}{0pt}%
\pgfsys@defobject{currentmarker}{\pgfqpoint{0.000000in}{-0.048611in}}{\pgfqpoint{0.000000in}{0.000000in}}{%
\pgfpathmoveto{\pgfqpoint{0.000000in}{0.000000in}}%
\pgfpathlineto{\pgfqpoint{0.000000in}{-0.048611in}}%
\pgfusepath{stroke,fill}%
}%
\begin{pgfscope}%
\pgfsys@transformshift{0.674954in}{0.862305in}%
\pgfsys@useobject{currentmarker}{}%
\end{pgfscope}%
\end{pgfscope}%
\begin{pgfscope}%
\definecolor{textcolor}{rgb}{0.000000,0.000000,0.000000}%
\pgfsetstrokecolor{textcolor}%
\pgfsetfillcolor{textcolor}%
\pgftext[x=0.674954in,y=0.765082in,,top]{\color{textcolor}{\rmfamily\fontsize{25.000000}{30.000000}\selectfont\catcode`\^=\active\def^{\ifmmode\sp\else\^{}\fi}\catcode`\%=\active\def
\end{pgfscope}%
\begin{pgfscope}%
\pgfpathrectangle{\pgfqpoint{0.674954in}{0.862305in}}{\pgfqpoint{6.472460in}{4.369081in}}%
\pgfusepath{clip}%
\pgfsetbuttcap%
\pgfsetroundjoin%
\pgfsetlinewidth{2.007500pt}%
\definecolor{currentstroke}{rgb}{0.501961,0.501961,0.501961}%
\pgfsetstrokecolor{currentstroke}%
\pgfsetstrokeopacity{0.300000}%
\pgfsetdash{{7.400000pt}{3.200000pt}}{0.000000pt}%
\pgfpathmoveto{\pgfqpoint{1.969446in}{0.862305in}}%
\pgfpathlineto{\pgfqpoint{1.969446in}{5.231386in}}%
\pgfusepath{stroke}%
\end{pgfscope}%
\begin{pgfscope}%
\pgfsetbuttcap%
\pgfsetroundjoin%
\definecolor{currentfill}{rgb}{0.000000,0.000000,0.000000}%
\pgfsetfillcolor{currentfill}%
\pgfsetlinewidth{0.803000pt}%
\definecolor{currentstroke}{rgb}{0.000000,0.000000,0.000000}%
\pgfsetstrokecolor{currentstroke}%
\pgfsetdash{}{0pt}%
\pgfsys@defobject{currentmarker}{\pgfqpoint{0.000000in}{-0.048611in}}{\pgfqpoint{0.000000in}{0.000000in}}{%
\pgfpathmoveto{\pgfqpoint{0.000000in}{0.000000in}}%
\pgfpathlineto{\pgfqpoint{0.000000in}{-0.048611in}}%
\pgfusepath{stroke,fill}%
}%
\begin{pgfscope}%
\pgfsys@transformshift{1.969446in}{0.862305in}%
\pgfsys@useobject{currentmarker}{}%
\end{pgfscope}%
\end{pgfscope}%
\begin{pgfscope}%
\definecolor{textcolor}{rgb}{0.000000,0.000000,0.000000}%
\pgfsetstrokecolor{textcolor}%
\pgfsetfillcolor{textcolor}%
\pgftext[x=1.969446in,y=0.765082in,,top]{\color{textcolor}{\rmfamily\fontsize{25.000000}{30.000000}\selectfont\catcode`\^=\active\def^{\ifmmode\sp\else\^{}\fi}\catcode`\%=\active\def
\end{pgfscope}%
\begin{pgfscope}%
\pgfpathrectangle{\pgfqpoint{0.674954in}{0.862305in}}{\pgfqpoint{6.472460in}{4.369081in}}%
\pgfusepath{clip}%
\pgfsetbuttcap%
\pgfsetroundjoin%
\pgfsetlinewidth{2.007500pt}%
\definecolor{currentstroke}{rgb}{0.501961,0.501961,0.501961}%
\pgfsetstrokecolor{currentstroke}%
\pgfsetstrokeopacity{0.300000}%
\pgfsetdash{{7.400000pt}{3.200000pt}}{0.000000pt}%
\pgfpathmoveto{\pgfqpoint{3.263938in}{0.862305in}}%
\pgfpathlineto{\pgfqpoint{3.263938in}{5.231386in}}%
\pgfusepath{stroke}%
\end{pgfscope}%
\begin{pgfscope}%
\pgfsetbuttcap%
\pgfsetroundjoin%
\definecolor{currentfill}{rgb}{0.000000,0.000000,0.000000}%
\pgfsetfillcolor{currentfill}%
\pgfsetlinewidth{0.803000pt}%
\definecolor{currentstroke}{rgb}{0.000000,0.000000,0.000000}%
\pgfsetstrokecolor{currentstroke}%
\pgfsetdash{}{0pt}%
\pgfsys@defobject{currentmarker}{\pgfqpoint{0.000000in}{-0.048611in}}{\pgfqpoint{0.000000in}{0.000000in}}{%
\pgfpathmoveto{\pgfqpoint{0.000000in}{0.000000in}}%
\pgfpathlineto{\pgfqpoint{0.000000in}{-0.048611in}}%
\pgfusepath{stroke,fill}%
}%
\begin{pgfscope}%
\pgfsys@transformshift{3.263938in}{0.862305in}%
\pgfsys@useobject{currentmarker}{}%
\end{pgfscope}%
\end{pgfscope}%
\begin{pgfscope}%
\definecolor{textcolor}{rgb}{0.000000,0.000000,0.000000}%
\pgfsetstrokecolor{textcolor}%
\pgfsetfillcolor{textcolor}%
\pgftext[x=3.263938in,y=0.765082in,,top]{\color{textcolor}{\rmfamily\fontsize{25.000000}{30.000000}\selectfont\catcode`\^=\active\def^{\ifmmode\sp\else\^{}\fi}\catcode`\%=\active\def
\end{pgfscope}%
\begin{pgfscope}%
\pgfpathrectangle{\pgfqpoint{0.674954in}{0.862305in}}{\pgfqpoint{6.472460in}{4.369081in}}%
\pgfusepath{clip}%
\pgfsetbuttcap%
\pgfsetroundjoin%
\pgfsetlinewidth{2.007500pt}%
\definecolor{currentstroke}{rgb}{0.501961,0.501961,0.501961}%
\pgfsetstrokecolor{currentstroke}%
\pgfsetstrokeopacity{0.300000}%
\pgfsetdash{{7.400000pt}{3.200000pt}}{0.000000pt}%
\pgfpathmoveto{\pgfqpoint{4.558430in}{0.862305in}}%
\pgfpathlineto{\pgfqpoint{4.558430in}{5.231386in}}%
\pgfusepath{stroke}%
\end{pgfscope}%
\begin{pgfscope}%
\pgfsetbuttcap%
\pgfsetroundjoin%
\definecolor{currentfill}{rgb}{0.000000,0.000000,0.000000}%
\pgfsetfillcolor{currentfill}%
\pgfsetlinewidth{0.803000pt}%
\definecolor{currentstroke}{rgb}{0.000000,0.000000,0.000000}%
\pgfsetstrokecolor{currentstroke}%
\pgfsetdash{}{0pt}%
\pgfsys@defobject{currentmarker}{\pgfqpoint{0.000000in}{-0.048611in}}{\pgfqpoint{0.000000in}{0.000000in}}{%
\pgfpathmoveto{\pgfqpoint{0.000000in}{0.000000in}}%
\pgfpathlineto{\pgfqpoint{0.000000in}{-0.048611in}}%
\pgfusepath{stroke,fill}%
}%
\begin{pgfscope}%
\pgfsys@transformshift{4.558430in}{0.862305in}%
\pgfsys@useobject{currentmarker}{}%
\end{pgfscope}%
\end{pgfscope}%
\begin{pgfscope}%
\definecolor{textcolor}{rgb}{0.000000,0.000000,0.000000}%
\pgfsetstrokecolor{textcolor}%
\pgfsetfillcolor{textcolor}%
\pgftext[x=4.558430in,y=0.765082in,,top]{\color{textcolor}{\rmfamily\fontsize{25.000000}{30.000000}\selectfont\catcode`\^=\active\def^{\ifmmode\sp\else\^{}\fi}\catcode`\%=\active\def
\end{pgfscope}%
\begin{pgfscope}%
\pgfpathrectangle{\pgfqpoint{0.674954in}{0.862305in}}{\pgfqpoint{6.472460in}{4.369081in}}%
\pgfusepath{clip}%
\pgfsetbuttcap%
\pgfsetroundjoin%
\pgfsetlinewidth{2.007500pt}%
\definecolor{currentstroke}{rgb}{0.501961,0.501961,0.501961}%
\pgfsetstrokecolor{currentstroke}%
\pgfsetstrokeopacity{0.300000}%
\pgfsetdash{{7.400000pt}{3.200000pt}}{0.000000pt}%
\pgfpathmoveto{\pgfqpoint{5.852922in}{0.862305in}}%
\pgfpathlineto{\pgfqpoint{5.852922in}{5.231386in}}%
\pgfusepath{stroke}%
\end{pgfscope}%
\begin{pgfscope}%
\pgfsetbuttcap%
\pgfsetroundjoin%
\definecolor{currentfill}{rgb}{0.000000,0.000000,0.000000}%
\pgfsetfillcolor{currentfill}%
\pgfsetlinewidth{0.803000pt}%
\definecolor{currentstroke}{rgb}{0.000000,0.000000,0.000000}%
\pgfsetstrokecolor{currentstroke}%
\pgfsetdash{}{0pt}%
\pgfsys@defobject{currentmarker}{\pgfqpoint{0.000000in}{-0.048611in}}{\pgfqpoint{0.000000in}{0.000000in}}{%
\pgfpathmoveto{\pgfqpoint{0.000000in}{0.000000in}}%
\pgfpathlineto{\pgfqpoint{0.000000in}{-0.048611in}}%
\pgfusepath{stroke,fill}%
}%
\begin{pgfscope}%
\pgfsys@transformshift{5.852922in}{0.862305in}%
\pgfsys@useobject{currentmarker}{}%
\end{pgfscope}%
\end{pgfscope}%
\begin{pgfscope}%
\definecolor{textcolor}{rgb}{0.000000,0.000000,0.000000}%
\pgfsetstrokecolor{textcolor}%
\pgfsetfillcolor{textcolor}%
\pgftext[x=5.852922in,y=0.765082in,,top]{\color{textcolor}{\rmfamily\fontsize{25.000000}{30.000000}\selectfont\catcode`\^=\active\def^{\ifmmode\sp\else\^{}\fi}\catcode`\%=\active\def
\end{pgfscope}%
\begin{pgfscope}%
\pgfpathrectangle{\pgfqpoint{0.674954in}{0.862305in}}{\pgfqpoint{6.472460in}{4.369081in}}%
\pgfusepath{clip}%
\pgfsetbuttcap%
\pgfsetroundjoin%
\pgfsetlinewidth{2.007500pt}%
\definecolor{currentstroke}{rgb}{0.501961,0.501961,0.501961}%
\pgfsetstrokecolor{currentstroke}%
\pgfsetstrokeopacity{0.300000}%
\pgfsetdash{{7.400000pt}{3.200000pt}}{0.000000pt}%
\pgfpathmoveto{\pgfqpoint{7.147414in}{0.862305in}}%
\pgfpathlineto{\pgfqpoint{7.147414in}{5.231386in}}%
\pgfusepath{stroke}%
\end{pgfscope}%
\begin{pgfscope}%
\pgfsetbuttcap%
\pgfsetroundjoin%
\definecolor{currentfill}{rgb}{0.000000,0.000000,0.000000}%
\pgfsetfillcolor{currentfill}%
\pgfsetlinewidth{0.803000pt}%
\definecolor{currentstroke}{rgb}{0.000000,0.000000,0.000000}%
\pgfsetstrokecolor{currentstroke}%
\pgfsetdash{}{0pt}%
\pgfsys@defobject{currentmarker}{\pgfqpoint{0.000000in}{-0.048611in}}{\pgfqpoint{0.000000in}{0.000000in}}{%
\pgfpathmoveto{\pgfqpoint{0.000000in}{0.000000in}}%
\pgfpathlineto{\pgfqpoint{0.000000in}{-0.048611in}}%
\pgfusepath{stroke,fill}%
}%
\begin{pgfscope}%
\pgfsys@transformshift{7.147414in}{0.862305in}%
\pgfsys@useobject{currentmarker}{}%
\end{pgfscope}%
\end{pgfscope}%
\begin{pgfscope}%
\definecolor{textcolor}{rgb}{0.000000,0.000000,0.000000}%
\pgfsetstrokecolor{textcolor}%
\pgfsetfillcolor{textcolor}%
\pgftext[x=7.147414in,y=0.765082in,,top]{\color{textcolor}{\rmfamily\fontsize{25.000000}{30.000000}\selectfont\catcode`\^=\active\def^{\ifmmode\sp\else\^{}\fi}\catcode`\%=\active\def
\end{pgfscope}%
\begin{pgfscope}%
\definecolor{textcolor}{rgb}{0.000000,0.000000,0.000000}%
\pgfsetstrokecolor{textcolor}%
\pgfsetfillcolor{textcolor}%
\pgftext[x=3.911184in,y=0.404763in,,top]{\color{textcolor}{\rmfamily\fontsize{25.000000}{30.000000}\selectfont\catcode`\^=\active\def^{\ifmmode\sp\else\^{}\fi}\catcode`\%=\active\def
\end{pgfscope}%
\begin{pgfscope}%
\pgfpathrectangle{\pgfqpoint{0.674954in}{0.862305in}}{\pgfqpoint{6.472460in}{4.369081in}}%
\pgfusepath{clip}%
\pgfsetbuttcap%
\pgfsetroundjoin%
\pgfsetlinewidth{2.007500pt}%
\definecolor{currentstroke}{rgb}{0.501961,0.501961,0.501961}%
\pgfsetstrokecolor{currentstroke}%
\pgfsetstrokeopacity{0.300000}%
\pgfsetdash{{7.400000pt}{3.200000pt}}{0.000000pt}%
\pgfpathmoveto{\pgfqpoint{0.674954in}{1.954575in}}%
\pgfpathlineto{\pgfqpoint{7.147414in}{1.954575in}}%
\pgfusepath{stroke}%
\end{pgfscope}%
\begin{pgfscope}%
\pgfsetbuttcap%
\pgfsetroundjoin%
\definecolor{currentfill}{rgb}{0.000000,0.000000,0.000000}%
\pgfsetfillcolor{currentfill}%
\pgfsetlinewidth{0.803000pt}%
\definecolor{currentstroke}{rgb}{0.000000,0.000000,0.000000}%
\pgfsetstrokecolor{currentstroke}%
\pgfsetdash{}{0pt}%
\pgfsys@defobject{currentmarker}{\pgfqpoint{-0.048611in}{0.000000in}}{\pgfqpoint{-0.000000in}{0.000000in}}{%
\pgfpathmoveto{\pgfqpoint{-0.000000in}{0.000000in}}%
\pgfpathlineto{\pgfqpoint{-0.048611in}{0.000000in}}%
\pgfusepath{stroke,fill}%
}%
\begin{pgfscope}%
\pgfsys@transformshift{0.674954in}{1.954575in}%
\pgfsys@useobject{currentmarker}{}%
\end{pgfscope}%
\end{pgfscope}%
\begin{pgfscope}%
\definecolor{textcolor}{rgb}{0.000000,0.000000,0.000000}%
\pgfsetstrokecolor{textcolor}%
\pgfsetfillcolor{textcolor}%
\pgftext[x=0.259244in, y=1.835961in, left, base]{\color{textcolor}{\rmfamily\fontsize{25.000000}{30.000000}\selectfont\catcode`\^=\active\def^{\ifmmode\sp\else\^{}\fi}\catcode`\%=\active\def
\end{pgfscope}%
\begin{pgfscope}%
\pgfpathrectangle{\pgfqpoint{0.674954in}{0.862305in}}{\pgfqpoint{6.472460in}{4.369081in}}%
\pgfusepath{clip}%
\pgfsetbuttcap%
\pgfsetroundjoin%
\pgfsetlinewidth{2.007500pt}%
\definecolor{currentstroke}{rgb}{0.501961,0.501961,0.501961}%
\pgfsetstrokecolor{currentstroke}%
\pgfsetstrokeopacity{0.300000}%
\pgfsetdash{{7.400000pt}{3.200000pt}}{0.000000pt}%
\pgfpathmoveto{\pgfqpoint{0.674954in}{3.046845in}}%
\pgfpathlineto{\pgfqpoint{7.147414in}{3.046845in}}%
\pgfusepath{stroke}%
\end{pgfscope}%
\begin{pgfscope}%
\pgfsetbuttcap%
\pgfsetroundjoin%
\definecolor{currentfill}{rgb}{0.000000,0.000000,0.000000}%
\pgfsetfillcolor{currentfill}%
\pgfsetlinewidth{0.803000pt}%
\definecolor{currentstroke}{rgb}{0.000000,0.000000,0.000000}%
\pgfsetstrokecolor{currentstroke}%
\pgfsetdash{}{0pt}%
\pgfsys@defobject{currentmarker}{\pgfqpoint{-0.048611in}{0.000000in}}{\pgfqpoint{-0.000000in}{0.000000in}}{%
\pgfpathmoveto{\pgfqpoint{-0.000000in}{0.000000in}}%
\pgfpathlineto{\pgfqpoint{-0.048611in}{0.000000in}}%
\pgfusepath{stroke,fill}%
}%
\begin{pgfscope}%
\pgfsys@transformshift{0.674954in}{3.046845in}%
\pgfsys@useobject{currentmarker}{}%
\end{pgfscope}%
\end{pgfscope}%
\begin{pgfscope}%
\definecolor{textcolor}{rgb}{0.000000,0.000000,0.000000}%
\pgfsetstrokecolor{textcolor}%
\pgfsetfillcolor{textcolor}%
\pgftext[x=0.259244in, y=2.928231in, left, base]{\color{textcolor}{\rmfamily\fontsize{25.000000}{30.000000}\selectfont\catcode`\^=\active\def^{\ifmmode\sp\else\^{}\fi}\catcode`\%=\active\def
\end{pgfscope}%
\begin{pgfscope}%
\pgfpathrectangle{\pgfqpoint{0.674954in}{0.862305in}}{\pgfqpoint{6.472460in}{4.369081in}}%
\pgfusepath{clip}%
\pgfsetbuttcap%
\pgfsetroundjoin%
\pgfsetlinewidth{2.007500pt}%
\definecolor{currentstroke}{rgb}{0.501961,0.501961,0.501961}%
\pgfsetstrokecolor{currentstroke}%
\pgfsetstrokeopacity{0.300000}%
\pgfsetdash{{7.400000pt}{3.200000pt}}{0.000000pt}%
\pgfpathmoveto{\pgfqpoint{0.674954in}{4.139116in}}%
\pgfpathlineto{\pgfqpoint{7.147414in}{4.139116in}}%
\pgfusepath{stroke}%
\end{pgfscope}%
\begin{pgfscope}%
\pgfsetbuttcap%
\pgfsetroundjoin%
\definecolor{currentfill}{rgb}{0.000000,0.000000,0.000000}%
\pgfsetfillcolor{currentfill}%
\pgfsetlinewidth{0.803000pt}%
\definecolor{currentstroke}{rgb}{0.000000,0.000000,0.000000}%
\pgfsetstrokecolor{currentstroke}%
\pgfsetdash{}{0pt}%
\pgfsys@defobject{currentmarker}{\pgfqpoint{-0.048611in}{0.000000in}}{\pgfqpoint{-0.000000in}{0.000000in}}{%
\pgfpathmoveto{\pgfqpoint{-0.000000in}{0.000000in}}%
\pgfpathlineto{\pgfqpoint{-0.048611in}{0.000000in}}%
\pgfusepath{stroke,fill}%
}%
\begin{pgfscope}%
\pgfsys@transformshift{0.674954in}{4.139116in}%
\pgfsys@useobject{currentmarker}{}%
\end{pgfscope}%
\end{pgfscope}%
\begin{pgfscope}%
\definecolor{textcolor}{rgb}{0.000000,0.000000,0.000000}%
\pgfsetstrokecolor{textcolor}%
\pgfsetfillcolor{textcolor}%
\pgftext[x=0.259244in, y=4.020501in, left, base]{\color{textcolor}{\rmfamily\fontsize{25.000000}{30.000000}\selectfont\catcode`\^=\active\def^{\ifmmode\sp\else\^{}\fi}\catcode`\%=\active\def
\end{pgfscope}%
\begin{pgfscope}%
\pgfpathrectangle{\pgfqpoint{0.674954in}{0.862305in}}{\pgfqpoint{6.472460in}{4.369081in}}%
\pgfusepath{clip}%
\pgfsetbuttcap%
\pgfsetroundjoin%
\pgfsetlinewidth{2.007500pt}%
\definecolor{currentstroke}{rgb}{0.501961,0.501961,0.501961}%
\pgfsetstrokecolor{currentstroke}%
\pgfsetstrokeopacity{0.300000}%
\pgfsetdash{{7.400000pt}{3.200000pt}}{0.000000pt}%
\pgfpathmoveto{\pgfqpoint{0.674954in}{5.231386in}}%
\pgfpathlineto{\pgfqpoint{7.147414in}{5.231386in}}%
\pgfusepath{stroke}%
\end{pgfscope}%
\begin{pgfscope}%
\pgfsetbuttcap%
\pgfsetroundjoin%
\definecolor{currentfill}{rgb}{0.000000,0.000000,0.000000}%
\pgfsetfillcolor{currentfill}%
\pgfsetlinewidth{0.803000pt}%
\definecolor{currentstroke}{rgb}{0.000000,0.000000,0.000000}%
\pgfsetstrokecolor{currentstroke}%
\pgfsetdash{}{0pt}%
\pgfsys@defobject{currentmarker}{\pgfqpoint{-0.048611in}{0.000000in}}{\pgfqpoint{-0.000000in}{0.000000in}}{%
\pgfpathmoveto{\pgfqpoint{-0.000000in}{0.000000in}}%
\pgfpathlineto{\pgfqpoint{-0.048611in}{0.000000in}}%
\pgfusepath{stroke,fill}%
}%
\begin{pgfscope}%
\pgfsys@transformshift{0.674954in}{5.231386in}%
\pgfsys@useobject{currentmarker}{}%
\end{pgfscope}%
\end{pgfscope}%
\begin{pgfscope}%
\definecolor{textcolor}{rgb}{0.000000,0.000000,0.000000}%
\pgfsetstrokecolor{textcolor}%
\pgfsetfillcolor{textcolor}%
\pgftext[x=0.100000in, y=5.112772in, left, base]{\color{textcolor}{\rmfamily\fontsize{25.000000}{30.000000}\selectfont\catcode`\^=\active\def^{\ifmmode\sp\else\^{}\fi}\catcode`\%=\active\def
\end{pgfscope}%
\begin{pgfscope}%
\pgfpathrectangle{\pgfqpoint{0.674954in}{0.862305in}}{\pgfqpoint{6.472460in}{4.369081in}}%
\pgfusepath{clip}%
\pgfsetrectcap%
\pgfsetroundjoin%
\pgfsetlinewidth{2.509375pt}%
\definecolor{currentstroke}{rgb}{0.050980,0.415686,0.509804}%
\pgfsetstrokecolor{currentstroke}%
\pgfsetdash{}{0pt}%
\pgfpathmoveto{\pgfqpoint{0.674954in}{5.153080in}}%
\pgfpathlineto{\pgfqpoint{0.998577in}{5.087427in}}%
\pgfpathlineto{\pgfqpoint{2.293069in}{4.912588in}}%
\pgfpathlineto{\pgfqpoint{3.911184in}{4.727877in}}%
\pgfpathlineto{\pgfqpoint{5.529299in}{4.548108in}}%
\pgfpathlineto{\pgfqpoint{6.823791in}{4.724238in}}%
\pgfpathlineto{\pgfqpoint{7.147414in}{4.841040in}}%
\pgfusepath{stroke}%
\end{pgfscope}%
\begin{pgfscope}%
\pgfpathrectangle{\pgfqpoint{0.674954in}{0.862305in}}{\pgfqpoint{6.472460in}{4.369081in}}%
\pgfusepath{clip}%
\pgfsetbuttcap%
\pgfsetroundjoin%
\definecolor{currentfill}{rgb}{0.050980,0.415686,0.509804}%
\pgfsetfillcolor{currentfill}%
\pgfsetlinewidth{1.003750pt}%
\definecolor{currentstroke}{rgb}{0.050980,0.415686,0.509804}%
\pgfsetstrokecolor{currentstroke}%
\pgfsetdash{}{0pt}%
\pgfsys@defobject{currentmarker}{\pgfqpoint{-0.055556in}{-0.055556in}}{\pgfqpoint{0.055556in}{0.055556in}}{%
\pgfpathmoveto{\pgfqpoint{0.000000in}{-0.055556in}}%
\pgfpathcurveto{\pgfqpoint{0.014734in}{-0.055556in}}{\pgfqpoint{0.028866in}{-0.049702in}}{\pgfqpoint{0.039284in}{-0.039284in}}%
\pgfpathcurveto{\pgfqpoint{0.049702in}{-0.028866in}}{\pgfqpoint{0.055556in}{-0.014734in}}{\pgfqpoint{0.055556in}{0.000000in}}%
\pgfpathcurveto{\pgfqpoint{0.055556in}{0.014734in}}{\pgfqpoint{0.049702in}{0.028866in}}{\pgfqpoint{0.039284in}{0.039284in}}%
\pgfpathcurveto{\pgfqpoint{0.028866in}{0.049702in}}{\pgfqpoint{0.014734in}{0.055556in}}{\pgfqpoint{0.000000in}{0.055556in}}%
\pgfpathcurveto{\pgfqpoint{-0.014734in}{0.055556in}}{\pgfqpoint{-0.028866in}{0.049702in}}{\pgfqpoint{-0.039284in}{0.039284in}}%
\pgfpathcurveto{\pgfqpoint{-0.049702in}{0.028866in}}{\pgfqpoint{-0.055556in}{0.014734in}}{\pgfqpoint{-0.055556in}{0.000000in}}%
\pgfpathcurveto{\pgfqpoint{-0.055556in}{-0.014734in}}{\pgfqpoint{-0.049702in}{-0.028866in}}{\pgfqpoint{-0.039284in}{-0.039284in}}%
\pgfpathcurveto{\pgfqpoint{-0.028866in}{-0.049702in}}{\pgfqpoint{-0.014734in}{-0.055556in}}{\pgfqpoint{0.000000in}{-0.055556in}}%
\pgfpathlineto{\pgfqpoint{0.000000in}{-0.055556in}}%
\pgfpathclose%
\pgfusepath{stroke,fill}%
}%
\begin{pgfscope}%
\pgfsys@transformshift{0.674954in}{5.153080in}%
\pgfsys@useobject{currentmarker}{}%
\end{pgfscope}%
\begin{pgfscope}%
\pgfsys@transformshift{0.998577in}{5.087427in}%
\pgfsys@useobject{currentmarker}{}%
\end{pgfscope}%
\begin{pgfscope}%
\pgfsys@transformshift{2.293069in}{4.912588in}%
\pgfsys@useobject{currentmarker}{}%
\end{pgfscope}%
\begin{pgfscope}%
\pgfsys@transformshift{3.911184in}{4.727877in}%
\pgfsys@useobject{currentmarker}{}%
\end{pgfscope}%
\begin{pgfscope}%
\pgfsys@transformshift{5.529299in}{4.548108in}%
\pgfsys@useobject{currentmarker}{}%
\end{pgfscope}%
\begin{pgfscope}%
\pgfsys@transformshift{6.823791in}{4.724238in}%
\pgfsys@useobject{currentmarker}{}%
\end{pgfscope}%
\begin{pgfscope}%
\pgfsys@transformshift{7.147414in}{4.841040in}%
\pgfsys@useobject{currentmarker}{}%
\end{pgfscope}%
\end{pgfscope}%
\begin{pgfscope}%
\pgfpathrectangle{\pgfqpoint{0.674954in}{0.862305in}}{\pgfqpoint{6.472460in}{4.369081in}}%
\pgfusepath{clip}%
\pgfsetbuttcap%
\pgfsetroundjoin%
\pgfsetlinewidth{2.509375pt}%
\definecolor{currentstroke}{rgb}{0.960784,0.462745,0.000000}%
\pgfsetstrokecolor{currentstroke}%
\pgfsetdash{{9.250000pt}{4.000000pt}}{0.000000pt}%
\pgfpathmoveto{\pgfqpoint{0.674954in}{5.048203in}}%
\pgfpathlineto{\pgfqpoint{0.998577in}{4.959726in}}%
\pgfpathlineto{\pgfqpoint{2.293069in}{4.665611in}}%
\pgfpathlineto{\pgfqpoint{3.911184in}{4.358948in}}%
\pgfpathlineto{\pgfqpoint{5.529299in}{4.126816in}}%
\pgfpathlineto{\pgfqpoint{6.823791in}{4.303514in}}%
\pgfpathlineto{\pgfqpoint{7.147414in}{4.470619in}}%
\pgfusepath{stroke}%
\end{pgfscope}%
\begin{pgfscope}%
\pgfpathrectangle{\pgfqpoint{0.674954in}{0.862305in}}{\pgfqpoint{6.472460in}{4.369081in}}%
\pgfusepath{clip}%
\pgfsetbuttcap%
\pgfsetmiterjoin%
\definecolor{currentfill}{rgb}{0.960784,0.462745,0.000000}%
\pgfsetfillcolor{currentfill}%
\pgfsetlinewidth{1.003750pt}%
\definecolor{currentstroke}{rgb}{0.960784,0.462745,0.000000}%
\pgfsetstrokecolor{currentstroke}%
\pgfsetdash{}{0pt}%
\pgfsys@defobject{currentmarker}{\pgfqpoint{-0.055556in}{-0.055556in}}{\pgfqpoint{0.055556in}{0.055556in}}{%
\pgfpathmoveto{\pgfqpoint{-0.055556in}{-0.055556in}}%
\pgfpathlineto{\pgfqpoint{0.055556in}{-0.055556in}}%
\pgfpathlineto{\pgfqpoint{0.055556in}{0.055556in}}%
\pgfpathlineto{\pgfqpoint{-0.055556in}{0.055556in}}%
\pgfpathlineto{\pgfqpoint{-0.055556in}{-0.055556in}}%
\pgfpathclose%
\pgfusepath{stroke,fill}%
}%
\begin{pgfscope}%
\pgfsys@transformshift{0.674954in}{5.048203in}%
\pgfsys@useobject{currentmarker}{}%
\end{pgfscope}%
\begin{pgfscope}%
\pgfsys@transformshift{0.998577in}{4.959726in}%
\pgfsys@useobject{currentmarker}{}%
\end{pgfscope}%
\begin{pgfscope}%
\pgfsys@transformshift{2.293069in}{4.665611in}%
\pgfsys@useobject{currentmarker}{}%
\end{pgfscope}%
\begin{pgfscope}%
\pgfsys@transformshift{3.911184in}{4.358948in}%
\pgfsys@useobject{currentmarker}{}%
\end{pgfscope}%
\begin{pgfscope}%
\pgfsys@transformshift{5.529299in}{4.126816in}%
\pgfsys@useobject{currentmarker}{}%
\end{pgfscope}%
\begin{pgfscope}%
\pgfsys@transformshift{6.823791in}{4.303514in}%
\pgfsys@useobject{currentmarker}{}%
\end{pgfscope}%
\begin{pgfscope}%
\pgfsys@transformshift{7.147414in}{4.470619in}%
\pgfsys@useobject{currentmarker}{}%
\end{pgfscope}%
\end{pgfscope}%
\begin{pgfscope}%
\pgfpathrectangle{\pgfqpoint{0.674954in}{0.862305in}}{\pgfqpoint{6.472460in}{4.369081in}}%
\pgfusepath{clip}%
\pgfsetbuttcap%
\pgfsetroundjoin%
\pgfsetlinewidth{2.509375pt}%
\definecolor{currentstroke}{rgb}{0.219608,0.219608,0.219608}%
\pgfsetstrokecolor{currentstroke}%
\pgfsetdash{{2.500000pt}{4.125000pt}}{0.000000pt}%
\pgfpathmoveto{\pgfqpoint{0.674954in}{4.506345in}}%
\pgfpathlineto{\pgfqpoint{0.998577in}{4.361553in}}%
\pgfpathlineto{\pgfqpoint{2.293069in}{3.913073in}}%
\pgfpathlineto{\pgfqpoint{3.911184in}{3.578515in}}%
\pgfpathlineto{\pgfqpoint{5.529299in}{3.417080in}}%
\pgfpathlineto{\pgfqpoint{6.823791in}{3.560346in}}%
\pgfpathlineto{\pgfqpoint{7.147414in}{3.666273in}}%
\pgfusepath{stroke}%
\end{pgfscope}%
\begin{pgfscope}%
\pgfpathrectangle{\pgfqpoint{0.674954in}{0.862305in}}{\pgfqpoint{6.472460in}{4.369081in}}%
\pgfusepath{clip}%
\pgfsetbuttcap%
\pgfsetmiterjoin%
\definecolor{currentfill}{rgb}{0.219608,0.219608,0.219608}%
\pgfsetfillcolor{currentfill}%
\pgfsetlinewidth{1.003750pt}%
\definecolor{currentstroke}{rgb}{0.219608,0.219608,0.219608}%
\pgfsetstrokecolor{currentstroke}%
\pgfsetdash{}{0pt}%
\pgfsys@defobject{currentmarker}{\pgfqpoint{-0.055556in}{-0.055556in}}{\pgfqpoint{0.055556in}{0.055556in}}{%
\pgfpathmoveto{\pgfqpoint{0.000000in}{0.055556in}}%
\pgfpathlineto{\pgfqpoint{-0.055556in}{-0.055556in}}%
\pgfpathlineto{\pgfqpoint{0.055556in}{-0.055556in}}%
\pgfpathlineto{\pgfqpoint{0.000000in}{0.055556in}}%
\pgfpathclose%
\pgfusepath{stroke,fill}%
}%
\begin{pgfscope}%
\pgfsys@transformshift{0.674954in}{4.506345in}%
\pgfsys@useobject{currentmarker}{}%
\end{pgfscope}%
\begin{pgfscope}%
\pgfsys@transformshift{0.998577in}{4.361553in}%
\pgfsys@useobject{currentmarker}{}%
\end{pgfscope}%
\begin{pgfscope}%
\pgfsys@transformshift{2.293069in}{3.913073in}%
\pgfsys@useobject{currentmarker}{}%
\end{pgfscope}%
\begin{pgfscope}%
\pgfsys@transformshift{3.911184in}{3.578515in}%
\pgfsys@useobject{currentmarker}{}%
\end{pgfscope}%
\begin{pgfscope}%
\pgfsys@transformshift{5.529299in}{3.417080in}%
\pgfsys@useobject{currentmarker}{}%
\end{pgfscope}%
\begin{pgfscope}%
\pgfsys@transformshift{6.823791in}{3.560346in}%
\pgfsys@useobject{currentmarker}{}%
\end{pgfscope}%
\begin{pgfscope}%
\pgfsys@transformshift{7.147414in}{3.666273in}%
\pgfsys@useobject{currentmarker}{}%
\end{pgfscope}%
\end{pgfscope}%
\begin{pgfscope}%
\pgfsetrectcap%
\pgfsetmiterjoin%
\pgfsetlinewidth{2.007500pt}%
\definecolor{currentstroke}{rgb}{0.000000,0.000000,0.000000}%
\pgfsetstrokecolor{currentstroke}%
\pgfsetdash{}{0pt}%
\pgfpathmoveto{\pgfqpoint{0.674954in}{0.862305in}}%
\pgfpathlineto{\pgfqpoint{0.674954in}{5.231386in}}%
\pgfusepath{stroke}%
\end{pgfscope}%
\begin{pgfscope}%
\pgfsetrectcap%
\pgfsetmiterjoin%
\pgfsetlinewidth{2.007500pt}%
\definecolor{currentstroke}{rgb}{0.000000,0.000000,0.000000}%
\pgfsetstrokecolor{currentstroke}%
\pgfsetdash{}{0pt}%
\pgfpathmoveto{\pgfqpoint{0.674954in}{0.862305in}}%
\pgfpathlineto{\pgfqpoint{7.147414in}{0.862305in}}%
\pgfusepath{stroke}%
\end{pgfscope}%
\begin{pgfscope}%
\pgfsetbuttcap%
\pgfsetmiterjoin%
\definecolor{currentfill}{rgb}{1.000000,1.000000,1.000000}%
\pgfsetfillcolor{currentfill}%
\pgfsetlinewidth{1.003750pt}%
\definecolor{currentstroke}{rgb}{0.800000,0.800000,0.800000}%
\pgfsetstrokecolor{currentstroke}%
\pgfsetdash{}{0pt}%
\pgfpathmoveto{\pgfqpoint{0.869398in}{1.001194in}}%
\pgfpathlineto{\pgfqpoint{2.213036in}{1.001194in}}%
\pgfpathquadraticcurveto{\pgfqpoint{2.268591in}{1.001194in}}{\pgfqpoint{2.268591in}{1.056749in}}%
\pgfpathlineto{\pgfqpoint{2.268591in}{2.191056in}}%
\pgfpathquadraticcurveto{\pgfqpoint{2.268591in}{2.246612in}}{\pgfqpoint{2.213036in}{2.246612in}}%
\pgfpathlineto{\pgfqpoint{0.869398in}{2.246612in}}%
\pgfpathquadraticcurveto{\pgfqpoint{0.813843in}{2.246612in}}{\pgfqpoint{0.813843in}{2.191056in}}%
\pgfpathlineto{\pgfqpoint{0.813843in}{1.056749in}}%
\pgfpathquadraticcurveto{\pgfqpoint{0.813843in}{1.001194in}}{\pgfqpoint{0.869398in}{1.001194in}}%
\pgfpathlineto{\pgfqpoint{0.869398in}{1.001194in}}%
\pgfpathclose%
\pgfusepath{stroke,fill}%
\end{pgfscope}%
\begin{pgfscope}%
\pgfsetrectcap%
\pgfsetroundjoin%
\pgfsetlinewidth{2.509375pt}%
\definecolor{currentstroke}{rgb}{0.050980,0.415686,0.509804}%
\pgfsetstrokecolor{currentstroke}%
\pgfsetdash{}{0pt}%
\pgfpathmoveto{\pgfqpoint{0.924954in}{2.038278in}}%
\pgfpathlineto{\pgfqpoint{1.202732in}{2.038278in}}%
\pgfpathlineto{\pgfqpoint{1.480509in}{2.038278in}}%
\pgfusepath{stroke}%
\end{pgfscope}%
\begin{pgfscope}%
\pgfsetbuttcap%
\pgfsetroundjoin%
\definecolor{currentfill}{rgb}{0.050980,0.415686,0.509804}%
\pgfsetfillcolor{currentfill}%
\pgfsetlinewidth{1.003750pt}%
\definecolor{currentstroke}{rgb}{0.050980,0.415686,0.509804}%
\pgfsetstrokecolor{currentstroke}%
\pgfsetdash{}{0pt}%
\pgfsys@defobject{currentmarker}{\pgfqpoint{-0.055556in}{-0.055556in}}{\pgfqpoint{0.055556in}{0.055556in}}{%
\pgfpathmoveto{\pgfqpoint{0.000000in}{-0.055556in}}%
\pgfpathcurveto{\pgfqpoint{0.014734in}{-0.055556in}}{\pgfqpoint{0.028866in}{-0.049702in}}{\pgfqpoint{0.039284in}{-0.039284in}}%
\pgfpathcurveto{\pgfqpoint{0.049702in}{-0.028866in}}{\pgfqpoint{0.055556in}{-0.014734in}}{\pgfqpoint{0.055556in}{0.000000in}}%
\pgfpathcurveto{\pgfqpoint{0.055556in}{0.014734in}}{\pgfqpoint{0.049702in}{0.028866in}}{\pgfqpoint{0.039284in}{0.039284in}}%
\pgfpathcurveto{\pgfqpoint{0.028866in}{0.049702in}}{\pgfqpoint{0.014734in}{0.055556in}}{\pgfqpoint{0.000000in}{0.055556in}}%
\pgfpathcurveto{\pgfqpoint{-0.014734in}{0.055556in}}{\pgfqpoint{-0.028866in}{0.049702in}}{\pgfqpoint{-0.039284in}{0.039284in}}%
\pgfpathcurveto{\pgfqpoint{-0.049702in}{0.028866in}}{\pgfqpoint{-0.055556in}{0.014734in}}{\pgfqpoint{-0.055556in}{0.000000in}}%
\pgfpathcurveto{\pgfqpoint{-0.055556in}{-0.014734in}}{\pgfqpoint{-0.049702in}{-0.028866in}}{\pgfqpoint{-0.039284in}{-0.039284in}}%
\pgfpathcurveto{\pgfqpoint{-0.028866in}{-0.049702in}}{\pgfqpoint{-0.014734in}{-0.055556in}}{\pgfqpoint{0.000000in}{-0.055556in}}%
\pgfpathlineto{\pgfqpoint{0.000000in}{-0.055556in}}%
\pgfpathclose%
\pgfusepath{stroke,fill}%
}%
\begin{pgfscope}%
\pgfsys@transformshift{1.202732in}{2.038278in}%
\pgfsys@useobject{currentmarker}{}%
\end{pgfscope}%
\end{pgfscope}%
\begin{pgfscope}%
\definecolor{textcolor}{rgb}{0.000000,0.000000,0.000000}%
\pgfsetstrokecolor{textcolor}%
\pgfsetfillcolor{textcolor}%
\pgftext[x=1.702732in,y=1.941056in,left,base]{\color{textcolor}{\rmfamily\fontsize{20.000000}{24.000000}\selectfont\catcode`\^=\active\def^{\ifmmode\sp\else\^{}\fi}\catcode`\%=\active\def
\end{pgfscope}%
\begin{pgfscope}%
\pgfsetbuttcap%
\pgfsetroundjoin%
\pgfsetlinewidth{2.509375pt}%
\definecolor{currentstroke}{rgb}{0.960784,0.462745,0.000000}%
\pgfsetstrokecolor{currentstroke}%
\pgfsetdash{{9.250000pt}{4.000000pt}}{0.000000pt}%
\pgfpathmoveto{\pgfqpoint{0.924954in}{1.650917in}}%
\pgfpathlineto{\pgfqpoint{1.202732in}{1.650917in}}%
\pgfpathlineto{\pgfqpoint{1.480509in}{1.650917in}}%
\pgfusepath{stroke}%
\end{pgfscope}%
\begin{pgfscope}%
\pgfsetbuttcap%
\pgfsetmiterjoin%
\definecolor{currentfill}{rgb}{0.960784,0.462745,0.000000}%
\pgfsetfillcolor{currentfill}%
\pgfsetlinewidth{1.003750pt}%
\definecolor{currentstroke}{rgb}{0.960784,0.462745,0.000000}%
\pgfsetstrokecolor{currentstroke}%
\pgfsetdash{}{0pt}%
\pgfsys@defobject{currentmarker}{\pgfqpoint{-0.055556in}{-0.055556in}}{\pgfqpoint{0.055556in}{0.055556in}}{%
\pgfpathmoveto{\pgfqpoint{-0.055556in}{-0.055556in}}%
\pgfpathlineto{\pgfqpoint{0.055556in}{-0.055556in}}%
\pgfpathlineto{\pgfqpoint{0.055556in}{0.055556in}}%
\pgfpathlineto{\pgfqpoint{-0.055556in}{0.055556in}}%
\pgfpathlineto{\pgfqpoint{-0.055556in}{-0.055556in}}%
\pgfpathclose%
\pgfusepath{stroke,fill}%
}%
\begin{pgfscope}%
\pgfsys@transformshift{1.202732in}{1.650917in}%
\pgfsys@useobject{currentmarker}{}%
\end{pgfscope}%
\end{pgfscope}%
\begin{pgfscope}%
\definecolor{textcolor}{rgb}{0.000000,0.000000,0.000000}%
\pgfsetstrokecolor{textcolor}%
\pgfsetfillcolor{textcolor}%
\pgftext[x=1.702732in,y=1.553694in,left,base]{\color{textcolor}{\rmfamily\fontsize{20.000000}{24.000000}\selectfont\catcode`\^=\active\def^{\ifmmode\sp\else\^{}\fi}\catcode`\%=\active\def
\end{pgfscope}%
\begin{pgfscope}%
\pgfsetbuttcap%
\pgfsetroundjoin%
\pgfsetlinewidth{2.509375pt}%
\definecolor{currentstroke}{rgb}{0.219608,0.219608,0.219608}%
\pgfsetstrokecolor{currentstroke}%
\pgfsetdash{{2.500000pt}{4.125000pt}}{0.000000pt}%
\pgfpathmoveto{\pgfqpoint{0.924954in}{1.263555in}}%
\pgfpathlineto{\pgfqpoint{1.202732in}{1.263555in}}%
\pgfpathlineto{\pgfqpoint{1.480509in}{1.263555in}}%
\pgfusepath{stroke}%
\end{pgfscope}%
\begin{pgfscope}%
\pgfsetbuttcap%
\pgfsetmiterjoin%
\definecolor{currentfill}{rgb}{0.219608,0.219608,0.219608}%
\pgfsetfillcolor{currentfill}%
\pgfsetlinewidth{1.003750pt}%
\definecolor{currentstroke}{rgb}{0.219608,0.219608,0.219608}%
\pgfsetstrokecolor{currentstroke}%
\pgfsetdash{}{0pt}%
\pgfsys@defobject{currentmarker}{\pgfqpoint{-0.055556in}{-0.055556in}}{\pgfqpoint{0.055556in}{0.055556in}}{%
\pgfpathmoveto{\pgfqpoint{0.000000in}{0.055556in}}%
\pgfpathlineto{\pgfqpoint{-0.055556in}{-0.055556in}}%
\pgfpathlineto{\pgfqpoint{0.055556in}{-0.055556in}}%
\pgfpathlineto{\pgfqpoint{0.000000in}{0.055556in}}%
\pgfpathclose%
\pgfusepath{stroke,fill}%
}%
\begin{pgfscope}%
\pgfsys@transformshift{1.202732in}{1.263555in}%
\pgfsys@useobject{currentmarker}{}%
\end{pgfscope}%
\end{pgfscope}%
\begin{pgfscope}%
\definecolor{textcolor}{rgb}{0.000000,0.000000,0.000000}%
\pgfsetstrokecolor{textcolor}%
\pgfsetfillcolor{textcolor}%
\pgftext[x=1.702732in,y=1.166333in,left,base]{\color{textcolor}{\rmfamily\fontsize{20.000000}{24.000000}\selectfont\catcode`\^=\active\def^{\ifmmode\sp\else\^{}\fi}\catcode`\%=\active\def
\end{pgfscope}%
\end{pgfpicture}%
\makeatother%
\endgroup%

%% file: images/RGB_Thermal_Features/distances_cosine.pgf
\begingroup%
\makeatletter%
\begin{pgfpicture}%
\pgfpathrectangle{\pgfpointorigin}{\pgfqpoint{11.450000in}{4.450000in}}%
\pgfusepath{use as bounding box, clip}%
\begin{pgfscope}%
\pgfsetbuttcap%
\pgfsetmiterjoin%
\definecolor{currentfill}{rgb}{1.000000,1.000000,1.000000}%
\pgfsetfillcolor{currentfill}%
\pgfsetlinewidth{0.000000pt}%
\definecolor{currentstroke}{rgb}{1.000000,1.000000,1.000000}%
\pgfsetstrokecolor{currentstroke}%
\pgfsetdash{}{0pt}%
\pgfpathmoveto{\pgfqpoint{0.000000in}{0.000000in}}%
\pgfpathlineto{\pgfqpoint{11.450000in}{0.000000in}}%
\pgfpathlineto{\pgfqpoint{11.450000in}{4.450000in}}%
\pgfpathlineto{\pgfqpoint{0.000000in}{4.450000in}}%
\pgfpathlineto{\pgfqpoint{0.000000in}{0.000000in}}%
\pgfpathclose%
\pgfusepath{fill}%
\end{pgfscope}%
\begin{pgfscope}%
\pgfsetbuttcap%
\pgfsetmiterjoin%
\definecolor{currentfill}{rgb}{1.000000,1.000000,1.000000}%
\pgfsetfillcolor{currentfill}%
\pgfsetlinewidth{0.000000pt}%
\definecolor{currentstroke}{rgb}{0.000000,0.000000,0.000000}%
\pgfsetstrokecolor{currentstroke}%
\pgfsetstrokeopacity{0.000000}%
\pgfsetdash{}{0pt}%
\pgfpathmoveto{\pgfqpoint{0.707138in}{0.642876in}}%
\pgfpathlineto{\pgfqpoint{11.350000in}{0.642876in}}%
\pgfpathlineto{\pgfqpoint{11.350000in}{4.350000in}}%
\pgfpathlineto{\pgfqpoint{0.707138in}{4.350000in}}%
\pgfpathlineto{\pgfqpoint{0.707138in}{0.642876in}}%
\pgfpathclose%
\pgfusepath{fill}%
\end{pgfscope}%
\begin{pgfscope}%
\pgfpathrectangle{\pgfqpoint{0.707138in}{0.642876in}}{\pgfqpoint{10.642862in}{3.707124in}}%
\pgfusepath{clip}%
\pgfsetbuttcap%
\pgfsetroundjoin%
\definecolor{currentfill}{rgb}{0.121569,0.466667,0.705882}%
\pgfsetfillcolor{currentfill}%
\pgfsetfillopacity{0.300000}%
\pgfsetlinewidth{1.003750pt}%
\definecolor{currentstroke}{rgb}{0.121569,0.466667,0.705882}%
\pgfsetstrokecolor{currentstroke}%
\pgfsetstrokeopacity{0.300000}%
\pgfsetdash{}{0pt}%
\pgfsys@defobject{currentmarker}{\pgfqpoint{0.707138in}{0.811381in}}{\pgfqpoint{11.350000in}{2.290673in}}{%
\pgfpathmoveto{\pgfqpoint{0.707138in}{1.644113in}}%
\pgfpathlineto{\pgfqpoint{0.707138in}{1.317353in}}%
\pgfpathlineto{\pgfqpoint{1.169871in}{1.681567in}}%
\pgfpathlineto{\pgfqpoint{1.632604in}{1.725588in}}%
\pgfpathlineto{\pgfqpoint{2.095337in}{1.691419in}}%
\pgfpathlineto{\pgfqpoint{2.558070in}{1.524356in}}%
\pgfpathlineto{\pgfqpoint{3.020803in}{1.581148in}}%
\pgfpathlineto{\pgfqpoint{3.483536in}{1.535794in}}%
\pgfpathlineto{\pgfqpoint{3.946270in}{1.492517in}}%
\pgfpathlineto{\pgfqpoint{4.409003in}{1.481127in}}%
\pgfpathlineto{\pgfqpoint{4.871736in}{1.439112in}}%
\pgfpathlineto{\pgfqpoint{5.334469in}{1.374172in}}%
\pgfpathlineto{\pgfqpoint{5.797202in}{1.314687in}}%
\pgfpathlineto{\pgfqpoint{6.259935in}{1.288551in}}%
\pgfpathlineto{\pgfqpoint{6.722668in}{1.175446in}}%
\pgfpathlineto{\pgfqpoint{7.185402in}{1.035902in}}%
\pgfpathlineto{\pgfqpoint{7.648135in}{1.068033in}}%
\pgfpathlineto{\pgfqpoint{8.110868in}{1.110710in}}%
\pgfpathlineto{\pgfqpoint{8.573601in}{0.972162in}}%
\pgfpathlineto{\pgfqpoint{9.036334in}{0.910149in}}%
\pgfpathlineto{\pgfqpoint{9.499067in}{0.951690in}}%
\pgfpathlineto{\pgfqpoint{9.961801in}{1.002584in}}%
\pgfpathlineto{\pgfqpoint{10.424534in}{0.918006in}}%
\pgfpathlineto{\pgfqpoint{10.887267in}{0.866861in}}%
\pgfpathlineto{\pgfqpoint{11.350000in}{0.811381in}}%
\pgfpathlineto{\pgfqpoint{11.350000in}{1.231039in}}%
\pgfpathlineto{\pgfqpoint{11.350000in}{1.231039in}}%
\pgfpathlineto{\pgfqpoint{10.887267in}{1.390398in}}%
\pgfpathlineto{\pgfqpoint{10.424534in}{1.434680in}}%
\pgfpathlineto{\pgfqpoint{9.961801in}{1.468449in}}%
\pgfpathlineto{\pgfqpoint{9.499067in}{1.406679in}}%
\pgfpathlineto{\pgfqpoint{9.036334in}{1.470473in}}%
\pgfpathlineto{\pgfqpoint{8.573601in}{1.473724in}}%
\pgfpathlineto{\pgfqpoint{8.110868in}{1.555529in}}%
\pgfpathlineto{\pgfqpoint{7.648135in}{1.474650in}}%
\pgfpathlineto{\pgfqpoint{7.185402in}{1.407071in}}%
\pgfpathlineto{\pgfqpoint{6.722668in}{1.650723in}}%
\pgfpathlineto{\pgfqpoint{6.259935in}{1.908980in}}%
\pgfpathlineto{\pgfqpoint{5.797202in}{1.953865in}}%
\pgfpathlineto{\pgfqpoint{5.334469in}{1.943343in}}%
\pgfpathlineto{\pgfqpoint{4.871736in}{1.970494in}}%
\pgfpathlineto{\pgfqpoint{4.409003in}{1.989580in}}%
\pgfpathlineto{\pgfqpoint{3.946270in}{1.996455in}}%
\pgfpathlineto{\pgfqpoint{3.483536in}{2.026973in}}%
\pgfpathlineto{\pgfqpoint{3.020803in}{2.064911in}}%
\pgfpathlineto{\pgfqpoint{2.558070in}{1.907252in}}%
\pgfpathlineto{\pgfqpoint{2.095337in}{2.212159in}}%
\pgfpathlineto{\pgfqpoint{1.632604in}{2.290673in}}%
\pgfpathlineto{\pgfqpoint{1.169871in}{2.208053in}}%
\pgfpathlineto{\pgfqpoint{0.707138in}{1.644113in}}%
\pgfpathlineto{\pgfqpoint{0.707138in}{1.644113in}}%
\pgfpathclose%
\pgfusepath{stroke,fill}%
}%
\begin{pgfscope}%
\pgfsys@transformshift{0.000000in}{0.000000in}%
\pgfsys@useobject{currentmarker}{}%
\end{pgfscope}%
\end{pgfscope}%
\begin{pgfscope}%
\pgfpathrectangle{\pgfqpoint{0.707138in}{0.642876in}}{\pgfqpoint{10.642862in}{3.707124in}}%
\pgfusepath{clip}%
\pgfsetbuttcap%
\pgfsetroundjoin%
\definecolor{currentfill}{rgb}{1.000000,0.498039,0.054902}%
\pgfsetfillcolor{currentfill}%
\pgfsetfillopacity{0.300000}%
\pgfsetlinewidth{1.003750pt}%
\definecolor{currentstroke}{rgb}{1.000000,0.498039,0.054902}%
\pgfsetstrokecolor{currentstroke}%
\pgfsetstrokeopacity{0.300000}%
\pgfsetdash{}{0pt}%
\pgfsys@defobject{currentmarker}{\pgfqpoint{0.707138in}{1.317487in}}{\pgfqpoint{11.350000in}{4.181494in}}{%
\pgfpathmoveto{\pgfqpoint{0.707138in}{1.652794in}}%
\pgfpathlineto{\pgfqpoint{0.707138in}{1.317487in}}%
\pgfpathlineto{\pgfqpoint{1.169871in}{1.743531in}}%
\pgfpathlineto{\pgfqpoint{1.632604in}{1.775070in}}%
\pgfpathlineto{\pgfqpoint{2.095337in}{1.772359in}}%
\pgfpathlineto{\pgfqpoint{2.558070in}{1.483035in}}%
\pgfpathlineto{\pgfqpoint{3.020803in}{1.651449in}}%
\pgfpathlineto{\pgfqpoint{3.483536in}{1.608524in}}%
\pgfpathlineto{\pgfqpoint{3.946270in}{1.535767in}}%
\pgfpathlineto{\pgfqpoint{4.409003in}{1.544001in}}%
\pgfpathlineto{\pgfqpoint{4.871736in}{1.430314in}}%
\pgfpathlineto{\pgfqpoint{5.334469in}{1.319856in}}%
\pgfpathlineto{\pgfqpoint{5.797202in}{1.361815in}}%
\pgfpathlineto{\pgfqpoint{6.259935in}{1.400422in}}%
\pgfpathlineto{\pgfqpoint{6.722668in}{1.464616in}}%
\pgfpathlineto{\pgfqpoint{7.185402in}{1.744376in}}%
\pgfpathlineto{\pgfqpoint{7.648135in}{1.724445in}}%
\pgfpathlineto{\pgfqpoint{8.110868in}{1.715025in}}%
\pgfpathlineto{\pgfqpoint{8.573601in}{1.676630in}}%
\pgfpathlineto{\pgfqpoint{9.036334in}{1.967524in}}%
\pgfpathlineto{\pgfqpoint{9.499067in}{1.916498in}}%
\pgfpathlineto{\pgfqpoint{9.961801in}{2.024672in}}%
\pgfpathlineto{\pgfqpoint{10.424534in}{2.040547in}}%
\pgfpathlineto{\pgfqpoint{10.887267in}{1.919803in}}%
\pgfpathlineto{\pgfqpoint{11.350000in}{1.766327in}}%
\pgfpathlineto{\pgfqpoint{11.350000in}{4.181494in}}%
\pgfpathlineto{\pgfqpoint{11.350000in}{4.181494in}}%
\pgfpathlineto{\pgfqpoint{10.887267in}{3.255488in}}%
\pgfpathlineto{\pgfqpoint{10.424534in}{3.231259in}}%
\pgfpathlineto{\pgfqpoint{9.961801in}{3.412038in}}%
\pgfpathlineto{\pgfqpoint{9.499067in}{3.313553in}}%
\pgfpathlineto{\pgfqpoint{9.036334in}{3.805691in}}%
\pgfpathlineto{\pgfqpoint{8.573601in}{3.250787in}}%
\pgfpathlineto{\pgfqpoint{8.110868in}{3.117007in}}%
\pgfpathlineto{\pgfqpoint{7.648135in}{2.945098in}}%
\pgfpathlineto{\pgfqpoint{7.185402in}{2.704854in}}%
\pgfpathlineto{\pgfqpoint{6.722668in}{2.147126in}}%
\pgfpathlineto{\pgfqpoint{6.259935in}{2.000042in}}%
\pgfpathlineto{\pgfqpoint{5.797202in}{1.790009in}}%
\pgfpathlineto{\pgfqpoint{5.334469in}{1.752323in}}%
\pgfpathlineto{\pgfqpoint{4.871736in}{1.854134in}}%
\pgfpathlineto{\pgfqpoint{4.409003in}{1.935494in}}%
\pgfpathlineto{\pgfqpoint{3.946270in}{1.948704in}}%
\pgfpathlineto{\pgfqpoint{3.483536in}{2.015497in}}%
\pgfpathlineto{\pgfqpoint{3.020803in}{2.078738in}}%
\pgfpathlineto{\pgfqpoint{2.558070in}{1.802129in}}%
\pgfpathlineto{\pgfqpoint{2.095337in}{2.296934in}}%
\pgfpathlineto{\pgfqpoint{1.632604in}{2.380463in}}%
\pgfpathlineto{\pgfqpoint{1.169871in}{2.266083in}}%
\pgfpathlineto{\pgfqpoint{0.707138in}{1.652794in}}%
\pgfpathlineto{\pgfqpoint{0.707138in}{1.652794in}}%
\pgfpathclose%
\pgfusepath{stroke,fill}%
}%
\begin{pgfscope}%
\pgfsys@transformshift{0.000000in}{0.000000in}%
\pgfsys@useobject{currentmarker}{}%
\end{pgfscope}%
\end{pgfscope}%
\begin{pgfscope}%
\pgfpathrectangle{\pgfqpoint{0.707138in}{0.642876in}}{\pgfqpoint{10.642862in}{3.707124in}}%
\pgfusepath{clip}%
\pgfsetbuttcap%
\pgfsetroundjoin%
\pgfsetlinewidth{2.007500pt}%
\definecolor{currentstroke}{rgb}{0.501961,0.501961,0.501961}%
\pgfsetstrokecolor{currentstroke}%
\pgfsetstrokeopacity{0.300000}%
\pgfsetdash{{7.400000pt}{3.200000pt}}{0.000000pt}%
\pgfpathmoveto{\pgfqpoint{0.707138in}{0.642876in}}%
\pgfpathlineto{\pgfqpoint{0.707138in}{4.350000in}}%
\pgfusepath{stroke}%
\end{pgfscope}%
\begin{pgfscope}%
\pgfsetbuttcap%
\pgfsetroundjoin%
\definecolor{currentfill}{rgb}{0.000000,0.000000,0.000000}%
\pgfsetfillcolor{currentfill}%
\pgfsetlinewidth{0.803000pt}%
\definecolor{currentstroke}{rgb}{0.000000,0.000000,0.000000}%
\pgfsetstrokecolor{currentstroke}%
\pgfsetdash{}{0pt}%
\pgfsys@defobject{currentmarker}{\pgfqpoint{0.000000in}{-0.048611in}}{\pgfqpoint{0.000000in}{0.000000in}}{%
\pgfpathmoveto{\pgfqpoint{0.000000in}{0.000000in}}%
\pgfpathlineto{\pgfqpoint{0.000000in}{-0.048611in}}%
\pgfusepath{stroke,fill}%
}%
\begin{pgfscope}%
\pgfsys@transformshift{0.707138in}{0.642876in}%
\pgfsys@useobject{currentmarker}{}%
\end{pgfscope}%
\end{pgfscope}%
\begin{pgfscope}%
\definecolor{textcolor}{rgb}{0.000000,0.000000,0.000000}%
\pgfsetstrokecolor{textcolor}%
\pgfsetfillcolor{textcolor}%
\pgftext[x=0.707138in,y=0.545653in,,top]{\color{textcolor}{\rmfamily\fontsize{16.000000}{19.200000}\selectfont\catcode`\^=\active\def^{\ifmmode\sp\else\^{}\fi}\catcode`\%=\active\def
\end{pgfscope}%
\begin{pgfscope}%
\pgfpathrectangle{\pgfqpoint{0.707138in}{0.642876in}}{\pgfqpoint{10.642862in}{3.707124in}}%
\pgfusepath{clip}%
\pgfsetbuttcap%
\pgfsetroundjoin%
\pgfsetlinewidth{2.007500pt}%
\definecolor{currentstroke}{rgb}{0.501961,0.501961,0.501961}%
\pgfsetstrokecolor{currentstroke}%
\pgfsetstrokeopacity{0.300000}%
\pgfsetdash{{7.400000pt}{3.200000pt}}{0.000000pt}%
\pgfpathmoveto{\pgfqpoint{3.020803in}{0.642876in}}%
\pgfpathlineto{\pgfqpoint{3.020803in}{4.350000in}}%
\pgfusepath{stroke}%
\end{pgfscope}%
\begin{pgfscope}%
\pgfsetbuttcap%
\pgfsetroundjoin%
\definecolor{currentfill}{rgb}{0.000000,0.000000,0.000000}%
\pgfsetfillcolor{currentfill}%
\pgfsetlinewidth{0.803000pt}%
\definecolor{currentstroke}{rgb}{0.000000,0.000000,0.000000}%
\pgfsetstrokecolor{currentstroke}%
\pgfsetdash{}{0pt}%
\pgfsys@defobject{currentmarker}{\pgfqpoint{0.000000in}{-0.048611in}}{\pgfqpoint{0.000000in}{0.000000in}}{%
\pgfpathmoveto{\pgfqpoint{0.000000in}{0.000000in}}%
\pgfpathlineto{\pgfqpoint{0.000000in}{-0.048611in}}%
\pgfusepath{stroke,fill}%
}%
\begin{pgfscope}%
\pgfsys@transformshift{3.020803in}{0.642876in}%
\pgfsys@useobject{currentmarker}{}%
\end{pgfscope}%
\end{pgfscope}%
\begin{pgfscope}%
\definecolor{textcolor}{rgb}{0.000000,0.000000,0.000000}%
\pgfsetstrokecolor{textcolor}%
\pgfsetfillcolor{textcolor}%
\pgftext[x=3.020803in,y=0.545653in,,top]{\color{textcolor}{\rmfamily\fontsize{16.000000}{19.200000}\selectfont\catcode`\^=\active\def^{\ifmmode\sp\else\^{}\fi}\catcode`\%=\active\def
\end{pgfscope}%
\begin{pgfscope}%
\pgfpathrectangle{\pgfqpoint{0.707138in}{0.642876in}}{\pgfqpoint{10.642862in}{3.707124in}}%
\pgfusepath{clip}%
\pgfsetbuttcap%
\pgfsetroundjoin%
\pgfsetlinewidth{2.007500pt}%
\definecolor{currentstroke}{rgb}{0.501961,0.501961,0.501961}%
\pgfsetstrokecolor{currentstroke}%
\pgfsetstrokeopacity{0.300000}%
\pgfsetdash{{7.400000pt}{3.200000pt}}{0.000000pt}%
\pgfpathmoveto{\pgfqpoint{5.334469in}{0.642876in}}%
\pgfpathlineto{\pgfqpoint{5.334469in}{4.350000in}}%
\pgfusepath{stroke}%
\end{pgfscope}%
\begin{pgfscope}%
\pgfsetbuttcap%
\pgfsetroundjoin%
\definecolor{currentfill}{rgb}{0.000000,0.000000,0.000000}%
\pgfsetfillcolor{currentfill}%
\pgfsetlinewidth{0.803000pt}%
\definecolor{currentstroke}{rgb}{0.000000,0.000000,0.000000}%
\pgfsetstrokecolor{currentstroke}%
\pgfsetdash{}{0pt}%
\pgfsys@defobject{currentmarker}{\pgfqpoint{0.000000in}{-0.048611in}}{\pgfqpoint{0.000000in}{0.000000in}}{%
\pgfpathmoveto{\pgfqpoint{0.000000in}{0.000000in}}%
\pgfpathlineto{\pgfqpoint{0.000000in}{-0.048611in}}%
\pgfusepath{stroke,fill}%
}%
\begin{pgfscope}%
\pgfsys@transformshift{5.334469in}{0.642876in}%
\pgfsys@useobject{currentmarker}{}%
\end{pgfscope}%
\end{pgfscope}%
\begin{pgfscope}%
\definecolor{textcolor}{rgb}{0.000000,0.000000,0.000000}%
\pgfsetstrokecolor{textcolor}%
\pgfsetfillcolor{textcolor}%
\pgftext[x=5.334469in,y=0.545653in,,top]{\color{textcolor}{\rmfamily\fontsize{16.000000}{19.200000}\selectfont\catcode`\^=\active\def^{\ifmmode\sp\else\^{}\fi}\catcode`\%=\active\def
\end{pgfscope}%
\begin{pgfscope}%
\pgfpathrectangle{\pgfqpoint{0.707138in}{0.642876in}}{\pgfqpoint{10.642862in}{3.707124in}}%
\pgfusepath{clip}%
\pgfsetbuttcap%
\pgfsetroundjoin%
\pgfsetlinewidth{2.007500pt}%
\definecolor{currentstroke}{rgb}{0.501961,0.501961,0.501961}%
\pgfsetstrokecolor{currentstroke}%
\pgfsetstrokeopacity{0.300000}%
\pgfsetdash{{7.400000pt}{3.200000pt}}{0.000000pt}%
\pgfpathmoveto{\pgfqpoint{7.648135in}{0.642876in}}%
\pgfpathlineto{\pgfqpoint{7.648135in}{4.350000in}}%
\pgfusepath{stroke}%
\end{pgfscope}%
\begin{pgfscope}%
\pgfsetbuttcap%
\pgfsetroundjoin%
\definecolor{currentfill}{rgb}{0.000000,0.000000,0.000000}%
\pgfsetfillcolor{currentfill}%
\pgfsetlinewidth{0.803000pt}%
\definecolor{currentstroke}{rgb}{0.000000,0.000000,0.000000}%
\pgfsetstrokecolor{currentstroke}%
\pgfsetdash{}{0pt}%
\pgfsys@defobject{currentmarker}{\pgfqpoint{0.000000in}{-0.048611in}}{\pgfqpoint{0.000000in}{0.000000in}}{%
\pgfpathmoveto{\pgfqpoint{0.000000in}{0.000000in}}%
\pgfpathlineto{\pgfqpoint{0.000000in}{-0.048611in}}%
\pgfusepath{stroke,fill}%
}%
\begin{pgfscope}%
\pgfsys@transformshift{7.648135in}{0.642876in}%
\pgfsys@useobject{currentmarker}{}%
\end{pgfscope}%
\end{pgfscope}%
\begin{pgfscope}%
\definecolor{textcolor}{rgb}{0.000000,0.000000,0.000000}%
\pgfsetstrokecolor{textcolor}%
\pgfsetfillcolor{textcolor}%
\pgftext[x=7.648135in,y=0.545653in,,top]{\color{textcolor}{\rmfamily\fontsize{16.000000}{19.200000}\selectfont\catcode`\^=\active\def^{\ifmmode\sp\else\^{}\fi}\catcode`\%=\active\def
\end{pgfscope}%
\begin{pgfscope}%
\pgfpathrectangle{\pgfqpoint{0.707138in}{0.642876in}}{\pgfqpoint{10.642862in}{3.707124in}}%
\pgfusepath{clip}%
\pgfsetbuttcap%
\pgfsetroundjoin%
\pgfsetlinewidth{2.007500pt}%
\definecolor{currentstroke}{rgb}{0.501961,0.501961,0.501961}%
\pgfsetstrokecolor{currentstroke}%
\pgfsetstrokeopacity{0.300000}%
\pgfsetdash{{7.400000pt}{3.200000pt}}{0.000000pt}%
\pgfpathmoveto{\pgfqpoint{9.961801in}{0.642876in}}%
\pgfpathlineto{\pgfqpoint{9.961801in}{4.350000in}}%
\pgfusepath{stroke}%
\end{pgfscope}%
\begin{pgfscope}%
\pgfsetbuttcap%
\pgfsetroundjoin%
\definecolor{currentfill}{rgb}{0.000000,0.000000,0.000000}%
\pgfsetfillcolor{currentfill}%
\pgfsetlinewidth{0.803000pt}%
\definecolor{currentstroke}{rgb}{0.000000,0.000000,0.000000}%
\pgfsetstrokecolor{currentstroke}%
\pgfsetdash{}{0pt}%
\pgfsys@defobject{currentmarker}{\pgfqpoint{0.000000in}{-0.048611in}}{\pgfqpoint{0.000000in}{0.000000in}}{%
\pgfpathmoveto{\pgfqpoint{0.000000in}{0.000000in}}%
\pgfpathlineto{\pgfqpoint{0.000000in}{-0.048611in}}%
\pgfusepath{stroke,fill}%
}%
\begin{pgfscope}%
\pgfsys@transformshift{9.961801in}{0.642876in}%
\pgfsys@useobject{currentmarker}{}%
\end{pgfscope}%
\end{pgfscope}%
\begin{pgfscope}%
\definecolor{textcolor}{rgb}{0.000000,0.000000,0.000000}%
\pgfsetstrokecolor{textcolor}%
\pgfsetfillcolor{textcolor}%
\pgftext[x=9.961801in,y=0.545653in,,top]{\color{textcolor}{\rmfamily\fontsize{16.000000}{19.200000}\selectfont\catcode`\^=\active\def^{\ifmmode\sp\else\^{}\fi}\catcode`\%=\active\def
\end{pgfscope}%
\begin{pgfscope}%
\definecolor{textcolor}{rgb}{0.000000,0.000000,0.000000}%
\pgfsetstrokecolor{textcolor}%
\pgfsetfillcolor{textcolor}%
\pgftext[x=6.028569in,y=0.295049in,,top]{\color{textcolor}{\rmfamily\fontsize{16.000000}{19.200000}\selectfont\catcode`\^=\active\def^{\ifmmode\sp\else\^{}\fi}\catcode`\%=\active\def
\end{pgfscope}%
\begin{pgfscope}%
\pgfpathrectangle{\pgfqpoint{0.707138in}{0.642876in}}{\pgfqpoint{10.642862in}{3.707124in}}%
\pgfusepath{clip}%
\pgfsetbuttcap%
\pgfsetroundjoin%
\pgfsetlinewidth{2.007500pt}%
\definecolor{currentstroke}{rgb}{0.501961,0.501961,0.501961}%
\pgfsetstrokecolor{currentstroke}%
\pgfsetstrokeopacity{0.300000}%
\pgfsetdash{{7.400000pt}{3.200000pt}}{0.000000pt}%
\pgfpathmoveto{\pgfqpoint{0.707138in}{0.835810in}}%
\pgfpathlineto{\pgfqpoint{11.350000in}{0.835810in}}%
\pgfusepath{stroke}%
\end{pgfscope}%
\begin{pgfscope}%
\pgfsetbuttcap%
\pgfsetroundjoin%
\definecolor{currentfill}{rgb}{0.000000,0.000000,0.000000}%
\pgfsetfillcolor{currentfill}%
\pgfsetlinewidth{0.803000pt}%
\definecolor{currentstroke}{rgb}{0.000000,0.000000,0.000000}%
\pgfsetstrokecolor{currentstroke}%
\pgfsetdash{}{0pt}%
\pgfsys@defobject{currentmarker}{\pgfqpoint{-0.048611in}{0.000000in}}{\pgfqpoint{-0.000000in}{0.000000in}}{%
\pgfpathmoveto{\pgfqpoint{-0.000000in}{0.000000in}}%
\pgfpathlineto{\pgfqpoint{-0.048611in}{0.000000in}}%
\pgfusepath{stroke,fill}%
}%
\begin{pgfscope}%
\pgfsys@transformshift{0.707138in}{0.835810in}%
\pgfsys@useobject{currentmarker}{}%
\end{pgfscope}%
\end{pgfscope}%
\begin{pgfscope}%
\definecolor{textcolor}{rgb}{0.000000,0.000000,0.000000}%
\pgfsetstrokecolor{textcolor}%
\pgfsetfillcolor{textcolor}%
\pgftext[x=0.350604in, y=0.759897in, left, base]{\color{textcolor}{\rmfamily\fontsize{16.000000}{19.200000}\selectfont\catcode`\^=\active\def^{\ifmmode\sp\else\^{}\fi}\catcode`\%=\active\def
\end{pgfscope}%
\begin{pgfscope}%
\pgfpathrectangle{\pgfqpoint{0.707138in}{0.642876in}}{\pgfqpoint{10.642862in}{3.707124in}}%
\pgfusepath{clip}%
\pgfsetbuttcap%
\pgfsetroundjoin%
\pgfsetlinewidth{2.007500pt}%
\definecolor{currentstroke}{rgb}{0.501961,0.501961,0.501961}%
\pgfsetstrokecolor{currentstroke}%
\pgfsetstrokeopacity{0.300000}%
\pgfsetdash{{7.400000pt}{3.200000pt}}{0.000000pt}%
\pgfpathmoveto{\pgfqpoint{0.707138in}{1.627576in}}%
\pgfpathlineto{\pgfqpoint{11.350000in}{1.627576in}}%
\pgfusepath{stroke}%
\end{pgfscope}%
\begin{pgfscope}%
\pgfsetbuttcap%
\pgfsetroundjoin%
\definecolor{currentfill}{rgb}{0.000000,0.000000,0.000000}%
\pgfsetfillcolor{currentfill}%
\pgfsetlinewidth{0.803000pt}%
\definecolor{currentstroke}{rgb}{0.000000,0.000000,0.000000}%
\pgfsetstrokecolor{currentstroke}%
\pgfsetdash{}{0pt}%
\pgfsys@defobject{currentmarker}{\pgfqpoint{-0.048611in}{0.000000in}}{\pgfqpoint{-0.000000in}{0.000000in}}{%
\pgfpathmoveto{\pgfqpoint{-0.000000in}{0.000000in}}%
\pgfpathlineto{\pgfqpoint{-0.048611in}{0.000000in}}%
\pgfusepath{stroke,fill}%
}%
\begin{pgfscope}%
\pgfsys@transformshift{0.707138in}{1.627576in}%
\pgfsys@useobject{currentmarker}{}%
\end{pgfscope}%
\end{pgfscope}%
\begin{pgfscope}%
\definecolor{textcolor}{rgb}{0.000000,0.000000,0.000000}%
\pgfsetstrokecolor{textcolor}%
\pgfsetfillcolor{textcolor}%
\pgftext[x=0.350604in, y=1.551663in, left, base]{\color{textcolor}{\rmfamily\fontsize{16.000000}{19.200000}\selectfont\catcode`\^=\active\def^{\ifmmode\sp\else\^{}\fi}\catcode`\%=\active\def
\end{pgfscope}%
\begin{pgfscope}%
\pgfpathrectangle{\pgfqpoint{0.707138in}{0.642876in}}{\pgfqpoint{10.642862in}{3.707124in}}%
\pgfusepath{clip}%
\pgfsetbuttcap%
\pgfsetroundjoin%
\pgfsetlinewidth{2.007500pt}%
\definecolor{currentstroke}{rgb}{0.501961,0.501961,0.501961}%
\pgfsetstrokecolor{currentstroke}%
\pgfsetstrokeopacity{0.300000}%
\pgfsetdash{{7.400000pt}{3.200000pt}}{0.000000pt}%
\pgfpathmoveto{\pgfqpoint{0.707138in}{2.419341in}}%
\pgfpathlineto{\pgfqpoint{11.350000in}{2.419341in}}%
\pgfusepath{stroke}%
\end{pgfscope}%
\begin{pgfscope}%
\pgfsetbuttcap%
\pgfsetroundjoin%
\definecolor{currentfill}{rgb}{0.000000,0.000000,0.000000}%
\pgfsetfillcolor{currentfill}%
\pgfsetlinewidth{0.803000pt}%
\definecolor{currentstroke}{rgb}{0.000000,0.000000,0.000000}%
\pgfsetstrokecolor{currentstroke}%
\pgfsetdash{}{0pt}%
\pgfsys@defobject{currentmarker}{\pgfqpoint{-0.048611in}{0.000000in}}{\pgfqpoint{-0.000000in}{0.000000in}}{%
\pgfpathmoveto{\pgfqpoint{-0.000000in}{0.000000in}}%
\pgfpathlineto{\pgfqpoint{-0.048611in}{0.000000in}}%
\pgfusepath{stroke,fill}%
}%
\begin{pgfscope}%
\pgfsys@transformshift{0.707138in}{2.419341in}%
\pgfsys@useobject{currentmarker}{}%
\end{pgfscope}%
\end{pgfscope}%
\begin{pgfscope}%
\definecolor{textcolor}{rgb}{0.000000,0.000000,0.000000}%
\pgfsetstrokecolor{textcolor}%
\pgfsetfillcolor{textcolor}%
\pgftext[x=0.350604in, y=2.343428in, left, base]{\color{textcolor}{\rmfamily\fontsize{16.000000}{19.200000}\selectfont\catcode`\^=\active\def^{\ifmmode\sp\else\^{}\fi}\catcode`\%=\active\def
\end{pgfscope}%
\begin{pgfscope}%
\pgfpathrectangle{\pgfqpoint{0.707138in}{0.642876in}}{\pgfqpoint{10.642862in}{3.707124in}}%
\pgfusepath{clip}%
\pgfsetbuttcap%
\pgfsetroundjoin%
\pgfsetlinewidth{2.007500pt}%
\definecolor{currentstroke}{rgb}{0.501961,0.501961,0.501961}%
\pgfsetstrokecolor{currentstroke}%
\pgfsetstrokeopacity{0.300000}%
\pgfsetdash{{7.400000pt}{3.200000pt}}{0.000000pt}%
\pgfpathmoveto{\pgfqpoint{0.707138in}{3.211107in}}%
\pgfpathlineto{\pgfqpoint{11.350000in}{3.211107in}}%
\pgfusepath{stroke}%
\end{pgfscope}%
\begin{pgfscope}%
\pgfsetbuttcap%
\pgfsetroundjoin%
\definecolor{currentfill}{rgb}{0.000000,0.000000,0.000000}%
\pgfsetfillcolor{currentfill}%
\pgfsetlinewidth{0.803000pt}%
\definecolor{currentstroke}{rgb}{0.000000,0.000000,0.000000}%
\pgfsetstrokecolor{currentstroke}%
\pgfsetdash{}{0pt}%
\pgfsys@defobject{currentmarker}{\pgfqpoint{-0.048611in}{0.000000in}}{\pgfqpoint{-0.000000in}{0.000000in}}{%
\pgfpathmoveto{\pgfqpoint{-0.000000in}{0.000000in}}%
\pgfpathlineto{\pgfqpoint{-0.048611in}{0.000000in}}%
\pgfusepath{stroke,fill}%
}%
\begin{pgfscope}%
\pgfsys@transformshift{0.707138in}{3.211107in}%
\pgfsys@useobject{currentmarker}{}%
\end{pgfscope}%
\end{pgfscope}%
\begin{pgfscope}%
\definecolor{textcolor}{rgb}{0.000000,0.000000,0.000000}%
\pgfsetstrokecolor{textcolor}%
\pgfsetfillcolor{textcolor}%
\pgftext[x=0.350604in, y=3.135193in, left, base]{\color{textcolor}{\rmfamily\fontsize{16.000000}{19.200000}\selectfont\catcode`\^=\active\def^{\ifmmode\sp\else\^{}\fi}\catcode`\%=\active\def
\end{pgfscope}%
\begin{pgfscope}%
\pgfpathrectangle{\pgfqpoint{0.707138in}{0.642876in}}{\pgfqpoint{10.642862in}{3.707124in}}%
\pgfusepath{clip}%
\pgfsetbuttcap%
\pgfsetroundjoin%
\pgfsetlinewidth{2.007500pt}%
\definecolor{currentstroke}{rgb}{0.501961,0.501961,0.501961}%
\pgfsetstrokecolor{currentstroke}%
\pgfsetstrokeopacity{0.300000}%
\pgfsetdash{{7.400000pt}{3.200000pt}}{0.000000pt}%
\pgfpathmoveto{\pgfqpoint{0.707138in}{4.002872in}}%
\pgfpathlineto{\pgfqpoint{11.350000in}{4.002872in}}%
\pgfusepath{stroke}%
\end{pgfscope}%
\begin{pgfscope}%
\pgfsetbuttcap%
\pgfsetroundjoin%
\definecolor{currentfill}{rgb}{0.000000,0.000000,0.000000}%
\pgfsetfillcolor{currentfill}%
\pgfsetlinewidth{0.803000pt}%
\definecolor{currentstroke}{rgb}{0.000000,0.000000,0.000000}%
\pgfsetstrokecolor{currentstroke}%
\pgfsetdash{}{0pt}%
\pgfsys@defobject{currentmarker}{\pgfqpoint{-0.048611in}{0.000000in}}{\pgfqpoint{-0.000000in}{0.000000in}}{%
\pgfpathmoveto{\pgfqpoint{-0.000000in}{0.000000in}}%
\pgfpathlineto{\pgfqpoint{-0.048611in}{0.000000in}}%
\pgfusepath{stroke,fill}%
}%
\begin{pgfscope}%
\pgfsys@transformshift{0.707138in}{4.002872in}%
\pgfsys@useobject{currentmarker}{}%
\end{pgfscope}%
\end{pgfscope}%
\begin{pgfscope}%
\definecolor{textcolor}{rgb}{0.000000,0.000000,0.000000}%
\pgfsetstrokecolor{textcolor}%
\pgfsetfillcolor{textcolor}%
\pgftext[x=0.350604in, y=3.926959in, left, base]{\color{textcolor}{\rmfamily\fontsize{16.000000}{19.200000}\selectfont\catcode`\^=\active\def^{\ifmmode\sp\else\^{}\fi}\catcode`\%=\active\def
\end{pgfscope}%
\begin{pgfscope}%
\definecolor{textcolor}{rgb}{0.000000,0.000000,0.000000}%
\pgfsetstrokecolor{textcolor}%
\pgfsetfillcolor{textcolor}%
\pgftext[x=0.295049in,y=2.496438in,,bottom,rotate=90.000000]{\color{textcolor}{\rmfamily\fontsize{16.000000}{19.200000}\selectfont\catcode`\^=\active\def^{\ifmmode\sp\else\^{}\fi}\catcode`\%=\active\def
\end{pgfscope}%
\begin{pgfscope}%
\pgfpathrectangle{\pgfqpoint{0.707138in}{0.642876in}}{\pgfqpoint{10.642862in}{3.707124in}}%
\pgfusepath{clip}%
\pgfsetrectcap%
\pgfsetroundjoin%
\pgfsetlinewidth{1.505625pt}%
\definecolor{currentstroke}{rgb}{0.121569,0.466667,0.705882}%
\pgfsetstrokecolor{currentstroke}%
\pgfsetdash{}{0pt}%
\pgfpathmoveto{\pgfqpoint{0.707138in}{1.420998in}}%
\pgfpathlineto{\pgfqpoint{1.169871in}{1.908457in}}%
\pgfpathlineto{\pgfqpoint{1.632604in}{2.019852in}}%
\pgfpathlineto{\pgfqpoint{2.095337in}{1.953893in}}%
\pgfpathlineto{\pgfqpoint{2.558070in}{1.742473in}}%
\pgfpathlineto{\pgfqpoint{3.020803in}{1.862455in}}%
\pgfpathlineto{\pgfqpoint{3.483536in}{1.814298in}}%
\pgfpathlineto{\pgfqpoint{3.946270in}{1.793281in}}%
\pgfpathlineto{\pgfqpoint{4.409003in}{1.799003in}}%
\pgfpathlineto{\pgfqpoint{4.871736in}{1.800417in}}%
\pgfpathlineto{\pgfqpoint{5.334469in}{1.759692in}}%
\pgfpathlineto{\pgfqpoint{5.797202in}{1.635038in}}%
\pgfpathlineto{\pgfqpoint{6.259935in}{1.598832in}}%
\pgfpathlineto{\pgfqpoint{6.722668in}{1.434455in}}%
\pgfpathlineto{\pgfqpoint{7.185402in}{1.232343in}}%
\pgfpathlineto{\pgfqpoint{7.648135in}{1.277422in}}%
\pgfpathlineto{\pgfqpoint{8.110868in}{1.359481in}}%
\pgfpathlineto{\pgfqpoint{8.573601in}{1.202629in}}%
\pgfpathlineto{\pgfqpoint{9.036334in}{1.116139in}}%
\pgfpathlineto{\pgfqpoint{9.499067in}{1.134138in}}%
\pgfpathlineto{\pgfqpoint{9.961801in}{1.182023in}}%
\pgfpathlineto{\pgfqpoint{10.424534in}{1.139173in}}%
\pgfpathlineto{\pgfqpoint{10.887267in}{1.065356in}}%
\pgfpathlineto{\pgfqpoint{11.350000in}{0.938651in}}%
\pgfusepath{stroke}%
\end{pgfscope}%
\begin{pgfscope}%
\pgfpathrectangle{\pgfqpoint{0.707138in}{0.642876in}}{\pgfqpoint{10.642862in}{3.707124in}}%
\pgfusepath{clip}%
\pgfsetrectcap%
\pgfsetroundjoin%
\pgfsetlinewidth{1.505625pt}%
\definecolor{currentstroke}{rgb}{1.000000,0.498039,0.054902}%
\pgfsetstrokecolor{currentstroke}%
\pgfsetdash{}{0pt}%
\pgfpathmoveto{\pgfqpoint{0.707138in}{1.419785in}}%
\pgfpathlineto{\pgfqpoint{1.169871in}{1.972585in}}%
\pgfpathlineto{\pgfqpoint{1.632604in}{2.062455in}}%
\pgfpathlineto{\pgfqpoint{2.095337in}{2.001329in}}%
\pgfpathlineto{\pgfqpoint{2.558070in}{1.638590in}}%
\pgfpathlineto{\pgfqpoint{3.020803in}{1.851143in}}%
\pgfpathlineto{\pgfqpoint{3.483536in}{1.783299in}}%
\pgfpathlineto{\pgfqpoint{3.946270in}{1.723998in}}%
\pgfpathlineto{\pgfqpoint{4.409003in}{1.710332in}}%
\pgfpathlineto{\pgfqpoint{4.871736in}{1.643309in}}%
\pgfpathlineto{\pgfqpoint{5.334469in}{1.585507in}}%
\pgfpathlineto{\pgfqpoint{5.797202in}{1.602453in}}%
\pgfpathlineto{\pgfqpoint{6.259935in}{1.677854in}}%
\pgfpathlineto{\pgfqpoint{6.722668in}{1.802483in}}%
\pgfpathlineto{\pgfqpoint{7.185402in}{2.150236in}}%
\pgfpathlineto{\pgfqpoint{7.648135in}{2.144159in}}%
\pgfpathlineto{\pgfqpoint{8.110868in}{2.247703in}}%
\pgfpathlineto{\pgfqpoint{8.573601in}{2.295756in}}%
\pgfpathlineto{\pgfqpoint{9.036334in}{2.731129in}}%
\pgfpathlineto{\pgfqpoint{9.499067in}{2.531573in}}%
\pgfpathlineto{\pgfqpoint{9.961801in}{2.643473in}}%
\pgfpathlineto{\pgfqpoint{10.424534in}{2.528515in}}%
\pgfpathlineto{\pgfqpoint{10.887267in}{2.673200in}}%
\pgfpathlineto{\pgfqpoint{11.350000in}{3.083495in}}%
\pgfusepath{stroke}%
\end{pgfscope}%
\begin{pgfscope}%
\pgfsetrectcap%
\pgfsetmiterjoin%
\pgfsetlinewidth{2.007500pt}%
\definecolor{currentstroke}{rgb}{0.000000,0.000000,0.000000}%
\pgfsetstrokecolor{currentstroke}%
\pgfsetdash{}{0pt}%
\pgfpathmoveto{\pgfqpoint{0.707138in}{0.642876in}}%
\pgfpathlineto{\pgfqpoint{0.707138in}{4.350000in}}%
\pgfusepath{stroke}%
\end{pgfscope}%
\begin{pgfscope}%
\pgfsetrectcap%
\pgfsetmiterjoin%
\pgfsetlinewidth{2.007500pt}%
\definecolor{currentstroke}{rgb}{0.000000,0.000000,0.000000}%
\pgfsetstrokecolor{currentstroke}%
\pgfsetdash{}{0pt}%
\pgfpathmoveto{\pgfqpoint{0.707138in}{0.642876in}}%
\pgfpathlineto{\pgfqpoint{11.350000in}{0.642876in}}%
\pgfusepath{stroke}%
\end{pgfscope}%
\begin{pgfscope}%
\pgfsetbuttcap%
\pgfsetmiterjoin%
\definecolor{currentfill}{rgb}{1.000000,1.000000,1.000000}%
\pgfsetfillcolor{currentfill}%
\pgfsetfillopacity{0.800000}%
\pgfsetlinewidth{1.003750pt}%
\definecolor{currentstroke}{rgb}{0.800000,0.800000,0.800000}%
\pgfsetstrokecolor{currentstroke}%
\pgfsetstrokeopacity{0.800000}%
\pgfsetdash{}{0pt}%
\pgfpathmoveto{\pgfqpoint{0.843249in}{3.652161in}}%
\pgfpathlineto{\pgfqpoint{2.779599in}{3.652161in}}%
\pgfpathquadraticcurveto{\pgfqpoint{2.818488in}{3.652161in}}{\pgfqpoint{2.818488in}{3.691050in}}%
\pgfpathlineto{\pgfqpoint{2.818488in}{4.213889in}}%
\pgfpathquadraticcurveto{\pgfqpoint{2.818488in}{4.252778in}}{\pgfqpoint{2.779599in}{4.252778in}}%
\pgfpathlineto{\pgfqpoint{0.843249in}{4.252778in}}%
\pgfpathquadraticcurveto{\pgfqpoint{0.804360in}{4.252778in}}{\pgfqpoint{0.804360in}{4.213889in}}%
\pgfpathlineto{\pgfqpoint{0.804360in}{3.691050in}}%
\pgfpathquadraticcurveto{\pgfqpoint{0.804360in}{3.652161in}}{\pgfqpoint{0.843249in}{3.652161in}}%
\pgfpathlineto{\pgfqpoint{0.843249in}{3.652161in}}%
\pgfpathclose%
\pgfusepath{stroke,fill}%
\end{pgfscope}%
\begin{pgfscope}%
\pgfsetrectcap%
\pgfsetroundjoin%
\pgfsetlinewidth{1.505625pt}%
\definecolor{currentstroke}{rgb}{0.121569,0.466667,0.705882}%
\pgfsetstrokecolor{currentstroke}%
\pgfsetdash{}{0pt}%
\pgfpathmoveto{\pgfqpoint{0.882138in}{4.106944in}}%
\pgfpathlineto{\pgfqpoint{1.076582in}{4.106944in}}%
\pgfpathlineto{\pgfqpoint{1.271026in}{4.106944in}}%
\pgfusepath{stroke}%
\end{pgfscope}%
\begin{pgfscope}%
\definecolor{textcolor}{rgb}{0.000000,0.000000,0.000000}%
\pgfsetstrokecolor{textcolor}%
\pgfsetfillcolor{textcolor}%
\pgftext[x=1.426582in,y=4.038889in,left,base]{\color{textcolor}{\rmfamily\fontsize{14.000000}{16.800000}\selectfont\catcode`\^=\active\def^{\ifmmode\sp\else\^{}\fi}\catcode`\%=\active\def
\end{pgfscope}%
\begin{pgfscope}%
\pgfsetrectcap%
\pgfsetroundjoin%
\pgfsetlinewidth{1.505625pt}%
\definecolor{currentstroke}{rgb}{1.000000,0.498039,0.054902}%
\pgfsetstrokecolor{currentstroke}%
\pgfsetdash{}{0pt}%
\pgfpathmoveto{\pgfqpoint{0.882138in}{3.835803in}}%
\pgfpathlineto{\pgfqpoint{1.076582in}{3.835803in}}%
\pgfpathlineto{\pgfqpoint{1.271026in}{3.835803in}}%
\pgfusepath{stroke}%
\end{pgfscope}%
\begin{pgfscope}%
\definecolor{textcolor}{rgb}{0.000000,0.000000,0.000000}%
\pgfsetstrokecolor{textcolor}%
\pgfsetfillcolor{textcolor}%
\pgftext[x=1.426582in,y=3.767747in,left,base]{\color{textcolor}{\rmfamily\fontsize{14.000000}{16.800000}\selectfont\catcode`\^=\active\def^{\ifmmode\sp\else\^{}\fi}\catcode`\%=\active\def
\end{pgfscope}%
\end{pgfpicture}%
\makeatother%
\endgroup%

%% file: tables/metrics_ablation.tex
\begin{table}[t]
  \centering
  \caption{
  This table shows the camera pose estimation metrics average and  95\% confidence interval for the different ablation studies of our paper.}
  \setlength{\tabcolsep}{6pt}
  \begin{tabular}{lccc}
    \toprule Method & AUC@30 & RRA@30 & RTA@30\\ \toprule
    No Camera Token & $67.8 \pm 1.4$ & $89.1 \pm 1.0$ & $86.6 \pm 1.3$ \\
    Thermal Projector & $67.2 \pm 1.1$ & $88.3 \pm 0.8$ & $86.0 \pm 0.8$ \\
    \makecell{Thermal Embedding} & $66.1 \pm 1.8$ & $87.6 \pm 1.0$ & $85.1 \pm 1.2$ \\\midrule
    Global Only & $69.1 \pm 1.5$ & $89.4 \pm 1.1$ & $87.3 \pm 1.6$ \\
    Frame Only & $62.4 \pm 0.6$ & $85.6 \pm 0.9$ & $83.0 \pm 1.0$ \\
    \ours & $68.4 \pm 0.7$ & $89.1 \pm 0.5$ & $86.9 \pm 0.4$ \\\bottomrule
  \end{tabular}
  \label{tab:ablation_study}
\end{table}

%% file: images/statistical_significance.tex
\begin{table}[t]
  \caption{\textbf{Pairwise statistical significance (AUC@30).} Each entry at position $(i,j)$ reports the p-value from a one-sided Welch's t-test for the null hypothesis that the row method outperforms the column method in AUC@30. Darker blue indicates smaller p-values (stronger evidence against the null).}
  \vspace{-5mm}
  \label{tab:pairwise-significance}
  \centering
  \renewcommand{\arraystretch}{1.20}
  \setlength{\extrarowheight}{1pt}
  \scalebox{0.8}{
    \begin{tabular}{>{\raggedright\arraybackslash}m{3.0cm}*{6}{>{\centering\arraybackslash}m{1.6cm}}}
      & \makecell{No Camera\\Token}
      & \makecell{Thermal\\Projector}
      & \makecell{Thermal\\Embedding}
      & Global Only
      & Frame Only
      & Ours \\
      \midrule
      No Camera Token
      & \cellcolor[HTML]{EFEFEF}{--} & \cellcolor[HTML]{8EB5D9}{0.72} & \cellcolor[HTML]{DCEAF6}{0.91} & \cellcolor[HTML]{2D6BAA}\textcolor{white}{0.12} & \cellcolor[HTML]{DCEAF6}{1.00} & \cellcolor[HTML]{2D6BAA}\textcolor{white}{0.24} \\
      Thermal Projector
      & \cellcolor[HTML]{2D6BAA}\textcolor{white}{0.28} & \cellcolor[HTML]{EFEFEF}{--} & \cellcolor[HTML]{8EB5D9}{0.83} & \cellcolor[HTML]{123B7A}\textcolor{white}{0.04} & \cellcolor[HTML]{DCEAF6}{1.00} & \cellcolor[HTML]{123B7A}\textcolor{white}{0.05} \\
      \makecell[l]{Thermal\\Embedding}
      & \cellcolor[HTML]{123B7A}\textcolor{white}{0.09} & \cellcolor[HTML]{2D6BAA}\textcolor{white}{0.17} & \cellcolor[HTML]{EFEFEF}{--} & \cellcolor[HTML]{123B7A}\textcolor{white}{0.02} & \cellcolor[HTML]{DCEAF6}{1.00} & \cellcolor[HTML]{123B7A}\textcolor{white}{0.02}\rule{0pt}{3.2ex} \\
      Global Only
      & \cellcolor[HTML]{8EB5D9}{0.88} & \cellcolor[HTML]{DCEAF6}{0.96} & \cellcolor[HTML]{DCEAF6}{0.98} & \cellcolor[HTML]{EFEFEF}{--} & \cellcolor[HTML]{DCEAF6}{1.00} & \cellcolor[HTML]{8EB5D9}{0.78} \\
      Frame Only
      & \cellcolor[HTML]{123B7A}\textcolor{white}{0.00} & \cellcolor[HTML]{123B7A}\textcolor{white}{0.00} & \cellcolor[HTML]{123B7A}\textcolor{white}{0.00} & \cellcolor[HTML]{123B7A}\textcolor{white}{0.00} & \cellcolor[HTML]{EFEFEF}{--} & \cellcolor[HTML]{123B7A}\textcolor{white}{0.00} \\
      Ours
      & \cellcolor[HTML]{8EB5D9}{0.76} & \cellcolor[HTML]{DCEAF6}{0.95} & \cellcolor[HTML]{DCEAF6}{0.98} & \cellcolor[HTML]{2D6BAA}\textcolor{white}{0.22} & \cellcolor[HTML]{DCEAF6}{1.00} & \cellcolor[HTML]{EFEFEF}{--} \\
      \bottomrule
    \end{tabular}
  }
  \vspace{-5mm}
\end{table}

%% file: images/lora_r_study/figure.tex
\begin{figure*}[t]
  \centering
  \begin{subfigure}[b]{0.32\textwidth}
    \centering
    \resizebox{\linewidth}{!}{\input{images/lora_r_study/r_results_mAA.pgf}}
    \caption{AUC $\uparrow$}
    \label{fig:thermal_ratio_maa}
  \end{subfigure}
  \begin{subfigure}[b]{0.32\textwidth}
    \centering
    \resizebox{\linewidth}{!}{\input{images/lora_r_study/r_results_RRA.pgf}}
    \caption{RRA $\uparrow$}
    \label{fig:thermal_ratio_rra}
  \end{subfigure}
  \begin{subfigure}[b]{0.32\textwidth}
    \centering
    \resizebox{\linewidth}{!}{\input{images/lora_r_study/r_results_RTA.pgf}}
    \caption{RTA $\uparrow$}
    \label{fig:thermal_ratio_rta}
  \end{subfigure}

  \caption{
    The AUC, RRA, RTA (errors <$30^\circ$, $15^\circ$, $5^\circ$) across varying LoRA $r$. The filled area represents the boundary from $0.25-$ to $0.75-$quantiles estimated by bootstrapping scenes $2000$ times.
  }
  \label{fig:lora_r}
  \vspace{-6mm}
\end{figure*}

%% file: images/lora_r_study/r_results_mAA.pgf
\begingroup%
\makeatletter%
\begin{pgfpicture}%
\pgfpathrectangle{\pgfpointorigin}{\pgfqpoint{7.450000in}{5.450000in}}%
\pgfusepath{use as bounding box, clip}%
\begin{pgfscope}%
\pgfsetbuttcap%
\pgfsetmiterjoin%
\definecolor{currentfill}{rgb}{1.000000,1.000000,1.000000}%
\pgfsetfillcolor{currentfill}%
\pgfsetlinewidth{0.000000pt}%
\definecolor{currentstroke}{rgb}{1.000000,1.000000,1.000000}%
\pgfsetstrokecolor{currentstroke}%
\pgfsetdash{}{0pt}%
\pgfpathmoveto{\pgfqpoint{0.000000in}{0.000000in}}%
\pgfpathlineto{\pgfqpoint{7.450000in}{0.000000in}}%
\pgfpathlineto{\pgfqpoint{7.450000in}{5.450000in}}%
\pgfpathlineto{\pgfqpoint{0.000000in}{5.450000in}}%
\pgfpathlineto{\pgfqpoint{0.000000in}{0.000000in}}%
\pgfpathclose%
\pgfusepath{fill}%
\end{pgfscope}%
\begin{pgfscope}%
\pgfsetbuttcap%
\pgfsetmiterjoin%
\definecolor{currentfill}{rgb}{1.000000,1.000000,1.000000}%
\pgfsetfillcolor{currentfill}%
\pgfsetlinewidth{0.000000pt}%
\definecolor{currentstroke}{rgb}{0.000000,0.000000,0.000000}%
\pgfsetstrokecolor{currentstroke}%
\pgfsetstrokeopacity{0.000000}%
\pgfsetdash{}{0pt}%
\pgfpathmoveto{\pgfqpoint{0.674954in}{0.862305in}}%
\pgfpathlineto{\pgfqpoint{7.190756in}{0.862305in}}%
\pgfpathlineto{\pgfqpoint{7.190756in}{5.231386in}}%
\pgfpathlineto{\pgfqpoint{0.674954in}{5.231386in}}%
\pgfpathlineto{\pgfqpoint{0.674954in}{0.862305in}}%
\pgfpathclose%
\pgfusepath{fill}%
\end{pgfscope}%
\begin{pgfscope}%
\pgfpathrectangle{\pgfqpoint{0.674954in}{0.862305in}}{\pgfqpoint{6.515802in}{4.369081in}}%
\pgfusepath{clip}%
\pgfsetbuttcap%
\pgfsetroundjoin%
\definecolor{currentfill}{rgb}{0.050980,0.415686,0.509804}%
\pgfsetfillcolor{currentfill}%
\pgfsetfillopacity{0.300000}%
\pgfsetlinewidth{1.003750pt}%
\definecolor{currentstroke}{rgb}{0.050980,0.415686,0.509804}%
\pgfsetstrokecolor{currentstroke}%
\pgfsetstrokeopacity{0.300000}%
\pgfsetdash{}{0pt}%
\pgfsys@defobject{currentmarker}{\pgfqpoint{0.674954in}{3.441278in}}{\pgfqpoint{7.190756in}{4.158651in}}{%
\pgfpathmoveto{\pgfqpoint{0.674954in}{3.785295in}}%
\pgfpathlineto{\pgfqpoint{0.674954in}{3.469042in}}%
\pgfpathlineto{\pgfqpoint{1.760921in}{3.619640in}}%
\pgfpathlineto{\pgfqpoint{2.846888in}{3.500736in}}%
\pgfpathlineto{\pgfqpoint{3.932855in}{3.441278in}}%
\pgfpathlineto{\pgfqpoint{5.018822in}{3.801395in}}%
\pgfpathlineto{\pgfqpoint{6.104789in}{3.642435in}}%
\pgfpathlineto{\pgfqpoint{6.740039in}{3.767442in}}%
\pgfpathlineto{\pgfqpoint{7.190756in}{3.760694in}}%
\pgfpathlineto{\pgfqpoint{7.190756in}{4.158651in}}%
\pgfpathlineto{\pgfqpoint{7.190756in}{4.158651in}}%
\pgfpathlineto{\pgfqpoint{6.740039in}{4.078813in}}%
\pgfpathlineto{\pgfqpoint{6.104789in}{3.952809in}}%
\pgfpathlineto{\pgfqpoint{5.018822in}{4.114106in}}%
\pgfpathlineto{\pgfqpoint{3.932855in}{3.739457in}}%
\pgfpathlineto{\pgfqpoint{2.846888in}{3.852493in}}%
\pgfpathlineto{\pgfqpoint{1.760921in}{3.898922in}}%
\pgfpathlineto{\pgfqpoint{0.674954in}{3.785295in}}%
\pgfpathlineto{\pgfqpoint{0.674954in}{3.785295in}}%
\pgfpathclose%
\pgfusepath{stroke,fill}%
}%
\begin{pgfscope}%
\pgfsys@transformshift{0.000000in}{0.000000in}%
\pgfsys@useobject{currentmarker}{}%
\end{pgfscope}%
\end{pgfscope}%
\begin{pgfscope}%
\pgfpathrectangle{\pgfqpoint{0.674954in}{0.862305in}}{\pgfqpoint{6.515802in}{4.369081in}}%
\pgfusepath{clip}%
\pgfsetbuttcap%
\pgfsetroundjoin%
\definecolor{currentfill}{rgb}{0.960784,0.462745,0.000000}%
\pgfsetfillcolor{currentfill}%
\pgfsetfillopacity{0.300000}%
\pgfsetlinewidth{1.003750pt}%
\definecolor{currentstroke}{rgb}{0.960784,0.462745,0.000000}%
\pgfsetstrokecolor{currentstroke}%
\pgfsetstrokeopacity{0.300000}%
\pgfsetdash{{3.700000pt}{1.600000pt}}{0.000000pt}%
\pgfpathmoveto{\pgfqpoint{0.674954in}{3.214120in}}%
\pgfpathlineto{\pgfqpoint{0.674954in}{2.905114in}}%
\pgfpathlineto{\pgfqpoint{1.760921in}{3.036691in}}%
\pgfpathlineto{\pgfqpoint{2.846888in}{3.004124in}}%
\pgfpathlineto{\pgfqpoint{3.932855in}{2.851625in}}%
\pgfpathlineto{\pgfqpoint{5.018822in}{3.252043in}}%
\pgfpathlineto{\pgfqpoint{6.104789in}{3.061328in}}%
\pgfpathlineto{\pgfqpoint{6.740039in}{3.217445in}}%
\pgfpathlineto{\pgfqpoint{7.190756in}{3.241529in}}%
\pgfpathlineto{\pgfqpoint{7.190756in}{3.645706in}}%
\pgfpathlineto{\pgfqpoint{7.190756in}{3.645706in}}%
\pgfpathlineto{\pgfqpoint{6.740039in}{3.519573in}}%
\pgfpathlineto{\pgfqpoint{6.104789in}{3.376709in}}%
\pgfpathlineto{\pgfqpoint{5.018822in}{3.593244in}}%
\pgfpathlineto{\pgfqpoint{3.932855in}{3.156219in}}%
\pgfpathlineto{\pgfqpoint{2.846888in}{3.361331in}}%
\pgfpathlineto{\pgfqpoint{1.760921in}{3.322555in}}%
\pgfpathlineto{\pgfqpoint{0.674954in}{3.214120in}}%
\pgfpathlineto{\pgfqpoint{0.674954in}{3.214120in}}%
\pgfpathclose%
\pgfusepath{stroke,fill}%
\end{pgfscope}%
\begin{pgfscope}%
\pgfpathrectangle{\pgfqpoint{0.674954in}{0.862305in}}{\pgfqpoint{6.515802in}{4.369081in}}%
\pgfusepath{clip}%
\pgfsetbuttcap%
\pgfsetroundjoin%
\definecolor{currentfill}{rgb}{0.219608,0.219608,0.219608}%
\pgfsetfillcolor{currentfill}%
\pgfsetfillopacity{0.300000}%
\pgfsetlinewidth{1.003750pt}%
\definecolor{currentstroke}{rgb}{0.219608,0.219608,0.219608}%
\pgfsetstrokecolor{currentstroke}%
\pgfsetstrokeopacity{0.300000}%
\pgfsetdash{{1.000000pt}{1.650000pt}}{0.000000pt}%
\pgfpathmoveto{\pgfqpoint{0.674954in}{2.141585in}}%
\pgfpathlineto{\pgfqpoint{0.674954in}{1.900751in}}%
\pgfpathlineto{\pgfqpoint{1.760921in}{1.947936in}}%
\pgfpathlineto{\pgfqpoint{2.846888in}{2.089938in}}%
\pgfpathlineto{\pgfqpoint{3.932855in}{1.894958in}}%
\pgfpathlineto{\pgfqpoint{5.018822in}{2.192546in}}%
\pgfpathlineto{\pgfqpoint{6.104789in}{2.035871in}}%
\pgfpathlineto{\pgfqpoint{6.740039in}{2.135038in}}%
\pgfpathlineto{\pgfqpoint{7.190756in}{2.181205in}}%
\pgfpathlineto{\pgfqpoint{7.190756in}{2.536187in}}%
\pgfpathlineto{\pgfqpoint{7.190756in}{2.536187in}}%
\pgfpathlineto{\pgfqpoint{6.740039in}{2.392125in}}%
\pgfpathlineto{\pgfqpoint{6.104789in}{2.285382in}}%
\pgfpathlineto{\pgfqpoint{5.018822in}{2.500270in}}%
\pgfpathlineto{\pgfqpoint{3.932855in}{2.152082in}}%
\pgfpathlineto{\pgfqpoint{2.846888in}{2.369406in}}%
\pgfpathlineto{\pgfqpoint{1.760921in}{2.210945in}}%
\pgfpathlineto{\pgfqpoint{0.674954in}{2.141585in}}%
\pgfpathlineto{\pgfqpoint{0.674954in}{2.141585in}}%
\pgfpathclose%
\pgfusepath{stroke,fill}%
\end{pgfscope}%
\begin{pgfscope}%
\pgfpathrectangle{\pgfqpoint{0.674954in}{0.862305in}}{\pgfqpoint{6.515802in}{4.369081in}}%
\pgfusepath{clip}%
\pgfsetbuttcap%
\pgfsetroundjoin%
\pgfsetlinewidth{2.007500pt}%
\definecolor{currentstroke}{rgb}{0.501961,0.501961,0.501961}%
\pgfsetstrokecolor{currentstroke}%
\pgfsetstrokeopacity{0.300000}%
\pgfsetdash{{7.400000pt}{3.200000pt}}{0.000000pt}%
\pgfpathmoveto{\pgfqpoint{0.674954in}{0.862305in}}%
\pgfpathlineto{\pgfqpoint{0.674954in}{5.231386in}}%
\pgfusepath{stroke}%
\end{pgfscope}%
\begin{pgfscope}%
\pgfsetbuttcap%
\pgfsetroundjoin%
\definecolor{currentfill}{rgb}{0.000000,0.000000,0.000000}%
\pgfsetfillcolor{currentfill}%
\pgfsetlinewidth{0.803000pt}%
\definecolor{currentstroke}{rgb}{0.000000,0.000000,0.000000}%
\pgfsetstrokecolor{currentstroke}%
\pgfsetdash{}{0pt}%
\pgfsys@defobject{currentmarker}{\pgfqpoint{0.000000in}{-0.048611in}}{\pgfqpoint{0.000000in}{0.000000in}}{%
\pgfpathmoveto{\pgfqpoint{0.000000in}{0.000000in}}%
\pgfpathlineto{\pgfqpoint{0.000000in}{-0.048611in}}%
\pgfusepath{stroke,fill}%
}%
\begin{pgfscope}%
\pgfsys@transformshift{0.674954in}{0.862305in}%
\pgfsys@useobject{currentmarker}{}%
\end{pgfscope}%
\end{pgfscope}%
\begin{pgfscope}%
\definecolor{textcolor}{rgb}{0.000000,0.000000,0.000000}%
\pgfsetstrokecolor{textcolor}%
\pgfsetfillcolor{textcolor}%
\pgftext[x=0.674954in,y=0.765082in,,top]{\color{textcolor}{\rmfamily\fontsize{25.000000}{30.000000}\selectfont\catcode`\^=\active\def^{\ifmmode\sp\else\^{}\fi}\catcode`\%=\active\def
\end{pgfscope}%
\begin{pgfscope}%
\pgfpathrectangle{\pgfqpoint{0.674954in}{0.862305in}}{\pgfqpoint{6.515802in}{4.369081in}}%
\pgfusepath{clip}%
\pgfsetbuttcap%
\pgfsetroundjoin%
\pgfsetlinewidth{2.007500pt}%
\definecolor{currentstroke}{rgb}{0.501961,0.501961,0.501961}%
\pgfsetstrokecolor{currentstroke}%
\pgfsetstrokeopacity{0.300000}%
\pgfsetdash{{7.400000pt}{3.200000pt}}{0.000000pt}%
\pgfpathmoveto{\pgfqpoint{1.760921in}{0.862305in}}%
\pgfpathlineto{\pgfqpoint{1.760921in}{5.231386in}}%
\pgfusepath{stroke}%
\end{pgfscope}%
\begin{pgfscope}%
\pgfsetbuttcap%
\pgfsetroundjoin%
\definecolor{currentfill}{rgb}{0.000000,0.000000,0.000000}%
\pgfsetfillcolor{currentfill}%
\pgfsetlinewidth{0.803000pt}%
\definecolor{currentstroke}{rgb}{0.000000,0.000000,0.000000}%
\pgfsetstrokecolor{currentstroke}%
\pgfsetdash{}{0pt}%
\pgfsys@defobject{currentmarker}{\pgfqpoint{0.000000in}{-0.048611in}}{\pgfqpoint{0.000000in}{0.000000in}}{%
\pgfpathmoveto{\pgfqpoint{0.000000in}{0.000000in}}%
\pgfpathlineto{\pgfqpoint{0.000000in}{-0.048611in}}%
\pgfusepath{stroke,fill}%
}%
\begin{pgfscope}%
\pgfsys@transformshift{1.760921in}{0.862305in}%
\pgfsys@useobject{currentmarker}{}%
\end{pgfscope}%
\end{pgfscope}%
\begin{pgfscope}%
\definecolor{textcolor}{rgb}{0.000000,0.000000,0.000000}%
\pgfsetstrokecolor{textcolor}%
\pgfsetfillcolor{textcolor}%
\pgftext[x=1.760921in,y=0.765082in,,top]{\color{textcolor}{\rmfamily\fontsize{25.000000}{30.000000}\selectfont\catcode`\^=\active\def^{\ifmmode\sp\else\^{}\fi}\catcode`\%=\active\def
\end{pgfscope}%
\begin{pgfscope}%
\pgfpathrectangle{\pgfqpoint{0.674954in}{0.862305in}}{\pgfqpoint{6.515802in}{4.369081in}}%
\pgfusepath{clip}%
\pgfsetbuttcap%
\pgfsetroundjoin%
\pgfsetlinewidth{2.007500pt}%
\definecolor{currentstroke}{rgb}{0.501961,0.501961,0.501961}%
\pgfsetstrokecolor{currentstroke}%
\pgfsetstrokeopacity{0.300000}%
\pgfsetdash{{7.400000pt}{3.200000pt}}{0.000000pt}%
\pgfpathmoveto{\pgfqpoint{2.846888in}{0.862305in}}%
\pgfpathlineto{\pgfqpoint{2.846888in}{5.231386in}}%
\pgfusepath{stroke}%
\end{pgfscope}%
\begin{pgfscope}%
\pgfsetbuttcap%
\pgfsetroundjoin%
\definecolor{currentfill}{rgb}{0.000000,0.000000,0.000000}%
\pgfsetfillcolor{currentfill}%
\pgfsetlinewidth{0.803000pt}%
\definecolor{currentstroke}{rgb}{0.000000,0.000000,0.000000}%
\pgfsetstrokecolor{currentstroke}%
\pgfsetdash{}{0pt}%
\pgfsys@defobject{currentmarker}{\pgfqpoint{0.000000in}{-0.048611in}}{\pgfqpoint{0.000000in}{0.000000in}}{%
\pgfpathmoveto{\pgfqpoint{0.000000in}{0.000000in}}%
\pgfpathlineto{\pgfqpoint{0.000000in}{-0.048611in}}%
\pgfusepath{stroke,fill}%
}%
\begin{pgfscope}%
\pgfsys@transformshift{2.846888in}{0.862305in}%
\pgfsys@useobject{currentmarker}{}%
\end{pgfscope}%
\end{pgfscope}%
\begin{pgfscope}%
\definecolor{textcolor}{rgb}{0.000000,0.000000,0.000000}%
\pgfsetstrokecolor{textcolor}%
\pgfsetfillcolor{textcolor}%
\pgftext[x=2.846888in,y=0.765082in,,top]{\color{textcolor}{\rmfamily\fontsize{25.000000}{30.000000}\selectfont\catcode`\^=\active\def^{\ifmmode\sp\else\^{}\fi}\catcode`\%=\active\def
\end{pgfscope}%
\begin{pgfscope}%
\pgfpathrectangle{\pgfqpoint{0.674954in}{0.862305in}}{\pgfqpoint{6.515802in}{4.369081in}}%
\pgfusepath{clip}%
\pgfsetbuttcap%
\pgfsetroundjoin%
\pgfsetlinewidth{2.007500pt}%
\definecolor{currentstroke}{rgb}{0.501961,0.501961,0.501961}%
\pgfsetstrokecolor{currentstroke}%
\pgfsetstrokeopacity{0.300000}%
\pgfsetdash{{7.400000pt}{3.200000pt}}{0.000000pt}%
\pgfpathmoveto{\pgfqpoint{3.932855in}{0.862305in}}%
\pgfpathlineto{\pgfqpoint{3.932855in}{5.231386in}}%
\pgfusepath{stroke}%
\end{pgfscope}%
\begin{pgfscope}%
\pgfsetbuttcap%
\pgfsetroundjoin%
\definecolor{currentfill}{rgb}{0.000000,0.000000,0.000000}%
\pgfsetfillcolor{currentfill}%
\pgfsetlinewidth{0.803000pt}%
\definecolor{currentstroke}{rgb}{0.000000,0.000000,0.000000}%
\pgfsetstrokecolor{currentstroke}%
\pgfsetdash{}{0pt}%
\pgfsys@defobject{currentmarker}{\pgfqpoint{0.000000in}{-0.048611in}}{\pgfqpoint{0.000000in}{0.000000in}}{%
\pgfpathmoveto{\pgfqpoint{0.000000in}{0.000000in}}%
\pgfpathlineto{\pgfqpoint{0.000000in}{-0.048611in}}%
\pgfusepath{stroke,fill}%
}%
\begin{pgfscope}%
\pgfsys@transformshift{3.932855in}{0.862305in}%
\pgfsys@useobject{currentmarker}{}%
\end{pgfscope}%
\end{pgfscope}%
\begin{pgfscope}%
\definecolor{textcolor}{rgb}{0.000000,0.000000,0.000000}%
\pgfsetstrokecolor{textcolor}%
\pgfsetfillcolor{textcolor}%
\pgftext[x=3.932855in,y=0.765082in,,top]{\color{textcolor}{\rmfamily\fontsize{25.000000}{30.000000}\selectfont\catcode`\^=\active\def^{\ifmmode\sp\else\^{}\fi}\catcode`\%=\active\def
\end{pgfscope}%
\begin{pgfscope}%
\pgfpathrectangle{\pgfqpoint{0.674954in}{0.862305in}}{\pgfqpoint{6.515802in}{4.369081in}}%
\pgfusepath{clip}%
\pgfsetbuttcap%
\pgfsetroundjoin%
\pgfsetlinewidth{2.007500pt}%
\definecolor{currentstroke}{rgb}{0.501961,0.501961,0.501961}%
\pgfsetstrokecolor{currentstroke}%
\pgfsetstrokeopacity{0.300000}%
\pgfsetdash{{7.400000pt}{3.200000pt}}{0.000000pt}%
\pgfpathmoveto{\pgfqpoint{5.018822in}{0.862305in}}%
\pgfpathlineto{\pgfqpoint{5.018822in}{5.231386in}}%
\pgfusepath{stroke}%
\end{pgfscope}%
\begin{pgfscope}%
\pgfsetbuttcap%
\pgfsetroundjoin%
\definecolor{currentfill}{rgb}{0.000000,0.000000,0.000000}%
\pgfsetfillcolor{currentfill}%
\pgfsetlinewidth{0.803000pt}%
\definecolor{currentstroke}{rgb}{0.000000,0.000000,0.000000}%
\pgfsetstrokecolor{currentstroke}%
\pgfsetdash{}{0pt}%
\pgfsys@defobject{currentmarker}{\pgfqpoint{0.000000in}{-0.048611in}}{\pgfqpoint{0.000000in}{0.000000in}}{%
\pgfpathmoveto{\pgfqpoint{0.000000in}{0.000000in}}%
\pgfpathlineto{\pgfqpoint{0.000000in}{-0.048611in}}%
\pgfusepath{stroke,fill}%
}%
\begin{pgfscope}%
\pgfsys@transformshift{5.018822in}{0.862305in}%
\pgfsys@useobject{currentmarker}{}%
\end{pgfscope}%
\end{pgfscope}%
\begin{pgfscope}%
\definecolor{textcolor}{rgb}{0.000000,0.000000,0.000000}%
\pgfsetstrokecolor{textcolor}%
\pgfsetfillcolor{textcolor}%
\pgftext[x=5.018822in,y=0.765082in,,top]{\color{textcolor}{\rmfamily\fontsize{25.000000}{30.000000}\selectfont\catcode`\^=\active\def^{\ifmmode\sp\else\^{}\fi}\catcode`\%=\active\def
\end{pgfscope}%
\begin{pgfscope}%
\pgfpathrectangle{\pgfqpoint{0.674954in}{0.862305in}}{\pgfqpoint{6.515802in}{4.369081in}}%
\pgfusepath{clip}%
\pgfsetbuttcap%
\pgfsetroundjoin%
\pgfsetlinewidth{2.007500pt}%
\definecolor{currentstroke}{rgb}{0.501961,0.501961,0.501961}%
\pgfsetstrokecolor{currentstroke}%
\pgfsetstrokeopacity{0.300000}%
\pgfsetdash{{7.400000pt}{3.200000pt}}{0.000000pt}%
\pgfpathmoveto{\pgfqpoint{6.104789in}{0.862305in}}%
\pgfpathlineto{\pgfqpoint{6.104789in}{5.231386in}}%
\pgfusepath{stroke}%
\end{pgfscope}%
\begin{pgfscope}%
\pgfsetbuttcap%
\pgfsetroundjoin%
\definecolor{currentfill}{rgb}{0.000000,0.000000,0.000000}%
\pgfsetfillcolor{currentfill}%
\pgfsetlinewidth{0.803000pt}%
\definecolor{currentstroke}{rgb}{0.000000,0.000000,0.000000}%
\pgfsetstrokecolor{currentstroke}%
\pgfsetdash{}{0pt}%
\pgfsys@defobject{currentmarker}{\pgfqpoint{0.000000in}{-0.048611in}}{\pgfqpoint{0.000000in}{0.000000in}}{%
\pgfpathmoveto{\pgfqpoint{0.000000in}{0.000000in}}%
\pgfpathlineto{\pgfqpoint{0.000000in}{-0.048611in}}%
\pgfusepath{stroke,fill}%
}%
\begin{pgfscope}%
\pgfsys@transformshift{6.104789in}{0.862305in}%
\pgfsys@useobject{currentmarker}{}%
\end{pgfscope}%
\end{pgfscope}%
\begin{pgfscope}%
\definecolor{textcolor}{rgb}{0.000000,0.000000,0.000000}%
\pgfsetstrokecolor{textcolor}%
\pgfsetfillcolor{textcolor}%
\pgftext[x=6.104789in,y=0.765082in,,top]{\color{textcolor}{\rmfamily\fontsize{25.000000}{30.000000}\selectfont\catcode`\^=\active\def^{\ifmmode\sp\else\^{}\fi}\catcode`\%=\active\def
\end{pgfscope}%
\begin{pgfscope}%
\pgfpathrectangle{\pgfqpoint{0.674954in}{0.862305in}}{\pgfqpoint{6.515802in}{4.369081in}}%
\pgfusepath{clip}%
\pgfsetbuttcap%
\pgfsetroundjoin%
\pgfsetlinewidth{2.007500pt}%
\definecolor{currentstroke}{rgb}{0.501961,0.501961,0.501961}%
\pgfsetstrokecolor{currentstroke}%
\pgfsetstrokeopacity{0.300000}%
\pgfsetdash{{7.400000pt}{3.200000pt}}{0.000000pt}%
\pgfpathmoveto{\pgfqpoint{6.740039in}{0.862305in}}%
\pgfpathlineto{\pgfqpoint{6.740039in}{5.231386in}}%
\pgfusepath{stroke}%
\end{pgfscope}%
\begin{pgfscope}%
\pgfsetbuttcap%
\pgfsetroundjoin%
\definecolor{currentfill}{rgb}{0.000000,0.000000,0.000000}%
\pgfsetfillcolor{currentfill}%
\pgfsetlinewidth{0.803000pt}%
\definecolor{currentstroke}{rgb}{0.000000,0.000000,0.000000}%
\pgfsetstrokecolor{currentstroke}%
\pgfsetdash{}{0pt}%
\pgfsys@defobject{currentmarker}{\pgfqpoint{0.000000in}{-0.048611in}}{\pgfqpoint{0.000000in}{0.000000in}}{%
\pgfpathmoveto{\pgfqpoint{0.000000in}{0.000000in}}%
\pgfpathlineto{\pgfqpoint{0.000000in}{-0.048611in}}%
\pgfusepath{stroke,fill}%
}%
\begin{pgfscope}%
\pgfsys@transformshift{6.740039in}{0.862305in}%
\pgfsys@useobject{currentmarker}{}%
\end{pgfscope}%
\end{pgfscope}%
\begin{pgfscope}%
\definecolor{textcolor}{rgb}{0.000000,0.000000,0.000000}%
\pgfsetstrokecolor{textcolor}%
\pgfsetfillcolor{textcolor}%
\pgftext[x=6.740039in,y=0.765082in,,top]{\color{textcolor}{\rmfamily\fontsize{25.000000}{30.000000}\selectfont\catcode`\^=\active\def^{\ifmmode\sp\else\^{}\fi}\catcode`\%=\active\def
\end{pgfscope}%
\begin{pgfscope}%
\pgfpathrectangle{\pgfqpoint{0.674954in}{0.862305in}}{\pgfqpoint{6.515802in}{4.369081in}}%
\pgfusepath{clip}%
\pgfsetbuttcap%
\pgfsetroundjoin%
\pgfsetlinewidth{2.007500pt}%
\definecolor{currentstroke}{rgb}{0.501961,0.501961,0.501961}%
\pgfsetstrokecolor{currentstroke}%
\pgfsetstrokeopacity{0.300000}%
\pgfsetdash{{7.400000pt}{3.200000pt}}{0.000000pt}%
\pgfpathmoveto{\pgfqpoint{7.190756in}{0.862305in}}%
\pgfpathlineto{\pgfqpoint{7.190756in}{5.231386in}}%
\pgfusepath{stroke}%
\end{pgfscope}%
\begin{pgfscope}%
\pgfsetbuttcap%
\pgfsetroundjoin%
\definecolor{currentfill}{rgb}{0.000000,0.000000,0.000000}%
\pgfsetfillcolor{currentfill}%
\pgfsetlinewidth{0.803000pt}%
\definecolor{currentstroke}{rgb}{0.000000,0.000000,0.000000}%
\pgfsetstrokecolor{currentstroke}%
\pgfsetdash{}{0pt}%
\pgfsys@defobject{currentmarker}{\pgfqpoint{0.000000in}{-0.048611in}}{\pgfqpoint{0.000000in}{0.000000in}}{%
\pgfpathmoveto{\pgfqpoint{0.000000in}{0.000000in}}%
\pgfpathlineto{\pgfqpoint{0.000000in}{-0.048611in}}%
\pgfusepath{stroke,fill}%
}%
\begin{pgfscope}%
\pgfsys@transformshift{7.190756in}{0.862305in}%
\pgfsys@useobject{currentmarker}{}%
\end{pgfscope}%
\end{pgfscope}%
\begin{pgfscope}%
\definecolor{textcolor}{rgb}{0.000000,0.000000,0.000000}%
\pgfsetstrokecolor{textcolor}%
\pgfsetfillcolor{textcolor}%
\pgftext[x=7.190756in,y=0.765082in,,top]{\color{textcolor}{\rmfamily\fontsize{25.000000}{30.000000}\selectfont\catcode`\^=\active\def^{\ifmmode\sp\else\^{}\fi}\catcode`\%=\active\def
\end{pgfscope}%
\begin{pgfscope}%
\definecolor{textcolor}{rgb}{0.000000,0.000000,0.000000}%
\pgfsetstrokecolor{textcolor}%
\pgfsetfillcolor{textcolor}%
\pgftext[x=3.932855in,y=0.404763in,,top]{\color{textcolor}{\rmfamily\fontsize{25.000000}{30.000000}\selectfont\catcode`\^=\active\def^{\ifmmode\sp\else\^{}\fi}\catcode`\%=\active\def
\end{pgfscope}%
\begin{pgfscope}%
\pgfpathrectangle{\pgfqpoint{0.674954in}{0.862305in}}{\pgfqpoint{6.515802in}{4.369081in}}%
\pgfusepath{clip}%
\pgfsetbuttcap%
\pgfsetroundjoin%
\pgfsetlinewidth{2.007500pt}%
\definecolor{currentstroke}{rgb}{0.501961,0.501961,0.501961}%
\pgfsetstrokecolor{currentstroke}%
\pgfsetstrokeopacity{0.300000}%
\pgfsetdash{{7.400000pt}{3.200000pt}}{0.000000pt}%
\pgfpathmoveto{\pgfqpoint{0.674954in}{1.954575in}}%
\pgfpathlineto{\pgfqpoint{7.190756in}{1.954575in}}%
\pgfusepath{stroke}%
\end{pgfscope}%
\begin{pgfscope}%
\pgfsetbuttcap%
\pgfsetroundjoin%
\definecolor{currentfill}{rgb}{0.000000,0.000000,0.000000}%
\pgfsetfillcolor{currentfill}%
\pgfsetlinewidth{0.803000pt}%
\definecolor{currentstroke}{rgb}{0.000000,0.000000,0.000000}%
\pgfsetstrokecolor{currentstroke}%
\pgfsetdash{}{0pt}%
\pgfsys@defobject{currentmarker}{\pgfqpoint{-0.048611in}{0.000000in}}{\pgfqpoint{-0.000000in}{0.000000in}}{%
\pgfpathmoveto{\pgfqpoint{-0.000000in}{0.000000in}}%
\pgfpathlineto{\pgfqpoint{-0.048611in}{0.000000in}}%
\pgfusepath{stroke,fill}%
}%
\begin{pgfscope}%
\pgfsys@transformshift{0.674954in}{1.954575in}%
\pgfsys@useobject{currentmarker}{}%
\end{pgfscope}%
\end{pgfscope}%
\begin{pgfscope}%
\definecolor{textcolor}{rgb}{0.000000,0.000000,0.000000}%
\pgfsetstrokecolor{textcolor}%
\pgfsetfillcolor{textcolor}%
\pgftext[x=0.259244in, y=1.835961in, left, base]{\color{textcolor}{\rmfamily\fontsize{25.000000}{30.000000}\selectfont\catcode`\^=\active\def^{\ifmmode\sp\else\^{}\fi}\catcode`\%=\active\def
\end{pgfscope}%
\begin{pgfscope}%
\pgfpathrectangle{\pgfqpoint{0.674954in}{0.862305in}}{\pgfqpoint{6.515802in}{4.369081in}}%
\pgfusepath{clip}%
\pgfsetbuttcap%
\pgfsetroundjoin%
\pgfsetlinewidth{2.007500pt}%
\definecolor{currentstroke}{rgb}{0.501961,0.501961,0.501961}%
\pgfsetstrokecolor{currentstroke}%
\pgfsetstrokeopacity{0.300000}%
\pgfsetdash{{7.400000pt}{3.200000pt}}{0.000000pt}%
\pgfpathmoveto{\pgfqpoint{0.674954in}{3.046845in}}%
\pgfpathlineto{\pgfqpoint{7.190756in}{3.046845in}}%
\pgfusepath{stroke}%
\end{pgfscope}%
\begin{pgfscope}%
\pgfsetbuttcap%
\pgfsetroundjoin%
\definecolor{currentfill}{rgb}{0.000000,0.000000,0.000000}%
\pgfsetfillcolor{currentfill}%
\pgfsetlinewidth{0.803000pt}%
\definecolor{currentstroke}{rgb}{0.000000,0.000000,0.000000}%
\pgfsetstrokecolor{currentstroke}%
\pgfsetdash{}{0pt}%
\pgfsys@defobject{currentmarker}{\pgfqpoint{-0.048611in}{0.000000in}}{\pgfqpoint{-0.000000in}{0.000000in}}{%
\pgfpathmoveto{\pgfqpoint{-0.000000in}{0.000000in}}%
\pgfpathlineto{\pgfqpoint{-0.048611in}{0.000000in}}%
\pgfusepath{stroke,fill}%
}%
\begin{pgfscope}%
\pgfsys@transformshift{0.674954in}{3.046845in}%
\pgfsys@useobject{currentmarker}{}%
\end{pgfscope}%
\end{pgfscope}%
\begin{pgfscope}%
\definecolor{textcolor}{rgb}{0.000000,0.000000,0.000000}%
\pgfsetstrokecolor{textcolor}%
\pgfsetfillcolor{textcolor}%
\pgftext[x=0.259244in, y=2.928231in, left, base]{\color{textcolor}{\rmfamily\fontsize{25.000000}{30.000000}\selectfont\catcode`\^=\active\def^{\ifmmode\sp\else\^{}\fi}\catcode`\%=\active\def
\end{pgfscope}%
\begin{pgfscope}%
\pgfpathrectangle{\pgfqpoint{0.674954in}{0.862305in}}{\pgfqpoint{6.515802in}{4.369081in}}%
\pgfusepath{clip}%
\pgfsetbuttcap%
\pgfsetroundjoin%
\pgfsetlinewidth{2.007500pt}%
\definecolor{currentstroke}{rgb}{0.501961,0.501961,0.501961}%
\pgfsetstrokecolor{currentstroke}%
\pgfsetstrokeopacity{0.300000}%
\pgfsetdash{{7.400000pt}{3.200000pt}}{0.000000pt}%
\pgfpathmoveto{\pgfqpoint{0.674954in}{4.139116in}}%
\pgfpathlineto{\pgfqpoint{7.190756in}{4.139116in}}%
\pgfusepath{stroke}%
\end{pgfscope}%
\begin{pgfscope}%
\pgfsetbuttcap%
\pgfsetroundjoin%
\definecolor{currentfill}{rgb}{0.000000,0.000000,0.000000}%
\pgfsetfillcolor{currentfill}%
\pgfsetlinewidth{0.803000pt}%
\definecolor{currentstroke}{rgb}{0.000000,0.000000,0.000000}%
\pgfsetstrokecolor{currentstroke}%
\pgfsetdash{}{0pt}%
\pgfsys@defobject{currentmarker}{\pgfqpoint{-0.048611in}{0.000000in}}{\pgfqpoint{-0.000000in}{0.000000in}}{%
\pgfpathmoveto{\pgfqpoint{-0.000000in}{0.000000in}}%
\pgfpathlineto{\pgfqpoint{-0.048611in}{0.000000in}}%
\pgfusepath{stroke,fill}%
}%
\begin{pgfscope}%
\pgfsys@transformshift{0.674954in}{4.139116in}%
\pgfsys@useobject{currentmarker}{}%
\end{pgfscope}%
\end{pgfscope}%
\begin{pgfscope}%
\definecolor{textcolor}{rgb}{0.000000,0.000000,0.000000}%
\pgfsetstrokecolor{textcolor}%
\pgfsetfillcolor{textcolor}%
\pgftext[x=0.259244in, y=4.020501in, left, base]{\color{textcolor}{\rmfamily\fontsize{25.000000}{30.000000}\selectfont\catcode`\^=\active\def^{\ifmmode\sp\else\^{}\fi}\catcode`\%=\active\def
\end{pgfscope}%
\begin{pgfscope}%
\pgfpathrectangle{\pgfqpoint{0.674954in}{0.862305in}}{\pgfqpoint{6.515802in}{4.369081in}}%
\pgfusepath{clip}%
\pgfsetbuttcap%
\pgfsetroundjoin%
\pgfsetlinewidth{2.007500pt}%
\definecolor{currentstroke}{rgb}{0.501961,0.501961,0.501961}%
\pgfsetstrokecolor{currentstroke}%
\pgfsetstrokeopacity{0.300000}%
\pgfsetdash{{7.400000pt}{3.200000pt}}{0.000000pt}%
\pgfpathmoveto{\pgfqpoint{0.674954in}{5.231386in}}%
\pgfpathlineto{\pgfqpoint{7.190756in}{5.231386in}}%
\pgfusepath{stroke}%
\end{pgfscope}%
\begin{pgfscope}%
\pgfsetbuttcap%
\pgfsetroundjoin%
\definecolor{currentfill}{rgb}{0.000000,0.000000,0.000000}%
\pgfsetfillcolor{currentfill}%
\pgfsetlinewidth{0.803000pt}%
\definecolor{currentstroke}{rgb}{0.000000,0.000000,0.000000}%
\pgfsetstrokecolor{currentstroke}%
\pgfsetdash{}{0pt}%
\pgfsys@defobject{currentmarker}{\pgfqpoint{-0.048611in}{0.000000in}}{\pgfqpoint{-0.000000in}{0.000000in}}{%
\pgfpathmoveto{\pgfqpoint{-0.000000in}{0.000000in}}%
\pgfpathlineto{\pgfqpoint{-0.048611in}{0.000000in}}%
\pgfusepath{stroke,fill}%
}%
\begin{pgfscope}%
\pgfsys@transformshift{0.674954in}{5.231386in}%
\pgfsys@useobject{currentmarker}{}%
\end{pgfscope}%
\end{pgfscope}%
\begin{pgfscope}%
\definecolor{textcolor}{rgb}{0.000000,0.000000,0.000000}%
\pgfsetstrokecolor{textcolor}%
\pgfsetfillcolor{textcolor}%
\pgftext[x=0.100000in, y=5.112772in, left, base]{\color{textcolor}{\rmfamily\fontsize{25.000000}{30.000000}\selectfont\catcode`\^=\active\def^{\ifmmode\sp\else\^{}\fi}\catcode`\%=\active\def
\end{pgfscope}%
\begin{pgfscope}%
\pgfpathrectangle{\pgfqpoint{0.674954in}{0.862305in}}{\pgfqpoint{6.515802in}{4.369081in}}%
\pgfusepath{clip}%
\pgfsetrectcap%
\pgfsetroundjoin%
\pgfsetlinewidth{2.509375pt}%
\definecolor{currentstroke}{rgb}{0.050980,0.415686,0.509804}%
\pgfsetstrokecolor{currentstroke}%
\pgfsetdash{}{0pt}%
\pgfpathmoveto{\pgfqpoint{0.674954in}{3.632494in}}%
\pgfpathlineto{\pgfqpoint{1.760921in}{3.771216in}}%
\pgfpathlineto{\pgfqpoint{2.846888in}{3.681854in}}%
\pgfpathlineto{\pgfqpoint{3.932855in}{3.594192in}}%
\pgfpathlineto{\pgfqpoint{5.018822in}{3.966982in}}%
\pgfpathlineto{\pgfqpoint{6.104789in}{3.807501in}}%
\pgfpathlineto{\pgfqpoint{6.740039in}{3.939016in}}%
\pgfpathlineto{\pgfqpoint{7.190756in}{3.967874in}}%
\pgfusepath{stroke}%
\end{pgfscope}%
\begin{pgfscope}%
\pgfpathrectangle{\pgfqpoint{0.674954in}{0.862305in}}{\pgfqpoint{6.515802in}{4.369081in}}%
\pgfusepath{clip}%
\pgfsetbuttcap%
\pgfsetroundjoin%
\definecolor{currentfill}{rgb}{0.050980,0.415686,0.509804}%
\pgfsetfillcolor{currentfill}%
\pgfsetlinewidth{1.003750pt}%
\definecolor{currentstroke}{rgb}{0.050980,0.415686,0.509804}%
\pgfsetstrokecolor{currentstroke}%
\pgfsetdash{}{0pt}%
\pgfsys@defobject{currentmarker}{\pgfqpoint{-0.055556in}{-0.055556in}}{\pgfqpoint{0.055556in}{0.055556in}}{%
\pgfpathmoveto{\pgfqpoint{0.000000in}{-0.055556in}}%
\pgfpathcurveto{\pgfqpoint{0.014734in}{-0.055556in}}{\pgfqpoint{0.028866in}{-0.049702in}}{\pgfqpoint{0.039284in}{-0.039284in}}%
\pgfpathcurveto{\pgfqpoint{0.049702in}{-0.028866in}}{\pgfqpoint{0.055556in}{-0.014734in}}{\pgfqpoint{0.055556in}{0.000000in}}%
\pgfpathcurveto{\pgfqpoint{0.055556in}{0.014734in}}{\pgfqpoint{0.049702in}{0.028866in}}{\pgfqpoint{0.039284in}{0.039284in}}%
\pgfpathcurveto{\pgfqpoint{0.028866in}{0.049702in}}{\pgfqpoint{0.014734in}{0.055556in}}{\pgfqpoint{0.000000in}{0.055556in}}%
\pgfpathcurveto{\pgfqpoint{-0.014734in}{0.055556in}}{\pgfqpoint{-0.028866in}{0.049702in}}{\pgfqpoint{-0.039284in}{0.039284in}}%
\pgfpathcurveto{\pgfqpoint{-0.049702in}{0.028866in}}{\pgfqpoint{-0.055556in}{0.014734in}}{\pgfqpoint{-0.055556in}{0.000000in}}%
\pgfpathcurveto{\pgfqpoint{-0.055556in}{-0.014734in}}{\pgfqpoint{-0.049702in}{-0.028866in}}{\pgfqpoint{-0.039284in}{-0.039284in}}%
\pgfpathcurveto{\pgfqpoint{-0.028866in}{-0.049702in}}{\pgfqpoint{-0.014734in}{-0.055556in}}{\pgfqpoint{0.000000in}{-0.055556in}}%
\pgfpathlineto{\pgfqpoint{0.000000in}{-0.055556in}}%
\pgfpathclose%
\pgfusepath{stroke,fill}%
}%
\begin{pgfscope}%
\pgfsys@transformshift{0.674954in}{3.632494in}%
\pgfsys@useobject{currentmarker}{}%
\end{pgfscope}%
\begin{pgfscope}%
\pgfsys@transformshift{1.760921in}{3.771216in}%
\pgfsys@useobject{currentmarker}{}%
\end{pgfscope}%
\begin{pgfscope}%
\pgfsys@transformshift{2.846888in}{3.681854in}%
\pgfsys@useobject{currentmarker}{}%
\end{pgfscope}%
\begin{pgfscope}%
\pgfsys@transformshift{3.932855in}{3.594192in}%
\pgfsys@useobject{currentmarker}{}%
\end{pgfscope}%
\begin{pgfscope}%
\pgfsys@transformshift{5.018822in}{3.966982in}%
\pgfsys@useobject{currentmarker}{}%
\end{pgfscope}%
\begin{pgfscope}%
\pgfsys@transformshift{6.104789in}{3.807501in}%
\pgfsys@useobject{currentmarker}{}%
\end{pgfscope}%
\begin{pgfscope}%
\pgfsys@transformshift{6.740039in}{3.939016in}%
\pgfsys@useobject{currentmarker}{}%
\end{pgfscope}%
\begin{pgfscope}%
\pgfsys@transformshift{7.190756in}{3.967874in}%
\pgfsys@useobject{currentmarker}{}%
\end{pgfscope}%
\end{pgfscope}%
\begin{pgfscope}%
\pgfpathrectangle{\pgfqpoint{0.674954in}{0.862305in}}{\pgfqpoint{6.515802in}{4.369081in}}%
\pgfusepath{clip}%
\pgfsetbuttcap%
\pgfsetroundjoin%
\pgfsetlinewidth{2.509375pt}%
\definecolor{currentstroke}{rgb}{0.960784,0.462745,0.000000}%
\pgfsetstrokecolor{currentstroke}%
\pgfsetdash{{9.250000pt}{4.000000pt}}{0.000000pt}%
\pgfpathmoveto{\pgfqpoint{0.674954in}{3.056815in}}%
\pgfpathlineto{\pgfqpoint{1.760921in}{3.186036in}}%
\pgfpathlineto{\pgfqpoint{2.846888in}{3.186013in}}%
\pgfpathlineto{\pgfqpoint{3.932855in}{3.009623in}}%
\pgfpathlineto{\pgfqpoint{5.018822in}{3.430278in}}%
\pgfpathlineto{\pgfqpoint{6.104789in}{3.219210in}}%
\pgfpathlineto{\pgfqpoint{6.740039in}{3.377633in}}%
\pgfpathlineto{\pgfqpoint{7.190756in}{3.443220in}}%
\pgfusepath{stroke}%
\end{pgfscope}%
\begin{pgfscope}%
\pgfpathrectangle{\pgfqpoint{0.674954in}{0.862305in}}{\pgfqpoint{6.515802in}{4.369081in}}%
\pgfusepath{clip}%
\pgfsetbuttcap%
\pgfsetmiterjoin%
\definecolor{currentfill}{rgb}{0.960784,0.462745,0.000000}%
\pgfsetfillcolor{currentfill}%
\pgfsetlinewidth{1.003750pt}%
\definecolor{currentstroke}{rgb}{0.960784,0.462745,0.000000}%
\pgfsetstrokecolor{currentstroke}%
\pgfsetdash{}{0pt}%
\pgfsys@defobject{currentmarker}{\pgfqpoint{-0.055556in}{-0.055556in}}{\pgfqpoint{0.055556in}{0.055556in}}{%
\pgfpathmoveto{\pgfqpoint{-0.055556in}{-0.055556in}}%
\pgfpathlineto{\pgfqpoint{0.055556in}{-0.055556in}}%
\pgfpathlineto{\pgfqpoint{0.055556in}{0.055556in}}%
\pgfpathlineto{\pgfqpoint{-0.055556in}{0.055556in}}%
\pgfpathlineto{\pgfqpoint{-0.055556in}{-0.055556in}}%
\pgfpathclose%
\pgfusepath{stroke,fill}%
}%
\begin{pgfscope}%
\pgfsys@transformshift{0.674954in}{3.056815in}%
\pgfsys@useobject{currentmarker}{}%
\end{pgfscope}%
\begin{pgfscope}%
\pgfsys@transformshift{1.760921in}{3.186036in}%
\pgfsys@useobject{currentmarker}{}%
\end{pgfscope}%
\begin{pgfscope}%
\pgfsys@transformshift{2.846888in}{3.186013in}%
\pgfsys@useobject{currentmarker}{}%
\end{pgfscope}%
\begin{pgfscope}%
\pgfsys@transformshift{3.932855in}{3.009623in}%
\pgfsys@useobject{currentmarker}{}%
\end{pgfscope}%
\begin{pgfscope}%
\pgfsys@transformshift{5.018822in}{3.430278in}%
\pgfsys@useobject{currentmarker}{}%
\end{pgfscope}%
\begin{pgfscope}%
\pgfsys@transformshift{6.104789in}{3.219210in}%
\pgfsys@useobject{currentmarker}{}%
\end{pgfscope}%
\begin{pgfscope}%
\pgfsys@transformshift{6.740039in}{3.377633in}%
\pgfsys@useobject{currentmarker}{}%
\end{pgfscope}%
\begin{pgfscope}%
\pgfsys@transformshift{7.190756in}{3.443220in}%
\pgfsys@useobject{currentmarker}{}%
\end{pgfscope}%
\end{pgfscope}%
\begin{pgfscope}%
\pgfpathrectangle{\pgfqpoint{0.674954in}{0.862305in}}{\pgfqpoint{6.515802in}{4.369081in}}%
\pgfusepath{clip}%
\pgfsetbuttcap%
\pgfsetroundjoin%
\pgfsetlinewidth{2.509375pt}%
\definecolor{currentstroke}{rgb}{0.219608,0.219608,0.219608}%
\pgfsetstrokecolor{currentstroke}%
\pgfsetdash{{2.500000pt}{4.125000pt}}{0.000000pt}%
\pgfpathmoveto{\pgfqpoint{0.674954in}{2.018722in}}%
\pgfpathlineto{\pgfqpoint{1.760921in}{2.077388in}}%
\pgfpathlineto{\pgfqpoint{2.846888in}{2.226079in}}%
\pgfpathlineto{\pgfqpoint{3.932855in}{2.020137in}}%
\pgfpathlineto{\pgfqpoint{5.018822in}{2.350322in}}%
\pgfpathlineto{\pgfqpoint{6.104789in}{2.163699in}}%
\pgfpathlineto{\pgfqpoint{6.740039in}{2.265104in}}%
\pgfpathlineto{\pgfqpoint{7.190756in}{2.361778in}}%
\pgfusepath{stroke}%
\end{pgfscope}%
\begin{pgfscope}%
\pgfpathrectangle{\pgfqpoint{0.674954in}{0.862305in}}{\pgfqpoint{6.515802in}{4.369081in}}%
\pgfusepath{clip}%
\pgfsetbuttcap%
\pgfsetmiterjoin%
\definecolor{currentfill}{rgb}{0.219608,0.219608,0.219608}%
\pgfsetfillcolor{currentfill}%
\pgfsetlinewidth{1.003750pt}%
\definecolor{currentstroke}{rgb}{0.219608,0.219608,0.219608}%
\pgfsetstrokecolor{currentstroke}%
\pgfsetdash{}{0pt}%
\pgfsys@defobject{currentmarker}{\pgfqpoint{-0.055556in}{-0.055556in}}{\pgfqpoint{0.055556in}{0.055556in}}{%
\pgfpathmoveto{\pgfqpoint{0.000000in}{0.055556in}}%
\pgfpathlineto{\pgfqpoint{-0.055556in}{-0.055556in}}%
\pgfpathlineto{\pgfqpoint{0.055556in}{-0.055556in}}%
\pgfpathlineto{\pgfqpoint{0.000000in}{0.055556in}}%
\pgfpathclose%
\pgfusepath{stroke,fill}%
}%
\begin{pgfscope}%
\pgfsys@transformshift{0.674954in}{2.018722in}%
\pgfsys@useobject{currentmarker}{}%
\end{pgfscope}%
\begin{pgfscope}%
\pgfsys@transformshift{1.760921in}{2.077388in}%
\pgfsys@useobject{currentmarker}{}%
\end{pgfscope}%
\begin{pgfscope}%
\pgfsys@transformshift{2.846888in}{2.226079in}%
\pgfsys@useobject{currentmarker}{}%
\end{pgfscope}%
\begin{pgfscope}%
\pgfsys@transformshift{3.932855in}{2.020137in}%
\pgfsys@useobject{currentmarker}{}%
\end{pgfscope}%
\begin{pgfscope}%
\pgfsys@transformshift{5.018822in}{2.350322in}%
\pgfsys@useobject{currentmarker}{}%
\end{pgfscope}%
\begin{pgfscope}%
\pgfsys@transformshift{6.104789in}{2.163699in}%
\pgfsys@useobject{currentmarker}{}%
\end{pgfscope}%
\begin{pgfscope}%
\pgfsys@transformshift{6.740039in}{2.265104in}%
\pgfsys@useobject{currentmarker}{}%
\end{pgfscope}%
\begin{pgfscope}%
\pgfsys@transformshift{7.190756in}{2.361778in}%
\pgfsys@useobject{currentmarker}{}%
\end{pgfscope}%
\end{pgfscope}%
\begin{pgfscope}%
\pgfsetrectcap%
\pgfsetmiterjoin%
\pgfsetlinewidth{2.007500pt}%
\definecolor{currentstroke}{rgb}{0.000000,0.000000,0.000000}%
\pgfsetstrokecolor{currentstroke}%
\pgfsetdash{}{0pt}%
\pgfpathmoveto{\pgfqpoint{0.674954in}{0.862305in}}%
\pgfpathlineto{\pgfqpoint{0.674954in}{5.231386in}}%
\pgfusepath{stroke}%
\end{pgfscope}%
\begin{pgfscope}%
\pgfsetrectcap%
\pgfsetmiterjoin%
\pgfsetlinewidth{2.007500pt}%
\definecolor{currentstroke}{rgb}{0.000000,0.000000,0.000000}%
\pgfsetstrokecolor{currentstroke}%
\pgfsetdash{}{0pt}%
\pgfpathmoveto{\pgfqpoint{0.674954in}{0.862305in}}%
\pgfpathlineto{\pgfqpoint{7.190756in}{0.862305in}}%
\pgfusepath{stroke}%
\end{pgfscope}%
\begin{pgfscope}%
\pgfsetbuttcap%
\pgfsetmiterjoin%
\definecolor{currentfill}{rgb}{1.000000,1.000000,1.000000}%
\pgfsetfillcolor{currentfill}%
\pgfsetlinewidth{1.003750pt}%
\definecolor{currentstroke}{rgb}{0.800000,0.800000,0.800000}%
\pgfsetstrokecolor{currentstroke}%
\pgfsetdash{}{0pt}%
\pgfpathmoveto{\pgfqpoint{0.869398in}{3.847079in}}%
\pgfpathlineto{\pgfqpoint{2.213036in}{3.847079in}}%
\pgfpathquadraticcurveto{\pgfqpoint{2.268591in}{3.847079in}}{\pgfqpoint{2.268591in}{3.902635in}}%
\pgfpathlineto{\pgfqpoint{2.268591in}{5.036941in}}%
\pgfpathquadraticcurveto{\pgfqpoint{2.268591in}{5.092497in}}{\pgfqpoint{2.213036in}{5.092497in}}%
\pgfpathlineto{\pgfqpoint{0.869398in}{5.092497in}}%
\pgfpathquadraticcurveto{\pgfqpoint{0.813843in}{5.092497in}}{\pgfqpoint{0.813843in}{5.036941in}}%
\pgfpathlineto{\pgfqpoint{0.813843in}{3.902635in}}%
\pgfpathquadraticcurveto{\pgfqpoint{0.813843in}{3.847079in}}{\pgfqpoint{0.869398in}{3.847079in}}%
\pgfpathlineto{\pgfqpoint{0.869398in}{3.847079in}}%
\pgfpathclose%
\pgfusepath{stroke,fill}%
\end{pgfscope}%
\begin{pgfscope}%
\pgfsetrectcap%
\pgfsetroundjoin%
\pgfsetlinewidth{2.509375pt}%
\definecolor{currentstroke}{rgb}{0.050980,0.415686,0.509804}%
\pgfsetstrokecolor{currentstroke}%
\pgfsetdash{}{0pt}%
\pgfpathmoveto{\pgfqpoint{0.924954in}{4.884164in}}%
\pgfpathlineto{\pgfqpoint{1.202732in}{4.884164in}}%
\pgfpathlineto{\pgfqpoint{1.480509in}{4.884164in}}%
\pgfusepath{stroke}%
\end{pgfscope}%
\begin{pgfscope}%
\pgfsetbuttcap%
\pgfsetroundjoin%
\definecolor{currentfill}{rgb}{0.050980,0.415686,0.509804}%
\pgfsetfillcolor{currentfill}%
\pgfsetlinewidth{1.003750pt}%
\definecolor{currentstroke}{rgb}{0.050980,0.415686,0.509804}%
\pgfsetstrokecolor{currentstroke}%
\pgfsetdash{}{0pt}%
\pgfsys@defobject{currentmarker}{\pgfqpoint{-0.055556in}{-0.055556in}}{\pgfqpoint{0.055556in}{0.055556in}}{%
\pgfpathmoveto{\pgfqpoint{0.000000in}{-0.055556in}}%
\pgfpathcurveto{\pgfqpoint{0.014734in}{-0.055556in}}{\pgfqpoint{0.028866in}{-0.049702in}}{\pgfqpoint{0.039284in}{-0.039284in}}%
\pgfpathcurveto{\pgfqpoint{0.049702in}{-0.028866in}}{\pgfqpoint{0.055556in}{-0.014734in}}{\pgfqpoint{0.055556in}{0.000000in}}%
\pgfpathcurveto{\pgfqpoint{0.055556in}{0.014734in}}{\pgfqpoint{0.049702in}{0.028866in}}{\pgfqpoint{0.039284in}{0.039284in}}%
\pgfpathcurveto{\pgfqpoint{0.028866in}{0.049702in}}{\pgfqpoint{0.014734in}{0.055556in}}{\pgfqpoint{0.000000in}{0.055556in}}%
\pgfpathcurveto{\pgfqpoint{-0.014734in}{0.055556in}}{\pgfqpoint{-0.028866in}{0.049702in}}{\pgfqpoint{-0.039284in}{0.039284in}}%
\pgfpathcurveto{\pgfqpoint{-0.049702in}{0.028866in}}{\pgfqpoint{-0.055556in}{0.014734in}}{\pgfqpoint{-0.055556in}{0.000000in}}%
\pgfpathcurveto{\pgfqpoint{-0.055556in}{-0.014734in}}{\pgfqpoint{-0.049702in}{-0.028866in}}{\pgfqpoint{-0.039284in}{-0.039284in}}%
\pgfpathcurveto{\pgfqpoint{-0.028866in}{-0.049702in}}{\pgfqpoint{-0.014734in}{-0.055556in}}{\pgfqpoint{0.000000in}{-0.055556in}}%
\pgfpathlineto{\pgfqpoint{0.000000in}{-0.055556in}}%
\pgfpathclose%
\pgfusepath{stroke,fill}%
}%
\begin{pgfscope}%
\pgfsys@transformshift{1.202732in}{4.884164in}%
\pgfsys@useobject{currentmarker}{}%
\end{pgfscope}%
\end{pgfscope}%
\begin{pgfscope}%
\definecolor{textcolor}{rgb}{0.000000,0.000000,0.000000}%
\pgfsetstrokecolor{textcolor}%
\pgfsetfillcolor{textcolor}%
\pgftext[x=1.702732in,y=4.786941in,left,base]{\color{textcolor}{\rmfamily\fontsize{20.000000}{24.000000}\selectfont\catcode`\^=\active\def^{\ifmmode\sp\else\^{}\fi}\catcode`\%=\active\def
\end{pgfscope}%
\begin{pgfscope}%
\pgfsetbuttcap%
\pgfsetroundjoin%
\pgfsetlinewidth{2.509375pt}%
\definecolor{currentstroke}{rgb}{0.960784,0.462745,0.000000}%
\pgfsetstrokecolor{currentstroke}%
\pgfsetdash{{9.250000pt}{4.000000pt}}{0.000000pt}%
\pgfpathmoveto{\pgfqpoint{0.924954in}{4.496802in}}%
\pgfpathlineto{\pgfqpoint{1.202732in}{4.496802in}}%
\pgfpathlineto{\pgfqpoint{1.480509in}{4.496802in}}%
\pgfusepath{stroke}%
\end{pgfscope}%
\begin{pgfscope}%
\pgfsetbuttcap%
\pgfsetmiterjoin%
\definecolor{currentfill}{rgb}{0.960784,0.462745,0.000000}%
\pgfsetfillcolor{currentfill}%
\pgfsetlinewidth{1.003750pt}%
\definecolor{currentstroke}{rgb}{0.960784,0.462745,0.000000}%
\pgfsetstrokecolor{currentstroke}%
\pgfsetdash{}{0pt}%
\pgfsys@defobject{currentmarker}{\pgfqpoint{-0.055556in}{-0.055556in}}{\pgfqpoint{0.055556in}{0.055556in}}{%
\pgfpathmoveto{\pgfqpoint{-0.055556in}{-0.055556in}}%
\pgfpathlineto{\pgfqpoint{0.055556in}{-0.055556in}}%
\pgfpathlineto{\pgfqpoint{0.055556in}{0.055556in}}%
\pgfpathlineto{\pgfqpoint{-0.055556in}{0.055556in}}%
\pgfpathlineto{\pgfqpoint{-0.055556in}{-0.055556in}}%
\pgfpathclose%
\pgfusepath{stroke,fill}%
}%
\begin{pgfscope}%
\pgfsys@transformshift{1.202732in}{4.496802in}%
\pgfsys@useobject{currentmarker}{}%
\end{pgfscope}%
\end{pgfscope}%
\begin{pgfscope}%
\definecolor{textcolor}{rgb}{0.000000,0.000000,0.000000}%
\pgfsetstrokecolor{textcolor}%
\pgfsetfillcolor{textcolor}%
\pgftext[x=1.702732in,y=4.399580in,left,base]{\color{textcolor}{\rmfamily\fontsize{20.000000}{24.000000}\selectfont\catcode`\^=\active\def^{\ifmmode\sp\else\^{}\fi}\catcode`\%=\active\def
\end{pgfscope}%
\begin{pgfscope}%
\pgfsetbuttcap%
\pgfsetroundjoin%
\pgfsetlinewidth{2.509375pt}%
\definecolor{currentstroke}{rgb}{0.219608,0.219608,0.219608}%
\pgfsetstrokecolor{currentstroke}%
\pgfsetdash{{2.500000pt}{4.125000pt}}{0.000000pt}%
\pgfpathmoveto{\pgfqpoint{0.924954in}{4.109441in}}%
\pgfpathlineto{\pgfqpoint{1.202732in}{4.109441in}}%
\pgfpathlineto{\pgfqpoint{1.480509in}{4.109441in}}%
\pgfusepath{stroke}%
\end{pgfscope}%
\begin{pgfscope}%
\pgfsetbuttcap%
\pgfsetmiterjoin%
\definecolor{currentfill}{rgb}{0.219608,0.219608,0.219608}%
\pgfsetfillcolor{currentfill}%
\pgfsetlinewidth{1.003750pt}%
\definecolor{currentstroke}{rgb}{0.219608,0.219608,0.219608}%
\pgfsetstrokecolor{currentstroke}%
\pgfsetdash{}{0pt}%
\pgfsys@defobject{currentmarker}{\pgfqpoint{-0.055556in}{-0.055556in}}{\pgfqpoint{0.055556in}{0.055556in}}{%
\pgfpathmoveto{\pgfqpoint{0.000000in}{0.055556in}}%
\pgfpathlineto{\pgfqpoint{-0.055556in}{-0.055556in}}%
\pgfpathlineto{\pgfqpoint{0.055556in}{-0.055556in}}%
\pgfpathlineto{\pgfqpoint{0.000000in}{0.055556in}}%
\pgfpathclose%
\pgfusepath{stroke,fill}%
}%
\begin{pgfscope}%
\pgfsys@transformshift{1.202732in}{4.109441in}%
\pgfsys@useobject{currentmarker}{}%
\end{pgfscope}%
\end{pgfscope}%
\begin{pgfscope}%
\definecolor{textcolor}{rgb}{0.000000,0.000000,0.000000}%
\pgfsetstrokecolor{textcolor}%
\pgfsetfillcolor{textcolor}%
\pgftext[x=1.702732in,y=4.012218in,left,base]{\color{textcolor}{\rmfamily\fontsize{20.000000}{24.000000}\selectfont\catcode`\^=\active\def^{\ifmmode\sp\else\^{}\fi}\catcode`\%=\active\def
\end{pgfscope}%
\end{pgfpicture}%
\makeatother%
\endgroup%

%% file: images/lora_r_study/r_results_RRA.pgf
\begingroup%
\makeatletter%
\begin{pgfpicture}%
\pgfpathrectangle{\pgfpointorigin}{\pgfqpoint{7.450000in}{5.450000in}}%
\pgfusepath{use as bounding box, clip}%
\begin{pgfscope}%
\pgfsetbuttcap%
\pgfsetmiterjoin%
\definecolor{currentfill}{rgb}{1.000000,1.000000,1.000000}%
\pgfsetfillcolor{currentfill}%
\pgfsetlinewidth{0.000000pt}%
\definecolor{currentstroke}{rgb}{1.000000,1.000000,1.000000}%
\pgfsetstrokecolor{currentstroke}%
\pgfsetdash{}{0pt}%
\pgfpathmoveto{\pgfqpoint{0.000000in}{0.000000in}}%
\pgfpathlineto{\pgfqpoint{7.450000in}{0.000000in}}%
\pgfpathlineto{\pgfqpoint{7.450000in}{5.450000in}}%
\pgfpathlineto{\pgfqpoint{0.000000in}{5.450000in}}%
\pgfpathlineto{\pgfqpoint{0.000000in}{0.000000in}}%
\pgfpathclose%
\pgfusepath{fill}%
\end{pgfscope}%
\begin{pgfscope}%
\pgfsetbuttcap%
\pgfsetmiterjoin%
\definecolor{currentfill}{rgb}{1.000000,1.000000,1.000000}%
\pgfsetfillcolor{currentfill}%
\pgfsetlinewidth{0.000000pt}%
\definecolor{currentstroke}{rgb}{0.000000,0.000000,0.000000}%
\pgfsetstrokecolor{currentstroke}%
\pgfsetstrokeopacity{0.000000}%
\pgfsetdash{}{0pt}%
\pgfpathmoveto{\pgfqpoint{0.674954in}{0.862305in}}%
\pgfpathlineto{\pgfqpoint{7.190756in}{0.862305in}}%
\pgfpathlineto{\pgfqpoint{7.190756in}{5.231386in}}%
\pgfpathlineto{\pgfqpoint{0.674954in}{5.231386in}}%
\pgfpathlineto{\pgfqpoint{0.674954in}{0.862305in}}%
\pgfpathclose%
\pgfusepath{fill}%
\end{pgfscope}%
\begin{pgfscope}%
\pgfpathrectangle{\pgfqpoint{0.674954in}{0.862305in}}{\pgfqpoint{6.515802in}{4.369081in}}%
\pgfusepath{clip}%
\pgfsetbuttcap%
\pgfsetroundjoin%
\definecolor{currentfill}{rgb}{0.050980,0.415686,0.509804}%
\pgfsetfillcolor{currentfill}%
\pgfsetfillopacity{0.300000}%
\pgfsetlinewidth{1.003750pt}%
\definecolor{currentstroke}{rgb}{0.050980,0.415686,0.509804}%
\pgfsetstrokecolor{currentstroke}%
\pgfsetstrokeopacity{0.300000}%
\pgfsetdash{}{0pt}%
\pgfsys@defobject{currentmarker}{\pgfqpoint{0.674954in}{4.450365in}}{\pgfqpoint{7.190756in}{5.008003in}}{%
\pgfpathmoveto{\pgfqpoint{0.674954in}{4.803816in}}%
\pgfpathlineto{\pgfqpoint{0.674954in}{4.484012in}}%
\pgfpathlineto{\pgfqpoint{1.760921in}{4.598077in}}%
\pgfpathlineto{\pgfqpoint{2.846888in}{4.450365in}}%
\pgfpathlineto{\pgfqpoint{3.932855in}{4.576469in}}%
\pgfpathlineto{\pgfqpoint{5.018822in}{4.720291in}}%
\pgfpathlineto{\pgfqpoint{6.104789in}{4.658378in}}%
\pgfpathlineto{\pgfqpoint{6.740039in}{4.626641in}}%
\pgfpathlineto{\pgfqpoint{7.190756in}{4.623616in}}%
\pgfpathlineto{\pgfqpoint{7.190756in}{5.008003in}}%
\pgfpathlineto{\pgfqpoint{7.190756in}{5.008003in}}%
\pgfpathlineto{\pgfqpoint{6.740039in}{4.937560in}}%
\pgfpathlineto{\pgfqpoint{6.104789in}{4.962086in}}%
\pgfpathlineto{\pgfqpoint{5.018822in}{5.003887in}}%
\pgfpathlineto{\pgfqpoint{3.932855in}{4.868778in}}%
\pgfpathlineto{\pgfqpoint{2.846888in}{4.788770in}}%
\pgfpathlineto{\pgfqpoint{1.760921in}{4.894324in}}%
\pgfpathlineto{\pgfqpoint{0.674954in}{4.803816in}}%
\pgfpathlineto{\pgfqpoint{0.674954in}{4.803816in}}%
\pgfpathclose%
\pgfusepath{stroke,fill}%
}%
\begin{pgfscope}%
\pgfsys@transformshift{0.000000in}{0.000000in}%
\pgfsys@useobject{currentmarker}{}%
\end{pgfscope}%
\end{pgfscope}%
\begin{pgfscope}%
\pgfpathrectangle{\pgfqpoint{0.674954in}{0.862305in}}{\pgfqpoint{6.515802in}{4.369081in}}%
\pgfusepath{clip}%
\pgfsetbuttcap%
\pgfsetroundjoin%
\definecolor{currentfill}{rgb}{0.960784,0.462745,0.000000}%
\pgfsetfillcolor{currentfill}%
\pgfsetfillopacity{0.300000}%
\pgfsetlinewidth{1.003750pt}%
\definecolor{currentstroke}{rgb}{0.960784,0.462745,0.000000}%
\pgfsetstrokecolor{currentstroke}%
\pgfsetstrokeopacity{0.300000}%
\pgfsetdash{{3.700000pt}{1.600000pt}}{0.000000pt}%
\pgfpathmoveto{\pgfqpoint{0.674954in}{4.664602in}}%
\pgfpathlineto{\pgfqpoint{0.674954in}{4.289079in}}%
\pgfpathlineto{\pgfqpoint{1.760921in}{4.481230in}}%
\pgfpathlineto{\pgfqpoint{2.846888in}{4.304691in}}%
\pgfpathlineto{\pgfqpoint{3.932855in}{4.328882in}}%
\pgfpathlineto{\pgfqpoint{5.018822in}{4.558393in}}%
\pgfpathlineto{\pgfqpoint{6.104789in}{4.490716in}}%
\pgfpathlineto{\pgfqpoint{6.740039in}{4.506859in}}%
\pgfpathlineto{\pgfqpoint{7.190756in}{4.501289in}}%
\pgfpathlineto{\pgfqpoint{7.190756in}{4.924759in}}%
\pgfpathlineto{\pgfqpoint{7.190756in}{4.924759in}}%
\pgfpathlineto{\pgfqpoint{6.740039in}{4.862011in}}%
\pgfpathlineto{\pgfqpoint{6.104789in}{4.816334in}}%
\pgfpathlineto{\pgfqpoint{5.018822in}{4.878344in}}%
\pgfpathlineto{\pgfqpoint{3.932855in}{4.667343in}}%
\pgfpathlineto{\pgfqpoint{2.846888in}{4.665707in}}%
\pgfpathlineto{\pgfqpoint{1.760921in}{4.821343in}}%
\pgfpathlineto{\pgfqpoint{0.674954in}{4.664602in}}%
\pgfpathlineto{\pgfqpoint{0.674954in}{4.664602in}}%
\pgfpathclose%
\pgfusepath{stroke,fill}%
\end{pgfscope}%
\begin{pgfscope}%
\pgfpathrectangle{\pgfqpoint{0.674954in}{0.862305in}}{\pgfqpoint{6.515802in}{4.369081in}}%
\pgfusepath{clip}%
\pgfsetbuttcap%
\pgfsetroundjoin%
\definecolor{currentfill}{rgb}{0.219608,0.219608,0.219608}%
\pgfsetfillcolor{currentfill}%
\pgfsetfillopacity{0.300000}%
\pgfsetlinewidth{1.003750pt}%
\definecolor{currentstroke}{rgb}{0.219608,0.219608,0.219608}%
\pgfsetstrokecolor{currentstroke}%
\pgfsetstrokeopacity{0.300000}%
\pgfsetdash{{1.000000pt}{1.650000pt}}{0.000000pt}%
\pgfpathmoveto{\pgfqpoint{0.674954in}{3.980412in}}%
\pgfpathlineto{\pgfqpoint{0.674954in}{3.630925in}}%
\pgfpathlineto{\pgfqpoint{1.760921in}{3.766175in}}%
\pgfpathlineto{\pgfqpoint{2.846888in}{3.767402in}}%
\pgfpathlineto{\pgfqpoint{3.932855in}{3.598929in}}%
\pgfpathlineto{\pgfqpoint{5.018822in}{3.976525in}}%
\pgfpathlineto{\pgfqpoint{6.104789in}{3.870159in}}%
\pgfpathlineto{\pgfqpoint{6.740039in}{4.012120in}}%
\pgfpathlineto{\pgfqpoint{7.190756in}{4.048732in}}%
\pgfpathlineto{\pgfqpoint{7.190756in}{4.517118in}}%
\pgfpathlineto{\pgfqpoint{7.190756in}{4.517118in}}%
\pgfpathlineto{\pgfqpoint{6.740039in}{4.385431in}}%
\pgfpathlineto{\pgfqpoint{6.104789in}{4.213473in}}%
\pgfpathlineto{\pgfqpoint{5.018822in}{4.388550in}}%
\pgfpathlineto{\pgfqpoint{3.932855in}{3.975070in}}%
\pgfpathlineto{\pgfqpoint{2.846888in}{4.184266in}}%
\pgfpathlineto{\pgfqpoint{1.760921in}{4.146609in}}%
\pgfpathlineto{\pgfqpoint{0.674954in}{3.980412in}}%
\pgfpathlineto{\pgfqpoint{0.674954in}{3.980412in}}%
\pgfpathclose%
\pgfusepath{stroke,fill}%
\end{pgfscope}%
\begin{pgfscope}%
\pgfpathrectangle{\pgfqpoint{0.674954in}{0.862305in}}{\pgfqpoint{6.515802in}{4.369081in}}%
\pgfusepath{clip}%
\pgfsetbuttcap%
\pgfsetroundjoin%
\pgfsetlinewidth{2.007500pt}%
\definecolor{currentstroke}{rgb}{0.501961,0.501961,0.501961}%
\pgfsetstrokecolor{currentstroke}%
\pgfsetstrokeopacity{0.300000}%
\pgfsetdash{{7.400000pt}{3.200000pt}}{0.000000pt}%
\pgfpathmoveto{\pgfqpoint{0.674954in}{0.862305in}}%
\pgfpathlineto{\pgfqpoint{0.674954in}{5.231386in}}%
\pgfusepath{stroke}%
\end{pgfscope}%
\begin{pgfscope}%
\pgfsetbuttcap%
\pgfsetroundjoin%
\definecolor{currentfill}{rgb}{0.000000,0.000000,0.000000}%
\pgfsetfillcolor{currentfill}%
\pgfsetlinewidth{0.803000pt}%
\definecolor{currentstroke}{rgb}{0.000000,0.000000,0.000000}%
\pgfsetstrokecolor{currentstroke}%
\pgfsetdash{}{0pt}%
\pgfsys@defobject{currentmarker}{\pgfqpoint{0.000000in}{-0.048611in}}{\pgfqpoint{0.000000in}{0.000000in}}{%
\pgfpathmoveto{\pgfqpoint{0.000000in}{0.000000in}}%
\pgfpathlineto{\pgfqpoint{0.000000in}{-0.048611in}}%
\pgfusepath{stroke,fill}%
}%
\begin{pgfscope}%
\pgfsys@transformshift{0.674954in}{0.862305in}%
\pgfsys@useobject{currentmarker}{}%
\end{pgfscope}%
\end{pgfscope}%
\begin{pgfscope}%
\definecolor{textcolor}{rgb}{0.000000,0.000000,0.000000}%
\pgfsetstrokecolor{textcolor}%
\pgfsetfillcolor{textcolor}%
\pgftext[x=0.674954in,y=0.765082in,,top]{\color{textcolor}{\rmfamily\fontsize{25.000000}{30.000000}\selectfont\catcode`\^=\active\def^{\ifmmode\sp\else\^{}\fi}\catcode`\%=\active\def
\end{pgfscope}%
\begin{pgfscope}%
\pgfpathrectangle{\pgfqpoint{0.674954in}{0.862305in}}{\pgfqpoint{6.515802in}{4.369081in}}%
\pgfusepath{clip}%
\pgfsetbuttcap%
\pgfsetroundjoin%
\pgfsetlinewidth{2.007500pt}%
\definecolor{currentstroke}{rgb}{0.501961,0.501961,0.501961}%
\pgfsetstrokecolor{currentstroke}%
\pgfsetstrokeopacity{0.300000}%
\pgfsetdash{{7.400000pt}{3.200000pt}}{0.000000pt}%
\pgfpathmoveto{\pgfqpoint{1.760921in}{0.862305in}}%
\pgfpathlineto{\pgfqpoint{1.760921in}{5.231386in}}%
\pgfusepath{stroke}%
\end{pgfscope}%
\begin{pgfscope}%
\pgfsetbuttcap%
\pgfsetroundjoin%
\definecolor{currentfill}{rgb}{0.000000,0.000000,0.000000}%
\pgfsetfillcolor{currentfill}%
\pgfsetlinewidth{0.803000pt}%
\definecolor{currentstroke}{rgb}{0.000000,0.000000,0.000000}%
\pgfsetstrokecolor{currentstroke}%
\pgfsetdash{}{0pt}%
\pgfsys@defobject{currentmarker}{\pgfqpoint{0.000000in}{-0.048611in}}{\pgfqpoint{0.000000in}{0.000000in}}{%
\pgfpathmoveto{\pgfqpoint{0.000000in}{0.000000in}}%
\pgfpathlineto{\pgfqpoint{0.000000in}{-0.048611in}}%
\pgfusepath{stroke,fill}%
}%
\begin{pgfscope}%
\pgfsys@transformshift{1.760921in}{0.862305in}%
\pgfsys@useobject{currentmarker}{}%
\end{pgfscope}%
\end{pgfscope}%
\begin{pgfscope}%
\definecolor{textcolor}{rgb}{0.000000,0.000000,0.000000}%
\pgfsetstrokecolor{textcolor}%
\pgfsetfillcolor{textcolor}%
\pgftext[x=1.760921in,y=0.765082in,,top]{\color{textcolor}{\rmfamily\fontsize{25.000000}{30.000000}\selectfont\catcode`\^=\active\def^{\ifmmode\sp\else\^{}\fi}\catcode`\%=\active\def
\end{pgfscope}%
\begin{pgfscope}%
\pgfpathrectangle{\pgfqpoint{0.674954in}{0.862305in}}{\pgfqpoint{6.515802in}{4.369081in}}%
\pgfusepath{clip}%
\pgfsetbuttcap%
\pgfsetroundjoin%
\pgfsetlinewidth{2.007500pt}%
\definecolor{currentstroke}{rgb}{0.501961,0.501961,0.501961}%
\pgfsetstrokecolor{currentstroke}%
\pgfsetstrokeopacity{0.300000}%
\pgfsetdash{{7.400000pt}{3.200000pt}}{0.000000pt}%
\pgfpathmoveto{\pgfqpoint{2.846888in}{0.862305in}}%
\pgfpathlineto{\pgfqpoint{2.846888in}{5.231386in}}%
\pgfusepath{stroke}%
\end{pgfscope}%
\begin{pgfscope}%
\pgfsetbuttcap%
\pgfsetroundjoin%
\definecolor{currentfill}{rgb}{0.000000,0.000000,0.000000}%
\pgfsetfillcolor{currentfill}%
\pgfsetlinewidth{0.803000pt}%
\definecolor{currentstroke}{rgb}{0.000000,0.000000,0.000000}%
\pgfsetstrokecolor{currentstroke}%
\pgfsetdash{}{0pt}%
\pgfsys@defobject{currentmarker}{\pgfqpoint{0.000000in}{-0.048611in}}{\pgfqpoint{0.000000in}{0.000000in}}{%
\pgfpathmoveto{\pgfqpoint{0.000000in}{0.000000in}}%
\pgfpathlineto{\pgfqpoint{0.000000in}{-0.048611in}}%
\pgfusepath{stroke,fill}%
}%
\begin{pgfscope}%
\pgfsys@transformshift{2.846888in}{0.862305in}%
\pgfsys@useobject{currentmarker}{}%
\end{pgfscope}%
\end{pgfscope}%
\begin{pgfscope}%
\definecolor{textcolor}{rgb}{0.000000,0.000000,0.000000}%
\pgfsetstrokecolor{textcolor}%
\pgfsetfillcolor{textcolor}%
\pgftext[x=2.846888in,y=0.765082in,,top]{\color{textcolor}{\rmfamily\fontsize{25.000000}{30.000000}\selectfont\catcode`\^=\active\def^{\ifmmode\sp\else\^{}\fi}\catcode`\%=\active\def
\end{pgfscope}%
\begin{pgfscope}%
\pgfpathrectangle{\pgfqpoint{0.674954in}{0.862305in}}{\pgfqpoint{6.515802in}{4.369081in}}%
\pgfusepath{clip}%
\pgfsetbuttcap%
\pgfsetroundjoin%
\pgfsetlinewidth{2.007500pt}%
\definecolor{currentstroke}{rgb}{0.501961,0.501961,0.501961}%
\pgfsetstrokecolor{currentstroke}%
\pgfsetstrokeopacity{0.300000}%
\pgfsetdash{{7.400000pt}{3.200000pt}}{0.000000pt}%
\pgfpathmoveto{\pgfqpoint{3.932855in}{0.862305in}}%
\pgfpathlineto{\pgfqpoint{3.932855in}{5.231386in}}%
\pgfusepath{stroke}%
\end{pgfscope}%
\begin{pgfscope}%
\pgfsetbuttcap%
\pgfsetroundjoin%
\definecolor{currentfill}{rgb}{0.000000,0.000000,0.000000}%
\pgfsetfillcolor{currentfill}%
\pgfsetlinewidth{0.803000pt}%
\definecolor{currentstroke}{rgb}{0.000000,0.000000,0.000000}%
\pgfsetstrokecolor{currentstroke}%
\pgfsetdash{}{0pt}%
\pgfsys@defobject{currentmarker}{\pgfqpoint{0.000000in}{-0.048611in}}{\pgfqpoint{0.000000in}{0.000000in}}{%
\pgfpathmoveto{\pgfqpoint{0.000000in}{0.000000in}}%
\pgfpathlineto{\pgfqpoint{0.000000in}{-0.048611in}}%
\pgfusepath{stroke,fill}%
}%
\begin{pgfscope}%
\pgfsys@transformshift{3.932855in}{0.862305in}%
\pgfsys@useobject{currentmarker}{}%
\end{pgfscope}%
\end{pgfscope}%
\begin{pgfscope}%
\definecolor{textcolor}{rgb}{0.000000,0.000000,0.000000}%
\pgfsetstrokecolor{textcolor}%
\pgfsetfillcolor{textcolor}%
\pgftext[x=3.932855in,y=0.765082in,,top]{\color{textcolor}{\rmfamily\fontsize{25.000000}{30.000000}\selectfont\catcode`\^=\active\def^{\ifmmode\sp\else\^{}\fi}\catcode`\%=\active\def
\end{pgfscope}%
\begin{pgfscope}%
\pgfpathrectangle{\pgfqpoint{0.674954in}{0.862305in}}{\pgfqpoint{6.515802in}{4.369081in}}%
\pgfusepath{clip}%
\pgfsetbuttcap%
\pgfsetroundjoin%
\pgfsetlinewidth{2.007500pt}%
\definecolor{currentstroke}{rgb}{0.501961,0.501961,0.501961}%
\pgfsetstrokecolor{currentstroke}%
\pgfsetstrokeopacity{0.300000}%
\pgfsetdash{{7.400000pt}{3.200000pt}}{0.000000pt}%
\pgfpathmoveto{\pgfqpoint{5.018822in}{0.862305in}}%
\pgfpathlineto{\pgfqpoint{5.018822in}{5.231386in}}%
\pgfusepath{stroke}%
\end{pgfscope}%
\begin{pgfscope}%
\pgfsetbuttcap%
\pgfsetroundjoin%
\definecolor{currentfill}{rgb}{0.000000,0.000000,0.000000}%
\pgfsetfillcolor{currentfill}%
\pgfsetlinewidth{0.803000pt}%
\definecolor{currentstroke}{rgb}{0.000000,0.000000,0.000000}%
\pgfsetstrokecolor{currentstroke}%
\pgfsetdash{}{0pt}%
\pgfsys@defobject{currentmarker}{\pgfqpoint{0.000000in}{-0.048611in}}{\pgfqpoint{0.000000in}{0.000000in}}{%
\pgfpathmoveto{\pgfqpoint{0.000000in}{0.000000in}}%
\pgfpathlineto{\pgfqpoint{0.000000in}{-0.048611in}}%
\pgfusepath{stroke,fill}%
}%
\begin{pgfscope}%
\pgfsys@transformshift{5.018822in}{0.862305in}%
\pgfsys@useobject{currentmarker}{}%
\end{pgfscope}%
\end{pgfscope}%
\begin{pgfscope}%
\definecolor{textcolor}{rgb}{0.000000,0.000000,0.000000}%
\pgfsetstrokecolor{textcolor}%
\pgfsetfillcolor{textcolor}%
\pgftext[x=5.018822in,y=0.765082in,,top]{\color{textcolor}{\rmfamily\fontsize{25.000000}{30.000000}\selectfont\catcode`\^=\active\def^{\ifmmode\sp\else\^{}\fi}\catcode`\%=\active\def
\end{pgfscope}%
\begin{pgfscope}%
\pgfpathrectangle{\pgfqpoint{0.674954in}{0.862305in}}{\pgfqpoint{6.515802in}{4.369081in}}%
\pgfusepath{clip}%
\pgfsetbuttcap%
\pgfsetroundjoin%
\pgfsetlinewidth{2.007500pt}%
\definecolor{currentstroke}{rgb}{0.501961,0.501961,0.501961}%
\pgfsetstrokecolor{currentstroke}%
\pgfsetstrokeopacity{0.300000}%
\pgfsetdash{{7.400000pt}{3.200000pt}}{0.000000pt}%
\pgfpathmoveto{\pgfqpoint{6.104789in}{0.862305in}}%
\pgfpathlineto{\pgfqpoint{6.104789in}{5.231386in}}%
\pgfusepath{stroke}%
\end{pgfscope}%
\begin{pgfscope}%
\pgfsetbuttcap%
\pgfsetroundjoin%
\definecolor{currentfill}{rgb}{0.000000,0.000000,0.000000}%
\pgfsetfillcolor{currentfill}%
\pgfsetlinewidth{0.803000pt}%
\definecolor{currentstroke}{rgb}{0.000000,0.000000,0.000000}%
\pgfsetstrokecolor{currentstroke}%
\pgfsetdash{}{0pt}%
\pgfsys@defobject{currentmarker}{\pgfqpoint{0.000000in}{-0.048611in}}{\pgfqpoint{0.000000in}{0.000000in}}{%
\pgfpathmoveto{\pgfqpoint{0.000000in}{0.000000in}}%
\pgfpathlineto{\pgfqpoint{0.000000in}{-0.048611in}}%
\pgfusepath{stroke,fill}%
}%
\begin{pgfscope}%
\pgfsys@transformshift{6.104789in}{0.862305in}%
\pgfsys@useobject{currentmarker}{}%
\end{pgfscope}%
\end{pgfscope}%
\begin{pgfscope}%
\definecolor{textcolor}{rgb}{0.000000,0.000000,0.000000}%
\pgfsetstrokecolor{textcolor}%
\pgfsetfillcolor{textcolor}%
\pgftext[x=6.104789in,y=0.765082in,,top]{\color{textcolor}{\rmfamily\fontsize{25.000000}{30.000000}\selectfont\catcode`\^=\active\def^{\ifmmode\sp\else\^{}\fi}\catcode`\%=\active\def
\end{pgfscope}%
\begin{pgfscope}%
\pgfpathrectangle{\pgfqpoint{0.674954in}{0.862305in}}{\pgfqpoint{6.515802in}{4.369081in}}%
\pgfusepath{clip}%
\pgfsetbuttcap%
\pgfsetroundjoin%
\pgfsetlinewidth{2.007500pt}%
\definecolor{currentstroke}{rgb}{0.501961,0.501961,0.501961}%
\pgfsetstrokecolor{currentstroke}%
\pgfsetstrokeopacity{0.300000}%
\pgfsetdash{{7.400000pt}{3.200000pt}}{0.000000pt}%
\pgfpathmoveto{\pgfqpoint{6.740039in}{0.862305in}}%
\pgfpathlineto{\pgfqpoint{6.740039in}{5.231386in}}%
\pgfusepath{stroke}%
\end{pgfscope}%
\begin{pgfscope}%
\pgfsetbuttcap%
\pgfsetroundjoin%
\definecolor{currentfill}{rgb}{0.000000,0.000000,0.000000}%
\pgfsetfillcolor{currentfill}%
\pgfsetlinewidth{0.803000pt}%
\definecolor{currentstroke}{rgb}{0.000000,0.000000,0.000000}%
\pgfsetstrokecolor{currentstroke}%
\pgfsetdash{}{0pt}%
\pgfsys@defobject{currentmarker}{\pgfqpoint{0.000000in}{-0.048611in}}{\pgfqpoint{0.000000in}{0.000000in}}{%
\pgfpathmoveto{\pgfqpoint{0.000000in}{0.000000in}}%
\pgfpathlineto{\pgfqpoint{0.000000in}{-0.048611in}}%
\pgfusepath{stroke,fill}%
}%
\begin{pgfscope}%
\pgfsys@transformshift{6.740039in}{0.862305in}%
\pgfsys@useobject{currentmarker}{}%
\end{pgfscope}%
\end{pgfscope}%
\begin{pgfscope}%
\definecolor{textcolor}{rgb}{0.000000,0.000000,0.000000}%
\pgfsetstrokecolor{textcolor}%
\pgfsetfillcolor{textcolor}%
\pgftext[x=6.740039in,y=0.765082in,,top]{\color{textcolor}{\rmfamily\fontsize{25.000000}{30.000000}\selectfont\catcode`\^=\active\def^{\ifmmode\sp\else\^{}\fi}\catcode`\%=\active\def
\end{pgfscope}%
\begin{pgfscope}%
\pgfpathrectangle{\pgfqpoint{0.674954in}{0.862305in}}{\pgfqpoint{6.515802in}{4.369081in}}%
\pgfusepath{clip}%
\pgfsetbuttcap%
\pgfsetroundjoin%
\pgfsetlinewidth{2.007500pt}%
\definecolor{currentstroke}{rgb}{0.501961,0.501961,0.501961}%
\pgfsetstrokecolor{currentstroke}%
\pgfsetstrokeopacity{0.300000}%
\pgfsetdash{{7.400000pt}{3.200000pt}}{0.000000pt}%
\pgfpathmoveto{\pgfqpoint{7.190756in}{0.862305in}}%
\pgfpathlineto{\pgfqpoint{7.190756in}{5.231386in}}%
\pgfusepath{stroke}%
\end{pgfscope}%
\begin{pgfscope}%
\pgfsetbuttcap%
\pgfsetroundjoin%
\definecolor{currentfill}{rgb}{0.000000,0.000000,0.000000}%
\pgfsetfillcolor{currentfill}%
\pgfsetlinewidth{0.803000pt}%
\definecolor{currentstroke}{rgb}{0.000000,0.000000,0.000000}%
\pgfsetstrokecolor{currentstroke}%
\pgfsetdash{}{0pt}%
\pgfsys@defobject{currentmarker}{\pgfqpoint{0.000000in}{-0.048611in}}{\pgfqpoint{0.000000in}{0.000000in}}{%
\pgfpathmoveto{\pgfqpoint{0.000000in}{0.000000in}}%
\pgfpathlineto{\pgfqpoint{0.000000in}{-0.048611in}}%
\pgfusepath{stroke,fill}%
}%
\begin{pgfscope}%
\pgfsys@transformshift{7.190756in}{0.862305in}%
\pgfsys@useobject{currentmarker}{}%
\end{pgfscope}%
\end{pgfscope}%
\begin{pgfscope}%
\definecolor{textcolor}{rgb}{0.000000,0.000000,0.000000}%
\pgfsetstrokecolor{textcolor}%
\pgfsetfillcolor{textcolor}%
\pgftext[x=7.190756in,y=0.765082in,,top]{\color{textcolor}{\rmfamily\fontsize{25.000000}{30.000000}\selectfont\catcode`\^=\active\def^{\ifmmode\sp\else\^{}\fi}\catcode`\%=\active\def
\end{pgfscope}%
\begin{pgfscope}%
\definecolor{textcolor}{rgb}{0.000000,0.000000,0.000000}%
\pgfsetstrokecolor{textcolor}%
\pgfsetfillcolor{textcolor}%
\pgftext[x=3.932855in,y=0.404763in,,top]{\color{textcolor}{\rmfamily\fontsize{25.000000}{30.000000}\selectfont\catcode`\^=\active\def^{\ifmmode\sp\else\^{}\fi}\catcode`\%=\active\def
\end{pgfscope}%
\begin{pgfscope}%
\pgfpathrectangle{\pgfqpoint{0.674954in}{0.862305in}}{\pgfqpoint{6.515802in}{4.369081in}}%
\pgfusepath{clip}%
\pgfsetbuttcap%
\pgfsetroundjoin%
\pgfsetlinewidth{2.007500pt}%
\definecolor{currentstroke}{rgb}{0.501961,0.501961,0.501961}%
\pgfsetstrokecolor{currentstroke}%
\pgfsetstrokeopacity{0.300000}%
\pgfsetdash{{7.400000pt}{3.200000pt}}{0.000000pt}%
\pgfpathmoveto{\pgfqpoint{0.674954in}{1.954575in}}%
\pgfpathlineto{\pgfqpoint{7.190756in}{1.954575in}}%
\pgfusepath{stroke}%
\end{pgfscope}%
\begin{pgfscope}%
\pgfsetbuttcap%
\pgfsetroundjoin%
\definecolor{currentfill}{rgb}{0.000000,0.000000,0.000000}%
\pgfsetfillcolor{currentfill}%
\pgfsetlinewidth{0.803000pt}%
\definecolor{currentstroke}{rgb}{0.000000,0.000000,0.000000}%
\pgfsetstrokecolor{currentstroke}%
\pgfsetdash{}{0pt}%
\pgfsys@defobject{currentmarker}{\pgfqpoint{-0.048611in}{0.000000in}}{\pgfqpoint{-0.000000in}{0.000000in}}{%
\pgfpathmoveto{\pgfqpoint{-0.000000in}{0.000000in}}%
\pgfpathlineto{\pgfqpoint{-0.048611in}{0.000000in}}%
\pgfusepath{stroke,fill}%
}%
\begin{pgfscope}%
\pgfsys@transformshift{0.674954in}{1.954575in}%
\pgfsys@useobject{currentmarker}{}%
\end{pgfscope}%
\end{pgfscope}%
\begin{pgfscope}%
\definecolor{textcolor}{rgb}{0.000000,0.000000,0.000000}%
\pgfsetstrokecolor{textcolor}%
\pgfsetfillcolor{textcolor}%
\pgftext[x=0.259244in, y=1.835961in, left, base]{\color{textcolor}{\rmfamily\fontsize{25.000000}{30.000000}\selectfont\catcode`\^=\active\def^{\ifmmode\sp\else\^{}\fi}\catcode`\%=\active\def
\end{pgfscope}%
\begin{pgfscope}%
\pgfpathrectangle{\pgfqpoint{0.674954in}{0.862305in}}{\pgfqpoint{6.515802in}{4.369081in}}%
\pgfusepath{clip}%
\pgfsetbuttcap%
\pgfsetroundjoin%
\pgfsetlinewidth{2.007500pt}%
\definecolor{currentstroke}{rgb}{0.501961,0.501961,0.501961}%
\pgfsetstrokecolor{currentstroke}%
\pgfsetstrokeopacity{0.300000}%
\pgfsetdash{{7.400000pt}{3.200000pt}}{0.000000pt}%
\pgfpathmoveto{\pgfqpoint{0.674954in}{3.046845in}}%
\pgfpathlineto{\pgfqpoint{7.190756in}{3.046845in}}%
\pgfusepath{stroke}%
\end{pgfscope}%
\begin{pgfscope}%
\pgfsetbuttcap%
\pgfsetroundjoin%
\definecolor{currentfill}{rgb}{0.000000,0.000000,0.000000}%
\pgfsetfillcolor{currentfill}%
\pgfsetlinewidth{0.803000pt}%
\definecolor{currentstroke}{rgb}{0.000000,0.000000,0.000000}%
\pgfsetstrokecolor{currentstroke}%
\pgfsetdash{}{0pt}%
\pgfsys@defobject{currentmarker}{\pgfqpoint{-0.048611in}{0.000000in}}{\pgfqpoint{-0.000000in}{0.000000in}}{%
\pgfpathmoveto{\pgfqpoint{-0.000000in}{0.000000in}}%
\pgfpathlineto{\pgfqpoint{-0.048611in}{0.000000in}}%
\pgfusepath{stroke,fill}%
}%
\begin{pgfscope}%
\pgfsys@transformshift{0.674954in}{3.046845in}%
\pgfsys@useobject{currentmarker}{}%
\end{pgfscope}%
\end{pgfscope}%
\begin{pgfscope}%
\definecolor{textcolor}{rgb}{0.000000,0.000000,0.000000}%
\pgfsetstrokecolor{textcolor}%
\pgfsetfillcolor{textcolor}%
\pgftext[x=0.259244in, y=2.928231in, left, base]{\color{textcolor}{\rmfamily\fontsize{25.000000}{30.000000}\selectfont\catcode`\^=\active\def^{\ifmmode\sp\else\^{}\fi}\catcode`\%=\active\def
\end{pgfscope}%
\begin{pgfscope}%
\pgfpathrectangle{\pgfqpoint{0.674954in}{0.862305in}}{\pgfqpoint{6.515802in}{4.369081in}}%
\pgfusepath{clip}%
\pgfsetbuttcap%
\pgfsetroundjoin%
\pgfsetlinewidth{2.007500pt}%
\definecolor{currentstroke}{rgb}{0.501961,0.501961,0.501961}%
\pgfsetstrokecolor{currentstroke}%
\pgfsetstrokeopacity{0.300000}%
\pgfsetdash{{7.400000pt}{3.200000pt}}{0.000000pt}%
\pgfpathmoveto{\pgfqpoint{0.674954in}{4.139116in}}%
\pgfpathlineto{\pgfqpoint{7.190756in}{4.139116in}}%
\pgfusepath{stroke}%
\end{pgfscope}%
\begin{pgfscope}%
\pgfsetbuttcap%
\pgfsetroundjoin%
\definecolor{currentfill}{rgb}{0.000000,0.000000,0.000000}%
\pgfsetfillcolor{currentfill}%
\pgfsetlinewidth{0.803000pt}%
\definecolor{currentstroke}{rgb}{0.000000,0.000000,0.000000}%
\pgfsetstrokecolor{currentstroke}%
\pgfsetdash{}{0pt}%
\pgfsys@defobject{currentmarker}{\pgfqpoint{-0.048611in}{0.000000in}}{\pgfqpoint{-0.000000in}{0.000000in}}{%
\pgfpathmoveto{\pgfqpoint{-0.000000in}{0.000000in}}%
\pgfpathlineto{\pgfqpoint{-0.048611in}{0.000000in}}%
\pgfusepath{stroke,fill}%
}%
\begin{pgfscope}%
\pgfsys@transformshift{0.674954in}{4.139116in}%
\pgfsys@useobject{currentmarker}{}%
\end{pgfscope}%
\end{pgfscope}%
\begin{pgfscope}%
\definecolor{textcolor}{rgb}{0.000000,0.000000,0.000000}%
\pgfsetstrokecolor{textcolor}%
\pgfsetfillcolor{textcolor}%
\pgftext[x=0.259244in, y=4.020501in, left, base]{\color{textcolor}{\rmfamily\fontsize{25.000000}{30.000000}\selectfont\catcode`\^=\active\def^{\ifmmode\sp\else\^{}\fi}\catcode`\%=\active\def
\end{pgfscope}%
\begin{pgfscope}%
\pgfpathrectangle{\pgfqpoint{0.674954in}{0.862305in}}{\pgfqpoint{6.515802in}{4.369081in}}%
\pgfusepath{clip}%
\pgfsetbuttcap%
\pgfsetroundjoin%
\pgfsetlinewidth{2.007500pt}%
\definecolor{currentstroke}{rgb}{0.501961,0.501961,0.501961}%
\pgfsetstrokecolor{currentstroke}%
\pgfsetstrokeopacity{0.300000}%
\pgfsetdash{{7.400000pt}{3.200000pt}}{0.000000pt}%
\pgfpathmoveto{\pgfqpoint{0.674954in}{5.231386in}}%
\pgfpathlineto{\pgfqpoint{7.190756in}{5.231386in}}%
\pgfusepath{stroke}%
\end{pgfscope}%
\begin{pgfscope}%
\pgfsetbuttcap%
\pgfsetroundjoin%
\definecolor{currentfill}{rgb}{0.000000,0.000000,0.000000}%
\pgfsetfillcolor{currentfill}%
\pgfsetlinewidth{0.803000pt}%
\definecolor{currentstroke}{rgb}{0.000000,0.000000,0.000000}%
\pgfsetstrokecolor{currentstroke}%
\pgfsetdash{}{0pt}%
\pgfsys@defobject{currentmarker}{\pgfqpoint{-0.048611in}{0.000000in}}{\pgfqpoint{-0.000000in}{0.000000in}}{%
\pgfpathmoveto{\pgfqpoint{-0.000000in}{0.000000in}}%
\pgfpathlineto{\pgfqpoint{-0.048611in}{0.000000in}}%
\pgfusepath{stroke,fill}%
}%
\begin{pgfscope}%
\pgfsys@transformshift{0.674954in}{5.231386in}%
\pgfsys@useobject{currentmarker}{}%
\end{pgfscope}%
\end{pgfscope}%
\begin{pgfscope}%
\definecolor{textcolor}{rgb}{0.000000,0.000000,0.000000}%
\pgfsetstrokecolor{textcolor}%
\pgfsetfillcolor{textcolor}%
\pgftext[x=0.100000in, y=5.112772in, left, base]{\color{textcolor}{\rmfamily\fontsize{25.000000}{30.000000}\selectfont\catcode`\^=\active\def^{\ifmmode\sp\else\^{}\fi}\catcode`\%=\active\def
\end{pgfscope}%
\begin{pgfscope}%
\pgfpathrectangle{\pgfqpoint{0.674954in}{0.862305in}}{\pgfqpoint{6.515802in}{4.369081in}}%
\pgfusepath{clip}%
\pgfsetrectcap%
\pgfsetroundjoin%
\pgfsetlinewidth{2.509375pt}%
\definecolor{currentstroke}{rgb}{0.050980,0.415686,0.509804}%
\pgfsetstrokecolor{currentstroke}%
\pgfsetdash{}{0pt}%
\pgfpathmoveto{\pgfqpoint{0.674954in}{4.642444in}}%
\pgfpathlineto{\pgfqpoint{1.760921in}{4.760496in}}%
\pgfpathlineto{\pgfqpoint{2.846888in}{4.626379in}}%
\pgfpathlineto{\pgfqpoint{3.932855in}{4.733685in}}%
\pgfpathlineto{\pgfqpoint{5.018822in}{4.873919in}}%
\pgfpathlineto{\pgfqpoint{6.104789in}{4.823700in}}%
\pgfpathlineto{\pgfqpoint{6.740039in}{4.797312in}}%
\pgfpathlineto{\pgfqpoint{7.190756in}{4.820665in}}%
\pgfusepath{stroke}%
\end{pgfscope}%
\begin{pgfscope}%
\pgfpathrectangle{\pgfqpoint{0.674954in}{0.862305in}}{\pgfqpoint{6.515802in}{4.369081in}}%
\pgfusepath{clip}%
\pgfsetbuttcap%
\pgfsetroundjoin%
\definecolor{currentfill}{rgb}{0.050980,0.415686,0.509804}%
\pgfsetfillcolor{currentfill}%
\pgfsetlinewidth{1.003750pt}%
\definecolor{currentstroke}{rgb}{0.050980,0.415686,0.509804}%
\pgfsetstrokecolor{currentstroke}%
\pgfsetdash{}{0pt}%
\pgfsys@defobject{currentmarker}{\pgfqpoint{-0.055556in}{-0.055556in}}{\pgfqpoint{0.055556in}{0.055556in}}{%
\pgfpathmoveto{\pgfqpoint{0.000000in}{-0.055556in}}%
\pgfpathcurveto{\pgfqpoint{0.014734in}{-0.055556in}}{\pgfqpoint{0.028866in}{-0.049702in}}{\pgfqpoint{0.039284in}{-0.039284in}}%
\pgfpathcurveto{\pgfqpoint{0.049702in}{-0.028866in}}{\pgfqpoint{0.055556in}{-0.014734in}}{\pgfqpoint{0.055556in}{0.000000in}}%
\pgfpathcurveto{\pgfqpoint{0.055556in}{0.014734in}}{\pgfqpoint{0.049702in}{0.028866in}}{\pgfqpoint{0.039284in}{0.039284in}}%
\pgfpathcurveto{\pgfqpoint{0.028866in}{0.049702in}}{\pgfqpoint{0.014734in}{0.055556in}}{\pgfqpoint{0.000000in}{0.055556in}}%
\pgfpathcurveto{\pgfqpoint{-0.014734in}{0.055556in}}{\pgfqpoint{-0.028866in}{0.049702in}}{\pgfqpoint{-0.039284in}{0.039284in}}%
\pgfpathcurveto{\pgfqpoint{-0.049702in}{0.028866in}}{\pgfqpoint{-0.055556in}{0.014734in}}{\pgfqpoint{-0.055556in}{0.000000in}}%
\pgfpathcurveto{\pgfqpoint{-0.055556in}{-0.014734in}}{\pgfqpoint{-0.049702in}{-0.028866in}}{\pgfqpoint{-0.039284in}{-0.039284in}}%
\pgfpathcurveto{\pgfqpoint{-0.028866in}{-0.049702in}}{\pgfqpoint{-0.014734in}{-0.055556in}}{\pgfqpoint{0.000000in}{-0.055556in}}%
\pgfpathlineto{\pgfqpoint{0.000000in}{-0.055556in}}%
\pgfpathclose%
\pgfusepath{stroke,fill}%
}%
\begin{pgfscope}%
\pgfsys@transformshift{0.674954in}{4.642444in}%
\pgfsys@useobject{currentmarker}{}%
\end{pgfscope}%
\begin{pgfscope}%
\pgfsys@transformshift{1.760921in}{4.760496in}%
\pgfsys@useobject{currentmarker}{}%
\end{pgfscope}%
\begin{pgfscope}%
\pgfsys@transformshift{2.846888in}{4.626379in}%
\pgfsys@useobject{currentmarker}{}%
\end{pgfscope}%
\begin{pgfscope}%
\pgfsys@transformshift{3.932855in}{4.733685in}%
\pgfsys@useobject{currentmarker}{}%
\end{pgfscope}%
\begin{pgfscope}%
\pgfsys@transformshift{5.018822in}{4.873919in}%
\pgfsys@useobject{currentmarker}{}%
\end{pgfscope}%
\begin{pgfscope}%
\pgfsys@transformshift{6.104789in}{4.823700in}%
\pgfsys@useobject{currentmarker}{}%
\end{pgfscope}%
\begin{pgfscope}%
\pgfsys@transformshift{6.740039in}{4.797312in}%
\pgfsys@useobject{currentmarker}{}%
\end{pgfscope}%
\begin{pgfscope}%
\pgfsys@transformshift{7.190756in}{4.820665in}%
\pgfsys@useobject{currentmarker}{}%
\end{pgfscope}%
\end{pgfscope}%
\begin{pgfscope}%
\pgfpathrectangle{\pgfqpoint{0.674954in}{0.862305in}}{\pgfqpoint{6.515802in}{4.369081in}}%
\pgfusepath{clip}%
\pgfsetbuttcap%
\pgfsetroundjoin%
\pgfsetlinewidth{2.509375pt}%
\definecolor{currentstroke}{rgb}{0.960784,0.462745,0.000000}%
\pgfsetstrokecolor{currentstroke}%
\pgfsetdash{{9.250000pt}{4.000000pt}}{0.000000pt}%
\pgfpathmoveto{\pgfqpoint{0.674954in}{4.482593in}}%
\pgfpathlineto{\pgfqpoint{1.760921in}{4.661557in}}%
\pgfpathlineto{\pgfqpoint{2.846888in}{4.492205in}}%
\pgfpathlineto{\pgfqpoint{3.932855in}{4.509040in}}%
\pgfpathlineto{\pgfqpoint{5.018822in}{4.726865in}}%
\pgfpathlineto{\pgfqpoint{6.104789in}{4.664258in}}%
\pgfpathlineto{\pgfqpoint{6.740039in}{4.700368in}}%
\pgfpathlineto{\pgfqpoint{7.190756in}{4.711241in}}%
\pgfusepath{stroke}%
\end{pgfscope}%
\begin{pgfscope}%
\pgfpathrectangle{\pgfqpoint{0.674954in}{0.862305in}}{\pgfqpoint{6.515802in}{4.369081in}}%
\pgfusepath{clip}%
\pgfsetbuttcap%
\pgfsetmiterjoin%
\definecolor{currentfill}{rgb}{0.960784,0.462745,0.000000}%
\pgfsetfillcolor{currentfill}%
\pgfsetlinewidth{1.003750pt}%
\definecolor{currentstroke}{rgb}{0.960784,0.462745,0.000000}%
\pgfsetstrokecolor{currentstroke}%
\pgfsetdash{}{0pt}%
\pgfsys@defobject{currentmarker}{\pgfqpoint{-0.055556in}{-0.055556in}}{\pgfqpoint{0.055556in}{0.055556in}}{%
\pgfpathmoveto{\pgfqpoint{-0.055556in}{-0.055556in}}%
\pgfpathlineto{\pgfqpoint{0.055556in}{-0.055556in}}%
\pgfpathlineto{\pgfqpoint{0.055556in}{0.055556in}}%
\pgfpathlineto{\pgfqpoint{-0.055556in}{0.055556in}}%
\pgfpathlineto{\pgfqpoint{-0.055556in}{-0.055556in}}%
\pgfpathclose%
\pgfusepath{stroke,fill}%
}%
\begin{pgfscope}%
\pgfsys@transformshift{0.674954in}{4.482593in}%
\pgfsys@useobject{currentmarker}{}%
\end{pgfscope}%
\begin{pgfscope}%
\pgfsys@transformshift{1.760921in}{4.661557in}%
\pgfsys@useobject{currentmarker}{}%
\end{pgfscope}%
\begin{pgfscope}%
\pgfsys@transformshift{2.846888in}{4.492205in}%
\pgfsys@useobject{currentmarker}{}%
\end{pgfscope}%
\begin{pgfscope}%
\pgfsys@transformshift{3.932855in}{4.509040in}%
\pgfsys@useobject{currentmarker}{}%
\end{pgfscope}%
\begin{pgfscope}%
\pgfsys@transformshift{5.018822in}{4.726865in}%
\pgfsys@useobject{currentmarker}{}%
\end{pgfscope}%
\begin{pgfscope}%
\pgfsys@transformshift{6.104789in}{4.664258in}%
\pgfsys@useobject{currentmarker}{}%
\end{pgfscope}%
\begin{pgfscope}%
\pgfsys@transformshift{6.740039in}{4.700368in}%
\pgfsys@useobject{currentmarker}{}%
\end{pgfscope}%
\begin{pgfscope}%
\pgfsys@transformshift{7.190756in}{4.711241in}%
\pgfsys@useobject{currentmarker}{}%
\end{pgfscope}%
\end{pgfscope}%
\begin{pgfscope}%
\pgfpathrectangle{\pgfqpoint{0.674954in}{0.862305in}}{\pgfqpoint{6.515802in}{4.369081in}}%
\pgfusepath{clip}%
\pgfsetbuttcap%
\pgfsetroundjoin%
\pgfsetlinewidth{2.509375pt}%
\definecolor{currentstroke}{rgb}{0.219608,0.219608,0.219608}%
\pgfsetstrokecolor{currentstroke}%
\pgfsetdash{{2.500000pt}{4.125000pt}}{0.000000pt}%
\pgfpathmoveto{\pgfqpoint{0.674954in}{3.802212in}}%
\pgfpathlineto{\pgfqpoint{1.760921in}{3.943038in}}%
\pgfpathlineto{\pgfqpoint{2.846888in}{3.982417in}}%
\pgfpathlineto{\pgfqpoint{3.932855in}{3.787671in}}%
\pgfpathlineto{\pgfqpoint{5.018822in}{4.196106in}}%
\pgfpathlineto{\pgfqpoint{6.104789in}{4.036967in}}%
\pgfpathlineto{\pgfqpoint{6.740039in}{4.215449in}}%
\pgfpathlineto{\pgfqpoint{7.190756in}{4.288028in}}%
\pgfusepath{stroke}%
\end{pgfscope}%
\begin{pgfscope}%
\pgfpathrectangle{\pgfqpoint{0.674954in}{0.862305in}}{\pgfqpoint{6.515802in}{4.369081in}}%
\pgfusepath{clip}%
\pgfsetbuttcap%
\pgfsetmiterjoin%
\definecolor{currentfill}{rgb}{0.219608,0.219608,0.219608}%
\pgfsetfillcolor{currentfill}%
\pgfsetlinewidth{1.003750pt}%
\definecolor{currentstroke}{rgb}{0.219608,0.219608,0.219608}%
\pgfsetstrokecolor{currentstroke}%
\pgfsetdash{}{0pt}%
\pgfsys@defobject{currentmarker}{\pgfqpoint{-0.055556in}{-0.055556in}}{\pgfqpoint{0.055556in}{0.055556in}}{%
\pgfpathmoveto{\pgfqpoint{0.000000in}{0.055556in}}%
\pgfpathlineto{\pgfqpoint{-0.055556in}{-0.055556in}}%
\pgfpathlineto{\pgfqpoint{0.055556in}{-0.055556in}}%
\pgfpathlineto{\pgfqpoint{0.000000in}{0.055556in}}%
\pgfpathclose%
\pgfusepath{stroke,fill}%
}%
\begin{pgfscope}%
\pgfsys@transformshift{0.674954in}{3.802212in}%
\pgfsys@useobject{currentmarker}{}%
\end{pgfscope}%
\begin{pgfscope}%
\pgfsys@transformshift{1.760921in}{3.943038in}%
\pgfsys@useobject{currentmarker}{}%
\end{pgfscope}%
\begin{pgfscope}%
\pgfsys@transformshift{2.846888in}{3.982417in}%
\pgfsys@useobject{currentmarker}{}%
\end{pgfscope}%
\begin{pgfscope}%
\pgfsys@transformshift{3.932855in}{3.787671in}%
\pgfsys@useobject{currentmarker}{}%
\end{pgfscope}%
\begin{pgfscope}%
\pgfsys@transformshift{5.018822in}{4.196106in}%
\pgfsys@useobject{currentmarker}{}%
\end{pgfscope}%
\begin{pgfscope}%
\pgfsys@transformshift{6.104789in}{4.036967in}%
\pgfsys@useobject{currentmarker}{}%
\end{pgfscope}%
\begin{pgfscope}%
\pgfsys@transformshift{6.740039in}{4.215449in}%
\pgfsys@useobject{currentmarker}{}%
\end{pgfscope}%
\begin{pgfscope}%
\pgfsys@transformshift{7.190756in}{4.288028in}%
\pgfsys@useobject{currentmarker}{}%
\end{pgfscope}%
\end{pgfscope}%
\begin{pgfscope}%
\pgfsetrectcap%
\pgfsetmiterjoin%
\pgfsetlinewidth{2.007500pt}%
\definecolor{currentstroke}{rgb}{0.000000,0.000000,0.000000}%
\pgfsetstrokecolor{currentstroke}%
\pgfsetdash{}{0pt}%
\pgfpathmoveto{\pgfqpoint{0.674954in}{0.862305in}}%
\pgfpathlineto{\pgfqpoint{0.674954in}{5.231386in}}%
\pgfusepath{stroke}%
\end{pgfscope}%
\begin{pgfscope}%
\pgfsetrectcap%
\pgfsetmiterjoin%
\pgfsetlinewidth{2.007500pt}%
\definecolor{currentstroke}{rgb}{0.000000,0.000000,0.000000}%
\pgfsetstrokecolor{currentstroke}%
\pgfsetdash{}{0pt}%
\pgfpathmoveto{\pgfqpoint{0.674954in}{0.862305in}}%
\pgfpathlineto{\pgfqpoint{7.190756in}{0.862305in}}%
\pgfusepath{stroke}%
\end{pgfscope}%
\begin{pgfscope}%
\pgfsetbuttcap%
\pgfsetmiterjoin%
\definecolor{currentfill}{rgb}{1.000000,1.000000,1.000000}%
\pgfsetfillcolor{currentfill}%
\pgfsetlinewidth{1.003750pt}%
\definecolor{currentstroke}{rgb}{0.800000,0.800000,0.800000}%
\pgfsetstrokecolor{currentstroke}%
\pgfsetdash{}{0pt}%
\pgfpathmoveto{\pgfqpoint{0.869398in}{1.001194in}}%
\pgfpathlineto{\pgfqpoint{2.213036in}{1.001194in}}%
\pgfpathquadraticcurveto{\pgfqpoint{2.268591in}{1.001194in}}{\pgfqpoint{2.268591in}{1.056749in}}%
\pgfpathlineto{\pgfqpoint{2.268591in}{2.191056in}}%
\pgfpathquadraticcurveto{\pgfqpoint{2.268591in}{2.246612in}}{\pgfqpoint{2.213036in}{2.246612in}}%
\pgfpathlineto{\pgfqpoint{0.869398in}{2.246612in}}%
\pgfpathquadraticcurveto{\pgfqpoint{0.813843in}{2.246612in}}{\pgfqpoint{0.813843in}{2.191056in}}%
\pgfpathlineto{\pgfqpoint{0.813843in}{1.056749in}}%
\pgfpathquadraticcurveto{\pgfqpoint{0.813843in}{1.001194in}}{\pgfqpoint{0.869398in}{1.001194in}}%
\pgfpathlineto{\pgfqpoint{0.869398in}{1.001194in}}%
\pgfpathclose%
\pgfusepath{stroke,fill}%
\end{pgfscope}%
\begin{pgfscope}%
\pgfsetrectcap%
\pgfsetroundjoin%
\pgfsetlinewidth{2.509375pt}%
\definecolor{currentstroke}{rgb}{0.050980,0.415686,0.509804}%
\pgfsetstrokecolor{currentstroke}%
\pgfsetdash{}{0pt}%
\pgfpathmoveto{\pgfqpoint{0.924954in}{2.038278in}}%
\pgfpathlineto{\pgfqpoint{1.202732in}{2.038278in}}%
\pgfpathlineto{\pgfqpoint{1.480509in}{2.038278in}}%
\pgfusepath{stroke}%
\end{pgfscope}%
\begin{pgfscope}%
\pgfsetbuttcap%
\pgfsetroundjoin%
\definecolor{currentfill}{rgb}{0.050980,0.415686,0.509804}%
\pgfsetfillcolor{currentfill}%
\pgfsetlinewidth{1.003750pt}%
\definecolor{currentstroke}{rgb}{0.050980,0.415686,0.509804}%
\pgfsetstrokecolor{currentstroke}%
\pgfsetdash{}{0pt}%
\pgfsys@defobject{currentmarker}{\pgfqpoint{-0.055556in}{-0.055556in}}{\pgfqpoint{0.055556in}{0.055556in}}{%
\pgfpathmoveto{\pgfqpoint{0.000000in}{-0.055556in}}%
\pgfpathcurveto{\pgfqpoint{0.014734in}{-0.055556in}}{\pgfqpoint{0.028866in}{-0.049702in}}{\pgfqpoint{0.039284in}{-0.039284in}}%
\pgfpathcurveto{\pgfqpoint{0.049702in}{-0.028866in}}{\pgfqpoint{0.055556in}{-0.014734in}}{\pgfqpoint{0.055556in}{0.000000in}}%
\pgfpathcurveto{\pgfqpoint{0.055556in}{0.014734in}}{\pgfqpoint{0.049702in}{0.028866in}}{\pgfqpoint{0.039284in}{0.039284in}}%
\pgfpathcurveto{\pgfqpoint{0.028866in}{0.049702in}}{\pgfqpoint{0.014734in}{0.055556in}}{\pgfqpoint{0.000000in}{0.055556in}}%
\pgfpathcurveto{\pgfqpoint{-0.014734in}{0.055556in}}{\pgfqpoint{-0.028866in}{0.049702in}}{\pgfqpoint{-0.039284in}{0.039284in}}%
\pgfpathcurveto{\pgfqpoint{-0.049702in}{0.028866in}}{\pgfqpoint{-0.055556in}{0.014734in}}{\pgfqpoint{-0.055556in}{0.000000in}}%
\pgfpathcurveto{\pgfqpoint{-0.055556in}{-0.014734in}}{\pgfqpoint{-0.049702in}{-0.028866in}}{\pgfqpoint{-0.039284in}{-0.039284in}}%
\pgfpathcurveto{\pgfqpoint{-0.028866in}{-0.049702in}}{\pgfqpoint{-0.014734in}{-0.055556in}}{\pgfqpoint{0.000000in}{-0.055556in}}%
\pgfpathlineto{\pgfqpoint{0.000000in}{-0.055556in}}%
\pgfpathclose%
\pgfusepath{stroke,fill}%
}%
\begin{pgfscope}%
\pgfsys@transformshift{1.202732in}{2.038278in}%
\pgfsys@useobject{currentmarker}{}%
\end{pgfscope}%
\end{pgfscope}%
\begin{pgfscope}%
\definecolor{textcolor}{rgb}{0.000000,0.000000,0.000000}%
\pgfsetstrokecolor{textcolor}%
\pgfsetfillcolor{textcolor}%
\pgftext[x=1.702732in,y=1.941056in,left,base]{\color{textcolor}{\rmfamily\fontsize{20.000000}{24.000000}\selectfont\catcode`\^=\active\def^{\ifmmode\sp\else\^{}\fi}\catcode`\%=\active\def
\end{pgfscope}%
\begin{pgfscope}%
\pgfsetbuttcap%
\pgfsetroundjoin%
\pgfsetlinewidth{2.509375pt}%
\definecolor{currentstroke}{rgb}{0.960784,0.462745,0.000000}%
\pgfsetstrokecolor{currentstroke}%
\pgfsetdash{{9.250000pt}{4.000000pt}}{0.000000pt}%
\pgfpathmoveto{\pgfqpoint{0.924954in}{1.650917in}}%
\pgfpathlineto{\pgfqpoint{1.202732in}{1.650917in}}%
\pgfpathlineto{\pgfqpoint{1.480509in}{1.650917in}}%
\pgfusepath{stroke}%
\end{pgfscope}%
\begin{pgfscope}%
\pgfsetbuttcap%
\pgfsetmiterjoin%
\definecolor{currentfill}{rgb}{0.960784,0.462745,0.000000}%
\pgfsetfillcolor{currentfill}%
\pgfsetlinewidth{1.003750pt}%
\definecolor{currentstroke}{rgb}{0.960784,0.462745,0.000000}%
\pgfsetstrokecolor{currentstroke}%
\pgfsetdash{}{0pt}%
\pgfsys@defobject{currentmarker}{\pgfqpoint{-0.055556in}{-0.055556in}}{\pgfqpoint{0.055556in}{0.055556in}}{%
\pgfpathmoveto{\pgfqpoint{-0.055556in}{-0.055556in}}%
\pgfpathlineto{\pgfqpoint{0.055556in}{-0.055556in}}%
\pgfpathlineto{\pgfqpoint{0.055556in}{0.055556in}}%
\pgfpathlineto{\pgfqpoint{-0.055556in}{0.055556in}}%
\pgfpathlineto{\pgfqpoint{-0.055556in}{-0.055556in}}%
\pgfpathclose%
\pgfusepath{stroke,fill}%
}%
\begin{pgfscope}%
\pgfsys@transformshift{1.202732in}{1.650917in}%
\pgfsys@useobject{currentmarker}{}%
\end{pgfscope}%
\end{pgfscope}%
\begin{pgfscope}%
\definecolor{textcolor}{rgb}{0.000000,0.000000,0.000000}%
\pgfsetstrokecolor{textcolor}%
\pgfsetfillcolor{textcolor}%
\pgftext[x=1.702732in,y=1.553694in,left,base]{\color{textcolor}{\rmfamily\fontsize{20.000000}{24.000000}\selectfont\catcode`\^=\active\def^{\ifmmode\sp\else\^{}\fi}\catcode`\%=\active\def
\end{pgfscope}%
\begin{pgfscope}%
\pgfsetbuttcap%
\pgfsetroundjoin%
\pgfsetlinewidth{2.509375pt}%
\definecolor{currentstroke}{rgb}{0.219608,0.219608,0.219608}%
\pgfsetstrokecolor{currentstroke}%
\pgfsetdash{{2.500000pt}{4.125000pt}}{0.000000pt}%
\pgfpathmoveto{\pgfqpoint{0.924954in}{1.263555in}}%
\pgfpathlineto{\pgfqpoint{1.202732in}{1.263555in}}%
\pgfpathlineto{\pgfqpoint{1.480509in}{1.263555in}}%
\pgfusepath{stroke}%
\end{pgfscope}%
\begin{pgfscope}%
\pgfsetbuttcap%
\pgfsetmiterjoin%
\definecolor{currentfill}{rgb}{0.219608,0.219608,0.219608}%
\pgfsetfillcolor{currentfill}%
\pgfsetlinewidth{1.003750pt}%
\definecolor{currentstroke}{rgb}{0.219608,0.219608,0.219608}%
\pgfsetstrokecolor{currentstroke}%
\pgfsetdash{}{0pt}%
\pgfsys@defobject{currentmarker}{\pgfqpoint{-0.055556in}{-0.055556in}}{\pgfqpoint{0.055556in}{0.055556in}}{%
\pgfpathmoveto{\pgfqpoint{0.000000in}{0.055556in}}%
\pgfpathlineto{\pgfqpoint{-0.055556in}{-0.055556in}}%
\pgfpathlineto{\pgfqpoint{0.055556in}{-0.055556in}}%
\pgfpathlineto{\pgfqpoint{0.000000in}{0.055556in}}%
\pgfpathclose%
\pgfusepath{stroke,fill}%
}%
\begin{pgfscope}%
\pgfsys@transformshift{1.202732in}{1.263555in}%
\pgfsys@useobject{currentmarker}{}%
\end{pgfscope}%
\end{pgfscope}%
\begin{pgfscope}%
\definecolor{textcolor}{rgb}{0.000000,0.000000,0.000000}%
\pgfsetstrokecolor{textcolor}%
\pgfsetfillcolor{textcolor}%
\pgftext[x=1.702732in,y=1.166333in,left,base]{\color{textcolor}{\rmfamily\fontsize{20.000000}{24.000000}\selectfont\catcode`\^=\active\def^{\ifmmode\sp\else\^{}\fi}\catcode`\%=\active\def
\end{pgfscope}%
\end{pgfpicture}%
\makeatother%
\endgroup%

%% file: images/lora_r_study/r_results_RTA.pgf
\begingroup%
\makeatletter%
\begin{pgfpicture}%
\pgfpathrectangle{\pgfpointorigin}{\pgfqpoint{7.450000in}{5.450000in}}%
\pgfusepath{use as bounding box, clip}%
\begin{pgfscope}%
\pgfsetbuttcap%
\pgfsetmiterjoin%
\definecolor{currentfill}{rgb}{1.000000,1.000000,1.000000}%
\pgfsetfillcolor{currentfill}%
\pgfsetlinewidth{0.000000pt}%
\definecolor{currentstroke}{rgb}{1.000000,1.000000,1.000000}%
\pgfsetstrokecolor{currentstroke}%
\pgfsetdash{}{0pt}%
\pgfpathmoveto{\pgfqpoint{0.000000in}{0.000000in}}%
\pgfpathlineto{\pgfqpoint{7.450000in}{0.000000in}}%
\pgfpathlineto{\pgfqpoint{7.450000in}{5.450000in}}%
\pgfpathlineto{\pgfqpoint{0.000000in}{5.450000in}}%
\pgfpathlineto{\pgfqpoint{0.000000in}{0.000000in}}%
\pgfpathclose%
\pgfusepath{fill}%
\end{pgfscope}%
\begin{pgfscope}%
\pgfsetbuttcap%
\pgfsetmiterjoin%
\definecolor{currentfill}{rgb}{1.000000,1.000000,1.000000}%
\pgfsetfillcolor{currentfill}%
\pgfsetlinewidth{0.000000pt}%
\definecolor{currentstroke}{rgb}{0.000000,0.000000,0.000000}%
\pgfsetstrokecolor{currentstroke}%
\pgfsetstrokeopacity{0.000000}%
\pgfsetdash{}{0pt}%
\pgfpathmoveto{\pgfqpoint{0.674954in}{0.862305in}}%
\pgfpathlineto{\pgfqpoint{7.190756in}{0.862305in}}%
\pgfpathlineto{\pgfqpoint{7.190756in}{5.231386in}}%
\pgfpathlineto{\pgfqpoint{0.674954in}{5.231386in}}%
\pgfpathlineto{\pgfqpoint{0.674954in}{0.862305in}}%
\pgfpathclose%
\pgfusepath{fill}%
\end{pgfscope}%
\begin{pgfscope}%
\pgfpathrectangle{\pgfqpoint{0.674954in}{0.862305in}}{\pgfqpoint{6.515802in}{4.369081in}}%
\pgfusepath{clip}%
\pgfsetbuttcap%
\pgfsetroundjoin%
\definecolor{currentfill}{rgb}{0.050980,0.415686,0.509804}%
\pgfsetfillcolor{currentfill}%
\pgfsetfillopacity{0.300000}%
\pgfsetlinewidth{1.003750pt}%
\definecolor{currentstroke}{rgb}{0.050980,0.415686,0.509804}%
\pgfsetstrokecolor{currentstroke}%
\pgfsetstrokeopacity{0.300000}%
\pgfsetdash{}{0pt}%
\pgfsys@defobject{currentmarker}{\pgfqpoint{0.674954in}{4.330237in}}{\pgfqpoint{7.190756in}{4.884136in}}{%
\pgfpathmoveto{\pgfqpoint{0.674954in}{4.577086in}}%
\pgfpathlineto{\pgfqpoint{0.674954in}{4.330237in}}%
\pgfpathlineto{\pgfqpoint{1.760921in}{4.481808in}}%
\pgfpathlineto{\pgfqpoint{2.846888in}{4.408156in}}%
\pgfpathlineto{\pgfqpoint{3.932855in}{4.415723in}}%
\pgfpathlineto{\pgfqpoint{5.018822in}{4.618334in}}%
\pgfpathlineto{\pgfqpoint{6.104789in}{4.508184in}}%
\pgfpathlineto{\pgfqpoint{6.740039in}{4.553878in}}%
\pgfpathlineto{\pgfqpoint{7.190756in}{4.570476in}}%
\pgfpathlineto{\pgfqpoint{7.190756in}{4.884136in}}%
\pgfpathlineto{\pgfqpoint{7.190756in}{4.884136in}}%
\pgfpathlineto{\pgfqpoint{6.740039in}{4.792236in}}%
\pgfpathlineto{\pgfqpoint{6.104789in}{4.757714in}}%
\pgfpathlineto{\pgfqpoint{5.018822in}{4.845521in}}%
\pgfpathlineto{\pgfqpoint{3.932855in}{4.653351in}}%
\pgfpathlineto{\pgfqpoint{2.846888in}{4.675463in}}%
\pgfpathlineto{\pgfqpoint{1.760921in}{4.712232in}}%
\pgfpathlineto{\pgfqpoint{0.674954in}{4.577086in}}%
\pgfpathlineto{\pgfqpoint{0.674954in}{4.577086in}}%
\pgfpathclose%
\pgfusepath{stroke,fill}%
}%
\begin{pgfscope}%
\pgfsys@transformshift{0.000000in}{0.000000in}%
\pgfsys@useobject{currentmarker}{}%
\end{pgfscope}%
\end{pgfscope}%
\begin{pgfscope}%
\pgfpathrectangle{\pgfqpoint{0.674954in}{0.862305in}}{\pgfqpoint{6.515802in}{4.369081in}}%
\pgfusepath{clip}%
\pgfsetbuttcap%
\pgfsetroundjoin%
\definecolor{currentfill}{rgb}{0.960784,0.462745,0.000000}%
\pgfsetfillcolor{currentfill}%
\pgfsetfillopacity{0.300000}%
\pgfsetlinewidth{1.003750pt}%
\definecolor{currentstroke}{rgb}{0.960784,0.462745,0.000000}%
\pgfsetstrokecolor{currentstroke}%
\pgfsetstrokeopacity{0.300000}%
\pgfsetdash{{3.700000pt}{1.600000pt}}{0.000000pt}%
\pgfpathmoveto{\pgfqpoint{0.674954in}{4.244345in}}%
\pgfpathlineto{\pgfqpoint{0.674954in}{3.921243in}}%
\pgfpathlineto{\pgfqpoint{1.760921in}{4.048756in}}%
\pgfpathlineto{\pgfqpoint{2.846888in}{3.956767in}}%
\pgfpathlineto{\pgfqpoint{3.932855in}{3.926558in}}%
\pgfpathlineto{\pgfqpoint{5.018822in}{4.216343in}}%
\pgfpathlineto{\pgfqpoint{6.104789in}{4.096354in}}%
\pgfpathlineto{\pgfqpoint{6.740039in}{4.219241in}}%
\pgfpathlineto{\pgfqpoint{7.190756in}{4.152195in}}%
\pgfpathlineto{\pgfqpoint{7.190756in}{4.552225in}}%
\pgfpathlineto{\pgfqpoint{7.190756in}{4.552225in}}%
\pgfpathlineto{\pgfqpoint{6.740039in}{4.521474in}}%
\pgfpathlineto{\pgfqpoint{6.104789in}{4.420260in}}%
\pgfpathlineto{\pgfqpoint{5.018822in}{4.515577in}}%
\pgfpathlineto{\pgfqpoint{3.932855in}{4.210439in}}%
\pgfpathlineto{\pgfqpoint{2.846888in}{4.298353in}}%
\pgfpathlineto{\pgfqpoint{1.760921in}{4.342406in}}%
\pgfpathlineto{\pgfqpoint{0.674954in}{4.244345in}}%
\pgfpathlineto{\pgfqpoint{0.674954in}{4.244345in}}%
\pgfpathclose%
\pgfusepath{stroke,fill}%
\end{pgfscope}%
\begin{pgfscope}%
\pgfpathrectangle{\pgfqpoint{0.674954in}{0.862305in}}{\pgfqpoint{6.515802in}{4.369081in}}%
\pgfusepath{clip}%
\pgfsetbuttcap%
\pgfsetroundjoin%
\definecolor{currentfill}{rgb}{0.219608,0.219608,0.219608}%
\pgfsetfillcolor{currentfill}%
\pgfsetfillopacity{0.300000}%
\pgfsetlinewidth{1.003750pt}%
\definecolor{currentstroke}{rgb}{0.219608,0.219608,0.219608}%
\pgfsetstrokecolor{currentstroke}%
\pgfsetstrokeopacity{0.300000}%
\pgfsetdash{{1.000000pt}{1.650000pt}}{0.000000pt}%
\pgfpathmoveto{\pgfqpoint{0.674954in}{3.388199in}}%
\pgfpathlineto{\pgfqpoint{0.674954in}{3.044553in}}%
\pgfpathlineto{\pgfqpoint{1.760921in}{3.125563in}}%
\pgfpathlineto{\pgfqpoint{2.846888in}{3.098142in}}%
\pgfpathlineto{\pgfqpoint{3.932855in}{2.952649in}}%
\pgfpathlineto{\pgfqpoint{5.018822in}{3.422165in}}%
\pgfpathlineto{\pgfqpoint{6.104789in}{3.205126in}}%
\pgfpathlineto{\pgfqpoint{6.740039in}{3.384779in}}%
\pgfpathlineto{\pgfqpoint{7.190756in}{3.348569in}}%
\pgfpathlineto{\pgfqpoint{7.190756in}{3.806573in}}%
\pgfpathlineto{\pgfqpoint{7.190756in}{3.806573in}}%
\pgfpathlineto{\pgfqpoint{6.740039in}{3.729915in}}%
\pgfpathlineto{\pgfqpoint{6.104789in}{3.560748in}}%
\pgfpathlineto{\pgfqpoint{5.018822in}{3.812901in}}%
\pgfpathlineto{\pgfqpoint{3.932855in}{3.296864in}}%
\pgfpathlineto{\pgfqpoint{2.846888in}{3.508253in}}%
\pgfpathlineto{\pgfqpoint{1.760921in}{3.449759in}}%
\pgfpathlineto{\pgfqpoint{0.674954in}{3.388199in}}%
\pgfpathlineto{\pgfqpoint{0.674954in}{3.388199in}}%
\pgfpathclose%
\pgfusepath{stroke,fill}%
\end{pgfscope}%
\begin{pgfscope}%
\pgfpathrectangle{\pgfqpoint{0.674954in}{0.862305in}}{\pgfqpoint{6.515802in}{4.369081in}}%
\pgfusepath{clip}%
\pgfsetbuttcap%
\pgfsetroundjoin%
\pgfsetlinewidth{2.007500pt}%
\definecolor{currentstroke}{rgb}{0.501961,0.501961,0.501961}%
\pgfsetstrokecolor{currentstroke}%
\pgfsetstrokeopacity{0.300000}%
\pgfsetdash{{7.400000pt}{3.200000pt}}{0.000000pt}%
\pgfpathmoveto{\pgfqpoint{0.674954in}{0.862305in}}%
\pgfpathlineto{\pgfqpoint{0.674954in}{5.231386in}}%
\pgfusepath{stroke}%
\end{pgfscope}%
\begin{pgfscope}%
\pgfsetbuttcap%
\pgfsetroundjoin%
\definecolor{currentfill}{rgb}{0.000000,0.000000,0.000000}%
\pgfsetfillcolor{currentfill}%
\pgfsetlinewidth{0.803000pt}%
\definecolor{currentstroke}{rgb}{0.000000,0.000000,0.000000}%
\pgfsetstrokecolor{currentstroke}%
\pgfsetdash{}{0pt}%
\pgfsys@defobject{currentmarker}{\pgfqpoint{0.000000in}{-0.048611in}}{\pgfqpoint{0.000000in}{0.000000in}}{%
\pgfpathmoveto{\pgfqpoint{0.000000in}{0.000000in}}%
\pgfpathlineto{\pgfqpoint{0.000000in}{-0.048611in}}%
\pgfusepath{stroke,fill}%
}%
\begin{pgfscope}%
\pgfsys@transformshift{0.674954in}{0.862305in}%
\pgfsys@useobject{currentmarker}{}%
\end{pgfscope}%
\end{pgfscope}%
\begin{pgfscope}%
\definecolor{textcolor}{rgb}{0.000000,0.000000,0.000000}%
\pgfsetstrokecolor{textcolor}%
\pgfsetfillcolor{textcolor}%
\pgftext[x=0.674954in,y=0.765082in,,top]{\color{textcolor}{\rmfamily\fontsize{25.000000}{30.000000}\selectfont\catcode`\^=\active\def^{\ifmmode\sp\else\^{}\fi}\catcode`\%=\active\def
\end{pgfscope}%
\begin{pgfscope}%
\pgfpathrectangle{\pgfqpoint{0.674954in}{0.862305in}}{\pgfqpoint{6.515802in}{4.369081in}}%
\pgfusepath{clip}%
\pgfsetbuttcap%
\pgfsetroundjoin%
\pgfsetlinewidth{2.007500pt}%
\definecolor{currentstroke}{rgb}{0.501961,0.501961,0.501961}%
\pgfsetstrokecolor{currentstroke}%
\pgfsetstrokeopacity{0.300000}%
\pgfsetdash{{7.400000pt}{3.200000pt}}{0.000000pt}%
\pgfpathmoveto{\pgfqpoint{1.760921in}{0.862305in}}%
\pgfpathlineto{\pgfqpoint{1.760921in}{5.231386in}}%
\pgfusepath{stroke}%
\end{pgfscope}%
\begin{pgfscope}%
\pgfsetbuttcap%
\pgfsetroundjoin%
\definecolor{currentfill}{rgb}{0.000000,0.000000,0.000000}%
\pgfsetfillcolor{currentfill}%
\pgfsetlinewidth{0.803000pt}%
\definecolor{currentstroke}{rgb}{0.000000,0.000000,0.000000}%
\pgfsetstrokecolor{currentstroke}%
\pgfsetdash{}{0pt}%
\pgfsys@defobject{currentmarker}{\pgfqpoint{0.000000in}{-0.048611in}}{\pgfqpoint{0.000000in}{0.000000in}}{%
\pgfpathmoveto{\pgfqpoint{0.000000in}{0.000000in}}%
\pgfpathlineto{\pgfqpoint{0.000000in}{-0.048611in}}%
\pgfusepath{stroke,fill}%
}%
\begin{pgfscope}%
\pgfsys@transformshift{1.760921in}{0.862305in}%
\pgfsys@useobject{currentmarker}{}%
\end{pgfscope}%
\end{pgfscope}%
\begin{pgfscope}%
\definecolor{textcolor}{rgb}{0.000000,0.000000,0.000000}%
\pgfsetstrokecolor{textcolor}%
\pgfsetfillcolor{textcolor}%
\pgftext[x=1.760921in,y=0.765082in,,top]{\color{textcolor}{\rmfamily\fontsize{25.000000}{30.000000}\selectfont\catcode`\^=\active\def^{\ifmmode\sp\else\^{}\fi}\catcode`\%=\active\def
\end{pgfscope}%
\begin{pgfscope}%
\pgfpathrectangle{\pgfqpoint{0.674954in}{0.862305in}}{\pgfqpoint{6.515802in}{4.369081in}}%
\pgfusepath{clip}%
\pgfsetbuttcap%
\pgfsetroundjoin%
\pgfsetlinewidth{2.007500pt}%
\definecolor{currentstroke}{rgb}{0.501961,0.501961,0.501961}%
\pgfsetstrokecolor{currentstroke}%
\pgfsetstrokeopacity{0.300000}%
\pgfsetdash{{7.400000pt}{3.200000pt}}{0.000000pt}%
\pgfpathmoveto{\pgfqpoint{2.846888in}{0.862305in}}%
\pgfpathlineto{\pgfqpoint{2.846888in}{5.231386in}}%
\pgfusepath{stroke}%
\end{pgfscope}%
\begin{pgfscope}%
\pgfsetbuttcap%
\pgfsetroundjoin%
\definecolor{currentfill}{rgb}{0.000000,0.000000,0.000000}%
\pgfsetfillcolor{currentfill}%
\pgfsetlinewidth{0.803000pt}%
\definecolor{currentstroke}{rgb}{0.000000,0.000000,0.000000}%
\pgfsetstrokecolor{currentstroke}%
\pgfsetdash{}{0pt}%
\pgfsys@defobject{currentmarker}{\pgfqpoint{0.000000in}{-0.048611in}}{\pgfqpoint{0.000000in}{0.000000in}}{%
\pgfpathmoveto{\pgfqpoint{0.000000in}{0.000000in}}%
\pgfpathlineto{\pgfqpoint{0.000000in}{-0.048611in}}%
\pgfusepath{stroke,fill}%
}%
\begin{pgfscope}%
\pgfsys@transformshift{2.846888in}{0.862305in}%
\pgfsys@useobject{currentmarker}{}%
\end{pgfscope}%
\end{pgfscope}%
\begin{pgfscope}%
\definecolor{textcolor}{rgb}{0.000000,0.000000,0.000000}%
\pgfsetstrokecolor{textcolor}%
\pgfsetfillcolor{textcolor}%
\pgftext[x=2.846888in,y=0.765082in,,top]{\color{textcolor}{\rmfamily\fontsize{25.000000}{30.000000}\selectfont\catcode`\^=\active\def^{\ifmmode\sp\else\^{}\fi}\catcode`\%=\active\def
\end{pgfscope}%
\begin{pgfscope}%
\pgfpathrectangle{\pgfqpoint{0.674954in}{0.862305in}}{\pgfqpoint{6.515802in}{4.369081in}}%
\pgfusepath{clip}%
\pgfsetbuttcap%
\pgfsetroundjoin%
\pgfsetlinewidth{2.007500pt}%
\definecolor{currentstroke}{rgb}{0.501961,0.501961,0.501961}%
\pgfsetstrokecolor{currentstroke}%
\pgfsetstrokeopacity{0.300000}%
\pgfsetdash{{7.400000pt}{3.200000pt}}{0.000000pt}%
\pgfpathmoveto{\pgfqpoint{3.932855in}{0.862305in}}%
\pgfpathlineto{\pgfqpoint{3.932855in}{5.231386in}}%
\pgfusepath{stroke}%
\end{pgfscope}%
\begin{pgfscope}%
\pgfsetbuttcap%
\pgfsetroundjoin%
\definecolor{currentfill}{rgb}{0.000000,0.000000,0.000000}%
\pgfsetfillcolor{currentfill}%
\pgfsetlinewidth{0.803000pt}%
\definecolor{currentstroke}{rgb}{0.000000,0.000000,0.000000}%
\pgfsetstrokecolor{currentstroke}%
\pgfsetdash{}{0pt}%
\pgfsys@defobject{currentmarker}{\pgfqpoint{0.000000in}{-0.048611in}}{\pgfqpoint{0.000000in}{0.000000in}}{%
\pgfpathmoveto{\pgfqpoint{0.000000in}{0.000000in}}%
\pgfpathlineto{\pgfqpoint{0.000000in}{-0.048611in}}%
\pgfusepath{stroke,fill}%
}%
\begin{pgfscope}%
\pgfsys@transformshift{3.932855in}{0.862305in}%
\pgfsys@useobject{currentmarker}{}%
\end{pgfscope}%
\end{pgfscope}%
\begin{pgfscope}%
\definecolor{textcolor}{rgb}{0.000000,0.000000,0.000000}%
\pgfsetstrokecolor{textcolor}%
\pgfsetfillcolor{textcolor}%
\pgftext[x=3.932855in,y=0.765082in,,top]{\color{textcolor}{\rmfamily\fontsize{25.000000}{30.000000}\selectfont\catcode`\^=\active\def^{\ifmmode\sp\else\^{}\fi}\catcode`\%=\active\def
\end{pgfscope}%
\begin{pgfscope}%
\pgfpathrectangle{\pgfqpoint{0.674954in}{0.862305in}}{\pgfqpoint{6.515802in}{4.369081in}}%
\pgfusepath{clip}%
\pgfsetbuttcap%
\pgfsetroundjoin%
\pgfsetlinewidth{2.007500pt}%
\definecolor{currentstroke}{rgb}{0.501961,0.501961,0.501961}%
\pgfsetstrokecolor{currentstroke}%
\pgfsetstrokeopacity{0.300000}%
\pgfsetdash{{7.400000pt}{3.200000pt}}{0.000000pt}%
\pgfpathmoveto{\pgfqpoint{5.018822in}{0.862305in}}%
\pgfpathlineto{\pgfqpoint{5.018822in}{5.231386in}}%
\pgfusepath{stroke}%
\end{pgfscope}%
\begin{pgfscope}%
\pgfsetbuttcap%
\pgfsetroundjoin%
\definecolor{currentfill}{rgb}{0.000000,0.000000,0.000000}%
\pgfsetfillcolor{currentfill}%
\pgfsetlinewidth{0.803000pt}%
\definecolor{currentstroke}{rgb}{0.000000,0.000000,0.000000}%
\pgfsetstrokecolor{currentstroke}%
\pgfsetdash{}{0pt}%
\pgfsys@defobject{currentmarker}{\pgfqpoint{0.000000in}{-0.048611in}}{\pgfqpoint{0.000000in}{0.000000in}}{%
\pgfpathmoveto{\pgfqpoint{0.000000in}{0.000000in}}%
\pgfpathlineto{\pgfqpoint{0.000000in}{-0.048611in}}%
\pgfusepath{stroke,fill}%
}%
\begin{pgfscope}%
\pgfsys@transformshift{5.018822in}{0.862305in}%
\pgfsys@useobject{currentmarker}{}%
\end{pgfscope}%
\end{pgfscope}%
\begin{pgfscope}%
\definecolor{textcolor}{rgb}{0.000000,0.000000,0.000000}%
\pgfsetstrokecolor{textcolor}%
\pgfsetfillcolor{textcolor}%
\pgftext[x=5.018822in,y=0.765082in,,top]{\color{textcolor}{\rmfamily\fontsize{25.000000}{30.000000}\selectfont\catcode`\^=\active\def^{\ifmmode\sp\else\^{}\fi}\catcode`\%=\active\def
\end{pgfscope}%
\begin{pgfscope}%
\pgfpathrectangle{\pgfqpoint{0.674954in}{0.862305in}}{\pgfqpoint{6.515802in}{4.369081in}}%
\pgfusepath{clip}%
\pgfsetbuttcap%
\pgfsetroundjoin%
\pgfsetlinewidth{2.007500pt}%
\definecolor{currentstroke}{rgb}{0.501961,0.501961,0.501961}%
\pgfsetstrokecolor{currentstroke}%
\pgfsetstrokeopacity{0.300000}%
\pgfsetdash{{7.400000pt}{3.200000pt}}{0.000000pt}%
\pgfpathmoveto{\pgfqpoint{6.104789in}{0.862305in}}%
\pgfpathlineto{\pgfqpoint{6.104789in}{5.231386in}}%
\pgfusepath{stroke}%
\end{pgfscope}%
\begin{pgfscope}%
\pgfsetbuttcap%
\pgfsetroundjoin%
\definecolor{currentfill}{rgb}{0.000000,0.000000,0.000000}%
\pgfsetfillcolor{currentfill}%
\pgfsetlinewidth{0.803000pt}%
\definecolor{currentstroke}{rgb}{0.000000,0.000000,0.000000}%
\pgfsetstrokecolor{currentstroke}%
\pgfsetdash{}{0pt}%
\pgfsys@defobject{currentmarker}{\pgfqpoint{0.000000in}{-0.048611in}}{\pgfqpoint{0.000000in}{0.000000in}}{%
\pgfpathmoveto{\pgfqpoint{0.000000in}{0.000000in}}%
\pgfpathlineto{\pgfqpoint{0.000000in}{-0.048611in}}%
\pgfusepath{stroke,fill}%
}%
\begin{pgfscope}%
\pgfsys@transformshift{6.104789in}{0.862305in}%
\pgfsys@useobject{currentmarker}{}%
\end{pgfscope}%
\end{pgfscope}%
\begin{pgfscope}%
\definecolor{textcolor}{rgb}{0.000000,0.000000,0.000000}%
\pgfsetstrokecolor{textcolor}%
\pgfsetfillcolor{textcolor}%
\pgftext[x=6.104789in,y=0.765082in,,top]{\color{textcolor}{\rmfamily\fontsize{25.000000}{30.000000}\selectfont\catcode`\^=\active\def^{\ifmmode\sp\else\^{}\fi}\catcode`\%=\active\def
\end{pgfscope}%
\begin{pgfscope}%
\pgfpathrectangle{\pgfqpoint{0.674954in}{0.862305in}}{\pgfqpoint{6.515802in}{4.369081in}}%
\pgfusepath{clip}%
\pgfsetbuttcap%
\pgfsetroundjoin%
\pgfsetlinewidth{2.007500pt}%
\definecolor{currentstroke}{rgb}{0.501961,0.501961,0.501961}%
\pgfsetstrokecolor{currentstroke}%
\pgfsetstrokeopacity{0.300000}%
\pgfsetdash{{7.400000pt}{3.200000pt}}{0.000000pt}%
\pgfpathmoveto{\pgfqpoint{6.740039in}{0.862305in}}%
\pgfpathlineto{\pgfqpoint{6.740039in}{5.231386in}}%
\pgfusepath{stroke}%
\end{pgfscope}%
\begin{pgfscope}%
\pgfsetbuttcap%
\pgfsetroundjoin%
\definecolor{currentfill}{rgb}{0.000000,0.000000,0.000000}%
\pgfsetfillcolor{currentfill}%
\pgfsetlinewidth{0.803000pt}%
\definecolor{currentstroke}{rgb}{0.000000,0.000000,0.000000}%
\pgfsetstrokecolor{currentstroke}%
\pgfsetdash{}{0pt}%
\pgfsys@defobject{currentmarker}{\pgfqpoint{0.000000in}{-0.048611in}}{\pgfqpoint{0.000000in}{0.000000in}}{%
\pgfpathmoveto{\pgfqpoint{0.000000in}{0.000000in}}%
\pgfpathlineto{\pgfqpoint{0.000000in}{-0.048611in}}%
\pgfusepath{stroke,fill}%
}%
\begin{pgfscope}%
\pgfsys@transformshift{6.740039in}{0.862305in}%
\pgfsys@useobject{currentmarker}{}%
\end{pgfscope}%
\end{pgfscope}%
\begin{pgfscope}%
\definecolor{textcolor}{rgb}{0.000000,0.000000,0.000000}%
\pgfsetstrokecolor{textcolor}%
\pgfsetfillcolor{textcolor}%
\pgftext[x=6.740039in,y=0.765082in,,top]{\color{textcolor}{\rmfamily\fontsize{25.000000}{30.000000}\selectfont\catcode`\^=\active\def^{\ifmmode\sp\else\^{}\fi}\catcode`\%=\active\def
\end{pgfscope}%
\begin{pgfscope}%
\pgfpathrectangle{\pgfqpoint{0.674954in}{0.862305in}}{\pgfqpoint{6.515802in}{4.369081in}}%
\pgfusepath{clip}%
\pgfsetbuttcap%
\pgfsetroundjoin%
\pgfsetlinewidth{2.007500pt}%
\definecolor{currentstroke}{rgb}{0.501961,0.501961,0.501961}%
\pgfsetstrokecolor{currentstroke}%
\pgfsetstrokeopacity{0.300000}%
\pgfsetdash{{7.400000pt}{3.200000pt}}{0.000000pt}%
\pgfpathmoveto{\pgfqpoint{7.190756in}{0.862305in}}%
\pgfpathlineto{\pgfqpoint{7.190756in}{5.231386in}}%
\pgfusepath{stroke}%
\end{pgfscope}%
\begin{pgfscope}%
\pgfsetbuttcap%
\pgfsetroundjoin%
\definecolor{currentfill}{rgb}{0.000000,0.000000,0.000000}%
\pgfsetfillcolor{currentfill}%
\pgfsetlinewidth{0.803000pt}%
\definecolor{currentstroke}{rgb}{0.000000,0.000000,0.000000}%
\pgfsetstrokecolor{currentstroke}%
\pgfsetdash{}{0pt}%
\pgfsys@defobject{currentmarker}{\pgfqpoint{0.000000in}{-0.048611in}}{\pgfqpoint{0.000000in}{0.000000in}}{%
\pgfpathmoveto{\pgfqpoint{0.000000in}{0.000000in}}%
\pgfpathlineto{\pgfqpoint{0.000000in}{-0.048611in}}%
\pgfusepath{stroke,fill}%
}%
\begin{pgfscope}%
\pgfsys@transformshift{7.190756in}{0.862305in}%
\pgfsys@useobject{currentmarker}{}%
\end{pgfscope}%
\end{pgfscope}%
\begin{pgfscope}%
\definecolor{textcolor}{rgb}{0.000000,0.000000,0.000000}%
\pgfsetstrokecolor{textcolor}%
\pgfsetfillcolor{textcolor}%
\pgftext[x=7.190756in,y=0.765082in,,top]{\color{textcolor}{\rmfamily\fontsize{25.000000}{30.000000}\selectfont\catcode`\^=\active\def^{\ifmmode\sp\else\^{}\fi}\catcode`\%=\active\def
\end{pgfscope}%
\begin{pgfscope}%
\definecolor{textcolor}{rgb}{0.000000,0.000000,0.000000}%
\pgfsetstrokecolor{textcolor}%
\pgfsetfillcolor{textcolor}%
\pgftext[x=3.932855in,y=0.404763in,,top]{\color{textcolor}{\rmfamily\fontsize{25.000000}{30.000000}\selectfont\catcode`\^=\active\def^{\ifmmode\sp\else\^{}\fi}\catcode`\%=\active\def
\end{pgfscope}%
\begin{pgfscope}%
\pgfpathrectangle{\pgfqpoint{0.674954in}{0.862305in}}{\pgfqpoint{6.515802in}{4.369081in}}%
\pgfusepath{clip}%
\pgfsetbuttcap%
\pgfsetroundjoin%
\pgfsetlinewidth{2.007500pt}%
\definecolor{currentstroke}{rgb}{0.501961,0.501961,0.501961}%
\pgfsetstrokecolor{currentstroke}%
\pgfsetstrokeopacity{0.300000}%
\pgfsetdash{{7.400000pt}{3.200000pt}}{0.000000pt}%
\pgfpathmoveto{\pgfqpoint{0.674954in}{1.954575in}}%
\pgfpathlineto{\pgfqpoint{7.190756in}{1.954575in}}%
\pgfusepath{stroke}%
\end{pgfscope}%
\begin{pgfscope}%
\pgfsetbuttcap%
\pgfsetroundjoin%
\definecolor{currentfill}{rgb}{0.000000,0.000000,0.000000}%
\pgfsetfillcolor{currentfill}%
\pgfsetlinewidth{0.803000pt}%
\definecolor{currentstroke}{rgb}{0.000000,0.000000,0.000000}%
\pgfsetstrokecolor{currentstroke}%
\pgfsetdash{}{0pt}%
\pgfsys@defobject{currentmarker}{\pgfqpoint{-0.048611in}{0.000000in}}{\pgfqpoint{-0.000000in}{0.000000in}}{%
\pgfpathmoveto{\pgfqpoint{-0.000000in}{0.000000in}}%
\pgfpathlineto{\pgfqpoint{-0.048611in}{0.000000in}}%
\pgfusepath{stroke,fill}%
}%
\begin{pgfscope}%
\pgfsys@transformshift{0.674954in}{1.954575in}%
\pgfsys@useobject{currentmarker}{}%
\end{pgfscope}%
\end{pgfscope}%
\begin{pgfscope}%
\definecolor{textcolor}{rgb}{0.000000,0.000000,0.000000}%
\pgfsetstrokecolor{textcolor}%
\pgfsetfillcolor{textcolor}%
\pgftext[x=0.259244in, y=1.835961in, left, base]{\color{textcolor}{\rmfamily\fontsize{25.000000}{30.000000}\selectfont\catcode`\^=\active\def^{\ifmmode\sp\else\^{}\fi}\catcode`\%=\active\def
\end{pgfscope}%
\begin{pgfscope}%
\pgfpathrectangle{\pgfqpoint{0.674954in}{0.862305in}}{\pgfqpoint{6.515802in}{4.369081in}}%
\pgfusepath{clip}%
\pgfsetbuttcap%
\pgfsetroundjoin%
\pgfsetlinewidth{2.007500pt}%
\definecolor{currentstroke}{rgb}{0.501961,0.501961,0.501961}%
\pgfsetstrokecolor{currentstroke}%
\pgfsetstrokeopacity{0.300000}%
\pgfsetdash{{7.400000pt}{3.200000pt}}{0.000000pt}%
\pgfpathmoveto{\pgfqpoint{0.674954in}{3.046845in}}%
\pgfpathlineto{\pgfqpoint{7.190756in}{3.046845in}}%
\pgfusepath{stroke}%
\end{pgfscope}%
\begin{pgfscope}%
\pgfsetbuttcap%
\pgfsetroundjoin%
\definecolor{currentfill}{rgb}{0.000000,0.000000,0.000000}%
\pgfsetfillcolor{currentfill}%
\pgfsetlinewidth{0.803000pt}%
\definecolor{currentstroke}{rgb}{0.000000,0.000000,0.000000}%
\pgfsetstrokecolor{currentstroke}%
\pgfsetdash{}{0pt}%
\pgfsys@defobject{currentmarker}{\pgfqpoint{-0.048611in}{0.000000in}}{\pgfqpoint{-0.000000in}{0.000000in}}{%
\pgfpathmoveto{\pgfqpoint{-0.000000in}{0.000000in}}%
\pgfpathlineto{\pgfqpoint{-0.048611in}{0.000000in}}%
\pgfusepath{stroke,fill}%
}%
\begin{pgfscope}%
\pgfsys@transformshift{0.674954in}{3.046845in}%
\pgfsys@useobject{currentmarker}{}%
\end{pgfscope}%
\end{pgfscope}%
\begin{pgfscope}%
\definecolor{textcolor}{rgb}{0.000000,0.000000,0.000000}%
\pgfsetstrokecolor{textcolor}%
\pgfsetfillcolor{textcolor}%
\pgftext[x=0.259244in, y=2.928231in, left, base]{\color{textcolor}{\rmfamily\fontsize{25.000000}{30.000000}\selectfont\catcode`\^=\active\def^{\ifmmode\sp\else\^{}\fi}\catcode`\%=\active\def
\end{pgfscope}%
\begin{pgfscope}%
\pgfpathrectangle{\pgfqpoint{0.674954in}{0.862305in}}{\pgfqpoint{6.515802in}{4.369081in}}%
\pgfusepath{clip}%
\pgfsetbuttcap%
\pgfsetroundjoin%
\pgfsetlinewidth{2.007500pt}%
\definecolor{currentstroke}{rgb}{0.501961,0.501961,0.501961}%
\pgfsetstrokecolor{currentstroke}%
\pgfsetstrokeopacity{0.300000}%
\pgfsetdash{{7.400000pt}{3.200000pt}}{0.000000pt}%
\pgfpathmoveto{\pgfqpoint{0.674954in}{4.139116in}}%
\pgfpathlineto{\pgfqpoint{7.190756in}{4.139116in}}%
\pgfusepath{stroke}%
\end{pgfscope}%
\begin{pgfscope}%
\pgfsetbuttcap%
\pgfsetroundjoin%
\definecolor{currentfill}{rgb}{0.000000,0.000000,0.000000}%
\pgfsetfillcolor{currentfill}%
\pgfsetlinewidth{0.803000pt}%
\definecolor{currentstroke}{rgb}{0.000000,0.000000,0.000000}%
\pgfsetstrokecolor{currentstroke}%
\pgfsetdash{}{0pt}%
\pgfsys@defobject{currentmarker}{\pgfqpoint{-0.048611in}{0.000000in}}{\pgfqpoint{-0.000000in}{0.000000in}}{%
\pgfpathmoveto{\pgfqpoint{-0.000000in}{0.000000in}}%
\pgfpathlineto{\pgfqpoint{-0.048611in}{0.000000in}}%
\pgfusepath{stroke,fill}%
}%
\begin{pgfscope}%
\pgfsys@transformshift{0.674954in}{4.139116in}%
\pgfsys@useobject{currentmarker}{}%
\end{pgfscope}%
\end{pgfscope}%
\begin{pgfscope}%
\definecolor{textcolor}{rgb}{0.000000,0.000000,0.000000}%
\pgfsetstrokecolor{textcolor}%
\pgfsetfillcolor{textcolor}%
\pgftext[x=0.259244in, y=4.020501in, left, base]{\color{textcolor}{\rmfamily\fontsize{25.000000}{30.000000}\selectfont\catcode`\^=\active\def^{\ifmmode\sp\else\^{}\fi}\catcode`\%=\active\def
\end{pgfscope}%
\begin{pgfscope}%
\pgfpathrectangle{\pgfqpoint{0.674954in}{0.862305in}}{\pgfqpoint{6.515802in}{4.369081in}}%
\pgfusepath{clip}%
\pgfsetbuttcap%
\pgfsetroundjoin%
\pgfsetlinewidth{2.007500pt}%
\definecolor{currentstroke}{rgb}{0.501961,0.501961,0.501961}%
\pgfsetstrokecolor{currentstroke}%
\pgfsetstrokeopacity{0.300000}%
\pgfsetdash{{7.400000pt}{3.200000pt}}{0.000000pt}%
\pgfpathmoveto{\pgfqpoint{0.674954in}{5.231386in}}%
\pgfpathlineto{\pgfqpoint{7.190756in}{5.231386in}}%
\pgfusepath{stroke}%
\end{pgfscope}%
\begin{pgfscope}%
\pgfsetbuttcap%
\pgfsetroundjoin%
\definecolor{currentfill}{rgb}{0.000000,0.000000,0.000000}%
\pgfsetfillcolor{currentfill}%
\pgfsetlinewidth{0.803000pt}%
\definecolor{currentstroke}{rgb}{0.000000,0.000000,0.000000}%
\pgfsetstrokecolor{currentstroke}%
\pgfsetdash{}{0pt}%
\pgfsys@defobject{currentmarker}{\pgfqpoint{-0.048611in}{0.000000in}}{\pgfqpoint{-0.000000in}{0.000000in}}{%
\pgfpathmoveto{\pgfqpoint{-0.000000in}{0.000000in}}%
\pgfpathlineto{\pgfqpoint{-0.048611in}{0.000000in}}%
\pgfusepath{stroke,fill}%
}%
\begin{pgfscope}%
\pgfsys@transformshift{0.674954in}{5.231386in}%
\pgfsys@useobject{currentmarker}{}%
\end{pgfscope}%
\end{pgfscope}%
\begin{pgfscope}%
\definecolor{textcolor}{rgb}{0.000000,0.000000,0.000000}%
\pgfsetstrokecolor{textcolor}%
\pgfsetfillcolor{textcolor}%
\pgftext[x=0.100000in, y=5.112772in, left, base]{\color{textcolor}{\rmfamily\fontsize{25.000000}{30.000000}\selectfont\catcode`\^=\active\def^{\ifmmode\sp\else\^{}\fi}\catcode`\%=\active\def
\end{pgfscope}%
\begin{pgfscope}%
\pgfpathrectangle{\pgfqpoint{0.674954in}{0.862305in}}{\pgfqpoint{6.515802in}{4.369081in}}%
\pgfusepath{clip}%
\pgfsetrectcap%
\pgfsetroundjoin%
\pgfsetlinewidth{2.509375pt}%
\definecolor{currentstroke}{rgb}{0.050980,0.415686,0.509804}%
\pgfsetstrokecolor{currentstroke}%
\pgfsetdash{}{0pt}%
\pgfpathmoveto{\pgfqpoint{0.674954in}{4.458208in}}%
\pgfpathlineto{\pgfqpoint{1.760921in}{4.604709in}}%
\pgfpathlineto{\pgfqpoint{2.846888in}{4.548782in}}%
\pgfpathlineto{\pgfqpoint{3.932855in}{4.541374in}}%
\pgfpathlineto{\pgfqpoint{5.018822in}{4.738066in}}%
\pgfpathlineto{\pgfqpoint{6.104789in}{4.641322in}}%
\pgfpathlineto{\pgfqpoint{6.740039in}{4.678612in}}%
\pgfpathlineto{\pgfqpoint{7.190756in}{4.727877in}}%
\pgfusepath{stroke}%
\end{pgfscope}%
\begin{pgfscope}%
\pgfpathrectangle{\pgfqpoint{0.674954in}{0.862305in}}{\pgfqpoint{6.515802in}{4.369081in}}%
\pgfusepath{clip}%
\pgfsetbuttcap%
\pgfsetroundjoin%
\definecolor{currentfill}{rgb}{0.050980,0.415686,0.509804}%
\pgfsetfillcolor{currentfill}%
\pgfsetlinewidth{1.003750pt}%
\definecolor{currentstroke}{rgb}{0.050980,0.415686,0.509804}%
\pgfsetstrokecolor{currentstroke}%
\pgfsetdash{}{0pt}%
\pgfsys@defobject{currentmarker}{\pgfqpoint{-0.055556in}{-0.055556in}}{\pgfqpoint{0.055556in}{0.055556in}}{%
\pgfpathmoveto{\pgfqpoint{0.000000in}{-0.055556in}}%
\pgfpathcurveto{\pgfqpoint{0.014734in}{-0.055556in}}{\pgfqpoint{0.028866in}{-0.049702in}}{\pgfqpoint{0.039284in}{-0.039284in}}%
\pgfpathcurveto{\pgfqpoint{0.049702in}{-0.028866in}}{\pgfqpoint{0.055556in}{-0.014734in}}{\pgfqpoint{0.055556in}{0.000000in}}%
\pgfpathcurveto{\pgfqpoint{0.055556in}{0.014734in}}{\pgfqpoint{0.049702in}{0.028866in}}{\pgfqpoint{0.039284in}{0.039284in}}%
\pgfpathcurveto{\pgfqpoint{0.028866in}{0.049702in}}{\pgfqpoint{0.014734in}{0.055556in}}{\pgfqpoint{0.000000in}{0.055556in}}%
\pgfpathcurveto{\pgfqpoint{-0.014734in}{0.055556in}}{\pgfqpoint{-0.028866in}{0.049702in}}{\pgfqpoint{-0.039284in}{0.039284in}}%
\pgfpathcurveto{\pgfqpoint{-0.049702in}{0.028866in}}{\pgfqpoint{-0.055556in}{0.014734in}}{\pgfqpoint{-0.055556in}{0.000000in}}%
\pgfpathcurveto{\pgfqpoint{-0.055556in}{-0.014734in}}{\pgfqpoint{-0.049702in}{-0.028866in}}{\pgfqpoint{-0.039284in}{-0.039284in}}%
\pgfpathcurveto{\pgfqpoint{-0.028866in}{-0.049702in}}{\pgfqpoint{-0.014734in}{-0.055556in}}{\pgfqpoint{0.000000in}{-0.055556in}}%
\pgfpathlineto{\pgfqpoint{0.000000in}{-0.055556in}}%
\pgfpathclose%
\pgfusepath{stroke,fill}%
}%
\begin{pgfscope}%
\pgfsys@transformshift{0.674954in}{4.458208in}%
\pgfsys@useobject{currentmarker}{}%
\end{pgfscope}%
\begin{pgfscope}%
\pgfsys@transformshift{1.760921in}{4.604709in}%
\pgfsys@useobject{currentmarker}{}%
\end{pgfscope}%
\begin{pgfscope}%
\pgfsys@transformshift{2.846888in}{4.548782in}%
\pgfsys@useobject{currentmarker}{}%
\end{pgfscope}%
\begin{pgfscope}%
\pgfsys@transformshift{3.932855in}{4.541374in}%
\pgfsys@useobject{currentmarker}{}%
\end{pgfscope}%
\begin{pgfscope}%
\pgfsys@transformshift{5.018822in}{4.738066in}%
\pgfsys@useobject{currentmarker}{}%
\end{pgfscope}%
\begin{pgfscope}%
\pgfsys@transformshift{6.104789in}{4.641322in}%
\pgfsys@useobject{currentmarker}{}%
\end{pgfscope}%
\begin{pgfscope}%
\pgfsys@transformshift{6.740039in}{4.678612in}%
\pgfsys@useobject{currentmarker}{}%
\end{pgfscope}%
\begin{pgfscope}%
\pgfsys@transformshift{7.190756in}{4.727877in}%
\pgfsys@useobject{currentmarker}{}%
\end{pgfscope}%
\end{pgfscope}%
\begin{pgfscope}%
\pgfpathrectangle{\pgfqpoint{0.674954in}{0.862305in}}{\pgfqpoint{6.515802in}{4.369081in}}%
\pgfusepath{clip}%
\pgfsetbuttcap%
\pgfsetroundjoin%
\pgfsetlinewidth{2.509375pt}%
\definecolor{currentstroke}{rgb}{0.960784,0.462745,0.000000}%
\pgfsetstrokecolor{currentstroke}%
\pgfsetdash{{9.250000pt}{4.000000pt}}{0.000000pt}%
\pgfpathmoveto{\pgfqpoint{0.674954in}{4.097410in}}%
\pgfpathlineto{\pgfqpoint{1.760921in}{4.207932in}}%
\pgfpathlineto{\pgfqpoint{2.846888in}{4.134733in}}%
\pgfpathlineto{\pgfqpoint{3.932855in}{4.073509in}}%
\pgfpathlineto{\pgfqpoint{5.018822in}{4.373850in}}%
\pgfpathlineto{\pgfqpoint{6.104789in}{4.266882in}}%
\pgfpathlineto{\pgfqpoint{6.740039in}{4.381144in}}%
\pgfpathlineto{\pgfqpoint{7.190756in}{4.358948in}}%
\pgfusepath{stroke}%
\end{pgfscope}%
\begin{pgfscope}%
\pgfpathrectangle{\pgfqpoint{0.674954in}{0.862305in}}{\pgfqpoint{6.515802in}{4.369081in}}%
\pgfusepath{clip}%
\pgfsetbuttcap%
\pgfsetmiterjoin%
\definecolor{currentfill}{rgb}{0.960784,0.462745,0.000000}%
\pgfsetfillcolor{currentfill}%
\pgfsetlinewidth{1.003750pt}%
\definecolor{currentstroke}{rgb}{0.960784,0.462745,0.000000}%
\pgfsetstrokecolor{currentstroke}%
\pgfsetdash{}{0pt}%
\pgfsys@defobject{currentmarker}{\pgfqpoint{-0.055556in}{-0.055556in}}{\pgfqpoint{0.055556in}{0.055556in}}{%
\pgfpathmoveto{\pgfqpoint{-0.055556in}{-0.055556in}}%
\pgfpathlineto{\pgfqpoint{0.055556in}{-0.055556in}}%
\pgfpathlineto{\pgfqpoint{0.055556in}{0.055556in}}%
\pgfpathlineto{\pgfqpoint{-0.055556in}{0.055556in}}%
\pgfpathlineto{\pgfqpoint{-0.055556in}{-0.055556in}}%
\pgfpathclose%
\pgfusepath{stroke,fill}%
}%
\begin{pgfscope}%
\pgfsys@transformshift{0.674954in}{4.097410in}%
\pgfsys@useobject{currentmarker}{}%
\end{pgfscope}%
\begin{pgfscope}%
\pgfsys@transformshift{1.760921in}{4.207932in}%
\pgfsys@useobject{currentmarker}{}%
\end{pgfscope}%
\begin{pgfscope}%
\pgfsys@transformshift{2.846888in}{4.134733in}%
\pgfsys@useobject{currentmarker}{}%
\end{pgfscope}%
\begin{pgfscope}%
\pgfsys@transformshift{3.932855in}{4.073509in}%
\pgfsys@useobject{currentmarker}{}%
\end{pgfscope}%
\begin{pgfscope}%
\pgfsys@transformshift{5.018822in}{4.373850in}%
\pgfsys@useobject{currentmarker}{}%
\end{pgfscope}%
\begin{pgfscope}%
\pgfsys@transformshift{6.104789in}{4.266882in}%
\pgfsys@useobject{currentmarker}{}%
\end{pgfscope}%
\begin{pgfscope}%
\pgfsys@transformshift{6.740039in}{4.381144in}%
\pgfsys@useobject{currentmarker}{}%
\end{pgfscope}%
\begin{pgfscope}%
\pgfsys@transformshift{7.190756in}{4.358948in}%
\pgfsys@useobject{currentmarker}{}%
\end{pgfscope}%
\end{pgfscope}%
\begin{pgfscope}%
\pgfpathrectangle{\pgfqpoint{0.674954in}{0.862305in}}{\pgfqpoint{6.515802in}{4.369081in}}%
\pgfusepath{clip}%
\pgfsetbuttcap%
\pgfsetroundjoin%
\pgfsetlinewidth{2.509375pt}%
\definecolor{currentstroke}{rgb}{0.219608,0.219608,0.219608}%
\pgfsetstrokecolor{currentstroke}%
\pgfsetdash{{2.500000pt}{4.125000pt}}{0.000000pt}%
\pgfpathmoveto{\pgfqpoint{0.674954in}{3.216004in}}%
\pgfpathlineto{\pgfqpoint{1.760921in}{3.295358in}}%
\pgfpathlineto{\pgfqpoint{2.846888in}{3.308342in}}%
\pgfpathlineto{\pgfqpoint{3.932855in}{3.124896in}}%
\pgfpathlineto{\pgfqpoint{5.018822in}{3.632677in}}%
\pgfpathlineto{\pgfqpoint{6.104789in}{3.389689in}}%
\pgfpathlineto{\pgfqpoint{6.740039in}{3.563882in}}%
\pgfpathlineto{\pgfqpoint{7.190756in}{3.578515in}}%
\pgfusepath{stroke}%
\end{pgfscope}%
\begin{pgfscope}%
\pgfpathrectangle{\pgfqpoint{0.674954in}{0.862305in}}{\pgfqpoint{6.515802in}{4.369081in}}%
\pgfusepath{clip}%
\pgfsetbuttcap%
\pgfsetmiterjoin%
\definecolor{currentfill}{rgb}{0.219608,0.219608,0.219608}%
\pgfsetfillcolor{currentfill}%
\pgfsetlinewidth{1.003750pt}%
\definecolor{currentstroke}{rgb}{0.219608,0.219608,0.219608}%
\pgfsetstrokecolor{currentstroke}%
\pgfsetdash{}{0pt}%
\pgfsys@defobject{currentmarker}{\pgfqpoint{-0.055556in}{-0.055556in}}{\pgfqpoint{0.055556in}{0.055556in}}{%
\pgfpathmoveto{\pgfqpoint{0.000000in}{0.055556in}}%
\pgfpathlineto{\pgfqpoint{-0.055556in}{-0.055556in}}%
\pgfpathlineto{\pgfqpoint{0.055556in}{-0.055556in}}%
\pgfpathlineto{\pgfqpoint{0.000000in}{0.055556in}}%
\pgfpathclose%
\pgfusepath{stroke,fill}%
}%
\begin{pgfscope}%
\pgfsys@transformshift{0.674954in}{3.216004in}%
\pgfsys@useobject{currentmarker}{}%
\end{pgfscope}%
\begin{pgfscope}%
\pgfsys@transformshift{1.760921in}{3.295358in}%
\pgfsys@useobject{currentmarker}{}%
\end{pgfscope}%
\begin{pgfscope}%
\pgfsys@transformshift{2.846888in}{3.308342in}%
\pgfsys@useobject{currentmarker}{}%
\end{pgfscope}%
\begin{pgfscope}%
\pgfsys@transformshift{3.932855in}{3.124896in}%
\pgfsys@useobject{currentmarker}{}%
\end{pgfscope}%
\begin{pgfscope}%
\pgfsys@transformshift{5.018822in}{3.632677in}%
\pgfsys@useobject{currentmarker}{}%
\end{pgfscope}%
\begin{pgfscope}%
\pgfsys@transformshift{6.104789in}{3.389689in}%
\pgfsys@useobject{currentmarker}{}%
\end{pgfscope}%
\begin{pgfscope}%
\pgfsys@transformshift{6.740039in}{3.563882in}%
\pgfsys@useobject{currentmarker}{}%
\end{pgfscope}%
\begin{pgfscope}%
\pgfsys@transformshift{7.190756in}{3.578515in}%
\pgfsys@useobject{currentmarker}{}%
\end{pgfscope}%
\end{pgfscope}%
\begin{pgfscope}%
\pgfsetrectcap%
\pgfsetmiterjoin%
\pgfsetlinewidth{2.007500pt}%
\definecolor{currentstroke}{rgb}{0.000000,0.000000,0.000000}%
\pgfsetstrokecolor{currentstroke}%
\pgfsetdash{}{0pt}%
\pgfpathmoveto{\pgfqpoint{0.674954in}{0.862305in}}%
\pgfpathlineto{\pgfqpoint{0.674954in}{5.231386in}}%
\pgfusepath{stroke}%
\end{pgfscope}%
\begin{pgfscope}%
\pgfsetrectcap%
\pgfsetmiterjoin%
\pgfsetlinewidth{2.007500pt}%
\definecolor{currentstroke}{rgb}{0.000000,0.000000,0.000000}%
\pgfsetstrokecolor{currentstroke}%
\pgfsetdash{}{0pt}%
\pgfpathmoveto{\pgfqpoint{0.674954in}{0.862305in}}%
\pgfpathlineto{\pgfqpoint{7.190756in}{0.862305in}}%
\pgfusepath{stroke}%
\end{pgfscope}%
\begin{pgfscope}%
\pgfsetbuttcap%
\pgfsetmiterjoin%
\definecolor{currentfill}{rgb}{1.000000,1.000000,1.000000}%
\pgfsetfillcolor{currentfill}%
\pgfsetlinewidth{1.003750pt}%
\definecolor{currentstroke}{rgb}{0.800000,0.800000,0.800000}%
\pgfsetstrokecolor{currentstroke}%
\pgfsetdash{}{0pt}%
\pgfpathmoveto{\pgfqpoint{0.869398in}{1.001194in}}%
\pgfpathlineto{\pgfqpoint{2.213036in}{1.001194in}}%
\pgfpathquadraticcurveto{\pgfqpoint{2.268591in}{1.001194in}}{\pgfqpoint{2.268591in}{1.056749in}}%
\pgfpathlineto{\pgfqpoint{2.268591in}{2.191056in}}%
\pgfpathquadraticcurveto{\pgfqpoint{2.268591in}{2.246612in}}{\pgfqpoint{2.213036in}{2.246612in}}%
\pgfpathlineto{\pgfqpoint{0.869398in}{2.246612in}}%
\pgfpathquadraticcurveto{\pgfqpoint{0.813843in}{2.246612in}}{\pgfqpoint{0.813843in}{2.191056in}}%
\pgfpathlineto{\pgfqpoint{0.813843in}{1.056749in}}%
\pgfpathquadraticcurveto{\pgfqpoint{0.813843in}{1.001194in}}{\pgfqpoint{0.869398in}{1.001194in}}%
\pgfpathlineto{\pgfqpoint{0.869398in}{1.001194in}}%
\pgfpathclose%
\pgfusepath{stroke,fill}%
\end{pgfscope}%
\begin{pgfscope}%
\pgfsetrectcap%
\pgfsetroundjoin%
\pgfsetlinewidth{2.509375pt}%
\definecolor{currentstroke}{rgb}{0.050980,0.415686,0.509804}%
\pgfsetstrokecolor{currentstroke}%
\pgfsetdash{}{0pt}%
\pgfpathmoveto{\pgfqpoint{0.924954in}{2.038278in}}%
\pgfpathlineto{\pgfqpoint{1.202732in}{2.038278in}}%
\pgfpathlineto{\pgfqpoint{1.480509in}{2.038278in}}%
\pgfusepath{stroke}%
\end{pgfscope}%
\begin{pgfscope}%
\pgfsetbuttcap%
\pgfsetroundjoin%
\definecolor{currentfill}{rgb}{0.050980,0.415686,0.509804}%
\pgfsetfillcolor{currentfill}%
\pgfsetlinewidth{1.003750pt}%
\definecolor{currentstroke}{rgb}{0.050980,0.415686,0.509804}%
\pgfsetstrokecolor{currentstroke}%
\pgfsetdash{}{0pt}%
\pgfsys@defobject{currentmarker}{\pgfqpoint{-0.055556in}{-0.055556in}}{\pgfqpoint{0.055556in}{0.055556in}}{%
\pgfpathmoveto{\pgfqpoint{0.000000in}{-0.055556in}}%
\pgfpathcurveto{\pgfqpoint{0.014734in}{-0.055556in}}{\pgfqpoint{0.028866in}{-0.049702in}}{\pgfqpoint{0.039284in}{-0.039284in}}%
\pgfpathcurveto{\pgfqpoint{0.049702in}{-0.028866in}}{\pgfqpoint{0.055556in}{-0.014734in}}{\pgfqpoint{0.055556in}{0.000000in}}%
\pgfpathcurveto{\pgfqpoint{0.055556in}{0.014734in}}{\pgfqpoint{0.049702in}{0.028866in}}{\pgfqpoint{0.039284in}{0.039284in}}%
\pgfpathcurveto{\pgfqpoint{0.028866in}{0.049702in}}{\pgfqpoint{0.014734in}{0.055556in}}{\pgfqpoint{0.000000in}{0.055556in}}%
\pgfpathcurveto{\pgfqpoint{-0.014734in}{0.055556in}}{\pgfqpoint{-0.028866in}{0.049702in}}{\pgfqpoint{-0.039284in}{0.039284in}}%
\pgfpathcurveto{\pgfqpoint{-0.049702in}{0.028866in}}{\pgfqpoint{-0.055556in}{0.014734in}}{\pgfqpoint{-0.055556in}{0.000000in}}%
\pgfpathcurveto{\pgfqpoint{-0.055556in}{-0.014734in}}{\pgfqpoint{-0.049702in}{-0.028866in}}{\pgfqpoint{-0.039284in}{-0.039284in}}%
\pgfpathcurveto{\pgfqpoint{-0.028866in}{-0.049702in}}{\pgfqpoint{-0.014734in}{-0.055556in}}{\pgfqpoint{0.000000in}{-0.055556in}}%
\pgfpathlineto{\pgfqpoint{0.000000in}{-0.055556in}}%
\pgfpathclose%
\pgfusepath{stroke,fill}%
}%
\begin{pgfscope}%
\pgfsys@transformshift{1.202732in}{2.038278in}%
\pgfsys@useobject{currentmarker}{}%
\end{pgfscope}%
\end{pgfscope}%
\begin{pgfscope}%
\definecolor{textcolor}{rgb}{0.000000,0.000000,0.000000}%
\pgfsetstrokecolor{textcolor}%
\pgfsetfillcolor{textcolor}%
\pgftext[x=1.702732in,y=1.941056in,left,base]{\color{textcolor}{\rmfamily\fontsize{20.000000}{24.000000}\selectfont\catcode`\^=\active\def^{\ifmmode\sp\else\^{}\fi}\catcode`\%=\active\def
\end{pgfscope}%
\begin{pgfscope}%
\pgfsetbuttcap%
\pgfsetroundjoin%
\pgfsetlinewidth{2.509375pt}%
\definecolor{currentstroke}{rgb}{0.960784,0.462745,0.000000}%
\pgfsetstrokecolor{currentstroke}%
\pgfsetdash{{9.250000pt}{4.000000pt}}{0.000000pt}%
\pgfpathmoveto{\pgfqpoint{0.924954in}{1.650917in}}%
\pgfpathlineto{\pgfqpoint{1.202732in}{1.650917in}}%
\pgfpathlineto{\pgfqpoint{1.480509in}{1.650917in}}%
\pgfusepath{stroke}%
\end{pgfscope}%
\begin{pgfscope}%
\pgfsetbuttcap%
\pgfsetmiterjoin%
\definecolor{currentfill}{rgb}{0.960784,0.462745,0.000000}%
\pgfsetfillcolor{currentfill}%
\pgfsetlinewidth{1.003750pt}%
\definecolor{currentstroke}{rgb}{0.960784,0.462745,0.000000}%
\pgfsetstrokecolor{currentstroke}%
\pgfsetdash{}{0pt}%
\pgfsys@defobject{currentmarker}{\pgfqpoint{-0.055556in}{-0.055556in}}{\pgfqpoint{0.055556in}{0.055556in}}{%
\pgfpathmoveto{\pgfqpoint{-0.055556in}{-0.055556in}}%
\pgfpathlineto{\pgfqpoint{0.055556in}{-0.055556in}}%
\pgfpathlineto{\pgfqpoint{0.055556in}{0.055556in}}%
\pgfpathlineto{\pgfqpoint{-0.055556in}{0.055556in}}%
\pgfpathlineto{\pgfqpoint{-0.055556in}{-0.055556in}}%
\pgfpathclose%
\pgfusepath{stroke,fill}%
}%
\begin{pgfscope}%
\pgfsys@transformshift{1.202732in}{1.650917in}%
\pgfsys@useobject{currentmarker}{}%
\end{pgfscope}%
\end{pgfscope}%
\begin{pgfscope}%
\definecolor{textcolor}{rgb}{0.000000,0.000000,0.000000}%
\pgfsetstrokecolor{textcolor}%
\pgfsetfillcolor{textcolor}%
\pgftext[x=1.702732in,y=1.553694in,left,base]{\color{textcolor}{\rmfamily\fontsize{20.000000}{24.000000}\selectfont\catcode`\^=\active\def^{\ifmmode\sp\else\^{}\fi}\catcode`\%=\active\def
\end{pgfscope}%
\begin{pgfscope}%
\pgfsetbuttcap%
\pgfsetroundjoin%
\pgfsetlinewidth{2.509375pt}%
\definecolor{currentstroke}{rgb}{0.219608,0.219608,0.219608}%
\pgfsetstrokecolor{currentstroke}%
\pgfsetdash{{2.500000pt}{4.125000pt}}{0.000000pt}%
\pgfpathmoveto{\pgfqpoint{0.924954in}{1.263555in}}%
\pgfpathlineto{\pgfqpoint{1.202732in}{1.263555in}}%
\pgfpathlineto{\pgfqpoint{1.480509in}{1.263555in}}%
\pgfusepath{stroke}%
\end{pgfscope}%
\begin{pgfscope}%
\pgfsetbuttcap%
\pgfsetmiterjoin%
\definecolor{currentfill}{rgb}{0.219608,0.219608,0.219608}%
\pgfsetfillcolor{currentfill}%
\pgfsetlinewidth{1.003750pt}%
\definecolor{currentstroke}{rgb}{0.219608,0.219608,0.219608}%
\pgfsetstrokecolor{currentstroke}%
\pgfsetdash{}{0pt}%
\pgfsys@defobject{currentmarker}{\pgfqpoint{-0.055556in}{-0.055556in}}{\pgfqpoint{0.055556in}{0.055556in}}{%
\pgfpathmoveto{\pgfqpoint{0.000000in}{0.055556in}}%
\pgfpathlineto{\pgfqpoint{-0.055556in}{-0.055556in}}%
\pgfpathlineto{\pgfqpoint{0.055556in}{-0.055556in}}%
\pgfpathlineto{\pgfqpoint{0.000000in}{0.055556in}}%
\pgfpathclose%
\pgfusepath{stroke,fill}%
}%
\begin{pgfscope}%
\pgfsys@transformshift{1.202732in}{1.263555in}%
\pgfsys@useobject{currentmarker}{}%
\end{pgfscope}%
\end{pgfscope}%
\begin{pgfscope}%
\definecolor{textcolor}{rgb}{0.000000,0.000000,0.000000}%
\pgfsetstrokecolor{textcolor}%
\pgfsetfillcolor{textcolor}%
\pgftext[x=1.702732in,y=1.166333in,left,base]{\color{textcolor}{\rmfamily\fontsize{20.000000}{24.000000}\selectfont\catcode`\^=\active\def^{\ifmmode\sp\else\^{}\fi}\catcode`\%=\active\def
\end{pgfscope}%
\end{pgfpicture}%
\makeatother%
\endgroup%

%% file: sec/6_conclusion.tex
\section{Conclusion}
\label{sec:conclusion}

We present \ours{}, a simple and efficient framework to adapt visual geometric grounded transformers for RGB-T camera pose estimation and 3D scene reconstruction.
By leveraging lightweight LoRA adapters and a carefully designed batching strategy, \ours{} achieves state-of-the-art performance with minimal computational overhead and a modest amount of data, making it a practical solution for real-world applications.
Our approach not only outperforms existing methods across all metrics but also demonstrates robustness in challenging real-world conditions such as low lighting, dense smoke, and asynchronous data capture.
While this study focuses on RGB-Thermal reconstruction, the methodology is possibly extensible to other modalities---such as multi-spectral imaging---which remains a subject for future work.
By showing that high-performance multimodal reconstruction is achievable with minimal adaptation and a relatively small amount of data, we hope to inspire broader adoption of parameter-efficient fine-tuning in multimodal geometric vision.

%% file: sec/acknowledgements.tex
We thank Ivan Skorokhodov for his assistance with writing, earlier drafts, and help with several experiments.
We also thank Martin Magnusson for valuable discussions and for suggesting several evaluation metrics and datasets.
Part of the experiments was run thanks to the facilities of the Scientific IT and Application Support Center of EPFL.

%% file: supplementary/2_impl_details.tex
\section{Implementation Details}\label{sec:impl_details}

We use the COLMAP + SuperPoint + SuperGlue pipeline provided in \texttt{nerfstudio}, with the default hyperparameters.

For DUSt3R and MASt3R, we use the official implementations from their respective GitHub repositories. These models first estimate relative camera poses and depths for pairs of input images, and then perform global optimization to recover the final camera poses and depths. The authors provide flexible control over the image pairing strategy. To avoid CUDA out-of-memory issues, we use the \texttt{noncyclic-logwin} scene graph with a window size of 5. Specifically, for each image at index $i$, this strategy creates image pairs
$(i - 2^5, i), (i - 2^4, i), \dots, (i + 2^4, i), (i + 2^5, i)$ instead of constructing a complete graph over all possible image pairs. This design preserves substantial overlap between paired views while significantly reducing computational and memory requirements.

MINIMA~\cite{ren2024minima} provides several multimodal image matching models trained with the authors' proposed strategy. Among them, we select the official ROMA~\cite{edstedt2024roma}-based checkpoint, denoted as $\text{MINIMA}_{\text{ROMA}}$, since it achieves the best performance. Match-Anything~\cite{he2025matchanything} reports results for multiple models trained using its proposed framework. However, among the officially released checkpoints, only the Efficient-LoFTR~\cite{eloftr}-based model, denoted as $\text{MA}_{\text{ELoFTR}}$, is publicly available. We therefore use this model as a baseline. For a fair comparison and to reduce computational cost, we adopt the same image pairing strategy for $\text{MA}_{\text{ELoFTR}}$ and $\text{MINIMA}_{\text{ROMA}}$ as used for MASt3R and DUSt3R.

For MP-SfM, we use MINIMA$_{\text{ROMA}}$ as the matching model and apply the same image-pairing strategy as for MASt3R and DUSt3R. For the depth and normal prediction models, we use the default configuration provided by MP-SfM.

%% file: supplementary/3_dataset.tex
\section{SEAR Dataset}

\input{tables/datasets}

We collect a dataset comprising nine scenes using the FLIR One Pro LT camera~\cite{FLIR_One_Pro_Product_Page_2026}. A comparison with other datasets is provided in~\cref{tab:datasets}. The camera operates within $[-20^\circ\text{C}, 120^\circ\text{C}]$ with a thermal accuracy of $\pm 3^\circ\text{C}$. We estimate the thermal precision following the procedure described in ThermoNeRF~\cite{hassanThermoNeRFMultimodalNeural2025}, and find that this characteristic matches the $\pm 0.14^\circ\text{C}$ reported in ThermoNeRF. We follow the postprocessing instructions from ThermoNeRF to extract RGB and thermal values from raw images.

\input{images/oursdataset_supp_samelight}
\input{images/oursdataset_supp_difflight}

Each scene capture consists of two trajectories. For six scenes (\texttt{conference-\allowbreak room}, \texttt{metallic-container},
  \texttt{parking}, \texttt{statue}, \texttt{telescope},
\texttt{old-drinking-foun\allowbreak tain}), the trajectories do not intersect and are recorded under the same favorable lighting conditions (see summary in \cref{tab:oursdataset_supp_samelight}). For the remaining three scenes (\texttt{red-container}, \texttt{house}, \texttt{messy-living-room}), the trajectories intersect and are captured under different lighting conditions---one trajectory is well lit, while the other is poorly lit (see summary in \cref{tab:oursdataset_supp_difflight}). In the latter case, the RGB images from the poorly lit trajectory are dark and therefore unsuitable for localization. For \texttt{telescope} and \texttt{house} scenes, the minimal temperature is lower than the camera's minimum operating bound of $-20^\circ \text{C}$ due to the atmospheric radiation in the sky(the same effect as in~\cite{hassanThermoNeRFMultimodalNeural2025})

%% file: tables/datasets.tex
\newcommand*{\belowrulesepcolor}[1]{%
  \noalign{%
    \kern-\belowrulesep
    \begingroup
    \color{#1}%
    \hrule height\belowrulesep
    \endgroup
  }%
}
\newcommand*{\aboverulesepcolor}[1]{%
  \noalign{%
    \begingroup
    \color{#1}%
    \hrule height\aboverulesep
    \endgroup
    \kern-\aboverulesep
  }%
}

\begin{table}[t]
  \centering
  \caption{Detailed description of the datasets and split used for training and for evaluation.}
  \label{tab:datasets}
  \resizebox{\textwidth}{!}{
    \begin{tabular}{lcccccc}
      \toprule
      Dataset  & \makecell[c]{\#Training \\Scenes} & \makecell[c]{\#Evaluation\\Scenes} & \makecell[c]{Aligned\\modalities} & \makecell[c]{Cross \\positional} & \makecell[c]{Cross\\temporal} & \makecell[c]{Ground\\Truth}  \\
      \midrule
      \multicolumn{7}{c}{For training and quantitative evaluation}\\\midrule
      ThermoScenes & 17  & 5   & True & True   & False & VGGT   \\
      ThermalGaussian    & 11   & 3    & True   & True  & False & VGGT           \\
      ThermalMix    & 4   & 2    & True  & True  & False & VGGT           \\
      ThermalNeRF     & 7        & 2    & False & True  & False & VGGT           \\
      \orebrodataset     & 9      & 6    & True     & True  & False & Motion Capture \\
      \rowcolor{gray!20} Total & 48 & 18 & & & &  \\
      \aboverulesepcolor{gray!20}
      \midrule
      \multicolumn{7}{c}{For qualitative evaluation only} \\\midrule
      SmokeSeer      & 0  &  2   & False  & True     & True & None           \\
      Ours (same light) & 0 & 6 & True & True & True & VGGT \\
      Ours (different lighting conditions)  & 0    & 3    & True   & False  & True  & None   \\
      \rowcolor{gray!20}  Total & 0 & 11 & & & & \\
      \aboverulesepcolor{gray!20}
      \bottomrule
    \end{tabular}
  }

\end{table}

%% file: images/oursdataset_supp_samelight.tex
\begin{table}[!t]
    \centering
    \caption{
    Summary of the \datasetname for well-lit scenes, where both trajectories are captured under the same bright illumination conditions. The figure includes RGB and thermal images, two camera trajectories obtained by running VGGT on the RGB images, the lengths of the two trajectories, and the minimum and maximum temperatures.
    }
    \label{tab:oursdataset_supp_samelight}
    \resizebox{0.9\linewidth}{!}{
    \begin{tabular}{lccccc}
\toprule Scene & RGB & Thermal & Trajectory & \makecell{Length 1 \\ Length 2} & \makecell {Temp. min\\ Temp. max} \\ \midrule
conference-room & \adjustbox{valign=c}{\includegraphics[height=40pt]{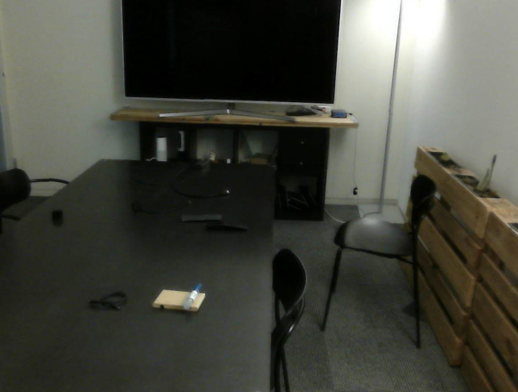} }&  \adjustbox{valign=c}{\includegraphics[height=40pt]{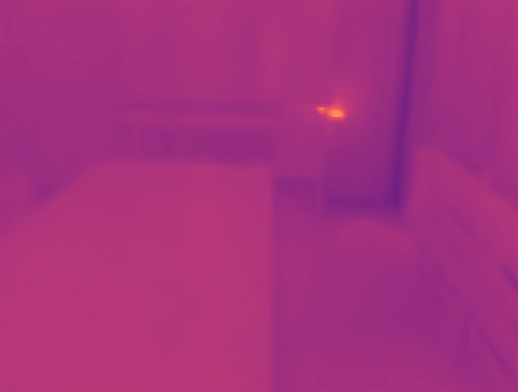} }&  \adjustbox{valign=c}{\includegraphics[height=40pt]{images/OursDataset/two_trajectories/conference-room_pose.png} }&  \makecell{34 \\ 35} &  \makecell{{12.9}$^\circ$C \\ {30.8}$^\circ$C} \\[10pt]\\
metallic-container & \adjustbox{valign=c}{\includegraphics[height=40pt]{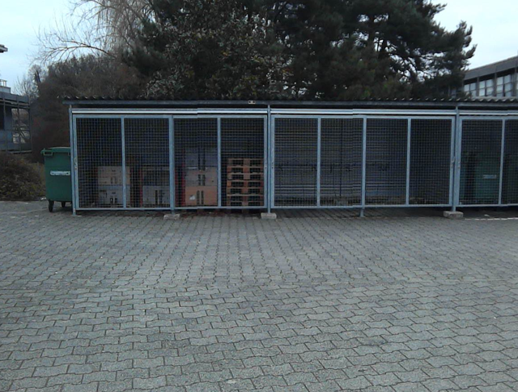} }&  \adjustbox{valign=c}{\includegraphics[height=40pt]{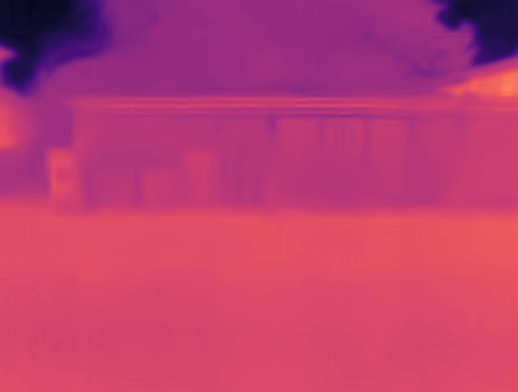} }&  \adjustbox{valign=c}{\includegraphics[height=40pt]{images/OursDataset/two_trajectories/metallic-container_pose.png} }&  \makecell{50 \\ 80} &  \makecell{{-6.8}$^\circ$C \\ {5.6}$^\circ$C} \\[10pt]\\
parking & \adjustbox{valign=c}{\includegraphics[height=40pt]{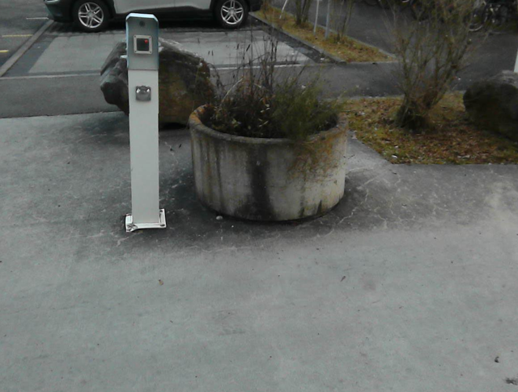} }&  \adjustbox{valign=c}{\includegraphics[height=40pt]{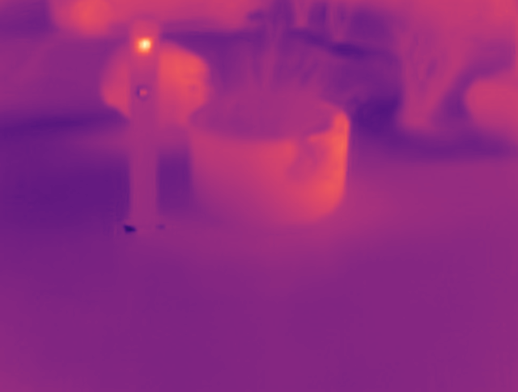} }&  \adjustbox{valign=c}{\includegraphics[height=40pt]{images/OursDataset/two_trajectories/parking_pose.png} }&  \makecell{40 \\ 43} &  \makecell{{-6.8}$^\circ$C \\ {13.1}$^\circ$C} \\[10pt]\\
statue & \adjustbox{valign=c}{\includegraphics[height=40pt]{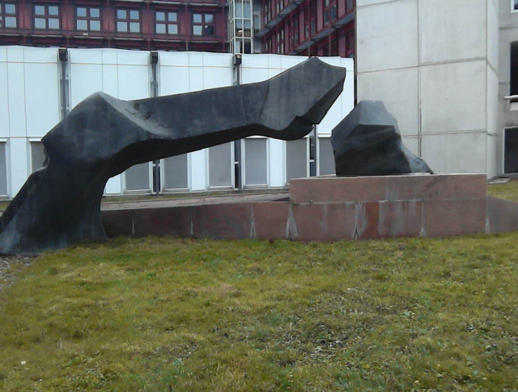} }&  \adjustbox{valign=c}{\includegraphics[height=40pt]{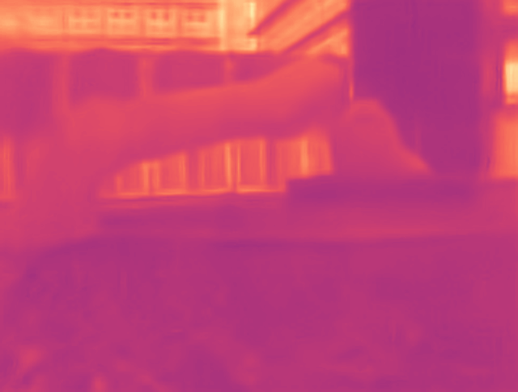} }&  \adjustbox{valign=c}{\includegraphics[height=40pt]{images/OursDataset/two_trajectories/statue_pose.png} }&  \makecell{70 \\ 37} &  \makecell{{-9.6}$^\circ$C \\ {4.9}$^\circ$C} \\[10pt]\\
telescope & \adjustbox{valign=c}{\includegraphics[height=40pt]{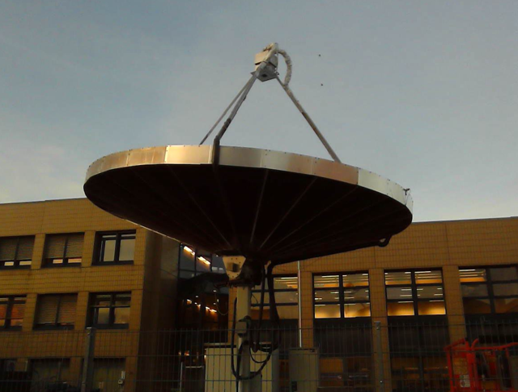} }&  \adjustbox{valign=c}{\includegraphics[height=40pt]{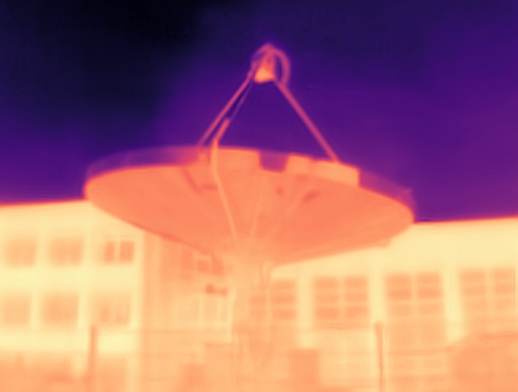} }&  \adjustbox{valign=c}{\includegraphics[height=40pt]{images/OursDataset/two_trajectories/telescope_pose.png} }&  \makecell{42 \\ 54} &  \makecell{{-46.3}$^\circ$C$^*$ \\ {13.4}$^\circ$C} \\[10pt]\\
old-drinking-fountain & \adjustbox{valign=c}{\includegraphics[height=40pt]{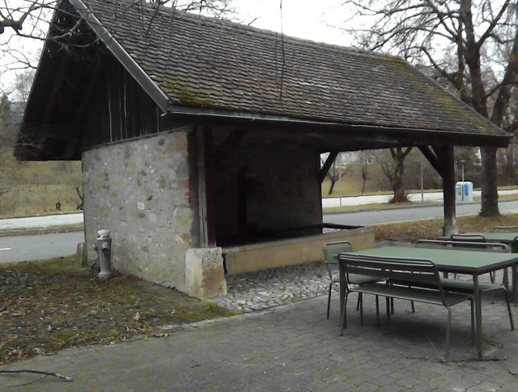} }&  \adjustbox{valign=c}{\includegraphics[height=40pt]{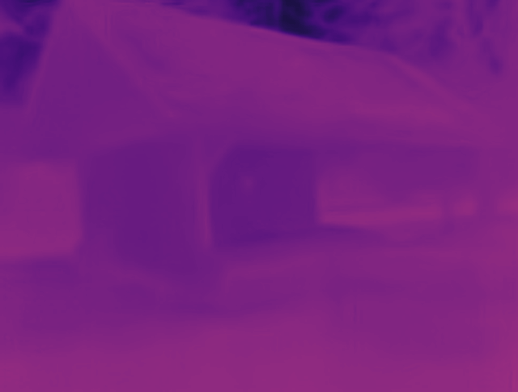} }&  \adjustbox{valign=c}{\includegraphics[height=40pt]{images/OursDataset/two_trajectories/old-drinking-fountain_pose.png} }&  \makecell{70 \\ 54} &  \makecell{{-8.0}$^\circ$C \\ {12.5}$^\circ$C} \\ \bottomrule
    \end{tabular}
    }
    
\end{table}

%% file: images/oursdataset_supp_difflight.tex
\begin{table}[!t]
    \centering
    \caption{
    Summary of the \datasetname for poorly lit scenes, where the trajectories are captured under different lighting conditions. The figure includes RGB and thermal images, the camera trajectory estimated by running VGGT on the thermal images (and is therefore less accurate), the lengths of the two trajectories, and the minimum and maximum temperatures for each trajectory.
    }
    \label{tab:oursdataset_supp_difflight}
    \resizebox{\linewidth}{!}{
    \begin{tabular}{lcccccc}
\toprule Scene & Well-Lit & RGB & Thermal & Trajectory & \makecell{Length} & \makecell {Temp. min\\ Temp. max} \\ \midrule
\multirow{3}{*}{\centering red-container} & True & \adjustbox{valign=c}{\includegraphics[height=40pt]{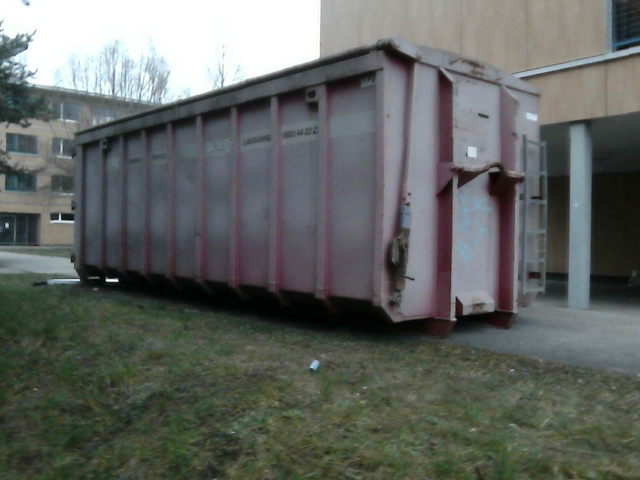} }&  \adjustbox{valign=c}{\includegraphics[height=40pt]{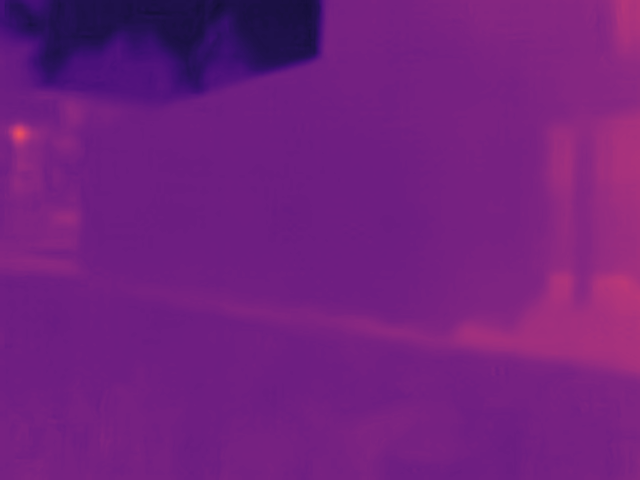} }&  \adjustbox{valign=c}{\includegraphics[height=40pt]{images/OursDataset/different_lighting/red-container_pose.png} }&  \makecell{55} &  \makecell{{-7.0}$^\circ$C \\ {6.8}$^\circ$C} \\[-9pt]\\
 & False &\adjustbox{valign=c}{\includegraphics[height=40pt]{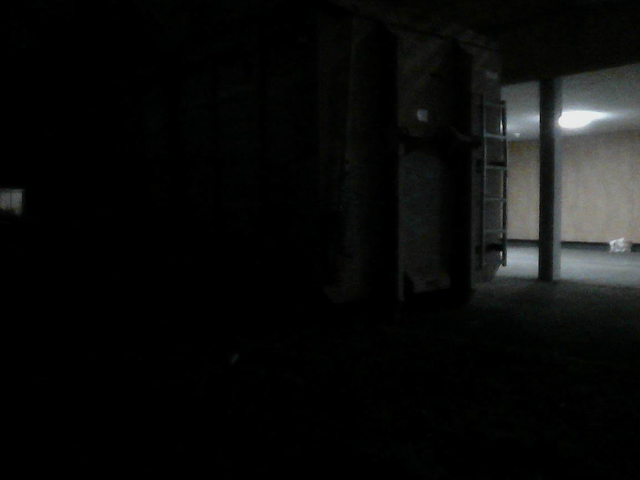} }&  \adjustbox{valign=c}{\includegraphics[height=40pt]{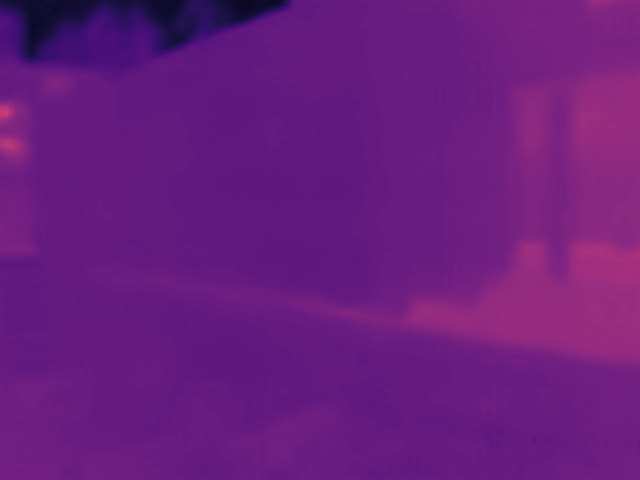} }&  \adjustbox{valign=c}{\includegraphics[height=40pt]{images/OursDataset/different_lighting/red-container_pose.png} }&  \makecell{60} &  \makecell{{-8.6}$^\circ$C \\ {6.4}$^\circ$C} \\[10pt]\midrule\\
\multirow{3}{*}{\centering house} & True & \adjustbox{valign=c}{\includegraphics[height=40pt]{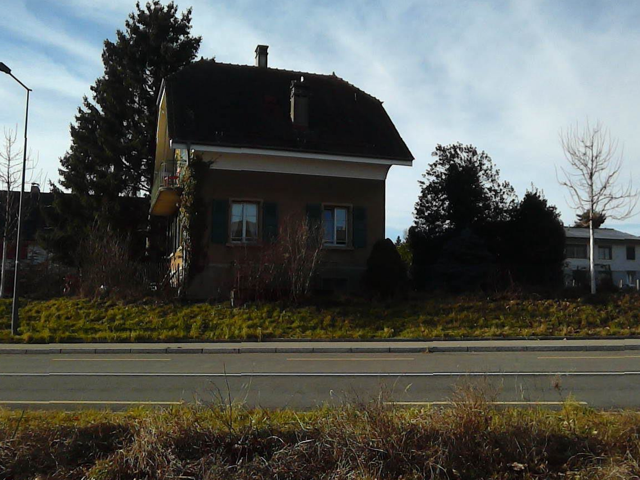} }&  \adjustbox{valign=c}{\includegraphics[height=40pt]{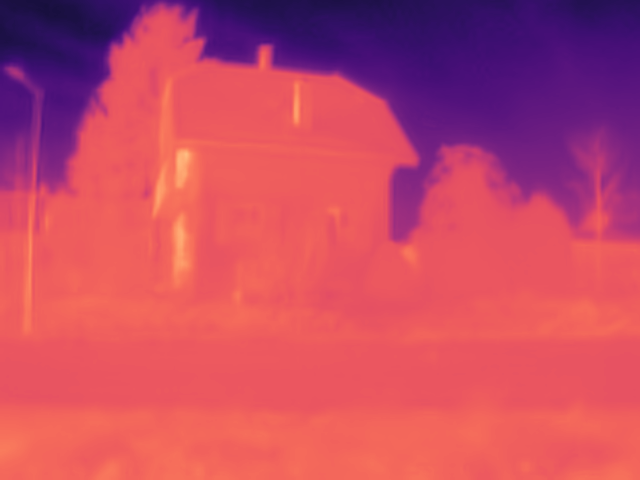} }&  \adjustbox{valign=c}{\includegraphics[height=40pt]{images/OursDataset/different_lighting/house_pose.png} }&  \makecell{60} &  \makecell{{-38.4}$^\circ$C$^*$ \\ {13.6}$^\circ$C} \\[-9pt]\\
 & False &\adjustbox{valign=c}{\includegraphics[height=40pt]{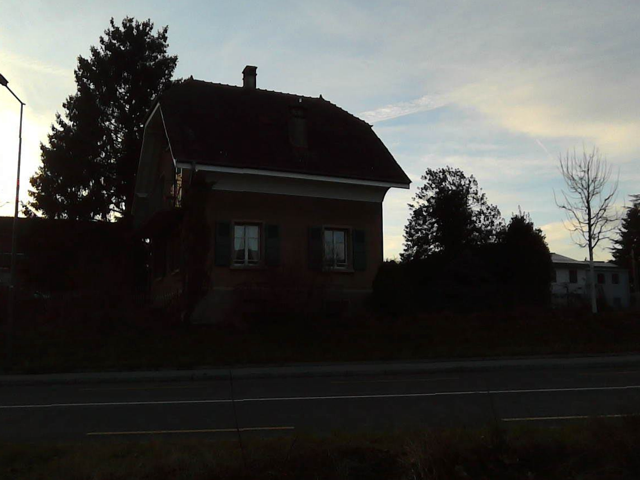} }&  \adjustbox{valign=c}{\includegraphics[height=40pt]{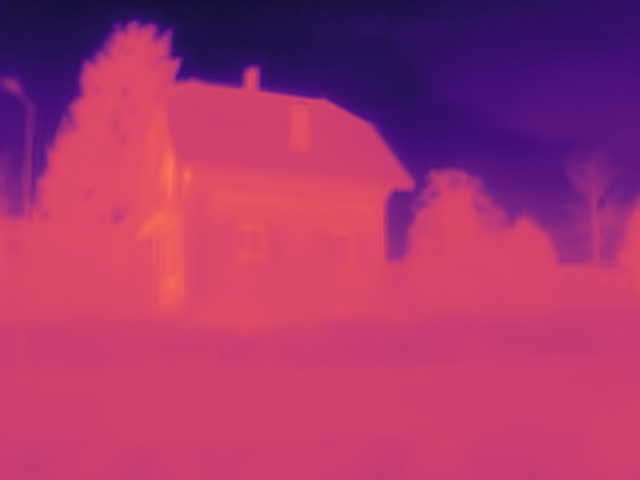} }&  \adjustbox{valign=c}{\includegraphics[height=40pt]{images/OursDataset/different_lighting/house_pose.png} }&  \makecell{60} &  \makecell{{-45.0}$^\circ$C$^*$ \\ {33.4}$^\circ$C} \\[10pt]\midrule\\
\multirow{3}{*}{\centering messy-living-room} & True & \adjustbox{valign=c}{\includegraphics[height=40pt]{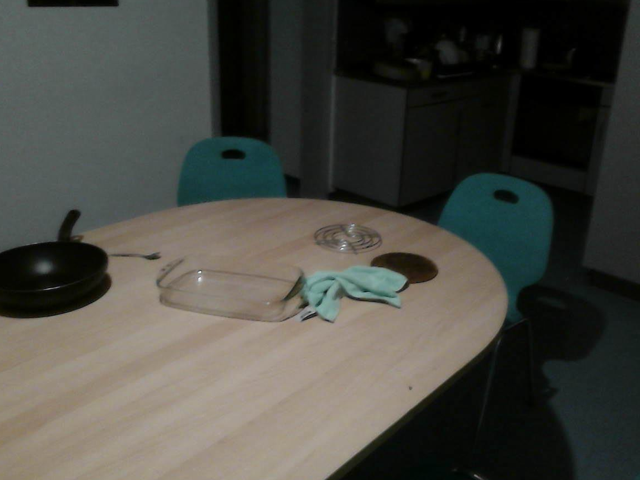} }&  \adjustbox{valign=c}{\includegraphics[height=40pt]{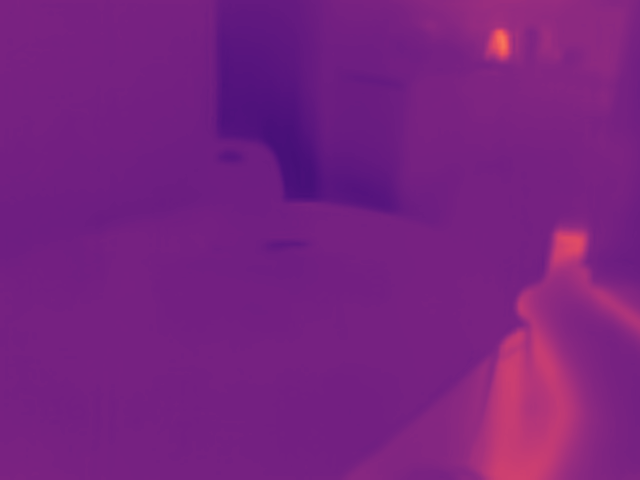} }&  \adjustbox{valign=c}{\includegraphics[height=40pt]{images/OursDataset/different_lighting/messy-living-room_pose.png} }&  \makecell{52} &  \makecell{{10.5}$^\circ$C \\ {36.4}$^\circ$C} \\[-9pt]\\
 & False &\adjustbox{valign=c}{\includegraphics[height=40pt]{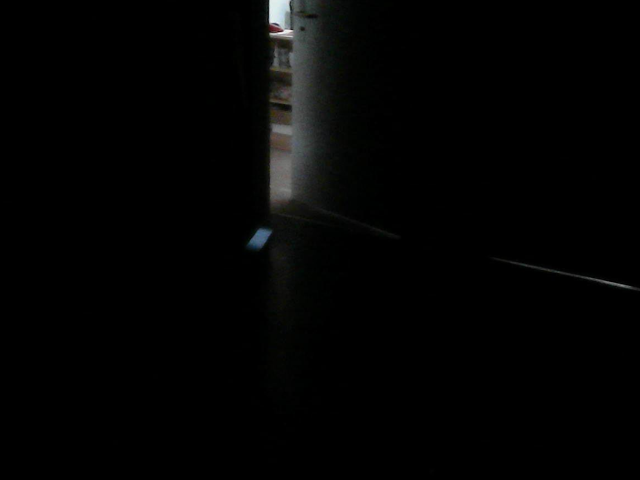} }&  \adjustbox{valign=c}{\includegraphics[height=40pt]{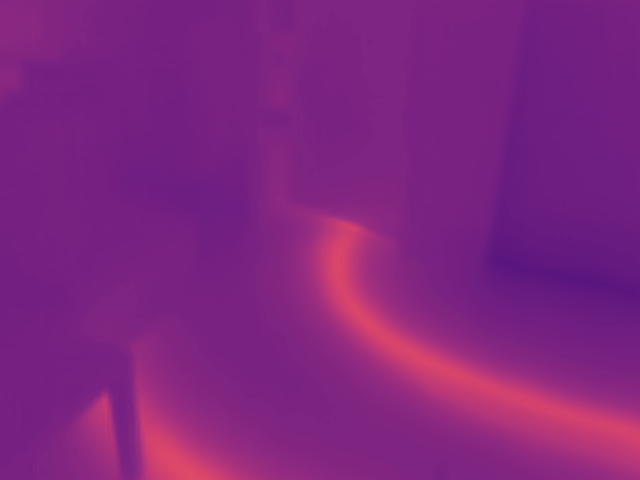} }&  \adjustbox{valign=c}{\includegraphics[height=40pt]{images/OursDataset/different_lighting/messy-living-room_pose.png} }&  \makecell{49} &  \makecell{{10.2}$^\circ$C \\ {38.8}$^\circ$C} \\ \bottomrule
    \end{tabular}
    }
    
\end{table}

%% file: supplementary/4_metrics_colmap.tex
\section{Using COLMAP to Estimate Ground Truth Camera Poses}\label{sec:colmap_as_gt}

\input{tables/supp_colmap_metrics}

While our work prioritizes improving VGGT’s multimodal reconstruction performance rather than its general pose estimation capabilities, we provide camera estimation metrics.
As stated in the paper, we benchmark against VGGT-derived poses---except for \orebrodataset Dataset, where motion-capture produces camera poses.
For each scene of N image pairs, we use $N$ RGB-only images for ground truth pose estimation, and $\frac{1}{2} N$ RGB and $\frac{1}{2} N$ thermal images in the test phase.
This approach guarantees both ground truth accuracy and sufficient differences between the test and ground truth sets.

We selected VGGT over alternatives like COLMAP for two key reasons: 1) VGGT authors' reported higher accuracy of their model on camera pose estimation, and 2)  empirical limitations of COLMAP.
For example, COLMAP estimated poses for the \texttt{INR-building} scene (ThermoScenes) are misaligned (visual inspection showed that many cameras are incorrectly oriented).
However, to avoid possible bias due to the use of VGGT for ground truth estimation, we include here comparative results using COLMAP-derived ground truth where available (i.e., for ThermoScenes~\cite{hassanThermoNeRFMultimodalNeural2025}, ThermalMix~\cite{thermalmix}, and ThermalGaussian~\cite{luThermalGaussianThermal3D2024a})---
per-scene results are provided in \cref{tab:supp_thermoscenes_colmap_gt}, \cref{tab:supp_thermalgaussian_colmap_gt}, and \cref{tab:supp_thermaxlmix_colmap_gt}.

Results on COLMAP-estimated camera poses are similar to those obtained using VGGT-estimated poses in the main paper: according to the quantitative metrics, \ours achieves improved multimodal geometry reconstruction compared with the baselines.
Our conclusions are consistent across different ground truth poses.

%% file: tables/supp_colmap_metrics.tex
\begin{table}[t]
    \centering
    
    \caption{
    Quantitative comparison of 3D reconstruction on \textbf{ThermoScenes}~\cite{hassanThermoNeRFMultimodalNeural2025} if COLMAP is used to estimate ground truth camera poses.
    }
    \label{tab:supp_thermoscenes_colmap_gt}

\resizebox{0.9\linewidth}{!}{
\begin{tabular}{l ccc cc@{\hspace{8pt}}ccc cc}
\toprule
& \multicolumn{5}{c}{prpt-cup}& \multicolumn{5}{c}{INR-building} \\\cmidrule(r){2-6} \cmidrule(l){7-11}

Method
& AUC $\uparrow$
& RRA $\uparrow$
& RTA $\uparrow$
& Reg (\%) $\uparrow$
& FPS $\uparrow$

& AUC $\uparrow$
& RRA $\uparrow$
& RTA $\uparrow$
& Reg (\%) $\uparrow$
& FPS $\uparrow$\\

\makecell[l]{COLMAP +\\SPSG} &  49.4 & 73.7 & 75.3 & \cellcolor{secondbest}{51.2} & 0.40 &  \cellcolor{best}{34.9} & \cellcolor{secondbest}{52.3} & \cellcolor{best}{57.6} &  \cellcolor{secondbest}{49.7} & 0.49 \\ \cmidrule(r){1-6}  \cmidrule(l){7-11}
$\text{MA}_{\text{ELoFTR}}$ &  16.3 & 70.0 & 45.0 &  4.0 & 0.18 &  2.6 & \cellcolor{best}{61.6} & 13.6 &  15.6 & 0.19 \\
$\text{MINIMA}_{\text{ROMA}}$ &  8.4 & 26.6 & 31.3 &  \cellcolor{best}{100.0} & 0.02 &  15.3 & 48.2 & 42.0 &  \cellcolor{best}{100.0} & 0.03 \\ 
MP-SfM & \cellcolor{secondbest}{59.4} & \cellcolor{secondbest}{93.7} & \cellcolor{secondbest}{86.6} &  \cellcolor{best}{100.0} & 0.04 &19.7 & 48.2 & 48.3 &  \cellcolor{best}{100.0} & 0.05\\
\cmidrule(r){1-6}  \cmidrule(l){7-11}
DUSt3R &  19.9 & 29.2 & 32.9 &  \cellcolor{best}{100.0} & 0.43 &  6.5 & 44.6 & 33.9 &  \cellcolor{best}{100.0} & 0.66 \\
MASt3R &  22.5 & 54.9 & 44.3 &  \cellcolor{best}{100.0} & 0.14 &  3.7 & 24.4 & 24.7 &  \cellcolor{best}{100.0} & 0.25 \\ \cmidrule(r){1-6}  \cmidrule(l){7-11}
VGGT &  15.9 & 59.0 & 37.2 &  \cellcolor{best}{100.0} & \cellcolor{best}{4.09} &  9.4 & 47.7 & 34.5 &  \cellcolor{best}{100.0} & \cellcolor{best}{4.91} \\
MapAnything & 14.7 & 44.1 & 36.5 &  \cellcolor{best}{100.0} & 0.56 &12.8 & 47.7 & 42.0 &  \cellcolor{best}{100.0} & 1.00\\
\ours &  \cellcolor{best}{82.4} & \cellcolor{best}{100.0} & \cellcolor{best}{97.8} &  \cellcolor{best}{100.0} & \cellcolor{secondbest}{2.65} &  \cellcolor{secondbest}{20.4} & 48.2 & \cellcolor{secondbest}{48.6} &  \cellcolor{best}{100.0} & \cellcolor{secondbest}{4.79} \\
\bottomrule
\end{tabular}}

\resizebox{0.9\linewidth}{!}{
\begin{tabular}{l ccc cc@{\hspace{8pt}}ccc cc}
\toprule
& \multicolumn{5}{c}{reflect-robot}& \multicolumn{5}{c}{melting\_ice\_cup} \\\cmidrule(r){2-6} \cmidrule(l){7-11}

Method
& AUC $\uparrow$
& RRA $\uparrow$
& RTA $\uparrow$
& Reg (\%) $\uparrow$
& FPS $\uparrow$

& AUC $\uparrow$
& RRA $\uparrow$
& RTA $\uparrow$
& Reg (\%) $\uparrow$
& FPS $\uparrow$\\

\makecell[l]{COLMAP +\\SPSG} &  6.2 & 66.3 & 28.4 &  \cellcolor{secondbest}{29.0} & 0.54 &  0.7 & 13.1 & 21.6 &  \cellcolor{secondbest}{25.8} & 0.56 \\ \cmidrule(r){1-6}  \cmidrule(l){7-11}
$\text{MA}_{\text{ELoFTR}}$ &  16.8 & 79.7 & 44.8 &  11.3 & 0.16 &  4.1 & 26.6 & 22.3 &  8.8 & 0.28 \\
$\text{MINIMA}_{\text{ROMA}}$ &  \cellcolor{secondbest}{60.2} & \cellcolor{secondbest}{89.8} & \cellcolor{secondbest}{85.6} &  \cellcolor{best}{100.0} & 0.04 &  6.2 & 16.9 & 21.7 &  \cellcolor{best}{100.0} & 0.03 \\ 
MP-SfM & 21.2 & 69.0 & 45.6 &  \cellcolor{best}{100.0} & 0.03 &7.8 & 18.1 & 28.0 &  \cellcolor{best}{100.0} & 0.05\\
\cmidrule(r){1-6}  \cmidrule(l){7-11}
DUSt3R &  13.4 & 37.1 & 29.2 &  \cellcolor{best}{100.0} & 0.66 &  17.1 & \cellcolor{secondbest}{39.6} & 40.7 &  \cellcolor{best}{100.0} & 0.68 \\
MASt3R &  23.9 & 79.9 & 33.2 &  \cellcolor{best}{100.0} & 0.24 &  \cellcolor{secondbest}{25.3} & 36.2 & \cellcolor{secondbest}{49.0} &  \cellcolor{best}{100.0} & 0.24 \\ \cmidrule(r){1-6}  \cmidrule(l){7-11}
VGGT &  16.6 & 45.3 & 38.4 &  \cellcolor{best}{100.0} & \cellcolor{best}{4.94} &  19.4 & 38.4 & 42.0 &  \cellcolor{best}{100.0} & \cellcolor{best}{6.81} \\
MapAnything & 17.2 & 65.1 & 31.2 &  \cellcolor{best}{100.0} & 1.00 &8.1 & 26.7 & 30.5 &  \cellcolor{best}{100.0} & 1.37\\
\ours &  \cellcolor{best}{64.5} & \cellcolor{best}{93.9} & \cellcolor{best}{93.2} &  \cellcolor{best}{100.0} & \cellcolor{secondbest}{4.81} &  \cellcolor{best}{35.5} & \cellcolor{best}{49.0} & \cellcolor{best}{64.2} &  \cellcolor{best}{100.0} & \cellcolor{secondbest}{6.59} \\
\bottomrule
\end{tabular}}

\resizebox{0.45\linewidth}{!}{
\begin{tabular}{l ccc cc}
\toprule
& \multicolumn{5}{c}{freezing\_ice\_cup} \\\cmidrule(r){2-6}

Method
& AUC $\uparrow$
& RRA $\uparrow$
& RTA $\uparrow$
& Reg (\%) $\uparrow$
& FPS $\uparrow$\\

\makecell[l]{COLMAP +\\SPSG} &  \cellcolor{best}{38.6} & \cellcolor{best}{57.1} & \cellcolor{best}{63.4} &  \cellcolor{secondbest}{50.0} & 0.52 \\ \cmidrule(r){1-6}
$\text{MA}_{\text{ELoFTR}}$ &  4.3 & 30.6 & 40.5 &  7.6 & 0.33 \\
$\text{MINIMA}_{\text{ROMA}}$ &  6.1 & 12.1 & 26.5 &  \cellcolor{best}{100.0} & 0.04 \\ 
MP-SfM & \cellcolor{secondbest}{19.1} & \cellcolor{secondbest}{39.1} & \cellcolor{secondbest}{54.3} &  \cellcolor{best}{100.0} & 0.04 \\
\cmidrule(r){1-6}
DUSt3R &  15.0 & 29.9 & 42.4 &  \cellcolor{best}{100.0} & 0.66 \\
MASt3R &  {18.6} & {38.6} & 40.8 &  \cellcolor{best}{100.0} & 0.22 \\ \cmidrule(r){1-6}
VGGT &  10.8 & 26.6 & 34.2 &  \cellcolor{best}{100.0} & \cellcolor{best}{4.89} \\
MapAnything & 9.0 & 18.9 & 28.7 &  \cellcolor{best}{100.0} & 1.00 \\
\ours &  18.3 & 33.5 & {45.8} &  \cellcolor{best}{100.0} & \cellcolor{secondbest}{4.76} \\
\bottomrule
\end{tabular}}

\end{table}

\begin{table}[t]
    \centering
    \caption{
    Quantitative comparison of 3D reconstruction on \textbf{ThermalMix}~\cite{thermalmix} if COLMAP is used to estimate ground truth camera poses.
    }
    \label{tab:supp_thermaxlmix_colmap_gt}
    
\resizebox{\linewidth}{!}{
\begin{tabular}{l ccc cc@{\hspace{8pt}}ccc cc}
\toprule
& \multicolumn{5}{c}{laptop}& \multicolumn{5}{c}{panel} \\\cmidrule(r){2-6} \cmidrule(l){7-11}

Method
& AUC $\uparrow$
& RRA $\uparrow$
& RTA $\uparrow$
& Reg (\%) $\uparrow$
& FPS $\uparrow$

& AUC $\uparrow$
& RRA $\uparrow$
& RTA $\uparrow$
& Reg (\%) $\uparrow$
& FPS $\uparrow$\\

\makecell[l]{COLMAP +\\SPSG} &  56.6 & \cellcolor{best}{100.0} & 90.9 &  10.8 & 0.28 &  17.6 & 32.2 & 48.3 &  28.0 & 0.32 \\ \cmidrule(r){1-6}  \cmidrule(l){7-11}
$\text{MA}_{\text{ELoFTR}}$ &  23.2 & 51.1 & 61.5 &  \cellcolor{secondbest}{48.9} & 0.15 &  46.0 & 73.6 & \cellcolor{secondbest}{98.1} &  \cellcolor{secondbest}{35.4} & 0.36 \\
$\text{MINIMA}_{\text{ROMA}}$ &  \cellcolor{secondbest}{78.5} & \cellcolor{secondbest}{94.8} & \cellcolor{secondbest}{94.3} &  \cellcolor{best}{100.0} & 0.04 &  \cellcolor{secondbest}{78.4} & \cellcolor{secondbest}{95.2} & 93.2 &  \cellcolor{best}{100.0} & 0.05 \\ 
MP-SfM & 47.3 & 64.3 & 65.8 &  \cellcolor{best}{100.0} & 0.04 &73.5 & 100.0 & 98.0 &  \cellcolor{best}{100.0} & 0.06\\
\cmidrule(r){1-6}  \cmidrule(l){7-11}
DUSt3R &  21.1 & 44.1 & 46.1 &  \cellcolor{best}{100.0} & 0.67 &  15.5 & 70.4 & 39.4 &  \cellcolor{best}{100.0} & 0.65 \\
MASt3R &  30.0 & 62.5 & 74.8 &  \cellcolor{best}{100.0} & 0.23 &  35.7 & 67.1 & 68.2 &  \cellcolor{best}{100.0} & 0.23 \\ \cmidrule(r){1-6}  \cmidrule(l){7-11}
VGGT &  15.3 & 30.8 & 41.3 &  \cellcolor{best}{100.0} & \cellcolor{best}{11.23} &  12.7 & 28.5 & 45.2 &  \cellcolor{best}{100.0} & \cellcolor{best}{17.04} \\
MapAnything & 11.6 & 27.2 & 34.4 &  \cellcolor{best}{100.0} & 2.22 &16.1 & 43.7 & 60.1 &  \cellcolor{best}{100.0} & 3.25\\
\ours &  \cellcolor{best}{90.5} & \cellcolor{best}{100.0} & \cellcolor{best}{99.5} &  \cellcolor{best}{100.0} & \cellcolor{secondbest}{10.70} &  \cellcolor{best}{82.2} & \cellcolor{best}{100.0} & \cellcolor{best}{99.3} &  \cellcolor{best}{100.0} & \cellcolor{secondbest}{15.72} \\
\bottomrule
\end{tabular}}
    
\end{table}

\begin{table}[t]
    \centering
     \caption{
    Quantitative comparison of 3D reconstruction on \textbf{ThermalGaussian}~\cite{luThermalGaussianThermal3D2024a} if COLMAP on RGB is ground truth.
    }
    \label{tab:supp_thermalgaussian_colmap_gt}
    
\resizebox{\linewidth}{!}{
\begin{tabular}{l ccc cc@{\hspace{8pt}}ccc cc}
\toprule
& \multicolumn{5}{c}{IronIngot}& \multicolumn{5}{c}{Parterre} \\\cmidrule(r){2-6} \cmidrule(l){7-11}

Method
& AUC $\uparrow$
& RRA $\uparrow$
& RTA $\uparrow$
& Reg (\%) $\uparrow$
& FPS $\uparrow$

& AUC $\uparrow$
& RRA $\uparrow$
& RTA $\uparrow$
& Reg (\%) $\uparrow$
& FPS $\uparrow$\\

\makecell[l]{COLMAP +\\SPSG} &  \cellcolor{secondbest}{96.0} & \cellcolor{best}{100.0} & \cellcolor{secondbest}{99.7} &  \cellcolor{best}{100.0} & 0.54 &  46.0 & \cellcolor{best}{100.0} & \cellcolor{best}{100.0} &  3.0 & 0.53 \\ \cmidrule(r){1-6}  \cmidrule(l){7-11}
$\text{MA}_{\text{ELoFTR}}$ &  13.7 & 56.0 & 58.5 &  \cellcolor{secondbest}{61.5} & 0.12 &  32.0 & 66.8 & 85.5 &  \cellcolor{secondbest}{24.2} & 0.23 \\
$\text{MINIMA}_{\text{ROMA}}$ &  \cellcolor{best}{97.4} & \cellcolor{best}{100.0} & \cellcolor{best}{99.9} &  \cellcolor{best}{100.0} & 0.06 &  34.4 & 47.5 & 64.1 &  \cellcolor{best}{100.0} & 0.04 \\ 
MP-SfM & 95.2 & \cellcolor{best}{100.0} & 99.6 &  \cellcolor{best}{100.0} & 0.06 &26.8 & 44.1 & 64.0 &  \cellcolor{best}{100.0} & 0.02\\
\cmidrule(r){1-6}  \cmidrule(l){7-11}
DUSt3R &  15.5 & \cellcolor{secondbest}{75.0} & 39.6 &  \cellcolor{best}{100.0} & 0.66 &  43.1 & \cellcolor{secondbest}{95.3} & 79.5 &  \cellcolor{best}{100.0} & 0.65 \\
MASt3R &  74.2 & \cellcolor{best}{100.0} & 96.7 &  \cellcolor{best}{100.0} & 0.23 &  \cellcolor{secondbest}{86.0} & \cellcolor{best}{100.0} & \cellcolor{secondbest}{98.8} &  \cellcolor{best}{100.0} & 0.22 \\ \cmidrule(r){1-6}  \cmidrule(l){7-11}
VGGT &  31.8 & 68.5 & 69.8 &  \cellcolor{best}{100.0} & \cellcolor{best}{14.13} &  65.5 & 87.2 & 85.9 &  \cellcolor{best}{100.0} & \cellcolor{best}{13.81} \\
MapAnything & 46.1 & 97.2 & 76.5 &  \cellcolor{best}{100.0} & 2.76 &33.8 & 69.7 & 67.1 &  \cellcolor{best}{100.0} & 2.66\\
\ours &  87.2 & \cellcolor{best}{100.0} & 98.1 &  \cellcolor{best}{100.0} & \cellcolor{secondbest}{13.34} &  \cellcolor{best}{94.6} & \cellcolor{best}{100.0} & \cellcolor{best}{100.0} &  \cellcolor{best}{100.0} & \cellcolor{secondbest}{13.01} \\
\bottomrule
\end{tabular}}

\scalebox{0.9}{
\begin{tabular}{l ccc cc}
\toprule
& \multicolumn{5}{c}{Ebike} \\\cmidrule(r){2-6}

Method
& AUC $\uparrow$
& RRA $\uparrow$
& RTA $\uparrow$
& Reg (\%) $\uparrow$
& FPS $\uparrow$\\

\makecell[l]{COLMAP +\\SPSG} &  65.0 & 70.2 & 70.4 &  \cellcolor{secondbest}{95.8} & 0.60 \\ \cmidrule(r){1-6}
$\text{MA}_{\text{ELoFTR}}$ &  44.6 & \cellcolor{secondbest}{77.8} & 66.7 &  7.3 & 0.38 \\
$\text{MINIMA}_{\text{ROMA}}$ &  \cellcolor{secondbest}{94.3} & \cellcolor{best}{100.0} & \cellcolor{best}{100.0} &  \cellcolor{best}{100.0} & 0.07 \\ 
MP-SfM & 93.9 & \cellcolor{best}{100.0} & 99.8 &  \cellcolor{best}{100.0} & 0.06 \\
\cmidrule(r){1-6}
DUSt3R &  20.0 & 29.7 & 35.9 &  \cellcolor{best}{100.0} & 0.65 \\
MASt3R &  90.3 & \cellcolor{best}{100.0} & \cellcolor{secondbest}{99.5} &  \cellcolor{best}{100.0} & 0.24 \\ \cmidrule(r){1-6}
VGGT &  83.9 & \cellcolor{best}{100.0} & 98.0 &  \cellcolor{best}{100.0} & \cellcolor{best}{16.06} \\
MapAnything & 14.6 & 29.4 & 41.3 &  \cellcolor{best}{100.0} & 3.06 \\
\ours &  \cellcolor{best}{95.2} & \cellcolor{best}{100.0} & \cellcolor{best}{100.0} &  \cellcolor{best}{100.0} & \cellcolor{secondbest}{15.07} \\
\bottomrule
\end{tabular}}
   
\end{table}

%% file: supplementary/5_more_visualization.tex
\section{Additional Visual Comparisons}

\begin{figure}[!t]
  \centering
  \vspace{3mm}
  \begin{overpic}[width=0.95\linewidth]{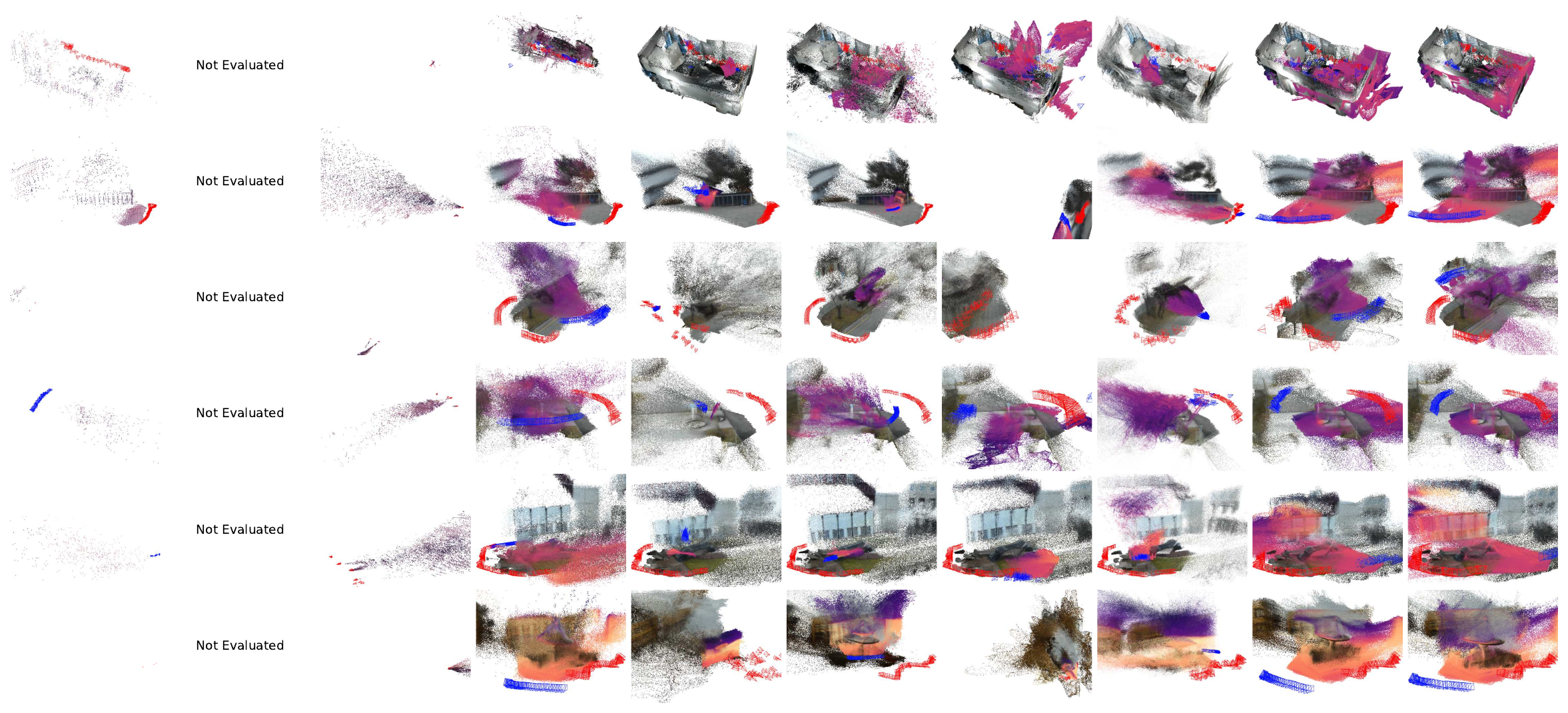}
    \put(2, 45){\scalebox{0.5}{COLMAP}}
    \put(12, 45){\scalebox{0.5}{MP-SfM}}
    \put(22, 45){\scalebox{0.5}{$\text{MA}_\text{ELoFTR}$}}
    \put(31, 45){\scalebox{0.5}{$\text{MINIMA}_\text{ROMA}$}}
    \put(42, 45){\scalebox{0.5}{DUSt3R}}
    \put(53, 45){\scalebox{0.5}{MASt3R}}
    \put(60, 45){\scalebox{0.5}{MapAnything}}
    \put(72, 45){\scalebox{0.5}{VGGT}}
    \put(82, 45){\scalebox{0.5}{\shortstack{\ours\\(ours)}}}
    \put(92, 45){\scalebox{0.5}{\shortstack{Ground\\Truth}}}
    \put(-2, 40){\scalebox{0.4}{\rotatebox{90}{\shortstack{conferen\\ce-room}}}}
    \put(-2, 32){\scalebox{0.4}{\rotatebox{90}{\shortstack{metallic-\\container}}}}
    \put(-2, 23){\scalebox{0.4}{\rotatebox{90}{\shortstack{old-drin-\\king-fountain}}}}
    \put(-2, 15){\scalebox{0.4}{\rotatebox{90}{parking}}}
    \put(-2, 10){\scalebox{0.4}{\rotatebox{90}{statue}}}
    \put(-2, 2){\scalebox{0.4}{\rotatebox{90}{telescope}}}
  \end{overpic}
  \caption{
    Qualitative results on \datasetname. Our method produces more accurate and structurally consistent point clouds than the baselines. All methods show limited performance on \texttt{conference-room} and \texttt{old-drinking-fountain}, likely because the thermal modality in these scenes provides weak localization cues.
  }
  \label{fig:vis_more_oursdataset}
\end{figure}

\begin{figure}[!t]
  \centering
  \begin{overpic}[width=0.95\linewidth]{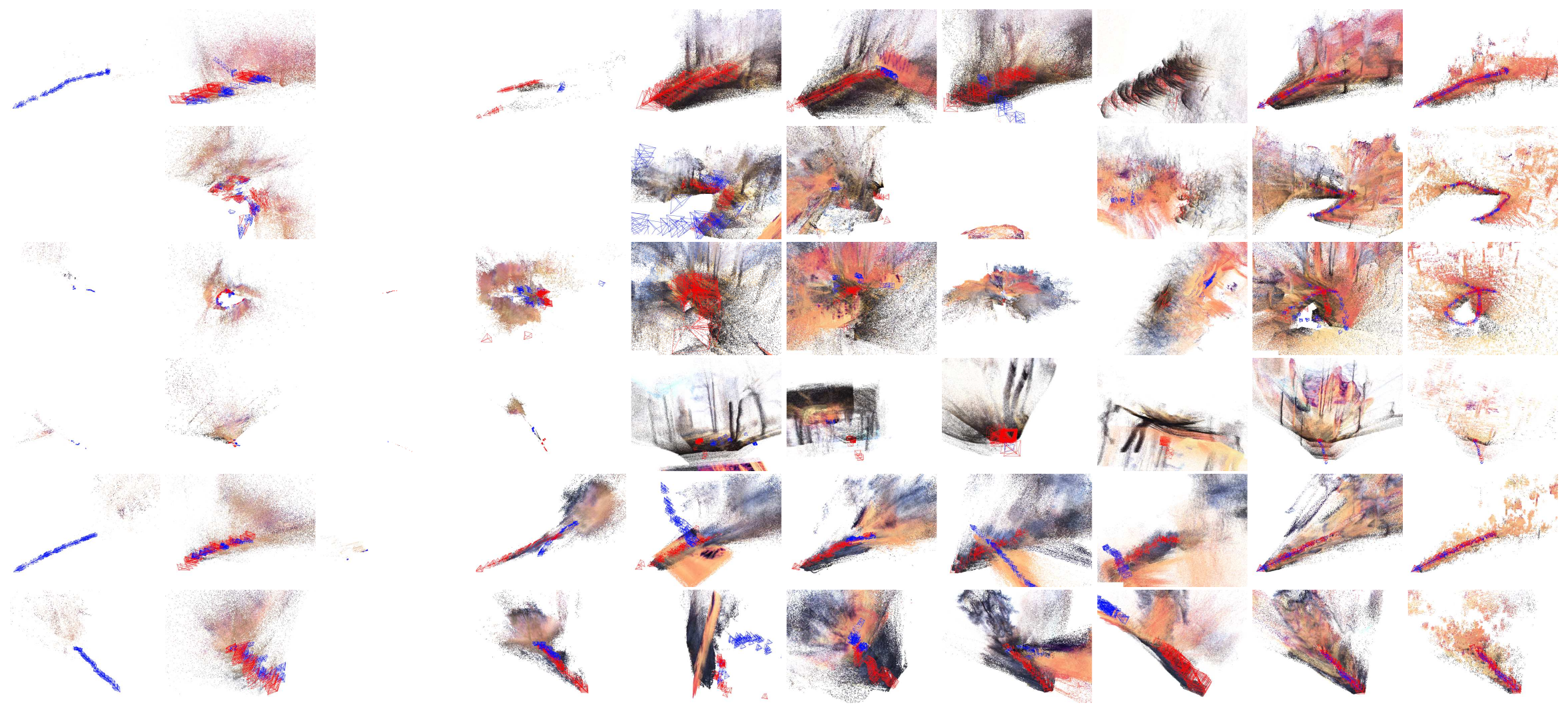}
    \put(2, 45){\scalebox{0.5}{COLMAP}}
    \put(12, 45){\scalebox{0.5}{MP-SfM}}
    \put(22, 45){\scalebox{0.5}{$\text{MA}_\text{ELoFTR}$}}
    \put(31, 45){\scalebox{0.5}{$\text{MINIMA}_\text{ROMA}$}}
    \put(42, 45){\scalebox{0.5}{DUSt3R}}
    \put(53, 45){\scalebox{0.5}{MASt3R}}
    \put(60, 45){\scalebox{0.5}{MapAnything}}
    \put(72, 45){\scalebox{0.5}{VGGT}}
    \put(82, 45){\scalebox{0.5}{\shortstack{\ours\\(ours)}}}
    \put(92, 45){\scalebox{0.5}{\shortstack{Ground\\Truth}}}

  \end{overpic}
  \caption{
    Qualitative results on the \orebrodataset dataset. Our method produces more accurate point clouds than the baselines. All methods show degraded performance on \texttt{01\_Annexet\_No\_Radars\_1} and \texttt{01\_Annexet\_No\_Radars\_2}, likely due to the complex camera motion in these sequences.
    From top to bottom the scenes are \texttt{01\_Annexet\_No\_Radars\_0}, \texttt{01\_Annexet\_No\_Radars\_1}, \texttt{01\_Annexet\_No\_Radars\_2}, \texttt{01\_Annexet\_No\_Radars\_3}, \texttt{04\_Forest\_pass\_no\_radars\_0}, and \texttt{04\_Forest\_pass\_no\_radars\_1}.
  }
  \label{fig:vis_more_oru}
\end{figure}

\begin{figure}[!t]
  \centering
  \vspace{3mm}
  \begin{overpic}[width=0.95\linewidth]{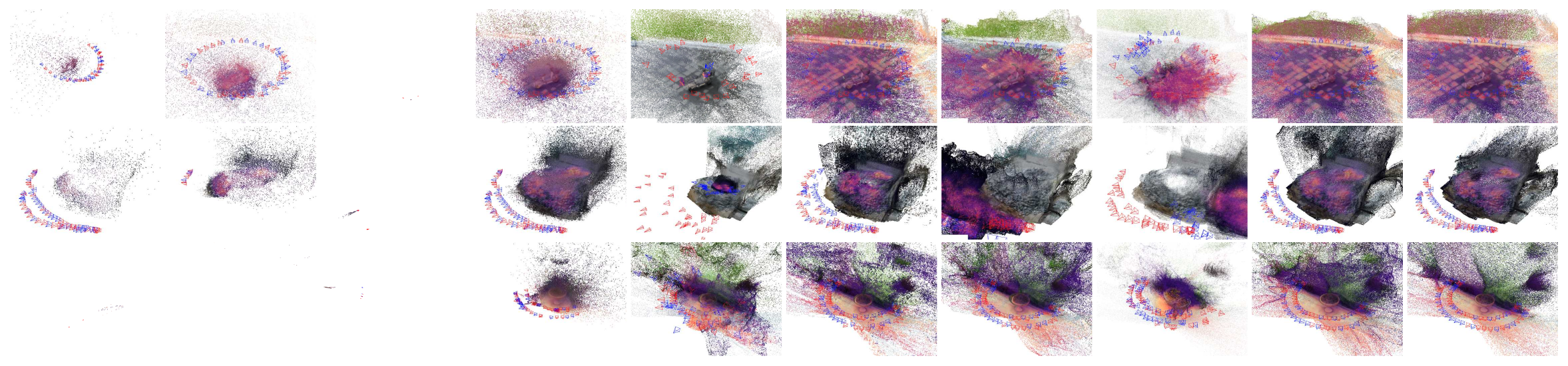}
    \put(2, 23){\scalebox{0.5}{COLMAP}}
    \put(12, 23){\scalebox{0.5}{MP-SfM}}
    \put(22, 23){\scalebox{0.5}{$\text{MA}_\text{ELoFTR}$}}
    \put(31, 23){\scalebox{0.5}{$\text{MINIMA}_\text{ROMA}$}}
    \put(42, 23){\scalebox{0.5}{DUSt3R}}
    \put(53, 23){\scalebox{0.5}{MASt3R}}
    \put(60, 23){\scalebox{0.5}{MapAnything}}
    \put(72, 23){\scalebox{0.5}{VGGT}}
    \put(82, 23){\scalebox{0.5}{\shortstack{\ours\\(ours)}}}
    \put(92, 23){\scalebox{0.5}{\shortstack{Ground\\Truth}}}
    \put(-2, 18){\scalebox{0.5}{\rotatebox{90}{Ebike}}}
    \put(-2, 9){\scalebox{0.5}{\rotatebox{90}{IronIngot}}}
    \put(-2, 1){\scalebox{0.5}{\rotatebox{90}{Parterre}}}
  \end{overpic}
  \caption{
    Qualitative results on the ThermalGaussian dataset. Our method produces more accurate point clouds than the baselines on \texttt{Ebike} and \texttt{Parterre}. For \texttt{Iron Ingot}, our reconstruction is visually highly accurate, although its quantitative metrics are lower than those of COLMAP and $\text{MINIMA}_\text{ROMA}$.
  }
  \label{fig:vis_more_thermalgaussian}
\end{figure}

\begin{figure}[!t]
  \centering
  \begin{overpic}[width=0.95\linewidth]{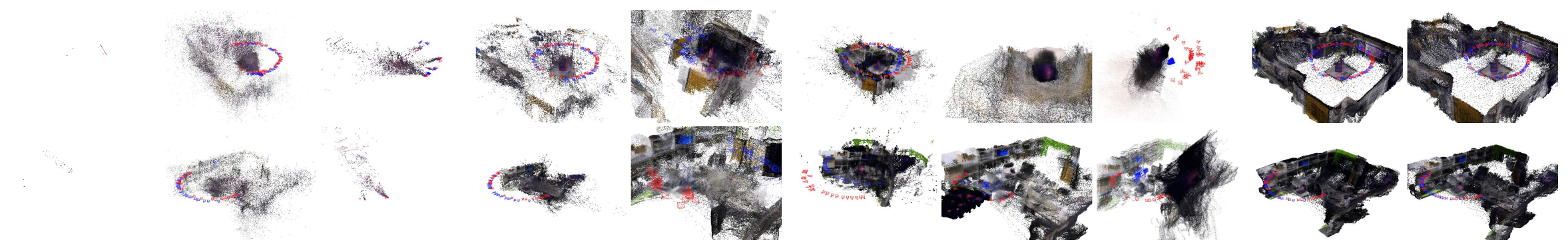}
    \put(2, 16){\scalebox{0.5}{COLMAP}}
    \put(12, 16){\scalebox{0.5}{MP-SfM}}
    \put(22, 16){\scalebox{0.5}{$\text{MA}_\text{ELoFTR}$}}
    \put(31, 16){\scalebox{0.5}{$\text{MINIMA}_\text{ROMA}$}}
    \put(42, 16){\scalebox{0.5}{DUSt3R}}
    \put(53, 16){\scalebox{0.5}{MASt3R}}
    \put(60, 16){\scalebox{0.5}{MapAnything}}
    \put(72, 16){\scalebox{0.5}{VGGT}}
    \put(82, 16){\scalebox{0.5}{\shortstack{\ours\\(ours)}}}
    \put(92, 16){\scalebox{0.5}{\shortstack{Ground\\Truth}}}
    \put(-2, 10){\scalebox{0.5}{\rotatebox{90}{laptop}}}
    \put(-2, 1){\scalebox{0.5}{\rotatebox{90}{panel}}}
  \end{overpic}
  \caption{
    Qualitative results on the ThermalMix dataset. Our method produces more accurate and complete point clouds than the baselines.
  }
  \label{fig:vis_more_thermalmix}
\end{figure}

\begin{figure}[!t]
  \centering
  \begin{overpic}[width=0.95\linewidth]{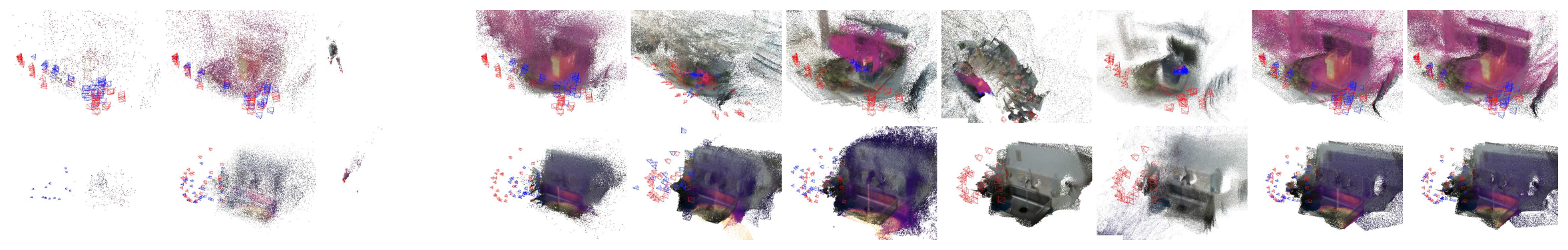}
    \put(2, 16){\scalebox{0.5}{COLMAP}}
    \put(12, 16){\scalebox{0.5}{MP-SfM}}
    \put(22, 16){\scalebox{0.5}{$\text{MA}_\text{ELoFTR}$}}
    \put(31, 16){\scalebox{0.5}{$\text{MINIMA}_\text{ROMA}$}}
    \put(42, 16){\scalebox{0.5}{DUSt3R}}
    \put(53, 16){\scalebox{0.5}{MASt3R}}
    \put(60, 16){\scalebox{0.5}{MapAnything}}
    \put(72, 16){\scalebox{0.5}{VGGT}}
    \put(82, 16){\scalebox{0.5}{\shortstack{\ours\\(ours)}}}
    \put(92, 16){\scalebox{0.5}{\shortstack{Ground\\Truth}}}
    \put(-2, 8){\scalebox{0.5}{\rotatebox{90}{generator}}}
    \put(-2, 3){\scalebox{0.5}{\rotatebox{90}{sink}}}
  \end{overpic}
  \caption{
    Qualitative results on the ThermalNeRF dataset. Our method produces more accurate and structurally coherent point clouds than the baselines.
  }
  \label{fig:vis_more_thermalnerf}
\end{figure}

\begin{figure}[!t]
  \centering
  \vspace{10pt}
  \begin{overpic}[width=0.95\linewidth]{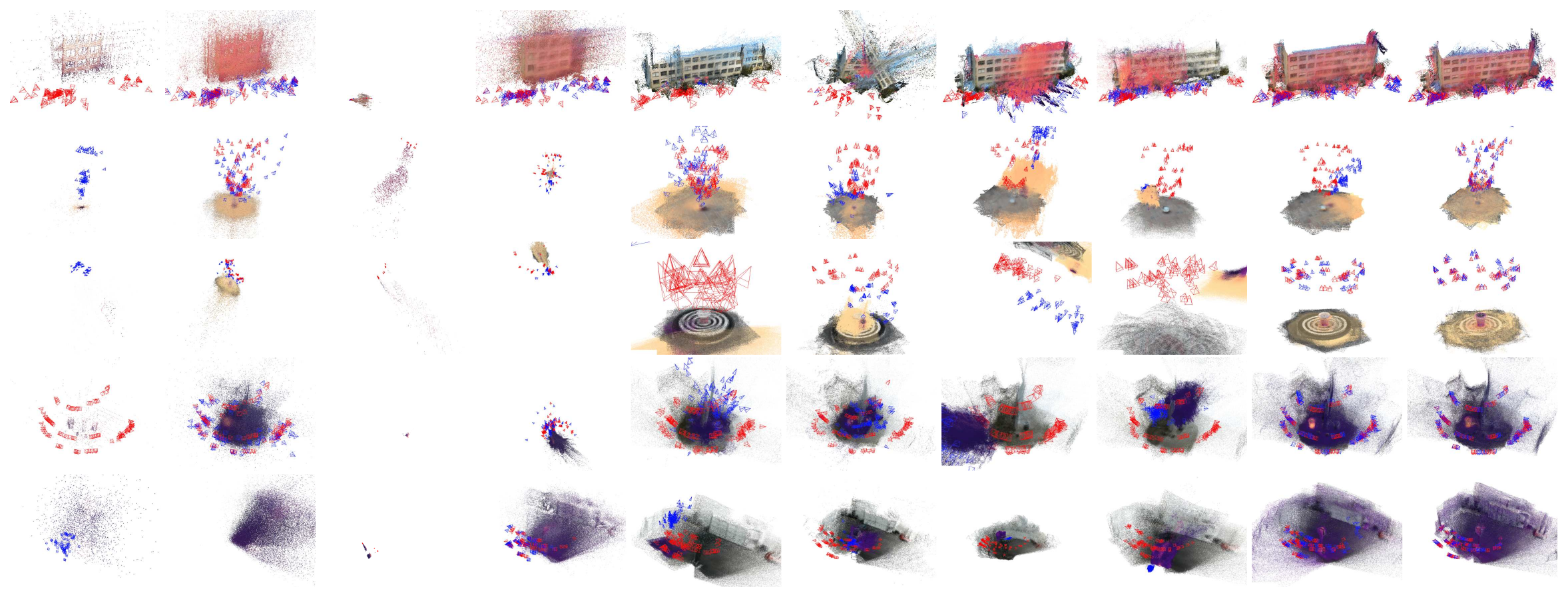}
    \put(2, 38){\scalebox{0.5}{COLMAP}}
    \put(12, 38){\scalebox{0.5}{MP-SfM}}
    \put(22, 38){\scalebox{0.5}{$\text{MA}_\text{ELoFTR}$}}
    \put(31, 38){\scalebox{0.5}{$\text{MINIMA}_\text{ROMA}$}}
    \put(42, 38){\scalebox{0.5}{DUSt3R}}
    \put(53, 38){\scalebox{0.5}{MASt3R}}
    \put(60, 38){\scalebox{0.5}{MapAnything}}
    \put(72, 38){\scalebox{0.5}{VGGT}}
    \put(82, 38){\scalebox{0.5}{\shortstack{\ours\\(ours)}}}
    \put(92, 38){\scalebox{0.5}{\shortstack{Ground\\Truth}}}
    \put(-2, 32){\scalebox{0.4}{\rotatebox{90}{INR-building}}}
    \put(-2, 25){\scalebox{0.4}{\rotatebox{90}{\shortstack{freezing\\ice cup}}}}
    \put(-2, 18){\scalebox{0.4}{\rotatebox{90}{\shortstack{melting\\ice cup}}}}
    \put(-2, 10.5){\scalebox{0.4}{\rotatebox{90}{prpt-cup}}}
    \put(-2, 2){\scalebox{0.4}{\rotatebox{90}{reflect-robot}}}
  \end{overpic}
  \caption{
    Qualitative results on the ThermoScenes dataset. Our method produces more accurate point clouds than the baselines. All methods demonstrate limited performance on \texttt{freezing\_ice\_cup}, likely because of the inherent difficulty of this scene, discussed in~\cref{sec:colmap_as_gt}.
  }
  \label{fig:vis_more_thermonerf}
\end{figure}

In this section, we provide additional visual comparisons with the baselines. For each evaluation scene, we randomly select either its first or second validation run. We present the results for \datasetname in \cref{fig:vis_more_oursdataset}, \orebrodataset in \cref{fig:vis_more_oru}, ThermalGaussian in \cref{fig:vis_more_thermalgaussian}, ThermalMix in \cref{fig:vis_more_thermalmix}, ThermalNeRF in \cref{fig:vis_more_thermalnerf}, and ThermoScenes in \cref{fig:vis_more_thermonerf}. COLMAP tends to reconstruct only one modality. $\text{MA}_\text{ELoFTR}$ provides an insufficient number of matches for reliable reconstruction, while $\text{MINIMA}_\text{ROMA}$ often produces noisy reconstructions. DUSt3R, MASt3R, and VGGT tend to reconstruct the two modalities as separate, unaligned point clouds. In contrast, the reconstructions produced by \ours are cleaner, more accurate, and contain fewer artifacts, with camera poses that are also closer to the ground truth.

\begin{figure}[!t]
  \centering
  \vspace{10pt}
  \begin{overpic}[width=0.95\linewidth]{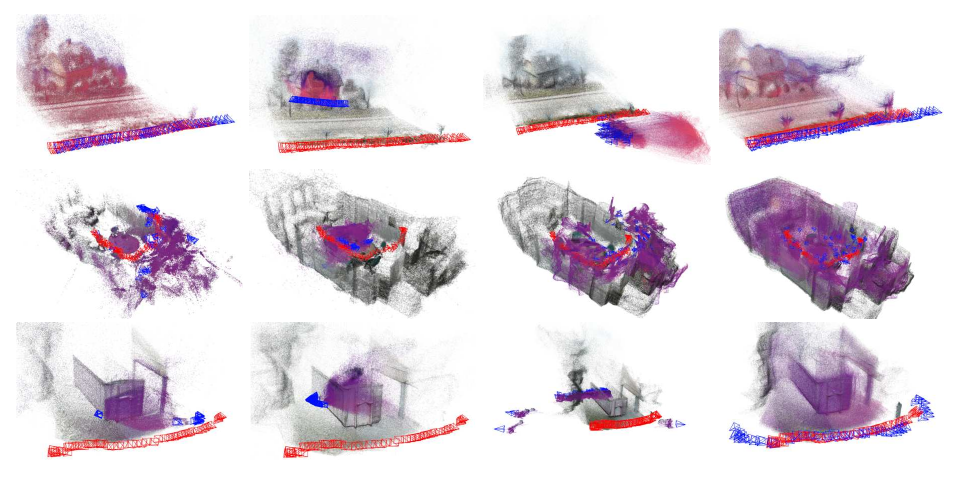}
    \put(5, 52){\scalebox{0.6}{$\text{MINIMA}_\text{ROMA}$}}
    \put(30, 52){\tiny{MASt3R}}
    \put(55, 52){\tiny{VGGT}}
    \put(80, 52){\tiny\shortstack{\ours\\(ours)}}
    \put(-2, 42){\scalebox{0.7}{\rotatebox{90}{house}}}
    \put(-2, 25){\scalebox{0.7}{\rotatebox{90}{\shortstack{messy-liv\\ing-room}}}}
    \put(-2, 7){\scalebox{0.7}{\rotatebox{90}{\shortstack{red-co\\ntainer}}}}
  \end{overpic}
  \caption{
    Qualitative results on \datasetname, where the trajectories were captured under different lighting conditions. Our method produces more accurate point clouds than the baselines. Although $\text{MINIMA}_\text{ROMA}$ achieves competitive performance on \texttt{house}, it fails on the remaining scenes.
  }
  \label{fig:vis_more_oursdataset_vis}
\end{figure}

\begin{figure}[!t]
  \centering
  \vspace{6mm}
  \begin{overpic}[width=0.95\linewidth]{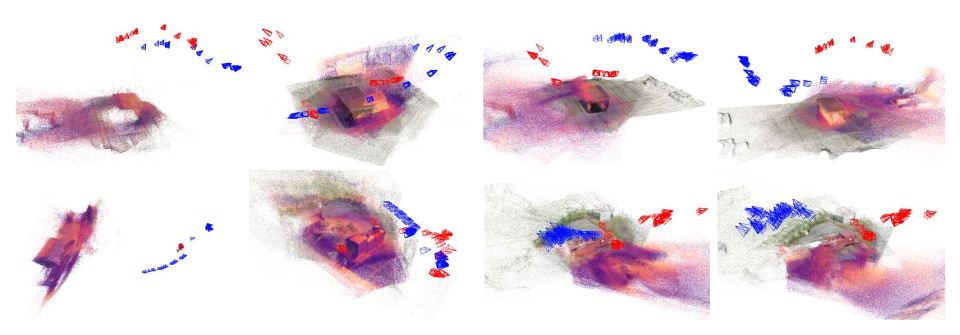}
    \put(5, 35){\scalebox{0.6}{$\text{MINIMA}_\text{ROMA}$}}
    \put(30, 35){\tiny{MASt3R}}
    \put(55, 35){\tiny{VGGT}}
    \put(80, 35){\tiny\shortstack{\ours\\(ours)}}
    \put(-2, 25){\scalebox{0.7}{\rotatebox{90}{\shortstack{bathroom}}}}
    \put(-2, 7){\scalebox{0.7}{\rotatebox{90}{\shortstack{red-co\\ntainer}}}}
  \end{overpic}
  \caption{
    Qualitative results on the SmokeSeer dataset. Our method produces more accurate point clouds than the baselines. None of the methods reconstruct \texttt{red-container}, likely because the overlaid text in the thermal images introduces misleading localization cues.
  }
  \label{fig:vis_more_smokeseer_vis}
\end{figure}

We also provide additional qualitative results for the best-performing methods on the qualitative-only datasets SmokeSeer in \cref{fig:vis_more_smokeseer_vis} and \datasetname (with trajectories captured at different times) in \cref{fig:vis_more_oursdataset_vis}. The behavior of our method and the baselines is consistent with the observations discussed in the previous paragraph. On SmokeSeer, all methods fail to reconstruct \texttt{red-container}. A possible reason is that the thermal images contain overlaid printed statistics, which may negatively affect the models since the text regions can be erroneously interpreted as localization features.

%% file: supplementary/6_per_scene_table.tex
\section{Per-Scene Comparison}

In this section, we present per-scene results of our method and the baselines. We present the results separately for \datasetname in \cref{tab:per_scene_oursdataset}, \orebrodataset in \cref{tab:per_scene_rlt}, ThermalGaussian in \cref{tab:per_scene_thermalgaussian}, ThermalMix in \cref{tab:per_scene_thermalmix}, ThermalNeRF in \cref{tab:per_scene_thermalnerf}, and ThermoScenes in \cref{tab:per_scene_thermoscenes}.

The per-scene results confirm the conclusions of the main evaluation. \ours is the only method that consistently achieves both high pose accuracy and point cloud quality across a wide range of scenes, which translates into reliable multimodal reconstruction. Although COLMAP obtains strong scores on some scenes, these results are often artificially inflated, since COLMAP frequently reconstructs only one modality and registers only a subset of the frames. Likewise, $\text{MA}_{\text{ELoFTR}}$ usually provides too few matches for robust estimation, resulting in low registration rates and weak 3D reconstruction.

Among the remaining baselines, $\text{MINIMA}_{\text{ROMA}}$ is the strongest competitor, but it still produces noisier and less stable reconstructions than \ours. DUSt3R, MASt3R, and VGGT are not designed for RGB-T reconstruction and tend to recover separate, unaligned point clouds for the two modalities instead of a single coherent scene geometry, which negatively influences the reconstruction scores. By contrast, \ours reconstructs cleaner geometry and estimates camera poses that are closer to the ground truth, leading to the most consistent overall performance in the per-scene comparison.

\input{tables/per_scene/difficultscenes}
\input{tables/per_scene/oru}
\input{tables/per_scene/thermalgaussian}
\input{tables/per_scene/thermalnerf}
\input{tables/per_scene/thermoscenes}
\input{tables/per_scene/thermalmix}

%% file: tables/per_scene/difficultscenes.tex
\begin{table}[t]
    \centering

\caption{
Per-scene metrics on \datasetname. Compared with the other methods, our approach achieves both a higher registration rate and better pose estimation accuracy. All methods obtain low scores on \texttt{conference-room} and \texttt{old-drinking-fountain}, likely because the thermal modality in these scenes provides weak localization cues.
}
    \label{tab:per_scene_oursdataset}

\resizebox{0.9\linewidth}{!}{
\begin{tabular}{l ccc cc@{\hspace{8pt}}ccc cc}
\toprule
& \multicolumn{5}{c}{metallic-container}& \multicolumn{5}{c}{conference-room} \\\cmidrule(r){2-6} \cmidrule(l){7-11}

Method
& AUC $\uparrow$
& RRA $\uparrow$
& RTA $\uparrow$
& Reg (\%) $\uparrow$
& FPS $\uparrow$

& AUC $\uparrow$
& RRA $\uparrow$
& RTA $\uparrow$
& Reg (\%) $\uparrow$
& FPS $\uparrow$\\

\makecell[l]{COLMAP +\\SPSG} &  \cellcolor{best}{75.8} & \cellcolor{best}{100.0} & \cellcolor{best}{96.8} &  \cellcolor{secondbest}{61.5} & 0.59 &  \cellcolor{best}{78.1} & \cellcolor{best}{99.6} & \cellcolor{best}{94.0} &  \cellcolor{secondbest}{25.4} & 0.84 \\ \cmidrule(r){1-6}  \cmidrule(l){7-11}
$\text{MA}_{\text{ELoFTR}}$ &  14.1 & \cellcolor{secondbest}{97.8} & 43.2 &  50.0 & 0.17 &  2.4 & 56.0 & 9.0 &  15.2 & 0.34 \\
$\text{MINIMA}_{\text{ROMA}}$ &  41.0 & 60.7 & 64.6 &  \cellcolor{best}{100.0} & 0.04 &  22.5 & 37.3 & 57.6 &  \cellcolor{best}{100.0} & 0.04 \\ \cmidrule(r){1-6}  \cmidrule(l){7-11}
DUSt3R &  23.7 & 75.4 & 54.3 &  \cellcolor{best}{100.0} & 0.67 &  15.9 & 34.0 & 43.1 &  \cellcolor{best}{100.0} & 0.66 \\
MASt3R &  41.2 & 60.9 & 57.3 &  \cellcolor{best}{100.0} & 0.27 &  25.6 & 38.0 & 56.8 &  \cellcolor{best}{100.0} & 0.26 \\ \cmidrule(r){1-6}  \cmidrule(l){7-11}
VGGT &  21.1 & 52.3 & 69.8 &  \cellcolor{best}{100.0} & \cellcolor{best}{8.81} &  18.6 & 38.7 & 44.5 &  \cellcolor{best}{100.0} & \cellcolor{best}{12.63} \\
MapAnything & 19.7 & 52.3 & 57.9 &  \cellcolor{best}{100.0} & 1.78 &14.5 & 36.9 & 49.5 &  \cellcolor{best}{100.0} & 2.54\\
\ours &  \cellcolor{secondbest}{67.7} & 95.9 & \cellcolor{secondbest}{91.9} &  \cellcolor{best}{100.0} & \cellcolor{secondbest}{8.64} &  \cellcolor{secondbest}{46.6} & \cellcolor{secondbest}{74.4} & \cellcolor{secondbest}{71.4} &  \cellcolor{best}{100.0} & \cellcolor{secondbest}{11.98} \\
\end{tabular}}

\resizebox{0.9\linewidth}{!}{
\begin{tabular}{l ccc cc@{\hspace{8pt}}ccc cc}
\toprule
& \multicolumn{5}{c}{statue}& \multicolumn{5}{c}{old-drinking-fountain} \\\cmidrule(r){2-6} \cmidrule(l){7-11}

Method
& AUC $\uparrow$
& RRA $\uparrow$
& RTA $\uparrow$
& Reg (\%) $\uparrow$
& FPS $\uparrow$

& AUC $\uparrow$
& RRA $\uparrow$
& RTA $\uparrow$
& Reg (\%) $\uparrow$
& FPS $\uparrow$\\

\makecell[l]{COLMAP +\\SPSG} &  54.8 & \cellcolor{best}{100.0} & \cellcolor{secondbest}{82.1} &  25.7 & 0.55 &  21.9 & 50.0 & \cellcolor{best}{75.0} &  2.0 & 0.74 \\ \cmidrule(r){1-6}  \cmidrule(l){7-11}
$\text{MA}_{\text{ELoFTR}}$ &  19.7 & 66.8 & 73.6 &  \cellcolor{secondbest}{50.5} & 0.11 &  21.1 & 55.9 & 56.9 &  \cellcolor{secondbest}{11.7} & 0.12 \\
$\text{MINIMA}_{\text{ROMA}}$ &  \cellcolor{secondbest}{58.3} & 76.2 & 75.9 &  \cellcolor{best}{100.0} & 0.04 &  \cellcolor{secondbest}{49.7} & \cellcolor{secondbest}{60.1} & 61.3 &  \cellcolor{best}{100.0} & 0.04 \\ \cmidrule(r){1-6}  \cmidrule(l){7-11}
DUSt3R &  23.4 & 41.8 & 34.5 &  \cellcolor{best}{100.0} & 0.67 &  16.8 & 43.2 & 33.6 &  \cellcolor{best}{100.0} & 0.67 \\
MASt3R &  35.7 & 49.8 & 57.3 &  \cellcolor{best}{100.0} & 0.25 &  37.5 & 45.8 & 52.6 &  \cellcolor{best}{100.0} & 0.25 \\ \cmidrule(r){1-6}  \cmidrule(l){7-11}
VGGT &  24.6 & 61.6 & 52.1 &  \cellcolor{best}{100.0} & \cellcolor{best}{10.02} &  22.4 & 43.8 & 51.1 &  \cellcolor{best}{100.0} & \cellcolor{best}{9.11} \\
MapAnything & 29.3 & 67.2 & 72.5 &  \cellcolor{best}{100.0} & 2.03 &22.8 & 49.6 & 40.2 &  \cellcolor{best}{100.0} & 1.84\\
\ours &  \cellcolor{best}{63.1} & \cellcolor{secondbest}{87.6} & \cellcolor{best}{94.0} &  \cellcolor{best}{100.0} & \cellcolor{secondbest}{9.81} &  \cellcolor{best}{50.9} & \cellcolor{best}{64.7} & \cellcolor{secondbest}{65.6} &  \cellcolor{best}{100.0} & \cellcolor{secondbest}{8.92} \\
\end{tabular}}

\resizebox{0.9\linewidth}{!}{
\begin{tabular}{l ccc cc@{\hspace{8pt}}ccc cc}
\toprule
& \multicolumn{5}{c}{parking}& \multicolumn{5}{c}{telescope} \\\cmidrule(r){2-6} \cmidrule(l){7-11}

Method
& AUC $\uparrow$
& RRA $\uparrow$
& RTA $\uparrow$
& Reg (\%) $\uparrow$
& FPS $\uparrow$

& AUC $\uparrow$
& RRA $\uparrow$
& RTA $\uparrow$
& Reg (\%) $\uparrow$
& FPS $\uparrow$\\

\makecell[l]{COLMAP +\\SPSG} &  \cellcolor{best}{77.9} & \cellcolor{best}{100.0} & \cellcolor{best}{99.9} &  \cellcolor{secondbest}{50.0} & 0.74 &  34.0 & 50.0 & 50.0 &  2.6 & 0.49 \\ \cmidrule(r){1-6}  \cmidrule(l){7-11}
$\text{MA}_{\text{ELoFTR}}$ &  22.4 & 74.5 & 79.5 &  42.8 & 0.15 &  11.4 & 42.8 & 34.7 &  \cellcolor{secondbest}{53.6} & 0.12 \\
$\text{MINIMA}_{\text{ROMA}}$ &  44.4 & 48.3 & 53.0 &  \cellcolor{best}{100.0} & 0.05 &  \cellcolor{secondbest}{62.4} & \cellcolor{best}{100.0} & \cellcolor{best}{98.5} &  \cellcolor{best}{100.0} & 0.04 \\ \cmidrule(r){1-6}  \cmidrule(l){7-11}
DUSt3R &  23.0 & 41.9 & 53.6 &  \cellcolor{best}{100.0} & 0.67 &  4.8 & 46.2 & 14.2 &  \cellcolor{best}{100.0} & 0.68 \\
MASt3R &  45.4 & 49.5 & 57.1 &  \cellcolor{best}{100.0} & 0.26 &  44.7 & 78.1 & 58.6 &  \cellcolor{best}{100.0} & 0.24 \\ \cmidrule(r){1-6}  \cmidrule(l){7-11}
VGGT &  36.0 & 62.6 & 67.3 &  \cellcolor{best}{100.0} & \cellcolor{best}{11.73} &  20.5 & 41.7 & 43.9 &  \cellcolor{best}{100.0} & \cellcolor{best}{10.78} \\
MapAnything & 22.0 & 38.8 & 47.1 &  \cellcolor{best}{100.0} & 2.34 &28.6 & 50.3 & 41.8 &  \cellcolor{best}{100.0} & 2.16\\

\ours &  \cellcolor{secondbest}{72.1} & \cellcolor{secondbest}{84.3} & \cellcolor{secondbest}{84.5} &  \cellcolor{best}{100.0} & \cellcolor{secondbest}{11.45} &  \cellcolor{best}{74.8} & \cellcolor{secondbest}{92.7} & \cellcolor{secondbest}{95.0} &  \cellcolor{best}{100.0} & \cellcolor{secondbest}{10.55} \\ \bottomrule
\end{tabular}}
    
\end{table}

%% file: tables/per_scene/oru.tex
\begin{table}[t]
    \centering
\caption{
Per-scene metrics on the \orebrodataset dataset. Our method achieves a superior registration rate and higher pose estimation accuracy than the competing methods. All methods obtain low scores on \texttt{01\_Annexet\_No\_Radars\_1} and \texttt{01\_Annexet\_No\_Radars\_2}, likely due to the complex camera motion in these sequences. In addition, the translation scores on \texttt{01\_Annexet\_No\_Radars\_3} are low because the trajectory contains frames with very small translational changes near the end, so even a slight deviation in translation can lead to large angular errors.
}
    \label{tab:per_scene_rlt}

\resizebox{\linewidth}{!}{
\begin{tabular}{l ccc ccc cc@{\hspace{8pt}}ccc ccc cc}
\toprule
& \multicolumn{8}{c}{01\_Annexet\_No\_Radars\_0}& \multicolumn{8}{c}{01\_Annexet\_No\_Radars\_1} \\\cmidrule(r){2-9} \cmidrule(l){10-14}

Method
& AUC $\uparrow$
& RRA $\uparrow$
& RTA $\uparrow$
& PCA $\downarrow$
& PCC $\downarrow$
& Chamfer $\downarrow$
& Reg (\%) $\uparrow$
& FPS $\uparrow$

& AUC $\uparrow$
& RRA $\uparrow$
& RTA $\uparrow$
& PCA $\downarrow$
& PCC $\downarrow$
& Chamfer $\downarrow$
& Reg (\%) $\uparrow$
& FPS $\uparrow$\\

\makecell[l]{COLMAP +\\SPSG} &  58.7 & \cellcolor{best}{100.0} & 76.3 &  3.17 & 1.56 & 2.36 &  49.6 & 0.24 &  \cellcolor{best}{86.1} & \cellcolor{best}{100.0} & \cellcolor{best}{99.9} &  1.44 & \cellcolor{secondbest}{1.03} & \cellcolor{secondbest}{1.24} &  \cellcolor{secondbest}{24.8} & 0.34 \\ \cmidrule(r){1-6}  \cmidrule(l){7-11}
$\text{MA}_{\text{ELoFTR}}$ &  15.4 & 96.6 & 31.5 &  4.27 & 8.47 & 6.37 &  10.3 & 0.12 &  3.8 & 79.1 & 10.4 &  \cellcolor{best}{0.68} & 3.05 & 1.87 &  13.9 & 0.16 \\
$\text{MINIMA}_{\text{ROMA}}$ &  \cellcolor{secondbest}{65.8} & \cellcolor{best}{100.0} & \cellcolor{secondbest}{87.3} &  0.60 & 0.57 & 0.59 &  \cellcolor{secondbest}{95.9} & 0.06 &  22.7 & 49.1 & 56.6 &  1.95 & 4.84 & 3.39 &  \cellcolor{best}{100.0} & 0.05 \\ 
MP-SfM & 7.4 & 89.2 & 26.0 &  1.2 & 0.5 & 0.8 &  \cellcolor{best}{100.0} & 0.03 &2.0 & 70.2 & 14.9 &  5.0 & 1.4 & 3.2 &  \cellcolor{best}{100.0} & 0.03\\
\cmidrule(r){1-6}  \cmidrule(l){7-11}
DUSt3R &  21.3 & 48.4 & 33.3 &  0.57 & 10.40 & 5.49 &  \cellcolor{best}{100.0} & 0.66 &  14.9 & 27.5 & 49.9 &  1.05 & 6.16 & 3.60 &  \cellcolor{best}{100.0} & 0.67 \\
MASt3R &  50.0 & \cellcolor{secondbest}{99.2} & 74.1 &  \cellcolor{best}{0.41} & \cellcolor{best}{0.08} & \cellcolor{best}{0.24} &  \cellcolor{best}{100.0} & 0.27 &  15.2 & 41.9 & 44.4 &  1.71 & 1.52 & 1.61 &  \cellcolor{best}{100.0} & 0.25 \\ \cmidrule(r){1-6}  \cmidrule(l){7-11}
VGGT &  24.4 & 49.5 & 50.3 &  1.19 & 0.71 & 0.95 &  \cellcolor{best}{100.0} & \cellcolor{best}{9.92} &  12.0 & 44.2 & 40.4 &  1.62 & 10.08 & 5.85 &  \cellcolor{best}{100.0} & \cellcolor{best}{10.39} \\
MapAnything & 21.1 & 49.1 & 48.3 &  0.4 & 6.7 & 3.5 &  \cellcolor{best}{100.0} & 1.78 &11.2 & 63.0 & 40.4 &  1.7 & 1.8 & 1.8 &  \cellcolor{best}{100.0} & 1.83\\
\ours &  \cellcolor{best}{78.5} & \cellcolor{best}{100.0} & \cellcolor{best}{94.2} &  \cellcolor{secondbest}{0.45} & \cellcolor{secondbest}{0.14} & \cellcolor{secondbest}{0.30} &  \cellcolor{best}{100.0} & \cellcolor{secondbest}{9.48} &  \cellcolor{secondbest}{53.7} & \cellcolor{secondbest}{99.9} & \cellcolor{secondbest}{84.4} &  \cellcolor{secondbest}{0.93} & \cellcolor{best}{0.12} & \cellcolor{best}{0.53} &  \cellcolor{best}{100.0} & \cellcolor{secondbest}{9.98} \\
\end{tabular}}

\resizebox{\linewidth}{!}{
\begin{tabular}{l ccc ccc cc@{\hspace{8pt}}ccc ccc cc}
\toprule
& \multicolumn{8}{c}{01\_Annexet\_No\_Radars\_2}& \multicolumn{8}{c}{01\_Annexet\_No\_Radars\_3} \\\cmidrule(r){2-9} \cmidrule(l){10-14}

Method
& AUC $\uparrow$
& RRA $\uparrow$
& RTA $\uparrow$
& PCA $\downarrow$
& PCC $\downarrow$
& Chamfer $\downarrow$
& Reg (\%) $\uparrow$
& FPS $\uparrow$

& AUC $\uparrow$
& RRA $\uparrow$
& RTA $\uparrow$
& PCA $\downarrow$
& PCC $\downarrow$
& Chamfer $\downarrow$
& Reg (\%) $\uparrow$
& FPS $\uparrow$\\

\makecell[l]{COLMAP +\\SPSG} &  \cellcolor{secondbest}{36.0} & 86.5 & \cellcolor{secondbest}{74.9} &  2.15 & 2.08 & 2.11 &  \cellcolor{secondbest}{20.2} & 0.35 &  15.1 & \cellcolor{best}{100.0} & 27.9 &  2.31 & 6.12 & 4.21 &  31.0 & 0.26 \\ \cmidrule(r){1-6}  \cmidrule(l){7-11}
$\text{MA}_{\text{ELoFTR}}$ &  4.0 & 51.8 & 21.7 &  \cellcolor{best}{0.13} & 4.68 & 2.41 &  13.4 & 0.12 &  10.0 & \cellcolor{best}{100.0} & 22.1 &  0.49 & 4.19 & 2.34 &  40.2 & 0.18 \\
$\text{MINIMA}_{\text{ROMA}}$ &  29.4 & 87.6 & 52.7 &  0.76 & 1.18 & 0.97 &  \cellcolor{best}{100.0} & 0.05 &  6.7 & \cellcolor{best}{100.0} & 16.1 &  1.20 & 5.18 & 3.19 &  \cellcolor{secondbest}{79.9} & 0.05 \\ 
MP-SfM & 1.2 & 45.8 & 15.5 &  0.9 & 1.2 & 1.1 &  \cellcolor{best}{100.0} & 0.03 &1.7 & 100.0 & 7.3 &  1.3 & 0.6 & 0.9 &  \cellcolor{best}{100.0} & 0.04\\
\cmidrule(r){1-6}  \cmidrule(l){7-11}
DUSt3R &  14.0 & 31.6 & 36.3 &  0.73 & 8.52 & 4.62 &  \cellcolor{best}{100.0} & 0.65 &  7.7 & \cellcolor{secondbest}{49.5} & 25.4 &  0.59 & 4.65 & 2.62 &  \cellcolor{best}{100.0} & 0.67 \\
MASt3R &  28.5 & \cellcolor{secondbest}{95.3} & 53.9 &  0.91 & \cellcolor{secondbest}{0.66} & 0.78 &  \cellcolor{best}{100.0} & 0.25 &  \cellcolor{secondbest}{16.8} & \cellcolor{best}{100.0} & 37.3 &  \cellcolor{secondbest}{0.41} & \cellcolor{secondbest}{3.40} & \cellcolor{secondbest}{1.90} &  \cellcolor{best}{100.0} & 0.28 \\ \cmidrule(r){1-6}  \cmidrule(l){7-11}
VGGT &  16.9 & 65.7 & 42.6 &  0.77 & 0.69 & \cellcolor{secondbest}{0.73} &  \cellcolor{best}{100.0} & \cellcolor{best}{10.19} &  14.3 & 46.9 & \cellcolor{secondbest}{43.1} &  0.68 & 3.64 & 2.16 &  \cellcolor{best}{100.0} & \cellcolor{best}{12.05} \\
MapAnything & 10.3 & 42.8 & 39.2 &  0.8 & 3.9 & 2.4 &  \cellcolor{best}{100.0} & 1.80 &12.4 & 49.5 & 28.1 &  0.6 & 4.3 & 2.5 &  \cellcolor{best}{100.0} & 2.09\\
\ours &  \cellcolor{best}{63.3} & \cellcolor{best}{99.9} & \cellcolor{best}{89.5} &  \cellcolor{secondbest}{0.68} & \cellcolor{best}{0.07} & \cellcolor{best}{0.38} &  \cellcolor{best}{100.0} & \cellcolor{secondbest}{9.78} &  \cellcolor{best}{51.7} & \cellcolor{best}{100.0} & \cellcolor{best}{78.4} &  \cellcolor{best}{0.38} & \cellcolor{best}{0.11} & \cellcolor{best}{0.25} &  \cellcolor{best}{100.0} & \cellcolor{secondbest}{11.49} \\
\end{tabular}}

\resizebox{\linewidth}{!}{
\begin{tabular}{l ccc ccc cc@{\hspace{8pt}}ccc ccc cc}
\toprule
& \multicolumn{8}{c}{04\_Forest\_pass\_no\_radars\_0}& \multicolumn{8}{c}{04\_Forest\_pass\_no\_radars\_1} \\\cmidrule(r){2-9} \cmidrule(l){10-14}

Method
& AUC $\uparrow$
& RRA $\uparrow$
& RTA $\uparrow$
& PCA $\downarrow$
& PCC $\downarrow$
& Chamfer $\downarrow$
& Reg (\%) $\uparrow$
& FPS $\uparrow$

& AUC $\uparrow$
& RRA $\uparrow$
& RTA $\uparrow$
& PCA $\downarrow$
& PCC $\downarrow$
& Chamfer $\downarrow$
& Reg (\%) $\uparrow$
& FPS $\uparrow$\\

\makecell[l]{COLMAP +\\SPSG} &  \cellcolor{best}{96.9} & \cellcolor{best}{100.0} & \cellcolor{best}{100.0} &  1.96 & 0.98 & 1.47 &  \cellcolor{secondbest}{50.0} & 0.25 &  \cellcolor{best}{96.3} & \cellcolor{best}{100.0} & \cellcolor{best}{99.9} &  \cellcolor{secondbest}{0.55} & 0.64 & 0.59 &  \cellcolor{secondbest}{50.0} & 0.31 \\ \cmidrule(r){1-6}  \cmidrule(l){7-11}
$\text{MA}_{\text{ELoFTR}}$ &  5.3 & \cellcolor{secondbest}{86.3} & 18.1 &  \cellcolor{best}{0.41} & 5.01 & 2.71 &  26.9 & 0.12 &  1.6 & \cellcolor{best}{100.0} & 6.6 &  1.49 & 5.23 & 3.36 &  7.3 & 0.14 \\
$\text{MINIMA}_{\text{ROMA}}$ &  76.7 & \cellcolor{best}{100.0} & 93.3 &  1.41 & 0.74 & 1.08 &  \cellcolor{best}{100.0} & 0.06 &  79.6 & \cellcolor{best}{100.0} & 96.0 &  0.71 & 0.37 & 0.54 &  \cellcolor{best}{100.0} & 0.06 \\ 
MP-SfM & 4.4 & 100.0 & 20.4 &  2.7 & 0.4 & 1.5 &  \cellcolor{best}{100.0} & 0.03 &8.8 & 100.0 & 34.6 &  1.4 & 0.3 & 0.8 &  \cellcolor{best}{100.0} & 0.05\\
\cmidrule(r){1-6}  \cmidrule(l){7-11}
DUSt3R &  23.2 & 49.6 & 61.0 &  1.31 & 1.84 & 1.57 &  \cellcolor{best}{100.0} & 0.66 &  20.5 & 49.0 & 44.4 &  0.97 & 2.23 & 1.60 &  \cellcolor{best}{100.0} & 0.67 \\
MASt3R &  62.1 & \cellcolor{best}{100.0} & 90.7 &  0.94 & \cellcolor{secondbest}{0.06} & \cellcolor{secondbest}{0.50} &  \cellcolor{best}{100.0} & 0.25 &  47.5 & \cellcolor{secondbest}{99.0} & 82.4 &  0.73 & \cellcolor{secondbest}{0.12} & \cellcolor{secondbest}{0.42} &  \cellcolor{best}{100.0} & 0.26 \\ \cmidrule(r){1-6}  \cmidrule(l){7-11}
VGGT &  31.9 & 49.6 & 59.3 &  3.86 & 1.05 & 2.46 &  \cellcolor{best}{100.0} & \cellcolor{best}{9.36} &  33.2 & 49.6 & 60.6 &  1.80 & 2.13 & 1.97 &  \cellcolor{best}{100.0} & \cellcolor{best}{10.30} \\
MapAnything & 28.3 & 49.6 & 53.2 &  1.0 & 2.1 & 1.6 &  \cellcolor{best}{100.0} & 1.67 &26.5 & 49.5 & 87.4 &  0.2 & 6.8 & 3.5 &  \cellcolor{best}{100.0} & 1.84\\
\ours &  \cellcolor{secondbest}{94.1} & \cellcolor{best}{100.0} & \cellcolor{secondbest}{99.3} &  \cellcolor{secondbest}{0.42} & \cellcolor{best}{0.02} & \cellcolor{best}{0.22} &  \cellcolor{best}{100.0} & \cellcolor{secondbest}{9.05} &  \cellcolor{secondbest}{92.5} & \cellcolor{best}{100.0} & \cellcolor{secondbest}{98.9} &  \cellcolor{best}{0.27} & \cellcolor{best}{0.03} & \cellcolor{best}{0.15} &  \cellcolor{best}{100.0} & \cellcolor{secondbest}{9.92} \\ \bottomrule
\end{tabular}}   

\end{table}

%% file: tables/per_scene/thermalgaussian.tex
\begin{table}[t]
    \centering
\caption{
Per-scene metrics on the ThermalGaussian dataset. Our method achieves the best pose estimation accuracy on \texttt{Ebike} and \texttt{Parterre}, while also delivering competitive performance on \texttt{Iron Ingot}.
}
    \label{tab:per_scene_thermalgaussian}

\resizebox{0.9\linewidth}{!}{
\begin{tabular}{l ccc cc@{\hspace{8pt}}ccc cc}
\toprule
& \multicolumn{5}{c}{Ebike}
& \multicolumn{5}{c}{IronIngot} \\
\cmidrule(r){2-6}
\cmidrule(l){7-11}

Method
& AUC $\uparrow$
& RRA $\uparrow$
& RTA $\uparrow$
& Reg (\%) $\uparrow$
& FPS $\uparrow$

& AUC $\uparrow$
& RRA $\uparrow$
& RTA $\uparrow$
& Reg (\%) $\uparrow$
& FPS $\uparrow$\\

\makecell[l]{COLMAP +\\SPSG} &  64.8 & 70.2 & 70.4 &  \cellcolor{secondbest}{95.8} & 0.60 & {92.5} & \cellcolor{best}{100.0} & \cellcolor{secondbest}{99.3} &  \cellcolor{best}{100.0} & 0.54 \\ \cmidrule(r){1-6} \cmidrule(l){7-11}
$\text{MA}_{\text{ELoFTR}}$ &  44.6 & \cellcolor{secondbest}{77.8} & 66.7 &  7.3 & 0.38 &  12.8 & 55.9 & 57.4 &  \cellcolor{secondbest}{61.5} & 0.12 \\
$\text{MINIMA}_{\text{ROMA}}$ &  \cellcolor{secondbest}{93.9} & \cellcolor{best}{100.0} & \cellcolor{best}{100.0} &  \cellcolor{best}{100.0} & 0.07 &  \cellcolor{best}{94.2} & \cellcolor{best}{100.0} & \cellcolor{best}{99.6} &  \cellcolor{best}{100.0} & 0.06 \\ 
MP-SfM & 93.6 & \cellcolor{best}{100.0} & \cellcolor{secondbest}{99.8} &  \cellcolor{best}{100.0} & 0.06 & \cellcolor{secondbest}{94.1} & \cellcolor{best}{100.0} & \cellcolor{secondbest}{99.3} &  \cellcolor{best}{100.0} & 0.06\\
\cmidrule(r){1-6} \cmidrule(l){7-11}
DUSt3R &  20.0 & 29.7 & 35.9 &  \cellcolor{best}{100.0} & 0.65 &  15.6 & \cellcolor{secondbest}{74.9} & 39.9 &  \cellcolor{best}{100.0} & 0.66 \\
MASt3R &  90.0 & \cellcolor{best}{100.0} & 99.5 &  \cellcolor{best}{100.0} & 0.24 &  73.3 & \cellcolor{best}{100.0} & 96.4 &  \cellcolor{best}{100.0} & 0.23 \\ \cmidrule(r){1-6} \cmidrule(l){7-11}
VGGT &  84.3 & \cellcolor{best}{100.0} & 98.2 &  \cellcolor{best}{100.0} & \cellcolor{best}{16.06} &  31.9 & 68.3 & 70.1 &  \cellcolor{best}{100.0} & \cellcolor{best}{14.13} \\
MapAnything & 14.6 & 29.2 & 41.5 &  \cellcolor{best}{100.0} & 3.06 &45.1 & 97.0 & 75.9 &  \cellcolor{best}{100.0} & 2.76\\

\ours &  \cellcolor{best}{95.2} & \cellcolor{best}{100.0} & \cellcolor{best}{100.0} &  \cellcolor{best}{100.0} & \cellcolor{secondbest}{15.07} &  82.1 & \cellcolor{best}{100.0} & 96.4 &  \cellcolor{best}{100.0} & \cellcolor{secondbest}{13.34} \\
\end{tabular}}

\resizebox{0.5\linewidth}{!}{
\begin{tabular}{l ccc cc@{\hspace{8pt}}ccc cc}
\toprule
& \multicolumn{5}{c}{Parterre}
& \multicolumn{5}{c}{ } \\
\cmidrule(r){2-6}

Method
& AUC $\uparrow$
& RRA $\uparrow$
& RTA $\uparrow$
& Reg (\%) $\uparrow$
& FPS $\uparrow$

& & & & & \\
\makecell[l]{COLMAP +\\SPSG} &  35.0 & \cellcolor{best}{100.0} & 50.0 &  3.0 & 0.53 &    &   &   &    &   \\ \cmidrule(r){1-6} 
$\text{MA}_{\text{ELoFTR}}$ &  32.1 & 66.4 & 85.5 &  \cellcolor{secondbest}{24.2} & 0.23 &    &   &   &    &   \\
$\text{MINIMA}_{\text{ROMA}}$ &  33.9 & 47.4 & 64.2 &  \cellcolor{best}{100.0} & 0.04 &    &   &   &    &   \\ 
MP-SfM & 26.9 & 44.0 & 63.8 &  \cellcolor{best}{100.0} & 0.02 & &  &  &   &  \\
\cmidrule(r){1-6} 
DUSt3R &  43.2 & \cellcolor{secondbest}{95.3} & 79.5 &  \cellcolor{best}{100.0} & 0.65 &    &   &   &    &   \\
MASt3R &  \cellcolor{secondbest}{85.6} & \cellcolor{best}{100.0} & \cellcolor{secondbest}{98.8} &  \cellcolor{best}{100.0} & 0.22 &    &   &   &    &   \\ \cmidrule(r){1-6} 
VGGT &  65.4 & 87.2 & 86.0 &  \cellcolor{best}{100.0} & \cellcolor{best}{13.81} &    &   &   &    &   \\
MapAnything & 33.6 & 69.6 & 66.9 &  \cellcolor{best}{100.0} & 2.66 & &  &  &   & \\

\ours &  \cellcolor{best}{93.7} & \cellcolor{best}{100.0} & \cellcolor{best}{100.0} &  \cellcolor{best}{100.0} & \cellcolor{secondbest}{13.01} &    &   &   &    &   \\ \bottomrule
\end{tabular}}

\end{table}

%% file: tables/per_scene/thermalnerf.tex
\begin{table}[t]
    \centering
    \caption{
    Per-scene metrics on ThermalNeRF. Our method achieves a superior registration rate and higher pose estimation accuracy than the competing methods.
    }
    \label{tab:per_scene_thermalnerf}
    
\resizebox{0.9\linewidth}{!}{
\begin{tabular}{l ccc cc@{\hspace{8pt}}ccc cc}
\toprule
& \multicolumn{5}{c}{sink}
& \multicolumn{5}{c}{generator} \\
\cmidrule(r){2-6}
\cmidrule(l){7-11}

Method
& AUC $\uparrow$
& RRA $\uparrow$
& RTA $\uparrow$
& Reg (\%) $\uparrow$
& FPS $\uparrow$

& AUC $\uparrow$
& RRA $\uparrow$
& RTA $\uparrow$
& Reg (\%) $\uparrow$
& FPS $\uparrow$\\

\makecell[l]{COLMAP +\\SPSG} &  \cellcolor{secondbest}{86.4} & \cellcolor{best}{100.0} & \cellcolor{secondbest}{98.5} &  \cellcolor{secondbest}{38.2} & 0.61 &  52.7 & 79.8 & 73.8 &  \cellcolor{best}{100.0} & 0.64 \\ \cmidrule(r){1-6} \cmidrule(l){7-11}
$\text{MA}_{\text{ELoFTR}}$ &  12.4 & \cellcolor{secondbest}{87.2} & 35.0 &  37.3 & 0.24 &  4.2 & 53.1 & 27.0 &  \cellcolor{secondbest}{19.7} & 0.22 \\
$\text{MINIMA}_{\text{ROMA}}$ &  48.8 & \cellcolor{best}{100.0} & 70.5 &  \cellcolor{best}{100.0} & 0.07 &  62.1 & \cellcolor{best}{100.0} & \cellcolor{secondbest}{84.5} &  \cellcolor{best}{100.0} & 0.06 \\
MP-SfM & 82.7 & 100.0 & 97.9 &  \cellcolor{best}{100.0} & 0.05 & \cellcolor{best}{91.5} & \cellcolor{best}{100.0} & 99.7 &  \cellcolor{best}{100.0} & 0.04\\
\cmidrule(r){1-6} \cmidrule(l){7-11}
DUSt3R &  33.6 & \cellcolor{best}{100.0} & 61.6 &  \cellcolor{best}{100.0} & 0.67 &  7.4 & 41.4 & 32.9 &  \cellcolor{best}{100.0} & 0.66 \\
MASt3R &  58.8 & \cellcolor{best}{100.0} & 85.6 &  \cellcolor{best}{100.0} & 0.24 &  32.3 & \cellcolor{secondbest}{84.8} & 47.3 &  \cellcolor{best}{100.0} & 0.24 \\ \cmidrule(r){1-6} \cmidrule(l){7-11}
VGGT &  28.5 & 43.3 & 53.1 &  \cellcolor{best}{100.0} & \cellcolor{best}{15.55} &  20.7 & 41.7 & 50.2 &  \cellcolor{best}{100.0} & \cellcolor{best}{12.60} \\
MapAnything & 31.7 & 100.0 & 53.6 &  \cellcolor{best}{100.0} & 2.96 &22.3 & 61.2 & 47.8 &  \cellcolor{best}{100.0} & 2.46\\
\ours &  \cellcolor{best}{88.1} & \cellcolor{best}{100.0} & \cellcolor{best}{99.5} &  \cellcolor{best}{100.0} & \cellcolor{secondbest}{14.42} &  \cellcolor{secondbest}{90.9} & \cellcolor{best}{100.0} & \cellcolor{best}{99.1} &  \cellcolor{best}{100.0} & \cellcolor{secondbest}{11.99} \\ \bottomrule
\end{tabular}}
\end{table}

%% file: tables/per_scene/thermoscenes.tex
\begin{table}[t]
\centering

    \caption{
    Per-scene metrics on the ThermoScenes dataset. Our method achieves a higher registration rate and better pose estimation accuracy than the competing methods. All methods achieve low scores on \texttt{freezing\_ice\_cup} and \texttt{melting\_ice\_cup}, likely due to the inherent difficulty of these scenes.
    }
    \label{tab:per_scene_thermoscenes}
    
\resizebox{0.9\linewidth}{!}{
\begin{tabular}{l ccc cc@{\hspace{8pt}}ccc cc}
\toprule
& \multicolumn{5}{c}{freezing\_ice\_cup}
& \multicolumn{5}{c}{prpt-cup} \\
\cmidrule(r){2-6}
\cmidrule(l){7-11}

Method
& AUC $\uparrow$
& RRA $\uparrow$
& RTA $\uparrow$
& Reg (\%) $\uparrow$
& FPS $\uparrow$

& AUC $\uparrow$
& RRA $\uparrow$
& RTA $\uparrow$
& Reg (\%) $\uparrow$
& FPS $\uparrow$\\

\makecell[l]{COLMAP +\\SPSG} &  \cellcolor{best}{34.5} & \cellcolor{best}{57.2} & \cellcolor{best}{60.5} &  \cellcolor{secondbest}{50.0} & 0.52 &  \cellcolor{secondbest}{51.4} & \cellcolor{secondbest}{72.0} & \cellcolor{secondbest}{75.3} &  \cellcolor{secondbest}{48.1} & 0.61 \\ \cmidrule(r){1-6} \cmidrule(l){7-11}
$\text{MA}_{\text{ELoFTR}}$ &  1.5 & 30.6 & 22.3 &  7.6 & 0.33 &  20.2 & 70.0 & 40.0 &  2.6 & 0.27 \\
$\text{MINIMA}_{\text{ROMA}}$ &  4.3 & 12.0 & 25.4 &  \cellcolor{best}{100.0} & 0.04 &  7.1 & 20.9 & 31.0 &  \cellcolor{best}{100.0} & 0.04 \\ 
MP-SfM & \cellcolor{secondbest}{23.9} & \cellcolor{secondbest}{39.4} & \cellcolor{secondbest}{58.2} &  \cellcolor{best}{100.0} & 0.04 &60.2 & 93.9 & 87.6 &  \cellcolor{best}{100.0} & 0.06\\

\cmidrule(r){1-6} \cmidrule(l){7-11}
DUSt3R &  14.6 & 30.1 & 42.4 &  \cellcolor{best}{100.0} & 0.66 &  19.9 & 28.2 & 33.6 &  \cellcolor{best}{100.0} & 0.65 \\
MASt3R &  15.3 & 38.0 & 39.0 &  \cellcolor{best}{100.0} & 0.22 &  24.6 & 55.9 & 50.3 &  \cellcolor{best}{100.0} & 0.22 \\ \cmidrule(r){1-6} \cmidrule(l){7-11}
VGGT &  13.2 & 27.2 & 35.9 &  \cellcolor{best}{100.0} & \cellcolor{best}{4.89} &  17.6 & 45.9 & 41.4 &  \cellcolor{best}{100.0} & \cellcolor{best}{4.09} \\
MapAnything & 8.8 & 18.9 & 28.8 &  \cellcolor{best}{100.0} & 1.00 &15.9 & 38.6 & 40.0 &  \cellcolor{best}{100.0} & 0.84\\
\ours &  18.3 & 35.1 & 45.2 &  \cellcolor{best}{100.0} & \cellcolor{secondbest}{4.76} &  \cellcolor{best}{81.2} & \cellcolor{best}{99.9} & \cellcolor{best}{97.1} &  \cellcolor{best}{100.0} & \cellcolor{secondbest}{4.00} \\
\end{tabular}}

\resizebox{0.9\linewidth}{!}{
\begin{tabular}{l ccc cc@{\hspace{8pt}}ccc cc}
\toprule
& \multicolumn{5}{c}{reflect-robot}
& \multicolumn{5}{c}{melting\_ice\_cup} \\
\cmidrule(r){2-6}
\cmidrule(l){7-11}

Method
& AUC $\uparrow$
& RRA $\uparrow$
& RTA $\uparrow$
& Reg (\%) $\uparrow$
& FPS $\uparrow$

& AUC $\uparrow$
& RRA $\uparrow$
& RTA $\uparrow$
& Reg (\%) $\uparrow$
& FPS $\uparrow$\\

\makecell[l]{COLMAP +\\SPSG} &  6.5 & 67.3 & 28.4 &  \cellcolor{secondbest}{29.0} & 0.54 &  0.7 & 13.1 & 19.8 &  \cellcolor{secondbest}{25.8} & 0.56 \\ \cmidrule(r){1-6} \cmidrule(l){7-11}
$\text{MA}_{\text{ELoFTR}}$ &  15.0 & 80.1 & 41.6 &  11.3 & 0.16 &  3.7 & 28.7 & 20.2 &  8.8 & 0.28 \\
$\text{MINIMA}_{\text{ROMA}}$ &  \cellcolor{secondbest}{59.0} & \cellcolor{secondbest}{90.0} & \cellcolor{secondbest}{85.4} &  \cellcolor{best}{100.0} & 0.04 &  5.5 & 16.9 & 20.3 &  \cellcolor{best}{100.0} & 0.03 \\ 
MP-SfM & 21.9 & 69.7 & 45.8 &  \cellcolor{best}{100.0} & 0.03 &8.7 & 17.9 & 28.6 &  \cellcolor{best}{100.0} & 0.05\\
\cmidrule(r){1-6} \cmidrule(l){7-11}
DUSt3R &  13.2 & 37.2 & 28.9 &  \cellcolor{best}{100.0} & 0.66 &  19.0 & \cellcolor{secondbest}{39.6} & 40.6 &  \cellcolor{best}{100.0} & 0.68 \\
MASt3R &  22.8 & 80.7 & 33.1 &  \cellcolor{best}{100.0} & 0.24 &  \cellcolor{secondbest}{25.6} & 36.8 & \cellcolor{secondbest}{47.0} &  \cellcolor{best}{100.0} & 0.24 \\ \cmidrule(r){1-6} \cmidrule(l){7-11}
VGGT &  17.3 & 45.5 & 38.5 &  \cellcolor{best}{100.0} & \cellcolor{best}{4.94} &  22.2 & 39.0 & 42.1 &  \cellcolor{best}{100.0} & \cellcolor{best}{6.81} \\
MapAnything & 17.0 & 65.8 & 30.9 &  \cellcolor{best}{100.0} & 1.00 &9.6 & 28.1 & 29.3 &  \cellcolor{best}{100.0} & 1.37\\

\ours &  \cellcolor{best}{63.0} & \cellcolor{best}{94.2} & \cellcolor{best}{88.7} &  \cellcolor{best}{100.0} & \cellcolor{secondbest}{4.81} &  \cellcolor{best}{37.0} & \cellcolor{best}{50.9} & \cellcolor{best}{59.4} &  \cellcolor{best}{100.0} & \cellcolor{secondbest}{6.59} \\
\end{tabular}}

\resizebox{0.5\linewidth}{!}{
\begin{tabular}{l ccc cc@{\hspace{8pt}}ccc cc}
\toprule
& \multicolumn{5}{c}{INR-building}
& \multicolumn{5}{c}{ } \\
\cmidrule(r){2-6}

Method
& AUC $\uparrow$
& RRA $\uparrow$
& RTA $\uparrow$
& Reg (\%) $\uparrow$
& FPS $\uparrow$

& & & & & \\
\makecell[l]{COLMAP +\\SPSG} &  58.0 & \cellcolor{best}{100.0} & 76.6 &  \cellcolor{secondbest}{49.7} & 0.49 &    &   &   &    &   \\ \cmidrule(r){1-6} 
$\text{MA}_{\text{ELoFTR}}$ &  8.7 & 73.0 & 37.2 &  15.6 & 0.19 &    &   &   &    &   \\
$\text{MINIMA}_{\text{ROMA}}$ &  54.3 & \cellcolor{best}{100.0} & 80.6 &  \cellcolor{best}{100.0} & 0.03 &    &   &   &    &   \\ 
MP-SfM & \cellcolor{secondbest}{63.2} & \cellcolor{best}{100.0} & {83.1} &  \cellcolor{best}{100.0} & 0.05 & &  &  &   &  \\
\cmidrule(r){1-6} 
DUSt3R &  27.7 & {94.2} & 55.7 &  \cellcolor{best}{100.0} & 0.66 &    &   &   &    &   \\
MASt3R &  9.0 & 38.0 & 39.3 &  \cellcolor{best}{100.0} & 0.25 &    &   &   &    &   \\ \cmidrule(r){1-6} 
VGGT &  33.7 & 87.3 & 64.9 &  \cellcolor{best}{100.0} & \cellcolor{best}{4.92} &    &   &   &    &   \\
MapAnything & 57.0 & \cellcolor{secondbest}{95.9} & \cellcolor{secondbest}{84.0} &  \cellcolor{best}{100.0} & 1.00 & &  &  & &   \\
\ours &  \cellcolor{best}{82.5} & \cellcolor{best}{100.0} & \cellcolor{best}{95.9} &  \cellcolor{best}{100.0} & \cellcolor{secondbest}{4.79} &    &   &   &    &   \\ \bottomrule
\end{tabular}}
\end{table}

%% file: tables/per_scene/thermalmix.tex
\begin{table}[]
    \centering
    \caption{
    Per-scene metrics on the ThermalMix dataset. Our method achieves a higher registration rate and better pose estimation accuracy than the competing methods.
    }
    \label{tab:per_scene_thermalmix}
    
\resizebox{0.9\linewidth}{!}{
\begin{tabular}{l ccc cc@{\hspace{8pt}}ccc cc}
\toprule
& \multicolumn{5}{c}{panel}
& \multicolumn{5}{c}{laptop} \\
\cmidrule(r){2-6}
\cmidrule(l){7-11}

Method
& AUC $\uparrow$
& RRA $\uparrow$
& RTA $\uparrow$
& Reg (\%) $\uparrow$
& FPS $\uparrow$

& AUC $\uparrow$
& RRA $\uparrow$
& RTA $\uparrow$
& Reg (\%) $\uparrow$
& FPS $\uparrow$\\

\makecell[l]{COLMAP +\\SPSG} &  15.8 & 31.8 & 46.4 &  28.0 & 0.32 &  56.1 & \cellcolor{best}{100.0} & 91.6 &  10.8 & 0.28 \\ \cmidrule(r){1-6} \cmidrule(l){7-11}
$\text{MA}_{\text{ELoFTR}}$ &  43.7 & 76.4 & \cellcolor{secondbest}{98.6} &  \cellcolor{secondbest}{35.4} & 0.36 &  20.6 & 51.0 & 56.4 &  \cellcolor{secondbest}{48.9} & 0.15 \\
$\text{MINIMA}_{\text{ROMA}}$ &  74.9 & 95.2 & 93.1 &  \cellcolor{best}{100.0} & 0.05 &  \cellcolor{secondbest}{77.1} & \cellcolor{secondbest}{94.8} & \cellcolor{secondbest}{94.2} &  \cellcolor{best}{100.0} & 0.04 \\ 
MP-SfM & \cellcolor{secondbest}{79.3} & \cellcolor{best}{100.0} & 98.1 &  \cellcolor{best}{100.0} & 0.06 &47.2 & 64.3 & 65.7 &  \cellcolor{best}{100.0} & 0.04\\
\cmidrule(r){1-6} \cmidrule(l){7-11}
DUSt3R &  15.9 & 70.4 & 39.5 &  \cellcolor{best}{100.0} & 0.65 &  21.1 & 44.0 & 45.9 &  \cellcolor{best}{100.0} & 0.67 \\
MASt3R &  34.9 & 64.5 & 66.2 &  \cellcolor{best}{100.0} & 0.23 &  29.6 & 61.7 & 73.8 &  \cellcolor{best}{100.0} & 0.23 \\ \cmidrule(r){1-6} \cmidrule(l){7-11}
VGGT &  13.0 & 28.3 & 44.0 &  \cellcolor{best}{100.0} & \cellcolor{best}{17.04} &  15.6 & 30.6 & 41.3 &  \cellcolor{best}{100.0} & \cellcolor{best}{11.23} \\
MapAnything & 16.1 & 42.6 & 58.5 &  \cellcolor{best}{100.0} & 3.25 &11.4 & 27.0 & 34.2 &  \cellcolor{best}{100.0} & 2.22\\
\ours &  \cellcolor{best}{83.3} & \cellcolor{secondbest}{99.3} & \cellcolor{best}{98.8} &  \cellcolor{best}{100.0} & \cellcolor{secondbest}{15.72} &  \cellcolor{best}{90.4} & \cellcolor{best}{100.0} & \cellcolor{best}{99.3} &  \cellcolor{best}{100.0} & \cellcolor{secondbest}{10.70} \\ \bottomrule
\end{tabular}}
    
\end{table}

%% file: supplementary/7_two_view_metrics.tex
\section{Two-View Camera Pose Estimation}

Because we vary the sequence length during training, our method naturally extends to the two-view setting for estimating the relative pose between a pair of multimodal images.
Unlike multimodal camera pose estimation on image sets (presented in the \textit{Multimodal Camera Pose Estimation} subsection of the main paper), evaluating $\text{MINIMA}_{\text{ROMA}}$ and $\text{MA}_{\text{ELoFTR}}$ in the two-view setting does not require \texttt{deep-image-matching}, since the relative pose can be recovered directly from 2D--2D correspondences via the essential matrix. This evaluation is important because it allows us to compare \ours and matching-based methods without relying on the \texttt{deep-image-matching} framework.

We use the widely adopted METU-VisTIR dataset~\cite{tuzcuouglu2024xoftr} to benchmark the methods.
In contrast, the collected \datasetname and Public datasets are not easily adaptable to two-view evaluation, because it is difficult to sample representative image pairs for a fair comparison.

To evaluate $\text{MINIMA}_{\text{ROMA}}$ and $\text{MA}_{\text{ELoFTR}}$, we first extract correspondences between a multimodal image pair, then estimate the essential matrix using the known camera intrinsics, following RoMA~\cite{edstedt2024roma}, and finally recover the relative camera pose.
For \ours, we feed the image pair directly into the model and estimate the relative pose without using known intrinsics, relying solely on the model predictions.

The results are presented in \cref{tab:relative_camera_poses}. \ours outperforms all competing methods by a large margin on nearly all metrics, with the exception of RRA@20, where it achieves a result slightly higher than $\text{MINIMA}_{\text{ROMA}}$.
VGGT is not designed for multimodal camera pose estimation and therefore performs poorly, while $\text{MA}_{\text{ELoFTR}}$ finds too few matches to enable accurate estimation.
These results show that \ours is not only effective for RGB-thermal geometry reconstruction on image sets, but also highly competitive for two-view relative camera pose estimation.

\input{tables/two_view_metrics.tex}

%% file: tables/two_view_metrics.tex
\begin{table}[!t]
  \centering
  \caption{
    Relative camera pose estimation results on METU-VisTIR for multimodal methods. Our method outperforms all competing approaches by a large margin on nearly all metrics. The only exception is RRA@20, where it achieves a result slightly higher than that of $\text{MINIMA}_{\text{ROMA}}$.
  }
  \label{tab:relative_camera_poses}
  \resizebox{0.95\linewidth}{!}{
    \begin{tabular}{c ccc ccc ccc }
      \toprule Method & AUC@5 $\uparrow$ & AUC@10 $\uparrow$ & AUC@20 $\uparrow$  & RRA@5 $\uparrow$ & RRA@10 $\uparrow$ & RRA@20 $\uparrow$  & RTA@5 $\uparrow$ & RTA@10 $\uparrow$ & RTA@20 $\uparrow$ \\ \midrule
      $\text{MA}_\text{ELoFTR}$ & 0.3 & 1.0 & 4.5 & 7.1 & 17.8 & 37.3 & 3.0 & 8.1 & 23.6 \\
      $\text{MINIMA}_\text{ROMA}$ & \cellcolor{secondbest}{9.4} & \cellcolor{secondbest}{27.2} & \cellcolor{secondbest}{50.9} & \cellcolor{secondbest}{62.7} & \cellcolor{secondbest}{88.6} & \cellcolor{secondbest}{98.0} & \cellcolor{secondbest}{37.1} & \cellcolor{secondbest}{63.9} & \cellcolor{secondbest}{84.7} \\
      VGGT & 1.8 & 8.2 & 21.9 & 60.4 & 76.2 & 84.9 & 7.4 & 24.6 & 47.6 \\
      \ours & \cellcolor{best}{23.6} & \cellcolor{best}{47.3} & \cellcolor{best}{68.0} & \cellcolor{best}{92.3} & \cellcolor{best}{97.5} & \cellcolor{best}{98.1} & \cellcolor{best}{57.7} & \cellcolor{best}{81.9} & \cellcolor{best}{92.8} \\ \bottomrule
    \end{tabular}
  }

\end{table}

%% file: supplementary/8_more_results_align_rgb_thermal.tex
\section{Thermal to RGB alignment}

In this section, we present additional results demonstrating the alignment between RGB and thermal features in our model.

\begin{figure}
  \centering
  \includegraphics[page=1,width=0.95\linewidth]{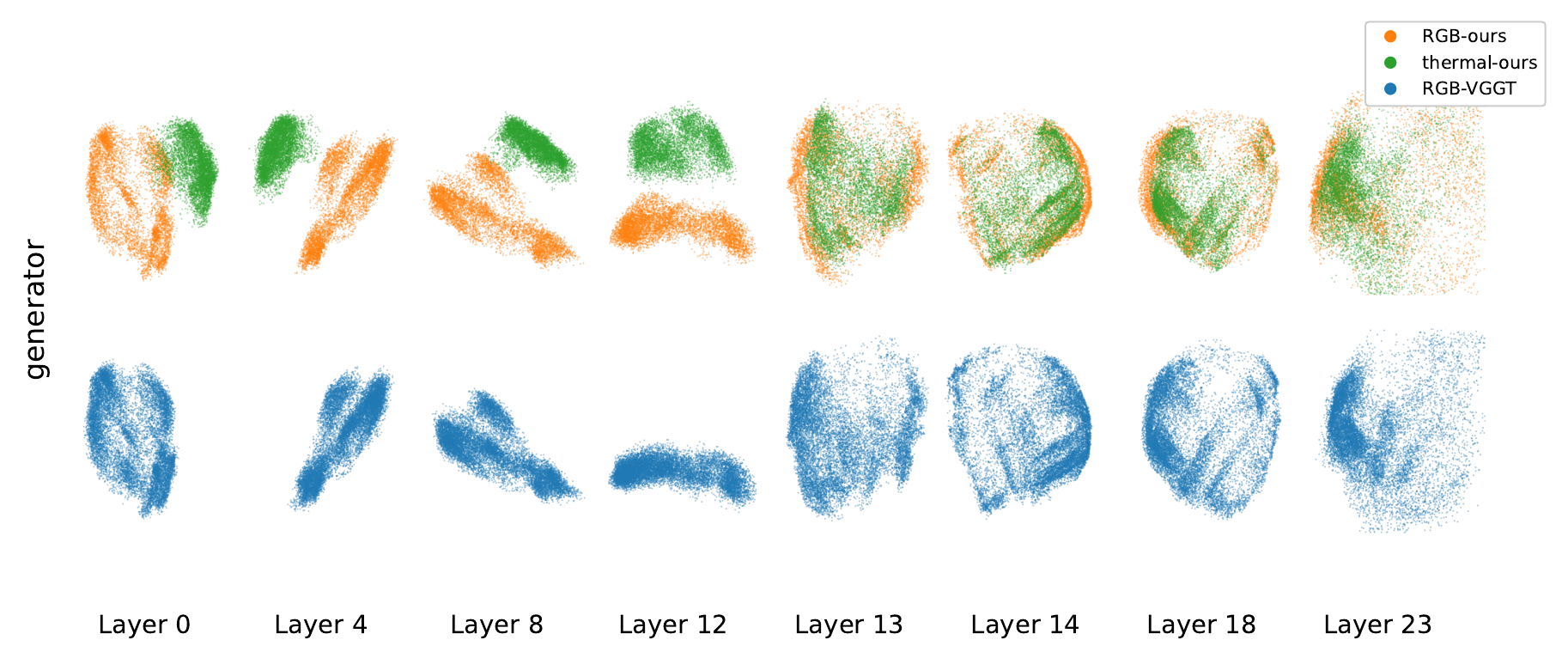}
  \includegraphics[page=1,width=0.95\linewidth]{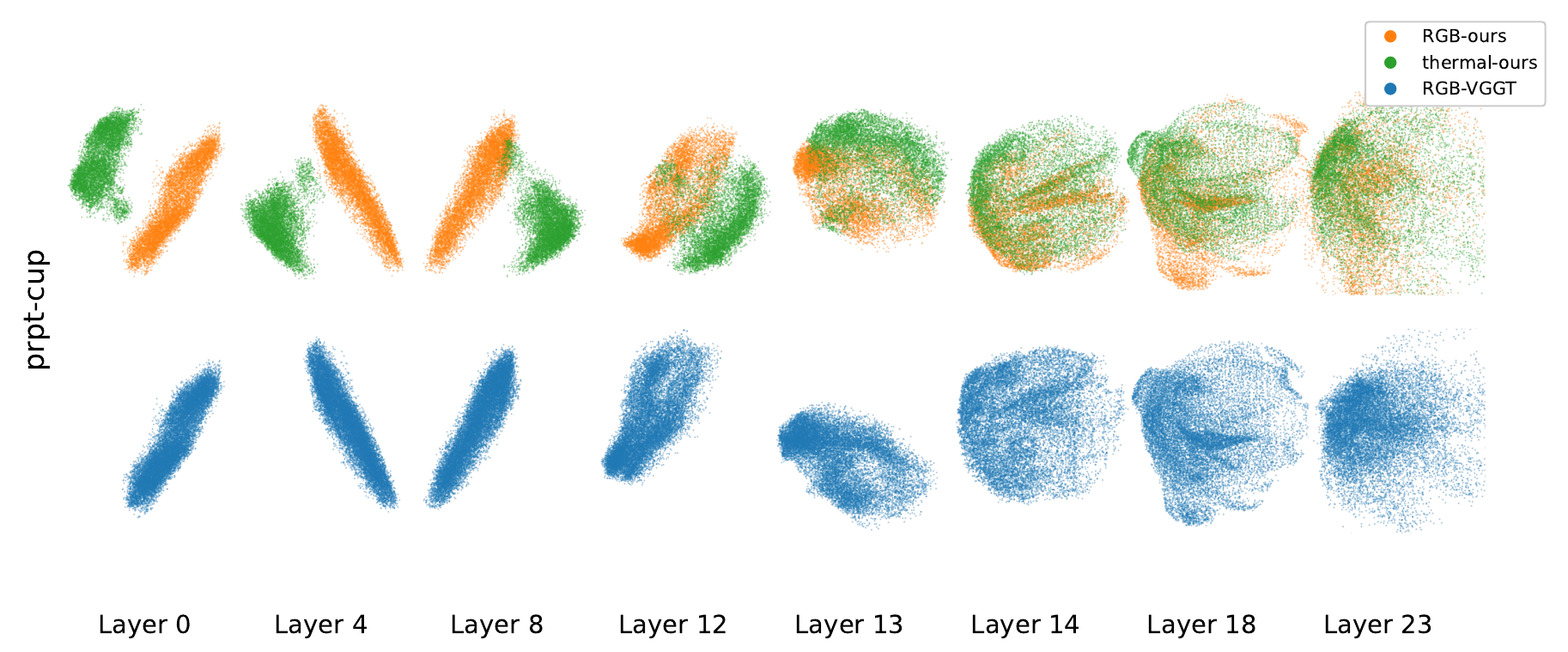}
  \includegraphics[page=1,width=0.95\linewidth]{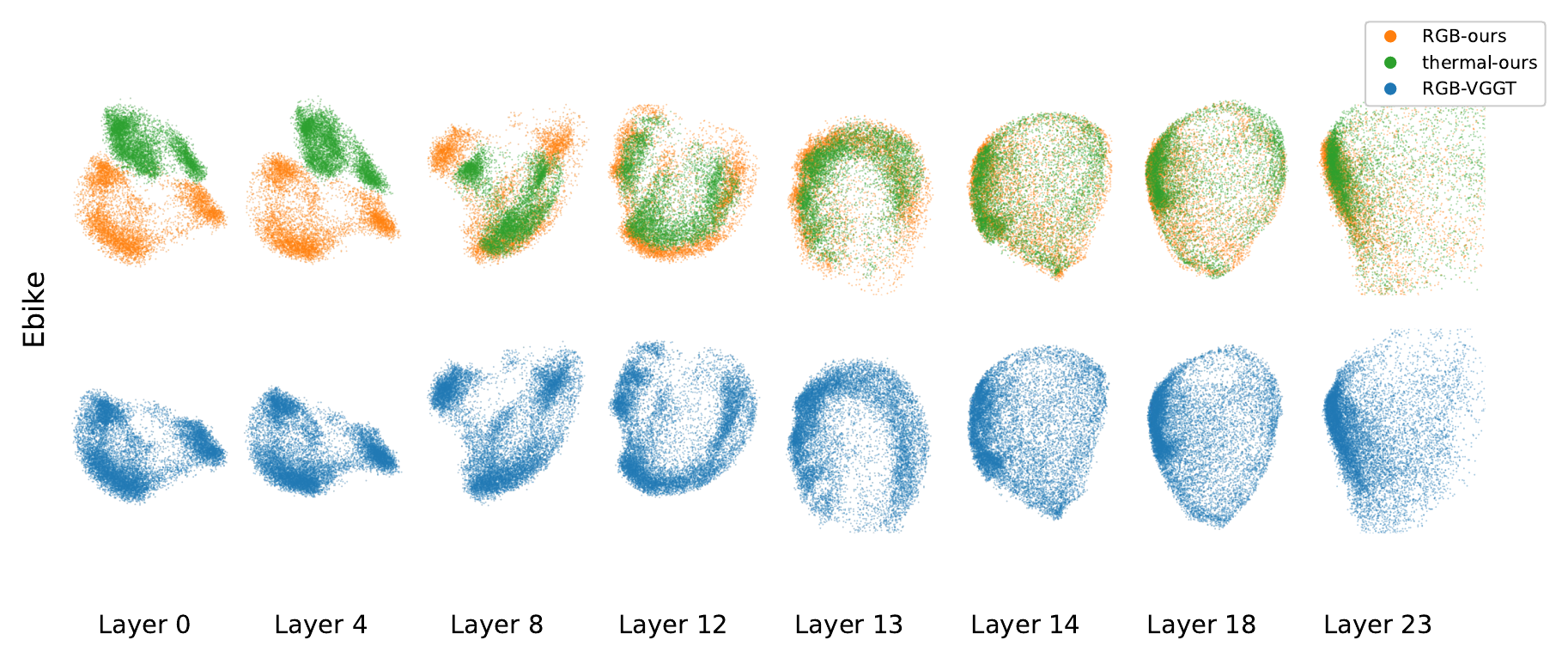}
  \caption{
    PCA-based analysis of RGB-thermal feature alignment. The figure shows three experiments: \texttt{generator} from ThermalNeRF (rows 1-2), \texttt{prpt-cup} from ThermoScenes (rows 3-4), and \texttt{Ebike} from ThermalGaussian (rows 5-6). For each experiment, the top row illustrates the feature evolution of our model on RGB-thermal inputs, while the bottom row shows the feature evolution of VGGT on the corresponding RGB-only inputs.
  }
  \label{fig:generator_ours}
\end{figure}

Our analysis suggests that the model does not learn a completely new representation for mixed-modality inputs, but instead operates largely within the original VGGT feature space. To support this claim, we run \ours on RGB-thermal inputs and the original VGGT on the corresponding RGB-only images and extract the intermediate outputs of the frame-attention layers in the AA modules. We then perform Principal Component Analysis (PCA) on features from the same layer and visualize the resulting embeddings in \cref{fig:generator_ours}. It can be observed that, at each layer, the RGB features produced by \ours and the original pre-trained VGGT model exhibit very similar structures. The \texttt{layers 12--14} clearly show how our model aligns RGB and thermal features. After \texttt{layer14}, the combined RGB and thermal features closely resemble the RGB features produced by the pre-trained VGGT model for RGB-only images.

We also measure the discrepancy between the distributions of thermal and RGB tokens using the KL divergence (specifically its symmetrized form, also referred to as the Jeffreys divergence).
We use the KL distance because the Wasserstein distance for non-parametric distributions is computationally prohibitive due to the large number of tokens and their high dimensionality, and the Maximum Mean Discrepancy (MMD) is known to be sensitive to the kernel choice.
Following a procedure analogous to the Fréchet Inception Distance, we model the tokens of each modality as samples from multivariate Gaussian distributions and estimate the corresponding distribution parameters.

\begin{figure}[!t]
  \centering
  \resizebox{0.95\textwidth}{!}{\input{images/RGB_Thermal_Features/distances_gauss.pgf}}
  \caption{
    The blue line represents the median symmetrized KL-divergence between RGB-thermal tokens dependency across layers for \ours method.
    The orange line represents the same dependency for the VGGT model.
    The filled area represents the boundary from $0.25-$ to $0.75-$quantiles.
  }
  \label{fig:distance_between_thermal_rgb_features_gauss}
\end{figure}
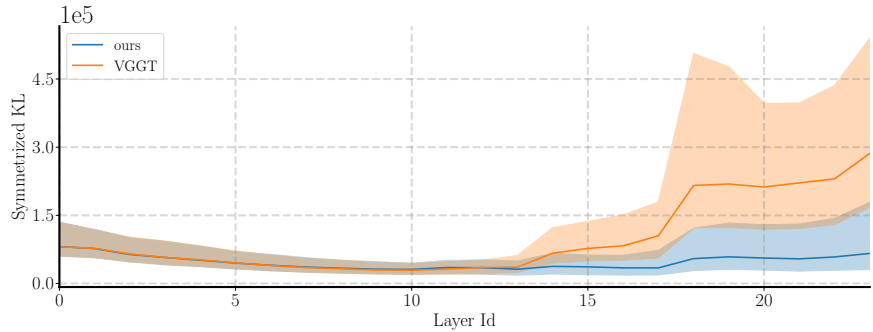

As for the experiments described in \textit{Thermal to RGB Alignment} of the main paper, we feed a mix of RGB-thermal images to \ours and VGGT and store the intermediate outputs of the frame-wise attention layers. We use a batch size of $12$, with the thermal ratio sampled uniformly from $[0.25, 0.75]$. The results in \cref{fig:distance_between_thermal_rgb_features_gauss} show the symmetrized KL divergence across layers. We observe that the divergence follows a similar trend in both models up to approximately \texttt{layer 13}. After that, the divergence rises sharply for VGGT, whereas it remains low for \ours.

%% file: images/RGB_Thermal_Features/distances_gauss.pgf
\begingroup%
\makeatletter%
\begin{pgfpicture}%
\pgfpathrectangle{\pgfpointorigin}{\pgfqpoint{11.450000in}{4.450000in}}%
\pgfusepath{use as bounding box, clip}%
\begin{pgfscope}%
\pgfsetbuttcap%
\pgfsetmiterjoin%
\definecolor{currentfill}{rgb}{1.000000,1.000000,1.000000}%
\pgfsetfillcolor{currentfill}%
\pgfsetlinewidth{0.000000pt}%
\definecolor{currentstroke}{rgb}{1.000000,1.000000,1.000000}%
\pgfsetstrokecolor{currentstroke}%
\pgfsetdash{}{0pt}%
\pgfpathmoveto{\pgfqpoint{0.000000in}{0.000000in}}%
\pgfpathlineto{\pgfqpoint{11.450000in}{0.000000in}}%
\pgfpathlineto{\pgfqpoint{11.450000in}{4.450000in}}%
\pgfpathlineto{\pgfqpoint{0.000000in}{4.450000in}}%
\pgfpathlineto{\pgfqpoint{0.000000in}{0.000000in}}%
\pgfpathclose%
\pgfusepath{fill}%
\end{pgfscope}%
\begin{pgfscope}%
\pgfsetbuttcap%
\pgfsetmiterjoin%
\definecolor{currentfill}{rgb}{1.000000,1.000000,1.000000}%
\pgfsetfillcolor{currentfill}%
\pgfsetlinewidth{0.000000pt}%
\definecolor{currentstroke}{rgb}{0.000000,0.000000,0.000000}%
\pgfsetstrokecolor{currentstroke}%
\pgfsetstrokeopacity{0.000000}%
\pgfsetdash{}{0pt}%
\pgfpathmoveto{\pgfqpoint{0.707138in}{0.642876in}}%
\pgfpathlineto{\pgfqpoint{11.350000in}{0.642876in}}%
\pgfpathlineto{\pgfqpoint{11.350000in}{4.071105in}}%
\pgfpathlineto{\pgfqpoint{0.707138in}{4.071105in}}%
\pgfpathlineto{\pgfqpoint{0.707138in}{0.642876in}}%
\pgfpathclose%
\pgfusepath{fill}%
\end{pgfscope}%
\begin{pgfscope}%
\pgfpathrectangle{\pgfqpoint{0.707138in}{0.642876in}}{\pgfqpoint{10.642862in}{3.428229in}}%
\pgfusepath{clip}%
\pgfsetbuttcap%
\pgfsetroundjoin%
\definecolor{currentfill}{rgb}{0.121569,0.466667,0.705882}%
\pgfsetfillcolor{currentfill}%
\pgfsetfillopacity{0.300000}%
\pgfsetlinewidth{1.003750pt}%
\definecolor{currentstroke}{rgb}{0.121569,0.466667,0.705882}%
\pgfsetstrokecolor{currentstroke}%
\pgfsetstrokeopacity{0.300000}%
\pgfsetdash{}{0pt}%
\pgfsys@defobject{currentmarker}{\pgfqpoint{0.707138in}{0.798704in}}{\pgfqpoint{11.350000in}{1.755575in}}{%
\pgfpathmoveto{\pgfqpoint{0.707138in}{1.494377in}}%
\pgfpathlineto{\pgfqpoint{0.707138in}{1.048815in}}%
\pgfpathlineto{\pgfqpoint{1.169871in}{1.026118in}}%
\pgfpathlineto{\pgfqpoint{1.632604in}{0.970981in}}%
\pgfpathlineto{\pgfqpoint{2.095337in}{0.933678in}}%
\pgfpathlineto{\pgfqpoint{2.558070in}{0.909910in}}%
\pgfpathlineto{\pgfqpoint{3.020803in}{0.880107in}}%
\pgfpathlineto{\pgfqpoint{3.483536in}{0.855808in}}%
\pgfpathlineto{\pgfqpoint{3.946270in}{0.838644in}}%
\pgfpathlineto{\pgfqpoint{4.409003in}{0.825731in}}%
\pgfpathlineto{\pgfqpoint{4.871736in}{0.816634in}}%
\pgfpathlineto{\pgfqpoint{5.334469in}{0.812291in}}%
\pgfpathlineto{\pgfqpoint{5.797202in}{0.820034in}}%
\pgfpathlineto{\pgfqpoint{6.259935in}{0.811882in}}%
\pgfpathlineto{\pgfqpoint{6.722668in}{0.799317in}}%
\pgfpathlineto{\pgfqpoint{7.185402in}{0.814030in}}%
\pgfpathlineto{\pgfqpoint{7.648135in}{0.805255in}}%
\pgfpathlineto{\pgfqpoint{8.110868in}{0.798704in}}%
\pgfpathlineto{\pgfqpoint{8.573601in}{0.801758in}}%
\pgfpathlineto{\pgfqpoint{9.036334in}{0.858689in}}%
\pgfpathlineto{\pgfqpoint{9.499067in}{0.876174in}}%
\pgfpathlineto{\pgfqpoint{9.961801in}{0.864950in}}%
\pgfpathlineto{\pgfqpoint{10.424534in}{0.852631in}}%
\pgfpathlineto{\pgfqpoint{10.887267in}{0.863023in}}%
\pgfpathlineto{\pgfqpoint{11.350000in}{0.871815in}}%
\pgfpathlineto{\pgfqpoint{11.350000in}{1.755575in}}%
\pgfpathlineto{\pgfqpoint{11.350000in}{1.755575in}}%
\pgfpathlineto{\pgfqpoint{10.887267in}{1.545889in}}%
\pgfpathlineto{\pgfqpoint{10.424534in}{1.470173in}}%
\pgfpathlineto{\pgfqpoint{9.961801in}{1.462216in}}%
\pgfpathlineto{\pgfqpoint{9.499067in}{1.483181in}}%
\pgfpathlineto{\pgfqpoint{9.036334in}{1.415294in}}%
\pgfpathlineto{\pgfqpoint{8.573601in}{1.124066in}}%
\pgfpathlineto{\pgfqpoint{8.110868in}{1.059847in}}%
\pgfpathlineto{\pgfqpoint{7.648135in}{1.062676in}}%
\pgfpathlineto{\pgfqpoint{7.185402in}{1.085305in}}%
\pgfpathlineto{\pgfqpoint{6.722668in}{0.984701in}}%
\pgfpathlineto{\pgfqpoint{6.259935in}{0.998401in}}%
\pgfpathlineto{\pgfqpoint{5.797202in}{0.992543in}}%
\pgfpathlineto{\pgfqpoint{5.334469in}{0.956534in}}%
\pgfpathlineto{\pgfqpoint{4.871736in}{0.975862in}}%
\pgfpathlineto{\pgfqpoint{4.409003in}{0.995922in}}%
\pgfpathlineto{\pgfqpoint{3.946270in}{1.028672in}}%
\pgfpathlineto{\pgfqpoint{3.483536in}{1.071404in}}%
\pgfpathlineto{\pgfqpoint{3.020803in}{1.107401in}}%
\pgfpathlineto{\pgfqpoint{2.558070in}{1.180184in}}%
\pgfpathlineto{\pgfqpoint{2.095337in}{1.242941in}}%
\pgfpathlineto{\pgfqpoint{1.632604in}{1.289105in}}%
\pgfpathlineto{\pgfqpoint{1.169871in}{1.400437in}}%
\pgfpathlineto{\pgfqpoint{0.707138in}{1.494377in}}%
\pgfpathlineto{\pgfqpoint{0.707138in}{1.494377in}}%
\pgfpathclose%
\pgfusepath{stroke,fill}%
}%
\begin{pgfscope}%
\pgfsys@transformshift{0.000000in}{0.000000in}%
\pgfsys@useobject{currentmarker}{}%
\end{pgfscope}%
\end{pgfscope}%
\begin{pgfscope}%
\pgfpathrectangle{\pgfqpoint{0.707138in}{0.642876in}}{\pgfqpoint{10.642862in}{3.428229in}}%
\pgfusepath{clip}%
\pgfsetbuttcap%
\pgfsetroundjoin%
\definecolor{currentfill}{rgb}{1.000000,0.498039,0.054902}%
\pgfsetfillcolor{currentfill}%
\pgfsetfillopacity{0.300000}%
\pgfsetlinewidth{1.003750pt}%
\definecolor{currentstroke}{rgb}{1.000000,0.498039,0.054902}%
\pgfsetstrokecolor{currentstroke}%
\pgfsetstrokeopacity{0.300000}%
\pgfsetdash{}{0pt}%
\pgfsys@defobject{currentmarker}{\pgfqpoint{0.707138in}{0.806740in}}{\pgfqpoint{11.350000in}{3.915276in}}{%
\pgfpathmoveto{\pgfqpoint{0.707138in}{1.494023in}}%
\pgfpathlineto{\pgfqpoint{0.707138in}{1.046881in}}%
\pgfpathlineto{\pgfqpoint{1.169871in}{1.032775in}}%
\pgfpathlineto{\pgfqpoint{1.632604in}{0.977336in}}%
\pgfpathlineto{\pgfqpoint{2.095337in}{0.939186in}}%
\pgfpathlineto{\pgfqpoint{2.558070in}{0.913208in}}%
\pgfpathlineto{\pgfqpoint{3.020803in}{0.882210in}}%
\pgfpathlineto{\pgfqpoint{3.483536in}{0.856726in}}%
\pgfpathlineto{\pgfqpoint{3.946270in}{0.836239in}}%
\pgfpathlineto{\pgfqpoint{4.409003in}{0.823967in}}%
\pgfpathlineto{\pgfqpoint{4.871736in}{0.811670in}}%
\pgfpathlineto{\pgfqpoint{5.334469in}{0.806740in}}%
\pgfpathlineto{\pgfqpoint{5.797202in}{0.808591in}}%
\pgfpathlineto{\pgfqpoint{6.259935in}{0.819639in}}%
\pgfpathlineto{\pgfqpoint{6.722668in}{0.834003in}}%
\pgfpathlineto{\pgfqpoint{7.185402in}{0.955946in}}%
\pgfpathlineto{\pgfqpoint{7.648135in}{0.988183in}}%
\pgfpathlineto{\pgfqpoint{8.110868in}{0.995197in}}%
\pgfpathlineto{\pgfqpoint{8.573601in}{1.026134in}}%
\pgfpathlineto{\pgfqpoint{9.036334in}{1.418887in}}%
\pgfpathlineto{\pgfqpoint{9.499067in}{1.425864in}}%
\pgfpathlineto{\pgfqpoint{9.961801in}{1.402083in}}%
\pgfpathlineto{\pgfqpoint{10.424534in}{1.412319in}}%
\pgfpathlineto{\pgfqpoint{10.887267in}{1.463062in}}%
\pgfpathlineto{\pgfqpoint{11.350000in}{1.689004in}}%
\pgfpathlineto{\pgfqpoint{11.350000in}{3.915276in}}%
\pgfpathlineto{\pgfqpoint{11.350000in}{3.915276in}}%
\pgfpathlineto{\pgfqpoint{10.887267in}{3.291930in}}%
\pgfpathlineto{\pgfqpoint{10.424534in}{3.060911in}}%
\pgfpathlineto{\pgfqpoint{9.961801in}{3.058540in}}%
\pgfpathlineto{\pgfqpoint{9.499067in}{3.537222in}}%
\pgfpathlineto{\pgfqpoint{9.036334in}{3.711482in}}%
\pgfpathlineto{\pgfqpoint{8.573601in}{1.760075in}}%
\pgfpathlineto{\pgfqpoint{8.110868in}{1.591323in}}%
\pgfpathlineto{\pgfqpoint{7.648135in}{1.504074in}}%
\pgfpathlineto{\pgfqpoint{7.185402in}{1.423698in}}%
\pgfpathlineto{\pgfqpoint{6.722668in}{1.057569in}}%
\pgfpathlineto{\pgfqpoint{6.259935in}{0.997501in}}%
\pgfpathlineto{\pgfqpoint{5.797202in}{0.974975in}}%
\pgfpathlineto{\pgfqpoint{5.334469in}{0.953246in}}%
\pgfpathlineto{\pgfqpoint{4.871736in}{0.964138in}}%
\pgfpathlineto{\pgfqpoint{4.409003in}{0.992498in}}%
\pgfpathlineto{\pgfqpoint{3.946270in}{1.020343in}}%
\pgfpathlineto{\pgfqpoint{3.483536in}{1.063000in}}%
\pgfpathlineto{\pgfqpoint{3.020803in}{1.117799in}}%
\pgfpathlineto{\pgfqpoint{2.558070in}{1.184472in}}%
\pgfpathlineto{\pgfqpoint{2.095337in}{1.248127in}}%
\pgfpathlineto{\pgfqpoint{1.632604in}{1.300648in}}%
\pgfpathlineto{\pgfqpoint{1.169871in}{1.394354in}}%
\pgfpathlineto{\pgfqpoint{0.707138in}{1.494023in}}%
\pgfpathlineto{\pgfqpoint{0.707138in}{1.494023in}}%
\pgfpathclose%
\pgfusepath{stroke,fill}%
}%
\begin{pgfscope}%
\pgfsys@transformshift{0.000000in}{0.000000in}%
\pgfsys@useobject{currentmarker}{}%
\end{pgfscope}%
\end{pgfscope}%
\begin{pgfscope}%
\pgfpathrectangle{\pgfqpoint{0.707138in}{0.642876in}}{\pgfqpoint{10.642862in}{3.428229in}}%
\pgfusepath{clip}%
\pgfsetbuttcap%
\pgfsetroundjoin%
\pgfsetlinewidth{2.007500pt}%
\definecolor{currentstroke}{rgb}{0.501961,0.501961,0.501961}%
\pgfsetstrokecolor{currentstroke}%
\pgfsetstrokeopacity{0.300000}%
\pgfsetdash{{7.400000pt}{3.200000pt}}{0.000000pt}%
\pgfpathmoveto{\pgfqpoint{0.707138in}{0.642876in}}%
\pgfpathlineto{\pgfqpoint{0.707138in}{4.071105in}}%
\pgfusepath{stroke}%
\end{pgfscope}%
\begin{pgfscope}%
\pgfsetbuttcap%
\pgfsetroundjoin%
\definecolor{currentfill}{rgb}{0.000000,0.000000,0.000000}%
\pgfsetfillcolor{currentfill}%
\pgfsetlinewidth{0.803000pt}%
\definecolor{currentstroke}{rgb}{0.000000,0.000000,0.000000}%
\pgfsetstrokecolor{currentstroke}%
\pgfsetdash{}{0pt}%
\pgfsys@defobject{currentmarker}{\pgfqpoint{0.000000in}{-0.048611in}}{\pgfqpoint{0.000000in}{0.000000in}}{%
\pgfpathmoveto{\pgfqpoint{0.000000in}{0.000000in}}%
\pgfpathlineto{\pgfqpoint{0.000000in}{-0.048611in}}%
\pgfusepath{stroke,fill}%
}%
\begin{pgfscope}%
\pgfsys@transformshift{0.707138in}{0.642876in}%
\pgfsys@useobject{currentmarker}{}%
\end{pgfscope}%
\end{pgfscope}%
\begin{pgfscope}%
\definecolor{textcolor}{rgb}{0.000000,0.000000,0.000000}%
\pgfsetstrokecolor{textcolor}%
\pgfsetfillcolor{textcolor}%
\pgftext[x=0.707138in,y=0.545653in,,top]{\color{textcolor}{\rmfamily\fontsize{16.000000}{19.200000}\selectfont\catcode`\^=\active\def^{\ifmmode\sp\else\^{}\fi}\catcode`\%=\active\def
\end{pgfscope}%
\begin{pgfscope}%
\pgfpathrectangle{\pgfqpoint{0.707138in}{0.642876in}}{\pgfqpoint{10.642862in}{3.428229in}}%
\pgfusepath{clip}%
\pgfsetbuttcap%
\pgfsetroundjoin%
\pgfsetlinewidth{2.007500pt}%
\definecolor{currentstroke}{rgb}{0.501961,0.501961,0.501961}%
\pgfsetstrokecolor{currentstroke}%
\pgfsetstrokeopacity{0.300000}%
\pgfsetdash{{7.400000pt}{3.200000pt}}{0.000000pt}%
\pgfpathmoveto{\pgfqpoint{3.020803in}{0.642876in}}%
\pgfpathlineto{\pgfqpoint{3.020803in}{4.071105in}}%
\pgfusepath{stroke}%
\end{pgfscope}%
\begin{pgfscope}%
\pgfsetbuttcap%
\pgfsetroundjoin%
\definecolor{currentfill}{rgb}{0.000000,0.000000,0.000000}%
\pgfsetfillcolor{currentfill}%
\pgfsetlinewidth{0.803000pt}%
\definecolor{currentstroke}{rgb}{0.000000,0.000000,0.000000}%
\pgfsetstrokecolor{currentstroke}%
\pgfsetdash{}{0pt}%
\pgfsys@defobject{currentmarker}{\pgfqpoint{0.000000in}{-0.048611in}}{\pgfqpoint{0.000000in}{0.000000in}}{%
\pgfpathmoveto{\pgfqpoint{0.000000in}{0.000000in}}%
\pgfpathlineto{\pgfqpoint{0.000000in}{-0.048611in}}%
\pgfusepath{stroke,fill}%
}%
\begin{pgfscope}%
\pgfsys@transformshift{3.020803in}{0.642876in}%
\pgfsys@useobject{currentmarker}{}%
\end{pgfscope}%
\end{pgfscope}%
\begin{pgfscope}%
\definecolor{textcolor}{rgb}{0.000000,0.000000,0.000000}%
\pgfsetstrokecolor{textcolor}%
\pgfsetfillcolor{textcolor}%
\pgftext[x=3.020803in,y=0.545653in,,top]{\color{textcolor}{\rmfamily\fontsize{16.000000}{19.200000}\selectfont\catcode`\^=\active\def^{\ifmmode\sp\else\^{}\fi}\catcode`\%=\active\def
\end{pgfscope}%
\begin{pgfscope}%
\pgfpathrectangle{\pgfqpoint{0.707138in}{0.642876in}}{\pgfqpoint{10.642862in}{3.428229in}}%
\pgfusepath{clip}%
\pgfsetbuttcap%
\pgfsetroundjoin%
\pgfsetlinewidth{2.007500pt}%
\definecolor{currentstroke}{rgb}{0.501961,0.501961,0.501961}%
\pgfsetstrokecolor{currentstroke}%
\pgfsetstrokeopacity{0.300000}%
\pgfsetdash{{7.400000pt}{3.200000pt}}{0.000000pt}%
\pgfpathmoveto{\pgfqpoint{5.334469in}{0.642876in}}%
\pgfpathlineto{\pgfqpoint{5.334469in}{4.071105in}}%
\pgfusepath{stroke}%
\end{pgfscope}%
\begin{pgfscope}%
\pgfsetbuttcap%
\pgfsetroundjoin%
\definecolor{currentfill}{rgb}{0.000000,0.000000,0.000000}%
\pgfsetfillcolor{currentfill}%
\pgfsetlinewidth{0.803000pt}%
\definecolor{currentstroke}{rgb}{0.000000,0.000000,0.000000}%
\pgfsetstrokecolor{currentstroke}%
\pgfsetdash{}{0pt}%
\pgfsys@defobject{currentmarker}{\pgfqpoint{0.000000in}{-0.048611in}}{\pgfqpoint{0.000000in}{0.000000in}}{%
\pgfpathmoveto{\pgfqpoint{0.000000in}{0.000000in}}%
\pgfpathlineto{\pgfqpoint{0.000000in}{-0.048611in}}%
\pgfusepath{stroke,fill}%
}%
\begin{pgfscope}%
\pgfsys@transformshift{5.334469in}{0.642876in}%
\pgfsys@useobject{currentmarker}{}%
\end{pgfscope}%
\end{pgfscope}%
\begin{pgfscope}%
\definecolor{textcolor}{rgb}{0.000000,0.000000,0.000000}%
\pgfsetstrokecolor{textcolor}%
\pgfsetfillcolor{textcolor}%
\pgftext[x=5.334469in,y=0.545653in,,top]{\color{textcolor}{\rmfamily\fontsize{16.000000}{19.200000}\selectfont\catcode`\^=\active\def^{\ifmmode\sp\else\^{}\fi}\catcode`\%=\active\def
\end{pgfscope}%
\begin{pgfscope}%
\pgfpathrectangle{\pgfqpoint{0.707138in}{0.642876in}}{\pgfqpoint{10.642862in}{3.428229in}}%
\pgfusepath{clip}%
\pgfsetbuttcap%
\pgfsetroundjoin%
\pgfsetlinewidth{2.007500pt}%
\definecolor{currentstroke}{rgb}{0.501961,0.501961,0.501961}%
\pgfsetstrokecolor{currentstroke}%
\pgfsetstrokeopacity{0.300000}%
\pgfsetdash{{7.400000pt}{3.200000pt}}{0.000000pt}%
\pgfpathmoveto{\pgfqpoint{7.648135in}{0.642876in}}%
\pgfpathlineto{\pgfqpoint{7.648135in}{4.071105in}}%
\pgfusepath{stroke}%
\end{pgfscope}%
\begin{pgfscope}%
\pgfsetbuttcap%
\pgfsetroundjoin%
\definecolor{currentfill}{rgb}{0.000000,0.000000,0.000000}%
\pgfsetfillcolor{currentfill}%
\pgfsetlinewidth{0.803000pt}%
\definecolor{currentstroke}{rgb}{0.000000,0.000000,0.000000}%
\pgfsetstrokecolor{currentstroke}%
\pgfsetdash{}{0pt}%
\pgfsys@defobject{currentmarker}{\pgfqpoint{0.000000in}{-0.048611in}}{\pgfqpoint{0.000000in}{0.000000in}}{%
\pgfpathmoveto{\pgfqpoint{0.000000in}{0.000000in}}%
\pgfpathlineto{\pgfqpoint{0.000000in}{-0.048611in}}%
\pgfusepath{stroke,fill}%
}%
\begin{pgfscope}%
\pgfsys@transformshift{7.648135in}{0.642876in}%
\pgfsys@useobject{currentmarker}{}%
\end{pgfscope}%
\end{pgfscope}%
\begin{pgfscope}%
\definecolor{textcolor}{rgb}{0.000000,0.000000,0.000000}%
\pgfsetstrokecolor{textcolor}%
\pgfsetfillcolor{textcolor}%
\pgftext[x=7.648135in,y=0.545653in,,top]{\color{textcolor}{\rmfamily\fontsize{16.000000}{19.200000}\selectfont\catcode`\^=\active\def^{\ifmmode\sp\else\^{}\fi}\catcode`\%=\active\def
\end{pgfscope}%
\begin{pgfscope}%
\pgfpathrectangle{\pgfqpoint{0.707138in}{0.642876in}}{\pgfqpoint{10.642862in}{3.428229in}}%
\pgfusepath{clip}%
\pgfsetbuttcap%
\pgfsetroundjoin%
\pgfsetlinewidth{2.007500pt}%
\definecolor{currentstroke}{rgb}{0.501961,0.501961,0.501961}%
\pgfsetstrokecolor{currentstroke}%
\pgfsetstrokeopacity{0.300000}%
\pgfsetdash{{7.400000pt}{3.200000pt}}{0.000000pt}%
\pgfpathmoveto{\pgfqpoint{9.961801in}{0.642876in}}%
\pgfpathlineto{\pgfqpoint{9.961801in}{4.071105in}}%
\pgfusepath{stroke}%
\end{pgfscope}%
\begin{pgfscope}%
\pgfsetbuttcap%
\pgfsetroundjoin%
\definecolor{currentfill}{rgb}{0.000000,0.000000,0.000000}%
\pgfsetfillcolor{currentfill}%
\pgfsetlinewidth{0.803000pt}%
\definecolor{currentstroke}{rgb}{0.000000,0.000000,0.000000}%
\pgfsetstrokecolor{currentstroke}%
\pgfsetdash{}{0pt}%
\pgfsys@defobject{currentmarker}{\pgfqpoint{0.000000in}{-0.048611in}}{\pgfqpoint{0.000000in}{0.000000in}}{%
\pgfpathmoveto{\pgfqpoint{0.000000in}{0.000000in}}%
\pgfpathlineto{\pgfqpoint{0.000000in}{-0.048611in}}%
\pgfusepath{stroke,fill}%
}%
\begin{pgfscope}%
\pgfsys@transformshift{9.961801in}{0.642876in}%
\pgfsys@useobject{currentmarker}{}%
\end{pgfscope}%
\end{pgfscope}%
\begin{pgfscope}%
\definecolor{textcolor}{rgb}{0.000000,0.000000,0.000000}%
\pgfsetstrokecolor{textcolor}%
\pgfsetfillcolor{textcolor}%
\pgftext[x=9.961801in,y=0.545653in,,top]{\color{textcolor}{\rmfamily\fontsize{16.000000}{19.200000}\selectfont\catcode`\^=\active\def^{\ifmmode\sp\else\^{}\fi}\catcode`\%=\active\def
\end{pgfscope}%
\begin{pgfscope}%
\definecolor{textcolor}{rgb}{0.000000,0.000000,0.000000}%
\pgfsetstrokecolor{textcolor}%
\pgfsetfillcolor{textcolor}%
\pgftext[x=6.028569in,y=0.295049in,,top]{\color{textcolor}{\rmfamily\fontsize{16.000000}{19.200000}\selectfont\catcode`\^=\active\def^{\ifmmode\sp\else\^{}\fi}\catcode`\%=\active\def
\end{pgfscope}%
\begin{pgfscope}%
\pgfpathrectangle{\pgfqpoint{0.707138in}{0.642876in}}{\pgfqpoint{10.642862in}{3.428229in}}%
\pgfusepath{clip}%
\pgfsetbuttcap%
\pgfsetroundjoin%
\pgfsetlinewidth{2.007500pt}%
\definecolor{currentstroke}{rgb}{0.501961,0.501961,0.501961}%
\pgfsetstrokecolor{currentstroke}%
\pgfsetstrokeopacity{0.300000}%
\pgfsetdash{{7.400000pt}{3.200000pt}}{0.000000pt}%
\pgfpathmoveto{\pgfqpoint{0.707138in}{0.689327in}}%
\pgfpathlineto{\pgfqpoint{11.350000in}{0.689327in}}%
\pgfusepath{stroke}%
\end{pgfscope}%
\begin{pgfscope}%
\pgfsetbuttcap%
\pgfsetroundjoin%
\definecolor{currentfill}{rgb}{0.000000,0.000000,0.000000}%
\pgfsetfillcolor{currentfill}%
\pgfsetlinewidth{0.803000pt}%
\definecolor{currentstroke}{rgb}{0.000000,0.000000,0.000000}%
\pgfsetstrokecolor{currentstroke}%
\pgfsetdash{}{0pt}%
\pgfsys@defobject{currentmarker}{\pgfqpoint{-0.048611in}{0.000000in}}{\pgfqpoint{-0.000000in}{0.000000in}}{%
\pgfpathmoveto{\pgfqpoint{-0.000000in}{0.000000in}}%
\pgfpathlineto{\pgfqpoint{-0.048611in}{0.000000in}}%
\pgfusepath{stroke,fill}%
}%
\begin{pgfscope}%
\pgfsys@transformshift{0.707138in}{0.689327in}%
\pgfsys@useobject{currentmarker}{}%
\end{pgfscope}%
\end{pgfscope}%
\begin{pgfscope}%
\definecolor{textcolor}{rgb}{0.000000,0.000000,0.000000}%
\pgfsetstrokecolor{textcolor}%
\pgfsetfillcolor{textcolor}%
\pgftext[x=0.350604in, y=0.613414in, left, base]{\color{textcolor}{\rmfamily\fontsize{16.000000}{19.200000}\selectfont\catcode`\^=\active\def^{\ifmmode\sp\else\^{}\fi}\catcode`\%=\active\def
\end{pgfscope}%
\begin{pgfscope}%
\pgfpathrectangle{\pgfqpoint{0.707138in}{0.642876in}}{\pgfqpoint{10.642862in}{3.428229in}}%
\pgfusepath{clip}%
\pgfsetbuttcap%
\pgfsetroundjoin%
\pgfsetlinewidth{2.007500pt}%
\definecolor{currentstroke}{rgb}{0.501961,0.501961,0.501961}%
\pgfsetstrokecolor{currentstroke}%
\pgfsetstrokeopacity{0.300000}%
\pgfsetdash{{7.400000pt}{3.200000pt}}{0.000000pt}%
\pgfpathmoveto{\pgfqpoint{0.707138in}{1.585171in}}%
\pgfpathlineto{\pgfqpoint{11.350000in}{1.585171in}}%
\pgfusepath{stroke}%
\end{pgfscope}%
\begin{pgfscope}%
\pgfsetbuttcap%
\pgfsetroundjoin%
\definecolor{currentfill}{rgb}{0.000000,0.000000,0.000000}%
\pgfsetfillcolor{currentfill}%
\pgfsetlinewidth{0.803000pt}%
\definecolor{currentstroke}{rgb}{0.000000,0.000000,0.000000}%
\pgfsetstrokecolor{currentstroke}%
\pgfsetdash{}{0pt}%
\pgfsys@defobject{currentmarker}{\pgfqpoint{-0.048611in}{0.000000in}}{\pgfqpoint{-0.000000in}{0.000000in}}{%
\pgfpathmoveto{\pgfqpoint{-0.000000in}{0.000000in}}%
\pgfpathlineto{\pgfqpoint{-0.048611in}{0.000000in}}%
\pgfusepath{stroke,fill}%
}%
\begin{pgfscope}%
\pgfsys@transformshift{0.707138in}{1.585171in}%
\pgfsys@useobject{currentmarker}{}%
\end{pgfscope}%
\end{pgfscope}%
\begin{pgfscope}%
\definecolor{textcolor}{rgb}{0.000000,0.000000,0.000000}%
\pgfsetstrokecolor{textcolor}%
\pgfsetfillcolor{textcolor}%
\pgftext[x=0.350604in, y=1.509258in, left, base]{\color{textcolor}{\rmfamily\fontsize{16.000000}{19.200000}\selectfont\catcode`\^=\active\def^{\ifmmode\sp\else\^{}\fi}\catcode`\%=\active\def
\end{pgfscope}%
\begin{pgfscope}%
\pgfpathrectangle{\pgfqpoint{0.707138in}{0.642876in}}{\pgfqpoint{10.642862in}{3.428229in}}%
\pgfusepath{clip}%
\pgfsetbuttcap%
\pgfsetroundjoin%
\pgfsetlinewidth{2.007500pt}%
\definecolor{currentstroke}{rgb}{0.501961,0.501961,0.501961}%
\pgfsetstrokecolor{currentstroke}%
\pgfsetstrokeopacity{0.300000}%
\pgfsetdash{{7.400000pt}{3.200000pt}}{0.000000pt}%
\pgfpathmoveto{\pgfqpoint{0.707138in}{2.481015in}}%
\pgfpathlineto{\pgfqpoint{11.350000in}{2.481015in}}%
\pgfusepath{stroke}%
\end{pgfscope}%
\begin{pgfscope}%
\pgfsetbuttcap%
\pgfsetroundjoin%
\definecolor{currentfill}{rgb}{0.000000,0.000000,0.000000}%
\pgfsetfillcolor{currentfill}%
\pgfsetlinewidth{0.803000pt}%
\definecolor{currentstroke}{rgb}{0.000000,0.000000,0.000000}%
\pgfsetstrokecolor{currentstroke}%
\pgfsetdash{}{0pt}%
\pgfsys@defobject{currentmarker}{\pgfqpoint{-0.048611in}{0.000000in}}{\pgfqpoint{-0.000000in}{0.000000in}}{%
\pgfpathmoveto{\pgfqpoint{-0.000000in}{0.000000in}}%
\pgfpathlineto{\pgfqpoint{-0.048611in}{0.000000in}}%
\pgfusepath{stroke,fill}%
}%
\begin{pgfscope}%
\pgfsys@transformshift{0.707138in}{2.481015in}%
\pgfsys@useobject{currentmarker}{}%
\end{pgfscope}%
\end{pgfscope}%
\begin{pgfscope}%
\definecolor{textcolor}{rgb}{0.000000,0.000000,0.000000}%
\pgfsetstrokecolor{textcolor}%
\pgfsetfillcolor{textcolor}%
\pgftext[x=0.350604in, y=2.405102in, left, base]{\color{textcolor}{\rmfamily\fontsize{16.000000}{19.200000}\selectfont\catcode`\^=\active\def^{\ifmmode\sp\else\^{}\fi}\catcode`\%=\active\def
\end{pgfscope}%
\begin{pgfscope}%
\pgfpathrectangle{\pgfqpoint{0.707138in}{0.642876in}}{\pgfqpoint{10.642862in}{3.428229in}}%
\pgfusepath{clip}%
\pgfsetbuttcap%
\pgfsetroundjoin%
\pgfsetlinewidth{2.007500pt}%
\definecolor{currentstroke}{rgb}{0.501961,0.501961,0.501961}%
\pgfsetstrokecolor{currentstroke}%
\pgfsetstrokeopacity{0.300000}%
\pgfsetdash{{7.400000pt}{3.200000pt}}{0.000000pt}%
\pgfpathmoveto{\pgfqpoint{0.707138in}{3.376859in}}%
\pgfpathlineto{\pgfqpoint{11.350000in}{3.376859in}}%
\pgfusepath{stroke}%
\end{pgfscope}%
\begin{pgfscope}%
\pgfsetbuttcap%
\pgfsetroundjoin%
\definecolor{currentfill}{rgb}{0.000000,0.000000,0.000000}%
\pgfsetfillcolor{currentfill}%
\pgfsetlinewidth{0.803000pt}%
\definecolor{currentstroke}{rgb}{0.000000,0.000000,0.000000}%
\pgfsetstrokecolor{currentstroke}%
\pgfsetdash{}{0pt}%
\pgfsys@defobject{currentmarker}{\pgfqpoint{-0.048611in}{0.000000in}}{\pgfqpoint{-0.000000in}{0.000000in}}{%
\pgfpathmoveto{\pgfqpoint{-0.000000in}{0.000000in}}%
\pgfpathlineto{\pgfqpoint{-0.048611in}{0.000000in}}%
\pgfusepath{stroke,fill}%
}%
\begin{pgfscope}%
\pgfsys@transformshift{0.707138in}{3.376859in}%
\pgfsys@useobject{currentmarker}{}%
\end{pgfscope}%
\end{pgfscope}%
\begin{pgfscope}%
\definecolor{textcolor}{rgb}{0.000000,0.000000,0.000000}%
\pgfsetstrokecolor{textcolor}%
\pgfsetfillcolor{textcolor}%
\pgftext[x=0.350604in, y=3.300946in, left, base]{\color{textcolor}{\rmfamily\fontsize{16.000000}{19.200000}\selectfont\catcode`\^=\active\def^{\ifmmode\sp\else\^{}\fi}\catcode`\%=\active\def
\end{pgfscope}%
\begin{pgfscope}%
\definecolor{textcolor}{rgb}{0.000000,0.000000,0.000000}%
\pgfsetstrokecolor{textcolor}%
\pgfsetfillcolor{textcolor}%
\pgftext[x=0.295049in,y=2.356990in,,bottom,rotate=90.000000]{\color{textcolor}{\rmfamily\fontsize{16.000000}{19.200000}\selectfont\catcode`\^=\active\def^{\ifmmode\sp\else\^{}\fi}\catcode`\%=\active\def
\end{pgfscope}%
\begin{pgfscope}%
\definecolor{textcolor}{rgb}{0.000000,0.000000,0.000000}%
\pgfsetstrokecolor{textcolor}%
\pgfsetfillcolor{textcolor}%
\pgftext[x=0.707138in,y=4.112772in,left,base]{\color{textcolor}{\rmfamily\fontsize{25.000000}{30.000000}\selectfont\catcode`\^=\active\def^{\ifmmode\sp\else\^{}\fi}\catcode`\%=\active\def
\end{pgfscope}%
\begin{pgfscope}%
\pgfpathrectangle{\pgfqpoint{0.707138in}{0.642876in}}{\pgfqpoint{10.642862in}{3.428229in}}%
\pgfusepath{clip}%
\pgfsetrectcap%
\pgfsetroundjoin%
\pgfsetlinewidth{1.505625pt}%
\definecolor{currentstroke}{rgb}{0.121569,0.466667,0.705882}%
\pgfsetstrokecolor{currentstroke}%
\pgfsetdash{}{0pt}%
\pgfpathmoveto{\pgfqpoint{0.707138in}{1.174658in}}%
\pgfpathlineto{\pgfqpoint{1.169871in}{1.147844in}}%
\pgfpathlineto{\pgfqpoint{1.632604in}{1.070869in}}%
\pgfpathlineto{\pgfqpoint{2.095337in}{1.031601in}}%
\pgfpathlineto{\pgfqpoint{2.558070in}{0.994921in}}%
\pgfpathlineto{\pgfqpoint{3.020803in}{0.957888in}}%
\pgfpathlineto{\pgfqpoint{3.483536in}{0.928604in}}%
\pgfpathlineto{\pgfqpoint{3.946270in}{0.904983in}}%
\pgfpathlineto{\pgfqpoint{4.409003in}{0.889924in}}%
\pgfpathlineto{\pgfqpoint{4.871736in}{0.877047in}}%
\pgfpathlineto{\pgfqpoint{5.334469in}{0.874204in}}%
\pgfpathlineto{\pgfqpoint{5.797202in}{0.898724in}}%
\pgfpathlineto{\pgfqpoint{6.259935in}{0.897289in}}%
\pgfpathlineto{\pgfqpoint{6.722668in}{0.876866in}}%
\pgfpathlineto{\pgfqpoint{7.185402in}{0.914350in}}%
\pgfpathlineto{\pgfqpoint{7.648135in}{0.908311in}}%
\pgfpathlineto{\pgfqpoint{8.110868in}{0.894962in}}%
\pgfpathlineto{\pgfqpoint{8.573601in}{0.894155in}}%
\pgfpathlineto{\pgfqpoint{9.036334in}{1.016354in}}%
\pgfpathlineto{\pgfqpoint{9.499067in}{1.039212in}}%
\pgfpathlineto{\pgfqpoint{9.961801in}{1.024095in}}%
\pgfpathlineto{\pgfqpoint{10.424534in}{1.013123in}}%
\pgfpathlineto{\pgfqpoint{10.887267in}{1.038118in}}%
\pgfpathlineto{\pgfqpoint{11.350000in}{1.084538in}}%
\pgfusepath{stroke}%
\end{pgfscope}%
\begin{pgfscope}%
\pgfpathrectangle{\pgfqpoint{0.707138in}{0.642876in}}{\pgfqpoint{10.642862in}{3.428229in}}%
\pgfusepath{clip}%
\pgfsetrectcap%
\pgfsetroundjoin%
\pgfsetlinewidth{1.505625pt}%
\definecolor{currentstroke}{rgb}{1.000000,0.498039,0.054902}%
\pgfsetstrokecolor{currentstroke}%
\pgfsetdash{}{0pt}%
\pgfpathmoveto{\pgfqpoint{0.707138in}{1.172668in}}%
\pgfpathlineto{\pgfqpoint{1.169871in}{1.151380in}}%
\pgfpathlineto{\pgfqpoint{1.632604in}{1.077954in}}%
\pgfpathlineto{\pgfqpoint{2.095337in}{1.031187in}}%
\pgfpathlineto{\pgfqpoint{2.558070in}{0.999942in}}%
\pgfpathlineto{\pgfqpoint{3.020803in}{0.959597in}}%
\pgfpathlineto{\pgfqpoint{3.483536in}{0.926859in}}%
\pgfpathlineto{\pgfqpoint{3.946270in}{0.898096in}}%
\pgfpathlineto{\pgfqpoint{4.409003in}{0.883415in}}%
\pgfpathlineto{\pgfqpoint{4.871736in}{0.866750in}}%
\pgfpathlineto{\pgfqpoint{5.334469in}{0.863971in}}%
\pgfpathlineto{\pgfqpoint{5.797202in}{0.878024in}}%
\pgfpathlineto{\pgfqpoint{6.259935in}{0.899633in}}%
\pgfpathlineto{\pgfqpoint{6.722668in}{0.903539in}}%
\pgfpathlineto{\pgfqpoint{7.185402in}{1.087154in}}%
\pgfpathlineto{\pgfqpoint{7.648135in}{1.150815in}}%
\pgfpathlineto{\pgfqpoint{8.110868in}{1.184410in}}%
\pgfpathlineto{\pgfqpoint{8.573601in}{1.315844in}}%
\pgfpathlineto{\pgfqpoint{9.036334in}{1.977618in}}%
\pgfpathlineto{\pgfqpoint{9.499067in}{1.997205in}}%
\pgfpathlineto{\pgfqpoint{9.961801in}{1.957100in}}%
\pgfpathlineto{\pgfqpoint{10.424534in}{2.011555in}}%
\pgfpathlineto{\pgfqpoint{10.887267in}{2.064064in}}%
\pgfpathlineto{\pgfqpoint{11.350000in}{2.398978in}}%
\pgfusepath{stroke}%
\end{pgfscope}%
\begin{pgfscope}%
\pgfsetrectcap%
\pgfsetmiterjoin%
\pgfsetlinewidth{2.007500pt}%
\definecolor{currentstroke}{rgb}{0.000000,0.000000,0.000000}%
\pgfsetstrokecolor{currentstroke}%
\pgfsetdash{}{0pt}%
\pgfpathmoveto{\pgfqpoint{0.707138in}{0.642876in}}%
\pgfpathlineto{\pgfqpoint{0.707138in}{4.071105in}}%
\pgfusepath{stroke}%
\end{pgfscope}%
\begin{pgfscope}%
\pgfsetrectcap%
\pgfsetmiterjoin%
\pgfsetlinewidth{2.007500pt}%
\definecolor{currentstroke}{rgb}{0.000000,0.000000,0.000000}%
\pgfsetstrokecolor{currentstroke}%
\pgfsetdash{}{0pt}%
\pgfpathmoveto{\pgfqpoint{0.707138in}{0.642876in}}%
\pgfpathlineto{\pgfqpoint{11.350000in}{0.642876in}}%
\pgfusepath{stroke}%
\end{pgfscope}%
\begin{pgfscope}%
\pgfsetbuttcap%
\pgfsetmiterjoin%
\definecolor{currentfill}{rgb}{1.000000,1.000000,1.000000}%
\pgfsetfillcolor{currentfill}%
\pgfsetfillopacity{0.800000}%
\pgfsetlinewidth{1.003750pt}%
\definecolor{currentstroke}{rgb}{0.800000,0.800000,0.800000}%
\pgfsetstrokecolor{currentstroke}%
\pgfsetstrokeopacity{0.800000}%
\pgfsetdash{}{0pt}%
\pgfpathmoveto{\pgfqpoint{0.843249in}{3.373266in}}%
\pgfpathlineto{\pgfqpoint{2.039172in}{3.373266in}}%
\pgfpathquadraticcurveto{\pgfqpoint{2.078061in}{3.373266in}}{\pgfqpoint{2.078061in}{3.412155in}}%
\pgfpathlineto{\pgfqpoint{2.078061in}{3.934994in}}%
\pgfpathquadraticcurveto{\pgfqpoint{2.078061in}{3.973883in}}{\pgfqpoint{2.039172in}{3.973883in}}%
\pgfpathlineto{\pgfqpoint{0.843249in}{3.973883in}}%
\pgfpathquadraticcurveto{\pgfqpoint{0.804360in}{3.973883in}}{\pgfqpoint{0.804360in}{3.934994in}}%
\pgfpathlineto{\pgfqpoint{0.804360in}{3.412155in}}%
\pgfpathquadraticcurveto{\pgfqpoint{0.804360in}{3.373266in}}{\pgfqpoint{0.843249in}{3.373266in}}%
\pgfpathlineto{\pgfqpoint{0.843249in}{3.373266in}}%
\pgfpathclose%
\pgfusepath{stroke,fill}%
\end{pgfscope}%
\begin{pgfscope}%
\pgfsetrectcap%
\pgfsetroundjoin%
\pgfsetlinewidth{1.505625pt}%
\definecolor{currentstroke}{rgb}{0.121569,0.466667,0.705882}%
\pgfsetstrokecolor{currentstroke}%
\pgfsetdash{}{0pt}%
\pgfpathmoveto{\pgfqpoint{0.882138in}{3.828049in}}%
\pgfpathlineto{\pgfqpoint{1.076582in}{3.828049in}}%
\pgfpathlineto{\pgfqpoint{1.271026in}{3.828049in}}%
\pgfusepath{stroke}%
\end{pgfscope}%
\begin{pgfscope}%
\definecolor{textcolor}{rgb}{0.000000,0.000000,0.000000}%
\pgfsetstrokecolor{textcolor}%
\pgfsetfillcolor{textcolor}%
\pgftext[x=1.426582in,y=3.759994in,left,base]{\color{textcolor}{\rmfamily\fontsize{14.000000}{16.800000}\selectfont\catcode`\^=\active\def^{\ifmmode\sp\else\^{}\fi}\catcode`\%=\active\def
\end{pgfscope}%
\begin{pgfscope}%
\pgfsetrectcap%
\pgfsetroundjoin%
\pgfsetlinewidth{1.505625pt}%
\definecolor{currentstroke}{rgb}{1.000000,0.498039,0.054902}%
\pgfsetstrokecolor{currentstroke}%
\pgfsetdash{}{0pt}%
\pgfpathmoveto{\pgfqpoint{0.882138in}{3.556908in}}%
\pgfpathlineto{\pgfqpoint{1.076582in}{3.556908in}}%
\pgfpathlineto{\pgfqpoint{1.271026in}{3.556908in}}%
\pgfusepath{stroke}%
\end{pgfscope}%
\begin{pgfscope}%
\definecolor{textcolor}{rgb}{0.000000,0.000000,0.000000}%
\pgfsetstrokecolor{textcolor}%
\pgfsetfillcolor{textcolor}%
\pgftext[x=1.426582in,y=3.488852in,left,base]{\color{textcolor}{\rmfamily\fontsize{14.000000}{16.800000}\selectfont\catcode`\^=\active\def^{\ifmmode\sp\else\^{}\fi}\catcode`\%=\active\def
\end{pgfscope}%
\end{pgfpicture}%
\makeatother%
\endgroup%

%% file: supplementary/11_thermal_only.tex
\section{Thermal-Only Input}

In this section, we compare the performance of VGGT and \ours when both models are applied to thermal-only inputs. We evaluate the methods on the Public Scenes dataset and report the results in \cref{tab:thermal_only_comparison}. These results show that our method significantly outperforms VGGT, achieving AUC@30 of $71.8$ vs $41.7$, indicating that \ours learned to process thermal images more effectively than the pre-trained VGGT model.

\begin{table}[!t]
  \centering
  \caption{
    Results on the Public Scenes dataset using only thermal modality. The table compares \ours and VGGT when only thermal images are passed to the networks.
  }
  \begin{tabular}{lcccccc}
    \toprule Method & AUC@30 $\uparrow$ & RRA@30 $\uparrow$ & RTA@30 $\uparrow$ & PCA $\downarrow$  & PCC $\downarrow$ & Chamfer $\downarrow$\\ \bottomrule
    VGGT & 41.7 & 75.8 & 67.9 & 0.96 & 0.20 & 0.58 \\
    \ours &  \cellcolor{best}{71.8} & \cellcolor{best}{93.7} & \cellcolor{best}{90.8} &  \cellcolor{best}{0.49} & \cellcolor{best}{0.07} & \cellcolor{best}{0.28} \\ \bottomrule
  \end{tabular}
  \label{tab:thermal_only_comparison}
\end{table}

%% file: supplementary/14_quality_thermal_and_rgb_independently.tex
\section{Varying thermal-to-RGB ratio: Per-Modality Quality Computation}

\input{images/thermal_ratio_per_modality/figure}

In this section, we further investigate the experiment described in Sec.~6.6 by analyzing the performance of the RGB and thermal modalities independently. The results are reported in~\cref{fig:thermal_ratio_per_modality}. We observe that the quality of the RGB metrics decreases monotonically as the thermal ratio increases, with the RGB and thermal performance becoming approximately equal at $\tau \approx 0.9$.

We hypothesize that the degradation in RGB performance is caused by two main factors. First, as the proportion of thermal images increases, the scene coverage provided by RGB images decreases. When this coverage becomes too limited, the RGB reconstruction quality drops substantially, as observed around $\tau \approx 0.75$. Second, thermal and RGB images interact through the global attention mechanism of the AA module. As a result, the model can still relate RGB images to thermal images. However, inaccuracies in the thermal modality, such as ghosting artifacts, may lead to less accurate thermal pose estimates and, in turn, reduce the final RGB reconstruction quality. This also explains why the 3D reconstruction quality for thermal poses decreases when $\tau$ decreases.

%% file: images/thermal_ratio_per_modality/figure.tex
\begin{figure*}[t]
  \centering
  \begin{subfigure}[b]{0.32\textwidth}
    \centering
    \resizebox{\linewidth}{!}{\input{images/thermal_ratio_per_modality/modality_results_mAA_5.pgf}}
    \caption{AUC@5 $\uparrow$}
    
  \end{subfigure}
  \begin{subfigure}[b]{0.32\textwidth}
    \centering
    \resizebox{\linewidth}{!}{\input{images/thermal_ratio_per_modality/modality_results_mAA_15.pgf}}
    \caption{AUC@15 $\uparrow$}
    
  \end{subfigure}
  \begin{subfigure}[b]{0.32\textwidth}
    \centering
    \resizebox{\linewidth}{!}{\input{images/thermal_ratio_per_modality/modality_results_mAA_30.pgf}}
    \caption{AUC@30 $\uparrow$}
  \end{subfigure}
  
  \hfill
  
  \begin{subfigure}[b]{0.32\textwidth}
    \centering
    \resizebox{\linewidth}{!}{\input{images/thermal_ratio_per_modality/modality_results_RRA_5.pgf}}
    \caption{RRA@5 $\uparrow$}
  \end{subfigure}
  \begin{subfigure}[b]{0.32\textwidth}
    \centering
    \resizebox{\linewidth}{!}{\input{images/thermal_ratio_per_modality/modality_results_RRA_15.pgf}}
    \caption{RRA@15 $\uparrow$}
    
  \end{subfigure}
  \begin{subfigure}[b]{0.32\textwidth}
    \centering
    \resizebox{\linewidth}{!}{\input{images/thermal_ratio_per_modality/modality_results_RRA_30.pgf}}
    \caption{RRA@30 $\uparrow$}
  \end{subfigure}

  \hfill

  \begin{subfigure}[b]{0.32\textwidth}
    \centering
    \resizebox{\linewidth}{!}{\input{images/thermal_ratio_per_modality/modality_results_RTA_5.pgf}}
    \caption{RTA@5 $\uparrow$}
  \end{subfigure}
  \begin{subfigure}[b]{0.32\textwidth}
    \centering
    \resizebox{\linewidth}{!}{\input{images/thermal_ratio_per_modality/modality_results_RTA_15.pgf}}
    \caption{RTA@15 $\uparrow$}
    
  \end{subfigure}
  \begin{subfigure}[b]{0.32\textwidth}
    \centering
    \resizebox{\linewidth}{!}{\input{images/thermal_ratio_per_modality/modality_results_RTA_30.pgf}}
    \caption{RTA@30 $\uparrow$}
  \end{subfigure}

  \caption{
    The AUC, RRA, RTA (errors <$30^\circ$, $15^\circ$, $5^\circ$) across modalities and varying thermal-to-RGB image ratios. The accuracy of RGB images decreases when decreasing the amount of RGB images similarly to thermal images.
  }
  \label{fig:thermal_ratio_per_modality}
  \vspace{-6mm}
\end{figure*}

%% file: images/thermal_ratio_per_modality/modality_results_mAA_5.pgf
\begingroup%
\makeatletter%
\begin{pgfpicture}%
\pgfpathrectangle{\pgfpointorigin}{\pgfqpoint{7.450000in}{5.450000in}}%
\pgfusepath{use as bounding box, clip}%
\begin{pgfscope}%
\pgfsetbuttcap%
\pgfsetmiterjoin%
\definecolor{currentfill}{rgb}{1.000000,1.000000,1.000000}%
\pgfsetfillcolor{currentfill}%
\pgfsetlinewidth{0.000000pt}%
\definecolor{currentstroke}{rgb}{1.000000,1.000000,1.000000}%
\pgfsetstrokecolor{currentstroke}%
\pgfsetdash{}{0pt}%
\pgfpathmoveto{\pgfqpoint{0.000000in}{0.000000in}}%
\pgfpathlineto{\pgfqpoint{7.450000in}{0.000000in}}%
\pgfpathlineto{\pgfqpoint{7.450000in}{5.450000in}}%
\pgfpathlineto{\pgfqpoint{0.000000in}{5.450000in}}%
\pgfpathlineto{\pgfqpoint{0.000000in}{0.000000in}}%
\pgfpathclose%
\pgfusepath{fill}%
\end{pgfscope}%
\begin{pgfscope}%
\pgfsetbuttcap%
\pgfsetmiterjoin%
\definecolor{currentfill}{rgb}{1.000000,1.000000,1.000000}%
\pgfsetfillcolor{currentfill}%
\pgfsetlinewidth{0.000000pt}%
\definecolor{currentstroke}{rgb}{0.000000,0.000000,0.000000}%
\pgfsetstrokecolor{currentstroke}%
\pgfsetstrokeopacity{0.000000}%
\pgfsetdash{}{0pt}%
\pgfpathmoveto{\pgfqpoint{0.674954in}{0.862305in}}%
\pgfpathlineto{\pgfqpoint{7.147414in}{0.862305in}}%
\pgfpathlineto{\pgfqpoint{7.147414in}{5.231386in}}%
\pgfpathlineto{\pgfqpoint{0.674954in}{5.231386in}}%
\pgfpathlineto{\pgfqpoint{0.674954in}{0.862305in}}%
\pgfpathclose%
\pgfusepath{fill}%
\end{pgfscope}%
\begin{pgfscope}%
\pgfpathrectangle{\pgfqpoint{0.674954in}{0.862305in}}{\pgfqpoint{6.472460in}{4.369081in}}%
\pgfusepath{clip}%
\pgfsetbuttcap%
\pgfsetroundjoin%
\pgfsetlinewidth{2.007500pt}%
\definecolor{currentstroke}{rgb}{0.501961,0.501961,0.501961}%
\pgfsetstrokecolor{currentstroke}%
\pgfsetstrokeopacity{0.300000}%
\pgfsetdash{{7.400000pt}{3.200000pt}}{0.000000pt}%
\pgfpathmoveto{\pgfqpoint{0.674954in}{0.862305in}}%
\pgfpathlineto{\pgfqpoint{0.674954in}{5.231386in}}%
\pgfusepath{stroke}%
\end{pgfscope}%
\begin{pgfscope}%
\pgfsetbuttcap%
\pgfsetroundjoin%
\definecolor{currentfill}{rgb}{0.000000,0.000000,0.000000}%
\pgfsetfillcolor{currentfill}%
\pgfsetlinewidth{0.803000pt}%
\definecolor{currentstroke}{rgb}{0.000000,0.000000,0.000000}%
\pgfsetstrokecolor{currentstroke}%
\pgfsetdash{}{0pt}%
\pgfsys@defobject{currentmarker}{\pgfqpoint{0.000000in}{-0.048611in}}{\pgfqpoint{0.000000in}{0.000000in}}{%
\pgfpathmoveto{\pgfqpoint{0.000000in}{0.000000in}}%
\pgfpathlineto{\pgfqpoint{0.000000in}{-0.048611in}}%
\pgfusepath{stroke,fill}%
}%
\begin{pgfscope}%
\pgfsys@transformshift{0.674954in}{0.862305in}%
\pgfsys@useobject{currentmarker}{}%
\end{pgfscope}%
\end{pgfscope}%
\begin{pgfscope}%
\definecolor{textcolor}{rgb}{0.000000,0.000000,0.000000}%
\pgfsetstrokecolor{textcolor}%
\pgfsetfillcolor{textcolor}%
\pgftext[x=0.674954in,y=0.765082in,,top]{\color{textcolor}{\rmfamily\fontsize{25.000000}{30.000000}\selectfont\catcode`\^=\active\def^{\ifmmode\sp\else\^{}\fi}\catcode`\%=\active\def
\end{pgfscope}%
\begin{pgfscope}%
\pgfpathrectangle{\pgfqpoint{0.674954in}{0.862305in}}{\pgfqpoint{6.472460in}{4.369081in}}%
\pgfusepath{clip}%
\pgfsetbuttcap%
\pgfsetroundjoin%
\pgfsetlinewidth{2.007500pt}%
\definecolor{currentstroke}{rgb}{0.501961,0.501961,0.501961}%
\pgfsetstrokecolor{currentstroke}%
\pgfsetstrokeopacity{0.300000}%
\pgfsetdash{{7.400000pt}{3.200000pt}}{0.000000pt}%
\pgfpathmoveto{\pgfqpoint{1.969446in}{0.862305in}}%
\pgfpathlineto{\pgfqpoint{1.969446in}{5.231386in}}%
\pgfusepath{stroke}%
\end{pgfscope}%
\begin{pgfscope}%
\pgfsetbuttcap%
\pgfsetroundjoin%
\definecolor{currentfill}{rgb}{0.000000,0.000000,0.000000}%
\pgfsetfillcolor{currentfill}%
\pgfsetlinewidth{0.803000pt}%
\definecolor{currentstroke}{rgb}{0.000000,0.000000,0.000000}%
\pgfsetstrokecolor{currentstroke}%
\pgfsetdash{}{0pt}%
\pgfsys@defobject{currentmarker}{\pgfqpoint{0.000000in}{-0.048611in}}{\pgfqpoint{0.000000in}{0.000000in}}{%
\pgfpathmoveto{\pgfqpoint{0.000000in}{0.000000in}}%
\pgfpathlineto{\pgfqpoint{0.000000in}{-0.048611in}}%
\pgfusepath{stroke,fill}%
}%
\begin{pgfscope}%
\pgfsys@transformshift{1.969446in}{0.862305in}%
\pgfsys@useobject{currentmarker}{}%
\end{pgfscope}%
\end{pgfscope}%
\begin{pgfscope}%
\definecolor{textcolor}{rgb}{0.000000,0.000000,0.000000}%
\pgfsetstrokecolor{textcolor}%
\pgfsetfillcolor{textcolor}%
\pgftext[x=1.969446in,y=0.765082in,,top]{\color{textcolor}{\rmfamily\fontsize{25.000000}{30.000000}\selectfont\catcode`\^=\active\def^{\ifmmode\sp\else\^{}\fi}\catcode`\%=\active\def
\end{pgfscope}%
\begin{pgfscope}%
\pgfpathrectangle{\pgfqpoint{0.674954in}{0.862305in}}{\pgfqpoint{6.472460in}{4.369081in}}%
\pgfusepath{clip}%
\pgfsetbuttcap%
\pgfsetroundjoin%
\pgfsetlinewidth{2.007500pt}%
\definecolor{currentstroke}{rgb}{0.501961,0.501961,0.501961}%
\pgfsetstrokecolor{currentstroke}%
\pgfsetstrokeopacity{0.300000}%
\pgfsetdash{{7.400000pt}{3.200000pt}}{0.000000pt}%
\pgfpathmoveto{\pgfqpoint{3.263938in}{0.862305in}}%
\pgfpathlineto{\pgfqpoint{3.263938in}{5.231386in}}%
\pgfusepath{stroke}%
\end{pgfscope}%
\begin{pgfscope}%
\pgfsetbuttcap%
\pgfsetroundjoin%
\definecolor{currentfill}{rgb}{0.000000,0.000000,0.000000}%
\pgfsetfillcolor{currentfill}%
\pgfsetlinewidth{0.803000pt}%
\definecolor{currentstroke}{rgb}{0.000000,0.000000,0.000000}%
\pgfsetstrokecolor{currentstroke}%
\pgfsetdash{}{0pt}%
\pgfsys@defobject{currentmarker}{\pgfqpoint{0.000000in}{-0.048611in}}{\pgfqpoint{0.000000in}{0.000000in}}{%
\pgfpathmoveto{\pgfqpoint{0.000000in}{0.000000in}}%
\pgfpathlineto{\pgfqpoint{0.000000in}{-0.048611in}}%
\pgfusepath{stroke,fill}%
}%
\begin{pgfscope}%
\pgfsys@transformshift{3.263938in}{0.862305in}%
\pgfsys@useobject{currentmarker}{}%
\end{pgfscope}%
\end{pgfscope}%
\begin{pgfscope}%
\definecolor{textcolor}{rgb}{0.000000,0.000000,0.000000}%
\pgfsetstrokecolor{textcolor}%
\pgfsetfillcolor{textcolor}%
\pgftext[x=3.263938in,y=0.765082in,,top]{\color{textcolor}{\rmfamily\fontsize{25.000000}{30.000000}\selectfont\catcode`\^=\active\def^{\ifmmode\sp\else\^{}\fi}\catcode`\%=\active\def
\end{pgfscope}%
\begin{pgfscope}%
\pgfpathrectangle{\pgfqpoint{0.674954in}{0.862305in}}{\pgfqpoint{6.472460in}{4.369081in}}%
\pgfusepath{clip}%
\pgfsetbuttcap%
\pgfsetroundjoin%
\pgfsetlinewidth{2.007500pt}%
\definecolor{currentstroke}{rgb}{0.501961,0.501961,0.501961}%
\pgfsetstrokecolor{currentstroke}%
\pgfsetstrokeopacity{0.300000}%
\pgfsetdash{{7.400000pt}{3.200000pt}}{0.000000pt}%
\pgfpathmoveto{\pgfqpoint{4.558430in}{0.862305in}}%
\pgfpathlineto{\pgfqpoint{4.558430in}{5.231386in}}%
\pgfusepath{stroke}%
\end{pgfscope}%
\begin{pgfscope}%
\pgfsetbuttcap%
\pgfsetroundjoin%
\definecolor{currentfill}{rgb}{0.000000,0.000000,0.000000}%
\pgfsetfillcolor{currentfill}%
\pgfsetlinewidth{0.803000pt}%
\definecolor{currentstroke}{rgb}{0.000000,0.000000,0.000000}%
\pgfsetstrokecolor{currentstroke}%
\pgfsetdash{}{0pt}%
\pgfsys@defobject{currentmarker}{\pgfqpoint{0.000000in}{-0.048611in}}{\pgfqpoint{0.000000in}{0.000000in}}{%
\pgfpathmoveto{\pgfqpoint{0.000000in}{0.000000in}}%
\pgfpathlineto{\pgfqpoint{0.000000in}{-0.048611in}}%
\pgfusepath{stroke,fill}%
}%
\begin{pgfscope}%
\pgfsys@transformshift{4.558430in}{0.862305in}%
\pgfsys@useobject{currentmarker}{}%
\end{pgfscope}%
\end{pgfscope}%
\begin{pgfscope}%
\definecolor{textcolor}{rgb}{0.000000,0.000000,0.000000}%
\pgfsetstrokecolor{textcolor}%
\pgfsetfillcolor{textcolor}%
\pgftext[x=4.558430in,y=0.765082in,,top]{\color{textcolor}{\rmfamily\fontsize{25.000000}{30.000000}\selectfont\catcode`\^=\active\def^{\ifmmode\sp\else\^{}\fi}\catcode`\%=\active\def
\end{pgfscope}%
\begin{pgfscope}%
\pgfpathrectangle{\pgfqpoint{0.674954in}{0.862305in}}{\pgfqpoint{6.472460in}{4.369081in}}%
\pgfusepath{clip}%
\pgfsetbuttcap%
\pgfsetroundjoin%
\pgfsetlinewidth{2.007500pt}%
\definecolor{currentstroke}{rgb}{0.501961,0.501961,0.501961}%
\pgfsetstrokecolor{currentstroke}%
\pgfsetstrokeopacity{0.300000}%
\pgfsetdash{{7.400000pt}{3.200000pt}}{0.000000pt}%
\pgfpathmoveto{\pgfqpoint{5.852922in}{0.862305in}}%
\pgfpathlineto{\pgfqpoint{5.852922in}{5.231386in}}%
\pgfusepath{stroke}%
\end{pgfscope}%
\begin{pgfscope}%
\pgfsetbuttcap%
\pgfsetroundjoin%
\definecolor{currentfill}{rgb}{0.000000,0.000000,0.000000}%
\pgfsetfillcolor{currentfill}%
\pgfsetlinewidth{0.803000pt}%
\definecolor{currentstroke}{rgb}{0.000000,0.000000,0.000000}%
\pgfsetstrokecolor{currentstroke}%
\pgfsetdash{}{0pt}%
\pgfsys@defobject{currentmarker}{\pgfqpoint{0.000000in}{-0.048611in}}{\pgfqpoint{0.000000in}{0.000000in}}{%
\pgfpathmoveto{\pgfqpoint{0.000000in}{0.000000in}}%
\pgfpathlineto{\pgfqpoint{0.000000in}{-0.048611in}}%
\pgfusepath{stroke,fill}%
}%
\begin{pgfscope}%
\pgfsys@transformshift{5.852922in}{0.862305in}%
\pgfsys@useobject{currentmarker}{}%
\end{pgfscope}%
\end{pgfscope}%
\begin{pgfscope}%
\definecolor{textcolor}{rgb}{0.000000,0.000000,0.000000}%
\pgfsetstrokecolor{textcolor}%
\pgfsetfillcolor{textcolor}%
\pgftext[x=5.852922in,y=0.765082in,,top]{\color{textcolor}{\rmfamily\fontsize{25.000000}{30.000000}\selectfont\catcode`\^=\active\def^{\ifmmode\sp\else\^{}\fi}\catcode`\%=\active\def
\end{pgfscope}%
\begin{pgfscope}%
\pgfpathrectangle{\pgfqpoint{0.674954in}{0.862305in}}{\pgfqpoint{6.472460in}{4.369081in}}%
\pgfusepath{clip}%
\pgfsetbuttcap%
\pgfsetroundjoin%
\pgfsetlinewidth{2.007500pt}%
\definecolor{currentstroke}{rgb}{0.501961,0.501961,0.501961}%
\pgfsetstrokecolor{currentstroke}%
\pgfsetstrokeopacity{0.300000}%
\pgfsetdash{{7.400000pt}{3.200000pt}}{0.000000pt}%
\pgfpathmoveto{\pgfqpoint{7.147414in}{0.862305in}}%
\pgfpathlineto{\pgfqpoint{7.147414in}{5.231386in}}%
\pgfusepath{stroke}%
\end{pgfscope}%
\begin{pgfscope}%
\pgfsetbuttcap%
\pgfsetroundjoin%
\definecolor{currentfill}{rgb}{0.000000,0.000000,0.000000}%
\pgfsetfillcolor{currentfill}%
\pgfsetlinewidth{0.803000pt}%
\definecolor{currentstroke}{rgb}{0.000000,0.000000,0.000000}%
\pgfsetstrokecolor{currentstroke}%
\pgfsetdash{}{0pt}%
\pgfsys@defobject{currentmarker}{\pgfqpoint{0.000000in}{-0.048611in}}{\pgfqpoint{0.000000in}{0.000000in}}{%
\pgfpathmoveto{\pgfqpoint{0.000000in}{0.000000in}}%
\pgfpathlineto{\pgfqpoint{0.000000in}{-0.048611in}}%
\pgfusepath{stroke,fill}%
}%
\begin{pgfscope}%
\pgfsys@transformshift{7.147414in}{0.862305in}%
\pgfsys@useobject{currentmarker}{}%
\end{pgfscope}%
\end{pgfscope}%
\begin{pgfscope}%
\definecolor{textcolor}{rgb}{0.000000,0.000000,0.000000}%
\pgfsetstrokecolor{textcolor}%
\pgfsetfillcolor{textcolor}%
\pgftext[x=7.147414in,y=0.765082in,,top]{\color{textcolor}{\rmfamily\fontsize{25.000000}{30.000000}\selectfont\catcode`\^=\active\def^{\ifmmode\sp\else\^{}\fi}\catcode`\%=\active\def
\end{pgfscope}%
\begin{pgfscope}%
\definecolor{textcolor}{rgb}{0.000000,0.000000,0.000000}%
\pgfsetstrokecolor{textcolor}%
\pgfsetfillcolor{textcolor}%
\pgftext[x=3.911184in,y=0.404763in,,top]{\color{textcolor}{\rmfamily\fontsize{25.000000}{30.000000}\selectfont\catcode`\^=\active\def^{\ifmmode\sp\else\^{}\fi}\catcode`\%=\active\def
\end{pgfscope}%
\begin{pgfscope}%
\pgfpathrectangle{\pgfqpoint{0.674954in}{0.862305in}}{\pgfqpoint{6.472460in}{4.369081in}}%
\pgfusepath{clip}%
\pgfsetbuttcap%
\pgfsetroundjoin%
\pgfsetlinewidth{2.007500pt}%
\definecolor{currentstroke}{rgb}{0.501961,0.501961,0.501961}%
\pgfsetstrokecolor{currentstroke}%
\pgfsetstrokeopacity{0.300000}%
\pgfsetdash{{7.400000pt}{3.200000pt}}{0.000000pt}%
\pgfpathmoveto{\pgfqpoint{0.674954in}{1.954575in}}%
\pgfpathlineto{\pgfqpoint{7.147414in}{1.954575in}}%
\pgfusepath{stroke}%
\end{pgfscope}%
\begin{pgfscope}%
\pgfsetbuttcap%
\pgfsetroundjoin%
\definecolor{currentfill}{rgb}{0.000000,0.000000,0.000000}%
\pgfsetfillcolor{currentfill}%
\pgfsetlinewidth{0.803000pt}%
\definecolor{currentstroke}{rgb}{0.000000,0.000000,0.000000}%
\pgfsetstrokecolor{currentstroke}%
\pgfsetdash{}{0pt}%
\pgfsys@defobject{currentmarker}{\pgfqpoint{-0.048611in}{0.000000in}}{\pgfqpoint{-0.000000in}{0.000000in}}{%
\pgfpathmoveto{\pgfqpoint{-0.000000in}{0.000000in}}%
\pgfpathlineto{\pgfqpoint{-0.048611in}{0.000000in}}%
\pgfusepath{stroke,fill}%
}%
\begin{pgfscope}%
\pgfsys@transformshift{0.674954in}{1.954575in}%
\pgfsys@useobject{currentmarker}{}%
\end{pgfscope}%
\end{pgfscope}%
\begin{pgfscope}%
\definecolor{textcolor}{rgb}{0.000000,0.000000,0.000000}%
\pgfsetstrokecolor{textcolor}%
\pgfsetfillcolor{textcolor}%
\pgftext[x=0.259244in, y=1.835961in, left, base]{\color{textcolor}{\rmfamily\fontsize{25.000000}{30.000000}\selectfont\catcode`\^=\active\def^{\ifmmode\sp\else\^{}\fi}\catcode`\%=\active\def
\end{pgfscope}%
\begin{pgfscope}%
\pgfpathrectangle{\pgfqpoint{0.674954in}{0.862305in}}{\pgfqpoint{6.472460in}{4.369081in}}%
\pgfusepath{clip}%
\pgfsetbuttcap%
\pgfsetroundjoin%
\pgfsetlinewidth{2.007500pt}%
\definecolor{currentstroke}{rgb}{0.501961,0.501961,0.501961}%
\pgfsetstrokecolor{currentstroke}%
\pgfsetstrokeopacity{0.300000}%
\pgfsetdash{{7.400000pt}{3.200000pt}}{0.000000pt}%
\pgfpathmoveto{\pgfqpoint{0.674954in}{3.046845in}}%
\pgfpathlineto{\pgfqpoint{7.147414in}{3.046845in}}%
\pgfusepath{stroke}%
\end{pgfscope}%
\begin{pgfscope}%
\pgfsetbuttcap%
\pgfsetroundjoin%
\definecolor{currentfill}{rgb}{0.000000,0.000000,0.000000}%
\pgfsetfillcolor{currentfill}%
\pgfsetlinewidth{0.803000pt}%
\definecolor{currentstroke}{rgb}{0.000000,0.000000,0.000000}%
\pgfsetstrokecolor{currentstroke}%
\pgfsetdash{}{0pt}%
\pgfsys@defobject{currentmarker}{\pgfqpoint{-0.048611in}{0.000000in}}{\pgfqpoint{-0.000000in}{0.000000in}}{%
\pgfpathmoveto{\pgfqpoint{-0.000000in}{0.000000in}}%
\pgfpathlineto{\pgfqpoint{-0.048611in}{0.000000in}}%
\pgfusepath{stroke,fill}%
}%
\begin{pgfscope}%
\pgfsys@transformshift{0.674954in}{3.046845in}%
\pgfsys@useobject{currentmarker}{}%
\end{pgfscope}%
\end{pgfscope}%
\begin{pgfscope}%
\definecolor{textcolor}{rgb}{0.000000,0.000000,0.000000}%
\pgfsetstrokecolor{textcolor}%
\pgfsetfillcolor{textcolor}%
\pgftext[x=0.259244in, y=2.928231in, left, base]{\color{textcolor}{\rmfamily\fontsize{25.000000}{30.000000}\selectfont\catcode`\^=\active\def^{\ifmmode\sp\else\^{}\fi}\catcode`\%=\active\def
\end{pgfscope}%
\begin{pgfscope}%
\pgfpathrectangle{\pgfqpoint{0.674954in}{0.862305in}}{\pgfqpoint{6.472460in}{4.369081in}}%
\pgfusepath{clip}%
\pgfsetbuttcap%
\pgfsetroundjoin%
\pgfsetlinewidth{2.007500pt}%
\definecolor{currentstroke}{rgb}{0.501961,0.501961,0.501961}%
\pgfsetstrokecolor{currentstroke}%
\pgfsetstrokeopacity{0.300000}%
\pgfsetdash{{7.400000pt}{3.200000pt}}{0.000000pt}%
\pgfpathmoveto{\pgfqpoint{0.674954in}{4.139116in}}%
\pgfpathlineto{\pgfqpoint{7.147414in}{4.139116in}}%
\pgfusepath{stroke}%
\end{pgfscope}%
\begin{pgfscope}%
\pgfsetbuttcap%
\pgfsetroundjoin%
\definecolor{currentfill}{rgb}{0.000000,0.000000,0.000000}%
\pgfsetfillcolor{currentfill}%
\pgfsetlinewidth{0.803000pt}%
\definecolor{currentstroke}{rgb}{0.000000,0.000000,0.000000}%
\pgfsetstrokecolor{currentstroke}%
\pgfsetdash{}{0pt}%
\pgfsys@defobject{currentmarker}{\pgfqpoint{-0.048611in}{0.000000in}}{\pgfqpoint{-0.000000in}{0.000000in}}{%
\pgfpathmoveto{\pgfqpoint{-0.000000in}{0.000000in}}%
\pgfpathlineto{\pgfqpoint{-0.048611in}{0.000000in}}%
\pgfusepath{stroke,fill}%
}%
\begin{pgfscope}%
\pgfsys@transformshift{0.674954in}{4.139116in}%
\pgfsys@useobject{currentmarker}{}%
\end{pgfscope}%
\end{pgfscope}%
\begin{pgfscope}%
\definecolor{textcolor}{rgb}{0.000000,0.000000,0.000000}%
\pgfsetstrokecolor{textcolor}%
\pgfsetfillcolor{textcolor}%
\pgftext[x=0.259244in, y=4.020501in, left, base]{\color{textcolor}{\rmfamily\fontsize{25.000000}{30.000000}\selectfont\catcode`\^=\active\def^{\ifmmode\sp\else\^{}\fi}\catcode`\%=\active\def
\end{pgfscope}%
\begin{pgfscope}%
\pgfpathrectangle{\pgfqpoint{0.674954in}{0.862305in}}{\pgfqpoint{6.472460in}{4.369081in}}%
\pgfusepath{clip}%
\pgfsetbuttcap%
\pgfsetroundjoin%
\pgfsetlinewidth{2.007500pt}%
\definecolor{currentstroke}{rgb}{0.501961,0.501961,0.501961}%
\pgfsetstrokecolor{currentstroke}%
\pgfsetstrokeopacity{0.300000}%
\pgfsetdash{{7.400000pt}{3.200000pt}}{0.000000pt}%
\pgfpathmoveto{\pgfqpoint{0.674954in}{5.231386in}}%
\pgfpathlineto{\pgfqpoint{7.147414in}{5.231386in}}%
\pgfusepath{stroke}%
\end{pgfscope}%
\begin{pgfscope}%
\pgfsetbuttcap%
\pgfsetroundjoin%
\definecolor{currentfill}{rgb}{0.000000,0.000000,0.000000}%
\pgfsetfillcolor{currentfill}%
\pgfsetlinewidth{0.803000pt}%
\definecolor{currentstroke}{rgb}{0.000000,0.000000,0.000000}%
\pgfsetstrokecolor{currentstroke}%
\pgfsetdash{}{0pt}%
\pgfsys@defobject{currentmarker}{\pgfqpoint{-0.048611in}{0.000000in}}{\pgfqpoint{-0.000000in}{0.000000in}}{%
\pgfpathmoveto{\pgfqpoint{-0.000000in}{0.000000in}}%
\pgfpathlineto{\pgfqpoint{-0.048611in}{0.000000in}}%
\pgfusepath{stroke,fill}%
}%
\begin{pgfscope}%
\pgfsys@transformshift{0.674954in}{5.231386in}%
\pgfsys@useobject{currentmarker}{}%
\end{pgfscope}%
\end{pgfscope}%
\begin{pgfscope}%
\definecolor{textcolor}{rgb}{0.000000,0.000000,0.000000}%
\pgfsetstrokecolor{textcolor}%
\pgfsetfillcolor{textcolor}%
\pgftext[x=0.100000in, y=5.112772in, left, base]{\color{textcolor}{\rmfamily\fontsize{25.000000}{30.000000}\selectfont\catcode`\^=\active\def^{\ifmmode\sp\else\^{}\fi}\catcode`\%=\active\def
\end{pgfscope}%
\begin{pgfscope}%
\pgfpathrectangle{\pgfqpoint{0.674954in}{0.862305in}}{\pgfqpoint{6.472460in}{4.369081in}}%
\pgfusepath{clip}%
\pgfsetrectcap%
\pgfsetroundjoin%
\pgfsetlinewidth{2.509375pt}%
\definecolor{currentstroke}{rgb}{0.050980,0.415686,0.509804}%
\pgfsetstrokecolor{currentstroke}%
\pgfsetdash{}{0pt}%
\pgfpathmoveto{\pgfqpoint{0.674954in}{3.329898in}}%
\pgfpathlineto{\pgfqpoint{0.998577in}{3.297888in}}%
\pgfpathlineto{\pgfqpoint{2.293069in}{3.095483in}}%
\pgfpathlineto{\pgfqpoint{3.911184in}{2.903986in}}%
\pgfpathlineto{\pgfqpoint{5.529299in}{2.764438in}}%
\pgfpathlineto{\pgfqpoint{6.823791in}{2.274721in}}%
\pgfusepath{stroke}%
\end{pgfscope}%
\begin{pgfscope}%
\pgfpathrectangle{\pgfqpoint{0.674954in}{0.862305in}}{\pgfqpoint{6.472460in}{4.369081in}}%
\pgfusepath{clip}%
\pgfsetbuttcap%
\pgfsetroundjoin%
\definecolor{currentfill}{rgb}{0.050980,0.415686,0.509804}%
\pgfsetfillcolor{currentfill}%
\pgfsetlinewidth{1.003750pt}%
\definecolor{currentstroke}{rgb}{0.050980,0.415686,0.509804}%
\pgfsetstrokecolor{currentstroke}%
\pgfsetdash{}{0pt}%
\pgfsys@defobject{currentmarker}{\pgfqpoint{-0.055556in}{-0.055556in}}{\pgfqpoint{0.055556in}{0.055556in}}{%
\pgfpathmoveto{\pgfqpoint{0.000000in}{-0.055556in}}%
\pgfpathcurveto{\pgfqpoint{0.014734in}{-0.055556in}}{\pgfqpoint{0.028866in}{-0.049702in}}{\pgfqpoint{0.039284in}{-0.039284in}}%
\pgfpathcurveto{\pgfqpoint{0.049702in}{-0.028866in}}{\pgfqpoint{0.055556in}{-0.014734in}}{\pgfqpoint{0.055556in}{0.000000in}}%
\pgfpathcurveto{\pgfqpoint{0.055556in}{0.014734in}}{\pgfqpoint{0.049702in}{0.028866in}}{\pgfqpoint{0.039284in}{0.039284in}}%
\pgfpathcurveto{\pgfqpoint{0.028866in}{0.049702in}}{\pgfqpoint{0.014734in}{0.055556in}}{\pgfqpoint{0.000000in}{0.055556in}}%
\pgfpathcurveto{\pgfqpoint{-0.014734in}{0.055556in}}{\pgfqpoint{-0.028866in}{0.049702in}}{\pgfqpoint{-0.039284in}{0.039284in}}%
\pgfpathcurveto{\pgfqpoint{-0.049702in}{0.028866in}}{\pgfqpoint{-0.055556in}{0.014734in}}{\pgfqpoint{-0.055556in}{0.000000in}}%
\pgfpathcurveto{\pgfqpoint{-0.055556in}{-0.014734in}}{\pgfqpoint{-0.049702in}{-0.028866in}}{\pgfqpoint{-0.039284in}{-0.039284in}}%
\pgfpathcurveto{\pgfqpoint{-0.028866in}{-0.049702in}}{\pgfqpoint{-0.014734in}{-0.055556in}}{\pgfqpoint{0.000000in}{-0.055556in}}%
\pgfpathlineto{\pgfqpoint{0.000000in}{-0.055556in}}%
\pgfpathclose%
\pgfusepath{stroke,fill}%
}%
\begin{pgfscope}%
\pgfsys@transformshift{0.674954in}{3.329898in}%
\pgfsys@useobject{currentmarker}{}%
\end{pgfscope}%
\begin{pgfscope}%
\pgfsys@transformshift{0.998577in}{3.297888in}%
\pgfsys@useobject{currentmarker}{}%
\end{pgfscope}%
\begin{pgfscope}%
\pgfsys@transformshift{2.293069in}{3.095483in}%
\pgfsys@useobject{currentmarker}{}%
\end{pgfscope}%
\begin{pgfscope}%
\pgfsys@transformshift{3.911184in}{2.903986in}%
\pgfsys@useobject{currentmarker}{}%
\end{pgfscope}%
\begin{pgfscope}%
\pgfsys@transformshift{5.529299in}{2.764438in}%
\pgfsys@useobject{currentmarker}{}%
\end{pgfscope}%
\begin{pgfscope}%
\pgfsys@transformshift{6.823791in}{2.274721in}%
\pgfsys@useobject{currentmarker}{}%
\end{pgfscope}%
\end{pgfscope}%
\begin{pgfscope}%
\pgfpathrectangle{\pgfqpoint{0.674954in}{0.862305in}}{\pgfqpoint{6.472460in}{4.369081in}}%
\pgfusepath{clip}%
\pgfsetbuttcap%
\pgfsetroundjoin%
\pgfsetlinewidth{2.509375pt}%
\definecolor{currentstroke}{rgb}{0.960784,0.462745,0.000000}%
\pgfsetstrokecolor{currentstroke}%
\pgfsetdash{{9.250000pt}{4.000000pt}}{0.000000pt}%
\pgfpathmoveto{\pgfqpoint{0.998577in}{2.006937in}}%
\pgfpathlineto{\pgfqpoint{2.293069in}{2.148293in}}%
\pgfpathlineto{\pgfqpoint{3.911184in}{2.225562in}}%
\pgfpathlineto{\pgfqpoint{5.529299in}{2.252906in}}%
\pgfpathlineto{\pgfqpoint{6.823791in}{2.337603in}}%
\pgfpathlineto{\pgfqpoint{7.147414in}{2.343134in}}%
\pgfusepath{stroke}%
\end{pgfscope}%
\begin{pgfscope}%
\pgfpathrectangle{\pgfqpoint{0.674954in}{0.862305in}}{\pgfqpoint{6.472460in}{4.369081in}}%
\pgfusepath{clip}%
\pgfsetbuttcap%
\pgfsetmiterjoin%
\definecolor{currentfill}{rgb}{0.960784,0.462745,0.000000}%
\pgfsetfillcolor{currentfill}%
\pgfsetlinewidth{1.003750pt}%
\definecolor{currentstroke}{rgb}{0.960784,0.462745,0.000000}%
\pgfsetstrokecolor{currentstroke}%
\pgfsetdash{}{0pt}%
\pgfsys@defobject{currentmarker}{\pgfqpoint{-0.055556in}{-0.055556in}}{\pgfqpoint{0.055556in}{0.055556in}}{%
\pgfpathmoveto{\pgfqpoint{-0.055556in}{-0.055556in}}%
\pgfpathlineto{\pgfqpoint{0.055556in}{-0.055556in}}%
\pgfpathlineto{\pgfqpoint{0.055556in}{0.055556in}}%
\pgfpathlineto{\pgfqpoint{-0.055556in}{0.055556in}}%
\pgfpathlineto{\pgfqpoint{-0.055556in}{-0.055556in}}%
\pgfpathclose%
\pgfusepath{stroke,fill}%
}%
\begin{pgfscope}%
\pgfsys@transformshift{0.998577in}{2.006937in}%
\pgfsys@useobject{currentmarker}{}%
\end{pgfscope}%
\begin{pgfscope}%
\pgfsys@transformshift{2.293069in}{2.148293in}%
\pgfsys@useobject{currentmarker}{}%
\end{pgfscope}%
\begin{pgfscope}%
\pgfsys@transformshift{3.911184in}{2.225562in}%
\pgfsys@useobject{currentmarker}{}%
\end{pgfscope}%
\begin{pgfscope}%
\pgfsys@transformshift{5.529299in}{2.252906in}%
\pgfsys@useobject{currentmarker}{}%
\end{pgfscope}%
\begin{pgfscope}%
\pgfsys@transformshift{6.823791in}{2.337603in}%
\pgfsys@useobject{currentmarker}{}%
\end{pgfscope}%
\begin{pgfscope}%
\pgfsys@transformshift{7.147414in}{2.343134in}%
\pgfsys@useobject{currentmarker}{}%
\end{pgfscope}%
\end{pgfscope}%
\begin{pgfscope}%
\pgfsetrectcap%
\pgfsetmiterjoin%
\pgfsetlinewidth{2.007500pt}%
\definecolor{currentstroke}{rgb}{0.000000,0.000000,0.000000}%
\pgfsetstrokecolor{currentstroke}%
\pgfsetdash{}{0pt}%
\pgfpathmoveto{\pgfqpoint{0.674954in}{0.862305in}}%
\pgfpathlineto{\pgfqpoint{0.674954in}{5.231386in}}%
\pgfusepath{stroke}%
\end{pgfscope}%
\begin{pgfscope}%
\pgfsetrectcap%
\pgfsetmiterjoin%
\pgfsetlinewidth{2.007500pt}%
\definecolor{currentstroke}{rgb}{0.000000,0.000000,0.000000}%
\pgfsetstrokecolor{currentstroke}%
\pgfsetdash{}{0pt}%
\pgfpathmoveto{\pgfqpoint{0.674954in}{0.862305in}}%
\pgfpathlineto{\pgfqpoint{7.147414in}{0.862305in}}%
\pgfusepath{stroke}%
\end{pgfscope}%
\begin{pgfscope}%
\pgfsetbuttcap%
\pgfsetmiterjoin%
\definecolor{currentfill}{rgb}{1.000000,1.000000,1.000000}%
\pgfsetfillcolor{currentfill}%
\pgfsetlinewidth{1.003750pt}%
\definecolor{currentstroke}{rgb}{0.800000,0.800000,0.800000}%
\pgfsetstrokecolor{currentstroke}%
\pgfsetdash{}{0pt}%
\pgfpathmoveto{\pgfqpoint{0.869398in}{1.001194in}}%
\pgfpathlineto{\pgfqpoint{2.708109in}{1.001194in}}%
\pgfpathquadraticcurveto{\pgfqpoint{2.763664in}{1.001194in}}{\pgfqpoint{2.763664in}{1.056749in}}%
\pgfpathlineto{\pgfqpoint{2.763664in}{1.803694in}}%
\pgfpathquadraticcurveto{\pgfqpoint{2.763664in}{1.859250in}}{\pgfqpoint{2.708109in}{1.859250in}}%
\pgfpathlineto{\pgfqpoint{0.869398in}{1.859250in}}%
\pgfpathquadraticcurveto{\pgfqpoint{0.813843in}{1.859250in}}{\pgfqpoint{0.813843in}{1.803694in}}%
\pgfpathlineto{\pgfqpoint{0.813843in}{1.056749in}}%
\pgfpathquadraticcurveto{\pgfqpoint{0.813843in}{1.001194in}}{\pgfqpoint{0.869398in}{1.001194in}}%
\pgfpathlineto{\pgfqpoint{0.869398in}{1.001194in}}%
\pgfpathclose%
\pgfusepath{stroke,fill}%
\end{pgfscope}%
\begin{pgfscope}%
\pgfsetrectcap%
\pgfsetroundjoin%
\pgfsetlinewidth{2.509375pt}%
\definecolor{currentstroke}{rgb}{0.050980,0.415686,0.509804}%
\pgfsetstrokecolor{currentstroke}%
\pgfsetdash{}{0pt}%
\pgfpathmoveto{\pgfqpoint{0.924954in}{1.650917in}}%
\pgfpathlineto{\pgfqpoint{1.202732in}{1.650917in}}%
\pgfpathlineto{\pgfqpoint{1.480509in}{1.650917in}}%
\pgfusepath{stroke}%
\end{pgfscope}%
\begin{pgfscope}%
\pgfsetbuttcap%
\pgfsetroundjoin%
\definecolor{currentfill}{rgb}{0.050980,0.415686,0.509804}%
\pgfsetfillcolor{currentfill}%
\pgfsetlinewidth{1.003750pt}%
\definecolor{currentstroke}{rgb}{0.050980,0.415686,0.509804}%
\pgfsetstrokecolor{currentstroke}%
\pgfsetdash{}{0pt}%
\pgfsys@defobject{currentmarker}{\pgfqpoint{-0.055556in}{-0.055556in}}{\pgfqpoint{0.055556in}{0.055556in}}{%
\pgfpathmoveto{\pgfqpoint{0.000000in}{-0.055556in}}%
\pgfpathcurveto{\pgfqpoint{0.014734in}{-0.055556in}}{\pgfqpoint{0.028866in}{-0.049702in}}{\pgfqpoint{0.039284in}{-0.039284in}}%
\pgfpathcurveto{\pgfqpoint{0.049702in}{-0.028866in}}{\pgfqpoint{0.055556in}{-0.014734in}}{\pgfqpoint{0.055556in}{0.000000in}}%
\pgfpathcurveto{\pgfqpoint{0.055556in}{0.014734in}}{\pgfqpoint{0.049702in}{0.028866in}}{\pgfqpoint{0.039284in}{0.039284in}}%
\pgfpathcurveto{\pgfqpoint{0.028866in}{0.049702in}}{\pgfqpoint{0.014734in}{0.055556in}}{\pgfqpoint{0.000000in}{0.055556in}}%
\pgfpathcurveto{\pgfqpoint{-0.014734in}{0.055556in}}{\pgfqpoint{-0.028866in}{0.049702in}}{\pgfqpoint{-0.039284in}{0.039284in}}%
\pgfpathcurveto{\pgfqpoint{-0.049702in}{0.028866in}}{\pgfqpoint{-0.055556in}{0.014734in}}{\pgfqpoint{-0.055556in}{0.000000in}}%
\pgfpathcurveto{\pgfqpoint{-0.055556in}{-0.014734in}}{\pgfqpoint{-0.049702in}{-0.028866in}}{\pgfqpoint{-0.039284in}{-0.039284in}}%
\pgfpathcurveto{\pgfqpoint{-0.028866in}{-0.049702in}}{\pgfqpoint{-0.014734in}{-0.055556in}}{\pgfqpoint{0.000000in}{-0.055556in}}%
\pgfpathlineto{\pgfqpoint{0.000000in}{-0.055556in}}%
\pgfpathclose%
\pgfusepath{stroke,fill}%
}%
\begin{pgfscope}%
\pgfsys@transformshift{1.202732in}{1.650917in}%
\pgfsys@useobject{currentmarker}{}%
\end{pgfscope}%
\end{pgfscope}%
\begin{pgfscope}%
\definecolor{textcolor}{rgb}{0.000000,0.000000,0.000000}%
\pgfsetstrokecolor{textcolor}%
\pgfsetfillcolor{textcolor}%
\pgftext[x=1.702732in,y=1.553694in,left,base]{\color{textcolor}{\rmfamily\fontsize{20.000000}{24.000000}\selectfont\catcode`\^=\active\def^{\ifmmode\sp\else\^{}\fi}\catcode`\%=\active\def
\end{pgfscope}%
\begin{pgfscope}%
\pgfsetbuttcap%
\pgfsetroundjoin%
\pgfsetlinewidth{2.509375pt}%
\definecolor{currentstroke}{rgb}{0.960784,0.462745,0.000000}%
\pgfsetstrokecolor{currentstroke}%
\pgfsetdash{{9.250000pt}{4.000000pt}}{0.000000pt}%
\pgfpathmoveto{\pgfqpoint{0.924954in}{1.263555in}}%
\pgfpathlineto{\pgfqpoint{1.202732in}{1.263555in}}%
\pgfpathlineto{\pgfqpoint{1.480509in}{1.263555in}}%
\pgfusepath{stroke}%
\end{pgfscope}%
\begin{pgfscope}%
\pgfsetbuttcap%
\pgfsetmiterjoin%
\definecolor{currentfill}{rgb}{0.960784,0.462745,0.000000}%
\pgfsetfillcolor{currentfill}%
\pgfsetlinewidth{1.003750pt}%
\definecolor{currentstroke}{rgb}{0.960784,0.462745,0.000000}%
\pgfsetstrokecolor{currentstroke}%
\pgfsetdash{}{0pt}%
\pgfsys@defobject{currentmarker}{\pgfqpoint{-0.055556in}{-0.055556in}}{\pgfqpoint{0.055556in}{0.055556in}}{%
\pgfpathmoveto{\pgfqpoint{-0.055556in}{-0.055556in}}%
\pgfpathlineto{\pgfqpoint{0.055556in}{-0.055556in}}%
\pgfpathlineto{\pgfqpoint{0.055556in}{0.055556in}}%
\pgfpathlineto{\pgfqpoint{-0.055556in}{0.055556in}}%
\pgfpathlineto{\pgfqpoint{-0.055556in}{-0.055556in}}%
\pgfpathclose%
\pgfusepath{stroke,fill}%
}%
\begin{pgfscope}%
\pgfsys@transformshift{1.202732in}{1.263555in}%
\pgfsys@useobject{currentmarker}{}%
\end{pgfscope}%
\end{pgfscope}%
\begin{pgfscope}%
\definecolor{textcolor}{rgb}{0.000000,0.000000,0.000000}%
\pgfsetstrokecolor{textcolor}%
\pgfsetfillcolor{textcolor}%
\pgftext[x=1.702732in,y=1.166333in,left,base]{\color{textcolor}{\rmfamily\fontsize{20.000000}{24.000000}\selectfont\catcode`\^=\active\def^{\ifmmode\sp\else\^{}\fi}\catcode`\%=\active\def
\end{pgfscope}%
\end{pgfpicture}%
\makeatother%
\endgroup%

%% file: images/thermal_ratio_per_modality/modality_results_mAA_15.pgf
\begingroup%
\makeatletter%
\begin{pgfpicture}%
\pgfpathrectangle{\pgfpointorigin}{\pgfqpoint{7.450000in}{5.450000in}}%
\pgfusepath{use as bounding box, clip}%
\begin{pgfscope}%
\pgfsetbuttcap%
\pgfsetmiterjoin%
\definecolor{currentfill}{rgb}{1.000000,1.000000,1.000000}%
\pgfsetfillcolor{currentfill}%
\pgfsetlinewidth{0.000000pt}%
\definecolor{currentstroke}{rgb}{1.000000,1.000000,1.000000}%
\pgfsetstrokecolor{currentstroke}%
\pgfsetdash{}{0pt}%
\pgfpathmoveto{\pgfqpoint{0.000000in}{0.000000in}}%
\pgfpathlineto{\pgfqpoint{7.450000in}{0.000000in}}%
\pgfpathlineto{\pgfqpoint{7.450000in}{5.450000in}}%
\pgfpathlineto{\pgfqpoint{0.000000in}{5.450000in}}%
\pgfpathlineto{\pgfqpoint{0.000000in}{0.000000in}}%
\pgfpathclose%
\pgfusepath{fill}%
\end{pgfscope}%
\begin{pgfscope}%
\pgfsetbuttcap%
\pgfsetmiterjoin%
\definecolor{currentfill}{rgb}{1.000000,1.000000,1.000000}%
\pgfsetfillcolor{currentfill}%
\pgfsetlinewidth{0.000000pt}%
\definecolor{currentstroke}{rgb}{0.000000,0.000000,0.000000}%
\pgfsetstrokecolor{currentstroke}%
\pgfsetstrokeopacity{0.000000}%
\pgfsetdash{}{0pt}%
\pgfpathmoveto{\pgfqpoint{0.674954in}{0.862305in}}%
\pgfpathlineto{\pgfqpoint{7.147414in}{0.862305in}}%
\pgfpathlineto{\pgfqpoint{7.147414in}{5.231386in}}%
\pgfpathlineto{\pgfqpoint{0.674954in}{5.231386in}}%
\pgfpathlineto{\pgfqpoint{0.674954in}{0.862305in}}%
\pgfpathclose%
\pgfusepath{fill}%
\end{pgfscope}%
\begin{pgfscope}%
\pgfpathrectangle{\pgfqpoint{0.674954in}{0.862305in}}{\pgfqpoint{6.472460in}{4.369081in}}%
\pgfusepath{clip}%
\pgfsetbuttcap%
\pgfsetroundjoin%
\pgfsetlinewidth{2.007500pt}%
\definecolor{currentstroke}{rgb}{0.501961,0.501961,0.501961}%
\pgfsetstrokecolor{currentstroke}%
\pgfsetstrokeopacity{0.300000}%
\pgfsetdash{{7.400000pt}{3.200000pt}}{0.000000pt}%
\pgfpathmoveto{\pgfqpoint{0.674954in}{0.862305in}}%
\pgfpathlineto{\pgfqpoint{0.674954in}{5.231386in}}%
\pgfusepath{stroke}%
\end{pgfscope}%
\begin{pgfscope}%
\pgfsetbuttcap%
\pgfsetroundjoin%
\definecolor{currentfill}{rgb}{0.000000,0.000000,0.000000}%
\pgfsetfillcolor{currentfill}%
\pgfsetlinewidth{0.803000pt}%
\definecolor{currentstroke}{rgb}{0.000000,0.000000,0.000000}%
\pgfsetstrokecolor{currentstroke}%
\pgfsetdash{}{0pt}%
\pgfsys@defobject{currentmarker}{\pgfqpoint{0.000000in}{-0.048611in}}{\pgfqpoint{0.000000in}{0.000000in}}{%
\pgfpathmoveto{\pgfqpoint{0.000000in}{0.000000in}}%
\pgfpathlineto{\pgfqpoint{0.000000in}{-0.048611in}}%
\pgfusepath{stroke,fill}%
}%
\begin{pgfscope}%
\pgfsys@transformshift{0.674954in}{0.862305in}%
\pgfsys@useobject{currentmarker}{}%
\end{pgfscope}%
\end{pgfscope}%
\begin{pgfscope}%
\definecolor{textcolor}{rgb}{0.000000,0.000000,0.000000}%
\pgfsetstrokecolor{textcolor}%
\pgfsetfillcolor{textcolor}%
\pgftext[x=0.674954in,y=0.765082in,,top]{\color{textcolor}{\rmfamily\fontsize{25.000000}{30.000000}\selectfont\catcode`\^=\active\def^{\ifmmode\sp\else\^{}\fi}\catcode`\%=\active\def
\end{pgfscope}%
\begin{pgfscope}%
\pgfpathrectangle{\pgfqpoint{0.674954in}{0.862305in}}{\pgfqpoint{6.472460in}{4.369081in}}%
\pgfusepath{clip}%
\pgfsetbuttcap%
\pgfsetroundjoin%
\pgfsetlinewidth{2.007500pt}%
\definecolor{currentstroke}{rgb}{0.501961,0.501961,0.501961}%
\pgfsetstrokecolor{currentstroke}%
\pgfsetstrokeopacity{0.300000}%
\pgfsetdash{{7.400000pt}{3.200000pt}}{0.000000pt}%
\pgfpathmoveto{\pgfqpoint{1.969446in}{0.862305in}}%
\pgfpathlineto{\pgfqpoint{1.969446in}{5.231386in}}%
\pgfusepath{stroke}%
\end{pgfscope}%
\begin{pgfscope}%
\pgfsetbuttcap%
\pgfsetroundjoin%
\definecolor{currentfill}{rgb}{0.000000,0.000000,0.000000}%
\pgfsetfillcolor{currentfill}%
\pgfsetlinewidth{0.803000pt}%
\definecolor{currentstroke}{rgb}{0.000000,0.000000,0.000000}%
\pgfsetstrokecolor{currentstroke}%
\pgfsetdash{}{0pt}%
\pgfsys@defobject{currentmarker}{\pgfqpoint{0.000000in}{-0.048611in}}{\pgfqpoint{0.000000in}{0.000000in}}{%
\pgfpathmoveto{\pgfqpoint{0.000000in}{0.000000in}}%
\pgfpathlineto{\pgfqpoint{0.000000in}{-0.048611in}}%
\pgfusepath{stroke,fill}%
}%
\begin{pgfscope}%
\pgfsys@transformshift{1.969446in}{0.862305in}%
\pgfsys@useobject{currentmarker}{}%
\end{pgfscope}%
\end{pgfscope}%
\begin{pgfscope}%
\definecolor{textcolor}{rgb}{0.000000,0.000000,0.000000}%
\pgfsetstrokecolor{textcolor}%
\pgfsetfillcolor{textcolor}%
\pgftext[x=1.969446in,y=0.765082in,,top]{\color{textcolor}{\rmfamily\fontsize{25.000000}{30.000000}\selectfont\catcode`\^=\active\def^{\ifmmode\sp\else\^{}\fi}\catcode`\%=\active\def
\end{pgfscope}%
\begin{pgfscope}%
\pgfpathrectangle{\pgfqpoint{0.674954in}{0.862305in}}{\pgfqpoint{6.472460in}{4.369081in}}%
\pgfusepath{clip}%
\pgfsetbuttcap%
\pgfsetroundjoin%
\pgfsetlinewidth{2.007500pt}%
\definecolor{currentstroke}{rgb}{0.501961,0.501961,0.501961}%
\pgfsetstrokecolor{currentstroke}%
\pgfsetstrokeopacity{0.300000}%
\pgfsetdash{{7.400000pt}{3.200000pt}}{0.000000pt}%
\pgfpathmoveto{\pgfqpoint{3.263938in}{0.862305in}}%
\pgfpathlineto{\pgfqpoint{3.263938in}{5.231386in}}%
\pgfusepath{stroke}%
\end{pgfscope}%
\begin{pgfscope}%
\pgfsetbuttcap%
\pgfsetroundjoin%
\definecolor{currentfill}{rgb}{0.000000,0.000000,0.000000}%
\pgfsetfillcolor{currentfill}%
\pgfsetlinewidth{0.803000pt}%
\definecolor{currentstroke}{rgb}{0.000000,0.000000,0.000000}%
\pgfsetstrokecolor{currentstroke}%
\pgfsetdash{}{0pt}%
\pgfsys@defobject{currentmarker}{\pgfqpoint{0.000000in}{-0.048611in}}{\pgfqpoint{0.000000in}{0.000000in}}{%
\pgfpathmoveto{\pgfqpoint{0.000000in}{0.000000in}}%
\pgfpathlineto{\pgfqpoint{0.000000in}{-0.048611in}}%
\pgfusepath{stroke,fill}%
}%
\begin{pgfscope}%
\pgfsys@transformshift{3.263938in}{0.862305in}%
\pgfsys@useobject{currentmarker}{}%
\end{pgfscope}%
\end{pgfscope}%
\begin{pgfscope}%
\definecolor{textcolor}{rgb}{0.000000,0.000000,0.000000}%
\pgfsetstrokecolor{textcolor}%
\pgfsetfillcolor{textcolor}%
\pgftext[x=3.263938in,y=0.765082in,,top]{\color{textcolor}{\rmfamily\fontsize{25.000000}{30.000000}\selectfont\catcode`\^=\active\def^{\ifmmode\sp\else\^{}\fi}\catcode`\%=\active\def
\end{pgfscope}%
\begin{pgfscope}%
\pgfpathrectangle{\pgfqpoint{0.674954in}{0.862305in}}{\pgfqpoint{6.472460in}{4.369081in}}%
\pgfusepath{clip}%
\pgfsetbuttcap%
\pgfsetroundjoin%
\pgfsetlinewidth{2.007500pt}%
\definecolor{currentstroke}{rgb}{0.501961,0.501961,0.501961}%
\pgfsetstrokecolor{currentstroke}%
\pgfsetstrokeopacity{0.300000}%
\pgfsetdash{{7.400000pt}{3.200000pt}}{0.000000pt}%
\pgfpathmoveto{\pgfqpoint{4.558430in}{0.862305in}}%
\pgfpathlineto{\pgfqpoint{4.558430in}{5.231386in}}%
\pgfusepath{stroke}%
\end{pgfscope}%
\begin{pgfscope}%
\pgfsetbuttcap%
\pgfsetroundjoin%
\definecolor{currentfill}{rgb}{0.000000,0.000000,0.000000}%
\pgfsetfillcolor{currentfill}%
\pgfsetlinewidth{0.803000pt}%
\definecolor{currentstroke}{rgb}{0.000000,0.000000,0.000000}%
\pgfsetstrokecolor{currentstroke}%
\pgfsetdash{}{0pt}%
\pgfsys@defobject{currentmarker}{\pgfqpoint{0.000000in}{-0.048611in}}{\pgfqpoint{0.000000in}{0.000000in}}{%
\pgfpathmoveto{\pgfqpoint{0.000000in}{0.000000in}}%
\pgfpathlineto{\pgfqpoint{0.000000in}{-0.048611in}}%
\pgfusepath{stroke,fill}%
}%
\begin{pgfscope}%
\pgfsys@transformshift{4.558430in}{0.862305in}%
\pgfsys@useobject{currentmarker}{}%
\end{pgfscope}%
\end{pgfscope}%
\begin{pgfscope}%
\definecolor{textcolor}{rgb}{0.000000,0.000000,0.000000}%
\pgfsetstrokecolor{textcolor}%
\pgfsetfillcolor{textcolor}%
\pgftext[x=4.558430in,y=0.765082in,,top]{\color{textcolor}{\rmfamily\fontsize{25.000000}{30.000000}\selectfont\catcode`\^=\active\def^{\ifmmode\sp\else\^{}\fi}\catcode`\%=\active\def
\end{pgfscope}%
\begin{pgfscope}%
\pgfpathrectangle{\pgfqpoint{0.674954in}{0.862305in}}{\pgfqpoint{6.472460in}{4.369081in}}%
\pgfusepath{clip}%
\pgfsetbuttcap%
\pgfsetroundjoin%
\pgfsetlinewidth{2.007500pt}%
\definecolor{currentstroke}{rgb}{0.501961,0.501961,0.501961}%
\pgfsetstrokecolor{currentstroke}%
\pgfsetstrokeopacity{0.300000}%
\pgfsetdash{{7.400000pt}{3.200000pt}}{0.000000pt}%
\pgfpathmoveto{\pgfqpoint{5.852922in}{0.862305in}}%
\pgfpathlineto{\pgfqpoint{5.852922in}{5.231386in}}%
\pgfusepath{stroke}%
\end{pgfscope}%
\begin{pgfscope}%
\pgfsetbuttcap%
\pgfsetroundjoin%
\definecolor{currentfill}{rgb}{0.000000,0.000000,0.000000}%
\pgfsetfillcolor{currentfill}%
\pgfsetlinewidth{0.803000pt}%
\definecolor{currentstroke}{rgb}{0.000000,0.000000,0.000000}%
\pgfsetstrokecolor{currentstroke}%
\pgfsetdash{}{0pt}%
\pgfsys@defobject{currentmarker}{\pgfqpoint{0.000000in}{-0.048611in}}{\pgfqpoint{0.000000in}{0.000000in}}{%
\pgfpathmoveto{\pgfqpoint{0.000000in}{0.000000in}}%
\pgfpathlineto{\pgfqpoint{0.000000in}{-0.048611in}}%
\pgfusepath{stroke,fill}%
}%
\begin{pgfscope}%
\pgfsys@transformshift{5.852922in}{0.862305in}%
\pgfsys@useobject{currentmarker}{}%
\end{pgfscope}%
\end{pgfscope}%
\begin{pgfscope}%
\definecolor{textcolor}{rgb}{0.000000,0.000000,0.000000}%
\pgfsetstrokecolor{textcolor}%
\pgfsetfillcolor{textcolor}%
\pgftext[x=5.852922in,y=0.765082in,,top]{\color{textcolor}{\rmfamily\fontsize{25.000000}{30.000000}\selectfont\catcode`\^=\active\def^{\ifmmode\sp\else\^{}\fi}\catcode`\%=\active\def
\end{pgfscope}%
\begin{pgfscope}%
\pgfpathrectangle{\pgfqpoint{0.674954in}{0.862305in}}{\pgfqpoint{6.472460in}{4.369081in}}%
\pgfusepath{clip}%
\pgfsetbuttcap%
\pgfsetroundjoin%
\pgfsetlinewidth{2.007500pt}%
\definecolor{currentstroke}{rgb}{0.501961,0.501961,0.501961}%
\pgfsetstrokecolor{currentstroke}%
\pgfsetstrokeopacity{0.300000}%
\pgfsetdash{{7.400000pt}{3.200000pt}}{0.000000pt}%
\pgfpathmoveto{\pgfqpoint{7.147414in}{0.862305in}}%
\pgfpathlineto{\pgfqpoint{7.147414in}{5.231386in}}%
\pgfusepath{stroke}%
\end{pgfscope}%
\begin{pgfscope}%
\pgfsetbuttcap%
\pgfsetroundjoin%
\definecolor{currentfill}{rgb}{0.000000,0.000000,0.000000}%
\pgfsetfillcolor{currentfill}%
\pgfsetlinewidth{0.803000pt}%
\definecolor{currentstroke}{rgb}{0.000000,0.000000,0.000000}%
\pgfsetstrokecolor{currentstroke}%
\pgfsetdash{}{0pt}%
\pgfsys@defobject{currentmarker}{\pgfqpoint{0.000000in}{-0.048611in}}{\pgfqpoint{0.000000in}{0.000000in}}{%
\pgfpathmoveto{\pgfqpoint{0.000000in}{0.000000in}}%
\pgfpathlineto{\pgfqpoint{0.000000in}{-0.048611in}}%
\pgfusepath{stroke,fill}%
}%
\begin{pgfscope}%
\pgfsys@transformshift{7.147414in}{0.862305in}%
\pgfsys@useobject{currentmarker}{}%
\end{pgfscope}%
\end{pgfscope}%
\begin{pgfscope}%
\definecolor{textcolor}{rgb}{0.000000,0.000000,0.000000}%
\pgfsetstrokecolor{textcolor}%
\pgfsetfillcolor{textcolor}%
\pgftext[x=7.147414in,y=0.765082in,,top]{\color{textcolor}{\rmfamily\fontsize{25.000000}{30.000000}\selectfont\catcode`\^=\active\def^{\ifmmode\sp\else\^{}\fi}\catcode`\%=\active\def
\end{pgfscope}%
\begin{pgfscope}%
\definecolor{textcolor}{rgb}{0.000000,0.000000,0.000000}%
\pgfsetstrokecolor{textcolor}%
\pgfsetfillcolor{textcolor}%
\pgftext[x=3.911184in,y=0.404763in,,top]{\color{textcolor}{\rmfamily\fontsize{25.000000}{30.000000}\selectfont\catcode`\^=\active\def^{\ifmmode\sp\else\^{}\fi}\catcode`\%=\active\def
\end{pgfscope}%
\begin{pgfscope}%
\pgfpathrectangle{\pgfqpoint{0.674954in}{0.862305in}}{\pgfqpoint{6.472460in}{4.369081in}}%
\pgfusepath{clip}%
\pgfsetbuttcap%
\pgfsetroundjoin%
\pgfsetlinewidth{2.007500pt}%
\definecolor{currentstroke}{rgb}{0.501961,0.501961,0.501961}%
\pgfsetstrokecolor{currentstroke}%
\pgfsetstrokeopacity{0.300000}%
\pgfsetdash{{7.400000pt}{3.200000pt}}{0.000000pt}%
\pgfpathmoveto{\pgfqpoint{0.674954in}{1.954575in}}%
\pgfpathlineto{\pgfqpoint{7.147414in}{1.954575in}}%
\pgfusepath{stroke}%
\end{pgfscope}%
\begin{pgfscope}%
\pgfsetbuttcap%
\pgfsetroundjoin%
\definecolor{currentfill}{rgb}{0.000000,0.000000,0.000000}%
\pgfsetfillcolor{currentfill}%
\pgfsetlinewidth{0.803000pt}%
\definecolor{currentstroke}{rgb}{0.000000,0.000000,0.000000}%
\pgfsetstrokecolor{currentstroke}%
\pgfsetdash{}{0pt}%
\pgfsys@defobject{currentmarker}{\pgfqpoint{-0.048611in}{0.000000in}}{\pgfqpoint{-0.000000in}{0.000000in}}{%
\pgfpathmoveto{\pgfqpoint{-0.000000in}{0.000000in}}%
\pgfpathlineto{\pgfqpoint{-0.048611in}{0.000000in}}%
\pgfusepath{stroke,fill}%
}%
\begin{pgfscope}%
\pgfsys@transformshift{0.674954in}{1.954575in}%
\pgfsys@useobject{currentmarker}{}%
\end{pgfscope}%
\end{pgfscope}%
\begin{pgfscope}%
\definecolor{textcolor}{rgb}{0.000000,0.000000,0.000000}%
\pgfsetstrokecolor{textcolor}%
\pgfsetfillcolor{textcolor}%
\pgftext[x=0.259244in, y=1.835961in, left, base]{\color{textcolor}{\rmfamily\fontsize{25.000000}{30.000000}\selectfont\catcode`\^=\active\def^{\ifmmode\sp\else\^{}\fi}\catcode`\%=\active\def
\end{pgfscope}%
\begin{pgfscope}%
\pgfpathrectangle{\pgfqpoint{0.674954in}{0.862305in}}{\pgfqpoint{6.472460in}{4.369081in}}%
\pgfusepath{clip}%
\pgfsetbuttcap%
\pgfsetroundjoin%
\pgfsetlinewidth{2.007500pt}%
\definecolor{currentstroke}{rgb}{0.501961,0.501961,0.501961}%
\pgfsetstrokecolor{currentstroke}%
\pgfsetstrokeopacity{0.300000}%
\pgfsetdash{{7.400000pt}{3.200000pt}}{0.000000pt}%
\pgfpathmoveto{\pgfqpoint{0.674954in}{3.046845in}}%
\pgfpathlineto{\pgfqpoint{7.147414in}{3.046845in}}%
\pgfusepath{stroke}%
\end{pgfscope}%
\begin{pgfscope}%
\pgfsetbuttcap%
\pgfsetroundjoin%
\definecolor{currentfill}{rgb}{0.000000,0.000000,0.000000}%
\pgfsetfillcolor{currentfill}%
\pgfsetlinewidth{0.803000pt}%
\definecolor{currentstroke}{rgb}{0.000000,0.000000,0.000000}%
\pgfsetstrokecolor{currentstroke}%
\pgfsetdash{}{0pt}%
\pgfsys@defobject{currentmarker}{\pgfqpoint{-0.048611in}{0.000000in}}{\pgfqpoint{-0.000000in}{0.000000in}}{%
\pgfpathmoveto{\pgfqpoint{-0.000000in}{0.000000in}}%
\pgfpathlineto{\pgfqpoint{-0.048611in}{0.000000in}}%
\pgfusepath{stroke,fill}%
}%
\begin{pgfscope}%
\pgfsys@transformshift{0.674954in}{3.046845in}%
\pgfsys@useobject{currentmarker}{}%
\end{pgfscope}%
\end{pgfscope}%
\begin{pgfscope}%
\definecolor{textcolor}{rgb}{0.000000,0.000000,0.000000}%
\pgfsetstrokecolor{textcolor}%
\pgfsetfillcolor{textcolor}%
\pgftext[x=0.259244in, y=2.928231in, left, base]{\color{textcolor}{\rmfamily\fontsize{25.000000}{30.000000}\selectfont\catcode`\^=\active\def^{\ifmmode\sp\else\^{}\fi}\catcode`\%=\active\def
\end{pgfscope}%
\begin{pgfscope}%
\pgfpathrectangle{\pgfqpoint{0.674954in}{0.862305in}}{\pgfqpoint{6.472460in}{4.369081in}}%
\pgfusepath{clip}%
\pgfsetbuttcap%
\pgfsetroundjoin%
\pgfsetlinewidth{2.007500pt}%
\definecolor{currentstroke}{rgb}{0.501961,0.501961,0.501961}%
\pgfsetstrokecolor{currentstroke}%
\pgfsetstrokeopacity{0.300000}%
\pgfsetdash{{7.400000pt}{3.200000pt}}{0.000000pt}%
\pgfpathmoveto{\pgfqpoint{0.674954in}{4.139116in}}%
\pgfpathlineto{\pgfqpoint{7.147414in}{4.139116in}}%
\pgfusepath{stroke}%
\end{pgfscope}%
\begin{pgfscope}%
\pgfsetbuttcap%
\pgfsetroundjoin%
\definecolor{currentfill}{rgb}{0.000000,0.000000,0.000000}%
\pgfsetfillcolor{currentfill}%
\pgfsetlinewidth{0.803000pt}%
\definecolor{currentstroke}{rgb}{0.000000,0.000000,0.000000}%
\pgfsetstrokecolor{currentstroke}%
\pgfsetdash{}{0pt}%
\pgfsys@defobject{currentmarker}{\pgfqpoint{-0.048611in}{0.000000in}}{\pgfqpoint{-0.000000in}{0.000000in}}{%
\pgfpathmoveto{\pgfqpoint{-0.000000in}{0.000000in}}%
\pgfpathlineto{\pgfqpoint{-0.048611in}{0.000000in}}%
\pgfusepath{stroke,fill}%
}%
\begin{pgfscope}%
\pgfsys@transformshift{0.674954in}{4.139116in}%
\pgfsys@useobject{currentmarker}{}%
\end{pgfscope}%
\end{pgfscope}%
\begin{pgfscope}%
\definecolor{textcolor}{rgb}{0.000000,0.000000,0.000000}%
\pgfsetstrokecolor{textcolor}%
\pgfsetfillcolor{textcolor}%
\pgftext[x=0.259244in, y=4.020501in, left, base]{\color{textcolor}{\rmfamily\fontsize{25.000000}{30.000000}\selectfont\catcode`\^=\active\def^{\ifmmode\sp\else\^{}\fi}\catcode`\%=\active\def
\end{pgfscope}%
\begin{pgfscope}%
\pgfpathrectangle{\pgfqpoint{0.674954in}{0.862305in}}{\pgfqpoint{6.472460in}{4.369081in}}%
\pgfusepath{clip}%
\pgfsetbuttcap%
\pgfsetroundjoin%
\pgfsetlinewidth{2.007500pt}%
\definecolor{currentstroke}{rgb}{0.501961,0.501961,0.501961}%
\pgfsetstrokecolor{currentstroke}%
\pgfsetstrokeopacity{0.300000}%
\pgfsetdash{{7.400000pt}{3.200000pt}}{0.000000pt}%
\pgfpathmoveto{\pgfqpoint{0.674954in}{5.231386in}}%
\pgfpathlineto{\pgfqpoint{7.147414in}{5.231386in}}%
\pgfusepath{stroke}%
\end{pgfscope}%
\begin{pgfscope}%
\pgfsetbuttcap%
\pgfsetroundjoin%
\definecolor{currentfill}{rgb}{0.000000,0.000000,0.000000}%
\pgfsetfillcolor{currentfill}%
\pgfsetlinewidth{0.803000pt}%
\definecolor{currentstroke}{rgb}{0.000000,0.000000,0.000000}%
\pgfsetstrokecolor{currentstroke}%
\pgfsetdash{}{0pt}%
\pgfsys@defobject{currentmarker}{\pgfqpoint{-0.048611in}{0.000000in}}{\pgfqpoint{-0.000000in}{0.000000in}}{%
\pgfpathmoveto{\pgfqpoint{-0.000000in}{0.000000in}}%
\pgfpathlineto{\pgfqpoint{-0.048611in}{0.000000in}}%
\pgfusepath{stroke,fill}%
}%
\begin{pgfscope}%
\pgfsys@transformshift{0.674954in}{5.231386in}%
\pgfsys@useobject{currentmarker}{}%
\end{pgfscope}%
\end{pgfscope}%
\begin{pgfscope}%
\definecolor{textcolor}{rgb}{0.000000,0.000000,0.000000}%
\pgfsetstrokecolor{textcolor}%
\pgfsetfillcolor{textcolor}%
\pgftext[x=0.100000in, y=5.112772in, left, base]{\color{textcolor}{\rmfamily\fontsize{25.000000}{30.000000}\selectfont\catcode`\^=\active\def^{\ifmmode\sp\else\^{}\fi}\catcode`\%=\active\def
\end{pgfscope}%
\begin{pgfscope}%
\pgfpathrectangle{\pgfqpoint{0.674954in}{0.862305in}}{\pgfqpoint{6.472460in}{4.369081in}}%
\pgfusepath{clip}%
\pgfsetrectcap%
\pgfsetroundjoin%
\pgfsetlinewidth{2.509375pt}%
\definecolor{currentstroke}{rgb}{0.050980,0.415686,0.509804}%
\pgfsetstrokecolor{currentstroke}%
\pgfsetdash{}{0pt}%
\pgfpathmoveto{\pgfqpoint{0.674954in}{4.351010in}}%
\pgfpathlineto{\pgfqpoint{0.998577in}{4.329639in}}%
\pgfpathlineto{\pgfqpoint{2.293069in}{4.168656in}}%
\pgfpathlineto{\pgfqpoint{3.911184in}{4.003376in}}%
\pgfpathlineto{\pgfqpoint{5.529299in}{3.739844in}}%
\pgfpathlineto{\pgfqpoint{6.823791in}{3.155189in}}%
\pgfusepath{stroke}%
\end{pgfscope}%
\begin{pgfscope}%
\pgfpathrectangle{\pgfqpoint{0.674954in}{0.862305in}}{\pgfqpoint{6.472460in}{4.369081in}}%
\pgfusepath{clip}%
\pgfsetbuttcap%
\pgfsetroundjoin%
\definecolor{currentfill}{rgb}{0.050980,0.415686,0.509804}%
\pgfsetfillcolor{currentfill}%
\pgfsetlinewidth{1.003750pt}%
\definecolor{currentstroke}{rgb}{0.050980,0.415686,0.509804}%
\pgfsetstrokecolor{currentstroke}%
\pgfsetdash{}{0pt}%
\pgfsys@defobject{currentmarker}{\pgfqpoint{-0.055556in}{-0.055556in}}{\pgfqpoint{0.055556in}{0.055556in}}{%
\pgfpathmoveto{\pgfqpoint{0.000000in}{-0.055556in}}%
\pgfpathcurveto{\pgfqpoint{0.014734in}{-0.055556in}}{\pgfqpoint{0.028866in}{-0.049702in}}{\pgfqpoint{0.039284in}{-0.039284in}}%
\pgfpathcurveto{\pgfqpoint{0.049702in}{-0.028866in}}{\pgfqpoint{0.055556in}{-0.014734in}}{\pgfqpoint{0.055556in}{0.000000in}}%
\pgfpathcurveto{\pgfqpoint{0.055556in}{0.014734in}}{\pgfqpoint{0.049702in}{0.028866in}}{\pgfqpoint{0.039284in}{0.039284in}}%
\pgfpathcurveto{\pgfqpoint{0.028866in}{0.049702in}}{\pgfqpoint{0.014734in}{0.055556in}}{\pgfqpoint{0.000000in}{0.055556in}}%
\pgfpathcurveto{\pgfqpoint{-0.014734in}{0.055556in}}{\pgfqpoint{-0.028866in}{0.049702in}}{\pgfqpoint{-0.039284in}{0.039284in}}%
\pgfpathcurveto{\pgfqpoint{-0.049702in}{0.028866in}}{\pgfqpoint{-0.055556in}{0.014734in}}{\pgfqpoint{-0.055556in}{0.000000in}}%
\pgfpathcurveto{\pgfqpoint{-0.055556in}{-0.014734in}}{\pgfqpoint{-0.049702in}{-0.028866in}}{\pgfqpoint{-0.039284in}{-0.039284in}}%
\pgfpathcurveto{\pgfqpoint{-0.028866in}{-0.049702in}}{\pgfqpoint{-0.014734in}{-0.055556in}}{\pgfqpoint{0.000000in}{-0.055556in}}%
\pgfpathlineto{\pgfqpoint{0.000000in}{-0.055556in}}%
\pgfpathclose%
\pgfusepath{stroke,fill}%
}%
\begin{pgfscope}%
\pgfsys@transformshift{0.674954in}{4.351010in}%
\pgfsys@useobject{currentmarker}{}%
\end{pgfscope}%
\begin{pgfscope}%
\pgfsys@transformshift{0.998577in}{4.329639in}%
\pgfsys@useobject{currentmarker}{}%
\end{pgfscope}%
\begin{pgfscope}%
\pgfsys@transformshift{2.293069in}{4.168656in}%
\pgfsys@useobject{currentmarker}{}%
\end{pgfscope}%
\begin{pgfscope}%
\pgfsys@transformshift{3.911184in}{4.003376in}%
\pgfsys@useobject{currentmarker}{}%
\end{pgfscope}%
\begin{pgfscope}%
\pgfsys@transformshift{5.529299in}{3.739844in}%
\pgfsys@useobject{currentmarker}{}%
\end{pgfscope}%
\begin{pgfscope}%
\pgfsys@transformshift{6.823791in}{3.155189in}%
\pgfsys@useobject{currentmarker}{}%
\end{pgfscope}%
\end{pgfscope}%
\begin{pgfscope}%
\pgfpathrectangle{\pgfqpoint{0.674954in}{0.862305in}}{\pgfqpoint{6.472460in}{4.369081in}}%
\pgfusepath{clip}%
\pgfsetbuttcap%
\pgfsetroundjoin%
\pgfsetlinewidth{2.509375pt}%
\definecolor{currentstroke}{rgb}{0.960784,0.462745,0.000000}%
\pgfsetstrokecolor{currentstroke}%
\pgfsetdash{{9.250000pt}{4.000000pt}}{0.000000pt}%
\pgfpathmoveto{\pgfqpoint{0.998577in}{3.153420in}}%
\pgfpathlineto{\pgfqpoint{2.293069in}{3.335972in}}%
\pgfpathlineto{\pgfqpoint{3.911184in}{3.344460in}}%
\pgfpathlineto{\pgfqpoint{5.529299in}{3.303697in}}%
\pgfpathlineto{\pgfqpoint{6.823791in}{3.385143in}}%
\pgfpathlineto{\pgfqpoint{7.147414in}{3.453731in}}%
\pgfusepath{stroke}%
\end{pgfscope}%
\begin{pgfscope}%
\pgfpathrectangle{\pgfqpoint{0.674954in}{0.862305in}}{\pgfqpoint{6.472460in}{4.369081in}}%
\pgfusepath{clip}%
\pgfsetbuttcap%
\pgfsetmiterjoin%
\definecolor{currentfill}{rgb}{0.960784,0.462745,0.000000}%
\pgfsetfillcolor{currentfill}%
\pgfsetlinewidth{1.003750pt}%
\definecolor{currentstroke}{rgb}{0.960784,0.462745,0.000000}%
\pgfsetstrokecolor{currentstroke}%
\pgfsetdash{}{0pt}%
\pgfsys@defobject{currentmarker}{\pgfqpoint{-0.055556in}{-0.055556in}}{\pgfqpoint{0.055556in}{0.055556in}}{%
\pgfpathmoveto{\pgfqpoint{-0.055556in}{-0.055556in}}%
\pgfpathlineto{\pgfqpoint{0.055556in}{-0.055556in}}%
\pgfpathlineto{\pgfqpoint{0.055556in}{0.055556in}}%
\pgfpathlineto{\pgfqpoint{-0.055556in}{0.055556in}}%
\pgfpathlineto{\pgfqpoint{-0.055556in}{-0.055556in}}%
\pgfpathclose%
\pgfusepath{stroke,fill}%
}%
\begin{pgfscope}%
\pgfsys@transformshift{0.998577in}{3.153420in}%
\pgfsys@useobject{currentmarker}{}%
\end{pgfscope}%
\begin{pgfscope}%
\pgfsys@transformshift{2.293069in}{3.335972in}%
\pgfsys@useobject{currentmarker}{}%
\end{pgfscope}%
\begin{pgfscope}%
\pgfsys@transformshift{3.911184in}{3.344460in}%
\pgfsys@useobject{currentmarker}{}%
\end{pgfscope}%
\begin{pgfscope}%
\pgfsys@transformshift{5.529299in}{3.303697in}%
\pgfsys@useobject{currentmarker}{}%
\end{pgfscope}%
\begin{pgfscope}%
\pgfsys@transformshift{6.823791in}{3.385143in}%
\pgfsys@useobject{currentmarker}{}%
\end{pgfscope}%
\begin{pgfscope}%
\pgfsys@transformshift{7.147414in}{3.453731in}%
\pgfsys@useobject{currentmarker}{}%
\end{pgfscope}%
\end{pgfscope}%
\begin{pgfscope}%
\pgfsetrectcap%
\pgfsetmiterjoin%
\pgfsetlinewidth{2.007500pt}%
\definecolor{currentstroke}{rgb}{0.000000,0.000000,0.000000}%
\pgfsetstrokecolor{currentstroke}%
\pgfsetdash{}{0pt}%
\pgfpathmoveto{\pgfqpoint{0.674954in}{0.862305in}}%
\pgfpathlineto{\pgfqpoint{0.674954in}{5.231386in}}%
\pgfusepath{stroke}%
\end{pgfscope}%
\begin{pgfscope}%
\pgfsetrectcap%
\pgfsetmiterjoin%
\pgfsetlinewidth{2.007500pt}%
\definecolor{currentstroke}{rgb}{0.000000,0.000000,0.000000}%
\pgfsetstrokecolor{currentstroke}%
\pgfsetdash{}{0pt}%
\pgfpathmoveto{\pgfqpoint{0.674954in}{0.862305in}}%
\pgfpathlineto{\pgfqpoint{7.147414in}{0.862305in}}%
\pgfusepath{stroke}%
\end{pgfscope}%
\begin{pgfscope}%
\pgfsetbuttcap%
\pgfsetmiterjoin%
\definecolor{currentfill}{rgb}{1.000000,1.000000,1.000000}%
\pgfsetfillcolor{currentfill}%
\pgfsetlinewidth{1.003750pt}%
\definecolor{currentstroke}{rgb}{0.800000,0.800000,0.800000}%
\pgfsetstrokecolor{currentstroke}%
\pgfsetdash{}{0pt}%
\pgfpathmoveto{\pgfqpoint{0.869398in}{1.001194in}}%
\pgfpathlineto{\pgfqpoint{2.708109in}{1.001194in}}%
\pgfpathquadraticcurveto{\pgfqpoint{2.763664in}{1.001194in}}{\pgfqpoint{2.763664in}{1.056749in}}%
\pgfpathlineto{\pgfqpoint{2.763664in}{1.803694in}}%
\pgfpathquadraticcurveto{\pgfqpoint{2.763664in}{1.859250in}}{\pgfqpoint{2.708109in}{1.859250in}}%
\pgfpathlineto{\pgfqpoint{0.869398in}{1.859250in}}%
\pgfpathquadraticcurveto{\pgfqpoint{0.813843in}{1.859250in}}{\pgfqpoint{0.813843in}{1.803694in}}%
\pgfpathlineto{\pgfqpoint{0.813843in}{1.056749in}}%
\pgfpathquadraticcurveto{\pgfqpoint{0.813843in}{1.001194in}}{\pgfqpoint{0.869398in}{1.001194in}}%
\pgfpathlineto{\pgfqpoint{0.869398in}{1.001194in}}%
\pgfpathclose%
\pgfusepath{stroke,fill}%
\end{pgfscope}%
\begin{pgfscope}%
\pgfsetrectcap%
\pgfsetroundjoin%
\pgfsetlinewidth{2.509375pt}%
\definecolor{currentstroke}{rgb}{0.050980,0.415686,0.509804}%
\pgfsetstrokecolor{currentstroke}%
\pgfsetdash{}{0pt}%
\pgfpathmoveto{\pgfqpoint{0.924954in}{1.650917in}}%
\pgfpathlineto{\pgfqpoint{1.202732in}{1.650917in}}%
\pgfpathlineto{\pgfqpoint{1.480509in}{1.650917in}}%
\pgfusepath{stroke}%
\end{pgfscope}%
\begin{pgfscope}%
\pgfsetbuttcap%
\pgfsetroundjoin%
\definecolor{currentfill}{rgb}{0.050980,0.415686,0.509804}%
\pgfsetfillcolor{currentfill}%
\pgfsetlinewidth{1.003750pt}%
\definecolor{currentstroke}{rgb}{0.050980,0.415686,0.509804}%
\pgfsetstrokecolor{currentstroke}%
\pgfsetdash{}{0pt}%
\pgfsys@defobject{currentmarker}{\pgfqpoint{-0.055556in}{-0.055556in}}{\pgfqpoint{0.055556in}{0.055556in}}{%
\pgfpathmoveto{\pgfqpoint{0.000000in}{-0.055556in}}%
\pgfpathcurveto{\pgfqpoint{0.014734in}{-0.055556in}}{\pgfqpoint{0.028866in}{-0.049702in}}{\pgfqpoint{0.039284in}{-0.039284in}}%
\pgfpathcurveto{\pgfqpoint{0.049702in}{-0.028866in}}{\pgfqpoint{0.055556in}{-0.014734in}}{\pgfqpoint{0.055556in}{0.000000in}}%
\pgfpathcurveto{\pgfqpoint{0.055556in}{0.014734in}}{\pgfqpoint{0.049702in}{0.028866in}}{\pgfqpoint{0.039284in}{0.039284in}}%
\pgfpathcurveto{\pgfqpoint{0.028866in}{0.049702in}}{\pgfqpoint{0.014734in}{0.055556in}}{\pgfqpoint{0.000000in}{0.055556in}}%
\pgfpathcurveto{\pgfqpoint{-0.014734in}{0.055556in}}{\pgfqpoint{-0.028866in}{0.049702in}}{\pgfqpoint{-0.039284in}{0.039284in}}%
\pgfpathcurveto{\pgfqpoint{-0.049702in}{0.028866in}}{\pgfqpoint{-0.055556in}{0.014734in}}{\pgfqpoint{-0.055556in}{0.000000in}}%
\pgfpathcurveto{\pgfqpoint{-0.055556in}{-0.014734in}}{\pgfqpoint{-0.049702in}{-0.028866in}}{\pgfqpoint{-0.039284in}{-0.039284in}}%
\pgfpathcurveto{\pgfqpoint{-0.028866in}{-0.049702in}}{\pgfqpoint{-0.014734in}{-0.055556in}}{\pgfqpoint{0.000000in}{-0.055556in}}%
\pgfpathlineto{\pgfqpoint{0.000000in}{-0.055556in}}%
\pgfpathclose%
\pgfusepath{stroke,fill}%
}%
\begin{pgfscope}%
\pgfsys@transformshift{1.202732in}{1.650917in}%
\pgfsys@useobject{currentmarker}{}%
\end{pgfscope}%
\end{pgfscope}%
\begin{pgfscope}%
\definecolor{textcolor}{rgb}{0.000000,0.000000,0.000000}%
\pgfsetstrokecolor{textcolor}%
\pgfsetfillcolor{textcolor}%
\pgftext[x=1.702732in,y=1.553694in,left,base]{\color{textcolor}{\rmfamily\fontsize{20.000000}{24.000000}\selectfont\catcode`\^=\active\def^{\ifmmode\sp\else\^{}\fi}\catcode`\%=\active\def
\end{pgfscope}%
\begin{pgfscope}%
\pgfsetbuttcap%
\pgfsetroundjoin%
\pgfsetlinewidth{2.509375pt}%
\definecolor{currentstroke}{rgb}{0.960784,0.462745,0.000000}%
\pgfsetstrokecolor{currentstroke}%
\pgfsetdash{{9.250000pt}{4.000000pt}}{0.000000pt}%
\pgfpathmoveto{\pgfqpoint{0.924954in}{1.263555in}}%
\pgfpathlineto{\pgfqpoint{1.202732in}{1.263555in}}%
\pgfpathlineto{\pgfqpoint{1.480509in}{1.263555in}}%
\pgfusepath{stroke}%
\end{pgfscope}%
\begin{pgfscope}%
\pgfsetbuttcap%
\pgfsetmiterjoin%
\definecolor{currentfill}{rgb}{0.960784,0.462745,0.000000}%
\pgfsetfillcolor{currentfill}%
\pgfsetlinewidth{1.003750pt}%
\definecolor{currentstroke}{rgb}{0.960784,0.462745,0.000000}%
\pgfsetstrokecolor{currentstroke}%
\pgfsetdash{}{0pt}%
\pgfsys@defobject{currentmarker}{\pgfqpoint{-0.055556in}{-0.055556in}}{\pgfqpoint{0.055556in}{0.055556in}}{%
\pgfpathmoveto{\pgfqpoint{-0.055556in}{-0.055556in}}%
\pgfpathlineto{\pgfqpoint{0.055556in}{-0.055556in}}%
\pgfpathlineto{\pgfqpoint{0.055556in}{0.055556in}}%
\pgfpathlineto{\pgfqpoint{-0.055556in}{0.055556in}}%
\pgfpathlineto{\pgfqpoint{-0.055556in}{-0.055556in}}%
\pgfpathclose%
\pgfusepath{stroke,fill}%
}%
\begin{pgfscope}%
\pgfsys@transformshift{1.202732in}{1.263555in}%
\pgfsys@useobject{currentmarker}{}%
\end{pgfscope}%
\end{pgfscope}%
\begin{pgfscope}%
\definecolor{textcolor}{rgb}{0.000000,0.000000,0.000000}%
\pgfsetstrokecolor{textcolor}%
\pgfsetfillcolor{textcolor}%
\pgftext[x=1.702732in,y=1.166333in,left,base]{\color{textcolor}{\rmfamily\fontsize{20.000000}{24.000000}\selectfont\catcode`\^=\active\def^{\ifmmode\sp\else\^{}\fi}\catcode`\%=\active\def
\end{pgfscope}%
\end{pgfpicture}%
\makeatother%
\endgroup%

%% file: images/thermal_ratio_per_modality/modality_results_mAA_30.pgf
\begingroup%
\makeatletter%
\begin{pgfpicture}%
\pgfpathrectangle{\pgfpointorigin}{\pgfqpoint{7.450000in}{5.450000in}}%
\pgfusepath{use as bounding box, clip}%
\begin{pgfscope}%
\pgfsetbuttcap%
\pgfsetmiterjoin%
\definecolor{currentfill}{rgb}{1.000000,1.000000,1.000000}%
\pgfsetfillcolor{currentfill}%
\pgfsetlinewidth{0.000000pt}%
\definecolor{currentstroke}{rgb}{1.000000,1.000000,1.000000}%
\pgfsetstrokecolor{currentstroke}%
\pgfsetdash{}{0pt}%
\pgfpathmoveto{\pgfqpoint{0.000000in}{0.000000in}}%
\pgfpathlineto{\pgfqpoint{7.450000in}{0.000000in}}%
\pgfpathlineto{\pgfqpoint{7.450000in}{5.450000in}}%
\pgfpathlineto{\pgfqpoint{0.000000in}{5.450000in}}%
\pgfpathlineto{\pgfqpoint{0.000000in}{0.000000in}}%
\pgfpathclose%
\pgfusepath{fill}%
\end{pgfscope}%
\begin{pgfscope}%
\pgfsetbuttcap%
\pgfsetmiterjoin%
\definecolor{currentfill}{rgb}{1.000000,1.000000,1.000000}%
\pgfsetfillcolor{currentfill}%
\pgfsetlinewidth{0.000000pt}%
\definecolor{currentstroke}{rgb}{0.000000,0.000000,0.000000}%
\pgfsetstrokecolor{currentstroke}%
\pgfsetstrokeopacity{0.000000}%
\pgfsetdash{}{0pt}%
\pgfpathmoveto{\pgfqpoint{0.674954in}{0.862305in}}%
\pgfpathlineto{\pgfqpoint{7.147414in}{0.862305in}}%
\pgfpathlineto{\pgfqpoint{7.147414in}{5.231386in}}%
\pgfpathlineto{\pgfqpoint{0.674954in}{5.231386in}}%
\pgfpathlineto{\pgfqpoint{0.674954in}{0.862305in}}%
\pgfpathclose%
\pgfusepath{fill}%
\end{pgfscope}%
\begin{pgfscope}%
\pgfpathrectangle{\pgfqpoint{0.674954in}{0.862305in}}{\pgfqpoint{6.472460in}{4.369081in}}%
\pgfusepath{clip}%
\pgfsetbuttcap%
\pgfsetroundjoin%
\pgfsetlinewidth{2.007500pt}%
\definecolor{currentstroke}{rgb}{0.501961,0.501961,0.501961}%
\pgfsetstrokecolor{currentstroke}%
\pgfsetstrokeopacity{0.300000}%
\pgfsetdash{{7.400000pt}{3.200000pt}}{0.000000pt}%
\pgfpathmoveto{\pgfqpoint{0.674954in}{0.862305in}}%
\pgfpathlineto{\pgfqpoint{0.674954in}{5.231386in}}%
\pgfusepath{stroke}%
\end{pgfscope}%
\begin{pgfscope}%
\pgfsetbuttcap%
\pgfsetroundjoin%
\definecolor{currentfill}{rgb}{0.000000,0.000000,0.000000}%
\pgfsetfillcolor{currentfill}%
\pgfsetlinewidth{0.803000pt}%
\definecolor{currentstroke}{rgb}{0.000000,0.000000,0.000000}%
\pgfsetstrokecolor{currentstroke}%
\pgfsetdash{}{0pt}%
\pgfsys@defobject{currentmarker}{\pgfqpoint{0.000000in}{-0.048611in}}{\pgfqpoint{0.000000in}{0.000000in}}{%
\pgfpathmoveto{\pgfqpoint{0.000000in}{0.000000in}}%
\pgfpathlineto{\pgfqpoint{0.000000in}{-0.048611in}}%
\pgfusepath{stroke,fill}%
}%
\begin{pgfscope}%
\pgfsys@transformshift{0.674954in}{0.862305in}%
\pgfsys@useobject{currentmarker}{}%
\end{pgfscope}%
\end{pgfscope}%
\begin{pgfscope}%
\definecolor{textcolor}{rgb}{0.000000,0.000000,0.000000}%
\pgfsetstrokecolor{textcolor}%
\pgfsetfillcolor{textcolor}%
\pgftext[x=0.674954in,y=0.765082in,,top]{\color{textcolor}{\rmfamily\fontsize{25.000000}{30.000000}\selectfont\catcode`\^=\active\def^{\ifmmode\sp\else\^{}\fi}\catcode`\%=\active\def
\end{pgfscope}%
\begin{pgfscope}%
\pgfpathrectangle{\pgfqpoint{0.674954in}{0.862305in}}{\pgfqpoint{6.472460in}{4.369081in}}%
\pgfusepath{clip}%
\pgfsetbuttcap%
\pgfsetroundjoin%
\pgfsetlinewidth{2.007500pt}%
\definecolor{currentstroke}{rgb}{0.501961,0.501961,0.501961}%
\pgfsetstrokecolor{currentstroke}%
\pgfsetstrokeopacity{0.300000}%
\pgfsetdash{{7.400000pt}{3.200000pt}}{0.000000pt}%
\pgfpathmoveto{\pgfqpoint{1.969446in}{0.862305in}}%
\pgfpathlineto{\pgfqpoint{1.969446in}{5.231386in}}%
\pgfusepath{stroke}%
\end{pgfscope}%
\begin{pgfscope}%
\pgfsetbuttcap%
\pgfsetroundjoin%
\definecolor{currentfill}{rgb}{0.000000,0.000000,0.000000}%
\pgfsetfillcolor{currentfill}%
\pgfsetlinewidth{0.803000pt}%
\definecolor{currentstroke}{rgb}{0.000000,0.000000,0.000000}%
\pgfsetstrokecolor{currentstroke}%
\pgfsetdash{}{0pt}%
\pgfsys@defobject{currentmarker}{\pgfqpoint{0.000000in}{-0.048611in}}{\pgfqpoint{0.000000in}{0.000000in}}{%
\pgfpathmoveto{\pgfqpoint{0.000000in}{0.000000in}}%
\pgfpathlineto{\pgfqpoint{0.000000in}{-0.048611in}}%
\pgfusepath{stroke,fill}%
}%
\begin{pgfscope}%
\pgfsys@transformshift{1.969446in}{0.862305in}%
\pgfsys@useobject{currentmarker}{}%
\end{pgfscope}%
\end{pgfscope}%
\begin{pgfscope}%
\definecolor{textcolor}{rgb}{0.000000,0.000000,0.000000}%
\pgfsetstrokecolor{textcolor}%
\pgfsetfillcolor{textcolor}%
\pgftext[x=1.969446in,y=0.765082in,,top]{\color{textcolor}{\rmfamily\fontsize{25.000000}{30.000000}\selectfont\catcode`\^=\active\def^{\ifmmode\sp\else\^{}\fi}\catcode`\%=\active\def
\end{pgfscope}%
\begin{pgfscope}%
\pgfpathrectangle{\pgfqpoint{0.674954in}{0.862305in}}{\pgfqpoint{6.472460in}{4.369081in}}%
\pgfusepath{clip}%
\pgfsetbuttcap%
\pgfsetroundjoin%
\pgfsetlinewidth{2.007500pt}%
\definecolor{currentstroke}{rgb}{0.501961,0.501961,0.501961}%
\pgfsetstrokecolor{currentstroke}%
\pgfsetstrokeopacity{0.300000}%
\pgfsetdash{{7.400000pt}{3.200000pt}}{0.000000pt}%
\pgfpathmoveto{\pgfqpoint{3.263938in}{0.862305in}}%
\pgfpathlineto{\pgfqpoint{3.263938in}{5.231386in}}%
\pgfusepath{stroke}%
\end{pgfscope}%
\begin{pgfscope}%
\pgfsetbuttcap%
\pgfsetroundjoin%
\definecolor{currentfill}{rgb}{0.000000,0.000000,0.000000}%
\pgfsetfillcolor{currentfill}%
\pgfsetlinewidth{0.803000pt}%
\definecolor{currentstroke}{rgb}{0.000000,0.000000,0.000000}%
\pgfsetstrokecolor{currentstroke}%
\pgfsetdash{}{0pt}%
\pgfsys@defobject{currentmarker}{\pgfqpoint{0.000000in}{-0.048611in}}{\pgfqpoint{0.000000in}{0.000000in}}{%
\pgfpathmoveto{\pgfqpoint{0.000000in}{0.000000in}}%
\pgfpathlineto{\pgfqpoint{0.000000in}{-0.048611in}}%
\pgfusepath{stroke,fill}%
}%
\begin{pgfscope}%
\pgfsys@transformshift{3.263938in}{0.862305in}%
\pgfsys@useobject{currentmarker}{}%
\end{pgfscope}%
\end{pgfscope}%
\begin{pgfscope}%
\definecolor{textcolor}{rgb}{0.000000,0.000000,0.000000}%
\pgfsetstrokecolor{textcolor}%
\pgfsetfillcolor{textcolor}%
\pgftext[x=3.263938in,y=0.765082in,,top]{\color{textcolor}{\rmfamily\fontsize{25.000000}{30.000000}\selectfont\catcode`\^=\active\def^{\ifmmode\sp\else\^{}\fi}\catcode`\%=\active\def
\end{pgfscope}%
\begin{pgfscope}%
\pgfpathrectangle{\pgfqpoint{0.674954in}{0.862305in}}{\pgfqpoint{6.472460in}{4.369081in}}%
\pgfusepath{clip}%
\pgfsetbuttcap%
\pgfsetroundjoin%
\pgfsetlinewidth{2.007500pt}%
\definecolor{currentstroke}{rgb}{0.501961,0.501961,0.501961}%
\pgfsetstrokecolor{currentstroke}%
\pgfsetstrokeopacity{0.300000}%
\pgfsetdash{{7.400000pt}{3.200000pt}}{0.000000pt}%
\pgfpathmoveto{\pgfqpoint{4.558430in}{0.862305in}}%
\pgfpathlineto{\pgfqpoint{4.558430in}{5.231386in}}%
\pgfusepath{stroke}%
\end{pgfscope}%
\begin{pgfscope}%
\pgfsetbuttcap%
\pgfsetroundjoin%
\definecolor{currentfill}{rgb}{0.000000,0.000000,0.000000}%
\pgfsetfillcolor{currentfill}%
\pgfsetlinewidth{0.803000pt}%
\definecolor{currentstroke}{rgb}{0.000000,0.000000,0.000000}%
\pgfsetstrokecolor{currentstroke}%
\pgfsetdash{}{0pt}%
\pgfsys@defobject{currentmarker}{\pgfqpoint{0.000000in}{-0.048611in}}{\pgfqpoint{0.000000in}{0.000000in}}{%
\pgfpathmoveto{\pgfqpoint{0.000000in}{0.000000in}}%
\pgfpathlineto{\pgfqpoint{0.000000in}{-0.048611in}}%
\pgfusepath{stroke,fill}%
}%
\begin{pgfscope}%
\pgfsys@transformshift{4.558430in}{0.862305in}%
\pgfsys@useobject{currentmarker}{}%
\end{pgfscope}%
\end{pgfscope}%
\begin{pgfscope}%
\definecolor{textcolor}{rgb}{0.000000,0.000000,0.000000}%
\pgfsetstrokecolor{textcolor}%
\pgfsetfillcolor{textcolor}%
\pgftext[x=4.558430in,y=0.765082in,,top]{\color{textcolor}{\rmfamily\fontsize{25.000000}{30.000000}\selectfont\catcode`\^=\active\def^{\ifmmode\sp\else\^{}\fi}\catcode`\%=\active\def
\end{pgfscope}%
\begin{pgfscope}%
\pgfpathrectangle{\pgfqpoint{0.674954in}{0.862305in}}{\pgfqpoint{6.472460in}{4.369081in}}%
\pgfusepath{clip}%
\pgfsetbuttcap%
\pgfsetroundjoin%
\pgfsetlinewidth{2.007500pt}%
\definecolor{currentstroke}{rgb}{0.501961,0.501961,0.501961}%
\pgfsetstrokecolor{currentstroke}%
\pgfsetstrokeopacity{0.300000}%
\pgfsetdash{{7.400000pt}{3.200000pt}}{0.000000pt}%
\pgfpathmoveto{\pgfqpoint{5.852922in}{0.862305in}}%
\pgfpathlineto{\pgfqpoint{5.852922in}{5.231386in}}%
\pgfusepath{stroke}%
\end{pgfscope}%
\begin{pgfscope}%
\pgfsetbuttcap%
\pgfsetroundjoin%
\definecolor{currentfill}{rgb}{0.000000,0.000000,0.000000}%
\pgfsetfillcolor{currentfill}%
\pgfsetlinewidth{0.803000pt}%
\definecolor{currentstroke}{rgb}{0.000000,0.000000,0.000000}%
\pgfsetstrokecolor{currentstroke}%
\pgfsetdash{}{0pt}%
\pgfsys@defobject{currentmarker}{\pgfqpoint{0.000000in}{-0.048611in}}{\pgfqpoint{0.000000in}{0.000000in}}{%
\pgfpathmoveto{\pgfqpoint{0.000000in}{0.000000in}}%
\pgfpathlineto{\pgfqpoint{0.000000in}{-0.048611in}}%
\pgfusepath{stroke,fill}%
}%
\begin{pgfscope}%
\pgfsys@transformshift{5.852922in}{0.862305in}%
\pgfsys@useobject{currentmarker}{}%
\end{pgfscope}%
\end{pgfscope}%
\begin{pgfscope}%
\definecolor{textcolor}{rgb}{0.000000,0.000000,0.000000}%
\pgfsetstrokecolor{textcolor}%
\pgfsetfillcolor{textcolor}%
\pgftext[x=5.852922in,y=0.765082in,,top]{\color{textcolor}{\rmfamily\fontsize{25.000000}{30.000000}\selectfont\catcode`\^=\active\def^{\ifmmode\sp\else\^{}\fi}\catcode`\%=\active\def
\end{pgfscope}%
\begin{pgfscope}%
\pgfpathrectangle{\pgfqpoint{0.674954in}{0.862305in}}{\pgfqpoint{6.472460in}{4.369081in}}%
\pgfusepath{clip}%
\pgfsetbuttcap%
\pgfsetroundjoin%
\pgfsetlinewidth{2.007500pt}%
\definecolor{currentstroke}{rgb}{0.501961,0.501961,0.501961}%
\pgfsetstrokecolor{currentstroke}%
\pgfsetstrokeopacity{0.300000}%
\pgfsetdash{{7.400000pt}{3.200000pt}}{0.000000pt}%
\pgfpathmoveto{\pgfqpoint{7.147414in}{0.862305in}}%
\pgfpathlineto{\pgfqpoint{7.147414in}{5.231386in}}%
\pgfusepath{stroke}%
\end{pgfscope}%
\begin{pgfscope}%
\pgfsetbuttcap%
\pgfsetroundjoin%
\definecolor{currentfill}{rgb}{0.000000,0.000000,0.000000}%
\pgfsetfillcolor{currentfill}%
\pgfsetlinewidth{0.803000pt}%
\definecolor{currentstroke}{rgb}{0.000000,0.000000,0.000000}%
\pgfsetstrokecolor{currentstroke}%
\pgfsetdash{}{0pt}%
\pgfsys@defobject{currentmarker}{\pgfqpoint{0.000000in}{-0.048611in}}{\pgfqpoint{0.000000in}{0.000000in}}{%
\pgfpathmoveto{\pgfqpoint{0.000000in}{0.000000in}}%
\pgfpathlineto{\pgfqpoint{0.000000in}{-0.048611in}}%
\pgfusepath{stroke,fill}%
}%
\begin{pgfscope}%
\pgfsys@transformshift{7.147414in}{0.862305in}%
\pgfsys@useobject{currentmarker}{}%
\end{pgfscope}%
\end{pgfscope}%
\begin{pgfscope}%
\definecolor{textcolor}{rgb}{0.000000,0.000000,0.000000}%
\pgfsetstrokecolor{textcolor}%
\pgfsetfillcolor{textcolor}%
\pgftext[x=7.147414in,y=0.765082in,,top]{\color{textcolor}{\rmfamily\fontsize{25.000000}{30.000000}\selectfont\catcode`\^=\active\def^{\ifmmode\sp\else\^{}\fi}\catcode`\%=\active\def
\end{pgfscope}%
\begin{pgfscope}%
\definecolor{textcolor}{rgb}{0.000000,0.000000,0.000000}%
\pgfsetstrokecolor{textcolor}%
\pgfsetfillcolor{textcolor}%
\pgftext[x=3.911184in,y=0.404763in,,top]{\color{textcolor}{\rmfamily\fontsize{25.000000}{30.000000}\selectfont\catcode`\^=\active\def^{\ifmmode\sp\else\^{}\fi}\catcode`\%=\active\def
\end{pgfscope}%
\begin{pgfscope}%
\pgfpathrectangle{\pgfqpoint{0.674954in}{0.862305in}}{\pgfqpoint{6.472460in}{4.369081in}}%
\pgfusepath{clip}%
\pgfsetbuttcap%
\pgfsetroundjoin%
\pgfsetlinewidth{2.007500pt}%
\definecolor{currentstroke}{rgb}{0.501961,0.501961,0.501961}%
\pgfsetstrokecolor{currentstroke}%
\pgfsetstrokeopacity{0.300000}%
\pgfsetdash{{7.400000pt}{3.200000pt}}{0.000000pt}%
\pgfpathmoveto{\pgfqpoint{0.674954in}{1.954575in}}%
\pgfpathlineto{\pgfqpoint{7.147414in}{1.954575in}}%
\pgfusepath{stroke}%
\end{pgfscope}%
\begin{pgfscope}%
\pgfsetbuttcap%
\pgfsetroundjoin%
\definecolor{currentfill}{rgb}{0.000000,0.000000,0.000000}%
\pgfsetfillcolor{currentfill}%
\pgfsetlinewidth{0.803000pt}%
\definecolor{currentstroke}{rgb}{0.000000,0.000000,0.000000}%
\pgfsetstrokecolor{currentstroke}%
\pgfsetdash{}{0pt}%
\pgfsys@defobject{currentmarker}{\pgfqpoint{-0.048611in}{0.000000in}}{\pgfqpoint{-0.000000in}{0.000000in}}{%
\pgfpathmoveto{\pgfqpoint{-0.000000in}{0.000000in}}%
\pgfpathlineto{\pgfqpoint{-0.048611in}{0.000000in}}%
\pgfusepath{stroke,fill}%
}%
\begin{pgfscope}%
\pgfsys@transformshift{0.674954in}{1.954575in}%
\pgfsys@useobject{currentmarker}{}%
\end{pgfscope}%
\end{pgfscope}%
\begin{pgfscope}%
\definecolor{textcolor}{rgb}{0.000000,0.000000,0.000000}%
\pgfsetstrokecolor{textcolor}%
\pgfsetfillcolor{textcolor}%
\pgftext[x=0.259244in, y=1.835961in, left, base]{\color{textcolor}{\rmfamily\fontsize{25.000000}{30.000000}\selectfont\catcode`\^=\active\def^{\ifmmode\sp\else\^{}\fi}\catcode`\%=\active\def
\end{pgfscope}%
\begin{pgfscope}%
\pgfpathrectangle{\pgfqpoint{0.674954in}{0.862305in}}{\pgfqpoint{6.472460in}{4.369081in}}%
\pgfusepath{clip}%
\pgfsetbuttcap%
\pgfsetroundjoin%
\pgfsetlinewidth{2.007500pt}%
\definecolor{currentstroke}{rgb}{0.501961,0.501961,0.501961}%
\pgfsetstrokecolor{currentstroke}%
\pgfsetstrokeopacity{0.300000}%
\pgfsetdash{{7.400000pt}{3.200000pt}}{0.000000pt}%
\pgfpathmoveto{\pgfqpoint{0.674954in}{3.046845in}}%
\pgfpathlineto{\pgfqpoint{7.147414in}{3.046845in}}%
\pgfusepath{stroke}%
\end{pgfscope}%
\begin{pgfscope}%
\pgfsetbuttcap%
\pgfsetroundjoin%
\definecolor{currentfill}{rgb}{0.000000,0.000000,0.000000}%
\pgfsetfillcolor{currentfill}%
\pgfsetlinewidth{0.803000pt}%
\definecolor{currentstroke}{rgb}{0.000000,0.000000,0.000000}%
\pgfsetstrokecolor{currentstroke}%
\pgfsetdash{}{0pt}%
\pgfsys@defobject{currentmarker}{\pgfqpoint{-0.048611in}{0.000000in}}{\pgfqpoint{-0.000000in}{0.000000in}}{%
\pgfpathmoveto{\pgfqpoint{-0.000000in}{0.000000in}}%
\pgfpathlineto{\pgfqpoint{-0.048611in}{0.000000in}}%
\pgfusepath{stroke,fill}%
}%
\begin{pgfscope}%
\pgfsys@transformshift{0.674954in}{3.046845in}%
\pgfsys@useobject{currentmarker}{}%
\end{pgfscope}%
\end{pgfscope}%
\begin{pgfscope}%
\definecolor{textcolor}{rgb}{0.000000,0.000000,0.000000}%
\pgfsetstrokecolor{textcolor}%
\pgfsetfillcolor{textcolor}%
\pgftext[x=0.259244in, y=2.928231in, left, base]{\color{textcolor}{\rmfamily\fontsize{25.000000}{30.000000}\selectfont\catcode`\^=\active\def^{\ifmmode\sp\else\^{}\fi}\catcode`\%=\active\def
\end{pgfscope}%
\begin{pgfscope}%
\pgfpathrectangle{\pgfqpoint{0.674954in}{0.862305in}}{\pgfqpoint{6.472460in}{4.369081in}}%
\pgfusepath{clip}%
\pgfsetbuttcap%
\pgfsetroundjoin%
\pgfsetlinewidth{2.007500pt}%
\definecolor{currentstroke}{rgb}{0.501961,0.501961,0.501961}%
\pgfsetstrokecolor{currentstroke}%
\pgfsetstrokeopacity{0.300000}%
\pgfsetdash{{7.400000pt}{3.200000pt}}{0.000000pt}%
\pgfpathmoveto{\pgfqpoint{0.674954in}{4.139116in}}%
\pgfpathlineto{\pgfqpoint{7.147414in}{4.139116in}}%
\pgfusepath{stroke}%
\end{pgfscope}%
\begin{pgfscope}%
\pgfsetbuttcap%
\pgfsetroundjoin%
\definecolor{currentfill}{rgb}{0.000000,0.000000,0.000000}%
\pgfsetfillcolor{currentfill}%
\pgfsetlinewidth{0.803000pt}%
\definecolor{currentstroke}{rgb}{0.000000,0.000000,0.000000}%
\pgfsetstrokecolor{currentstroke}%
\pgfsetdash{}{0pt}%
\pgfsys@defobject{currentmarker}{\pgfqpoint{-0.048611in}{0.000000in}}{\pgfqpoint{-0.000000in}{0.000000in}}{%
\pgfpathmoveto{\pgfqpoint{-0.000000in}{0.000000in}}%
\pgfpathlineto{\pgfqpoint{-0.048611in}{0.000000in}}%
\pgfusepath{stroke,fill}%
}%
\begin{pgfscope}%
\pgfsys@transformshift{0.674954in}{4.139116in}%
\pgfsys@useobject{currentmarker}{}%
\end{pgfscope}%
\end{pgfscope}%
\begin{pgfscope}%
\definecolor{textcolor}{rgb}{0.000000,0.000000,0.000000}%
\pgfsetstrokecolor{textcolor}%
\pgfsetfillcolor{textcolor}%
\pgftext[x=0.259244in, y=4.020501in, left, base]{\color{textcolor}{\rmfamily\fontsize{25.000000}{30.000000}\selectfont\catcode`\^=\active\def^{\ifmmode\sp\else\^{}\fi}\catcode`\%=\active\def
\end{pgfscope}%
\begin{pgfscope}%
\pgfpathrectangle{\pgfqpoint{0.674954in}{0.862305in}}{\pgfqpoint{6.472460in}{4.369081in}}%
\pgfusepath{clip}%
\pgfsetbuttcap%
\pgfsetroundjoin%
\pgfsetlinewidth{2.007500pt}%
\definecolor{currentstroke}{rgb}{0.501961,0.501961,0.501961}%
\pgfsetstrokecolor{currentstroke}%
\pgfsetstrokeopacity{0.300000}%
\pgfsetdash{{7.400000pt}{3.200000pt}}{0.000000pt}%
\pgfpathmoveto{\pgfqpoint{0.674954in}{5.231386in}}%
\pgfpathlineto{\pgfqpoint{7.147414in}{5.231386in}}%
\pgfusepath{stroke}%
\end{pgfscope}%
\begin{pgfscope}%
\pgfsetbuttcap%
\pgfsetroundjoin%
\definecolor{currentfill}{rgb}{0.000000,0.000000,0.000000}%
\pgfsetfillcolor{currentfill}%
\pgfsetlinewidth{0.803000pt}%
\definecolor{currentstroke}{rgb}{0.000000,0.000000,0.000000}%
\pgfsetstrokecolor{currentstroke}%
\pgfsetdash{}{0pt}%
\pgfsys@defobject{currentmarker}{\pgfqpoint{-0.048611in}{0.000000in}}{\pgfqpoint{-0.000000in}{0.000000in}}{%
\pgfpathmoveto{\pgfqpoint{-0.000000in}{0.000000in}}%
\pgfpathlineto{\pgfqpoint{-0.048611in}{0.000000in}}%
\pgfusepath{stroke,fill}%
}%
\begin{pgfscope}%
\pgfsys@transformshift{0.674954in}{5.231386in}%
\pgfsys@useobject{currentmarker}{}%
\end{pgfscope}%
\end{pgfscope}%
\begin{pgfscope}%
\definecolor{textcolor}{rgb}{0.000000,0.000000,0.000000}%
\pgfsetstrokecolor{textcolor}%
\pgfsetfillcolor{textcolor}%
\pgftext[x=0.100000in, y=5.112772in, left, base]{\color{textcolor}{\rmfamily\fontsize{25.000000}{30.000000}\selectfont\catcode`\^=\active\def^{\ifmmode\sp\else\^{}\fi}\catcode`\%=\active\def
\end{pgfscope}%
\begin{pgfscope}%
\pgfpathrectangle{\pgfqpoint{0.674954in}{0.862305in}}{\pgfqpoint{6.472460in}{4.369081in}}%
\pgfusepath{clip}%
\pgfsetrectcap%
\pgfsetroundjoin%
\pgfsetlinewidth{2.509375pt}%
\definecolor{currentstroke}{rgb}{0.050980,0.415686,0.509804}%
\pgfsetstrokecolor{currentstroke}%
\pgfsetdash{}{0pt}%
\pgfpathmoveto{\pgfqpoint{0.674954in}{4.730650in}}%
\pgfpathlineto{\pgfqpoint{0.998577in}{4.718620in}}%
\pgfpathlineto{\pgfqpoint{2.293069in}{4.610217in}}%
\pgfpathlineto{\pgfqpoint{3.911184in}{4.512449in}}%
\pgfpathlineto{\pgfqpoint{5.529299in}{4.236181in}}%
\pgfpathlineto{\pgfqpoint{6.823791in}{3.617352in}}%
\pgfusepath{stroke}%
\end{pgfscope}%
\begin{pgfscope}%
\pgfpathrectangle{\pgfqpoint{0.674954in}{0.862305in}}{\pgfqpoint{6.472460in}{4.369081in}}%
\pgfusepath{clip}%
\pgfsetbuttcap%
\pgfsetroundjoin%
\definecolor{currentfill}{rgb}{0.050980,0.415686,0.509804}%
\pgfsetfillcolor{currentfill}%
\pgfsetlinewidth{1.003750pt}%
\definecolor{currentstroke}{rgb}{0.050980,0.415686,0.509804}%
\pgfsetstrokecolor{currentstroke}%
\pgfsetdash{}{0pt}%
\pgfsys@defobject{currentmarker}{\pgfqpoint{-0.055556in}{-0.055556in}}{\pgfqpoint{0.055556in}{0.055556in}}{%
\pgfpathmoveto{\pgfqpoint{0.000000in}{-0.055556in}}%
\pgfpathcurveto{\pgfqpoint{0.014734in}{-0.055556in}}{\pgfqpoint{0.028866in}{-0.049702in}}{\pgfqpoint{0.039284in}{-0.039284in}}%
\pgfpathcurveto{\pgfqpoint{0.049702in}{-0.028866in}}{\pgfqpoint{0.055556in}{-0.014734in}}{\pgfqpoint{0.055556in}{0.000000in}}%
\pgfpathcurveto{\pgfqpoint{0.055556in}{0.014734in}}{\pgfqpoint{0.049702in}{0.028866in}}{\pgfqpoint{0.039284in}{0.039284in}}%
\pgfpathcurveto{\pgfqpoint{0.028866in}{0.049702in}}{\pgfqpoint{0.014734in}{0.055556in}}{\pgfqpoint{0.000000in}{0.055556in}}%
\pgfpathcurveto{\pgfqpoint{-0.014734in}{0.055556in}}{\pgfqpoint{-0.028866in}{0.049702in}}{\pgfqpoint{-0.039284in}{0.039284in}}%
\pgfpathcurveto{\pgfqpoint{-0.049702in}{0.028866in}}{\pgfqpoint{-0.055556in}{0.014734in}}{\pgfqpoint{-0.055556in}{0.000000in}}%
\pgfpathcurveto{\pgfqpoint{-0.055556in}{-0.014734in}}{\pgfqpoint{-0.049702in}{-0.028866in}}{\pgfqpoint{-0.039284in}{-0.039284in}}%
\pgfpathcurveto{\pgfqpoint{-0.028866in}{-0.049702in}}{\pgfqpoint{-0.014734in}{-0.055556in}}{\pgfqpoint{0.000000in}{-0.055556in}}%
\pgfpathlineto{\pgfqpoint{0.000000in}{-0.055556in}}%
\pgfpathclose%
\pgfusepath{stroke,fill}%
}%
\begin{pgfscope}%
\pgfsys@transformshift{0.674954in}{4.730650in}%
\pgfsys@useobject{currentmarker}{}%
\end{pgfscope}%
\begin{pgfscope}%
\pgfsys@transformshift{0.998577in}{4.718620in}%
\pgfsys@useobject{currentmarker}{}%
\end{pgfscope}%
\begin{pgfscope}%
\pgfsys@transformshift{2.293069in}{4.610217in}%
\pgfsys@useobject{currentmarker}{}%
\end{pgfscope}%
\begin{pgfscope}%
\pgfsys@transformshift{3.911184in}{4.512449in}%
\pgfsys@useobject{currentmarker}{}%
\end{pgfscope}%
\begin{pgfscope}%
\pgfsys@transformshift{5.529299in}{4.236181in}%
\pgfsys@useobject{currentmarker}{}%
\end{pgfscope}%
\begin{pgfscope}%
\pgfsys@transformshift{6.823791in}{3.617352in}%
\pgfsys@useobject{currentmarker}{}%
\end{pgfscope}%
\end{pgfscope}%
\begin{pgfscope}%
\pgfpathrectangle{\pgfqpoint{0.674954in}{0.862305in}}{\pgfqpoint{6.472460in}{4.369081in}}%
\pgfusepath{clip}%
\pgfsetbuttcap%
\pgfsetroundjoin%
\pgfsetlinewidth{2.509375pt}%
\definecolor{currentstroke}{rgb}{0.960784,0.462745,0.000000}%
\pgfsetstrokecolor{currentstroke}%
\pgfsetdash{{9.250000pt}{4.000000pt}}{0.000000pt}%
\pgfpathmoveto{\pgfqpoint{0.998577in}{3.743510in}}%
\pgfpathlineto{\pgfqpoint{2.293069in}{3.927041in}}%
\pgfpathlineto{\pgfqpoint{3.911184in}{3.877264in}}%
\pgfpathlineto{\pgfqpoint{5.529299in}{3.818975in}}%
\pgfpathlineto{\pgfqpoint{6.823791in}{3.894803in}}%
\pgfpathlineto{\pgfqpoint{7.147414in}{4.000678in}}%
\pgfusepath{stroke}%
\end{pgfscope}%
\begin{pgfscope}%
\pgfpathrectangle{\pgfqpoint{0.674954in}{0.862305in}}{\pgfqpoint{6.472460in}{4.369081in}}%
\pgfusepath{clip}%
\pgfsetbuttcap%
\pgfsetmiterjoin%
\definecolor{currentfill}{rgb}{0.960784,0.462745,0.000000}%
\pgfsetfillcolor{currentfill}%
\pgfsetlinewidth{1.003750pt}%
\definecolor{currentstroke}{rgb}{0.960784,0.462745,0.000000}%
\pgfsetstrokecolor{currentstroke}%
\pgfsetdash{}{0pt}%
\pgfsys@defobject{currentmarker}{\pgfqpoint{-0.055556in}{-0.055556in}}{\pgfqpoint{0.055556in}{0.055556in}}{%
\pgfpathmoveto{\pgfqpoint{-0.055556in}{-0.055556in}}%
\pgfpathlineto{\pgfqpoint{0.055556in}{-0.055556in}}%
\pgfpathlineto{\pgfqpoint{0.055556in}{0.055556in}}%
\pgfpathlineto{\pgfqpoint{-0.055556in}{0.055556in}}%
\pgfpathlineto{\pgfqpoint{-0.055556in}{-0.055556in}}%
\pgfpathclose%
\pgfusepath{stroke,fill}%
}%
\begin{pgfscope}%
\pgfsys@transformshift{0.998577in}{3.743510in}%
\pgfsys@useobject{currentmarker}{}%
\end{pgfscope}%
\begin{pgfscope}%
\pgfsys@transformshift{2.293069in}{3.927041in}%
\pgfsys@useobject{currentmarker}{}%
\end{pgfscope}%
\begin{pgfscope}%
\pgfsys@transformshift{3.911184in}{3.877264in}%
\pgfsys@useobject{currentmarker}{}%
\end{pgfscope}%
\begin{pgfscope}%
\pgfsys@transformshift{5.529299in}{3.818975in}%
\pgfsys@useobject{currentmarker}{}%
\end{pgfscope}%
\begin{pgfscope}%
\pgfsys@transformshift{6.823791in}{3.894803in}%
\pgfsys@useobject{currentmarker}{}%
\end{pgfscope}%
\begin{pgfscope}%
\pgfsys@transformshift{7.147414in}{4.000678in}%
\pgfsys@useobject{currentmarker}{}%
\end{pgfscope}%
\end{pgfscope}%
\begin{pgfscope}%
\pgfsetrectcap%
\pgfsetmiterjoin%
\pgfsetlinewidth{2.007500pt}%
\definecolor{currentstroke}{rgb}{0.000000,0.000000,0.000000}%
\pgfsetstrokecolor{currentstroke}%
\pgfsetdash{}{0pt}%
\pgfpathmoveto{\pgfqpoint{0.674954in}{0.862305in}}%
\pgfpathlineto{\pgfqpoint{0.674954in}{5.231386in}}%
\pgfusepath{stroke}%
\end{pgfscope}%
\begin{pgfscope}%
\pgfsetrectcap%
\pgfsetmiterjoin%
\pgfsetlinewidth{2.007500pt}%
\definecolor{currentstroke}{rgb}{0.000000,0.000000,0.000000}%
\pgfsetstrokecolor{currentstroke}%
\pgfsetdash{}{0pt}%
\pgfpathmoveto{\pgfqpoint{0.674954in}{0.862305in}}%
\pgfpathlineto{\pgfqpoint{7.147414in}{0.862305in}}%
\pgfusepath{stroke}%
\end{pgfscope}%
\begin{pgfscope}%
\pgfsetbuttcap%
\pgfsetmiterjoin%
\definecolor{currentfill}{rgb}{1.000000,1.000000,1.000000}%
\pgfsetfillcolor{currentfill}%
\pgfsetlinewidth{1.003750pt}%
\definecolor{currentstroke}{rgb}{0.800000,0.800000,0.800000}%
\pgfsetstrokecolor{currentstroke}%
\pgfsetdash{}{0pt}%
\pgfpathmoveto{\pgfqpoint{0.869398in}{1.001194in}}%
\pgfpathlineto{\pgfqpoint{2.708109in}{1.001194in}}%
\pgfpathquadraticcurveto{\pgfqpoint{2.763664in}{1.001194in}}{\pgfqpoint{2.763664in}{1.056749in}}%
\pgfpathlineto{\pgfqpoint{2.763664in}{1.803694in}}%
\pgfpathquadraticcurveto{\pgfqpoint{2.763664in}{1.859250in}}{\pgfqpoint{2.708109in}{1.859250in}}%
\pgfpathlineto{\pgfqpoint{0.869398in}{1.859250in}}%
\pgfpathquadraticcurveto{\pgfqpoint{0.813843in}{1.859250in}}{\pgfqpoint{0.813843in}{1.803694in}}%
\pgfpathlineto{\pgfqpoint{0.813843in}{1.056749in}}%
\pgfpathquadraticcurveto{\pgfqpoint{0.813843in}{1.001194in}}{\pgfqpoint{0.869398in}{1.001194in}}%
\pgfpathlineto{\pgfqpoint{0.869398in}{1.001194in}}%
\pgfpathclose%
\pgfusepath{stroke,fill}%
\end{pgfscope}%
\begin{pgfscope}%
\pgfsetrectcap%
\pgfsetroundjoin%
\pgfsetlinewidth{2.509375pt}%
\definecolor{currentstroke}{rgb}{0.050980,0.415686,0.509804}%
\pgfsetstrokecolor{currentstroke}%
\pgfsetdash{}{0pt}%
\pgfpathmoveto{\pgfqpoint{0.924954in}{1.650917in}}%
\pgfpathlineto{\pgfqpoint{1.202732in}{1.650917in}}%
\pgfpathlineto{\pgfqpoint{1.480509in}{1.650917in}}%
\pgfusepath{stroke}%
\end{pgfscope}%
\begin{pgfscope}%
\pgfsetbuttcap%
\pgfsetroundjoin%
\definecolor{currentfill}{rgb}{0.050980,0.415686,0.509804}%
\pgfsetfillcolor{currentfill}%
\pgfsetlinewidth{1.003750pt}%
\definecolor{currentstroke}{rgb}{0.050980,0.415686,0.509804}%
\pgfsetstrokecolor{currentstroke}%
\pgfsetdash{}{0pt}%
\pgfsys@defobject{currentmarker}{\pgfqpoint{-0.055556in}{-0.055556in}}{\pgfqpoint{0.055556in}{0.055556in}}{%
\pgfpathmoveto{\pgfqpoint{0.000000in}{-0.055556in}}%
\pgfpathcurveto{\pgfqpoint{0.014734in}{-0.055556in}}{\pgfqpoint{0.028866in}{-0.049702in}}{\pgfqpoint{0.039284in}{-0.039284in}}%
\pgfpathcurveto{\pgfqpoint{0.049702in}{-0.028866in}}{\pgfqpoint{0.055556in}{-0.014734in}}{\pgfqpoint{0.055556in}{0.000000in}}%
\pgfpathcurveto{\pgfqpoint{0.055556in}{0.014734in}}{\pgfqpoint{0.049702in}{0.028866in}}{\pgfqpoint{0.039284in}{0.039284in}}%
\pgfpathcurveto{\pgfqpoint{0.028866in}{0.049702in}}{\pgfqpoint{0.014734in}{0.055556in}}{\pgfqpoint{0.000000in}{0.055556in}}%
\pgfpathcurveto{\pgfqpoint{-0.014734in}{0.055556in}}{\pgfqpoint{-0.028866in}{0.049702in}}{\pgfqpoint{-0.039284in}{0.039284in}}%
\pgfpathcurveto{\pgfqpoint{-0.049702in}{0.028866in}}{\pgfqpoint{-0.055556in}{0.014734in}}{\pgfqpoint{-0.055556in}{0.000000in}}%
\pgfpathcurveto{\pgfqpoint{-0.055556in}{-0.014734in}}{\pgfqpoint{-0.049702in}{-0.028866in}}{\pgfqpoint{-0.039284in}{-0.039284in}}%
\pgfpathcurveto{\pgfqpoint{-0.028866in}{-0.049702in}}{\pgfqpoint{-0.014734in}{-0.055556in}}{\pgfqpoint{0.000000in}{-0.055556in}}%
\pgfpathlineto{\pgfqpoint{0.000000in}{-0.055556in}}%
\pgfpathclose%
\pgfusepath{stroke,fill}%
}%
\begin{pgfscope}%
\pgfsys@transformshift{1.202732in}{1.650917in}%
\pgfsys@useobject{currentmarker}{}%
\end{pgfscope}%
\end{pgfscope}%
\begin{pgfscope}%
\definecolor{textcolor}{rgb}{0.000000,0.000000,0.000000}%
\pgfsetstrokecolor{textcolor}%
\pgfsetfillcolor{textcolor}%
\pgftext[x=1.702732in,y=1.553694in,left,base]{\color{textcolor}{\rmfamily\fontsize{20.000000}{24.000000}\selectfont\catcode`\^=\active\def^{\ifmmode\sp\else\^{}\fi}\catcode`\%=\active\def
\end{pgfscope}%
\begin{pgfscope}%
\pgfsetbuttcap%
\pgfsetroundjoin%
\pgfsetlinewidth{2.509375pt}%
\definecolor{currentstroke}{rgb}{0.960784,0.462745,0.000000}%
\pgfsetstrokecolor{currentstroke}%
\pgfsetdash{{9.250000pt}{4.000000pt}}{0.000000pt}%
\pgfpathmoveto{\pgfqpoint{0.924954in}{1.263555in}}%
\pgfpathlineto{\pgfqpoint{1.202732in}{1.263555in}}%
\pgfpathlineto{\pgfqpoint{1.480509in}{1.263555in}}%
\pgfusepath{stroke}%
\end{pgfscope}%
\begin{pgfscope}%
\pgfsetbuttcap%
\pgfsetmiterjoin%
\definecolor{currentfill}{rgb}{0.960784,0.462745,0.000000}%
\pgfsetfillcolor{currentfill}%
\pgfsetlinewidth{1.003750pt}%
\definecolor{currentstroke}{rgb}{0.960784,0.462745,0.000000}%
\pgfsetstrokecolor{currentstroke}%
\pgfsetdash{}{0pt}%
\pgfsys@defobject{currentmarker}{\pgfqpoint{-0.055556in}{-0.055556in}}{\pgfqpoint{0.055556in}{0.055556in}}{%
\pgfpathmoveto{\pgfqpoint{-0.055556in}{-0.055556in}}%
\pgfpathlineto{\pgfqpoint{0.055556in}{-0.055556in}}%
\pgfpathlineto{\pgfqpoint{0.055556in}{0.055556in}}%
\pgfpathlineto{\pgfqpoint{-0.055556in}{0.055556in}}%
\pgfpathlineto{\pgfqpoint{-0.055556in}{-0.055556in}}%
\pgfpathclose%
\pgfusepath{stroke,fill}%
}%
\begin{pgfscope}%
\pgfsys@transformshift{1.202732in}{1.263555in}%
\pgfsys@useobject{currentmarker}{}%
\end{pgfscope}%
\end{pgfscope}%
\begin{pgfscope}%
\definecolor{textcolor}{rgb}{0.000000,0.000000,0.000000}%
\pgfsetstrokecolor{textcolor}%
\pgfsetfillcolor{textcolor}%
\pgftext[x=1.702732in,y=1.166333in,left,base]{\color{textcolor}{\rmfamily\fontsize{20.000000}{24.000000}\selectfont\catcode`\^=\active\def^{\ifmmode\sp\else\^{}\fi}\catcode`\%=\active\def
\end{pgfscope}%
\end{pgfpicture}%
\makeatother%
\endgroup%

%% file: images/thermal_ratio_per_modality/modality_results_RRA_5.pgf
\begingroup%
\makeatletter%
\begin{pgfpicture}%
\pgfpathrectangle{\pgfpointorigin}{\pgfqpoint{7.450000in}{5.450000in}}%
\pgfusepath{use as bounding box, clip}%
\begin{pgfscope}%
\pgfsetbuttcap%
\pgfsetmiterjoin%
\definecolor{currentfill}{rgb}{1.000000,1.000000,1.000000}%
\pgfsetfillcolor{currentfill}%
\pgfsetlinewidth{0.000000pt}%
\definecolor{currentstroke}{rgb}{1.000000,1.000000,1.000000}%
\pgfsetstrokecolor{currentstroke}%
\pgfsetdash{}{0pt}%
\pgfpathmoveto{\pgfqpoint{0.000000in}{0.000000in}}%
\pgfpathlineto{\pgfqpoint{7.450000in}{0.000000in}}%
\pgfpathlineto{\pgfqpoint{7.450000in}{5.450000in}}%
\pgfpathlineto{\pgfqpoint{0.000000in}{5.450000in}}%
\pgfpathlineto{\pgfqpoint{0.000000in}{0.000000in}}%
\pgfpathclose%
\pgfusepath{fill}%
\end{pgfscope}%
\begin{pgfscope}%
\pgfsetbuttcap%
\pgfsetmiterjoin%
\definecolor{currentfill}{rgb}{1.000000,1.000000,1.000000}%
\pgfsetfillcolor{currentfill}%
\pgfsetlinewidth{0.000000pt}%
\definecolor{currentstroke}{rgb}{0.000000,0.000000,0.000000}%
\pgfsetstrokecolor{currentstroke}%
\pgfsetstrokeopacity{0.000000}%
\pgfsetdash{}{0pt}%
\pgfpathmoveto{\pgfqpoint{0.674954in}{0.862305in}}%
\pgfpathlineto{\pgfqpoint{7.147414in}{0.862305in}}%
\pgfpathlineto{\pgfqpoint{7.147414in}{5.231386in}}%
\pgfpathlineto{\pgfqpoint{0.674954in}{5.231386in}}%
\pgfpathlineto{\pgfqpoint{0.674954in}{0.862305in}}%
\pgfpathclose%
\pgfusepath{fill}%
\end{pgfscope}%
\begin{pgfscope}%
\pgfpathrectangle{\pgfqpoint{0.674954in}{0.862305in}}{\pgfqpoint{6.472460in}{4.369081in}}%
\pgfusepath{clip}%
\pgfsetbuttcap%
\pgfsetroundjoin%
\pgfsetlinewidth{2.007500pt}%
\definecolor{currentstroke}{rgb}{0.501961,0.501961,0.501961}%
\pgfsetstrokecolor{currentstroke}%
\pgfsetstrokeopacity{0.300000}%
\pgfsetdash{{7.400000pt}{3.200000pt}}{0.000000pt}%
\pgfpathmoveto{\pgfqpoint{0.674954in}{0.862305in}}%
\pgfpathlineto{\pgfqpoint{0.674954in}{5.231386in}}%
\pgfusepath{stroke}%
\end{pgfscope}%
\begin{pgfscope}%
\pgfsetbuttcap%
\pgfsetroundjoin%
\definecolor{currentfill}{rgb}{0.000000,0.000000,0.000000}%
\pgfsetfillcolor{currentfill}%
\pgfsetlinewidth{0.803000pt}%
\definecolor{currentstroke}{rgb}{0.000000,0.000000,0.000000}%
\pgfsetstrokecolor{currentstroke}%
\pgfsetdash{}{0pt}%
\pgfsys@defobject{currentmarker}{\pgfqpoint{0.000000in}{-0.048611in}}{\pgfqpoint{0.000000in}{0.000000in}}{%
\pgfpathmoveto{\pgfqpoint{0.000000in}{0.000000in}}%
\pgfpathlineto{\pgfqpoint{0.000000in}{-0.048611in}}%
\pgfusepath{stroke,fill}%
}%
\begin{pgfscope}%
\pgfsys@transformshift{0.674954in}{0.862305in}%
\pgfsys@useobject{currentmarker}{}%
\end{pgfscope}%
\end{pgfscope}%
\begin{pgfscope}%
\definecolor{textcolor}{rgb}{0.000000,0.000000,0.000000}%
\pgfsetstrokecolor{textcolor}%
\pgfsetfillcolor{textcolor}%
\pgftext[x=0.674954in,y=0.765082in,,top]{\color{textcolor}{\rmfamily\fontsize{25.000000}{30.000000}\selectfont\catcode`\^=\active\def^{\ifmmode\sp\else\^{}\fi}\catcode`\%=\active\def
\end{pgfscope}%
\begin{pgfscope}%
\pgfpathrectangle{\pgfqpoint{0.674954in}{0.862305in}}{\pgfqpoint{6.472460in}{4.369081in}}%
\pgfusepath{clip}%
\pgfsetbuttcap%
\pgfsetroundjoin%
\pgfsetlinewidth{2.007500pt}%
\definecolor{currentstroke}{rgb}{0.501961,0.501961,0.501961}%
\pgfsetstrokecolor{currentstroke}%
\pgfsetstrokeopacity{0.300000}%
\pgfsetdash{{7.400000pt}{3.200000pt}}{0.000000pt}%
\pgfpathmoveto{\pgfqpoint{1.969446in}{0.862305in}}%
\pgfpathlineto{\pgfqpoint{1.969446in}{5.231386in}}%
\pgfusepath{stroke}%
\end{pgfscope}%
\begin{pgfscope}%
\pgfsetbuttcap%
\pgfsetroundjoin%
\definecolor{currentfill}{rgb}{0.000000,0.000000,0.000000}%
\pgfsetfillcolor{currentfill}%
\pgfsetlinewidth{0.803000pt}%
\definecolor{currentstroke}{rgb}{0.000000,0.000000,0.000000}%
\pgfsetstrokecolor{currentstroke}%
\pgfsetdash{}{0pt}%
\pgfsys@defobject{currentmarker}{\pgfqpoint{0.000000in}{-0.048611in}}{\pgfqpoint{0.000000in}{0.000000in}}{%
\pgfpathmoveto{\pgfqpoint{0.000000in}{0.000000in}}%
\pgfpathlineto{\pgfqpoint{0.000000in}{-0.048611in}}%
\pgfusepath{stroke,fill}%
}%
\begin{pgfscope}%
\pgfsys@transformshift{1.969446in}{0.862305in}%
\pgfsys@useobject{currentmarker}{}%
\end{pgfscope}%
\end{pgfscope}%
\begin{pgfscope}%
\definecolor{textcolor}{rgb}{0.000000,0.000000,0.000000}%
\pgfsetstrokecolor{textcolor}%
\pgfsetfillcolor{textcolor}%
\pgftext[x=1.969446in,y=0.765082in,,top]{\color{textcolor}{\rmfamily\fontsize{25.000000}{30.000000}\selectfont\catcode`\^=\active\def^{\ifmmode\sp\else\^{}\fi}\catcode`\%=\active\def
\end{pgfscope}%
\begin{pgfscope}%
\pgfpathrectangle{\pgfqpoint{0.674954in}{0.862305in}}{\pgfqpoint{6.472460in}{4.369081in}}%
\pgfusepath{clip}%
\pgfsetbuttcap%
\pgfsetroundjoin%
\pgfsetlinewidth{2.007500pt}%
\definecolor{currentstroke}{rgb}{0.501961,0.501961,0.501961}%
\pgfsetstrokecolor{currentstroke}%
\pgfsetstrokeopacity{0.300000}%
\pgfsetdash{{7.400000pt}{3.200000pt}}{0.000000pt}%
\pgfpathmoveto{\pgfqpoint{3.263938in}{0.862305in}}%
\pgfpathlineto{\pgfqpoint{3.263938in}{5.231386in}}%
\pgfusepath{stroke}%
\end{pgfscope}%
\begin{pgfscope}%
\pgfsetbuttcap%
\pgfsetroundjoin%
\definecolor{currentfill}{rgb}{0.000000,0.000000,0.000000}%
\pgfsetfillcolor{currentfill}%
\pgfsetlinewidth{0.803000pt}%
\definecolor{currentstroke}{rgb}{0.000000,0.000000,0.000000}%
\pgfsetstrokecolor{currentstroke}%
\pgfsetdash{}{0pt}%
\pgfsys@defobject{currentmarker}{\pgfqpoint{0.000000in}{-0.048611in}}{\pgfqpoint{0.000000in}{0.000000in}}{%
\pgfpathmoveto{\pgfqpoint{0.000000in}{0.000000in}}%
\pgfpathlineto{\pgfqpoint{0.000000in}{-0.048611in}}%
\pgfusepath{stroke,fill}%
}%
\begin{pgfscope}%
\pgfsys@transformshift{3.263938in}{0.862305in}%
\pgfsys@useobject{currentmarker}{}%
\end{pgfscope}%
\end{pgfscope}%
\begin{pgfscope}%
\definecolor{textcolor}{rgb}{0.000000,0.000000,0.000000}%
\pgfsetstrokecolor{textcolor}%
\pgfsetfillcolor{textcolor}%
\pgftext[x=3.263938in,y=0.765082in,,top]{\color{textcolor}{\rmfamily\fontsize{25.000000}{30.000000}\selectfont\catcode`\^=\active\def^{\ifmmode\sp\else\^{}\fi}\catcode`\%=\active\def
\end{pgfscope}%
\begin{pgfscope}%
\pgfpathrectangle{\pgfqpoint{0.674954in}{0.862305in}}{\pgfqpoint{6.472460in}{4.369081in}}%
\pgfusepath{clip}%
\pgfsetbuttcap%
\pgfsetroundjoin%
\pgfsetlinewidth{2.007500pt}%
\definecolor{currentstroke}{rgb}{0.501961,0.501961,0.501961}%
\pgfsetstrokecolor{currentstroke}%
\pgfsetstrokeopacity{0.300000}%
\pgfsetdash{{7.400000pt}{3.200000pt}}{0.000000pt}%
\pgfpathmoveto{\pgfqpoint{4.558430in}{0.862305in}}%
\pgfpathlineto{\pgfqpoint{4.558430in}{5.231386in}}%
\pgfusepath{stroke}%
\end{pgfscope}%
\begin{pgfscope}%
\pgfsetbuttcap%
\pgfsetroundjoin%
\definecolor{currentfill}{rgb}{0.000000,0.000000,0.000000}%
\pgfsetfillcolor{currentfill}%
\pgfsetlinewidth{0.803000pt}%
\definecolor{currentstroke}{rgb}{0.000000,0.000000,0.000000}%
\pgfsetstrokecolor{currentstroke}%
\pgfsetdash{}{0pt}%
\pgfsys@defobject{currentmarker}{\pgfqpoint{0.000000in}{-0.048611in}}{\pgfqpoint{0.000000in}{0.000000in}}{%
\pgfpathmoveto{\pgfqpoint{0.000000in}{0.000000in}}%
\pgfpathlineto{\pgfqpoint{0.000000in}{-0.048611in}}%
\pgfusepath{stroke,fill}%
}%
\begin{pgfscope}%
\pgfsys@transformshift{4.558430in}{0.862305in}%
\pgfsys@useobject{currentmarker}{}%
\end{pgfscope}%
\end{pgfscope}%
\begin{pgfscope}%
\definecolor{textcolor}{rgb}{0.000000,0.000000,0.000000}%
\pgfsetstrokecolor{textcolor}%
\pgfsetfillcolor{textcolor}%
\pgftext[x=4.558430in,y=0.765082in,,top]{\color{textcolor}{\rmfamily\fontsize{25.000000}{30.000000}\selectfont\catcode`\^=\active\def^{\ifmmode\sp\else\^{}\fi}\catcode`\%=\active\def
\end{pgfscope}%
\begin{pgfscope}%
\pgfpathrectangle{\pgfqpoint{0.674954in}{0.862305in}}{\pgfqpoint{6.472460in}{4.369081in}}%
\pgfusepath{clip}%
\pgfsetbuttcap%
\pgfsetroundjoin%
\pgfsetlinewidth{2.007500pt}%
\definecolor{currentstroke}{rgb}{0.501961,0.501961,0.501961}%
\pgfsetstrokecolor{currentstroke}%
\pgfsetstrokeopacity{0.300000}%
\pgfsetdash{{7.400000pt}{3.200000pt}}{0.000000pt}%
\pgfpathmoveto{\pgfqpoint{5.852922in}{0.862305in}}%
\pgfpathlineto{\pgfqpoint{5.852922in}{5.231386in}}%
\pgfusepath{stroke}%
\end{pgfscope}%
\begin{pgfscope}%
\pgfsetbuttcap%
\pgfsetroundjoin%
\definecolor{currentfill}{rgb}{0.000000,0.000000,0.000000}%
\pgfsetfillcolor{currentfill}%
\pgfsetlinewidth{0.803000pt}%
\definecolor{currentstroke}{rgb}{0.000000,0.000000,0.000000}%
\pgfsetstrokecolor{currentstroke}%
\pgfsetdash{}{0pt}%
\pgfsys@defobject{currentmarker}{\pgfqpoint{0.000000in}{-0.048611in}}{\pgfqpoint{0.000000in}{0.000000in}}{%
\pgfpathmoveto{\pgfqpoint{0.000000in}{0.000000in}}%
\pgfpathlineto{\pgfqpoint{0.000000in}{-0.048611in}}%
\pgfusepath{stroke,fill}%
}%
\begin{pgfscope}%
\pgfsys@transformshift{5.852922in}{0.862305in}%
\pgfsys@useobject{currentmarker}{}%
\end{pgfscope}%
\end{pgfscope}%
\begin{pgfscope}%
\definecolor{textcolor}{rgb}{0.000000,0.000000,0.000000}%
\pgfsetstrokecolor{textcolor}%
\pgfsetfillcolor{textcolor}%
\pgftext[x=5.852922in,y=0.765082in,,top]{\color{textcolor}{\rmfamily\fontsize{25.000000}{30.000000}\selectfont\catcode`\^=\active\def^{\ifmmode\sp\else\^{}\fi}\catcode`\%=\active\def
\end{pgfscope}%
\begin{pgfscope}%
\pgfpathrectangle{\pgfqpoint{0.674954in}{0.862305in}}{\pgfqpoint{6.472460in}{4.369081in}}%
\pgfusepath{clip}%
\pgfsetbuttcap%
\pgfsetroundjoin%
\pgfsetlinewidth{2.007500pt}%
\definecolor{currentstroke}{rgb}{0.501961,0.501961,0.501961}%
\pgfsetstrokecolor{currentstroke}%
\pgfsetstrokeopacity{0.300000}%
\pgfsetdash{{7.400000pt}{3.200000pt}}{0.000000pt}%
\pgfpathmoveto{\pgfqpoint{7.147414in}{0.862305in}}%
\pgfpathlineto{\pgfqpoint{7.147414in}{5.231386in}}%
\pgfusepath{stroke}%
\end{pgfscope}%
\begin{pgfscope}%
\pgfsetbuttcap%
\pgfsetroundjoin%
\definecolor{currentfill}{rgb}{0.000000,0.000000,0.000000}%
\pgfsetfillcolor{currentfill}%
\pgfsetlinewidth{0.803000pt}%
\definecolor{currentstroke}{rgb}{0.000000,0.000000,0.000000}%
\pgfsetstrokecolor{currentstroke}%
\pgfsetdash{}{0pt}%
\pgfsys@defobject{currentmarker}{\pgfqpoint{0.000000in}{-0.048611in}}{\pgfqpoint{0.000000in}{0.000000in}}{%
\pgfpathmoveto{\pgfqpoint{0.000000in}{0.000000in}}%
\pgfpathlineto{\pgfqpoint{0.000000in}{-0.048611in}}%
\pgfusepath{stroke,fill}%
}%
\begin{pgfscope}%
\pgfsys@transformshift{7.147414in}{0.862305in}%
\pgfsys@useobject{currentmarker}{}%
\end{pgfscope}%
\end{pgfscope}%
\begin{pgfscope}%
\definecolor{textcolor}{rgb}{0.000000,0.000000,0.000000}%
\pgfsetstrokecolor{textcolor}%
\pgfsetfillcolor{textcolor}%
\pgftext[x=7.147414in,y=0.765082in,,top]{\color{textcolor}{\rmfamily\fontsize{25.000000}{30.000000}\selectfont\catcode`\^=\active\def^{\ifmmode\sp\else\^{}\fi}\catcode`\%=\active\def
\end{pgfscope}%
\begin{pgfscope}%
\definecolor{textcolor}{rgb}{0.000000,0.000000,0.000000}%
\pgfsetstrokecolor{textcolor}%
\pgfsetfillcolor{textcolor}%
\pgftext[x=3.911184in,y=0.404763in,,top]{\color{textcolor}{\rmfamily\fontsize{25.000000}{30.000000}\selectfont\catcode`\^=\active\def^{\ifmmode\sp\else\^{}\fi}\catcode`\%=\active\def
\end{pgfscope}%
\begin{pgfscope}%
\pgfpathrectangle{\pgfqpoint{0.674954in}{0.862305in}}{\pgfqpoint{6.472460in}{4.369081in}}%
\pgfusepath{clip}%
\pgfsetbuttcap%
\pgfsetroundjoin%
\pgfsetlinewidth{2.007500pt}%
\definecolor{currentstroke}{rgb}{0.501961,0.501961,0.501961}%
\pgfsetstrokecolor{currentstroke}%
\pgfsetstrokeopacity{0.300000}%
\pgfsetdash{{7.400000pt}{3.200000pt}}{0.000000pt}%
\pgfpathmoveto{\pgfqpoint{0.674954in}{1.954575in}}%
\pgfpathlineto{\pgfqpoint{7.147414in}{1.954575in}}%
\pgfusepath{stroke}%
\end{pgfscope}%
\begin{pgfscope}%
\pgfsetbuttcap%
\pgfsetroundjoin%
\definecolor{currentfill}{rgb}{0.000000,0.000000,0.000000}%
\pgfsetfillcolor{currentfill}%
\pgfsetlinewidth{0.803000pt}%
\definecolor{currentstroke}{rgb}{0.000000,0.000000,0.000000}%
\pgfsetstrokecolor{currentstroke}%
\pgfsetdash{}{0pt}%
\pgfsys@defobject{currentmarker}{\pgfqpoint{-0.048611in}{0.000000in}}{\pgfqpoint{-0.000000in}{0.000000in}}{%
\pgfpathmoveto{\pgfqpoint{-0.000000in}{0.000000in}}%
\pgfpathlineto{\pgfqpoint{-0.048611in}{0.000000in}}%
\pgfusepath{stroke,fill}%
}%
\begin{pgfscope}%
\pgfsys@transformshift{0.674954in}{1.954575in}%
\pgfsys@useobject{currentmarker}{}%
\end{pgfscope}%
\end{pgfscope}%
\begin{pgfscope}%
\definecolor{textcolor}{rgb}{0.000000,0.000000,0.000000}%
\pgfsetstrokecolor{textcolor}%
\pgfsetfillcolor{textcolor}%
\pgftext[x=0.259244in, y=1.835961in, left, base]{\color{textcolor}{\rmfamily\fontsize{25.000000}{30.000000}\selectfont\catcode`\^=\active\def^{\ifmmode\sp\else\^{}\fi}\catcode`\%=\active\def
\end{pgfscope}%
\begin{pgfscope}%
\pgfpathrectangle{\pgfqpoint{0.674954in}{0.862305in}}{\pgfqpoint{6.472460in}{4.369081in}}%
\pgfusepath{clip}%
\pgfsetbuttcap%
\pgfsetroundjoin%
\pgfsetlinewidth{2.007500pt}%
\definecolor{currentstroke}{rgb}{0.501961,0.501961,0.501961}%
\pgfsetstrokecolor{currentstroke}%
\pgfsetstrokeopacity{0.300000}%
\pgfsetdash{{7.400000pt}{3.200000pt}}{0.000000pt}%
\pgfpathmoveto{\pgfqpoint{0.674954in}{3.046845in}}%
\pgfpathlineto{\pgfqpoint{7.147414in}{3.046845in}}%
\pgfusepath{stroke}%
\end{pgfscope}%
\begin{pgfscope}%
\pgfsetbuttcap%
\pgfsetroundjoin%
\definecolor{currentfill}{rgb}{0.000000,0.000000,0.000000}%
\pgfsetfillcolor{currentfill}%
\pgfsetlinewidth{0.803000pt}%
\definecolor{currentstroke}{rgb}{0.000000,0.000000,0.000000}%
\pgfsetstrokecolor{currentstroke}%
\pgfsetdash{}{0pt}%
\pgfsys@defobject{currentmarker}{\pgfqpoint{-0.048611in}{0.000000in}}{\pgfqpoint{-0.000000in}{0.000000in}}{%
\pgfpathmoveto{\pgfqpoint{-0.000000in}{0.000000in}}%
\pgfpathlineto{\pgfqpoint{-0.048611in}{0.000000in}}%
\pgfusepath{stroke,fill}%
}%
\begin{pgfscope}%
\pgfsys@transformshift{0.674954in}{3.046845in}%
\pgfsys@useobject{currentmarker}{}%
\end{pgfscope}%
\end{pgfscope}%
\begin{pgfscope}%
\definecolor{textcolor}{rgb}{0.000000,0.000000,0.000000}%
\pgfsetstrokecolor{textcolor}%
\pgfsetfillcolor{textcolor}%
\pgftext[x=0.259244in, y=2.928231in, left, base]{\color{textcolor}{\rmfamily\fontsize{25.000000}{30.000000}\selectfont\catcode`\^=\active\def^{\ifmmode\sp\else\^{}\fi}\catcode`\%=\active\def
\end{pgfscope}%
\begin{pgfscope}%
\pgfpathrectangle{\pgfqpoint{0.674954in}{0.862305in}}{\pgfqpoint{6.472460in}{4.369081in}}%
\pgfusepath{clip}%
\pgfsetbuttcap%
\pgfsetroundjoin%
\pgfsetlinewidth{2.007500pt}%
\definecolor{currentstroke}{rgb}{0.501961,0.501961,0.501961}%
\pgfsetstrokecolor{currentstroke}%
\pgfsetstrokeopacity{0.300000}%
\pgfsetdash{{7.400000pt}{3.200000pt}}{0.000000pt}%
\pgfpathmoveto{\pgfqpoint{0.674954in}{4.139116in}}%
\pgfpathlineto{\pgfqpoint{7.147414in}{4.139116in}}%
\pgfusepath{stroke}%
\end{pgfscope}%
\begin{pgfscope}%
\pgfsetbuttcap%
\pgfsetroundjoin%
\definecolor{currentfill}{rgb}{0.000000,0.000000,0.000000}%
\pgfsetfillcolor{currentfill}%
\pgfsetlinewidth{0.803000pt}%
\definecolor{currentstroke}{rgb}{0.000000,0.000000,0.000000}%
\pgfsetstrokecolor{currentstroke}%
\pgfsetdash{}{0pt}%
\pgfsys@defobject{currentmarker}{\pgfqpoint{-0.048611in}{0.000000in}}{\pgfqpoint{-0.000000in}{0.000000in}}{%
\pgfpathmoveto{\pgfqpoint{-0.000000in}{0.000000in}}%
\pgfpathlineto{\pgfqpoint{-0.048611in}{0.000000in}}%
\pgfusepath{stroke,fill}%
}%
\begin{pgfscope}%
\pgfsys@transformshift{0.674954in}{4.139116in}%
\pgfsys@useobject{currentmarker}{}%
\end{pgfscope}%
\end{pgfscope}%
\begin{pgfscope}%
\definecolor{textcolor}{rgb}{0.000000,0.000000,0.000000}%
\pgfsetstrokecolor{textcolor}%
\pgfsetfillcolor{textcolor}%
\pgftext[x=0.259244in, y=4.020501in, left, base]{\color{textcolor}{\rmfamily\fontsize{25.000000}{30.000000}\selectfont\catcode`\^=\active\def^{\ifmmode\sp\else\^{}\fi}\catcode`\%=\active\def
\end{pgfscope}%
\begin{pgfscope}%
\pgfpathrectangle{\pgfqpoint{0.674954in}{0.862305in}}{\pgfqpoint{6.472460in}{4.369081in}}%
\pgfusepath{clip}%
\pgfsetbuttcap%
\pgfsetroundjoin%
\pgfsetlinewidth{2.007500pt}%
\definecolor{currentstroke}{rgb}{0.501961,0.501961,0.501961}%
\pgfsetstrokecolor{currentstroke}%
\pgfsetstrokeopacity{0.300000}%
\pgfsetdash{{7.400000pt}{3.200000pt}}{0.000000pt}%
\pgfpathmoveto{\pgfqpoint{0.674954in}{5.231386in}}%
\pgfpathlineto{\pgfqpoint{7.147414in}{5.231386in}}%
\pgfusepath{stroke}%
\end{pgfscope}%
\begin{pgfscope}%
\pgfsetbuttcap%
\pgfsetroundjoin%
\definecolor{currentfill}{rgb}{0.000000,0.000000,0.000000}%
\pgfsetfillcolor{currentfill}%
\pgfsetlinewidth{0.803000pt}%
\definecolor{currentstroke}{rgb}{0.000000,0.000000,0.000000}%
\pgfsetstrokecolor{currentstroke}%
\pgfsetdash{}{0pt}%
\pgfsys@defobject{currentmarker}{\pgfqpoint{-0.048611in}{0.000000in}}{\pgfqpoint{-0.000000in}{0.000000in}}{%
\pgfpathmoveto{\pgfqpoint{-0.000000in}{0.000000in}}%
\pgfpathlineto{\pgfqpoint{-0.048611in}{0.000000in}}%
\pgfusepath{stroke,fill}%
}%
\begin{pgfscope}%
\pgfsys@transformshift{0.674954in}{5.231386in}%
\pgfsys@useobject{currentmarker}{}%
\end{pgfscope}%
\end{pgfscope}%
\begin{pgfscope}%
\definecolor{textcolor}{rgb}{0.000000,0.000000,0.000000}%
\pgfsetstrokecolor{textcolor}%
\pgfsetfillcolor{textcolor}%
\pgftext[x=0.100000in, y=5.112772in, left, base]{\color{textcolor}{\rmfamily\fontsize{25.000000}{30.000000}\selectfont\catcode`\^=\active\def^{\ifmmode\sp\else\^{}\fi}\catcode`\%=\active\def
\end{pgfscope}%
\begin{pgfscope}%
\pgfpathrectangle{\pgfqpoint{0.674954in}{0.862305in}}{\pgfqpoint{6.472460in}{4.369081in}}%
\pgfusepath{clip}%
\pgfsetrectcap%
\pgfsetroundjoin%
\pgfsetlinewidth{2.509375pt}%
\definecolor{currentstroke}{rgb}{0.050980,0.415686,0.509804}%
\pgfsetstrokecolor{currentstroke}%
\pgfsetdash{}{0pt}%
\pgfpathmoveto{\pgfqpoint{0.674954in}{5.221702in}}%
\pgfpathlineto{\pgfqpoint{0.998577in}{5.208089in}}%
\pgfpathlineto{\pgfqpoint{2.293069in}{4.963336in}}%
\pgfpathlineto{\pgfqpoint{3.911184in}{4.785987in}}%
\pgfpathlineto{\pgfqpoint{5.529299in}{4.528929in}}%
\pgfpathlineto{\pgfqpoint{6.823791in}{3.844376in}}%
\pgfusepath{stroke}%
\end{pgfscope}%
\begin{pgfscope}%
\pgfpathrectangle{\pgfqpoint{0.674954in}{0.862305in}}{\pgfqpoint{6.472460in}{4.369081in}}%
\pgfusepath{clip}%
\pgfsetbuttcap%
\pgfsetroundjoin%
\definecolor{currentfill}{rgb}{0.050980,0.415686,0.509804}%
\pgfsetfillcolor{currentfill}%
\pgfsetlinewidth{1.003750pt}%
\definecolor{currentstroke}{rgb}{0.050980,0.415686,0.509804}%
\pgfsetstrokecolor{currentstroke}%
\pgfsetdash{}{0pt}%
\pgfsys@defobject{currentmarker}{\pgfqpoint{-0.055556in}{-0.055556in}}{\pgfqpoint{0.055556in}{0.055556in}}{%
\pgfpathmoveto{\pgfqpoint{0.000000in}{-0.055556in}}%
\pgfpathcurveto{\pgfqpoint{0.014734in}{-0.055556in}}{\pgfqpoint{0.028866in}{-0.049702in}}{\pgfqpoint{0.039284in}{-0.039284in}}%
\pgfpathcurveto{\pgfqpoint{0.049702in}{-0.028866in}}{\pgfqpoint{0.055556in}{-0.014734in}}{\pgfqpoint{0.055556in}{0.000000in}}%
\pgfpathcurveto{\pgfqpoint{0.055556in}{0.014734in}}{\pgfqpoint{0.049702in}{0.028866in}}{\pgfqpoint{0.039284in}{0.039284in}}%
\pgfpathcurveto{\pgfqpoint{0.028866in}{0.049702in}}{\pgfqpoint{0.014734in}{0.055556in}}{\pgfqpoint{0.000000in}{0.055556in}}%
\pgfpathcurveto{\pgfqpoint{-0.014734in}{0.055556in}}{\pgfqpoint{-0.028866in}{0.049702in}}{\pgfqpoint{-0.039284in}{0.039284in}}%
\pgfpathcurveto{\pgfqpoint{-0.049702in}{0.028866in}}{\pgfqpoint{-0.055556in}{0.014734in}}{\pgfqpoint{-0.055556in}{0.000000in}}%
\pgfpathcurveto{\pgfqpoint{-0.055556in}{-0.014734in}}{\pgfqpoint{-0.049702in}{-0.028866in}}{\pgfqpoint{-0.039284in}{-0.039284in}}%
\pgfpathcurveto{\pgfqpoint{-0.028866in}{-0.049702in}}{\pgfqpoint{-0.014734in}{-0.055556in}}{\pgfqpoint{0.000000in}{-0.055556in}}%
\pgfpathlineto{\pgfqpoint{0.000000in}{-0.055556in}}%
\pgfpathclose%
\pgfusepath{stroke,fill}%
}%
\begin{pgfscope}%
\pgfsys@transformshift{0.674954in}{5.221702in}%
\pgfsys@useobject{currentmarker}{}%
\end{pgfscope}%
\begin{pgfscope}%
\pgfsys@transformshift{0.998577in}{5.208089in}%
\pgfsys@useobject{currentmarker}{}%
\end{pgfscope}%
\begin{pgfscope}%
\pgfsys@transformshift{2.293069in}{4.963336in}%
\pgfsys@useobject{currentmarker}{}%
\end{pgfscope}%
\begin{pgfscope}%
\pgfsys@transformshift{3.911184in}{4.785987in}%
\pgfsys@useobject{currentmarker}{}%
\end{pgfscope}%
\begin{pgfscope}%
\pgfsys@transformshift{5.529299in}{4.528929in}%
\pgfsys@useobject{currentmarker}{}%
\end{pgfscope}%
\begin{pgfscope}%
\pgfsys@transformshift{6.823791in}{3.844376in}%
\pgfsys@useobject{currentmarker}{}%
\end{pgfscope}%
\end{pgfscope}%
\begin{pgfscope}%
\pgfpathrectangle{\pgfqpoint{0.674954in}{0.862305in}}{\pgfqpoint{6.472460in}{4.369081in}}%
\pgfusepath{clip}%
\pgfsetbuttcap%
\pgfsetroundjoin%
\pgfsetlinewidth{2.509375pt}%
\definecolor{currentstroke}{rgb}{0.960784,0.462745,0.000000}%
\pgfsetstrokecolor{currentstroke}%
\pgfsetdash{{9.250000pt}{4.000000pt}}{0.000000pt}%
\pgfpathmoveto{\pgfqpoint{0.998577in}{3.801699in}}%
\pgfpathlineto{\pgfqpoint{2.293069in}{4.110692in}}%
\pgfpathlineto{\pgfqpoint{3.911184in}{4.142429in}}%
\pgfpathlineto{\pgfqpoint{5.529299in}{4.082631in}}%
\pgfpathlineto{\pgfqpoint{6.823791in}{4.142579in}}%
\pgfpathlineto{\pgfqpoint{7.147414in}{4.165490in}}%
\pgfusepath{stroke}%
\end{pgfscope}%
\begin{pgfscope}%
\pgfpathrectangle{\pgfqpoint{0.674954in}{0.862305in}}{\pgfqpoint{6.472460in}{4.369081in}}%
\pgfusepath{clip}%
\pgfsetbuttcap%
\pgfsetmiterjoin%
\definecolor{currentfill}{rgb}{0.960784,0.462745,0.000000}%
\pgfsetfillcolor{currentfill}%
\pgfsetlinewidth{1.003750pt}%
\definecolor{currentstroke}{rgb}{0.960784,0.462745,0.000000}%
\pgfsetstrokecolor{currentstroke}%
\pgfsetdash{}{0pt}%
\pgfsys@defobject{currentmarker}{\pgfqpoint{-0.055556in}{-0.055556in}}{\pgfqpoint{0.055556in}{0.055556in}}{%
\pgfpathmoveto{\pgfqpoint{-0.055556in}{-0.055556in}}%
\pgfpathlineto{\pgfqpoint{0.055556in}{-0.055556in}}%
\pgfpathlineto{\pgfqpoint{0.055556in}{0.055556in}}%
\pgfpathlineto{\pgfqpoint{-0.055556in}{0.055556in}}%
\pgfpathlineto{\pgfqpoint{-0.055556in}{-0.055556in}}%
\pgfpathclose%
\pgfusepath{stroke,fill}%
}%
\begin{pgfscope}%
\pgfsys@transformshift{0.998577in}{3.801699in}%
\pgfsys@useobject{currentmarker}{}%
\end{pgfscope}%
\begin{pgfscope}%
\pgfsys@transformshift{2.293069in}{4.110692in}%
\pgfsys@useobject{currentmarker}{}%
\end{pgfscope}%
\begin{pgfscope}%
\pgfsys@transformshift{3.911184in}{4.142429in}%
\pgfsys@useobject{currentmarker}{}%
\end{pgfscope}%
\begin{pgfscope}%
\pgfsys@transformshift{5.529299in}{4.082631in}%
\pgfsys@useobject{currentmarker}{}%
\end{pgfscope}%
\begin{pgfscope}%
\pgfsys@transformshift{6.823791in}{4.142579in}%
\pgfsys@useobject{currentmarker}{}%
\end{pgfscope}%
\begin{pgfscope}%
\pgfsys@transformshift{7.147414in}{4.165490in}%
\pgfsys@useobject{currentmarker}{}%
\end{pgfscope}%
\end{pgfscope}%
\begin{pgfscope}%
\pgfsetrectcap%
\pgfsetmiterjoin%
\pgfsetlinewidth{2.007500pt}%
\definecolor{currentstroke}{rgb}{0.000000,0.000000,0.000000}%
\pgfsetstrokecolor{currentstroke}%
\pgfsetdash{}{0pt}%
\pgfpathmoveto{\pgfqpoint{0.674954in}{0.862305in}}%
\pgfpathlineto{\pgfqpoint{0.674954in}{5.231386in}}%
\pgfusepath{stroke}%
\end{pgfscope}%
\begin{pgfscope}%
\pgfsetrectcap%
\pgfsetmiterjoin%
\pgfsetlinewidth{2.007500pt}%
\definecolor{currentstroke}{rgb}{0.000000,0.000000,0.000000}%
\pgfsetstrokecolor{currentstroke}%
\pgfsetdash{}{0pt}%
\pgfpathmoveto{\pgfqpoint{0.674954in}{0.862305in}}%
\pgfpathlineto{\pgfqpoint{7.147414in}{0.862305in}}%
\pgfusepath{stroke}%
\end{pgfscope}%
\begin{pgfscope}%
\pgfsetbuttcap%
\pgfsetmiterjoin%
\definecolor{currentfill}{rgb}{1.000000,1.000000,1.000000}%
\pgfsetfillcolor{currentfill}%
\pgfsetlinewidth{1.003750pt}%
\definecolor{currentstroke}{rgb}{0.800000,0.800000,0.800000}%
\pgfsetstrokecolor{currentstroke}%
\pgfsetdash{}{0pt}%
\pgfpathmoveto{\pgfqpoint{0.869398in}{1.001194in}}%
\pgfpathlineto{\pgfqpoint{2.708109in}{1.001194in}}%
\pgfpathquadraticcurveto{\pgfqpoint{2.763664in}{1.001194in}}{\pgfqpoint{2.763664in}{1.056749in}}%
\pgfpathlineto{\pgfqpoint{2.763664in}{1.803694in}}%
\pgfpathquadraticcurveto{\pgfqpoint{2.763664in}{1.859250in}}{\pgfqpoint{2.708109in}{1.859250in}}%
\pgfpathlineto{\pgfqpoint{0.869398in}{1.859250in}}%
\pgfpathquadraticcurveto{\pgfqpoint{0.813843in}{1.859250in}}{\pgfqpoint{0.813843in}{1.803694in}}%
\pgfpathlineto{\pgfqpoint{0.813843in}{1.056749in}}%
\pgfpathquadraticcurveto{\pgfqpoint{0.813843in}{1.001194in}}{\pgfqpoint{0.869398in}{1.001194in}}%
\pgfpathlineto{\pgfqpoint{0.869398in}{1.001194in}}%
\pgfpathclose%
\pgfusepath{stroke,fill}%
\end{pgfscope}%
\begin{pgfscope}%
\pgfsetrectcap%
\pgfsetroundjoin%
\pgfsetlinewidth{2.509375pt}%
\definecolor{currentstroke}{rgb}{0.050980,0.415686,0.509804}%
\pgfsetstrokecolor{currentstroke}%
\pgfsetdash{}{0pt}%
\pgfpathmoveto{\pgfqpoint{0.924954in}{1.650917in}}%
\pgfpathlineto{\pgfqpoint{1.202732in}{1.650917in}}%
\pgfpathlineto{\pgfqpoint{1.480509in}{1.650917in}}%
\pgfusepath{stroke}%
\end{pgfscope}%
\begin{pgfscope}%
\pgfsetbuttcap%
\pgfsetroundjoin%
\definecolor{currentfill}{rgb}{0.050980,0.415686,0.509804}%
\pgfsetfillcolor{currentfill}%
\pgfsetlinewidth{1.003750pt}%
\definecolor{currentstroke}{rgb}{0.050980,0.415686,0.509804}%
\pgfsetstrokecolor{currentstroke}%
\pgfsetdash{}{0pt}%
\pgfsys@defobject{currentmarker}{\pgfqpoint{-0.055556in}{-0.055556in}}{\pgfqpoint{0.055556in}{0.055556in}}{%
\pgfpathmoveto{\pgfqpoint{0.000000in}{-0.055556in}}%
\pgfpathcurveto{\pgfqpoint{0.014734in}{-0.055556in}}{\pgfqpoint{0.028866in}{-0.049702in}}{\pgfqpoint{0.039284in}{-0.039284in}}%
\pgfpathcurveto{\pgfqpoint{0.049702in}{-0.028866in}}{\pgfqpoint{0.055556in}{-0.014734in}}{\pgfqpoint{0.055556in}{0.000000in}}%
\pgfpathcurveto{\pgfqpoint{0.055556in}{0.014734in}}{\pgfqpoint{0.049702in}{0.028866in}}{\pgfqpoint{0.039284in}{0.039284in}}%
\pgfpathcurveto{\pgfqpoint{0.028866in}{0.049702in}}{\pgfqpoint{0.014734in}{0.055556in}}{\pgfqpoint{0.000000in}{0.055556in}}%
\pgfpathcurveto{\pgfqpoint{-0.014734in}{0.055556in}}{\pgfqpoint{-0.028866in}{0.049702in}}{\pgfqpoint{-0.039284in}{0.039284in}}%
\pgfpathcurveto{\pgfqpoint{-0.049702in}{0.028866in}}{\pgfqpoint{-0.055556in}{0.014734in}}{\pgfqpoint{-0.055556in}{0.000000in}}%
\pgfpathcurveto{\pgfqpoint{-0.055556in}{-0.014734in}}{\pgfqpoint{-0.049702in}{-0.028866in}}{\pgfqpoint{-0.039284in}{-0.039284in}}%
\pgfpathcurveto{\pgfqpoint{-0.028866in}{-0.049702in}}{\pgfqpoint{-0.014734in}{-0.055556in}}{\pgfqpoint{0.000000in}{-0.055556in}}%
\pgfpathlineto{\pgfqpoint{0.000000in}{-0.055556in}}%
\pgfpathclose%
\pgfusepath{stroke,fill}%
}%
\begin{pgfscope}%
\pgfsys@transformshift{1.202732in}{1.650917in}%
\pgfsys@useobject{currentmarker}{}%
\end{pgfscope}%
\end{pgfscope}%
\begin{pgfscope}%
\definecolor{textcolor}{rgb}{0.000000,0.000000,0.000000}%
\pgfsetstrokecolor{textcolor}%
\pgfsetfillcolor{textcolor}%
\pgftext[x=1.702732in,y=1.553694in,left,base]{\color{textcolor}{\rmfamily\fontsize{20.000000}{24.000000}\selectfont\catcode`\^=\active\def^{\ifmmode\sp\else\^{}\fi}\catcode`\%=\active\def
\end{pgfscope}%
\begin{pgfscope}%
\pgfsetbuttcap%
\pgfsetroundjoin%
\pgfsetlinewidth{2.509375pt}%
\definecolor{currentstroke}{rgb}{0.960784,0.462745,0.000000}%
\pgfsetstrokecolor{currentstroke}%
\pgfsetdash{{9.250000pt}{4.000000pt}}{0.000000pt}%
\pgfpathmoveto{\pgfqpoint{0.924954in}{1.263555in}}%
\pgfpathlineto{\pgfqpoint{1.202732in}{1.263555in}}%
\pgfpathlineto{\pgfqpoint{1.480509in}{1.263555in}}%
\pgfusepath{stroke}%
\end{pgfscope}%
\begin{pgfscope}%
\pgfsetbuttcap%
\pgfsetmiterjoin%
\definecolor{currentfill}{rgb}{0.960784,0.462745,0.000000}%
\pgfsetfillcolor{currentfill}%
\pgfsetlinewidth{1.003750pt}%
\definecolor{currentstroke}{rgb}{0.960784,0.462745,0.000000}%
\pgfsetstrokecolor{currentstroke}%
\pgfsetdash{}{0pt}%
\pgfsys@defobject{currentmarker}{\pgfqpoint{-0.055556in}{-0.055556in}}{\pgfqpoint{0.055556in}{0.055556in}}{%
\pgfpathmoveto{\pgfqpoint{-0.055556in}{-0.055556in}}%
\pgfpathlineto{\pgfqpoint{0.055556in}{-0.055556in}}%
\pgfpathlineto{\pgfqpoint{0.055556in}{0.055556in}}%
\pgfpathlineto{\pgfqpoint{-0.055556in}{0.055556in}}%
\pgfpathlineto{\pgfqpoint{-0.055556in}{-0.055556in}}%
\pgfpathclose%
\pgfusepath{stroke,fill}%
}%
\begin{pgfscope}%
\pgfsys@transformshift{1.202732in}{1.263555in}%
\pgfsys@useobject{currentmarker}{}%
\end{pgfscope}%
\end{pgfscope}%
\begin{pgfscope}%
\definecolor{textcolor}{rgb}{0.000000,0.000000,0.000000}%
\pgfsetstrokecolor{textcolor}%
\pgfsetfillcolor{textcolor}%
\pgftext[x=1.702732in,y=1.166333in,left,base]{\color{textcolor}{\rmfamily\fontsize{20.000000}{24.000000}\selectfont\catcode`\^=\active\def^{\ifmmode\sp\else\^{}\fi}\catcode`\%=\active\def
\end{pgfscope}%
\end{pgfpicture}%
\makeatother%
\endgroup%

%% file: images/thermal_ratio_per_modality/modality_results_RRA_15.pgf
\begingroup%
\makeatletter%
\begin{pgfpicture}%
\pgfpathrectangle{\pgfpointorigin}{\pgfqpoint{7.450000in}{5.450000in}}%
\pgfusepath{use as bounding box, clip}%
\begin{pgfscope}%
\pgfsetbuttcap%
\pgfsetmiterjoin%
\definecolor{currentfill}{rgb}{1.000000,1.000000,1.000000}%
\pgfsetfillcolor{currentfill}%
\pgfsetlinewidth{0.000000pt}%
\definecolor{currentstroke}{rgb}{1.000000,1.000000,1.000000}%
\pgfsetstrokecolor{currentstroke}%
\pgfsetdash{}{0pt}%
\pgfpathmoveto{\pgfqpoint{0.000000in}{0.000000in}}%
\pgfpathlineto{\pgfqpoint{7.450000in}{0.000000in}}%
\pgfpathlineto{\pgfqpoint{7.450000in}{5.450000in}}%
\pgfpathlineto{\pgfqpoint{0.000000in}{5.450000in}}%
\pgfpathlineto{\pgfqpoint{0.000000in}{0.000000in}}%
\pgfpathclose%
\pgfusepath{fill}%
\end{pgfscope}%
\begin{pgfscope}%
\pgfsetbuttcap%
\pgfsetmiterjoin%
\definecolor{currentfill}{rgb}{1.000000,1.000000,1.000000}%
\pgfsetfillcolor{currentfill}%
\pgfsetlinewidth{0.000000pt}%
\definecolor{currentstroke}{rgb}{0.000000,0.000000,0.000000}%
\pgfsetstrokecolor{currentstroke}%
\pgfsetstrokeopacity{0.000000}%
\pgfsetdash{}{0pt}%
\pgfpathmoveto{\pgfqpoint{0.674954in}{0.862305in}}%
\pgfpathlineto{\pgfqpoint{7.147414in}{0.862305in}}%
\pgfpathlineto{\pgfqpoint{7.147414in}{5.231386in}}%
\pgfpathlineto{\pgfqpoint{0.674954in}{5.231386in}}%
\pgfpathlineto{\pgfqpoint{0.674954in}{0.862305in}}%
\pgfpathclose%
\pgfusepath{fill}%
\end{pgfscope}%
\begin{pgfscope}%
\pgfpathrectangle{\pgfqpoint{0.674954in}{0.862305in}}{\pgfqpoint{6.472460in}{4.369081in}}%
\pgfusepath{clip}%
\pgfsetbuttcap%
\pgfsetroundjoin%
\pgfsetlinewidth{2.007500pt}%
\definecolor{currentstroke}{rgb}{0.501961,0.501961,0.501961}%
\pgfsetstrokecolor{currentstroke}%
\pgfsetstrokeopacity{0.300000}%
\pgfsetdash{{7.400000pt}{3.200000pt}}{0.000000pt}%
\pgfpathmoveto{\pgfqpoint{0.674954in}{0.862305in}}%
\pgfpathlineto{\pgfqpoint{0.674954in}{5.231386in}}%
\pgfusepath{stroke}%
\end{pgfscope}%
\begin{pgfscope}%
\pgfsetbuttcap%
\pgfsetroundjoin%
\definecolor{currentfill}{rgb}{0.000000,0.000000,0.000000}%
\pgfsetfillcolor{currentfill}%
\pgfsetlinewidth{0.803000pt}%
\definecolor{currentstroke}{rgb}{0.000000,0.000000,0.000000}%
\pgfsetstrokecolor{currentstroke}%
\pgfsetdash{}{0pt}%
\pgfsys@defobject{currentmarker}{\pgfqpoint{0.000000in}{-0.048611in}}{\pgfqpoint{0.000000in}{0.000000in}}{%
\pgfpathmoveto{\pgfqpoint{0.000000in}{0.000000in}}%
\pgfpathlineto{\pgfqpoint{0.000000in}{-0.048611in}}%
\pgfusepath{stroke,fill}%
}%
\begin{pgfscope}%
\pgfsys@transformshift{0.674954in}{0.862305in}%
\pgfsys@useobject{currentmarker}{}%
\end{pgfscope}%
\end{pgfscope}%
\begin{pgfscope}%
\definecolor{textcolor}{rgb}{0.000000,0.000000,0.000000}%
\pgfsetstrokecolor{textcolor}%
\pgfsetfillcolor{textcolor}%
\pgftext[x=0.674954in,y=0.765082in,,top]{\color{textcolor}{\rmfamily\fontsize{25.000000}{30.000000}\selectfont\catcode`\^=\active\def^{\ifmmode\sp\else\^{}\fi}\catcode`\%=\active\def
\end{pgfscope}%
\begin{pgfscope}%
\pgfpathrectangle{\pgfqpoint{0.674954in}{0.862305in}}{\pgfqpoint{6.472460in}{4.369081in}}%
\pgfusepath{clip}%
\pgfsetbuttcap%
\pgfsetroundjoin%
\pgfsetlinewidth{2.007500pt}%
\definecolor{currentstroke}{rgb}{0.501961,0.501961,0.501961}%
\pgfsetstrokecolor{currentstroke}%
\pgfsetstrokeopacity{0.300000}%
\pgfsetdash{{7.400000pt}{3.200000pt}}{0.000000pt}%
\pgfpathmoveto{\pgfqpoint{1.969446in}{0.862305in}}%
\pgfpathlineto{\pgfqpoint{1.969446in}{5.231386in}}%
\pgfusepath{stroke}%
\end{pgfscope}%
\begin{pgfscope}%
\pgfsetbuttcap%
\pgfsetroundjoin%
\definecolor{currentfill}{rgb}{0.000000,0.000000,0.000000}%
\pgfsetfillcolor{currentfill}%
\pgfsetlinewidth{0.803000pt}%
\definecolor{currentstroke}{rgb}{0.000000,0.000000,0.000000}%
\pgfsetstrokecolor{currentstroke}%
\pgfsetdash{}{0pt}%
\pgfsys@defobject{currentmarker}{\pgfqpoint{0.000000in}{-0.048611in}}{\pgfqpoint{0.000000in}{0.000000in}}{%
\pgfpathmoveto{\pgfqpoint{0.000000in}{0.000000in}}%
\pgfpathlineto{\pgfqpoint{0.000000in}{-0.048611in}}%
\pgfusepath{stroke,fill}%
}%
\begin{pgfscope}%
\pgfsys@transformshift{1.969446in}{0.862305in}%
\pgfsys@useobject{currentmarker}{}%
\end{pgfscope}%
\end{pgfscope}%
\begin{pgfscope}%
\definecolor{textcolor}{rgb}{0.000000,0.000000,0.000000}%
\pgfsetstrokecolor{textcolor}%
\pgfsetfillcolor{textcolor}%
\pgftext[x=1.969446in,y=0.765082in,,top]{\color{textcolor}{\rmfamily\fontsize{25.000000}{30.000000}\selectfont\catcode`\^=\active\def^{\ifmmode\sp\else\^{}\fi}\catcode`\%=\active\def
\end{pgfscope}%
\begin{pgfscope}%
\pgfpathrectangle{\pgfqpoint{0.674954in}{0.862305in}}{\pgfqpoint{6.472460in}{4.369081in}}%
\pgfusepath{clip}%
\pgfsetbuttcap%
\pgfsetroundjoin%
\pgfsetlinewidth{2.007500pt}%
\definecolor{currentstroke}{rgb}{0.501961,0.501961,0.501961}%
\pgfsetstrokecolor{currentstroke}%
\pgfsetstrokeopacity{0.300000}%
\pgfsetdash{{7.400000pt}{3.200000pt}}{0.000000pt}%
\pgfpathmoveto{\pgfqpoint{3.263938in}{0.862305in}}%
\pgfpathlineto{\pgfqpoint{3.263938in}{5.231386in}}%
\pgfusepath{stroke}%
\end{pgfscope}%
\begin{pgfscope}%
\pgfsetbuttcap%
\pgfsetroundjoin%
\definecolor{currentfill}{rgb}{0.000000,0.000000,0.000000}%
\pgfsetfillcolor{currentfill}%
\pgfsetlinewidth{0.803000pt}%
\definecolor{currentstroke}{rgb}{0.000000,0.000000,0.000000}%
\pgfsetstrokecolor{currentstroke}%
\pgfsetdash{}{0pt}%
\pgfsys@defobject{currentmarker}{\pgfqpoint{0.000000in}{-0.048611in}}{\pgfqpoint{0.000000in}{0.000000in}}{%
\pgfpathmoveto{\pgfqpoint{0.000000in}{0.000000in}}%
\pgfpathlineto{\pgfqpoint{0.000000in}{-0.048611in}}%
\pgfusepath{stroke,fill}%
}%
\begin{pgfscope}%
\pgfsys@transformshift{3.263938in}{0.862305in}%
\pgfsys@useobject{currentmarker}{}%
\end{pgfscope}%
\end{pgfscope}%
\begin{pgfscope}%
\definecolor{textcolor}{rgb}{0.000000,0.000000,0.000000}%
\pgfsetstrokecolor{textcolor}%
\pgfsetfillcolor{textcolor}%
\pgftext[x=3.263938in,y=0.765082in,,top]{\color{textcolor}{\rmfamily\fontsize{25.000000}{30.000000}\selectfont\catcode`\^=\active\def^{\ifmmode\sp\else\^{}\fi}\catcode`\%=\active\def
\end{pgfscope}%
\begin{pgfscope}%
\pgfpathrectangle{\pgfqpoint{0.674954in}{0.862305in}}{\pgfqpoint{6.472460in}{4.369081in}}%
\pgfusepath{clip}%
\pgfsetbuttcap%
\pgfsetroundjoin%
\pgfsetlinewidth{2.007500pt}%
\definecolor{currentstroke}{rgb}{0.501961,0.501961,0.501961}%
\pgfsetstrokecolor{currentstroke}%
\pgfsetstrokeopacity{0.300000}%
\pgfsetdash{{7.400000pt}{3.200000pt}}{0.000000pt}%
\pgfpathmoveto{\pgfqpoint{4.558430in}{0.862305in}}%
\pgfpathlineto{\pgfqpoint{4.558430in}{5.231386in}}%
\pgfusepath{stroke}%
\end{pgfscope}%
\begin{pgfscope}%
\pgfsetbuttcap%
\pgfsetroundjoin%
\definecolor{currentfill}{rgb}{0.000000,0.000000,0.000000}%
\pgfsetfillcolor{currentfill}%
\pgfsetlinewidth{0.803000pt}%
\definecolor{currentstroke}{rgb}{0.000000,0.000000,0.000000}%
\pgfsetstrokecolor{currentstroke}%
\pgfsetdash{}{0pt}%
\pgfsys@defobject{currentmarker}{\pgfqpoint{0.000000in}{-0.048611in}}{\pgfqpoint{0.000000in}{0.000000in}}{%
\pgfpathmoveto{\pgfqpoint{0.000000in}{0.000000in}}%
\pgfpathlineto{\pgfqpoint{0.000000in}{-0.048611in}}%
\pgfusepath{stroke,fill}%
}%
\begin{pgfscope}%
\pgfsys@transformshift{4.558430in}{0.862305in}%
\pgfsys@useobject{currentmarker}{}%
\end{pgfscope}%
\end{pgfscope}%
\begin{pgfscope}%
\definecolor{textcolor}{rgb}{0.000000,0.000000,0.000000}%
\pgfsetstrokecolor{textcolor}%
\pgfsetfillcolor{textcolor}%
\pgftext[x=4.558430in,y=0.765082in,,top]{\color{textcolor}{\rmfamily\fontsize{25.000000}{30.000000}\selectfont\catcode`\^=\active\def^{\ifmmode\sp\else\^{}\fi}\catcode`\%=\active\def
\end{pgfscope}%
\begin{pgfscope}%
\pgfpathrectangle{\pgfqpoint{0.674954in}{0.862305in}}{\pgfqpoint{6.472460in}{4.369081in}}%
\pgfusepath{clip}%
\pgfsetbuttcap%
\pgfsetroundjoin%
\pgfsetlinewidth{2.007500pt}%
\definecolor{currentstroke}{rgb}{0.501961,0.501961,0.501961}%
\pgfsetstrokecolor{currentstroke}%
\pgfsetstrokeopacity{0.300000}%
\pgfsetdash{{7.400000pt}{3.200000pt}}{0.000000pt}%
\pgfpathmoveto{\pgfqpoint{5.852922in}{0.862305in}}%
\pgfpathlineto{\pgfqpoint{5.852922in}{5.231386in}}%
\pgfusepath{stroke}%
\end{pgfscope}%
\begin{pgfscope}%
\pgfsetbuttcap%
\pgfsetroundjoin%
\definecolor{currentfill}{rgb}{0.000000,0.000000,0.000000}%
\pgfsetfillcolor{currentfill}%
\pgfsetlinewidth{0.803000pt}%
\definecolor{currentstroke}{rgb}{0.000000,0.000000,0.000000}%
\pgfsetstrokecolor{currentstroke}%
\pgfsetdash{}{0pt}%
\pgfsys@defobject{currentmarker}{\pgfqpoint{0.000000in}{-0.048611in}}{\pgfqpoint{0.000000in}{0.000000in}}{%
\pgfpathmoveto{\pgfqpoint{0.000000in}{0.000000in}}%
\pgfpathlineto{\pgfqpoint{0.000000in}{-0.048611in}}%
\pgfusepath{stroke,fill}%
}%
\begin{pgfscope}%
\pgfsys@transformshift{5.852922in}{0.862305in}%
\pgfsys@useobject{currentmarker}{}%
\end{pgfscope}%
\end{pgfscope}%
\begin{pgfscope}%
\definecolor{textcolor}{rgb}{0.000000,0.000000,0.000000}%
\pgfsetstrokecolor{textcolor}%
\pgfsetfillcolor{textcolor}%
\pgftext[x=5.852922in,y=0.765082in,,top]{\color{textcolor}{\rmfamily\fontsize{25.000000}{30.000000}\selectfont\catcode`\^=\active\def^{\ifmmode\sp\else\^{}\fi}\catcode`\%=\active\def
\end{pgfscope}%
\begin{pgfscope}%
\pgfpathrectangle{\pgfqpoint{0.674954in}{0.862305in}}{\pgfqpoint{6.472460in}{4.369081in}}%
\pgfusepath{clip}%
\pgfsetbuttcap%
\pgfsetroundjoin%
\pgfsetlinewidth{2.007500pt}%
\definecolor{currentstroke}{rgb}{0.501961,0.501961,0.501961}%
\pgfsetstrokecolor{currentstroke}%
\pgfsetstrokeopacity{0.300000}%
\pgfsetdash{{7.400000pt}{3.200000pt}}{0.000000pt}%
\pgfpathmoveto{\pgfqpoint{7.147414in}{0.862305in}}%
\pgfpathlineto{\pgfqpoint{7.147414in}{5.231386in}}%
\pgfusepath{stroke}%
\end{pgfscope}%
\begin{pgfscope}%
\pgfsetbuttcap%
\pgfsetroundjoin%
\definecolor{currentfill}{rgb}{0.000000,0.000000,0.000000}%
\pgfsetfillcolor{currentfill}%
\pgfsetlinewidth{0.803000pt}%
\definecolor{currentstroke}{rgb}{0.000000,0.000000,0.000000}%
\pgfsetstrokecolor{currentstroke}%
\pgfsetdash{}{0pt}%
\pgfsys@defobject{currentmarker}{\pgfqpoint{0.000000in}{-0.048611in}}{\pgfqpoint{0.000000in}{0.000000in}}{%
\pgfpathmoveto{\pgfqpoint{0.000000in}{0.000000in}}%
\pgfpathlineto{\pgfqpoint{0.000000in}{-0.048611in}}%
\pgfusepath{stroke,fill}%
}%
\begin{pgfscope}%
\pgfsys@transformshift{7.147414in}{0.862305in}%
\pgfsys@useobject{currentmarker}{}%
\end{pgfscope}%
\end{pgfscope}%
\begin{pgfscope}%
\definecolor{textcolor}{rgb}{0.000000,0.000000,0.000000}%
\pgfsetstrokecolor{textcolor}%
\pgfsetfillcolor{textcolor}%
\pgftext[x=7.147414in,y=0.765082in,,top]{\color{textcolor}{\rmfamily\fontsize{25.000000}{30.000000}\selectfont\catcode`\^=\active\def^{\ifmmode\sp\else\^{}\fi}\catcode`\%=\active\def
\end{pgfscope}%
\begin{pgfscope}%
\definecolor{textcolor}{rgb}{0.000000,0.000000,0.000000}%
\pgfsetstrokecolor{textcolor}%
\pgfsetfillcolor{textcolor}%
\pgftext[x=3.911184in,y=0.404763in,,top]{\color{textcolor}{\rmfamily\fontsize{25.000000}{30.000000}\selectfont\catcode`\^=\active\def^{\ifmmode\sp\else\^{}\fi}\catcode`\%=\active\def
\end{pgfscope}%
\begin{pgfscope}%
\pgfpathrectangle{\pgfqpoint{0.674954in}{0.862305in}}{\pgfqpoint{6.472460in}{4.369081in}}%
\pgfusepath{clip}%
\pgfsetbuttcap%
\pgfsetroundjoin%
\pgfsetlinewidth{2.007500pt}%
\definecolor{currentstroke}{rgb}{0.501961,0.501961,0.501961}%
\pgfsetstrokecolor{currentstroke}%
\pgfsetstrokeopacity{0.300000}%
\pgfsetdash{{7.400000pt}{3.200000pt}}{0.000000pt}%
\pgfpathmoveto{\pgfqpoint{0.674954in}{1.954575in}}%
\pgfpathlineto{\pgfqpoint{7.147414in}{1.954575in}}%
\pgfusepath{stroke}%
\end{pgfscope}%
\begin{pgfscope}%
\pgfsetbuttcap%
\pgfsetroundjoin%
\definecolor{currentfill}{rgb}{0.000000,0.000000,0.000000}%
\pgfsetfillcolor{currentfill}%
\pgfsetlinewidth{0.803000pt}%
\definecolor{currentstroke}{rgb}{0.000000,0.000000,0.000000}%
\pgfsetstrokecolor{currentstroke}%
\pgfsetdash{}{0pt}%
\pgfsys@defobject{currentmarker}{\pgfqpoint{-0.048611in}{0.000000in}}{\pgfqpoint{-0.000000in}{0.000000in}}{%
\pgfpathmoveto{\pgfqpoint{-0.000000in}{0.000000in}}%
\pgfpathlineto{\pgfqpoint{-0.048611in}{0.000000in}}%
\pgfusepath{stroke,fill}%
}%
\begin{pgfscope}%
\pgfsys@transformshift{0.674954in}{1.954575in}%
\pgfsys@useobject{currentmarker}{}%
\end{pgfscope}%
\end{pgfscope}%
\begin{pgfscope}%
\definecolor{textcolor}{rgb}{0.000000,0.000000,0.000000}%
\pgfsetstrokecolor{textcolor}%
\pgfsetfillcolor{textcolor}%
\pgftext[x=0.259244in, y=1.835961in, left, base]{\color{textcolor}{\rmfamily\fontsize{25.000000}{30.000000}\selectfont\catcode`\^=\active\def^{\ifmmode\sp\else\^{}\fi}\catcode`\%=\active\def
\end{pgfscope}%
\begin{pgfscope}%
\pgfpathrectangle{\pgfqpoint{0.674954in}{0.862305in}}{\pgfqpoint{6.472460in}{4.369081in}}%
\pgfusepath{clip}%
\pgfsetbuttcap%
\pgfsetroundjoin%
\pgfsetlinewidth{2.007500pt}%
\definecolor{currentstroke}{rgb}{0.501961,0.501961,0.501961}%
\pgfsetstrokecolor{currentstroke}%
\pgfsetstrokeopacity{0.300000}%
\pgfsetdash{{7.400000pt}{3.200000pt}}{0.000000pt}%
\pgfpathmoveto{\pgfqpoint{0.674954in}{3.046845in}}%
\pgfpathlineto{\pgfqpoint{7.147414in}{3.046845in}}%
\pgfusepath{stroke}%
\end{pgfscope}%
\begin{pgfscope}%
\pgfsetbuttcap%
\pgfsetroundjoin%
\definecolor{currentfill}{rgb}{0.000000,0.000000,0.000000}%
\pgfsetfillcolor{currentfill}%
\pgfsetlinewidth{0.803000pt}%
\definecolor{currentstroke}{rgb}{0.000000,0.000000,0.000000}%
\pgfsetstrokecolor{currentstroke}%
\pgfsetdash{}{0pt}%
\pgfsys@defobject{currentmarker}{\pgfqpoint{-0.048611in}{0.000000in}}{\pgfqpoint{-0.000000in}{0.000000in}}{%
\pgfpathmoveto{\pgfqpoint{-0.000000in}{0.000000in}}%
\pgfpathlineto{\pgfqpoint{-0.048611in}{0.000000in}}%
\pgfusepath{stroke,fill}%
}%
\begin{pgfscope}%
\pgfsys@transformshift{0.674954in}{3.046845in}%
\pgfsys@useobject{currentmarker}{}%
\end{pgfscope}%
\end{pgfscope}%
\begin{pgfscope}%
\definecolor{textcolor}{rgb}{0.000000,0.000000,0.000000}%
\pgfsetstrokecolor{textcolor}%
\pgfsetfillcolor{textcolor}%
\pgftext[x=0.259244in, y=2.928231in, left, base]{\color{textcolor}{\rmfamily\fontsize{25.000000}{30.000000}\selectfont\catcode`\^=\active\def^{\ifmmode\sp\else\^{}\fi}\catcode`\%=\active\def
\end{pgfscope}%
\begin{pgfscope}%
\pgfpathrectangle{\pgfqpoint{0.674954in}{0.862305in}}{\pgfqpoint{6.472460in}{4.369081in}}%
\pgfusepath{clip}%
\pgfsetbuttcap%
\pgfsetroundjoin%
\pgfsetlinewidth{2.007500pt}%
\definecolor{currentstroke}{rgb}{0.501961,0.501961,0.501961}%
\pgfsetstrokecolor{currentstroke}%
\pgfsetstrokeopacity{0.300000}%
\pgfsetdash{{7.400000pt}{3.200000pt}}{0.000000pt}%
\pgfpathmoveto{\pgfqpoint{0.674954in}{4.139116in}}%
\pgfpathlineto{\pgfqpoint{7.147414in}{4.139116in}}%
\pgfusepath{stroke}%
\end{pgfscope}%
\begin{pgfscope}%
\pgfsetbuttcap%
\pgfsetroundjoin%
\definecolor{currentfill}{rgb}{0.000000,0.000000,0.000000}%
\pgfsetfillcolor{currentfill}%
\pgfsetlinewidth{0.803000pt}%
\definecolor{currentstroke}{rgb}{0.000000,0.000000,0.000000}%
\pgfsetstrokecolor{currentstroke}%
\pgfsetdash{}{0pt}%
\pgfsys@defobject{currentmarker}{\pgfqpoint{-0.048611in}{0.000000in}}{\pgfqpoint{-0.000000in}{0.000000in}}{%
\pgfpathmoveto{\pgfqpoint{-0.000000in}{0.000000in}}%
\pgfpathlineto{\pgfqpoint{-0.048611in}{0.000000in}}%
\pgfusepath{stroke,fill}%
}%
\begin{pgfscope}%
\pgfsys@transformshift{0.674954in}{4.139116in}%
\pgfsys@useobject{currentmarker}{}%
\end{pgfscope}%
\end{pgfscope}%
\begin{pgfscope}%
\definecolor{textcolor}{rgb}{0.000000,0.000000,0.000000}%
\pgfsetstrokecolor{textcolor}%
\pgfsetfillcolor{textcolor}%
\pgftext[x=0.259244in, y=4.020501in, left, base]{\color{textcolor}{\rmfamily\fontsize{25.000000}{30.000000}\selectfont\catcode`\^=\active\def^{\ifmmode\sp\else\^{}\fi}\catcode`\%=\active\def
\end{pgfscope}%
\begin{pgfscope}%
\pgfpathrectangle{\pgfqpoint{0.674954in}{0.862305in}}{\pgfqpoint{6.472460in}{4.369081in}}%
\pgfusepath{clip}%
\pgfsetbuttcap%
\pgfsetroundjoin%
\pgfsetlinewidth{2.007500pt}%
\definecolor{currentstroke}{rgb}{0.501961,0.501961,0.501961}%
\pgfsetstrokecolor{currentstroke}%
\pgfsetstrokeopacity{0.300000}%
\pgfsetdash{{7.400000pt}{3.200000pt}}{0.000000pt}%
\pgfpathmoveto{\pgfqpoint{0.674954in}{5.231386in}}%
\pgfpathlineto{\pgfqpoint{7.147414in}{5.231386in}}%
\pgfusepath{stroke}%
\end{pgfscope}%
\begin{pgfscope}%
\pgfsetbuttcap%
\pgfsetroundjoin%
\definecolor{currentfill}{rgb}{0.000000,0.000000,0.000000}%
\pgfsetfillcolor{currentfill}%
\pgfsetlinewidth{0.803000pt}%
\definecolor{currentstroke}{rgb}{0.000000,0.000000,0.000000}%
\pgfsetstrokecolor{currentstroke}%
\pgfsetdash{}{0pt}%
\pgfsys@defobject{currentmarker}{\pgfqpoint{-0.048611in}{0.000000in}}{\pgfqpoint{-0.000000in}{0.000000in}}{%
\pgfpathmoveto{\pgfqpoint{-0.000000in}{0.000000in}}%
\pgfpathlineto{\pgfqpoint{-0.048611in}{0.000000in}}%
\pgfusepath{stroke,fill}%
}%
\begin{pgfscope}%
\pgfsys@transformshift{0.674954in}{5.231386in}%
\pgfsys@useobject{currentmarker}{}%
\end{pgfscope}%
\end{pgfscope}%
\begin{pgfscope}%
\definecolor{textcolor}{rgb}{0.000000,0.000000,0.000000}%
\pgfsetstrokecolor{textcolor}%
\pgfsetfillcolor{textcolor}%
\pgftext[x=0.100000in, y=5.112772in, left, base]{\color{textcolor}{\rmfamily\fontsize{25.000000}{30.000000}\selectfont\catcode`\^=\active\def^{\ifmmode\sp\else\^{}\fi}\catcode`\%=\active\def
\end{pgfscope}%
\begin{pgfscope}%
\pgfpathrectangle{\pgfqpoint{0.674954in}{0.862305in}}{\pgfqpoint{6.472460in}{4.369081in}}%
\pgfusepath{clip}%
\pgfsetrectcap%
\pgfsetroundjoin%
\pgfsetlinewidth{2.509375pt}%
\definecolor{currentstroke}{rgb}{0.050980,0.415686,0.509804}%
\pgfsetstrokecolor{currentstroke}%
\pgfsetdash{}{0pt}%
\pgfpathmoveto{\pgfqpoint{0.674954in}{5.231386in}}%
\pgfpathlineto{\pgfqpoint{0.998577in}{5.231386in}}%
\pgfpathlineto{\pgfqpoint{2.293069in}{5.161744in}}%
\pgfpathlineto{\pgfqpoint{3.911184in}{5.147494in}}%
\pgfpathlineto{\pgfqpoint{5.529299in}{4.896317in}}%
\pgfpathlineto{\pgfqpoint{6.823791in}{4.221951in}}%
\pgfusepath{stroke}%
\end{pgfscope}%
\begin{pgfscope}%
\pgfpathrectangle{\pgfqpoint{0.674954in}{0.862305in}}{\pgfqpoint{6.472460in}{4.369081in}}%
\pgfusepath{clip}%
\pgfsetbuttcap%
\pgfsetroundjoin%
\definecolor{currentfill}{rgb}{0.050980,0.415686,0.509804}%
\pgfsetfillcolor{currentfill}%
\pgfsetlinewidth{1.003750pt}%
\definecolor{currentstroke}{rgb}{0.050980,0.415686,0.509804}%
\pgfsetstrokecolor{currentstroke}%
\pgfsetdash{}{0pt}%
\pgfsys@defobject{currentmarker}{\pgfqpoint{-0.055556in}{-0.055556in}}{\pgfqpoint{0.055556in}{0.055556in}}{%
\pgfpathmoveto{\pgfqpoint{0.000000in}{-0.055556in}}%
\pgfpathcurveto{\pgfqpoint{0.014734in}{-0.055556in}}{\pgfqpoint{0.028866in}{-0.049702in}}{\pgfqpoint{0.039284in}{-0.039284in}}%
\pgfpathcurveto{\pgfqpoint{0.049702in}{-0.028866in}}{\pgfqpoint{0.055556in}{-0.014734in}}{\pgfqpoint{0.055556in}{0.000000in}}%
\pgfpathcurveto{\pgfqpoint{0.055556in}{0.014734in}}{\pgfqpoint{0.049702in}{0.028866in}}{\pgfqpoint{0.039284in}{0.039284in}}%
\pgfpathcurveto{\pgfqpoint{0.028866in}{0.049702in}}{\pgfqpoint{0.014734in}{0.055556in}}{\pgfqpoint{0.000000in}{0.055556in}}%
\pgfpathcurveto{\pgfqpoint{-0.014734in}{0.055556in}}{\pgfqpoint{-0.028866in}{0.049702in}}{\pgfqpoint{-0.039284in}{0.039284in}}%
\pgfpathcurveto{\pgfqpoint{-0.049702in}{0.028866in}}{\pgfqpoint{-0.055556in}{0.014734in}}{\pgfqpoint{-0.055556in}{0.000000in}}%
\pgfpathcurveto{\pgfqpoint{-0.055556in}{-0.014734in}}{\pgfqpoint{-0.049702in}{-0.028866in}}{\pgfqpoint{-0.039284in}{-0.039284in}}%
\pgfpathcurveto{\pgfqpoint{-0.028866in}{-0.049702in}}{\pgfqpoint{-0.014734in}{-0.055556in}}{\pgfqpoint{0.000000in}{-0.055556in}}%
\pgfpathlineto{\pgfqpoint{0.000000in}{-0.055556in}}%
\pgfpathclose%
\pgfusepath{stroke,fill}%
}%
\begin{pgfscope}%
\pgfsys@transformshift{0.674954in}{5.231386in}%
\pgfsys@useobject{currentmarker}{}%
\end{pgfscope}%
\begin{pgfscope}%
\pgfsys@transformshift{0.998577in}{5.231386in}%
\pgfsys@useobject{currentmarker}{}%
\end{pgfscope}%
\begin{pgfscope}%
\pgfsys@transformshift{2.293069in}{5.161744in}%
\pgfsys@useobject{currentmarker}{}%
\end{pgfscope}%
\begin{pgfscope}%
\pgfsys@transformshift{3.911184in}{5.147494in}%
\pgfsys@useobject{currentmarker}{}%
\end{pgfscope}%
\begin{pgfscope}%
\pgfsys@transformshift{5.529299in}{4.896317in}%
\pgfsys@useobject{currentmarker}{}%
\end{pgfscope}%
\begin{pgfscope}%
\pgfsys@transformshift{6.823791in}{4.221951in}%
\pgfsys@useobject{currentmarker}{}%
\end{pgfscope}%
\end{pgfscope}%
\begin{pgfscope}%
\pgfpathrectangle{\pgfqpoint{0.674954in}{0.862305in}}{\pgfqpoint{6.472460in}{4.369081in}}%
\pgfusepath{clip}%
\pgfsetbuttcap%
\pgfsetroundjoin%
\pgfsetlinewidth{2.509375pt}%
\definecolor{currentstroke}{rgb}{0.960784,0.462745,0.000000}%
\pgfsetstrokecolor{currentstroke}%
\pgfsetdash{{9.250000pt}{4.000000pt}}{0.000000pt}%
\pgfpathmoveto{\pgfqpoint{0.998577in}{4.601897in}}%
\pgfpathlineto{\pgfqpoint{2.293069in}{4.755789in}}%
\pgfpathlineto{\pgfqpoint{3.911184in}{4.641360in}}%
\pgfpathlineto{\pgfqpoint{5.529299in}{4.548419in}}%
\pgfpathlineto{\pgfqpoint{6.823791in}{4.569468in}}%
\pgfpathlineto{\pgfqpoint{7.147414in}{4.709847in}}%
\pgfusepath{stroke}%
\end{pgfscope}%
\begin{pgfscope}%
\pgfpathrectangle{\pgfqpoint{0.674954in}{0.862305in}}{\pgfqpoint{6.472460in}{4.369081in}}%
\pgfusepath{clip}%
\pgfsetbuttcap%
\pgfsetmiterjoin%
\definecolor{currentfill}{rgb}{0.960784,0.462745,0.000000}%
\pgfsetfillcolor{currentfill}%
\pgfsetlinewidth{1.003750pt}%
\definecolor{currentstroke}{rgb}{0.960784,0.462745,0.000000}%
\pgfsetstrokecolor{currentstroke}%
\pgfsetdash{}{0pt}%
\pgfsys@defobject{currentmarker}{\pgfqpoint{-0.055556in}{-0.055556in}}{\pgfqpoint{0.055556in}{0.055556in}}{%
\pgfpathmoveto{\pgfqpoint{-0.055556in}{-0.055556in}}%
\pgfpathlineto{\pgfqpoint{0.055556in}{-0.055556in}}%
\pgfpathlineto{\pgfqpoint{0.055556in}{0.055556in}}%
\pgfpathlineto{\pgfqpoint{-0.055556in}{0.055556in}}%
\pgfpathlineto{\pgfqpoint{-0.055556in}{-0.055556in}}%
\pgfpathclose%
\pgfusepath{stroke,fill}%
}%
\begin{pgfscope}%
\pgfsys@transformshift{0.998577in}{4.601897in}%
\pgfsys@useobject{currentmarker}{}%
\end{pgfscope}%
\begin{pgfscope}%
\pgfsys@transformshift{2.293069in}{4.755789in}%
\pgfsys@useobject{currentmarker}{}%
\end{pgfscope}%
\begin{pgfscope}%
\pgfsys@transformshift{3.911184in}{4.641360in}%
\pgfsys@useobject{currentmarker}{}%
\end{pgfscope}%
\begin{pgfscope}%
\pgfsys@transformshift{5.529299in}{4.548419in}%
\pgfsys@useobject{currentmarker}{}%
\end{pgfscope}%
\begin{pgfscope}%
\pgfsys@transformshift{6.823791in}{4.569468in}%
\pgfsys@useobject{currentmarker}{}%
\end{pgfscope}%
\begin{pgfscope}%
\pgfsys@transformshift{7.147414in}{4.709847in}%
\pgfsys@useobject{currentmarker}{}%
\end{pgfscope}%
\end{pgfscope}%
\begin{pgfscope}%
\pgfsetrectcap%
\pgfsetmiterjoin%
\pgfsetlinewidth{2.007500pt}%
\definecolor{currentstroke}{rgb}{0.000000,0.000000,0.000000}%
\pgfsetstrokecolor{currentstroke}%
\pgfsetdash{}{0pt}%
\pgfpathmoveto{\pgfqpoint{0.674954in}{0.862305in}}%
\pgfpathlineto{\pgfqpoint{0.674954in}{5.231386in}}%
\pgfusepath{stroke}%
\end{pgfscope}%
\begin{pgfscope}%
\pgfsetrectcap%
\pgfsetmiterjoin%
\pgfsetlinewidth{2.007500pt}%
\definecolor{currentstroke}{rgb}{0.000000,0.000000,0.000000}%
\pgfsetstrokecolor{currentstroke}%
\pgfsetdash{}{0pt}%
\pgfpathmoveto{\pgfqpoint{0.674954in}{0.862305in}}%
\pgfpathlineto{\pgfqpoint{7.147414in}{0.862305in}}%
\pgfusepath{stroke}%
\end{pgfscope}%
\begin{pgfscope}%
\pgfsetbuttcap%
\pgfsetmiterjoin%
\definecolor{currentfill}{rgb}{1.000000,1.000000,1.000000}%
\pgfsetfillcolor{currentfill}%
\pgfsetlinewidth{1.003750pt}%
\definecolor{currentstroke}{rgb}{0.800000,0.800000,0.800000}%
\pgfsetstrokecolor{currentstroke}%
\pgfsetdash{}{0pt}%
\pgfpathmoveto{\pgfqpoint{0.869398in}{1.001194in}}%
\pgfpathlineto{\pgfqpoint{2.708109in}{1.001194in}}%
\pgfpathquadraticcurveto{\pgfqpoint{2.763664in}{1.001194in}}{\pgfqpoint{2.763664in}{1.056749in}}%
\pgfpathlineto{\pgfqpoint{2.763664in}{1.803694in}}%
\pgfpathquadraticcurveto{\pgfqpoint{2.763664in}{1.859250in}}{\pgfqpoint{2.708109in}{1.859250in}}%
\pgfpathlineto{\pgfqpoint{0.869398in}{1.859250in}}%
\pgfpathquadraticcurveto{\pgfqpoint{0.813843in}{1.859250in}}{\pgfqpoint{0.813843in}{1.803694in}}%
\pgfpathlineto{\pgfqpoint{0.813843in}{1.056749in}}%
\pgfpathquadraticcurveto{\pgfqpoint{0.813843in}{1.001194in}}{\pgfqpoint{0.869398in}{1.001194in}}%
\pgfpathlineto{\pgfqpoint{0.869398in}{1.001194in}}%
\pgfpathclose%
\pgfusepath{stroke,fill}%
\end{pgfscope}%
\begin{pgfscope}%
\pgfsetrectcap%
\pgfsetroundjoin%
\pgfsetlinewidth{2.509375pt}%
\definecolor{currentstroke}{rgb}{0.050980,0.415686,0.509804}%
\pgfsetstrokecolor{currentstroke}%
\pgfsetdash{}{0pt}%
\pgfpathmoveto{\pgfqpoint{0.924954in}{1.650917in}}%
\pgfpathlineto{\pgfqpoint{1.202732in}{1.650917in}}%
\pgfpathlineto{\pgfqpoint{1.480509in}{1.650917in}}%
\pgfusepath{stroke}%
\end{pgfscope}%
\begin{pgfscope}%
\pgfsetbuttcap%
\pgfsetroundjoin%
\definecolor{currentfill}{rgb}{0.050980,0.415686,0.509804}%
\pgfsetfillcolor{currentfill}%
\pgfsetlinewidth{1.003750pt}%
\definecolor{currentstroke}{rgb}{0.050980,0.415686,0.509804}%
\pgfsetstrokecolor{currentstroke}%
\pgfsetdash{}{0pt}%
\pgfsys@defobject{currentmarker}{\pgfqpoint{-0.055556in}{-0.055556in}}{\pgfqpoint{0.055556in}{0.055556in}}{%
\pgfpathmoveto{\pgfqpoint{0.000000in}{-0.055556in}}%
\pgfpathcurveto{\pgfqpoint{0.014734in}{-0.055556in}}{\pgfqpoint{0.028866in}{-0.049702in}}{\pgfqpoint{0.039284in}{-0.039284in}}%
\pgfpathcurveto{\pgfqpoint{0.049702in}{-0.028866in}}{\pgfqpoint{0.055556in}{-0.014734in}}{\pgfqpoint{0.055556in}{0.000000in}}%
\pgfpathcurveto{\pgfqpoint{0.055556in}{0.014734in}}{\pgfqpoint{0.049702in}{0.028866in}}{\pgfqpoint{0.039284in}{0.039284in}}%
\pgfpathcurveto{\pgfqpoint{0.028866in}{0.049702in}}{\pgfqpoint{0.014734in}{0.055556in}}{\pgfqpoint{0.000000in}{0.055556in}}%
\pgfpathcurveto{\pgfqpoint{-0.014734in}{0.055556in}}{\pgfqpoint{-0.028866in}{0.049702in}}{\pgfqpoint{-0.039284in}{0.039284in}}%
\pgfpathcurveto{\pgfqpoint{-0.049702in}{0.028866in}}{\pgfqpoint{-0.055556in}{0.014734in}}{\pgfqpoint{-0.055556in}{0.000000in}}%
\pgfpathcurveto{\pgfqpoint{-0.055556in}{-0.014734in}}{\pgfqpoint{-0.049702in}{-0.028866in}}{\pgfqpoint{-0.039284in}{-0.039284in}}%
\pgfpathcurveto{\pgfqpoint{-0.028866in}{-0.049702in}}{\pgfqpoint{-0.014734in}{-0.055556in}}{\pgfqpoint{0.000000in}{-0.055556in}}%
\pgfpathlineto{\pgfqpoint{0.000000in}{-0.055556in}}%
\pgfpathclose%
\pgfusepath{stroke,fill}%
}%
\begin{pgfscope}%
\pgfsys@transformshift{1.202732in}{1.650917in}%
\pgfsys@useobject{currentmarker}{}%
\end{pgfscope}%
\end{pgfscope}%
\begin{pgfscope}%
\definecolor{textcolor}{rgb}{0.000000,0.000000,0.000000}%
\pgfsetstrokecolor{textcolor}%
\pgfsetfillcolor{textcolor}%
\pgftext[x=1.702732in,y=1.553694in,left,base]{\color{textcolor}{\rmfamily\fontsize{20.000000}{24.000000}\selectfont\catcode`\^=\active\def^{\ifmmode\sp\else\^{}\fi}\catcode`\%=\active\def
\end{pgfscope}%
\begin{pgfscope}%
\pgfsetbuttcap%
\pgfsetroundjoin%
\pgfsetlinewidth{2.509375pt}%
\definecolor{currentstroke}{rgb}{0.960784,0.462745,0.000000}%
\pgfsetstrokecolor{currentstroke}%
\pgfsetdash{{9.250000pt}{4.000000pt}}{0.000000pt}%
\pgfpathmoveto{\pgfqpoint{0.924954in}{1.263555in}}%
\pgfpathlineto{\pgfqpoint{1.202732in}{1.263555in}}%
\pgfpathlineto{\pgfqpoint{1.480509in}{1.263555in}}%
\pgfusepath{stroke}%
\end{pgfscope}%
\begin{pgfscope}%
\pgfsetbuttcap%
\pgfsetmiterjoin%
\definecolor{currentfill}{rgb}{0.960784,0.462745,0.000000}%
\pgfsetfillcolor{currentfill}%
\pgfsetlinewidth{1.003750pt}%
\definecolor{currentstroke}{rgb}{0.960784,0.462745,0.000000}%
\pgfsetstrokecolor{currentstroke}%
\pgfsetdash{}{0pt}%
\pgfsys@defobject{currentmarker}{\pgfqpoint{-0.055556in}{-0.055556in}}{\pgfqpoint{0.055556in}{0.055556in}}{%
\pgfpathmoveto{\pgfqpoint{-0.055556in}{-0.055556in}}%
\pgfpathlineto{\pgfqpoint{0.055556in}{-0.055556in}}%
\pgfpathlineto{\pgfqpoint{0.055556in}{0.055556in}}%
\pgfpathlineto{\pgfqpoint{-0.055556in}{0.055556in}}%
\pgfpathlineto{\pgfqpoint{-0.055556in}{-0.055556in}}%
\pgfpathclose%
\pgfusepath{stroke,fill}%
}%
\begin{pgfscope}%
\pgfsys@transformshift{1.202732in}{1.263555in}%
\pgfsys@useobject{currentmarker}{}%
\end{pgfscope}%
\end{pgfscope}%
\begin{pgfscope}%
\definecolor{textcolor}{rgb}{0.000000,0.000000,0.000000}%
\pgfsetstrokecolor{textcolor}%
\pgfsetfillcolor{textcolor}%
\pgftext[x=1.702732in,y=1.166333in,left,base]{\color{textcolor}{\rmfamily\fontsize{20.000000}{24.000000}\selectfont\catcode`\^=\active\def^{\ifmmode\sp\else\^{}\fi}\catcode`\%=\active\def
\end{pgfscope}%
\end{pgfpicture}%
\makeatother%
\endgroup%

%% file: images/thermal_ratio_per_modality/modality_results_RRA_30.pgf
\begingroup%
\makeatletter%
\begin{pgfpicture}%
\pgfpathrectangle{\pgfpointorigin}{\pgfqpoint{7.450000in}{5.450000in}}%
\pgfusepath{use as bounding box, clip}%
\begin{pgfscope}%
\pgfsetbuttcap%
\pgfsetmiterjoin%
\definecolor{currentfill}{rgb}{1.000000,1.000000,1.000000}%
\pgfsetfillcolor{currentfill}%
\pgfsetlinewidth{0.000000pt}%
\definecolor{currentstroke}{rgb}{1.000000,1.000000,1.000000}%
\pgfsetstrokecolor{currentstroke}%
\pgfsetdash{}{0pt}%
\pgfpathmoveto{\pgfqpoint{0.000000in}{0.000000in}}%
\pgfpathlineto{\pgfqpoint{7.450000in}{0.000000in}}%
\pgfpathlineto{\pgfqpoint{7.450000in}{5.450000in}}%
\pgfpathlineto{\pgfqpoint{0.000000in}{5.450000in}}%
\pgfpathlineto{\pgfqpoint{0.000000in}{0.000000in}}%
\pgfpathclose%
\pgfusepath{fill}%
\end{pgfscope}%
\begin{pgfscope}%
\pgfsetbuttcap%
\pgfsetmiterjoin%
\definecolor{currentfill}{rgb}{1.000000,1.000000,1.000000}%
\pgfsetfillcolor{currentfill}%
\pgfsetlinewidth{0.000000pt}%
\definecolor{currentstroke}{rgb}{0.000000,0.000000,0.000000}%
\pgfsetstrokecolor{currentstroke}%
\pgfsetstrokeopacity{0.000000}%
\pgfsetdash{}{0pt}%
\pgfpathmoveto{\pgfqpoint{0.674954in}{0.862305in}}%
\pgfpathlineto{\pgfqpoint{7.147414in}{0.862305in}}%
\pgfpathlineto{\pgfqpoint{7.147414in}{5.231386in}}%
\pgfpathlineto{\pgfqpoint{0.674954in}{5.231386in}}%
\pgfpathlineto{\pgfqpoint{0.674954in}{0.862305in}}%
\pgfpathclose%
\pgfusepath{fill}%
\end{pgfscope}%
\begin{pgfscope}%
\pgfpathrectangle{\pgfqpoint{0.674954in}{0.862305in}}{\pgfqpoint{6.472460in}{4.369081in}}%
\pgfusepath{clip}%
\pgfsetbuttcap%
\pgfsetroundjoin%
\pgfsetlinewidth{2.007500pt}%
\definecolor{currentstroke}{rgb}{0.501961,0.501961,0.501961}%
\pgfsetstrokecolor{currentstroke}%
\pgfsetstrokeopacity{0.300000}%
\pgfsetdash{{7.400000pt}{3.200000pt}}{0.000000pt}%
\pgfpathmoveto{\pgfqpoint{0.674954in}{0.862305in}}%
\pgfpathlineto{\pgfqpoint{0.674954in}{5.231386in}}%
\pgfusepath{stroke}%
\end{pgfscope}%
\begin{pgfscope}%
\pgfsetbuttcap%
\pgfsetroundjoin%
\definecolor{currentfill}{rgb}{0.000000,0.000000,0.000000}%
\pgfsetfillcolor{currentfill}%
\pgfsetlinewidth{0.803000pt}%
\definecolor{currentstroke}{rgb}{0.000000,0.000000,0.000000}%
\pgfsetstrokecolor{currentstroke}%
\pgfsetdash{}{0pt}%
\pgfsys@defobject{currentmarker}{\pgfqpoint{0.000000in}{-0.048611in}}{\pgfqpoint{0.000000in}{0.000000in}}{%
\pgfpathmoveto{\pgfqpoint{0.000000in}{0.000000in}}%
\pgfpathlineto{\pgfqpoint{0.000000in}{-0.048611in}}%
\pgfusepath{stroke,fill}%
}%
\begin{pgfscope}%
\pgfsys@transformshift{0.674954in}{0.862305in}%
\pgfsys@useobject{currentmarker}{}%
\end{pgfscope}%
\end{pgfscope}%
\begin{pgfscope}%
\definecolor{textcolor}{rgb}{0.000000,0.000000,0.000000}%
\pgfsetstrokecolor{textcolor}%
\pgfsetfillcolor{textcolor}%
\pgftext[x=0.674954in,y=0.765082in,,top]{\color{textcolor}{\rmfamily\fontsize{25.000000}{30.000000}\selectfont\catcode`\^=\active\def^{\ifmmode\sp\else\^{}\fi}\catcode`\%=\active\def
\end{pgfscope}%
\begin{pgfscope}%
\pgfpathrectangle{\pgfqpoint{0.674954in}{0.862305in}}{\pgfqpoint{6.472460in}{4.369081in}}%
\pgfusepath{clip}%
\pgfsetbuttcap%
\pgfsetroundjoin%
\pgfsetlinewidth{2.007500pt}%
\definecolor{currentstroke}{rgb}{0.501961,0.501961,0.501961}%
\pgfsetstrokecolor{currentstroke}%
\pgfsetstrokeopacity{0.300000}%
\pgfsetdash{{7.400000pt}{3.200000pt}}{0.000000pt}%
\pgfpathmoveto{\pgfqpoint{1.969446in}{0.862305in}}%
\pgfpathlineto{\pgfqpoint{1.969446in}{5.231386in}}%
\pgfusepath{stroke}%
\end{pgfscope}%
\begin{pgfscope}%
\pgfsetbuttcap%
\pgfsetroundjoin%
\definecolor{currentfill}{rgb}{0.000000,0.000000,0.000000}%
\pgfsetfillcolor{currentfill}%
\pgfsetlinewidth{0.803000pt}%
\definecolor{currentstroke}{rgb}{0.000000,0.000000,0.000000}%
\pgfsetstrokecolor{currentstroke}%
\pgfsetdash{}{0pt}%
\pgfsys@defobject{currentmarker}{\pgfqpoint{0.000000in}{-0.048611in}}{\pgfqpoint{0.000000in}{0.000000in}}{%
\pgfpathmoveto{\pgfqpoint{0.000000in}{0.000000in}}%
\pgfpathlineto{\pgfqpoint{0.000000in}{-0.048611in}}%
\pgfusepath{stroke,fill}%
}%
\begin{pgfscope}%
\pgfsys@transformshift{1.969446in}{0.862305in}%
\pgfsys@useobject{currentmarker}{}%
\end{pgfscope}%
\end{pgfscope}%
\begin{pgfscope}%
\definecolor{textcolor}{rgb}{0.000000,0.000000,0.000000}%
\pgfsetstrokecolor{textcolor}%
\pgfsetfillcolor{textcolor}%
\pgftext[x=1.969446in,y=0.765082in,,top]{\color{textcolor}{\rmfamily\fontsize{25.000000}{30.000000}\selectfont\catcode`\^=\active\def^{\ifmmode\sp\else\^{}\fi}\catcode`\%=\active\def
\end{pgfscope}%
\begin{pgfscope}%
\pgfpathrectangle{\pgfqpoint{0.674954in}{0.862305in}}{\pgfqpoint{6.472460in}{4.369081in}}%
\pgfusepath{clip}%
\pgfsetbuttcap%
\pgfsetroundjoin%
\pgfsetlinewidth{2.007500pt}%
\definecolor{currentstroke}{rgb}{0.501961,0.501961,0.501961}%
\pgfsetstrokecolor{currentstroke}%
\pgfsetstrokeopacity{0.300000}%
\pgfsetdash{{7.400000pt}{3.200000pt}}{0.000000pt}%
\pgfpathmoveto{\pgfqpoint{3.263938in}{0.862305in}}%
\pgfpathlineto{\pgfqpoint{3.263938in}{5.231386in}}%
\pgfusepath{stroke}%
\end{pgfscope}%
\begin{pgfscope}%
\pgfsetbuttcap%
\pgfsetroundjoin%
\definecolor{currentfill}{rgb}{0.000000,0.000000,0.000000}%
\pgfsetfillcolor{currentfill}%
\pgfsetlinewidth{0.803000pt}%
\definecolor{currentstroke}{rgb}{0.000000,0.000000,0.000000}%
\pgfsetstrokecolor{currentstroke}%
\pgfsetdash{}{0pt}%
\pgfsys@defobject{currentmarker}{\pgfqpoint{0.000000in}{-0.048611in}}{\pgfqpoint{0.000000in}{0.000000in}}{%
\pgfpathmoveto{\pgfqpoint{0.000000in}{0.000000in}}%
\pgfpathlineto{\pgfqpoint{0.000000in}{-0.048611in}}%
\pgfusepath{stroke,fill}%
}%
\begin{pgfscope}%
\pgfsys@transformshift{3.263938in}{0.862305in}%
\pgfsys@useobject{currentmarker}{}%
\end{pgfscope}%
\end{pgfscope}%
\begin{pgfscope}%
\definecolor{textcolor}{rgb}{0.000000,0.000000,0.000000}%
\pgfsetstrokecolor{textcolor}%
\pgfsetfillcolor{textcolor}%
\pgftext[x=3.263938in,y=0.765082in,,top]{\color{textcolor}{\rmfamily\fontsize{25.000000}{30.000000}\selectfont\catcode`\^=\active\def^{\ifmmode\sp\else\^{}\fi}\catcode`\%=\active\def
\end{pgfscope}%
\begin{pgfscope}%
\pgfpathrectangle{\pgfqpoint{0.674954in}{0.862305in}}{\pgfqpoint{6.472460in}{4.369081in}}%
\pgfusepath{clip}%
\pgfsetbuttcap%
\pgfsetroundjoin%
\pgfsetlinewidth{2.007500pt}%
\definecolor{currentstroke}{rgb}{0.501961,0.501961,0.501961}%
\pgfsetstrokecolor{currentstroke}%
\pgfsetstrokeopacity{0.300000}%
\pgfsetdash{{7.400000pt}{3.200000pt}}{0.000000pt}%
\pgfpathmoveto{\pgfqpoint{4.558430in}{0.862305in}}%
\pgfpathlineto{\pgfqpoint{4.558430in}{5.231386in}}%
\pgfusepath{stroke}%
\end{pgfscope}%
\begin{pgfscope}%
\pgfsetbuttcap%
\pgfsetroundjoin%
\definecolor{currentfill}{rgb}{0.000000,0.000000,0.000000}%
\pgfsetfillcolor{currentfill}%
\pgfsetlinewidth{0.803000pt}%
\definecolor{currentstroke}{rgb}{0.000000,0.000000,0.000000}%
\pgfsetstrokecolor{currentstroke}%
\pgfsetdash{}{0pt}%
\pgfsys@defobject{currentmarker}{\pgfqpoint{0.000000in}{-0.048611in}}{\pgfqpoint{0.000000in}{0.000000in}}{%
\pgfpathmoveto{\pgfqpoint{0.000000in}{0.000000in}}%
\pgfpathlineto{\pgfqpoint{0.000000in}{-0.048611in}}%
\pgfusepath{stroke,fill}%
}%
\begin{pgfscope}%
\pgfsys@transformshift{4.558430in}{0.862305in}%
\pgfsys@useobject{currentmarker}{}%
\end{pgfscope}%
\end{pgfscope}%
\begin{pgfscope}%
\definecolor{textcolor}{rgb}{0.000000,0.000000,0.000000}%
\pgfsetstrokecolor{textcolor}%
\pgfsetfillcolor{textcolor}%
\pgftext[x=4.558430in,y=0.765082in,,top]{\color{textcolor}{\rmfamily\fontsize{25.000000}{30.000000}\selectfont\catcode`\^=\active\def^{\ifmmode\sp\else\^{}\fi}\catcode`\%=\active\def
\end{pgfscope}%
\begin{pgfscope}%
\pgfpathrectangle{\pgfqpoint{0.674954in}{0.862305in}}{\pgfqpoint{6.472460in}{4.369081in}}%
\pgfusepath{clip}%
\pgfsetbuttcap%
\pgfsetroundjoin%
\pgfsetlinewidth{2.007500pt}%
\definecolor{currentstroke}{rgb}{0.501961,0.501961,0.501961}%
\pgfsetstrokecolor{currentstroke}%
\pgfsetstrokeopacity{0.300000}%
\pgfsetdash{{7.400000pt}{3.200000pt}}{0.000000pt}%
\pgfpathmoveto{\pgfqpoint{5.852922in}{0.862305in}}%
\pgfpathlineto{\pgfqpoint{5.852922in}{5.231386in}}%
\pgfusepath{stroke}%
\end{pgfscope}%
\begin{pgfscope}%
\pgfsetbuttcap%
\pgfsetroundjoin%
\definecolor{currentfill}{rgb}{0.000000,0.000000,0.000000}%
\pgfsetfillcolor{currentfill}%
\pgfsetlinewidth{0.803000pt}%
\definecolor{currentstroke}{rgb}{0.000000,0.000000,0.000000}%
\pgfsetstrokecolor{currentstroke}%
\pgfsetdash{}{0pt}%
\pgfsys@defobject{currentmarker}{\pgfqpoint{0.000000in}{-0.048611in}}{\pgfqpoint{0.000000in}{0.000000in}}{%
\pgfpathmoveto{\pgfqpoint{0.000000in}{0.000000in}}%
\pgfpathlineto{\pgfqpoint{0.000000in}{-0.048611in}}%
\pgfusepath{stroke,fill}%
}%
\begin{pgfscope}%
\pgfsys@transformshift{5.852922in}{0.862305in}%
\pgfsys@useobject{currentmarker}{}%
\end{pgfscope}%
\end{pgfscope}%
\begin{pgfscope}%
\definecolor{textcolor}{rgb}{0.000000,0.000000,0.000000}%
\pgfsetstrokecolor{textcolor}%
\pgfsetfillcolor{textcolor}%
\pgftext[x=5.852922in,y=0.765082in,,top]{\color{textcolor}{\rmfamily\fontsize{25.000000}{30.000000}\selectfont\catcode`\^=\active\def^{\ifmmode\sp\else\^{}\fi}\catcode`\%=\active\def
\end{pgfscope}%
\begin{pgfscope}%
\pgfpathrectangle{\pgfqpoint{0.674954in}{0.862305in}}{\pgfqpoint{6.472460in}{4.369081in}}%
\pgfusepath{clip}%
\pgfsetbuttcap%
\pgfsetroundjoin%
\pgfsetlinewidth{2.007500pt}%
\definecolor{currentstroke}{rgb}{0.501961,0.501961,0.501961}%
\pgfsetstrokecolor{currentstroke}%
\pgfsetstrokeopacity{0.300000}%
\pgfsetdash{{7.400000pt}{3.200000pt}}{0.000000pt}%
\pgfpathmoveto{\pgfqpoint{7.147414in}{0.862305in}}%
\pgfpathlineto{\pgfqpoint{7.147414in}{5.231386in}}%
\pgfusepath{stroke}%
\end{pgfscope}%
\begin{pgfscope}%
\pgfsetbuttcap%
\pgfsetroundjoin%
\definecolor{currentfill}{rgb}{0.000000,0.000000,0.000000}%
\pgfsetfillcolor{currentfill}%
\pgfsetlinewidth{0.803000pt}%
\definecolor{currentstroke}{rgb}{0.000000,0.000000,0.000000}%
\pgfsetstrokecolor{currentstroke}%
\pgfsetdash{}{0pt}%
\pgfsys@defobject{currentmarker}{\pgfqpoint{0.000000in}{-0.048611in}}{\pgfqpoint{0.000000in}{0.000000in}}{%
\pgfpathmoveto{\pgfqpoint{0.000000in}{0.000000in}}%
\pgfpathlineto{\pgfqpoint{0.000000in}{-0.048611in}}%
\pgfusepath{stroke,fill}%
}%
\begin{pgfscope}%
\pgfsys@transformshift{7.147414in}{0.862305in}%
\pgfsys@useobject{currentmarker}{}%
\end{pgfscope}%
\end{pgfscope}%
\begin{pgfscope}%
\definecolor{textcolor}{rgb}{0.000000,0.000000,0.000000}%
\pgfsetstrokecolor{textcolor}%
\pgfsetfillcolor{textcolor}%
\pgftext[x=7.147414in,y=0.765082in,,top]{\color{textcolor}{\rmfamily\fontsize{25.000000}{30.000000}\selectfont\catcode`\^=\active\def^{\ifmmode\sp\else\^{}\fi}\catcode`\%=\active\def
\end{pgfscope}%
\begin{pgfscope}%
\definecolor{textcolor}{rgb}{0.000000,0.000000,0.000000}%
\pgfsetstrokecolor{textcolor}%
\pgfsetfillcolor{textcolor}%
\pgftext[x=3.911184in,y=0.404763in,,top]{\color{textcolor}{\rmfamily\fontsize{25.000000}{30.000000}\selectfont\catcode`\^=\active\def^{\ifmmode\sp\else\^{}\fi}\catcode`\%=\active\def
\end{pgfscope}%
\begin{pgfscope}%
\pgfpathrectangle{\pgfqpoint{0.674954in}{0.862305in}}{\pgfqpoint{6.472460in}{4.369081in}}%
\pgfusepath{clip}%
\pgfsetbuttcap%
\pgfsetroundjoin%
\pgfsetlinewidth{2.007500pt}%
\definecolor{currentstroke}{rgb}{0.501961,0.501961,0.501961}%
\pgfsetstrokecolor{currentstroke}%
\pgfsetstrokeopacity{0.300000}%
\pgfsetdash{{7.400000pt}{3.200000pt}}{0.000000pt}%
\pgfpathmoveto{\pgfqpoint{0.674954in}{1.954575in}}%
\pgfpathlineto{\pgfqpoint{7.147414in}{1.954575in}}%
\pgfusepath{stroke}%
\end{pgfscope}%
\begin{pgfscope}%
\pgfsetbuttcap%
\pgfsetroundjoin%
\definecolor{currentfill}{rgb}{0.000000,0.000000,0.000000}%
\pgfsetfillcolor{currentfill}%
\pgfsetlinewidth{0.803000pt}%
\definecolor{currentstroke}{rgb}{0.000000,0.000000,0.000000}%
\pgfsetstrokecolor{currentstroke}%
\pgfsetdash{}{0pt}%
\pgfsys@defobject{currentmarker}{\pgfqpoint{-0.048611in}{0.000000in}}{\pgfqpoint{-0.000000in}{0.000000in}}{%
\pgfpathmoveto{\pgfqpoint{-0.000000in}{0.000000in}}%
\pgfpathlineto{\pgfqpoint{-0.048611in}{0.000000in}}%
\pgfusepath{stroke,fill}%
}%
\begin{pgfscope}%
\pgfsys@transformshift{0.674954in}{1.954575in}%
\pgfsys@useobject{currentmarker}{}%
\end{pgfscope}%
\end{pgfscope}%
\begin{pgfscope}%
\definecolor{textcolor}{rgb}{0.000000,0.000000,0.000000}%
\pgfsetstrokecolor{textcolor}%
\pgfsetfillcolor{textcolor}%
\pgftext[x=0.259244in, y=1.835961in, left, base]{\color{textcolor}{\rmfamily\fontsize{25.000000}{30.000000}\selectfont\catcode`\^=\active\def^{\ifmmode\sp\else\^{}\fi}\catcode`\%=\active\def
\end{pgfscope}%
\begin{pgfscope}%
\pgfpathrectangle{\pgfqpoint{0.674954in}{0.862305in}}{\pgfqpoint{6.472460in}{4.369081in}}%
\pgfusepath{clip}%
\pgfsetbuttcap%
\pgfsetroundjoin%
\pgfsetlinewidth{2.007500pt}%
\definecolor{currentstroke}{rgb}{0.501961,0.501961,0.501961}%
\pgfsetstrokecolor{currentstroke}%
\pgfsetstrokeopacity{0.300000}%
\pgfsetdash{{7.400000pt}{3.200000pt}}{0.000000pt}%
\pgfpathmoveto{\pgfqpoint{0.674954in}{3.046845in}}%
\pgfpathlineto{\pgfqpoint{7.147414in}{3.046845in}}%
\pgfusepath{stroke}%
\end{pgfscope}%
\begin{pgfscope}%
\pgfsetbuttcap%
\pgfsetroundjoin%
\definecolor{currentfill}{rgb}{0.000000,0.000000,0.000000}%
\pgfsetfillcolor{currentfill}%
\pgfsetlinewidth{0.803000pt}%
\definecolor{currentstroke}{rgb}{0.000000,0.000000,0.000000}%
\pgfsetstrokecolor{currentstroke}%
\pgfsetdash{}{0pt}%
\pgfsys@defobject{currentmarker}{\pgfqpoint{-0.048611in}{0.000000in}}{\pgfqpoint{-0.000000in}{0.000000in}}{%
\pgfpathmoveto{\pgfqpoint{-0.000000in}{0.000000in}}%
\pgfpathlineto{\pgfqpoint{-0.048611in}{0.000000in}}%
\pgfusepath{stroke,fill}%
}%
\begin{pgfscope}%
\pgfsys@transformshift{0.674954in}{3.046845in}%
\pgfsys@useobject{currentmarker}{}%
\end{pgfscope}%
\end{pgfscope}%
\begin{pgfscope}%
\definecolor{textcolor}{rgb}{0.000000,0.000000,0.000000}%
\pgfsetstrokecolor{textcolor}%
\pgfsetfillcolor{textcolor}%
\pgftext[x=0.259244in, y=2.928231in, left, base]{\color{textcolor}{\rmfamily\fontsize{25.000000}{30.000000}\selectfont\catcode`\^=\active\def^{\ifmmode\sp\else\^{}\fi}\catcode`\%=\active\def
\end{pgfscope}%
\begin{pgfscope}%
\pgfpathrectangle{\pgfqpoint{0.674954in}{0.862305in}}{\pgfqpoint{6.472460in}{4.369081in}}%
\pgfusepath{clip}%
\pgfsetbuttcap%
\pgfsetroundjoin%
\pgfsetlinewidth{2.007500pt}%
\definecolor{currentstroke}{rgb}{0.501961,0.501961,0.501961}%
\pgfsetstrokecolor{currentstroke}%
\pgfsetstrokeopacity{0.300000}%
\pgfsetdash{{7.400000pt}{3.200000pt}}{0.000000pt}%
\pgfpathmoveto{\pgfqpoint{0.674954in}{4.139116in}}%
\pgfpathlineto{\pgfqpoint{7.147414in}{4.139116in}}%
\pgfusepath{stroke}%
\end{pgfscope}%
\begin{pgfscope}%
\pgfsetbuttcap%
\pgfsetroundjoin%
\definecolor{currentfill}{rgb}{0.000000,0.000000,0.000000}%
\pgfsetfillcolor{currentfill}%
\pgfsetlinewidth{0.803000pt}%
\definecolor{currentstroke}{rgb}{0.000000,0.000000,0.000000}%
\pgfsetstrokecolor{currentstroke}%
\pgfsetdash{}{0pt}%
\pgfsys@defobject{currentmarker}{\pgfqpoint{-0.048611in}{0.000000in}}{\pgfqpoint{-0.000000in}{0.000000in}}{%
\pgfpathmoveto{\pgfqpoint{-0.000000in}{0.000000in}}%
\pgfpathlineto{\pgfqpoint{-0.048611in}{0.000000in}}%
\pgfusepath{stroke,fill}%
}%
\begin{pgfscope}%
\pgfsys@transformshift{0.674954in}{4.139116in}%
\pgfsys@useobject{currentmarker}{}%
\end{pgfscope}%
\end{pgfscope}%
\begin{pgfscope}%
\definecolor{textcolor}{rgb}{0.000000,0.000000,0.000000}%
\pgfsetstrokecolor{textcolor}%
\pgfsetfillcolor{textcolor}%
\pgftext[x=0.259244in, y=4.020501in, left, base]{\color{textcolor}{\rmfamily\fontsize{25.000000}{30.000000}\selectfont\catcode`\^=\active\def^{\ifmmode\sp\else\^{}\fi}\catcode`\%=\active\def
\end{pgfscope}%
\begin{pgfscope}%
\pgfpathrectangle{\pgfqpoint{0.674954in}{0.862305in}}{\pgfqpoint{6.472460in}{4.369081in}}%
\pgfusepath{clip}%
\pgfsetbuttcap%
\pgfsetroundjoin%
\pgfsetlinewidth{2.007500pt}%
\definecolor{currentstroke}{rgb}{0.501961,0.501961,0.501961}%
\pgfsetstrokecolor{currentstroke}%
\pgfsetstrokeopacity{0.300000}%
\pgfsetdash{{7.400000pt}{3.200000pt}}{0.000000pt}%
\pgfpathmoveto{\pgfqpoint{0.674954in}{5.231386in}}%
\pgfpathlineto{\pgfqpoint{7.147414in}{5.231386in}}%
\pgfusepath{stroke}%
\end{pgfscope}%
\begin{pgfscope}%
\pgfsetbuttcap%
\pgfsetroundjoin%
\definecolor{currentfill}{rgb}{0.000000,0.000000,0.000000}%
\pgfsetfillcolor{currentfill}%
\pgfsetlinewidth{0.803000pt}%
\definecolor{currentstroke}{rgb}{0.000000,0.000000,0.000000}%
\pgfsetstrokecolor{currentstroke}%
\pgfsetdash{}{0pt}%
\pgfsys@defobject{currentmarker}{\pgfqpoint{-0.048611in}{0.000000in}}{\pgfqpoint{-0.000000in}{0.000000in}}{%
\pgfpathmoveto{\pgfqpoint{-0.000000in}{0.000000in}}%
\pgfpathlineto{\pgfqpoint{-0.048611in}{0.000000in}}%
\pgfusepath{stroke,fill}%
}%
\begin{pgfscope}%
\pgfsys@transformshift{0.674954in}{5.231386in}%
\pgfsys@useobject{currentmarker}{}%
\end{pgfscope}%
\end{pgfscope}%
\begin{pgfscope}%
\definecolor{textcolor}{rgb}{0.000000,0.000000,0.000000}%
\pgfsetstrokecolor{textcolor}%
\pgfsetfillcolor{textcolor}%
\pgftext[x=0.100000in, y=5.112772in, left, base]{\color{textcolor}{\rmfamily\fontsize{25.000000}{30.000000}\selectfont\catcode`\^=\active\def^{\ifmmode\sp\else\^{}\fi}\catcode`\%=\active\def
\end{pgfscope}%
\begin{pgfscope}%
\pgfpathrectangle{\pgfqpoint{0.674954in}{0.862305in}}{\pgfqpoint{6.472460in}{4.369081in}}%
\pgfusepath{clip}%
\pgfsetrectcap%
\pgfsetroundjoin%
\pgfsetlinewidth{2.509375pt}%
\definecolor{currentstroke}{rgb}{0.050980,0.415686,0.509804}%
\pgfsetstrokecolor{currentstroke}%
\pgfsetdash{}{0pt}%
\pgfpathmoveto{\pgfqpoint{0.674954in}{5.231386in}}%
\pgfpathlineto{\pgfqpoint{0.998577in}{5.231386in}}%
\pgfpathlineto{\pgfqpoint{2.293069in}{5.213881in}}%
\pgfpathlineto{\pgfqpoint{3.911184in}{5.227764in}}%
\pgfpathlineto{\pgfqpoint{5.529299in}{5.119257in}}%
\pgfpathlineto{\pgfqpoint{6.823791in}{4.684287in}}%
\pgfusepath{stroke}%
\end{pgfscope}%
\begin{pgfscope}%
\pgfpathrectangle{\pgfqpoint{0.674954in}{0.862305in}}{\pgfqpoint{6.472460in}{4.369081in}}%
\pgfusepath{clip}%
\pgfsetbuttcap%
\pgfsetroundjoin%
\definecolor{currentfill}{rgb}{0.050980,0.415686,0.509804}%
\pgfsetfillcolor{currentfill}%
\pgfsetlinewidth{1.003750pt}%
\definecolor{currentstroke}{rgb}{0.050980,0.415686,0.509804}%
\pgfsetstrokecolor{currentstroke}%
\pgfsetdash{}{0pt}%
\pgfsys@defobject{currentmarker}{\pgfqpoint{-0.055556in}{-0.055556in}}{\pgfqpoint{0.055556in}{0.055556in}}{%
\pgfpathmoveto{\pgfqpoint{0.000000in}{-0.055556in}}%
\pgfpathcurveto{\pgfqpoint{0.014734in}{-0.055556in}}{\pgfqpoint{0.028866in}{-0.049702in}}{\pgfqpoint{0.039284in}{-0.039284in}}%
\pgfpathcurveto{\pgfqpoint{0.049702in}{-0.028866in}}{\pgfqpoint{0.055556in}{-0.014734in}}{\pgfqpoint{0.055556in}{0.000000in}}%
\pgfpathcurveto{\pgfqpoint{0.055556in}{0.014734in}}{\pgfqpoint{0.049702in}{0.028866in}}{\pgfqpoint{0.039284in}{0.039284in}}%
\pgfpathcurveto{\pgfqpoint{0.028866in}{0.049702in}}{\pgfqpoint{0.014734in}{0.055556in}}{\pgfqpoint{0.000000in}{0.055556in}}%
\pgfpathcurveto{\pgfqpoint{-0.014734in}{0.055556in}}{\pgfqpoint{-0.028866in}{0.049702in}}{\pgfqpoint{-0.039284in}{0.039284in}}%
\pgfpathcurveto{\pgfqpoint{-0.049702in}{0.028866in}}{\pgfqpoint{-0.055556in}{0.014734in}}{\pgfqpoint{-0.055556in}{0.000000in}}%
\pgfpathcurveto{\pgfqpoint{-0.055556in}{-0.014734in}}{\pgfqpoint{-0.049702in}{-0.028866in}}{\pgfqpoint{-0.039284in}{-0.039284in}}%
\pgfpathcurveto{\pgfqpoint{-0.028866in}{-0.049702in}}{\pgfqpoint{-0.014734in}{-0.055556in}}{\pgfqpoint{0.000000in}{-0.055556in}}%
\pgfpathlineto{\pgfqpoint{0.000000in}{-0.055556in}}%
\pgfpathclose%
\pgfusepath{stroke,fill}%
}%
\begin{pgfscope}%
\pgfsys@transformshift{0.674954in}{5.231386in}%
\pgfsys@useobject{currentmarker}{}%
\end{pgfscope}%
\begin{pgfscope}%
\pgfsys@transformshift{0.998577in}{5.231386in}%
\pgfsys@useobject{currentmarker}{}%
\end{pgfscope}%
\begin{pgfscope}%
\pgfsys@transformshift{2.293069in}{5.213881in}%
\pgfsys@useobject{currentmarker}{}%
\end{pgfscope}%
\begin{pgfscope}%
\pgfsys@transformshift{3.911184in}{5.227764in}%
\pgfsys@useobject{currentmarker}{}%
\end{pgfscope}%
\begin{pgfscope}%
\pgfsys@transformshift{5.529299in}{5.119257in}%
\pgfsys@useobject{currentmarker}{}%
\end{pgfscope}%
\begin{pgfscope}%
\pgfsys@transformshift{6.823791in}{4.684287in}%
\pgfsys@useobject{currentmarker}{}%
\end{pgfscope}%
\end{pgfscope}%
\begin{pgfscope}%
\pgfpathrectangle{\pgfqpoint{0.674954in}{0.862305in}}{\pgfqpoint{6.472460in}{4.369081in}}%
\pgfusepath{clip}%
\pgfsetbuttcap%
\pgfsetroundjoin%
\pgfsetlinewidth{2.509375pt}%
\definecolor{currentstroke}{rgb}{0.960784,0.462745,0.000000}%
\pgfsetstrokecolor{currentstroke}%
\pgfsetdash{{9.250000pt}{4.000000pt}}{0.000000pt}%
\pgfpathmoveto{\pgfqpoint{0.998577in}{4.809948in}}%
\pgfpathlineto{\pgfqpoint{2.293069in}{4.854490in}}%
\pgfpathlineto{\pgfqpoint{3.911184in}{4.779254in}}%
\pgfpathlineto{\pgfqpoint{5.529299in}{4.775743in}}%
\pgfpathlineto{\pgfqpoint{6.823791in}{4.841702in}}%
\pgfpathlineto{\pgfqpoint{7.147414in}{4.954444in}}%
\pgfusepath{stroke}%
\end{pgfscope}%
\begin{pgfscope}%
\pgfpathrectangle{\pgfqpoint{0.674954in}{0.862305in}}{\pgfqpoint{6.472460in}{4.369081in}}%
\pgfusepath{clip}%
\pgfsetbuttcap%
\pgfsetmiterjoin%
\definecolor{currentfill}{rgb}{0.960784,0.462745,0.000000}%
\pgfsetfillcolor{currentfill}%
\pgfsetlinewidth{1.003750pt}%
\definecolor{currentstroke}{rgb}{0.960784,0.462745,0.000000}%
\pgfsetstrokecolor{currentstroke}%
\pgfsetdash{}{0pt}%
\pgfsys@defobject{currentmarker}{\pgfqpoint{-0.055556in}{-0.055556in}}{\pgfqpoint{0.055556in}{0.055556in}}{%
\pgfpathmoveto{\pgfqpoint{-0.055556in}{-0.055556in}}%
\pgfpathlineto{\pgfqpoint{0.055556in}{-0.055556in}}%
\pgfpathlineto{\pgfqpoint{0.055556in}{0.055556in}}%
\pgfpathlineto{\pgfqpoint{-0.055556in}{0.055556in}}%
\pgfpathlineto{\pgfqpoint{-0.055556in}{-0.055556in}}%
\pgfpathclose%
\pgfusepath{stroke,fill}%
}%
\begin{pgfscope}%
\pgfsys@transformshift{0.998577in}{4.809948in}%
\pgfsys@useobject{currentmarker}{}%
\end{pgfscope}%
\begin{pgfscope}%
\pgfsys@transformshift{2.293069in}{4.854490in}%
\pgfsys@useobject{currentmarker}{}%
\end{pgfscope}%
\begin{pgfscope}%
\pgfsys@transformshift{3.911184in}{4.779254in}%
\pgfsys@useobject{currentmarker}{}%
\end{pgfscope}%
\begin{pgfscope}%
\pgfsys@transformshift{5.529299in}{4.775743in}%
\pgfsys@useobject{currentmarker}{}%
\end{pgfscope}%
\begin{pgfscope}%
\pgfsys@transformshift{6.823791in}{4.841702in}%
\pgfsys@useobject{currentmarker}{}%
\end{pgfscope}%
\begin{pgfscope}%
\pgfsys@transformshift{7.147414in}{4.954444in}%
\pgfsys@useobject{currentmarker}{}%
\end{pgfscope}%
\end{pgfscope}%
\begin{pgfscope}%
\pgfsetrectcap%
\pgfsetmiterjoin%
\pgfsetlinewidth{2.007500pt}%
\definecolor{currentstroke}{rgb}{0.000000,0.000000,0.000000}%
\pgfsetstrokecolor{currentstroke}%
\pgfsetdash{}{0pt}%
\pgfpathmoveto{\pgfqpoint{0.674954in}{0.862305in}}%
\pgfpathlineto{\pgfqpoint{0.674954in}{5.231386in}}%
\pgfusepath{stroke}%
\end{pgfscope}%
\begin{pgfscope}%
\pgfsetrectcap%
\pgfsetmiterjoin%
\pgfsetlinewidth{2.007500pt}%
\definecolor{currentstroke}{rgb}{0.000000,0.000000,0.000000}%
\pgfsetstrokecolor{currentstroke}%
\pgfsetdash{}{0pt}%
\pgfpathmoveto{\pgfqpoint{0.674954in}{0.862305in}}%
\pgfpathlineto{\pgfqpoint{7.147414in}{0.862305in}}%
\pgfusepath{stroke}%
\end{pgfscope}%
\begin{pgfscope}%
\pgfsetbuttcap%
\pgfsetmiterjoin%
\definecolor{currentfill}{rgb}{1.000000,1.000000,1.000000}%
\pgfsetfillcolor{currentfill}%
\pgfsetlinewidth{1.003750pt}%
\definecolor{currentstroke}{rgb}{0.800000,0.800000,0.800000}%
\pgfsetstrokecolor{currentstroke}%
\pgfsetdash{}{0pt}%
\pgfpathmoveto{\pgfqpoint{0.869398in}{1.001194in}}%
\pgfpathlineto{\pgfqpoint{2.708109in}{1.001194in}}%
\pgfpathquadraticcurveto{\pgfqpoint{2.763664in}{1.001194in}}{\pgfqpoint{2.763664in}{1.056749in}}%
\pgfpathlineto{\pgfqpoint{2.763664in}{1.803694in}}%
\pgfpathquadraticcurveto{\pgfqpoint{2.763664in}{1.859250in}}{\pgfqpoint{2.708109in}{1.859250in}}%
\pgfpathlineto{\pgfqpoint{0.869398in}{1.859250in}}%
\pgfpathquadraticcurveto{\pgfqpoint{0.813843in}{1.859250in}}{\pgfqpoint{0.813843in}{1.803694in}}%
\pgfpathlineto{\pgfqpoint{0.813843in}{1.056749in}}%
\pgfpathquadraticcurveto{\pgfqpoint{0.813843in}{1.001194in}}{\pgfqpoint{0.869398in}{1.001194in}}%
\pgfpathlineto{\pgfqpoint{0.869398in}{1.001194in}}%
\pgfpathclose%
\pgfusepath{stroke,fill}%
\end{pgfscope}%
\begin{pgfscope}%
\pgfsetrectcap%
\pgfsetroundjoin%
\pgfsetlinewidth{2.509375pt}%
\definecolor{currentstroke}{rgb}{0.050980,0.415686,0.509804}%
\pgfsetstrokecolor{currentstroke}%
\pgfsetdash{}{0pt}%
\pgfpathmoveto{\pgfqpoint{0.924954in}{1.650917in}}%
\pgfpathlineto{\pgfqpoint{1.202732in}{1.650917in}}%
\pgfpathlineto{\pgfqpoint{1.480509in}{1.650917in}}%
\pgfusepath{stroke}%
\end{pgfscope}%
\begin{pgfscope}%
\pgfsetbuttcap%
\pgfsetroundjoin%
\definecolor{currentfill}{rgb}{0.050980,0.415686,0.509804}%
\pgfsetfillcolor{currentfill}%
\pgfsetlinewidth{1.003750pt}%
\definecolor{currentstroke}{rgb}{0.050980,0.415686,0.509804}%
\pgfsetstrokecolor{currentstroke}%
\pgfsetdash{}{0pt}%
\pgfsys@defobject{currentmarker}{\pgfqpoint{-0.055556in}{-0.055556in}}{\pgfqpoint{0.055556in}{0.055556in}}{%
\pgfpathmoveto{\pgfqpoint{0.000000in}{-0.055556in}}%
\pgfpathcurveto{\pgfqpoint{0.014734in}{-0.055556in}}{\pgfqpoint{0.028866in}{-0.049702in}}{\pgfqpoint{0.039284in}{-0.039284in}}%
\pgfpathcurveto{\pgfqpoint{0.049702in}{-0.028866in}}{\pgfqpoint{0.055556in}{-0.014734in}}{\pgfqpoint{0.055556in}{0.000000in}}%
\pgfpathcurveto{\pgfqpoint{0.055556in}{0.014734in}}{\pgfqpoint{0.049702in}{0.028866in}}{\pgfqpoint{0.039284in}{0.039284in}}%
\pgfpathcurveto{\pgfqpoint{0.028866in}{0.049702in}}{\pgfqpoint{0.014734in}{0.055556in}}{\pgfqpoint{0.000000in}{0.055556in}}%
\pgfpathcurveto{\pgfqpoint{-0.014734in}{0.055556in}}{\pgfqpoint{-0.028866in}{0.049702in}}{\pgfqpoint{-0.039284in}{0.039284in}}%
\pgfpathcurveto{\pgfqpoint{-0.049702in}{0.028866in}}{\pgfqpoint{-0.055556in}{0.014734in}}{\pgfqpoint{-0.055556in}{0.000000in}}%
\pgfpathcurveto{\pgfqpoint{-0.055556in}{-0.014734in}}{\pgfqpoint{-0.049702in}{-0.028866in}}{\pgfqpoint{-0.039284in}{-0.039284in}}%
\pgfpathcurveto{\pgfqpoint{-0.028866in}{-0.049702in}}{\pgfqpoint{-0.014734in}{-0.055556in}}{\pgfqpoint{0.000000in}{-0.055556in}}%
\pgfpathlineto{\pgfqpoint{0.000000in}{-0.055556in}}%
\pgfpathclose%
\pgfusepath{stroke,fill}%
}%
\begin{pgfscope}%
\pgfsys@transformshift{1.202732in}{1.650917in}%
\pgfsys@useobject{currentmarker}{}%
\end{pgfscope}%
\end{pgfscope}%
\begin{pgfscope}%
\definecolor{textcolor}{rgb}{0.000000,0.000000,0.000000}%
\pgfsetstrokecolor{textcolor}%
\pgfsetfillcolor{textcolor}%
\pgftext[x=1.702732in,y=1.553694in,left,base]{\color{textcolor}{\rmfamily\fontsize{20.000000}{24.000000}\selectfont\catcode`\^=\active\def^{\ifmmode\sp\else\^{}\fi}\catcode`\%=\active\def
\end{pgfscope}%
\begin{pgfscope}%
\pgfsetbuttcap%
\pgfsetroundjoin%
\pgfsetlinewidth{2.509375pt}%
\definecolor{currentstroke}{rgb}{0.960784,0.462745,0.000000}%
\pgfsetstrokecolor{currentstroke}%
\pgfsetdash{{9.250000pt}{4.000000pt}}{0.000000pt}%
\pgfpathmoveto{\pgfqpoint{0.924954in}{1.263555in}}%
\pgfpathlineto{\pgfqpoint{1.202732in}{1.263555in}}%
\pgfpathlineto{\pgfqpoint{1.480509in}{1.263555in}}%
\pgfusepath{stroke}%
\end{pgfscope}%
\begin{pgfscope}%
\pgfsetbuttcap%
\pgfsetmiterjoin%
\definecolor{currentfill}{rgb}{0.960784,0.462745,0.000000}%
\pgfsetfillcolor{currentfill}%
\pgfsetlinewidth{1.003750pt}%
\definecolor{currentstroke}{rgb}{0.960784,0.462745,0.000000}%
\pgfsetstrokecolor{currentstroke}%
\pgfsetdash{}{0pt}%
\pgfsys@defobject{currentmarker}{\pgfqpoint{-0.055556in}{-0.055556in}}{\pgfqpoint{0.055556in}{0.055556in}}{%
\pgfpathmoveto{\pgfqpoint{-0.055556in}{-0.055556in}}%
\pgfpathlineto{\pgfqpoint{0.055556in}{-0.055556in}}%
\pgfpathlineto{\pgfqpoint{0.055556in}{0.055556in}}%
\pgfpathlineto{\pgfqpoint{-0.055556in}{0.055556in}}%
\pgfpathlineto{\pgfqpoint{-0.055556in}{-0.055556in}}%
\pgfpathclose%
\pgfusepath{stroke,fill}%
}%
\begin{pgfscope}%
\pgfsys@transformshift{1.202732in}{1.263555in}%
\pgfsys@useobject{currentmarker}{}%
\end{pgfscope}%
\end{pgfscope}%
\begin{pgfscope}%
\definecolor{textcolor}{rgb}{0.000000,0.000000,0.000000}%
\pgfsetstrokecolor{textcolor}%
\pgfsetfillcolor{textcolor}%
\pgftext[x=1.702732in,y=1.166333in,left,base]{\color{textcolor}{\rmfamily\fontsize{20.000000}{24.000000}\selectfont\catcode`\^=\active\def^{\ifmmode\sp\else\^{}\fi}\catcode`\%=\active\def
\end{pgfscope}%
\end{pgfpicture}%
\makeatother%
\endgroup%

%% file: images/thermal_ratio_per_modality/modality_results_RTA_5.pgf
\begingroup%
\makeatletter%
\begin{pgfpicture}%
\pgfpathrectangle{\pgfpointorigin}{\pgfqpoint{7.450000in}{5.450000in}}%
\pgfusepath{use as bounding box, clip}%
\begin{pgfscope}%
\pgfsetbuttcap%
\pgfsetmiterjoin%
\definecolor{currentfill}{rgb}{1.000000,1.000000,1.000000}%
\pgfsetfillcolor{currentfill}%
\pgfsetlinewidth{0.000000pt}%
\definecolor{currentstroke}{rgb}{1.000000,1.000000,1.000000}%
\pgfsetstrokecolor{currentstroke}%
\pgfsetdash{}{0pt}%
\pgfpathmoveto{\pgfqpoint{0.000000in}{0.000000in}}%
\pgfpathlineto{\pgfqpoint{7.450000in}{0.000000in}}%
\pgfpathlineto{\pgfqpoint{7.450000in}{5.450000in}}%
\pgfpathlineto{\pgfqpoint{0.000000in}{5.450000in}}%
\pgfpathlineto{\pgfqpoint{0.000000in}{0.000000in}}%
\pgfpathclose%
\pgfusepath{fill}%
\end{pgfscope}%
\begin{pgfscope}%
\pgfsetbuttcap%
\pgfsetmiterjoin%
\definecolor{currentfill}{rgb}{1.000000,1.000000,1.000000}%
\pgfsetfillcolor{currentfill}%
\pgfsetlinewidth{0.000000pt}%
\definecolor{currentstroke}{rgb}{0.000000,0.000000,0.000000}%
\pgfsetstrokecolor{currentstroke}%
\pgfsetstrokeopacity{0.000000}%
\pgfsetdash{}{0pt}%
\pgfpathmoveto{\pgfqpoint{0.674954in}{0.862305in}}%
\pgfpathlineto{\pgfqpoint{7.147414in}{0.862305in}}%
\pgfpathlineto{\pgfqpoint{7.147414in}{5.231386in}}%
\pgfpathlineto{\pgfqpoint{0.674954in}{5.231386in}}%
\pgfpathlineto{\pgfqpoint{0.674954in}{0.862305in}}%
\pgfpathclose%
\pgfusepath{fill}%
\end{pgfscope}%
\begin{pgfscope}%
\pgfpathrectangle{\pgfqpoint{0.674954in}{0.862305in}}{\pgfqpoint{6.472460in}{4.369081in}}%
\pgfusepath{clip}%
\pgfsetbuttcap%
\pgfsetroundjoin%
\pgfsetlinewidth{2.007500pt}%
\definecolor{currentstroke}{rgb}{0.501961,0.501961,0.501961}%
\pgfsetstrokecolor{currentstroke}%
\pgfsetstrokeopacity{0.300000}%
\pgfsetdash{{7.400000pt}{3.200000pt}}{0.000000pt}%
\pgfpathmoveto{\pgfqpoint{0.674954in}{0.862305in}}%
\pgfpathlineto{\pgfqpoint{0.674954in}{5.231386in}}%
\pgfusepath{stroke}%
\end{pgfscope}%
\begin{pgfscope}%
\pgfsetbuttcap%
\pgfsetroundjoin%
\definecolor{currentfill}{rgb}{0.000000,0.000000,0.000000}%
\pgfsetfillcolor{currentfill}%
\pgfsetlinewidth{0.803000pt}%
\definecolor{currentstroke}{rgb}{0.000000,0.000000,0.000000}%
\pgfsetstrokecolor{currentstroke}%
\pgfsetdash{}{0pt}%
\pgfsys@defobject{currentmarker}{\pgfqpoint{0.000000in}{-0.048611in}}{\pgfqpoint{0.000000in}{0.000000in}}{%
\pgfpathmoveto{\pgfqpoint{0.000000in}{0.000000in}}%
\pgfpathlineto{\pgfqpoint{0.000000in}{-0.048611in}}%
\pgfusepath{stroke,fill}%
}%
\begin{pgfscope}%
\pgfsys@transformshift{0.674954in}{0.862305in}%
\pgfsys@useobject{currentmarker}{}%
\end{pgfscope}%
\end{pgfscope}%
\begin{pgfscope}%
\definecolor{textcolor}{rgb}{0.000000,0.000000,0.000000}%
\pgfsetstrokecolor{textcolor}%
\pgfsetfillcolor{textcolor}%
\pgftext[x=0.674954in,y=0.765082in,,top]{\color{textcolor}{\rmfamily\fontsize{25.000000}{30.000000}\selectfont\catcode`\^=\active\def^{\ifmmode\sp\else\^{}\fi}\catcode`\%=\active\def
\end{pgfscope}%
\begin{pgfscope}%
\pgfpathrectangle{\pgfqpoint{0.674954in}{0.862305in}}{\pgfqpoint{6.472460in}{4.369081in}}%
\pgfusepath{clip}%
\pgfsetbuttcap%
\pgfsetroundjoin%
\pgfsetlinewidth{2.007500pt}%
\definecolor{currentstroke}{rgb}{0.501961,0.501961,0.501961}%
\pgfsetstrokecolor{currentstroke}%
\pgfsetstrokeopacity{0.300000}%
\pgfsetdash{{7.400000pt}{3.200000pt}}{0.000000pt}%
\pgfpathmoveto{\pgfqpoint{1.969446in}{0.862305in}}%
\pgfpathlineto{\pgfqpoint{1.969446in}{5.231386in}}%
\pgfusepath{stroke}%
\end{pgfscope}%
\begin{pgfscope}%
\pgfsetbuttcap%
\pgfsetroundjoin%
\definecolor{currentfill}{rgb}{0.000000,0.000000,0.000000}%
\pgfsetfillcolor{currentfill}%
\pgfsetlinewidth{0.803000pt}%
\definecolor{currentstroke}{rgb}{0.000000,0.000000,0.000000}%
\pgfsetstrokecolor{currentstroke}%
\pgfsetdash{}{0pt}%
\pgfsys@defobject{currentmarker}{\pgfqpoint{0.000000in}{-0.048611in}}{\pgfqpoint{0.000000in}{0.000000in}}{%
\pgfpathmoveto{\pgfqpoint{0.000000in}{0.000000in}}%
\pgfpathlineto{\pgfqpoint{0.000000in}{-0.048611in}}%
\pgfusepath{stroke,fill}%
}%
\begin{pgfscope}%
\pgfsys@transformshift{1.969446in}{0.862305in}%
\pgfsys@useobject{currentmarker}{}%
\end{pgfscope}%
\end{pgfscope}%
\begin{pgfscope}%
\definecolor{textcolor}{rgb}{0.000000,0.000000,0.000000}%
\pgfsetstrokecolor{textcolor}%
\pgfsetfillcolor{textcolor}%
\pgftext[x=1.969446in,y=0.765082in,,top]{\color{textcolor}{\rmfamily\fontsize{25.000000}{30.000000}\selectfont\catcode`\^=\active\def^{\ifmmode\sp\else\^{}\fi}\catcode`\%=\active\def
\end{pgfscope}%
\begin{pgfscope}%
\pgfpathrectangle{\pgfqpoint{0.674954in}{0.862305in}}{\pgfqpoint{6.472460in}{4.369081in}}%
\pgfusepath{clip}%
\pgfsetbuttcap%
\pgfsetroundjoin%
\pgfsetlinewidth{2.007500pt}%
\definecolor{currentstroke}{rgb}{0.501961,0.501961,0.501961}%
\pgfsetstrokecolor{currentstroke}%
\pgfsetstrokeopacity{0.300000}%
\pgfsetdash{{7.400000pt}{3.200000pt}}{0.000000pt}%
\pgfpathmoveto{\pgfqpoint{3.263938in}{0.862305in}}%
\pgfpathlineto{\pgfqpoint{3.263938in}{5.231386in}}%
\pgfusepath{stroke}%
\end{pgfscope}%
\begin{pgfscope}%
\pgfsetbuttcap%
\pgfsetroundjoin%
\definecolor{currentfill}{rgb}{0.000000,0.000000,0.000000}%
\pgfsetfillcolor{currentfill}%
\pgfsetlinewidth{0.803000pt}%
\definecolor{currentstroke}{rgb}{0.000000,0.000000,0.000000}%
\pgfsetstrokecolor{currentstroke}%
\pgfsetdash{}{0pt}%
\pgfsys@defobject{currentmarker}{\pgfqpoint{0.000000in}{-0.048611in}}{\pgfqpoint{0.000000in}{0.000000in}}{%
\pgfpathmoveto{\pgfqpoint{0.000000in}{0.000000in}}%
\pgfpathlineto{\pgfqpoint{0.000000in}{-0.048611in}}%
\pgfusepath{stroke,fill}%
}%
\begin{pgfscope}%
\pgfsys@transformshift{3.263938in}{0.862305in}%
\pgfsys@useobject{currentmarker}{}%
\end{pgfscope}%
\end{pgfscope}%
\begin{pgfscope}%
\definecolor{textcolor}{rgb}{0.000000,0.000000,0.000000}%
\pgfsetstrokecolor{textcolor}%
\pgfsetfillcolor{textcolor}%
\pgftext[x=3.263938in,y=0.765082in,,top]{\color{textcolor}{\rmfamily\fontsize{25.000000}{30.000000}\selectfont\catcode`\^=\active\def^{\ifmmode\sp\else\^{}\fi}\catcode`\%=\active\def
\end{pgfscope}%
\begin{pgfscope}%
\pgfpathrectangle{\pgfqpoint{0.674954in}{0.862305in}}{\pgfqpoint{6.472460in}{4.369081in}}%
\pgfusepath{clip}%
\pgfsetbuttcap%
\pgfsetroundjoin%
\pgfsetlinewidth{2.007500pt}%
\definecolor{currentstroke}{rgb}{0.501961,0.501961,0.501961}%
\pgfsetstrokecolor{currentstroke}%
\pgfsetstrokeopacity{0.300000}%
\pgfsetdash{{7.400000pt}{3.200000pt}}{0.000000pt}%
\pgfpathmoveto{\pgfqpoint{4.558430in}{0.862305in}}%
\pgfpathlineto{\pgfqpoint{4.558430in}{5.231386in}}%
\pgfusepath{stroke}%
\end{pgfscope}%
\begin{pgfscope}%
\pgfsetbuttcap%
\pgfsetroundjoin%
\definecolor{currentfill}{rgb}{0.000000,0.000000,0.000000}%
\pgfsetfillcolor{currentfill}%
\pgfsetlinewidth{0.803000pt}%
\definecolor{currentstroke}{rgb}{0.000000,0.000000,0.000000}%
\pgfsetstrokecolor{currentstroke}%
\pgfsetdash{}{0pt}%
\pgfsys@defobject{currentmarker}{\pgfqpoint{0.000000in}{-0.048611in}}{\pgfqpoint{0.000000in}{0.000000in}}{%
\pgfpathmoveto{\pgfqpoint{0.000000in}{0.000000in}}%
\pgfpathlineto{\pgfqpoint{0.000000in}{-0.048611in}}%
\pgfusepath{stroke,fill}%
}%
\begin{pgfscope}%
\pgfsys@transformshift{4.558430in}{0.862305in}%
\pgfsys@useobject{currentmarker}{}%
\end{pgfscope}%
\end{pgfscope}%
\begin{pgfscope}%
\definecolor{textcolor}{rgb}{0.000000,0.000000,0.000000}%
\pgfsetstrokecolor{textcolor}%
\pgfsetfillcolor{textcolor}%
\pgftext[x=4.558430in,y=0.765082in,,top]{\color{textcolor}{\rmfamily\fontsize{25.000000}{30.000000}\selectfont\catcode`\^=\active\def^{\ifmmode\sp\else\^{}\fi}\catcode`\%=\active\def
\end{pgfscope}%
\begin{pgfscope}%
\pgfpathrectangle{\pgfqpoint{0.674954in}{0.862305in}}{\pgfqpoint{6.472460in}{4.369081in}}%
\pgfusepath{clip}%
\pgfsetbuttcap%
\pgfsetroundjoin%
\pgfsetlinewidth{2.007500pt}%
\definecolor{currentstroke}{rgb}{0.501961,0.501961,0.501961}%
\pgfsetstrokecolor{currentstroke}%
\pgfsetstrokeopacity{0.300000}%
\pgfsetdash{{7.400000pt}{3.200000pt}}{0.000000pt}%
\pgfpathmoveto{\pgfqpoint{5.852922in}{0.862305in}}%
\pgfpathlineto{\pgfqpoint{5.852922in}{5.231386in}}%
\pgfusepath{stroke}%
\end{pgfscope}%
\begin{pgfscope}%
\pgfsetbuttcap%
\pgfsetroundjoin%
\definecolor{currentfill}{rgb}{0.000000,0.000000,0.000000}%
\pgfsetfillcolor{currentfill}%
\pgfsetlinewidth{0.803000pt}%
\definecolor{currentstroke}{rgb}{0.000000,0.000000,0.000000}%
\pgfsetstrokecolor{currentstroke}%
\pgfsetdash{}{0pt}%
\pgfsys@defobject{currentmarker}{\pgfqpoint{0.000000in}{-0.048611in}}{\pgfqpoint{0.000000in}{0.000000in}}{%
\pgfpathmoveto{\pgfqpoint{0.000000in}{0.000000in}}%
\pgfpathlineto{\pgfqpoint{0.000000in}{-0.048611in}}%
\pgfusepath{stroke,fill}%
}%
\begin{pgfscope}%
\pgfsys@transformshift{5.852922in}{0.862305in}%
\pgfsys@useobject{currentmarker}{}%
\end{pgfscope}%
\end{pgfscope}%
\begin{pgfscope}%
\definecolor{textcolor}{rgb}{0.000000,0.000000,0.000000}%
\pgfsetstrokecolor{textcolor}%
\pgfsetfillcolor{textcolor}%
\pgftext[x=5.852922in,y=0.765082in,,top]{\color{textcolor}{\rmfamily\fontsize{25.000000}{30.000000}\selectfont\catcode`\^=\active\def^{\ifmmode\sp\else\^{}\fi}\catcode`\%=\active\def
\end{pgfscope}%
\begin{pgfscope}%
\pgfpathrectangle{\pgfqpoint{0.674954in}{0.862305in}}{\pgfqpoint{6.472460in}{4.369081in}}%
\pgfusepath{clip}%
\pgfsetbuttcap%
\pgfsetroundjoin%
\pgfsetlinewidth{2.007500pt}%
\definecolor{currentstroke}{rgb}{0.501961,0.501961,0.501961}%
\pgfsetstrokecolor{currentstroke}%
\pgfsetstrokeopacity{0.300000}%
\pgfsetdash{{7.400000pt}{3.200000pt}}{0.000000pt}%
\pgfpathmoveto{\pgfqpoint{7.147414in}{0.862305in}}%
\pgfpathlineto{\pgfqpoint{7.147414in}{5.231386in}}%
\pgfusepath{stroke}%
\end{pgfscope}%
\begin{pgfscope}%
\pgfsetbuttcap%
\pgfsetroundjoin%
\definecolor{currentfill}{rgb}{0.000000,0.000000,0.000000}%
\pgfsetfillcolor{currentfill}%
\pgfsetlinewidth{0.803000pt}%
\definecolor{currentstroke}{rgb}{0.000000,0.000000,0.000000}%
\pgfsetstrokecolor{currentstroke}%
\pgfsetdash{}{0pt}%
\pgfsys@defobject{currentmarker}{\pgfqpoint{0.000000in}{-0.048611in}}{\pgfqpoint{0.000000in}{0.000000in}}{%
\pgfpathmoveto{\pgfqpoint{0.000000in}{0.000000in}}%
\pgfpathlineto{\pgfqpoint{0.000000in}{-0.048611in}}%
\pgfusepath{stroke,fill}%
}%
\begin{pgfscope}%
\pgfsys@transformshift{7.147414in}{0.862305in}%
\pgfsys@useobject{currentmarker}{}%
\end{pgfscope}%
\end{pgfscope}%
\begin{pgfscope}%
\definecolor{textcolor}{rgb}{0.000000,0.000000,0.000000}%
\pgfsetstrokecolor{textcolor}%
\pgfsetfillcolor{textcolor}%
\pgftext[x=7.147414in,y=0.765082in,,top]{\color{textcolor}{\rmfamily\fontsize{25.000000}{30.000000}\selectfont\catcode`\^=\active\def^{\ifmmode\sp\else\^{}\fi}\catcode`\%=\active\def
\end{pgfscope}%
\begin{pgfscope}%
\definecolor{textcolor}{rgb}{0.000000,0.000000,0.000000}%
\pgfsetstrokecolor{textcolor}%
\pgfsetfillcolor{textcolor}%
\pgftext[x=3.911184in,y=0.404763in,,top]{\color{textcolor}{\rmfamily\fontsize{25.000000}{30.000000}\selectfont\catcode`\^=\active\def^{\ifmmode\sp\else\^{}\fi}\catcode`\%=\active\def
\end{pgfscope}%
\begin{pgfscope}%
\pgfpathrectangle{\pgfqpoint{0.674954in}{0.862305in}}{\pgfqpoint{6.472460in}{4.369081in}}%
\pgfusepath{clip}%
\pgfsetbuttcap%
\pgfsetroundjoin%
\pgfsetlinewidth{2.007500pt}%
\definecolor{currentstroke}{rgb}{0.501961,0.501961,0.501961}%
\pgfsetstrokecolor{currentstroke}%
\pgfsetstrokeopacity{0.300000}%
\pgfsetdash{{7.400000pt}{3.200000pt}}{0.000000pt}%
\pgfpathmoveto{\pgfqpoint{0.674954in}{1.954575in}}%
\pgfpathlineto{\pgfqpoint{7.147414in}{1.954575in}}%
\pgfusepath{stroke}%
\end{pgfscope}%
\begin{pgfscope}%
\pgfsetbuttcap%
\pgfsetroundjoin%
\definecolor{currentfill}{rgb}{0.000000,0.000000,0.000000}%
\pgfsetfillcolor{currentfill}%
\pgfsetlinewidth{0.803000pt}%
\definecolor{currentstroke}{rgb}{0.000000,0.000000,0.000000}%
\pgfsetstrokecolor{currentstroke}%
\pgfsetdash{}{0pt}%
\pgfsys@defobject{currentmarker}{\pgfqpoint{-0.048611in}{0.000000in}}{\pgfqpoint{-0.000000in}{0.000000in}}{%
\pgfpathmoveto{\pgfqpoint{-0.000000in}{0.000000in}}%
\pgfpathlineto{\pgfqpoint{-0.048611in}{0.000000in}}%
\pgfusepath{stroke,fill}%
}%
\begin{pgfscope}%
\pgfsys@transformshift{0.674954in}{1.954575in}%
\pgfsys@useobject{currentmarker}{}%
\end{pgfscope}%
\end{pgfscope}%
\begin{pgfscope}%
\definecolor{textcolor}{rgb}{0.000000,0.000000,0.000000}%
\pgfsetstrokecolor{textcolor}%
\pgfsetfillcolor{textcolor}%
\pgftext[x=0.259244in, y=1.835961in, left, base]{\color{textcolor}{\rmfamily\fontsize{25.000000}{30.000000}\selectfont\catcode`\^=\active\def^{\ifmmode\sp\else\^{}\fi}\catcode`\%=\active\def
\end{pgfscope}%
\begin{pgfscope}%
\pgfpathrectangle{\pgfqpoint{0.674954in}{0.862305in}}{\pgfqpoint{6.472460in}{4.369081in}}%
\pgfusepath{clip}%
\pgfsetbuttcap%
\pgfsetroundjoin%
\pgfsetlinewidth{2.007500pt}%
\definecolor{currentstroke}{rgb}{0.501961,0.501961,0.501961}%
\pgfsetstrokecolor{currentstroke}%
\pgfsetstrokeopacity{0.300000}%
\pgfsetdash{{7.400000pt}{3.200000pt}}{0.000000pt}%
\pgfpathmoveto{\pgfqpoint{0.674954in}{3.046845in}}%
\pgfpathlineto{\pgfqpoint{7.147414in}{3.046845in}}%
\pgfusepath{stroke}%
\end{pgfscope}%
\begin{pgfscope}%
\pgfsetbuttcap%
\pgfsetroundjoin%
\definecolor{currentfill}{rgb}{0.000000,0.000000,0.000000}%
\pgfsetfillcolor{currentfill}%
\pgfsetlinewidth{0.803000pt}%
\definecolor{currentstroke}{rgb}{0.000000,0.000000,0.000000}%
\pgfsetstrokecolor{currentstroke}%
\pgfsetdash{}{0pt}%
\pgfsys@defobject{currentmarker}{\pgfqpoint{-0.048611in}{0.000000in}}{\pgfqpoint{-0.000000in}{0.000000in}}{%
\pgfpathmoveto{\pgfqpoint{-0.000000in}{0.000000in}}%
\pgfpathlineto{\pgfqpoint{-0.048611in}{0.000000in}}%
\pgfusepath{stroke,fill}%
}%
\begin{pgfscope}%
\pgfsys@transformshift{0.674954in}{3.046845in}%
\pgfsys@useobject{currentmarker}{}%
\end{pgfscope}%
\end{pgfscope}%
\begin{pgfscope}%
\definecolor{textcolor}{rgb}{0.000000,0.000000,0.000000}%
\pgfsetstrokecolor{textcolor}%
\pgfsetfillcolor{textcolor}%
\pgftext[x=0.259244in, y=2.928231in, left, base]{\color{textcolor}{\rmfamily\fontsize{25.000000}{30.000000}\selectfont\catcode`\^=\active\def^{\ifmmode\sp\else\^{}\fi}\catcode`\%=\active\def
\end{pgfscope}%
\begin{pgfscope}%
\pgfpathrectangle{\pgfqpoint{0.674954in}{0.862305in}}{\pgfqpoint{6.472460in}{4.369081in}}%
\pgfusepath{clip}%
\pgfsetbuttcap%
\pgfsetroundjoin%
\pgfsetlinewidth{2.007500pt}%
\definecolor{currentstroke}{rgb}{0.501961,0.501961,0.501961}%
\pgfsetstrokecolor{currentstroke}%
\pgfsetstrokeopacity{0.300000}%
\pgfsetdash{{7.400000pt}{3.200000pt}}{0.000000pt}%
\pgfpathmoveto{\pgfqpoint{0.674954in}{4.139116in}}%
\pgfpathlineto{\pgfqpoint{7.147414in}{4.139116in}}%
\pgfusepath{stroke}%
\end{pgfscope}%
\begin{pgfscope}%
\pgfsetbuttcap%
\pgfsetroundjoin%
\definecolor{currentfill}{rgb}{0.000000,0.000000,0.000000}%
\pgfsetfillcolor{currentfill}%
\pgfsetlinewidth{0.803000pt}%
\definecolor{currentstroke}{rgb}{0.000000,0.000000,0.000000}%
\pgfsetstrokecolor{currentstroke}%
\pgfsetdash{}{0pt}%
\pgfsys@defobject{currentmarker}{\pgfqpoint{-0.048611in}{0.000000in}}{\pgfqpoint{-0.000000in}{0.000000in}}{%
\pgfpathmoveto{\pgfqpoint{-0.000000in}{0.000000in}}%
\pgfpathlineto{\pgfqpoint{-0.048611in}{0.000000in}}%
\pgfusepath{stroke,fill}%
}%
\begin{pgfscope}%
\pgfsys@transformshift{0.674954in}{4.139116in}%
\pgfsys@useobject{currentmarker}{}%
\end{pgfscope}%
\end{pgfscope}%
\begin{pgfscope}%
\definecolor{textcolor}{rgb}{0.000000,0.000000,0.000000}%
\pgfsetstrokecolor{textcolor}%
\pgfsetfillcolor{textcolor}%
\pgftext[x=0.259244in, y=4.020501in, left, base]{\color{textcolor}{\rmfamily\fontsize{25.000000}{30.000000}\selectfont\catcode`\^=\active\def^{\ifmmode\sp\else\^{}\fi}\catcode`\%=\active\def
\end{pgfscope}%
\begin{pgfscope}%
\pgfpathrectangle{\pgfqpoint{0.674954in}{0.862305in}}{\pgfqpoint{6.472460in}{4.369081in}}%
\pgfusepath{clip}%
\pgfsetbuttcap%
\pgfsetroundjoin%
\pgfsetlinewidth{2.007500pt}%
\definecolor{currentstroke}{rgb}{0.501961,0.501961,0.501961}%
\pgfsetstrokecolor{currentstroke}%
\pgfsetstrokeopacity{0.300000}%
\pgfsetdash{{7.400000pt}{3.200000pt}}{0.000000pt}%
\pgfpathmoveto{\pgfqpoint{0.674954in}{5.231386in}}%
\pgfpathlineto{\pgfqpoint{7.147414in}{5.231386in}}%
\pgfusepath{stroke}%
\end{pgfscope}%
\begin{pgfscope}%
\pgfsetbuttcap%
\pgfsetroundjoin%
\definecolor{currentfill}{rgb}{0.000000,0.000000,0.000000}%
\pgfsetfillcolor{currentfill}%
\pgfsetlinewidth{0.803000pt}%
\definecolor{currentstroke}{rgb}{0.000000,0.000000,0.000000}%
\pgfsetstrokecolor{currentstroke}%
\pgfsetdash{}{0pt}%
\pgfsys@defobject{currentmarker}{\pgfqpoint{-0.048611in}{0.000000in}}{\pgfqpoint{-0.000000in}{0.000000in}}{%
\pgfpathmoveto{\pgfqpoint{-0.000000in}{0.000000in}}%
\pgfpathlineto{\pgfqpoint{-0.048611in}{0.000000in}}%
\pgfusepath{stroke,fill}%
}%
\begin{pgfscope}%
\pgfsys@transformshift{0.674954in}{5.231386in}%
\pgfsys@useobject{currentmarker}{}%
\end{pgfscope}%
\end{pgfscope}%
\begin{pgfscope}%
\definecolor{textcolor}{rgb}{0.000000,0.000000,0.000000}%
\pgfsetstrokecolor{textcolor}%
\pgfsetfillcolor{textcolor}%
\pgftext[x=0.100000in, y=5.112772in, left, base]{\color{textcolor}{\rmfamily\fontsize{25.000000}{30.000000}\selectfont\catcode`\^=\active\def^{\ifmmode\sp\else\^{}\fi}\catcode`\%=\active\def
\end{pgfscope}%
\begin{pgfscope}%
\pgfpathrectangle{\pgfqpoint{0.674954in}{0.862305in}}{\pgfqpoint{6.472460in}{4.369081in}}%
\pgfusepath{clip}%
\pgfsetrectcap%
\pgfsetroundjoin%
\pgfsetlinewidth{2.509375pt}%
\definecolor{currentstroke}{rgb}{0.050980,0.415686,0.509804}%
\pgfsetstrokecolor{currentstroke}%
\pgfsetdash{}{0pt}%
\pgfpathmoveto{\pgfqpoint{0.674954in}{4.496064in}}%
\pgfpathlineto{\pgfqpoint{0.998577in}{4.475566in}}%
\pgfpathlineto{\pgfqpoint{2.293069in}{4.342535in}}%
\pgfpathlineto{\pgfqpoint{3.911184in}{4.147718in}}%
\pgfpathlineto{\pgfqpoint{5.529299in}{3.921233in}}%
\pgfpathlineto{\pgfqpoint{6.823791in}{3.320394in}}%
\pgfusepath{stroke}%
\end{pgfscope}%
\begin{pgfscope}%
\pgfpathrectangle{\pgfqpoint{0.674954in}{0.862305in}}{\pgfqpoint{6.472460in}{4.369081in}}%
\pgfusepath{clip}%
\pgfsetbuttcap%
\pgfsetroundjoin%
\definecolor{currentfill}{rgb}{0.050980,0.415686,0.509804}%
\pgfsetfillcolor{currentfill}%
\pgfsetlinewidth{1.003750pt}%
\definecolor{currentstroke}{rgb}{0.050980,0.415686,0.509804}%
\pgfsetstrokecolor{currentstroke}%
\pgfsetdash{}{0pt}%
\pgfsys@defobject{currentmarker}{\pgfqpoint{-0.055556in}{-0.055556in}}{\pgfqpoint{0.055556in}{0.055556in}}{%
\pgfpathmoveto{\pgfqpoint{0.000000in}{-0.055556in}}%
\pgfpathcurveto{\pgfqpoint{0.014734in}{-0.055556in}}{\pgfqpoint{0.028866in}{-0.049702in}}{\pgfqpoint{0.039284in}{-0.039284in}}%
\pgfpathcurveto{\pgfqpoint{0.049702in}{-0.028866in}}{\pgfqpoint{0.055556in}{-0.014734in}}{\pgfqpoint{0.055556in}{0.000000in}}%
\pgfpathcurveto{\pgfqpoint{0.055556in}{0.014734in}}{\pgfqpoint{0.049702in}{0.028866in}}{\pgfqpoint{0.039284in}{0.039284in}}%
\pgfpathcurveto{\pgfqpoint{0.028866in}{0.049702in}}{\pgfqpoint{0.014734in}{0.055556in}}{\pgfqpoint{0.000000in}{0.055556in}}%
\pgfpathcurveto{\pgfqpoint{-0.014734in}{0.055556in}}{\pgfqpoint{-0.028866in}{0.049702in}}{\pgfqpoint{-0.039284in}{0.039284in}}%
\pgfpathcurveto{\pgfqpoint{-0.049702in}{0.028866in}}{\pgfqpoint{-0.055556in}{0.014734in}}{\pgfqpoint{-0.055556in}{0.000000in}}%
\pgfpathcurveto{\pgfqpoint{-0.055556in}{-0.014734in}}{\pgfqpoint{-0.049702in}{-0.028866in}}{\pgfqpoint{-0.039284in}{-0.039284in}}%
\pgfpathcurveto{\pgfqpoint{-0.028866in}{-0.049702in}}{\pgfqpoint{-0.014734in}{-0.055556in}}{\pgfqpoint{0.000000in}{-0.055556in}}%
\pgfpathlineto{\pgfqpoint{0.000000in}{-0.055556in}}%
\pgfpathclose%
\pgfusepath{stroke,fill}%
}%
\begin{pgfscope}%
\pgfsys@transformshift{0.674954in}{4.496064in}%
\pgfsys@useobject{currentmarker}{}%
\end{pgfscope}%
\begin{pgfscope}%
\pgfsys@transformshift{0.998577in}{4.475566in}%
\pgfsys@useobject{currentmarker}{}%
\end{pgfscope}%
\begin{pgfscope}%
\pgfsys@transformshift{2.293069in}{4.342535in}%
\pgfsys@useobject{currentmarker}{}%
\end{pgfscope}%
\begin{pgfscope}%
\pgfsys@transformshift{3.911184in}{4.147718in}%
\pgfsys@useobject{currentmarker}{}%
\end{pgfscope}%
\begin{pgfscope}%
\pgfsys@transformshift{5.529299in}{3.921233in}%
\pgfsys@useobject{currentmarker}{}%
\end{pgfscope}%
\begin{pgfscope}%
\pgfsys@transformshift{6.823791in}{3.320394in}%
\pgfsys@useobject{currentmarker}{}%
\end{pgfscope}%
\end{pgfscope}%
\begin{pgfscope}%
\pgfpathrectangle{\pgfqpoint{0.674954in}{0.862305in}}{\pgfqpoint{6.472460in}{4.369081in}}%
\pgfusepath{clip}%
\pgfsetbuttcap%
\pgfsetroundjoin%
\pgfsetlinewidth{2.509375pt}%
\definecolor{currentstroke}{rgb}{0.960784,0.462745,0.000000}%
\pgfsetstrokecolor{currentstroke}%
\pgfsetdash{{9.250000pt}{4.000000pt}}{0.000000pt}%
\pgfpathmoveto{\pgfqpoint{0.998577in}{3.225556in}}%
\pgfpathlineto{\pgfqpoint{2.293069in}{3.429780in}}%
\pgfpathlineto{\pgfqpoint{3.911184in}{3.505238in}}%
\pgfpathlineto{\pgfqpoint{5.529299in}{3.486485in}}%
\pgfpathlineto{\pgfqpoint{6.823791in}{3.592934in}}%
\pgfpathlineto{\pgfqpoint{7.147414in}{3.645363in}}%
\pgfusepath{stroke}%
\end{pgfscope}%
\begin{pgfscope}%
\pgfpathrectangle{\pgfqpoint{0.674954in}{0.862305in}}{\pgfqpoint{6.472460in}{4.369081in}}%
\pgfusepath{clip}%
\pgfsetbuttcap%
\pgfsetmiterjoin%
\definecolor{currentfill}{rgb}{0.960784,0.462745,0.000000}%
\pgfsetfillcolor{currentfill}%
\pgfsetlinewidth{1.003750pt}%
\definecolor{currentstroke}{rgb}{0.960784,0.462745,0.000000}%
\pgfsetstrokecolor{currentstroke}%
\pgfsetdash{}{0pt}%
\pgfsys@defobject{currentmarker}{\pgfqpoint{-0.055556in}{-0.055556in}}{\pgfqpoint{0.055556in}{0.055556in}}{%
\pgfpathmoveto{\pgfqpoint{-0.055556in}{-0.055556in}}%
\pgfpathlineto{\pgfqpoint{0.055556in}{-0.055556in}}%
\pgfpathlineto{\pgfqpoint{0.055556in}{0.055556in}}%
\pgfpathlineto{\pgfqpoint{-0.055556in}{0.055556in}}%
\pgfpathlineto{\pgfqpoint{-0.055556in}{-0.055556in}}%
\pgfpathclose%
\pgfusepath{stroke,fill}%
}%
\begin{pgfscope}%
\pgfsys@transformshift{0.998577in}{3.225556in}%
\pgfsys@useobject{currentmarker}{}%
\end{pgfscope}%
\begin{pgfscope}%
\pgfsys@transformshift{2.293069in}{3.429780in}%
\pgfsys@useobject{currentmarker}{}%
\end{pgfscope}%
\begin{pgfscope}%
\pgfsys@transformshift{3.911184in}{3.505238in}%
\pgfsys@useobject{currentmarker}{}%
\end{pgfscope}%
\begin{pgfscope}%
\pgfsys@transformshift{5.529299in}{3.486485in}%
\pgfsys@useobject{currentmarker}{}%
\end{pgfscope}%
\begin{pgfscope}%
\pgfsys@transformshift{6.823791in}{3.592934in}%
\pgfsys@useobject{currentmarker}{}%
\end{pgfscope}%
\begin{pgfscope}%
\pgfsys@transformshift{7.147414in}{3.645363in}%
\pgfsys@useobject{currentmarker}{}%
\end{pgfscope}%
\end{pgfscope}%
\begin{pgfscope}%
\pgfsetrectcap%
\pgfsetmiterjoin%
\pgfsetlinewidth{2.007500pt}%
\definecolor{currentstroke}{rgb}{0.000000,0.000000,0.000000}%
\pgfsetstrokecolor{currentstroke}%
\pgfsetdash{}{0pt}%
\pgfpathmoveto{\pgfqpoint{0.674954in}{0.862305in}}%
\pgfpathlineto{\pgfqpoint{0.674954in}{5.231386in}}%
\pgfusepath{stroke}%
\end{pgfscope}%
\begin{pgfscope}%
\pgfsetrectcap%
\pgfsetmiterjoin%
\pgfsetlinewidth{2.007500pt}%
\definecolor{currentstroke}{rgb}{0.000000,0.000000,0.000000}%
\pgfsetstrokecolor{currentstroke}%
\pgfsetdash{}{0pt}%
\pgfpathmoveto{\pgfqpoint{0.674954in}{0.862305in}}%
\pgfpathlineto{\pgfqpoint{7.147414in}{0.862305in}}%
\pgfusepath{stroke}%
\end{pgfscope}%
\begin{pgfscope}%
\pgfsetbuttcap%
\pgfsetmiterjoin%
\definecolor{currentfill}{rgb}{1.000000,1.000000,1.000000}%
\pgfsetfillcolor{currentfill}%
\pgfsetlinewidth{1.003750pt}%
\definecolor{currentstroke}{rgb}{0.800000,0.800000,0.800000}%
\pgfsetstrokecolor{currentstroke}%
\pgfsetdash{}{0pt}%
\pgfpathmoveto{\pgfqpoint{0.869398in}{1.001194in}}%
\pgfpathlineto{\pgfqpoint{2.708109in}{1.001194in}}%
\pgfpathquadraticcurveto{\pgfqpoint{2.763664in}{1.001194in}}{\pgfqpoint{2.763664in}{1.056749in}}%
\pgfpathlineto{\pgfqpoint{2.763664in}{1.803694in}}%
\pgfpathquadraticcurveto{\pgfqpoint{2.763664in}{1.859250in}}{\pgfqpoint{2.708109in}{1.859250in}}%
\pgfpathlineto{\pgfqpoint{0.869398in}{1.859250in}}%
\pgfpathquadraticcurveto{\pgfqpoint{0.813843in}{1.859250in}}{\pgfqpoint{0.813843in}{1.803694in}}%
\pgfpathlineto{\pgfqpoint{0.813843in}{1.056749in}}%
\pgfpathquadraticcurveto{\pgfqpoint{0.813843in}{1.001194in}}{\pgfqpoint{0.869398in}{1.001194in}}%
\pgfpathlineto{\pgfqpoint{0.869398in}{1.001194in}}%
\pgfpathclose%
\pgfusepath{stroke,fill}%
\end{pgfscope}%
\begin{pgfscope}%
\pgfsetrectcap%
\pgfsetroundjoin%
\pgfsetlinewidth{2.509375pt}%
\definecolor{currentstroke}{rgb}{0.050980,0.415686,0.509804}%
\pgfsetstrokecolor{currentstroke}%
\pgfsetdash{}{0pt}%
\pgfpathmoveto{\pgfqpoint{0.924954in}{1.650917in}}%
\pgfpathlineto{\pgfqpoint{1.202732in}{1.650917in}}%
\pgfpathlineto{\pgfqpoint{1.480509in}{1.650917in}}%
\pgfusepath{stroke}%
\end{pgfscope}%
\begin{pgfscope}%
\pgfsetbuttcap%
\pgfsetroundjoin%
\definecolor{currentfill}{rgb}{0.050980,0.415686,0.509804}%
\pgfsetfillcolor{currentfill}%
\pgfsetlinewidth{1.003750pt}%
\definecolor{currentstroke}{rgb}{0.050980,0.415686,0.509804}%
\pgfsetstrokecolor{currentstroke}%
\pgfsetdash{}{0pt}%
\pgfsys@defobject{currentmarker}{\pgfqpoint{-0.055556in}{-0.055556in}}{\pgfqpoint{0.055556in}{0.055556in}}{%
\pgfpathmoveto{\pgfqpoint{0.000000in}{-0.055556in}}%
\pgfpathcurveto{\pgfqpoint{0.014734in}{-0.055556in}}{\pgfqpoint{0.028866in}{-0.049702in}}{\pgfqpoint{0.039284in}{-0.039284in}}%
\pgfpathcurveto{\pgfqpoint{0.049702in}{-0.028866in}}{\pgfqpoint{0.055556in}{-0.014734in}}{\pgfqpoint{0.055556in}{0.000000in}}%
\pgfpathcurveto{\pgfqpoint{0.055556in}{0.014734in}}{\pgfqpoint{0.049702in}{0.028866in}}{\pgfqpoint{0.039284in}{0.039284in}}%
\pgfpathcurveto{\pgfqpoint{0.028866in}{0.049702in}}{\pgfqpoint{0.014734in}{0.055556in}}{\pgfqpoint{0.000000in}{0.055556in}}%
\pgfpathcurveto{\pgfqpoint{-0.014734in}{0.055556in}}{\pgfqpoint{-0.028866in}{0.049702in}}{\pgfqpoint{-0.039284in}{0.039284in}}%
\pgfpathcurveto{\pgfqpoint{-0.049702in}{0.028866in}}{\pgfqpoint{-0.055556in}{0.014734in}}{\pgfqpoint{-0.055556in}{0.000000in}}%
\pgfpathcurveto{\pgfqpoint{-0.055556in}{-0.014734in}}{\pgfqpoint{-0.049702in}{-0.028866in}}{\pgfqpoint{-0.039284in}{-0.039284in}}%
\pgfpathcurveto{\pgfqpoint{-0.028866in}{-0.049702in}}{\pgfqpoint{-0.014734in}{-0.055556in}}{\pgfqpoint{0.000000in}{-0.055556in}}%
\pgfpathlineto{\pgfqpoint{0.000000in}{-0.055556in}}%
\pgfpathclose%
\pgfusepath{stroke,fill}%
}%
\begin{pgfscope}%
\pgfsys@transformshift{1.202732in}{1.650917in}%
\pgfsys@useobject{currentmarker}{}%
\end{pgfscope}%
\end{pgfscope}%
\begin{pgfscope}%
\definecolor{textcolor}{rgb}{0.000000,0.000000,0.000000}%
\pgfsetstrokecolor{textcolor}%
\pgfsetfillcolor{textcolor}%
\pgftext[x=1.702732in,y=1.553694in,left,base]{\color{textcolor}{\rmfamily\fontsize{20.000000}{24.000000}\selectfont\catcode`\^=\active\def^{\ifmmode\sp\else\^{}\fi}\catcode`\%=\active\def
\end{pgfscope}%
\begin{pgfscope}%
\pgfsetbuttcap%
\pgfsetroundjoin%
\pgfsetlinewidth{2.509375pt}%
\definecolor{currentstroke}{rgb}{0.960784,0.462745,0.000000}%
\pgfsetstrokecolor{currentstroke}%
\pgfsetdash{{9.250000pt}{4.000000pt}}{0.000000pt}%
\pgfpathmoveto{\pgfqpoint{0.924954in}{1.263555in}}%
\pgfpathlineto{\pgfqpoint{1.202732in}{1.263555in}}%
\pgfpathlineto{\pgfqpoint{1.480509in}{1.263555in}}%
\pgfusepath{stroke}%
\end{pgfscope}%
\begin{pgfscope}%
\pgfsetbuttcap%
\pgfsetmiterjoin%
\definecolor{currentfill}{rgb}{0.960784,0.462745,0.000000}%
\pgfsetfillcolor{currentfill}%
\pgfsetlinewidth{1.003750pt}%
\definecolor{currentstroke}{rgb}{0.960784,0.462745,0.000000}%
\pgfsetstrokecolor{currentstroke}%
\pgfsetdash{}{0pt}%
\pgfsys@defobject{currentmarker}{\pgfqpoint{-0.055556in}{-0.055556in}}{\pgfqpoint{0.055556in}{0.055556in}}{%
\pgfpathmoveto{\pgfqpoint{-0.055556in}{-0.055556in}}%
\pgfpathlineto{\pgfqpoint{0.055556in}{-0.055556in}}%
\pgfpathlineto{\pgfqpoint{0.055556in}{0.055556in}}%
\pgfpathlineto{\pgfqpoint{-0.055556in}{0.055556in}}%
\pgfpathlineto{\pgfqpoint{-0.055556in}{-0.055556in}}%
\pgfpathclose%
\pgfusepath{stroke,fill}%
}%
\begin{pgfscope}%
\pgfsys@transformshift{1.202732in}{1.263555in}%
\pgfsys@useobject{currentmarker}{}%
\end{pgfscope}%
\end{pgfscope}%
\begin{pgfscope}%
\definecolor{textcolor}{rgb}{0.000000,0.000000,0.000000}%
\pgfsetstrokecolor{textcolor}%
\pgfsetfillcolor{textcolor}%
\pgftext[x=1.702732in,y=1.166333in,left,base]{\color{textcolor}{\rmfamily\fontsize{20.000000}{24.000000}\selectfont\catcode`\^=\active\def^{\ifmmode\sp\else\^{}\fi}\catcode`\%=\active\def
\end{pgfscope}%
\end{pgfpicture}%
\makeatother%
\endgroup%

%% file: images/thermal_ratio_per_modality/modality_results_RTA_15.pgf
\begingroup%
\makeatletter%
\begin{pgfpicture}%
\pgfpathrectangle{\pgfpointorigin}{\pgfqpoint{7.450000in}{5.450000in}}%
\pgfusepath{use as bounding box, clip}%
\begin{pgfscope}%
\pgfsetbuttcap%
\pgfsetmiterjoin%
\definecolor{currentfill}{rgb}{1.000000,1.000000,1.000000}%
\pgfsetfillcolor{currentfill}%
\pgfsetlinewidth{0.000000pt}%
\definecolor{currentstroke}{rgb}{1.000000,1.000000,1.000000}%
\pgfsetstrokecolor{currentstroke}%
\pgfsetdash{}{0pt}%
\pgfpathmoveto{\pgfqpoint{0.000000in}{0.000000in}}%
\pgfpathlineto{\pgfqpoint{7.450000in}{0.000000in}}%
\pgfpathlineto{\pgfqpoint{7.450000in}{5.450000in}}%
\pgfpathlineto{\pgfqpoint{0.000000in}{5.450000in}}%
\pgfpathlineto{\pgfqpoint{0.000000in}{0.000000in}}%
\pgfpathclose%
\pgfusepath{fill}%
\end{pgfscope}%
\begin{pgfscope}%
\pgfsetbuttcap%
\pgfsetmiterjoin%
\definecolor{currentfill}{rgb}{1.000000,1.000000,1.000000}%
\pgfsetfillcolor{currentfill}%
\pgfsetlinewidth{0.000000pt}%
\definecolor{currentstroke}{rgb}{0.000000,0.000000,0.000000}%
\pgfsetstrokecolor{currentstroke}%
\pgfsetstrokeopacity{0.000000}%
\pgfsetdash{}{0pt}%
\pgfpathmoveto{\pgfqpoint{0.674954in}{0.862305in}}%
\pgfpathlineto{\pgfqpoint{7.147414in}{0.862305in}}%
\pgfpathlineto{\pgfqpoint{7.147414in}{5.231386in}}%
\pgfpathlineto{\pgfqpoint{0.674954in}{5.231386in}}%
\pgfpathlineto{\pgfqpoint{0.674954in}{0.862305in}}%
\pgfpathclose%
\pgfusepath{fill}%
\end{pgfscope}%
\begin{pgfscope}%
\pgfpathrectangle{\pgfqpoint{0.674954in}{0.862305in}}{\pgfqpoint{6.472460in}{4.369081in}}%
\pgfusepath{clip}%
\pgfsetbuttcap%
\pgfsetroundjoin%
\pgfsetlinewidth{2.007500pt}%
\definecolor{currentstroke}{rgb}{0.501961,0.501961,0.501961}%
\pgfsetstrokecolor{currentstroke}%
\pgfsetstrokeopacity{0.300000}%
\pgfsetdash{{7.400000pt}{3.200000pt}}{0.000000pt}%
\pgfpathmoveto{\pgfqpoint{0.674954in}{0.862305in}}%
\pgfpathlineto{\pgfqpoint{0.674954in}{5.231386in}}%
\pgfusepath{stroke}%
\end{pgfscope}%
\begin{pgfscope}%
\pgfsetbuttcap%
\pgfsetroundjoin%
\definecolor{currentfill}{rgb}{0.000000,0.000000,0.000000}%
\pgfsetfillcolor{currentfill}%
\pgfsetlinewidth{0.803000pt}%
\definecolor{currentstroke}{rgb}{0.000000,0.000000,0.000000}%
\pgfsetstrokecolor{currentstroke}%
\pgfsetdash{}{0pt}%
\pgfsys@defobject{currentmarker}{\pgfqpoint{0.000000in}{-0.048611in}}{\pgfqpoint{0.000000in}{0.000000in}}{%
\pgfpathmoveto{\pgfqpoint{0.000000in}{0.000000in}}%
\pgfpathlineto{\pgfqpoint{0.000000in}{-0.048611in}}%
\pgfusepath{stroke,fill}%
}%
\begin{pgfscope}%
\pgfsys@transformshift{0.674954in}{0.862305in}%
\pgfsys@useobject{currentmarker}{}%
\end{pgfscope}%
\end{pgfscope}%
\begin{pgfscope}%
\definecolor{textcolor}{rgb}{0.000000,0.000000,0.000000}%
\pgfsetstrokecolor{textcolor}%
\pgfsetfillcolor{textcolor}%
\pgftext[x=0.674954in,y=0.765082in,,top]{\color{textcolor}{\rmfamily\fontsize{25.000000}{30.000000}\selectfont\catcode`\^=\active\def^{\ifmmode\sp\else\^{}\fi}\catcode`\%=\active\def
\end{pgfscope}%
\begin{pgfscope}%
\pgfpathrectangle{\pgfqpoint{0.674954in}{0.862305in}}{\pgfqpoint{6.472460in}{4.369081in}}%
\pgfusepath{clip}%
\pgfsetbuttcap%
\pgfsetroundjoin%
\pgfsetlinewidth{2.007500pt}%
\definecolor{currentstroke}{rgb}{0.501961,0.501961,0.501961}%
\pgfsetstrokecolor{currentstroke}%
\pgfsetstrokeopacity{0.300000}%
\pgfsetdash{{7.400000pt}{3.200000pt}}{0.000000pt}%
\pgfpathmoveto{\pgfqpoint{1.969446in}{0.862305in}}%
\pgfpathlineto{\pgfqpoint{1.969446in}{5.231386in}}%
\pgfusepath{stroke}%
\end{pgfscope}%
\begin{pgfscope}%
\pgfsetbuttcap%
\pgfsetroundjoin%
\definecolor{currentfill}{rgb}{0.000000,0.000000,0.000000}%
\pgfsetfillcolor{currentfill}%
\pgfsetlinewidth{0.803000pt}%
\definecolor{currentstroke}{rgb}{0.000000,0.000000,0.000000}%
\pgfsetstrokecolor{currentstroke}%
\pgfsetdash{}{0pt}%
\pgfsys@defobject{currentmarker}{\pgfqpoint{0.000000in}{-0.048611in}}{\pgfqpoint{0.000000in}{0.000000in}}{%
\pgfpathmoveto{\pgfqpoint{0.000000in}{0.000000in}}%
\pgfpathlineto{\pgfqpoint{0.000000in}{-0.048611in}}%
\pgfusepath{stroke,fill}%
}%
\begin{pgfscope}%
\pgfsys@transformshift{1.969446in}{0.862305in}%
\pgfsys@useobject{currentmarker}{}%
\end{pgfscope}%
\end{pgfscope}%
\begin{pgfscope}%
\definecolor{textcolor}{rgb}{0.000000,0.000000,0.000000}%
\pgfsetstrokecolor{textcolor}%
\pgfsetfillcolor{textcolor}%
\pgftext[x=1.969446in,y=0.765082in,,top]{\color{textcolor}{\rmfamily\fontsize{25.000000}{30.000000}\selectfont\catcode`\^=\active\def^{\ifmmode\sp\else\^{}\fi}\catcode`\%=\active\def
\end{pgfscope}%
\begin{pgfscope}%
\pgfpathrectangle{\pgfqpoint{0.674954in}{0.862305in}}{\pgfqpoint{6.472460in}{4.369081in}}%
\pgfusepath{clip}%
\pgfsetbuttcap%
\pgfsetroundjoin%
\pgfsetlinewidth{2.007500pt}%
\definecolor{currentstroke}{rgb}{0.501961,0.501961,0.501961}%
\pgfsetstrokecolor{currentstroke}%
\pgfsetstrokeopacity{0.300000}%
\pgfsetdash{{7.400000pt}{3.200000pt}}{0.000000pt}%
\pgfpathmoveto{\pgfqpoint{3.263938in}{0.862305in}}%
\pgfpathlineto{\pgfqpoint{3.263938in}{5.231386in}}%
\pgfusepath{stroke}%
\end{pgfscope}%
\begin{pgfscope}%
\pgfsetbuttcap%
\pgfsetroundjoin%
\definecolor{currentfill}{rgb}{0.000000,0.000000,0.000000}%
\pgfsetfillcolor{currentfill}%
\pgfsetlinewidth{0.803000pt}%
\definecolor{currentstroke}{rgb}{0.000000,0.000000,0.000000}%
\pgfsetstrokecolor{currentstroke}%
\pgfsetdash{}{0pt}%
\pgfsys@defobject{currentmarker}{\pgfqpoint{0.000000in}{-0.048611in}}{\pgfqpoint{0.000000in}{0.000000in}}{%
\pgfpathmoveto{\pgfqpoint{0.000000in}{0.000000in}}%
\pgfpathlineto{\pgfqpoint{0.000000in}{-0.048611in}}%
\pgfusepath{stroke,fill}%
}%
\begin{pgfscope}%
\pgfsys@transformshift{3.263938in}{0.862305in}%
\pgfsys@useobject{currentmarker}{}%
\end{pgfscope}%
\end{pgfscope}%
\begin{pgfscope}%
\definecolor{textcolor}{rgb}{0.000000,0.000000,0.000000}%
\pgfsetstrokecolor{textcolor}%
\pgfsetfillcolor{textcolor}%
\pgftext[x=3.263938in,y=0.765082in,,top]{\color{textcolor}{\rmfamily\fontsize{25.000000}{30.000000}\selectfont\catcode`\^=\active\def^{\ifmmode\sp\else\^{}\fi}\catcode`\%=\active\def
\end{pgfscope}%
\begin{pgfscope}%
\pgfpathrectangle{\pgfqpoint{0.674954in}{0.862305in}}{\pgfqpoint{6.472460in}{4.369081in}}%
\pgfusepath{clip}%
\pgfsetbuttcap%
\pgfsetroundjoin%
\pgfsetlinewidth{2.007500pt}%
\definecolor{currentstroke}{rgb}{0.501961,0.501961,0.501961}%
\pgfsetstrokecolor{currentstroke}%
\pgfsetstrokeopacity{0.300000}%
\pgfsetdash{{7.400000pt}{3.200000pt}}{0.000000pt}%
\pgfpathmoveto{\pgfqpoint{4.558430in}{0.862305in}}%
\pgfpathlineto{\pgfqpoint{4.558430in}{5.231386in}}%
\pgfusepath{stroke}%
\end{pgfscope}%
\begin{pgfscope}%
\pgfsetbuttcap%
\pgfsetroundjoin%
\definecolor{currentfill}{rgb}{0.000000,0.000000,0.000000}%
\pgfsetfillcolor{currentfill}%
\pgfsetlinewidth{0.803000pt}%
\definecolor{currentstroke}{rgb}{0.000000,0.000000,0.000000}%
\pgfsetstrokecolor{currentstroke}%
\pgfsetdash{}{0pt}%
\pgfsys@defobject{currentmarker}{\pgfqpoint{0.000000in}{-0.048611in}}{\pgfqpoint{0.000000in}{0.000000in}}{%
\pgfpathmoveto{\pgfqpoint{0.000000in}{0.000000in}}%
\pgfpathlineto{\pgfqpoint{0.000000in}{-0.048611in}}%
\pgfusepath{stroke,fill}%
}%
\begin{pgfscope}%
\pgfsys@transformshift{4.558430in}{0.862305in}%
\pgfsys@useobject{currentmarker}{}%
\end{pgfscope}%
\end{pgfscope}%
\begin{pgfscope}%
\definecolor{textcolor}{rgb}{0.000000,0.000000,0.000000}%
\pgfsetstrokecolor{textcolor}%
\pgfsetfillcolor{textcolor}%
\pgftext[x=4.558430in,y=0.765082in,,top]{\color{textcolor}{\rmfamily\fontsize{25.000000}{30.000000}\selectfont\catcode`\^=\active\def^{\ifmmode\sp\else\^{}\fi}\catcode`\%=\active\def
\end{pgfscope}%
\begin{pgfscope}%
\pgfpathrectangle{\pgfqpoint{0.674954in}{0.862305in}}{\pgfqpoint{6.472460in}{4.369081in}}%
\pgfusepath{clip}%
\pgfsetbuttcap%
\pgfsetroundjoin%
\pgfsetlinewidth{2.007500pt}%
\definecolor{currentstroke}{rgb}{0.501961,0.501961,0.501961}%
\pgfsetstrokecolor{currentstroke}%
\pgfsetstrokeopacity{0.300000}%
\pgfsetdash{{7.400000pt}{3.200000pt}}{0.000000pt}%
\pgfpathmoveto{\pgfqpoint{5.852922in}{0.862305in}}%
\pgfpathlineto{\pgfqpoint{5.852922in}{5.231386in}}%
\pgfusepath{stroke}%
\end{pgfscope}%
\begin{pgfscope}%
\pgfsetbuttcap%
\pgfsetroundjoin%
\definecolor{currentfill}{rgb}{0.000000,0.000000,0.000000}%
\pgfsetfillcolor{currentfill}%
\pgfsetlinewidth{0.803000pt}%
\definecolor{currentstroke}{rgb}{0.000000,0.000000,0.000000}%
\pgfsetstrokecolor{currentstroke}%
\pgfsetdash{}{0pt}%
\pgfsys@defobject{currentmarker}{\pgfqpoint{0.000000in}{-0.048611in}}{\pgfqpoint{0.000000in}{0.000000in}}{%
\pgfpathmoveto{\pgfqpoint{0.000000in}{0.000000in}}%
\pgfpathlineto{\pgfqpoint{0.000000in}{-0.048611in}}%
\pgfusepath{stroke,fill}%
}%
\begin{pgfscope}%
\pgfsys@transformshift{5.852922in}{0.862305in}%
\pgfsys@useobject{currentmarker}{}%
\end{pgfscope}%
\end{pgfscope}%
\begin{pgfscope}%
\definecolor{textcolor}{rgb}{0.000000,0.000000,0.000000}%
\pgfsetstrokecolor{textcolor}%
\pgfsetfillcolor{textcolor}%
\pgftext[x=5.852922in,y=0.765082in,,top]{\color{textcolor}{\rmfamily\fontsize{25.000000}{30.000000}\selectfont\catcode`\^=\active\def^{\ifmmode\sp\else\^{}\fi}\catcode`\%=\active\def
\end{pgfscope}%
\begin{pgfscope}%
\pgfpathrectangle{\pgfqpoint{0.674954in}{0.862305in}}{\pgfqpoint{6.472460in}{4.369081in}}%
\pgfusepath{clip}%
\pgfsetbuttcap%
\pgfsetroundjoin%
\pgfsetlinewidth{2.007500pt}%
\definecolor{currentstroke}{rgb}{0.501961,0.501961,0.501961}%
\pgfsetstrokecolor{currentstroke}%
\pgfsetstrokeopacity{0.300000}%
\pgfsetdash{{7.400000pt}{3.200000pt}}{0.000000pt}%
\pgfpathmoveto{\pgfqpoint{7.147414in}{0.862305in}}%
\pgfpathlineto{\pgfqpoint{7.147414in}{5.231386in}}%
\pgfusepath{stroke}%
\end{pgfscope}%
\begin{pgfscope}%
\pgfsetbuttcap%
\pgfsetroundjoin%
\definecolor{currentfill}{rgb}{0.000000,0.000000,0.000000}%
\pgfsetfillcolor{currentfill}%
\pgfsetlinewidth{0.803000pt}%
\definecolor{currentstroke}{rgb}{0.000000,0.000000,0.000000}%
\pgfsetstrokecolor{currentstroke}%
\pgfsetdash{}{0pt}%
\pgfsys@defobject{currentmarker}{\pgfqpoint{0.000000in}{-0.048611in}}{\pgfqpoint{0.000000in}{0.000000in}}{%
\pgfpathmoveto{\pgfqpoint{0.000000in}{0.000000in}}%
\pgfpathlineto{\pgfqpoint{0.000000in}{-0.048611in}}%
\pgfusepath{stroke,fill}%
}%
\begin{pgfscope}%
\pgfsys@transformshift{7.147414in}{0.862305in}%
\pgfsys@useobject{currentmarker}{}%
\end{pgfscope}%
\end{pgfscope}%
\begin{pgfscope}%
\definecolor{textcolor}{rgb}{0.000000,0.000000,0.000000}%
\pgfsetstrokecolor{textcolor}%
\pgfsetfillcolor{textcolor}%
\pgftext[x=7.147414in,y=0.765082in,,top]{\color{textcolor}{\rmfamily\fontsize{25.000000}{30.000000}\selectfont\catcode`\^=\active\def^{\ifmmode\sp\else\^{}\fi}\catcode`\%=\active\def
\end{pgfscope}%
\begin{pgfscope}%
\definecolor{textcolor}{rgb}{0.000000,0.000000,0.000000}%
\pgfsetstrokecolor{textcolor}%
\pgfsetfillcolor{textcolor}%
\pgftext[x=3.911184in,y=0.404763in,,top]{\color{textcolor}{\rmfamily\fontsize{25.000000}{30.000000}\selectfont\catcode`\^=\active\def^{\ifmmode\sp\else\^{}\fi}\catcode`\%=\active\def
\end{pgfscope}%
\begin{pgfscope}%
\pgfpathrectangle{\pgfqpoint{0.674954in}{0.862305in}}{\pgfqpoint{6.472460in}{4.369081in}}%
\pgfusepath{clip}%
\pgfsetbuttcap%
\pgfsetroundjoin%
\pgfsetlinewidth{2.007500pt}%
\definecolor{currentstroke}{rgb}{0.501961,0.501961,0.501961}%
\pgfsetstrokecolor{currentstroke}%
\pgfsetstrokeopacity{0.300000}%
\pgfsetdash{{7.400000pt}{3.200000pt}}{0.000000pt}%
\pgfpathmoveto{\pgfqpoint{0.674954in}{1.954575in}}%
\pgfpathlineto{\pgfqpoint{7.147414in}{1.954575in}}%
\pgfusepath{stroke}%
\end{pgfscope}%
\begin{pgfscope}%
\pgfsetbuttcap%
\pgfsetroundjoin%
\definecolor{currentfill}{rgb}{0.000000,0.000000,0.000000}%
\pgfsetfillcolor{currentfill}%
\pgfsetlinewidth{0.803000pt}%
\definecolor{currentstroke}{rgb}{0.000000,0.000000,0.000000}%
\pgfsetstrokecolor{currentstroke}%
\pgfsetdash{}{0pt}%
\pgfsys@defobject{currentmarker}{\pgfqpoint{-0.048611in}{0.000000in}}{\pgfqpoint{-0.000000in}{0.000000in}}{%
\pgfpathmoveto{\pgfqpoint{-0.000000in}{0.000000in}}%
\pgfpathlineto{\pgfqpoint{-0.048611in}{0.000000in}}%
\pgfusepath{stroke,fill}%
}%
\begin{pgfscope}%
\pgfsys@transformshift{0.674954in}{1.954575in}%
\pgfsys@useobject{currentmarker}{}%
\end{pgfscope}%
\end{pgfscope}%
\begin{pgfscope}%
\definecolor{textcolor}{rgb}{0.000000,0.000000,0.000000}%
\pgfsetstrokecolor{textcolor}%
\pgfsetfillcolor{textcolor}%
\pgftext[x=0.259244in, y=1.835961in, left, base]{\color{textcolor}{\rmfamily\fontsize{25.000000}{30.000000}\selectfont\catcode`\^=\active\def^{\ifmmode\sp\else\^{}\fi}\catcode`\%=\active\def
\end{pgfscope}%
\begin{pgfscope}%
\pgfpathrectangle{\pgfqpoint{0.674954in}{0.862305in}}{\pgfqpoint{6.472460in}{4.369081in}}%
\pgfusepath{clip}%
\pgfsetbuttcap%
\pgfsetroundjoin%
\pgfsetlinewidth{2.007500pt}%
\definecolor{currentstroke}{rgb}{0.501961,0.501961,0.501961}%
\pgfsetstrokecolor{currentstroke}%
\pgfsetstrokeopacity{0.300000}%
\pgfsetdash{{7.400000pt}{3.200000pt}}{0.000000pt}%
\pgfpathmoveto{\pgfqpoint{0.674954in}{3.046845in}}%
\pgfpathlineto{\pgfqpoint{7.147414in}{3.046845in}}%
\pgfusepath{stroke}%
\end{pgfscope}%
\begin{pgfscope}%
\pgfsetbuttcap%
\pgfsetroundjoin%
\definecolor{currentfill}{rgb}{0.000000,0.000000,0.000000}%
\pgfsetfillcolor{currentfill}%
\pgfsetlinewidth{0.803000pt}%
\definecolor{currentstroke}{rgb}{0.000000,0.000000,0.000000}%
\pgfsetstrokecolor{currentstroke}%
\pgfsetdash{}{0pt}%
\pgfsys@defobject{currentmarker}{\pgfqpoint{-0.048611in}{0.000000in}}{\pgfqpoint{-0.000000in}{0.000000in}}{%
\pgfpathmoveto{\pgfqpoint{-0.000000in}{0.000000in}}%
\pgfpathlineto{\pgfqpoint{-0.048611in}{0.000000in}}%
\pgfusepath{stroke,fill}%
}%
\begin{pgfscope}%
\pgfsys@transformshift{0.674954in}{3.046845in}%
\pgfsys@useobject{currentmarker}{}%
\end{pgfscope}%
\end{pgfscope}%
\begin{pgfscope}%
\definecolor{textcolor}{rgb}{0.000000,0.000000,0.000000}%
\pgfsetstrokecolor{textcolor}%
\pgfsetfillcolor{textcolor}%
\pgftext[x=0.259244in, y=2.928231in, left, base]{\color{textcolor}{\rmfamily\fontsize{25.000000}{30.000000}\selectfont\catcode`\^=\active\def^{\ifmmode\sp\else\^{}\fi}\catcode`\%=\active\def
\end{pgfscope}%
\begin{pgfscope}%
\pgfpathrectangle{\pgfqpoint{0.674954in}{0.862305in}}{\pgfqpoint{6.472460in}{4.369081in}}%
\pgfusepath{clip}%
\pgfsetbuttcap%
\pgfsetroundjoin%
\pgfsetlinewidth{2.007500pt}%
\definecolor{currentstroke}{rgb}{0.501961,0.501961,0.501961}%
\pgfsetstrokecolor{currentstroke}%
\pgfsetstrokeopacity{0.300000}%
\pgfsetdash{{7.400000pt}{3.200000pt}}{0.000000pt}%
\pgfpathmoveto{\pgfqpoint{0.674954in}{4.139116in}}%
\pgfpathlineto{\pgfqpoint{7.147414in}{4.139116in}}%
\pgfusepath{stroke}%
\end{pgfscope}%
\begin{pgfscope}%
\pgfsetbuttcap%
\pgfsetroundjoin%
\definecolor{currentfill}{rgb}{0.000000,0.000000,0.000000}%
\pgfsetfillcolor{currentfill}%
\pgfsetlinewidth{0.803000pt}%
\definecolor{currentstroke}{rgb}{0.000000,0.000000,0.000000}%
\pgfsetstrokecolor{currentstroke}%
\pgfsetdash{}{0pt}%
\pgfsys@defobject{currentmarker}{\pgfqpoint{-0.048611in}{0.000000in}}{\pgfqpoint{-0.000000in}{0.000000in}}{%
\pgfpathmoveto{\pgfqpoint{-0.000000in}{0.000000in}}%
\pgfpathlineto{\pgfqpoint{-0.048611in}{0.000000in}}%
\pgfusepath{stroke,fill}%
}%
\begin{pgfscope}%
\pgfsys@transformshift{0.674954in}{4.139116in}%
\pgfsys@useobject{currentmarker}{}%
\end{pgfscope}%
\end{pgfscope}%
\begin{pgfscope}%
\definecolor{textcolor}{rgb}{0.000000,0.000000,0.000000}%
\pgfsetstrokecolor{textcolor}%
\pgfsetfillcolor{textcolor}%
\pgftext[x=0.259244in, y=4.020501in, left, base]{\color{textcolor}{\rmfamily\fontsize{25.000000}{30.000000}\selectfont\catcode`\^=\active\def^{\ifmmode\sp\else\^{}\fi}\catcode`\%=\active\def
\end{pgfscope}%
\begin{pgfscope}%
\pgfpathrectangle{\pgfqpoint{0.674954in}{0.862305in}}{\pgfqpoint{6.472460in}{4.369081in}}%
\pgfusepath{clip}%
\pgfsetbuttcap%
\pgfsetroundjoin%
\pgfsetlinewidth{2.007500pt}%
\definecolor{currentstroke}{rgb}{0.501961,0.501961,0.501961}%
\pgfsetstrokecolor{currentstroke}%
\pgfsetstrokeopacity{0.300000}%
\pgfsetdash{{7.400000pt}{3.200000pt}}{0.000000pt}%
\pgfpathmoveto{\pgfqpoint{0.674954in}{5.231386in}}%
\pgfpathlineto{\pgfqpoint{7.147414in}{5.231386in}}%
\pgfusepath{stroke}%
\end{pgfscope}%
\begin{pgfscope}%
\pgfsetbuttcap%
\pgfsetroundjoin%
\definecolor{currentfill}{rgb}{0.000000,0.000000,0.000000}%
\pgfsetfillcolor{currentfill}%
\pgfsetlinewidth{0.803000pt}%
\definecolor{currentstroke}{rgb}{0.000000,0.000000,0.000000}%
\pgfsetstrokecolor{currentstroke}%
\pgfsetdash{}{0pt}%
\pgfsys@defobject{currentmarker}{\pgfqpoint{-0.048611in}{0.000000in}}{\pgfqpoint{-0.000000in}{0.000000in}}{%
\pgfpathmoveto{\pgfqpoint{-0.000000in}{0.000000in}}%
\pgfpathlineto{\pgfqpoint{-0.048611in}{0.000000in}}%
\pgfusepath{stroke,fill}%
}%
\begin{pgfscope}%
\pgfsys@transformshift{0.674954in}{5.231386in}%
\pgfsys@useobject{currentmarker}{}%
\end{pgfscope}%
\end{pgfscope}%
\begin{pgfscope}%
\definecolor{textcolor}{rgb}{0.000000,0.000000,0.000000}%
\pgfsetstrokecolor{textcolor}%
\pgfsetfillcolor{textcolor}%
\pgftext[x=0.100000in, y=5.112772in, left, base]{\color{textcolor}{\rmfamily\fontsize{25.000000}{30.000000}\selectfont\catcode`\^=\active\def^{\ifmmode\sp\else\^{}\fi}\catcode`\%=\active\def
\end{pgfscope}%
\begin{pgfscope}%
\pgfpathrectangle{\pgfqpoint{0.674954in}{0.862305in}}{\pgfqpoint{6.472460in}{4.369081in}}%
\pgfusepath{clip}%
\pgfsetrectcap%
\pgfsetroundjoin%
\pgfsetlinewidth{2.509375pt}%
\definecolor{currentstroke}{rgb}{0.050980,0.415686,0.509804}%
\pgfsetstrokecolor{currentstroke}%
\pgfsetdash{}{0pt}%
\pgfpathmoveto{\pgfqpoint{0.674954in}{5.043831in}}%
\pgfpathlineto{\pgfqpoint{0.998577in}{5.039479in}}%
\pgfpathlineto{\pgfqpoint{2.293069in}{4.982064in}}%
\pgfpathlineto{\pgfqpoint{3.911184in}{4.905140in}}%
\pgfpathlineto{\pgfqpoint{5.529299in}{4.592469in}}%
\pgfpathlineto{\pgfqpoint{6.823791in}{4.083250in}}%
\pgfusepath{stroke}%
\end{pgfscope}%
\begin{pgfscope}%
\pgfpathrectangle{\pgfqpoint{0.674954in}{0.862305in}}{\pgfqpoint{6.472460in}{4.369081in}}%
\pgfusepath{clip}%
\pgfsetbuttcap%
\pgfsetroundjoin%
\definecolor{currentfill}{rgb}{0.050980,0.415686,0.509804}%
\pgfsetfillcolor{currentfill}%
\pgfsetlinewidth{1.003750pt}%
\definecolor{currentstroke}{rgb}{0.050980,0.415686,0.509804}%
\pgfsetstrokecolor{currentstroke}%
\pgfsetdash{}{0pt}%
\pgfsys@defobject{currentmarker}{\pgfqpoint{-0.055556in}{-0.055556in}}{\pgfqpoint{0.055556in}{0.055556in}}{%
\pgfpathmoveto{\pgfqpoint{0.000000in}{-0.055556in}}%
\pgfpathcurveto{\pgfqpoint{0.014734in}{-0.055556in}}{\pgfqpoint{0.028866in}{-0.049702in}}{\pgfqpoint{0.039284in}{-0.039284in}}%
\pgfpathcurveto{\pgfqpoint{0.049702in}{-0.028866in}}{\pgfqpoint{0.055556in}{-0.014734in}}{\pgfqpoint{0.055556in}{0.000000in}}%
\pgfpathcurveto{\pgfqpoint{0.055556in}{0.014734in}}{\pgfqpoint{0.049702in}{0.028866in}}{\pgfqpoint{0.039284in}{0.039284in}}%
\pgfpathcurveto{\pgfqpoint{0.028866in}{0.049702in}}{\pgfqpoint{0.014734in}{0.055556in}}{\pgfqpoint{0.000000in}{0.055556in}}%
\pgfpathcurveto{\pgfqpoint{-0.014734in}{0.055556in}}{\pgfqpoint{-0.028866in}{0.049702in}}{\pgfqpoint{-0.039284in}{0.039284in}}%
\pgfpathcurveto{\pgfqpoint{-0.049702in}{0.028866in}}{\pgfqpoint{-0.055556in}{0.014734in}}{\pgfqpoint{-0.055556in}{0.000000in}}%
\pgfpathcurveto{\pgfqpoint{-0.055556in}{-0.014734in}}{\pgfqpoint{-0.049702in}{-0.028866in}}{\pgfqpoint{-0.039284in}{-0.039284in}}%
\pgfpathcurveto{\pgfqpoint{-0.028866in}{-0.049702in}}{\pgfqpoint{-0.014734in}{-0.055556in}}{\pgfqpoint{0.000000in}{-0.055556in}}%
\pgfpathlineto{\pgfqpoint{0.000000in}{-0.055556in}}%
\pgfpathclose%
\pgfusepath{stroke,fill}%
}%
\begin{pgfscope}%
\pgfsys@transformshift{0.674954in}{5.043831in}%
\pgfsys@useobject{currentmarker}{}%
\end{pgfscope}%
\begin{pgfscope}%
\pgfsys@transformshift{0.998577in}{5.039479in}%
\pgfsys@useobject{currentmarker}{}%
\end{pgfscope}%
\begin{pgfscope}%
\pgfsys@transformshift{2.293069in}{4.982064in}%
\pgfsys@useobject{currentmarker}{}%
\end{pgfscope}%
\begin{pgfscope}%
\pgfsys@transformshift{3.911184in}{4.905140in}%
\pgfsys@useobject{currentmarker}{}%
\end{pgfscope}%
\begin{pgfscope}%
\pgfsys@transformshift{5.529299in}{4.592469in}%
\pgfsys@useobject{currentmarker}{}%
\end{pgfscope}%
\begin{pgfscope}%
\pgfsys@transformshift{6.823791in}{4.083250in}%
\pgfsys@useobject{currentmarker}{}%
\end{pgfscope}%
\end{pgfscope}%
\begin{pgfscope}%
\pgfpathrectangle{\pgfqpoint{0.674954in}{0.862305in}}{\pgfqpoint{6.472460in}{4.369081in}}%
\pgfusepath{clip}%
\pgfsetbuttcap%
\pgfsetroundjoin%
\pgfsetlinewidth{2.509375pt}%
\definecolor{currentstroke}{rgb}{0.960784,0.462745,0.000000}%
\pgfsetstrokecolor{currentstroke}%
\pgfsetdash{{9.250000pt}{4.000000pt}}{0.000000pt}%
\pgfpathmoveto{\pgfqpoint{0.998577in}{4.329829in}}%
\pgfpathlineto{\pgfqpoint{2.293069in}{4.417623in}}%
\pgfpathlineto{\pgfqpoint{3.911184in}{4.313191in}}%
\pgfpathlineto{\pgfqpoint{5.529299in}{4.244487in}}%
\pgfpathlineto{\pgfqpoint{6.823791in}{4.342527in}}%
\pgfpathlineto{\pgfqpoint{7.147414in}{4.442942in}}%
\pgfusepath{stroke}%
\end{pgfscope}%
\begin{pgfscope}%
\pgfpathrectangle{\pgfqpoint{0.674954in}{0.862305in}}{\pgfqpoint{6.472460in}{4.369081in}}%
\pgfusepath{clip}%
\pgfsetbuttcap%
\pgfsetmiterjoin%
\definecolor{currentfill}{rgb}{0.960784,0.462745,0.000000}%
\pgfsetfillcolor{currentfill}%
\pgfsetlinewidth{1.003750pt}%
\definecolor{currentstroke}{rgb}{0.960784,0.462745,0.000000}%
\pgfsetstrokecolor{currentstroke}%
\pgfsetdash{}{0pt}%
\pgfsys@defobject{currentmarker}{\pgfqpoint{-0.055556in}{-0.055556in}}{\pgfqpoint{0.055556in}{0.055556in}}{%
\pgfpathmoveto{\pgfqpoint{-0.055556in}{-0.055556in}}%
\pgfpathlineto{\pgfqpoint{0.055556in}{-0.055556in}}%
\pgfpathlineto{\pgfqpoint{0.055556in}{0.055556in}}%
\pgfpathlineto{\pgfqpoint{-0.055556in}{0.055556in}}%
\pgfpathlineto{\pgfqpoint{-0.055556in}{-0.055556in}}%
\pgfpathclose%
\pgfusepath{stroke,fill}%
}%
\begin{pgfscope}%
\pgfsys@transformshift{0.998577in}{4.329829in}%
\pgfsys@useobject{currentmarker}{}%
\end{pgfscope}%
\begin{pgfscope}%
\pgfsys@transformshift{2.293069in}{4.417623in}%
\pgfsys@useobject{currentmarker}{}%
\end{pgfscope}%
\begin{pgfscope}%
\pgfsys@transformshift{3.911184in}{4.313191in}%
\pgfsys@useobject{currentmarker}{}%
\end{pgfscope}%
\begin{pgfscope}%
\pgfsys@transformshift{5.529299in}{4.244487in}%
\pgfsys@useobject{currentmarker}{}%
\end{pgfscope}%
\begin{pgfscope}%
\pgfsys@transformshift{6.823791in}{4.342527in}%
\pgfsys@useobject{currentmarker}{}%
\end{pgfscope}%
\begin{pgfscope}%
\pgfsys@transformshift{7.147414in}{4.442942in}%
\pgfsys@useobject{currentmarker}{}%
\end{pgfscope}%
\end{pgfscope}%
\begin{pgfscope}%
\pgfsetrectcap%
\pgfsetmiterjoin%
\pgfsetlinewidth{2.007500pt}%
\definecolor{currentstroke}{rgb}{0.000000,0.000000,0.000000}%
\pgfsetstrokecolor{currentstroke}%
\pgfsetdash{}{0pt}%
\pgfpathmoveto{\pgfqpoint{0.674954in}{0.862305in}}%
\pgfpathlineto{\pgfqpoint{0.674954in}{5.231386in}}%
\pgfusepath{stroke}%
\end{pgfscope}%
\begin{pgfscope}%
\pgfsetrectcap%
\pgfsetmiterjoin%
\pgfsetlinewidth{2.007500pt}%
\definecolor{currentstroke}{rgb}{0.000000,0.000000,0.000000}%
\pgfsetstrokecolor{currentstroke}%
\pgfsetdash{}{0pt}%
\pgfpathmoveto{\pgfqpoint{0.674954in}{0.862305in}}%
\pgfpathlineto{\pgfqpoint{7.147414in}{0.862305in}}%
\pgfusepath{stroke}%
\end{pgfscope}%
\begin{pgfscope}%
\pgfsetbuttcap%
\pgfsetmiterjoin%
\definecolor{currentfill}{rgb}{1.000000,1.000000,1.000000}%
\pgfsetfillcolor{currentfill}%
\pgfsetlinewidth{1.003750pt}%
\definecolor{currentstroke}{rgb}{0.800000,0.800000,0.800000}%
\pgfsetstrokecolor{currentstroke}%
\pgfsetdash{}{0pt}%
\pgfpathmoveto{\pgfqpoint{0.869398in}{1.001194in}}%
\pgfpathlineto{\pgfqpoint{2.708109in}{1.001194in}}%
\pgfpathquadraticcurveto{\pgfqpoint{2.763664in}{1.001194in}}{\pgfqpoint{2.763664in}{1.056749in}}%
\pgfpathlineto{\pgfqpoint{2.763664in}{1.803694in}}%
\pgfpathquadraticcurveto{\pgfqpoint{2.763664in}{1.859250in}}{\pgfqpoint{2.708109in}{1.859250in}}%
\pgfpathlineto{\pgfqpoint{0.869398in}{1.859250in}}%
\pgfpathquadraticcurveto{\pgfqpoint{0.813843in}{1.859250in}}{\pgfqpoint{0.813843in}{1.803694in}}%
\pgfpathlineto{\pgfqpoint{0.813843in}{1.056749in}}%
\pgfpathquadraticcurveto{\pgfqpoint{0.813843in}{1.001194in}}{\pgfqpoint{0.869398in}{1.001194in}}%
\pgfpathlineto{\pgfqpoint{0.869398in}{1.001194in}}%
\pgfpathclose%
\pgfusepath{stroke,fill}%
\end{pgfscope}%
\begin{pgfscope}%
\pgfsetrectcap%
\pgfsetroundjoin%
\pgfsetlinewidth{2.509375pt}%
\definecolor{currentstroke}{rgb}{0.050980,0.415686,0.509804}%
\pgfsetstrokecolor{currentstroke}%
\pgfsetdash{}{0pt}%
\pgfpathmoveto{\pgfqpoint{0.924954in}{1.650917in}}%
\pgfpathlineto{\pgfqpoint{1.202732in}{1.650917in}}%
\pgfpathlineto{\pgfqpoint{1.480509in}{1.650917in}}%
\pgfusepath{stroke}%
\end{pgfscope}%
\begin{pgfscope}%
\pgfsetbuttcap%
\pgfsetroundjoin%
\definecolor{currentfill}{rgb}{0.050980,0.415686,0.509804}%
\pgfsetfillcolor{currentfill}%
\pgfsetlinewidth{1.003750pt}%
\definecolor{currentstroke}{rgb}{0.050980,0.415686,0.509804}%
\pgfsetstrokecolor{currentstroke}%
\pgfsetdash{}{0pt}%
\pgfsys@defobject{currentmarker}{\pgfqpoint{-0.055556in}{-0.055556in}}{\pgfqpoint{0.055556in}{0.055556in}}{%
\pgfpathmoveto{\pgfqpoint{0.000000in}{-0.055556in}}%
\pgfpathcurveto{\pgfqpoint{0.014734in}{-0.055556in}}{\pgfqpoint{0.028866in}{-0.049702in}}{\pgfqpoint{0.039284in}{-0.039284in}}%
\pgfpathcurveto{\pgfqpoint{0.049702in}{-0.028866in}}{\pgfqpoint{0.055556in}{-0.014734in}}{\pgfqpoint{0.055556in}{0.000000in}}%
\pgfpathcurveto{\pgfqpoint{0.055556in}{0.014734in}}{\pgfqpoint{0.049702in}{0.028866in}}{\pgfqpoint{0.039284in}{0.039284in}}%
\pgfpathcurveto{\pgfqpoint{0.028866in}{0.049702in}}{\pgfqpoint{0.014734in}{0.055556in}}{\pgfqpoint{0.000000in}{0.055556in}}%
\pgfpathcurveto{\pgfqpoint{-0.014734in}{0.055556in}}{\pgfqpoint{-0.028866in}{0.049702in}}{\pgfqpoint{-0.039284in}{0.039284in}}%
\pgfpathcurveto{\pgfqpoint{-0.049702in}{0.028866in}}{\pgfqpoint{-0.055556in}{0.014734in}}{\pgfqpoint{-0.055556in}{0.000000in}}%
\pgfpathcurveto{\pgfqpoint{-0.055556in}{-0.014734in}}{\pgfqpoint{-0.049702in}{-0.028866in}}{\pgfqpoint{-0.039284in}{-0.039284in}}%
\pgfpathcurveto{\pgfqpoint{-0.028866in}{-0.049702in}}{\pgfqpoint{-0.014734in}{-0.055556in}}{\pgfqpoint{0.000000in}{-0.055556in}}%
\pgfpathlineto{\pgfqpoint{0.000000in}{-0.055556in}}%
\pgfpathclose%
\pgfusepath{stroke,fill}%
}%
\begin{pgfscope}%
\pgfsys@transformshift{1.202732in}{1.650917in}%
\pgfsys@useobject{currentmarker}{}%
\end{pgfscope}%
\end{pgfscope}%
\begin{pgfscope}%
\definecolor{textcolor}{rgb}{0.000000,0.000000,0.000000}%
\pgfsetstrokecolor{textcolor}%
\pgfsetfillcolor{textcolor}%
\pgftext[x=1.702732in,y=1.553694in,left,base]{\color{textcolor}{\rmfamily\fontsize{20.000000}{24.000000}\selectfont\catcode`\^=\active\def^{\ifmmode\sp\else\^{}\fi}\catcode`\%=\active\def
\end{pgfscope}%
\begin{pgfscope}%
\pgfsetbuttcap%
\pgfsetroundjoin%
\pgfsetlinewidth{2.509375pt}%
\definecolor{currentstroke}{rgb}{0.960784,0.462745,0.000000}%
\pgfsetstrokecolor{currentstroke}%
\pgfsetdash{{9.250000pt}{4.000000pt}}{0.000000pt}%
\pgfpathmoveto{\pgfqpoint{0.924954in}{1.263555in}}%
\pgfpathlineto{\pgfqpoint{1.202732in}{1.263555in}}%
\pgfpathlineto{\pgfqpoint{1.480509in}{1.263555in}}%
\pgfusepath{stroke}%
\end{pgfscope}%
\begin{pgfscope}%
\pgfsetbuttcap%
\pgfsetmiterjoin%
\definecolor{currentfill}{rgb}{0.960784,0.462745,0.000000}%
\pgfsetfillcolor{currentfill}%
\pgfsetlinewidth{1.003750pt}%
\definecolor{currentstroke}{rgb}{0.960784,0.462745,0.000000}%
\pgfsetstrokecolor{currentstroke}%
\pgfsetdash{}{0pt}%
\pgfsys@defobject{currentmarker}{\pgfqpoint{-0.055556in}{-0.055556in}}{\pgfqpoint{0.055556in}{0.055556in}}{%
\pgfpathmoveto{\pgfqpoint{-0.055556in}{-0.055556in}}%
\pgfpathlineto{\pgfqpoint{0.055556in}{-0.055556in}}%
\pgfpathlineto{\pgfqpoint{0.055556in}{0.055556in}}%
\pgfpathlineto{\pgfqpoint{-0.055556in}{0.055556in}}%
\pgfpathlineto{\pgfqpoint{-0.055556in}{-0.055556in}}%
\pgfpathclose%
\pgfusepath{stroke,fill}%
}%
\begin{pgfscope}%
\pgfsys@transformshift{1.202732in}{1.263555in}%
\pgfsys@useobject{currentmarker}{}%
\end{pgfscope}%
\end{pgfscope}%
\begin{pgfscope}%
\definecolor{textcolor}{rgb}{0.000000,0.000000,0.000000}%
\pgfsetstrokecolor{textcolor}%
\pgfsetfillcolor{textcolor}%
\pgftext[x=1.702732in,y=1.166333in,left,base]{\color{textcolor}{\rmfamily\fontsize{20.000000}{24.000000}\selectfont\catcode`\^=\active\def^{\ifmmode\sp\else\^{}\fi}\catcode`\%=\active\def
\end{pgfscope}%
\end{pgfpicture}%
\makeatother%
\endgroup%

%% file: images/thermal_ratio_per_modality/modality_results_RTA_30.pgf
\begingroup%
\makeatletter%
\begin{pgfpicture}%
\pgfpathrectangle{\pgfpointorigin}{\pgfqpoint{7.450000in}{5.450000in}}%
\pgfusepath{use as bounding box, clip}%
\begin{pgfscope}%
\pgfsetbuttcap%
\pgfsetmiterjoin%
\definecolor{currentfill}{rgb}{1.000000,1.000000,1.000000}%
\pgfsetfillcolor{currentfill}%
\pgfsetlinewidth{0.000000pt}%
\definecolor{currentstroke}{rgb}{1.000000,1.000000,1.000000}%
\pgfsetstrokecolor{currentstroke}%
\pgfsetdash{}{0pt}%
\pgfpathmoveto{\pgfqpoint{0.000000in}{0.000000in}}%
\pgfpathlineto{\pgfqpoint{7.450000in}{0.000000in}}%
\pgfpathlineto{\pgfqpoint{7.450000in}{5.450000in}}%
\pgfpathlineto{\pgfqpoint{0.000000in}{5.450000in}}%
\pgfpathlineto{\pgfqpoint{0.000000in}{0.000000in}}%
\pgfpathclose%
\pgfusepath{fill}%
\end{pgfscope}%
\begin{pgfscope}%
\pgfsetbuttcap%
\pgfsetmiterjoin%
\definecolor{currentfill}{rgb}{1.000000,1.000000,1.000000}%
\pgfsetfillcolor{currentfill}%
\pgfsetlinewidth{0.000000pt}%
\definecolor{currentstroke}{rgb}{0.000000,0.000000,0.000000}%
\pgfsetstrokecolor{currentstroke}%
\pgfsetstrokeopacity{0.000000}%
\pgfsetdash{}{0pt}%
\pgfpathmoveto{\pgfqpoint{0.674954in}{0.862305in}}%
\pgfpathlineto{\pgfqpoint{7.147414in}{0.862305in}}%
\pgfpathlineto{\pgfqpoint{7.147414in}{5.231386in}}%
\pgfpathlineto{\pgfqpoint{0.674954in}{5.231386in}}%
\pgfpathlineto{\pgfqpoint{0.674954in}{0.862305in}}%
\pgfpathclose%
\pgfusepath{fill}%
\end{pgfscope}%
\begin{pgfscope}%
\pgfpathrectangle{\pgfqpoint{0.674954in}{0.862305in}}{\pgfqpoint{6.472460in}{4.369081in}}%
\pgfusepath{clip}%
\pgfsetbuttcap%
\pgfsetroundjoin%
\pgfsetlinewidth{2.007500pt}%
\definecolor{currentstroke}{rgb}{0.501961,0.501961,0.501961}%
\pgfsetstrokecolor{currentstroke}%
\pgfsetstrokeopacity{0.300000}%
\pgfsetdash{{7.400000pt}{3.200000pt}}{0.000000pt}%
\pgfpathmoveto{\pgfqpoint{0.674954in}{0.862305in}}%
\pgfpathlineto{\pgfqpoint{0.674954in}{5.231386in}}%
\pgfusepath{stroke}%
\end{pgfscope}%
\begin{pgfscope}%
\pgfsetbuttcap%
\pgfsetroundjoin%
\definecolor{currentfill}{rgb}{0.000000,0.000000,0.000000}%
\pgfsetfillcolor{currentfill}%
\pgfsetlinewidth{0.803000pt}%
\definecolor{currentstroke}{rgb}{0.000000,0.000000,0.000000}%
\pgfsetstrokecolor{currentstroke}%
\pgfsetdash{}{0pt}%
\pgfsys@defobject{currentmarker}{\pgfqpoint{0.000000in}{-0.048611in}}{\pgfqpoint{0.000000in}{0.000000in}}{%
\pgfpathmoveto{\pgfqpoint{0.000000in}{0.000000in}}%
\pgfpathlineto{\pgfqpoint{0.000000in}{-0.048611in}}%
\pgfusepath{stroke,fill}%
}%
\begin{pgfscope}%
\pgfsys@transformshift{0.674954in}{0.862305in}%
\pgfsys@useobject{currentmarker}{}%
\end{pgfscope}%
\end{pgfscope}%
\begin{pgfscope}%
\definecolor{textcolor}{rgb}{0.000000,0.000000,0.000000}%
\pgfsetstrokecolor{textcolor}%
\pgfsetfillcolor{textcolor}%
\pgftext[x=0.674954in,y=0.765082in,,top]{\color{textcolor}{\rmfamily\fontsize{25.000000}{30.000000}\selectfont\catcode`\^=\active\def^{\ifmmode\sp\else\^{}\fi}\catcode`\%=\active\def
\end{pgfscope}%
\begin{pgfscope}%
\pgfpathrectangle{\pgfqpoint{0.674954in}{0.862305in}}{\pgfqpoint{6.472460in}{4.369081in}}%
\pgfusepath{clip}%
\pgfsetbuttcap%
\pgfsetroundjoin%
\pgfsetlinewidth{2.007500pt}%
\definecolor{currentstroke}{rgb}{0.501961,0.501961,0.501961}%
\pgfsetstrokecolor{currentstroke}%
\pgfsetstrokeopacity{0.300000}%
\pgfsetdash{{7.400000pt}{3.200000pt}}{0.000000pt}%
\pgfpathmoveto{\pgfqpoint{1.969446in}{0.862305in}}%
\pgfpathlineto{\pgfqpoint{1.969446in}{5.231386in}}%
\pgfusepath{stroke}%
\end{pgfscope}%
\begin{pgfscope}%
\pgfsetbuttcap%
\pgfsetroundjoin%
\definecolor{currentfill}{rgb}{0.000000,0.000000,0.000000}%
\pgfsetfillcolor{currentfill}%
\pgfsetlinewidth{0.803000pt}%
\definecolor{currentstroke}{rgb}{0.000000,0.000000,0.000000}%
\pgfsetstrokecolor{currentstroke}%
\pgfsetdash{}{0pt}%
\pgfsys@defobject{currentmarker}{\pgfqpoint{0.000000in}{-0.048611in}}{\pgfqpoint{0.000000in}{0.000000in}}{%
\pgfpathmoveto{\pgfqpoint{0.000000in}{0.000000in}}%
\pgfpathlineto{\pgfqpoint{0.000000in}{-0.048611in}}%
\pgfusepath{stroke,fill}%
}%
\begin{pgfscope}%
\pgfsys@transformshift{1.969446in}{0.862305in}%
\pgfsys@useobject{currentmarker}{}%
\end{pgfscope}%
\end{pgfscope}%
\begin{pgfscope}%
\definecolor{textcolor}{rgb}{0.000000,0.000000,0.000000}%
\pgfsetstrokecolor{textcolor}%
\pgfsetfillcolor{textcolor}%
\pgftext[x=1.969446in,y=0.765082in,,top]{\color{textcolor}{\rmfamily\fontsize{25.000000}{30.000000}\selectfont\catcode`\^=\active\def^{\ifmmode\sp\else\^{}\fi}\catcode`\%=\active\def
\end{pgfscope}%
\begin{pgfscope}%
\pgfpathrectangle{\pgfqpoint{0.674954in}{0.862305in}}{\pgfqpoint{6.472460in}{4.369081in}}%
\pgfusepath{clip}%
\pgfsetbuttcap%
\pgfsetroundjoin%
\pgfsetlinewidth{2.007500pt}%
\definecolor{currentstroke}{rgb}{0.501961,0.501961,0.501961}%
\pgfsetstrokecolor{currentstroke}%
\pgfsetstrokeopacity{0.300000}%
\pgfsetdash{{7.400000pt}{3.200000pt}}{0.000000pt}%
\pgfpathmoveto{\pgfqpoint{3.263938in}{0.862305in}}%
\pgfpathlineto{\pgfqpoint{3.263938in}{5.231386in}}%
\pgfusepath{stroke}%
\end{pgfscope}%
\begin{pgfscope}%
\pgfsetbuttcap%
\pgfsetroundjoin%
\definecolor{currentfill}{rgb}{0.000000,0.000000,0.000000}%
\pgfsetfillcolor{currentfill}%
\pgfsetlinewidth{0.803000pt}%
\definecolor{currentstroke}{rgb}{0.000000,0.000000,0.000000}%
\pgfsetstrokecolor{currentstroke}%
\pgfsetdash{}{0pt}%
\pgfsys@defobject{currentmarker}{\pgfqpoint{0.000000in}{-0.048611in}}{\pgfqpoint{0.000000in}{0.000000in}}{%
\pgfpathmoveto{\pgfqpoint{0.000000in}{0.000000in}}%
\pgfpathlineto{\pgfqpoint{0.000000in}{-0.048611in}}%
\pgfusepath{stroke,fill}%
}%
\begin{pgfscope}%
\pgfsys@transformshift{3.263938in}{0.862305in}%
\pgfsys@useobject{currentmarker}{}%
\end{pgfscope}%
\end{pgfscope}%
\begin{pgfscope}%
\definecolor{textcolor}{rgb}{0.000000,0.000000,0.000000}%
\pgfsetstrokecolor{textcolor}%
\pgfsetfillcolor{textcolor}%
\pgftext[x=3.263938in,y=0.765082in,,top]{\color{textcolor}{\rmfamily\fontsize{25.000000}{30.000000}\selectfont\catcode`\^=\active\def^{\ifmmode\sp\else\^{}\fi}\catcode`\%=\active\def
\end{pgfscope}%
\begin{pgfscope}%
\pgfpathrectangle{\pgfqpoint{0.674954in}{0.862305in}}{\pgfqpoint{6.472460in}{4.369081in}}%
\pgfusepath{clip}%
\pgfsetbuttcap%
\pgfsetroundjoin%
\pgfsetlinewidth{2.007500pt}%
\definecolor{currentstroke}{rgb}{0.501961,0.501961,0.501961}%
\pgfsetstrokecolor{currentstroke}%
\pgfsetstrokeopacity{0.300000}%
\pgfsetdash{{7.400000pt}{3.200000pt}}{0.000000pt}%
\pgfpathmoveto{\pgfqpoint{4.558430in}{0.862305in}}%
\pgfpathlineto{\pgfqpoint{4.558430in}{5.231386in}}%
\pgfusepath{stroke}%
\end{pgfscope}%
\begin{pgfscope}%
\pgfsetbuttcap%
\pgfsetroundjoin%
\definecolor{currentfill}{rgb}{0.000000,0.000000,0.000000}%
\pgfsetfillcolor{currentfill}%
\pgfsetlinewidth{0.803000pt}%
\definecolor{currentstroke}{rgb}{0.000000,0.000000,0.000000}%
\pgfsetstrokecolor{currentstroke}%
\pgfsetdash{}{0pt}%
\pgfsys@defobject{currentmarker}{\pgfqpoint{0.000000in}{-0.048611in}}{\pgfqpoint{0.000000in}{0.000000in}}{%
\pgfpathmoveto{\pgfqpoint{0.000000in}{0.000000in}}%
\pgfpathlineto{\pgfqpoint{0.000000in}{-0.048611in}}%
\pgfusepath{stroke,fill}%
}%
\begin{pgfscope}%
\pgfsys@transformshift{4.558430in}{0.862305in}%
\pgfsys@useobject{currentmarker}{}%
\end{pgfscope}%
\end{pgfscope}%
\begin{pgfscope}%
\definecolor{textcolor}{rgb}{0.000000,0.000000,0.000000}%
\pgfsetstrokecolor{textcolor}%
\pgfsetfillcolor{textcolor}%
\pgftext[x=4.558430in,y=0.765082in,,top]{\color{textcolor}{\rmfamily\fontsize{25.000000}{30.000000}\selectfont\catcode`\^=\active\def^{\ifmmode\sp\else\^{}\fi}\catcode`\%=\active\def
\end{pgfscope}%
\begin{pgfscope}%
\pgfpathrectangle{\pgfqpoint{0.674954in}{0.862305in}}{\pgfqpoint{6.472460in}{4.369081in}}%
\pgfusepath{clip}%
\pgfsetbuttcap%
\pgfsetroundjoin%
\pgfsetlinewidth{2.007500pt}%
\definecolor{currentstroke}{rgb}{0.501961,0.501961,0.501961}%
\pgfsetstrokecolor{currentstroke}%
\pgfsetstrokeopacity{0.300000}%
\pgfsetdash{{7.400000pt}{3.200000pt}}{0.000000pt}%
\pgfpathmoveto{\pgfqpoint{5.852922in}{0.862305in}}%
\pgfpathlineto{\pgfqpoint{5.852922in}{5.231386in}}%
\pgfusepath{stroke}%
\end{pgfscope}%
\begin{pgfscope}%
\pgfsetbuttcap%
\pgfsetroundjoin%
\definecolor{currentfill}{rgb}{0.000000,0.000000,0.000000}%
\pgfsetfillcolor{currentfill}%
\pgfsetlinewidth{0.803000pt}%
\definecolor{currentstroke}{rgb}{0.000000,0.000000,0.000000}%
\pgfsetstrokecolor{currentstroke}%
\pgfsetdash{}{0pt}%
\pgfsys@defobject{currentmarker}{\pgfqpoint{0.000000in}{-0.048611in}}{\pgfqpoint{0.000000in}{0.000000in}}{%
\pgfpathmoveto{\pgfqpoint{0.000000in}{0.000000in}}%
\pgfpathlineto{\pgfqpoint{0.000000in}{-0.048611in}}%
\pgfusepath{stroke,fill}%
}%
\begin{pgfscope}%
\pgfsys@transformshift{5.852922in}{0.862305in}%
\pgfsys@useobject{currentmarker}{}%
\end{pgfscope}%
\end{pgfscope}%
\begin{pgfscope}%
\definecolor{textcolor}{rgb}{0.000000,0.000000,0.000000}%
\pgfsetstrokecolor{textcolor}%
\pgfsetfillcolor{textcolor}%
\pgftext[x=5.852922in,y=0.765082in,,top]{\color{textcolor}{\rmfamily\fontsize{25.000000}{30.000000}\selectfont\catcode`\^=\active\def^{\ifmmode\sp\else\^{}\fi}\catcode`\%=\active\def
\end{pgfscope}%
\begin{pgfscope}%
\pgfpathrectangle{\pgfqpoint{0.674954in}{0.862305in}}{\pgfqpoint{6.472460in}{4.369081in}}%
\pgfusepath{clip}%
\pgfsetbuttcap%
\pgfsetroundjoin%
\pgfsetlinewidth{2.007500pt}%
\definecolor{currentstroke}{rgb}{0.501961,0.501961,0.501961}%
\pgfsetstrokecolor{currentstroke}%
\pgfsetstrokeopacity{0.300000}%
\pgfsetdash{{7.400000pt}{3.200000pt}}{0.000000pt}%
\pgfpathmoveto{\pgfqpoint{7.147414in}{0.862305in}}%
\pgfpathlineto{\pgfqpoint{7.147414in}{5.231386in}}%
\pgfusepath{stroke}%
\end{pgfscope}%
\begin{pgfscope}%
\pgfsetbuttcap%
\pgfsetroundjoin%
\definecolor{currentfill}{rgb}{0.000000,0.000000,0.000000}%
\pgfsetfillcolor{currentfill}%
\pgfsetlinewidth{0.803000pt}%
\definecolor{currentstroke}{rgb}{0.000000,0.000000,0.000000}%
\pgfsetstrokecolor{currentstroke}%
\pgfsetdash{}{0pt}%
\pgfsys@defobject{currentmarker}{\pgfqpoint{0.000000in}{-0.048611in}}{\pgfqpoint{0.000000in}{0.000000in}}{%
\pgfpathmoveto{\pgfqpoint{0.000000in}{0.000000in}}%
\pgfpathlineto{\pgfqpoint{0.000000in}{-0.048611in}}%
\pgfusepath{stroke,fill}%
}%
\begin{pgfscope}%
\pgfsys@transformshift{7.147414in}{0.862305in}%
\pgfsys@useobject{currentmarker}{}%
\end{pgfscope}%
\end{pgfscope}%
\begin{pgfscope}%
\definecolor{textcolor}{rgb}{0.000000,0.000000,0.000000}%
\pgfsetstrokecolor{textcolor}%
\pgfsetfillcolor{textcolor}%
\pgftext[x=7.147414in,y=0.765082in,,top]{\color{textcolor}{\rmfamily\fontsize{25.000000}{30.000000}\selectfont\catcode`\^=\active\def^{\ifmmode\sp\else\^{}\fi}\catcode`\%=\active\def
\end{pgfscope}%
\begin{pgfscope}%
\definecolor{textcolor}{rgb}{0.000000,0.000000,0.000000}%
\pgfsetstrokecolor{textcolor}%
\pgfsetfillcolor{textcolor}%
\pgftext[x=3.911184in,y=0.404763in,,top]{\color{textcolor}{\rmfamily\fontsize{25.000000}{30.000000}\selectfont\catcode`\^=\active\def^{\ifmmode\sp\else\^{}\fi}\catcode`\%=\active\def
\end{pgfscope}%
\begin{pgfscope}%
\pgfpathrectangle{\pgfqpoint{0.674954in}{0.862305in}}{\pgfqpoint{6.472460in}{4.369081in}}%
\pgfusepath{clip}%
\pgfsetbuttcap%
\pgfsetroundjoin%
\pgfsetlinewidth{2.007500pt}%
\definecolor{currentstroke}{rgb}{0.501961,0.501961,0.501961}%
\pgfsetstrokecolor{currentstroke}%
\pgfsetstrokeopacity{0.300000}%
\pgfsetdash{{7.400000pt}{3.200000pt}}{0.000000pt}%
\pgfpathmoveto{\pgfqpoint{0.674954in}{1.954575in}}%
\pgfpathlineto{\pgfqpoint{7.147414in}{1.954575in}}%
\pgfusepath{stroke}%
\end{pgfscope}%
\begin{pgfscope}%
\pgfsetbuttcap%
\pgfsetroundjoin%
\definecolor{currentfill}{rgb}{0.000000,0.000000,0.000000}%
\pgfsetfillcolor{currentfill}%
\pgfsetlinewidth{0.803000pt}%
\definecolor{currentstroke}{rgb}{0.000000,0.000000,0.000000}%
\pgfsetstrokecolor{currentstroke}%
\pgfsetdash{}{0pt}%
\pgfsys@defobject{currentmarker}{\pgfqpoint{-0.048611in}{0.000000in}}{\pgfqpoint{-0.000000in}{0.000000in}}{%
\pgfpathmoveto{\pgfqpoint{-0.000000in}{0.000000in}}%
\pgfpathlineto{\pgfqpoint{-0.048611in}{0.000000in}}%
\pgfusepath{stroke,fill}%
}%
\begin{pgfscope}%
\pgfsys@transformshift{0.674954in}{1.954575in}%
\pgfsys@useobject{currentmarker}{}%
\end{pgfscope}%
\end{pgfscope}%
\begin{pgfscope}%
\definecolor{textcolor}{rgb}{0.000000,0.000000,0.000000}%
\pgfsetstrokecolor{textcolor}%
\pgfsetfillcolor{textcolor}%
\pgftext[x=0.259244in, y=1.835961in, left, base]{\color{textcolor}{\rmfamily\fontsize{25.000000}{30.000000}\selectfont\catcode`\^=\active\def^{\ifmmode\sp\else\^{}\fi}\catcode`\%=\active\def
\end{pgfscope}%
\begin{pgfscope}%
\pgfpathrectangle{\pgfqpoint{0.674954in}{0.862305in}}{\pgfqpoint{6.472460in}{4.369081in}}%
\pgfusepath{clip}%
\pgfsetbuttcap%
\pgfsetroundjoin%
\pgfsetlinewidth{2.007500pt}%
\definecolor{currentstroke}{rgb}{0.501961,0.501961,0.501961}%
\pgfsetstrokecolor{currentstroke}%
\pgfsetstrokeopacity{0.300000}%
\pgfsetdash{{7.400000pt}{3.200000pt}}{0.000000pt}%
\pgfpathmoveto{\pgfqpoint{0.674954in}{3.046845in}}%
\pgfpathlineto{\pgfqpoint{7.147414in}{3.046845in}}%
\pgfusepath{stroke}%
\end{pgfscope}%
\begin{pgfscope}%
\pgfsetbuttcap%
\pgfsetroundjoin%
\definecolor{currentfill}{rgb}{0.000000,0.000000,0.000000}%
\pgfsetfillcolor{currentfill}%
\pgfsetlinewidth{0.803000pt}%
\definecolor{currentstroke}{rgb}{0.000000,0.000000,0.000000}%
\pgfsetstrokecolor{currentstroke}%
\pgfsetdash{}{0pt}%
\pgfsys@defobject{currentmarker}{\pgfqpoint{-0.048611in}{0.000000in}}{\pgfqpoint{-0.000000in}{0.000000in}}{%
\pgfpathmoveto{\pgfqpoint{-0.000000in}{0.000000in}}%
\pgfpathlineto{\pgfqpoint{-0.048611in}{0.000000in}}%
\pgfusepath{stroke,fill}%
}%
\begin{pgfscope}%
\pgfsys@transformshift{0.674954in}{3.046845in}%
\pgfsys@useobject{currentmarker}{}%
\end{pgfscope}%
\end{pgfscope}%
\begin{pgfscope}%
\definecolor{textcolor}{rgb}{0.000000,0.000000,0.000000}%
\pgfsetstrokecolor{textcolor}%
\pgfsetfillcolor{textcolor}%
\pgftext[x=0.259244in, y=2.928231in, left, base]{\color{textcolor}{\rmfamily\fontsize{25.000000}{30.000000}\selectfont\catcode`\^=\active\def^{\ifmmode\sp\else\^{}\fi}\catcode`\%=\active\def
\end{pgfscope}%
\begin{pgfscope}%
\pgfpathrectangle{\pgfqpoint{0.674954in}{0.862305in}}{\pgfqpoint{6.472460in}{4.369081in}}%
\pgfusepath{clip}%
\pgfsetbuttcap%
\pgfsetroundjoin%
\pgfsetlinewidth{2.007500pt}%
\definecolor{currentstroke}{rgb}{0.501961,0.501961,0.501961}%
\pgfsetstrokecolor{currentstroke}%
\pgfsetstrokeopacity{0.300000}%
\pgfsetdash{{7.400000pt}{3.200000pt}}{0.000000pt}%
\pgfpathmoveto{\pgfqpoint{0.674954in}{4.139116in}}%
\pgfpathlineto{\pgfqpoint{7.147414in}{4.139116in}}%
\pgfusepath{stroke}%
\end{pgfscope}%
\begin{pgfscope}%
\pgfsetbuttcap%
\pgfsetroundjoin%
\definecolor{currentfill}{rgb}{0.000000,0.000000,0.000000}%
\pgfsetfillcolor{currentfill}%
\pgfsetlinewidth{0.803000pt}%
\definecolor{currentstroke}{rgb}{0.000000,0.000000,0.000000}%
\pgfsetstrokecolor{currentstroke}%
\pgfsetdash{}{0pt}%
\pgfsys@defobject{currentmarker}{\pgfqpoint{-0.048611in}{0.000000in}}{\pgfqpoint{-0.000000in}{0.000000in}}{%
\pgfpathmoveto{\pgfqpoint{-0.000000in}{0.000000in}}%
\pgfpathlineto{\pgfqpoint{-0.048611in}{0.000000in}}%
\pgfusepath{stroke,fill}%
}%
\begin{pgfscope}%
\pgfsys@transformshift{0.674954in}{4.139116in}%
\pgfsys@useobject{currentmarker}{}%
\end{pgfscope}%
\end{pgfscope}%
\begin{pgfscope}%
\definecolor{textcolor}{rgb}{0.000000,0.000000,0.000000}%
\pgfsetstrokecolor{textcolor}%
\pgfsetfillcolor{textcolor}%
\pgftext[x=0.259244in, y=4.020501in, left, base]{\color{textcolor}{\rmfamily\fontsize{25.000000}{30.000000}\selectfont\catcode`\^=\active\def^{\ifmmode\sp\else\^{}\fi}\catcode`\%=\active\def
\end{pgfscope}%
\begin{pgfscope}%
\pgfpathrectangle{\pgfqpoint{0.674954in}{0.862305in}}{\pgfqpoint{6.472460in}{4.369081in}}%
\pgfusepath{clip}%
\pgfsetbuttcap%
\pgfsetroundjoin%
\pgfsetlinewidth{2.007500pt}%
\definecolor{currentstroke}{rgb}{0.501961,0.501961,0.501961}%
\pgfsetstrokecolor{currentstroke}%
\pgfsetstrokeopacity{0.300000}%
\pgfsetdash{{7.400000pt}{3.200000pt}}{0.000000pt}%
\pgfpathmoveto{\pgfqpoint{0.674954in}{5.231386in}}%
\pgfpathlineto{\pgfqpoint{7.147414in}{5.231386in}}%
\pgfusepath{stroke}%
\end{pgfscope}%
\begin{pgfscope}%
\pgfsetbuttcap%
\pgfsetroundjoin%
\definecolor{currentfill}{rgb}{0.000000,0.000000,0.000000}%
\pgfsetfillcolor{currentfill}%
\pgfsetlinewidth{0.803000pt}%
\definecolor{currentstroke}{rgb}{0.000000,0.000000,0.000000}%
\pgfsetstrokecolor{currentstroke}%
\pgfsetdash{}{0pt}%
\pgfsys@defobject{currentmarker}{\pgfqpoint{-0.048611in}{0.000000in}}{\pgfqpoint{-0.000000in}{0.000000in}}{%
\pgfpathmoveto{\pgfqpoint{-0.000000in}{0.000000in}}%
\pgfpathlineto{\pgfqpoint{-0.048611in}{0.000000in}}%
\pgfusepath{stroke,fill}%
}%
\begin{pgfscope}%
\pgfsys@transformshift{0.674954in}{5.231386in}%
\pgfsys@useobject{currentmarker}{}%
\end{pgfscope}%
\end{pgfscope}%
\begin{pgfscope}%
\definecolor{textcolor}{rgb}{0.000000,0.000000,0.000000}%
\pgfsetstrokecolor{textcolor}%
\pgfsetfillcolor{textcolor}%
\pgftext[x=0.100000in, y=5.112772in, left, base]{\color{textcolor}{\rmfamily\fontsize{25.000000}{30.000000}\selectfont\catcode`\^=\active\def^{\ifmmode\sp\else\^{}\fi}\catcode`\%=\active\def
\end{pgfscope}%
\begin{pgfscope}%
\pgfpathrectangle{\pgfqpoint{0.674954in}{0.862305in}}{\pgfqpoint{6.472460in}{4.369081in}}%
\pgfusepath{clip}%
\pgfsetrectcap%
\pgfsetroundjoin%
\pgfsetlinewidth{2.509375pt}%
\definecolor{currentstroke}{rgb}{0.050980,0.415686,0.509804}%
\pgfsetstrokecolor{currentstroke}%
\pgfsetdash{}{0pt}%
\pgfpathmoveto{\pgfqpoint{0.674954in}{5.148860in}}%
\pgfpathlineto{\pgfqpoint{0.998577in}{5.146125in}}%
\pgfpathlineto{\pgfqpoint{2.293069in}{5.131088in}}%
\pgfpathlineto{\pgfqpoint{3.911184in}{5.114736in}}%
\pgfpathlineto{\pgfqpoint{5.529299in}{4.912147in}}%
\pgfpathlineto{\pgfqpoint{6.823791in}{4.491647in}}%
\pgfusepath{stroke}%
\end{pgfscope}%
\begin{pgfscope}%
\pgfpathrectangle{\pgfqpoint{0.674954in}{0.862305in}}{\pgfqpoint{6.472460in}{4.369081in}}%
\pgfusepath{clip}%
\pgfsetbuttcap%
\pgfsetroundjoin%
\definecolor{currentfill}{rgb}{0.050980,0.415686,0.509804}%
\pgfsetfillcolor{currentfill}%
\pgfsetlinewidth{1.003750pt}%
\definecolor{currentstroke}{rgb}{0.050980,0.415686,0.509804}%
\pgfsetstrokecolor{currentstroke}%
\pgfsetdash{}{0pt}%
\pgfsys@defobject{currentmarker}{\pgfqpoint{-0.055556in}{-0.055556in}}{\pgfqpoint{0.055556in}{0.055556in}}{%
\pgfpathmoveto{\pgfqpoint{0.000000in}{-0.055556in}}%
\pgfpathcurveto{\pgfqpoint{0.014734in}{-0.055556in}}{\pgfqpoint{0.028866in}{-0.049702in}}{\pgfqpoint{0.039284in}{-0.039284in}}%
\pgfpathcurveto{\pgfqpoint{0.049702in}{-0.028866in}}{\pgfqpoint{0.055556in}{-0.014734in}}{\pgfqpoint{0.055556in}{0.000000in}}%
\pgfpathcurveto{\pgfqpoint{0.055556in}{0.014734in}}{\pgfqpoint{0.049702in}{0.028866in}}{\pgfqpoint{0.039284in}{0.039284in}}%
\pgfpathcurveto{\pgfqpoint{0.028866in}{0.049702in}}{\pgfqpoint{0.014734in}{0.055556in}}{\pgfqpoint{0.000000in}{0.055556in}}%
\pgfpathcurveto{\pgfqpoint{-0.014734in}{0.055556in}}{\pgfqpoint{-0.028866in}{0.049702in}}{\pgfqpoint{-0.039284in}{0.039284in}}%
\pgfpathcurveto{\pgfqpoint{-0.049702in}{0.028866in}}{\pgfqpoint{-0.055556in}{0.014734in}}{\pgfqpoint{-0.055556in}{0.000000in}}%
\pgfpathcurveto{\pgfqpoint{-0.055556in}{-0.014734in}}{\pgfqpoint{-0.049702in}{-0.028866in}}{\pgfqpoint{-0.039284in}{-0.039284in}}%
\pgfpathcurveto{\pgfqpoint{-0.028866in}{-0.049702in}}{\pgfqpoint{-0.014734in}{-0.055556in}}{\pgfqpoint{0.000000in}{-0.055556in}}%
\pgfpathlineto{\pgfqpoint{0.000000in}{-0.055556in}}%
\pgfpathclose%
\pgfusepath{stroke,fill}%
}%
\begin{pgfscope}%
\pgfsys@transformshift{0.674954in}{5.148860in}%
\pgfsys@useobject{currentmarker}{}%
\end{pgfscope}%
\begin{pgfscope}%
\pgfsys@transformshift{0.998577in}{5.146125in}%
\pgfsys@useobject{currentmarker}{}%
\end{pgfscope}%
\begin{pgfscope}%
\pgfsys@transformshift{2.293069in}{5.131088in}%
\pgfsys@useobject{currentmarker}{}%
\end{pgfscope}%
\begin{pgfscope}%
\pgfsys@transformshift{3.911184in}{5.114736in}%
\pgfsys@useobject{currentmarker}{}%
\end{pgfscope}%
\begin{pgfscope}%
\pgfsys@transformshift{5.529299in}{4.912147in}%
\pgfsys@useobject{currentmarker}{}%
\end{pgfscope}%
\begin{pgfscope}%
\pgfsys@transformshift{6.823791in}{4.491647in}%
\pgfsys@useobject{currentmarker}{}%
\end{pgfscope}%
\end{pgfscope}%
\begin{pgfscope}%
\pgfpathrectangle{\pgfqpoint{0.674954in}{0.862305in}}{\pgfqpoint{6.472460in}{4.369081in}}%
\pgfusepath{clip}%
\pgfsetbuttcap%
\pgfsetroundjoin%
\pgfsetlinewidth{2.509375pt}%
\definecolor{currentstroke}{rgb}{0.960784,0.462745,0.000000}%
\pgfsetstrokecolor{currentstroke}%
\pgfsetdash{{9.250000pt}{4.000000pt}}{0.000000pt}%
\pgfpathmoveto{\pgfqpoint{0.998577in}{4.607231in}}%
\pgfpathlineto{\pgfqpoint{2.293069in}{4.769532in}}%
\pgfpathlineto{\pgfqpoint{3.911184in}{4.706420in}}%
\pgfpathlineto{\pgfqpoint{5.529299in}{4.665397in}}%
\pgfpathlineto{\pgfqpoint{6.823791in}{4.768369in}}%
\pgfpathlineto{\pgfqpoint{7.147414in}{4.828480in}}%
\pgfusepath{stroke}%
\end{pgfscope}%
\begin{pgfscope}%
\pgfpathrectangle{\pgfqpoint{0.674954in}{0.862305in}}{\pgfqpoint{6.472460in}{4.369081in}}%
\pgfusepath{clip}%
\pgfsetbuttcap%
\pgfsetmiterjoin%
\definecolor{currentfill}{rgb}{0.960784,0.462745,0.000000}%
\pgfsetfillcolor{currentfill}%
\pgfsetlinewidth{1.003750pt}%
\definecolor{currentstroke}{rgb}{0.960784,0.462745,0.000000}%
\pgfsetstrokecolor{currentstroke}%
\pgfsetdash{}{0pt}%
\pgfsys@defobject{currentmarker}{\pgfqpoint{-0.055556in}{-0.055556in}}{\pgfqpoint{0.055556in}{0.055556in}}{%
\pgfpathmoveto{\pgfqpoint{-0.055556in}{-0.055556in}}%
\pgfpathlineto{\pgfqpoint{0.055556in}{-0.055556in}}%
\pgfpathlineto{\pgfqpoint{0.055556in}{0.055556in}}%
\pgfpathlineto{\pgfqpoint{-0.055556in}{0.055556in}}%
\pgfpathlineto{\pgfqpoint{-0.055556in}{-0.055556in}}%
\pgfpathclose%
\pgfusepath{stroke,fill}%
}%
\begin{pgfscope}%
\pgfsys@transformshift{0.998577in}{4.607231in}%
\pgfsys@useobject{currentmarker}{}%
\end{pgfscope}%
\begin{pgfscope}%
\pgfsys@transformshift{2.293069in}{4.769532in}%
\pgfsys@useobject{currentmarker}{}%
\end{pgfscope}%
\begin{pgfscope}%
\pgfsys@transformshift{3.911184in}{4.706420in}%
\pgfsys@useobject{currentmarker}{}%
\end{pgfscope}%
\begin{pgfscope}%
\pgfsys@transformshift{5.529299in}{4.665397in}%
\pgfsys@useobject{currentmarker}{}%
\end{pgfscope}%
\begin{pgfscope}%
\pgfsys@transformshift{6.823791in}{4.768369in}%
\pgfsys@useobject{currentmarker}{}%
\end{pgfscope}%
\begin{pgfscope}%
\pgfsys@transformshift{7.147414in}{4.828480in}%
\pgfsys@useobject{currentmarker}{}%
\end{pgfscope}%
\end{pgfscope}%
\begin{pgfscope}%
\pgfsetrectcap%
\pgfsetmiterjoin%
\pgfsetlinewidth{2.007500pt}%
\definecolor{currentstroke}{rgb}{0.000000,0.000000,0.000000}%
\pgfsetstrokecolor{currentstroke}%
\pgfsetdash{}{0pt}%
\pgfpathmoveto{\pgfqpoint{0.674954in}{0.862305in}}%
\pgfpathlineto{\pgfqpoint{0.674954in}{5.231386in}}%
\pgfusepath{stroke}%
\end{pgfscope}%
\begin{pgfscope}%
\pgfsetrectcap%
\pgfsetmiterjoin%
\pgfsetlinewidth{2.007500pt}%
\definecolor{currentstroke}{rgb}{0.000000,0.000000,0.000000}%
\pgfsetstrokecolor{currentstroke}%
\pgfsetdash{}{0pt}%
\pgfpathmoveto{\pgfqpoint{0.674954in}{0.862305in}}%
\pgfpathlineto{\pgfqpoint{7.147414in}{0.862305in}}%
\pgfusepath{stroke}%
\end{pgfscope}%
\begin{pgfscope}%
\pgfsetbuttcap%
\pgfsetmiterjoin%
\definecolor{currentfill}{rgb}{1.000000,1.000000,1.000000}%
\pgfsetfillcolor{currentfill}%
\pgfsetlinewidth{1.003750pt}%
\definecolor{currentstroke}{rgb}{0.800000,0.800000,0.800000}%
\pgfsetstrokecolor{currentstroke}%
\pgfsetdash{}{0pt}%
\pgfpathmoveto{\pgfqpoint{0.869398in}{1.001194in}}%
\pgfpathlineto{\pgfqpoint{2.708109in}{1.001194in}}%
\pgfpathquadraticcurveto{\pgfqpoint{2.763664in}{1.001194in}}{\pgfqpoint{2.763664in}{1.056749in}}%
\pgfpathlineto{\pgfqpoint{2.763664in}{1.803694in}}%
\pgfpathquadraticcurveto{\pgfqpoint{2.763664in}{1.859250in}}{\pgfqpoint{2.708109in}{1.859250in}}%
\pgfpathlineto{\pgfqpoint{0.869398in}{1.859250in}}%
\pgfpathquadraticcurveto{\pgfqpoint{0.813843in}{1.859250in}}{\pgfqpoint{0.813843in}{1.803694in}}%
\pgfpathlineto{\pgfqpoint{0.813843in}{1.056749in}}%
\pgfpathquadraticcurveto{\pgfqpoint{0.813843in}{1.001194in}}{\pgfqpoint{0.869398in}{1.001194in}}%
\pgfpathlineto{\pgfqpoint{0.869398in}{1.001194in}}%
\pgfpathclose%
\pgfusepath{stroke,fill}%
\end{pgfscope}%
\begin{pgfscope}%
\pgfsetrectcap%
\pgfsetroundjoin%
\pgfsetlinewidth{2.509375pt}%
\definecolor{currentstroke}{rgb}{0.050980,0.415686,0.509804}%
\pgfsetstrokecolor{currentstroke}%
\pgfsetdash{}{0pt}%
\pgfpathmoveto{\pgfqpoint{0.924954in}{1.650917in}}%
\pgfpathlineto{\pgfqpoint{1.202732in}{1.650917in}}%
\pgfpathlineto{\pgfqpoint{1.480509in}{1.650917in}}%
\pgfusepath{stroke}%
\end{pgfscope}%
\begin{pgfscope}%
\pgfsetbuttcap%
\pgfsetroundjoin%
\definecolor{currentfill}{rgb}{0.050980,0.415686,0.509804}%
\pgfsetfillcolor{currentfill}%
\pgfsetlinewidth{1.003750pt}%
\definecolor{currentstroke}{rgb}{0.050980,0.415686,0.509804}%
\pgfsetstrokecolor{currentstroke}%
\pgfsetdash{}{0pt}%
\pgfsys@defobject{currentmarker}{\pgfqpoint{-0.055556in}{-0.055556in}}{\pgfqpoint{0.055556in}{0.055556in}}{%
\pgfpathmoveto{\pgfqpoint{0.000000in}{-0.055556in}}%
\pgfpathcurveto{\pgfqpoint{0.014734in}{-0.055556in}}{\pgfqpoint{0.028866in}{-0.049702in}}{\pgfqpoint{0.039284in}{-0.039284in}}%
\pgfpathcurveto{\pgfqpoint{0.049702in}{-0.028866in}}{\pgfqpoint{0.055556in}{-0.014734in}}{\pgfqpoint{0.055556in}{0.000000in}}%
\pgfpathcurveto{\pgfqpoint{0.055556in}{0.014734in}}{\pgfqpoint{0.049702in}{0.028866in}}{\pgfqpoint{0.039284in}{0.039284in}}%
\pgfpathcurveto{\pgfqpoint{0.028866in}{0.049702in}}{\pgfqpoint{0.014734in}{0.055556in}}{\pgfqpoint{0.000000in}{0.055556in}}%
\pgfpathcurveto{\pgfqpoint{-0.014734in}{0.055556in}}{\pgfqpoint{-0.028866in}{0.049702in}}{\pgfqpoint{-0.039284in}{0.039284in}}%
\pgfpathcurveto{\pgfqpoint{-0.049702in}{0.028866in}}{\pgfqpoint{-0.055556in}{0.014734in}}{\pgfqpoint{-0.055556in}{0.000000in}}%
\pgfpathcurveto{\pgfqpoint{-0.055556in}{-0.014734in}}{\pgfqpoint{-0.049702in}{-0.028866in}}{\pgfqpoint{-0.039284in}{-0.039284in}}%
\pgfpathcurveto{\pgfqpoint{-0.028866in}{-0.049702in}}{\pgfqpoint{-0.014734in}{-0.055556in}}{\pgfqpoint{0.000000in}{-0.055556in}}%
\pgfpathlineto{\pgfqpoint{0.000000in}{-0.055556in}}%
\pgfpathclose%
\pgfusepath{stroke,fill}%
}%
\begin{pgfscope}%
\pgfsys@transformshift{1.202732in}{1.650917in}%
\pgfsys@useobject{currentmarker}{}%
\end{pgfscope}%
\end{pgfscope}%
\begin{pgfscope}%
\definecolor{textcolor}{rgb}{0.000000,0.000000,0.000000}%
\pgfsetstrokecolor{textcolor}%
\pgfsetfillcolor{textcolor}%
\pgftext[x=1.702732in,y=1.553694in,left,base]{\color{textcolor}{\rmfamily\fontsize{20.000000}{24.000000}\selectfont\catcode`\^=\active\def^{\ifmmode\sp\else\^{}\fi}\catcode`\%=\active\def
\end{pgfscope}%
\begin{pgfscope}%
\pgfsetbuttcap%
\pgfsetroundjoin%
\pgfsetlinewidth{2.509375pt}%
\definecolor{currentstroke}{rgb}{0.960784,0.462745,0.000000}%
\pgfsetstrokecolor{currentstroke}%
\pgfsetdash{{9.250000pt}{4.000000pt}}{0.000000pt}%
\pgfpathmoveto{\pgfqpoint{0.924954in}{1.263555in}}%
\pgfpathlineto{\pgfqpoint{1.202732in}{1.263555in}}%
\pgfpathlineto{\pgfqpoint{1.480509in}{1.263555in}}%
\pgfusepath{stroke}%
\end{pgfscope}%
\begin{pgfscope}%
\pgfsetbuttcap%
\pgfsetmiterjoin%
\definecolor{currentfill}{rgb}{0.960784,0.462745,0.000000}%
\pgfsetfillcolor{currentfill}%
\pgfsetlinewidth{1.003750pt}%
\definecolor{currentstroke}{rgb}{0.960784,0.462745,0.000000}%
\pgfsetstrokecolor{currentstroke}%
\pgfsetdash{}{0pt}%
\pgfsys@defobject{currentmarker}{\pgfqpoint{-0.055556in}{-0.055556in}}{\pgfqpoint{0.055556in}{0.055556in}}{%
\pgfpathmoveto{\pgfqpoint{-0.055556in}{-0.055556in}}%
\pgfpathlineto{\pgfqpoint{0.055556in}{-0.055556in}}%
\pgfpathlineto{\pgfqpoint{0.055556in}{0.055556in}}%
\pgfpathlineto{\pgfqpoint{-0.055556in}{0.055556in}}%
\pgfpathlineto{\pgfqpoint{-0.055556in}{-0.055556in}}%
\pgfpathclose%
\pgfusepath{stroke,fill}%
}%
\begin{pgfscope}%
\pgfsys@transformshift{1.202732in}{1.263555in}%
\pgfsys@useobject{currentmarker}{}%
\end{pgfscope}%
\end{pgfscope}%
\begin{pgfscope}%
\definecolor{textcolor}{rgb}{0.000000,0.000000,0.000000}%
\pgfsetstrokecolor{textcolor}%
\pgfsetfillcolor{textcolor}%
\pgftext[x=1.702732in,y=1.166333in,left,base]{\color{textcolor}{\rmfamily\fontsize{20.000000}{24.000000}\selectfont\catcode`\^=\active\def^{\ifmmode\sp\else\^{}\fi}\catcode`\%=\active\def
\end{pgfscope}%
\end{pgfpicture}%
\makeatother%
\endgroup%

%% file: supplementary/15_results_full_average.tex
\section{Aggregate Over All Datasets}

In this section, we report the metrics obtained by aggregating the errors across all scenes, in contrast to the evaluation protocol used in in the main paper, where the results are reported separately for the Public Datasets and \datasetname.
As shown in~\cref{tab:all_aggregated}, our method achieves the highest overall performance in this combined setting. MP-SfM is not presented in the table because it is not validated on \datasetname.

\input{tables/metrics_aggregated_all}

%% file: tables/metrics_aggregated_all.tex
\begin{table}[t]
  \centering
  \caption{
    Quantitative comparison of 3D reconstruction methods for all datasets. The results are aggregated across Public Datasets and \datasetname. Our method yields better scores than the baselines.
  }
  \scalebox{0.9}{
    \begin{tabular}{lcccccccc}
      \toprule Method & AUC@30 $\uparrow$ & RRA@30 $\uparrow$ & RTA@30 $\uparrow$ & PCA $\downarrow$ & PCC $\downarrow$ & Chamfer $\downarrow$ & Reg., \% $\uparrow$ & FPS $\uparrow$ \\ \midrule
      \makecell[l]{COLMAP +\\SPSG} & \cellcolor{secondbest}{60.2} & \cellcolor{secondbest}{85.3} & \cellcolor{secondbest}{78.0} & 1.64 & 1.20 & 1.42 & 40.5 & 0.5 \\ \midrule
      $\text{MA}_{\text{ELoFTR}}$ & 14.7 & 72.1 & 47.8 & \cellcolor{secondbest}{0.49} & 5.16 & 2.82 & 25.7 & 0.2 \\
      $\text{MINIMA}_{\text{ROMA}}$ & 42.6 & 67.7 & 64.3 & 0.97 & 1.09 & 1.03 & \cellcolor{secondbest}{99.0} & 0.0 \\ \midrule
      DUSt3R & 18.9 & 47.0 & 41.4 & 0.72 & 4.72 & 2.72 & \cellcolor{best}{100.0} & 0.7 \\
      MASt3R & 32.7 & 66.4 & 55.9 & 0.66 & \cellcolor{secondbest}{0.27} & \cellcolor{secondbest}{0.46} & \cellcolor{best}{100.0} & 0.2 \\ \midrule
      VGGT & 23.0 & 50.6 & 50.3 & 1.22 & 2.60 & 1.91 & \cellcolor{best}{100.0} & \cellcolor{best}{10.47} \\
      MapAnything & 21.8 & 51.4 & 48.4 & 0.69 & 3.75 & 2.22 & \cellcolor{best}{100.0} & 2.01 \\
      \ours & \cellcolor{best}{68.4} & \cellcolor{best}{89.1} & \cellcolor{best}{86.8} & \cellcolor{best}{0.47} & \cellcolor{best}{0.06} & \cellcolor{best}{0.27} & \cellcolor{best}{100.0} & \cellcolor{secondbest}{10.01} \\ \bottomrule
    \end{tabular}
  }
  \label{tab:all_aggregated}
\end{table}

%% file: supplementary/16_data_split.tex
\section{Dataset Splits}

We randomly split scenes into train and test sets, selecting at least 20\% of scenes from each dataset for testing. Table~\ref{tab:dataset_splits} reports the resulting split for each dataset separately. For every dataset, the Train column lists scenes used for training, while the Test column lists held-out scenes used only for evaluation.

\begin{table}[t]
  \centering
  \caption{Train/test splits for each dataset. The Train column contains scenes used for model training, while the Test column contains held-out scenes used for evaluation.}
  \label{tab:dataset_splits}
  \resizebox{\textwidth}{!}{%
    \begin{tabular}{l l l}
      \toprule
      Dataset & Train & Test \\
      \midrule

      ThermalGaussian &
      \makecell[l]{RotaryKiln, DarkScenes\\
        Truck, PlantEquipment\\
        DailyStuff, LandScape\\
        RoadBlock, Dimsum\\
        Building, GlassCup\\
      TransmissionTower} &
      \makecell[l]{Parterre, IronIngot\\
      Ebike} \\
      \midrule

      ThermalNeRF &
      \makecell[l]{engine, heater\\
        trace, sheet\\
        generators, charger\\
      pyrex} &
      \makecell[l]{sink, generator} \\
      \midrule

      ThermoScenes &
      \makecell[l]{trees, heater\_water\_cup\\
        reflect-laptop, double\_robot\\
        dorm1, raspberrypi\\
        buildingA\_spring, MED-building\\
        buildingA\_winter, reflect-cup\\
        exhibition\_building, dorm2\\
        reflect-kettle, building-sunrise\\
        BI-building, prpt-fridge\\
      heater\_water\_kettle} &
      \makecell[l]{reflect-robot, freezing\_ice\_cup\\
        prpt-cup, melting\_ice\_cup\\
      INR-building} \\
      \midrule

      ThermalMix &
      \makecell[l]{hand, lion\\
      face, pan} &
      \makecell[l]{panel, laptop} \\
      \midrule

      \orebrodataset &
      \makecell[l]{02\_Fox\_area\_no\_radars\_0\\
        02\_Fox\_area\_no\_radars\_1\\
        02\_Fox\_area\_no\_radars\_2\\
        02\_Fox\_area\_no\_radars\_3\\
        05\_Return\_from\_the\_forest\_no\_radar\_0\\
        05\_Return\_from\_the\_forest\_no\_radar\_1\\
        06\_Sunset\_area\_no\_radars\_0\\
        06\_Sunset\_area\_no\_radars\_1\\
      06\_Sunset\_area\_no\_radars\_2} &
      \makecell[l]{01\_Annexet\_No\_Radars\_0\\
        01\_Annexet\_No\_Radars\_1\\
        01\_Annexet\_No\_Radars\_2\\
        01\_Annexet\_No\_Radars\_3\\
        04\_Forest\_pass\_no\_radars\_0\\
      04\_Forest\_pass\_no\_radars\_1} \\
      \midrule

      \datasetname &
      -- &
      \makecell[l]{conference-room, metallic-container\\
        old-drinking-fountain, parking\\
      statue, telescope} \\

      \bottomrule
    \end{tabular}%
  }
\end{table}